\documentclass{article}

\usepackage{arxiv}

\usepackage[numbers]{natbib}
\usepackage[utf8]{inputenc} 
\usepackage[T1]{fontenc}    
\usepackage{hyperref}       
\usepackage{url}            
\usepackage{booktabs}       
\usepackage{amsfonts}       
\usepackage{nicefrac}       
\usepackage{microtype}      
\usepackage{xcolor}         

\usepackage{graphicx}
\usepackage{amssymb}
\usepackage{pgfplots}
\usepackage{subcaption} 
\usepackage{tabularx}   
\usepackage{array}      
\usepackage{caption}    
\usepackage{float}      
\usepackage{rotating}
\usepackage{svg}
\usepackage{soul}
\usepackage{chngcntr}   
\usepackage{multirow}
\usepackage[export]{adjustbox} 

\usepackage[accsupp]{axessibility}  
\usepackage[final]{changes}
\usepackage[capitalize]{cleveref}
\usepackage{enumitem}

\title{Unlocking Diffusion Hierarchies: Adaptive Timestep Selection for Zero-Shot Segmentation}

\author{%
  \textbf{Ramin Nakhli} $\cdot$ \textbf{Mahesh Ramachandran} $\cdot$ \textbf{Luca Ballan} \\[0.5em]
  Google
}

\begin{document}

\maketitle

\begin{figure*}[h]
    \centering
    
    \begin{minipage}[c]{0.05\textwidth}
        \rotatebox{90}{\textbf{Input}}
    \end{minipage}%
    \begin{minipage}[c]{0.105\textwidth}
        \includegraphics[width=\textwidth]{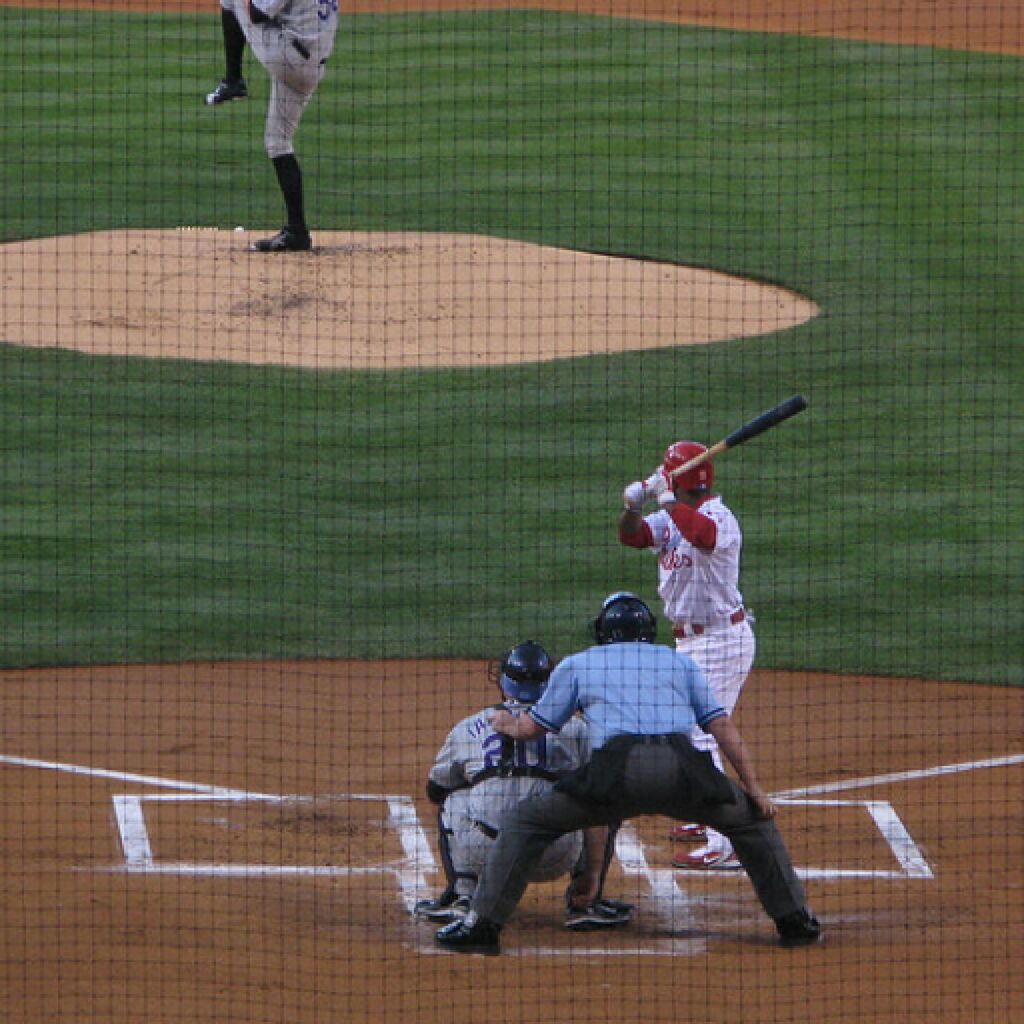}
    \end{minipage}%
    \begin{minipage}[c]{0.105\textwidth}
        \includegraphics[width=\textwidth]{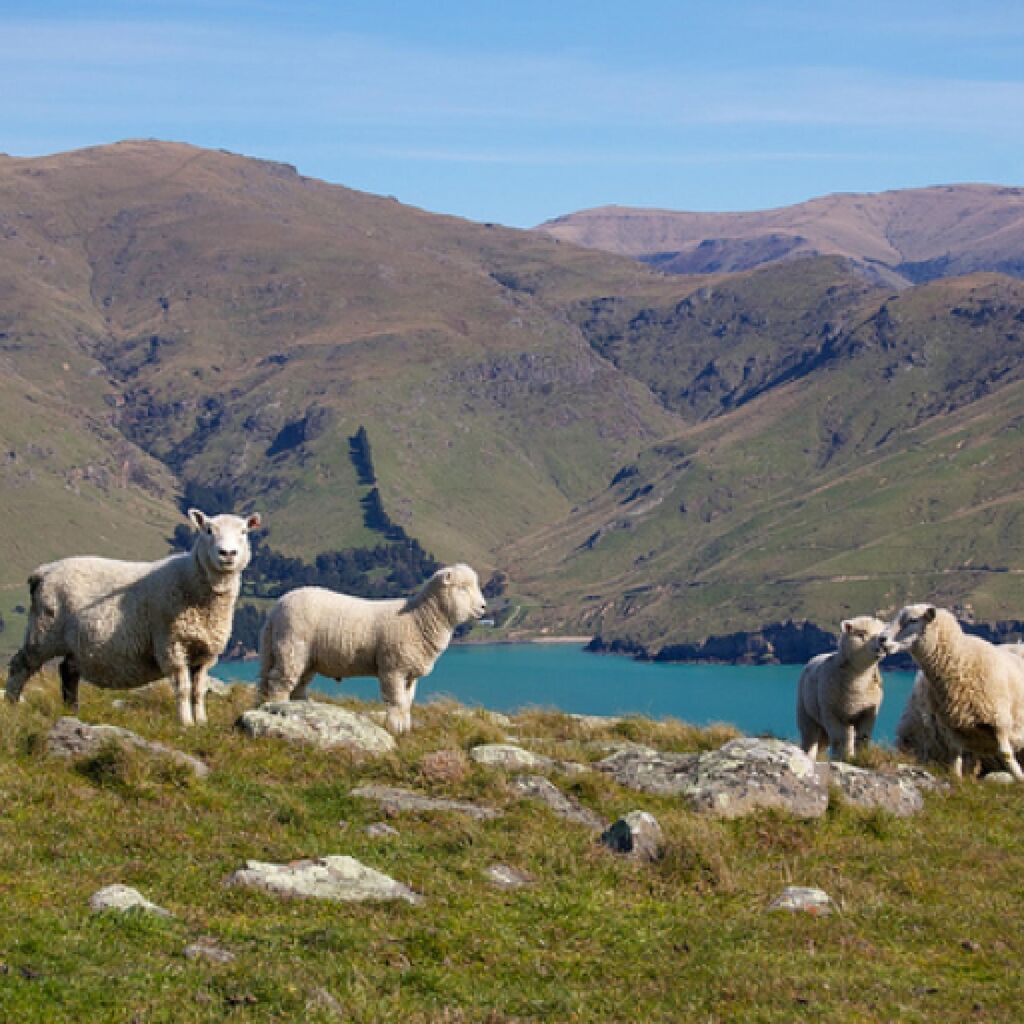}
    \end{minipage}%
    \begin{minipage}[c]{0.105\textwidth}
        \includegraphics[width=\textwidth]{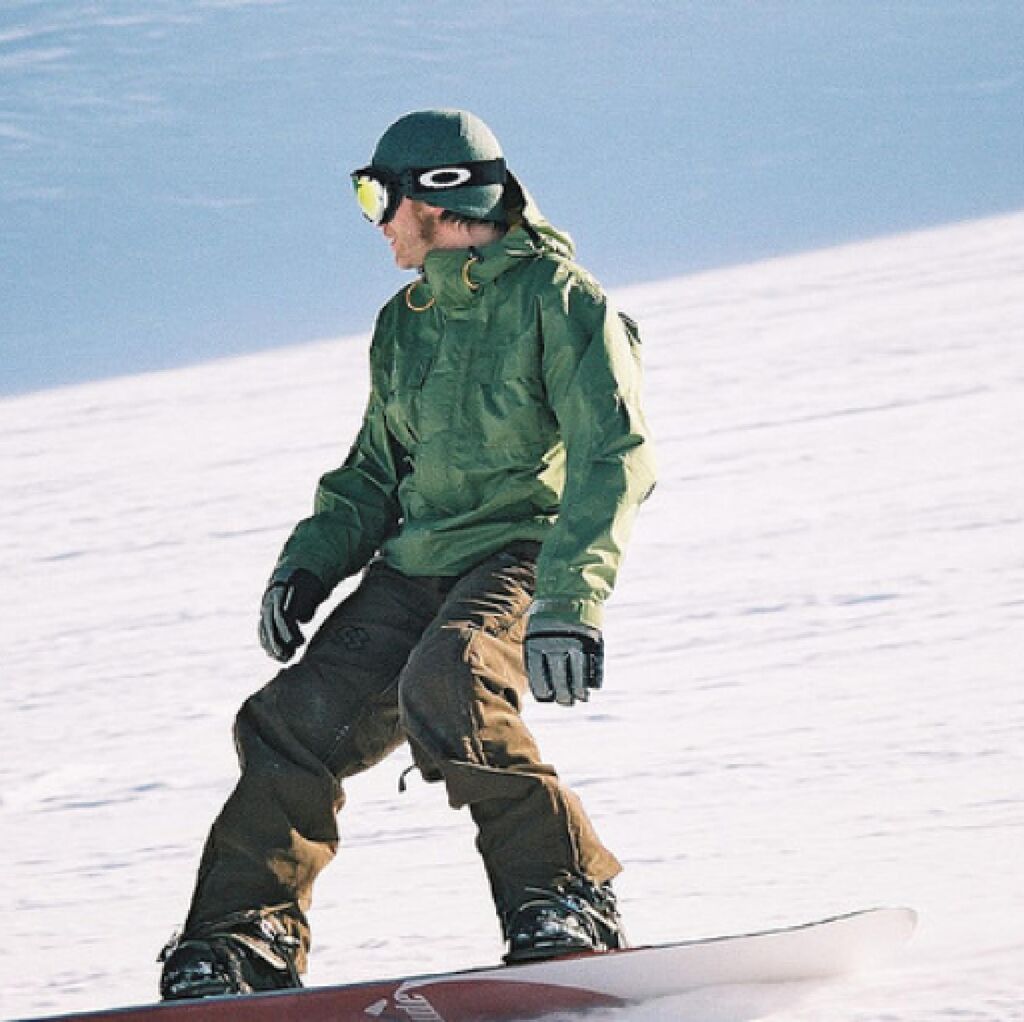}
    \end{minipage}%
    \begin{minipage}[c]{0.105\textwidth}
        \includegraphics[width=\textwidth]{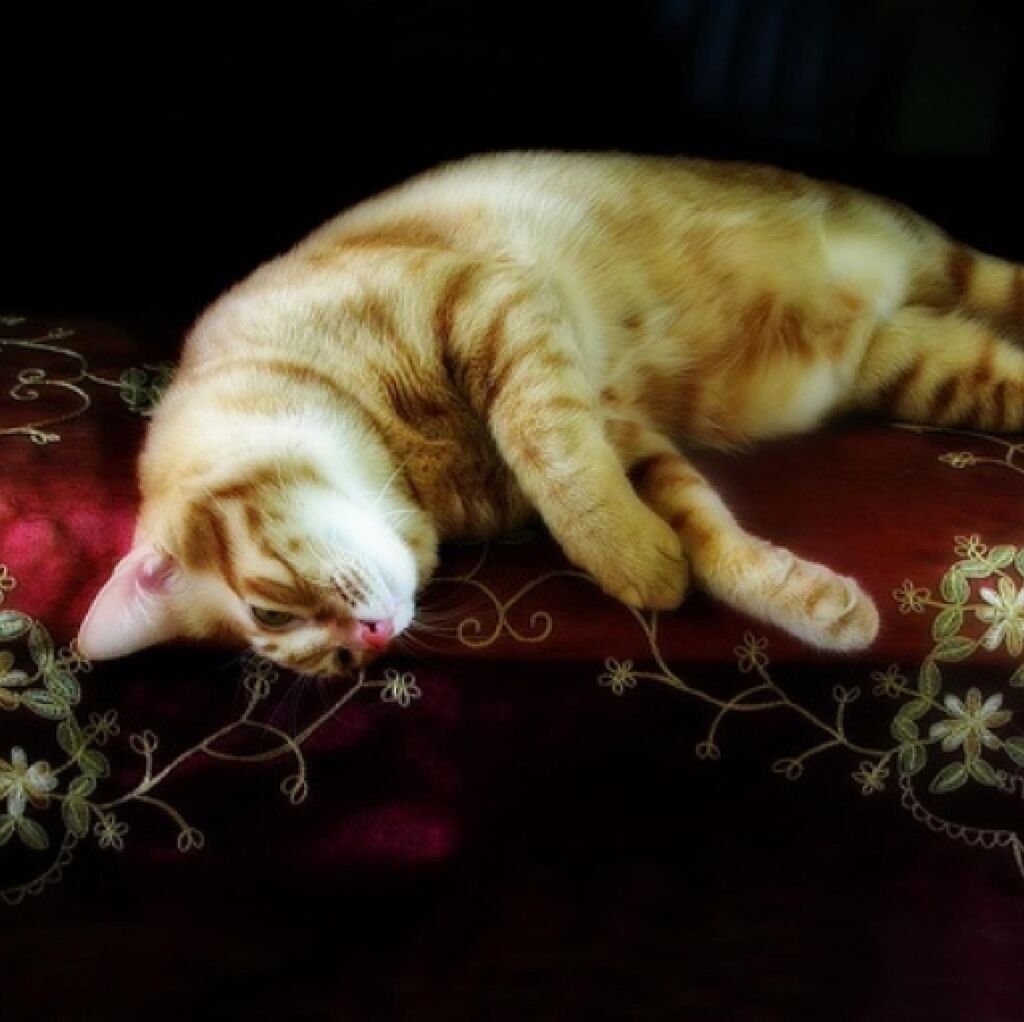}
    \end{minipage}%
    \begin{minipage}[c]{0.105\textwidth}
        \includegraphics[width=\textwidth]{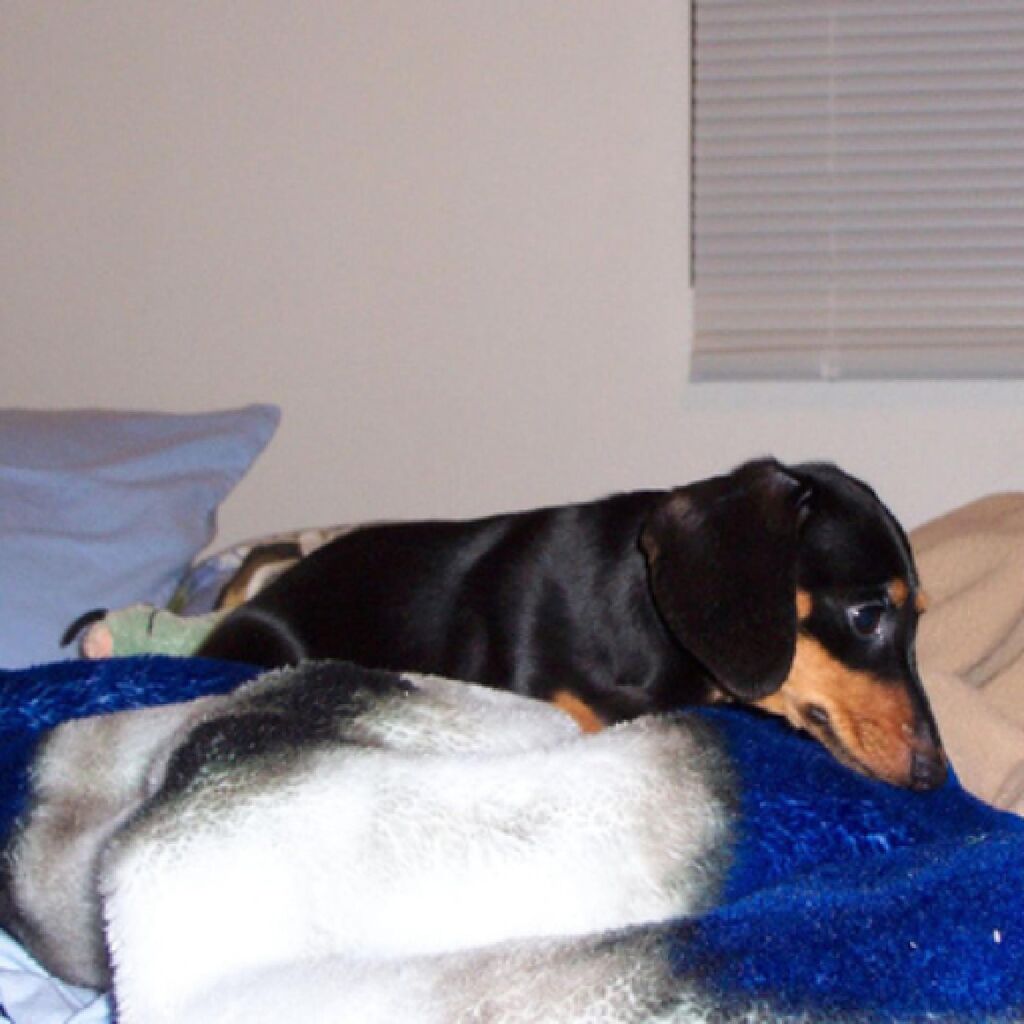}
    \end{minipage}%
    \begin{minipage}[c]{0.105\textwidth}
        \includegraphics[width=\textwidth]{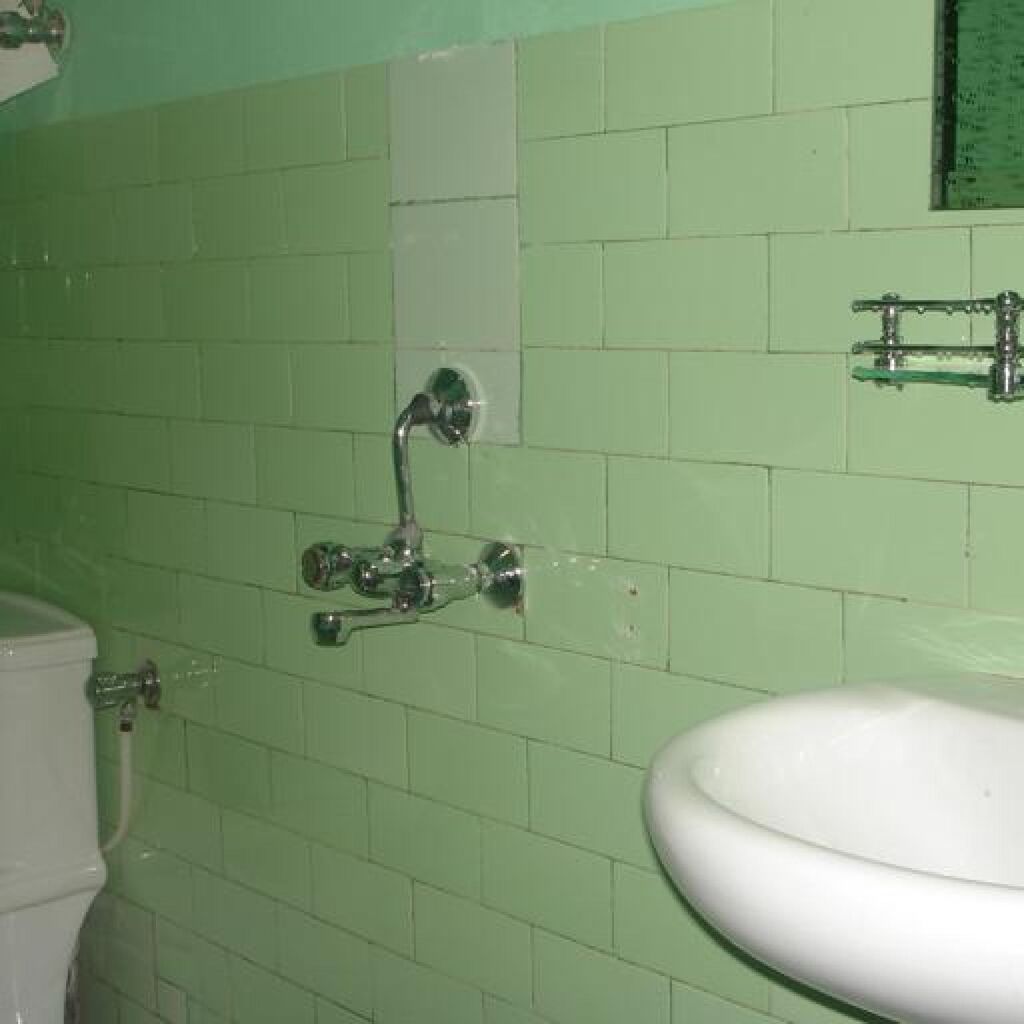}
    \end{minipage}%
    \begin{minipage}[c]{0.105\textwidth}
        \includegraphics[width=\textwidth]{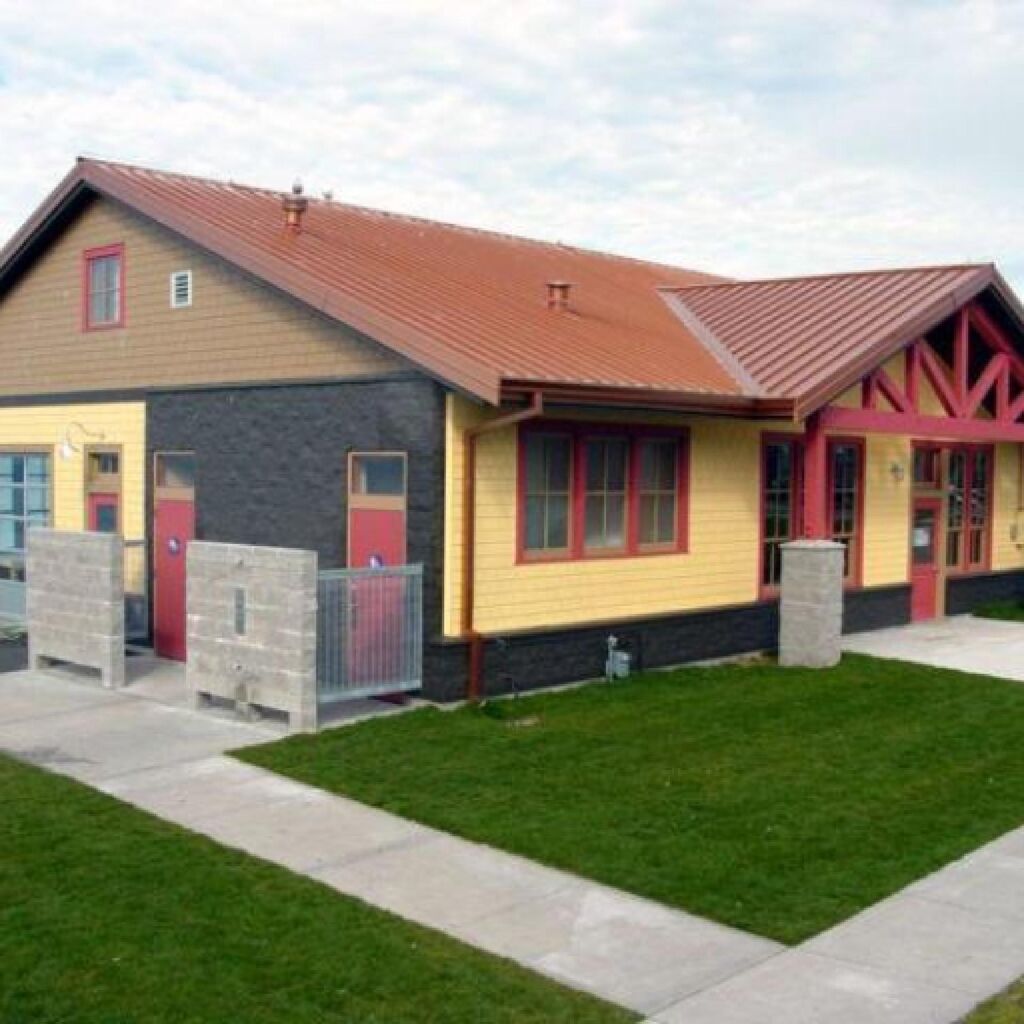}
    \end{minipage}%
    \begin{minipage}[c]{0.105\textwidth}
        \includegraphics[width=\textwidth]{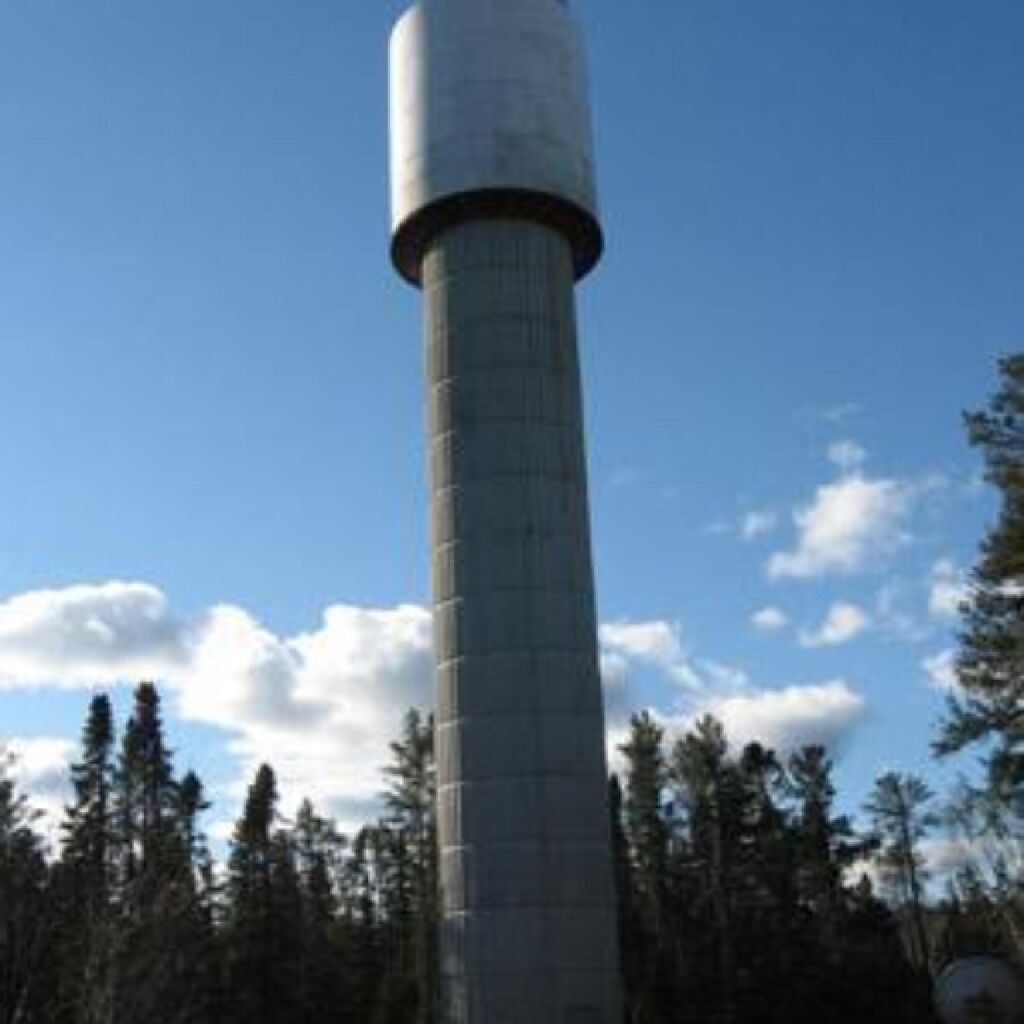}
    \end{minipage}%
    \begin{minipage}[c]{0.105\textwidth}
        \includegraphics[width=\textwidth]{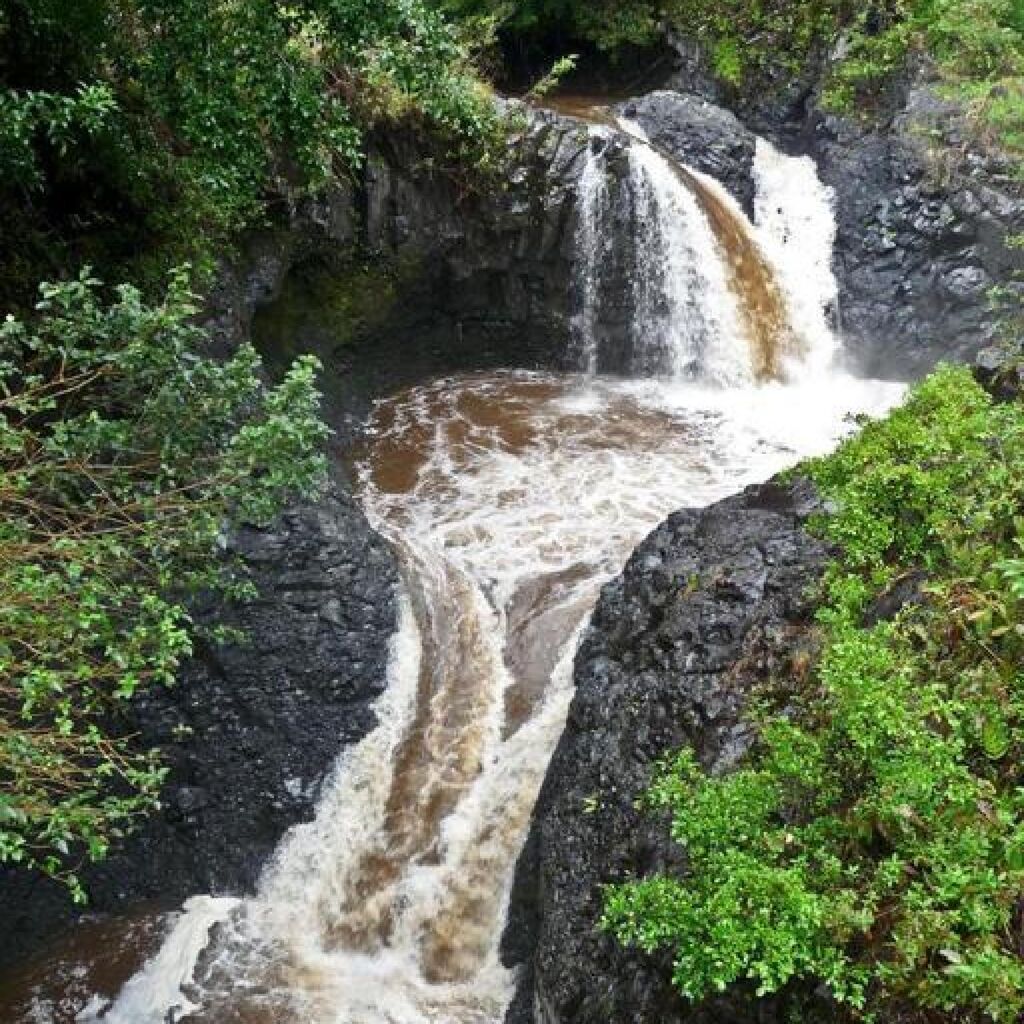}
    \end{minipage}%

    \vspace{-0.5em}

    \begin{minipage}[c]{0.05\textwidth}
        \rotatebox{90}{\textbf{DiffCut}}
    \end{minipage}%
    \begin{minipage}[c]{0.105\textwidth}
        \includegraphics[width=\textwidth]{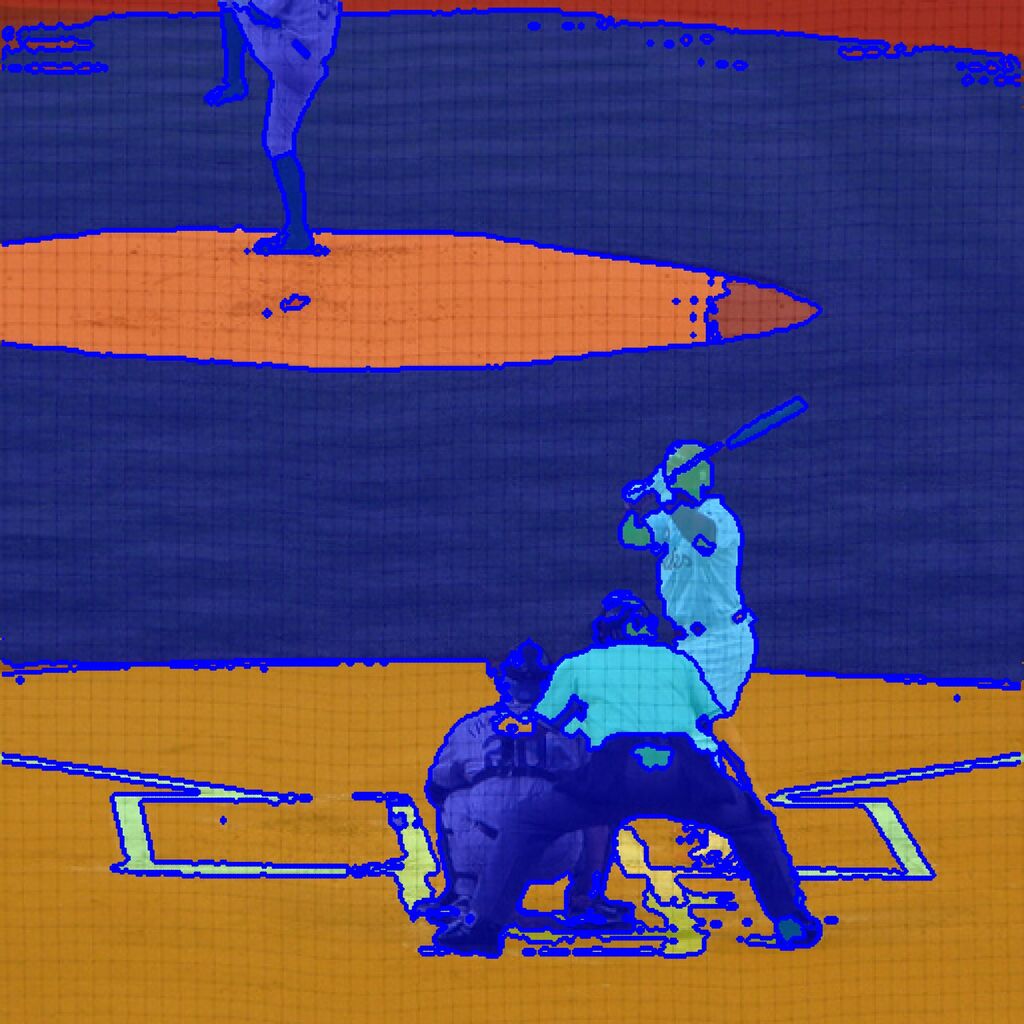}
    \end{minipage}%
    \begin{minipage}[c]{0.105\textwidth}
        \includegraphics[width=\textwidth]{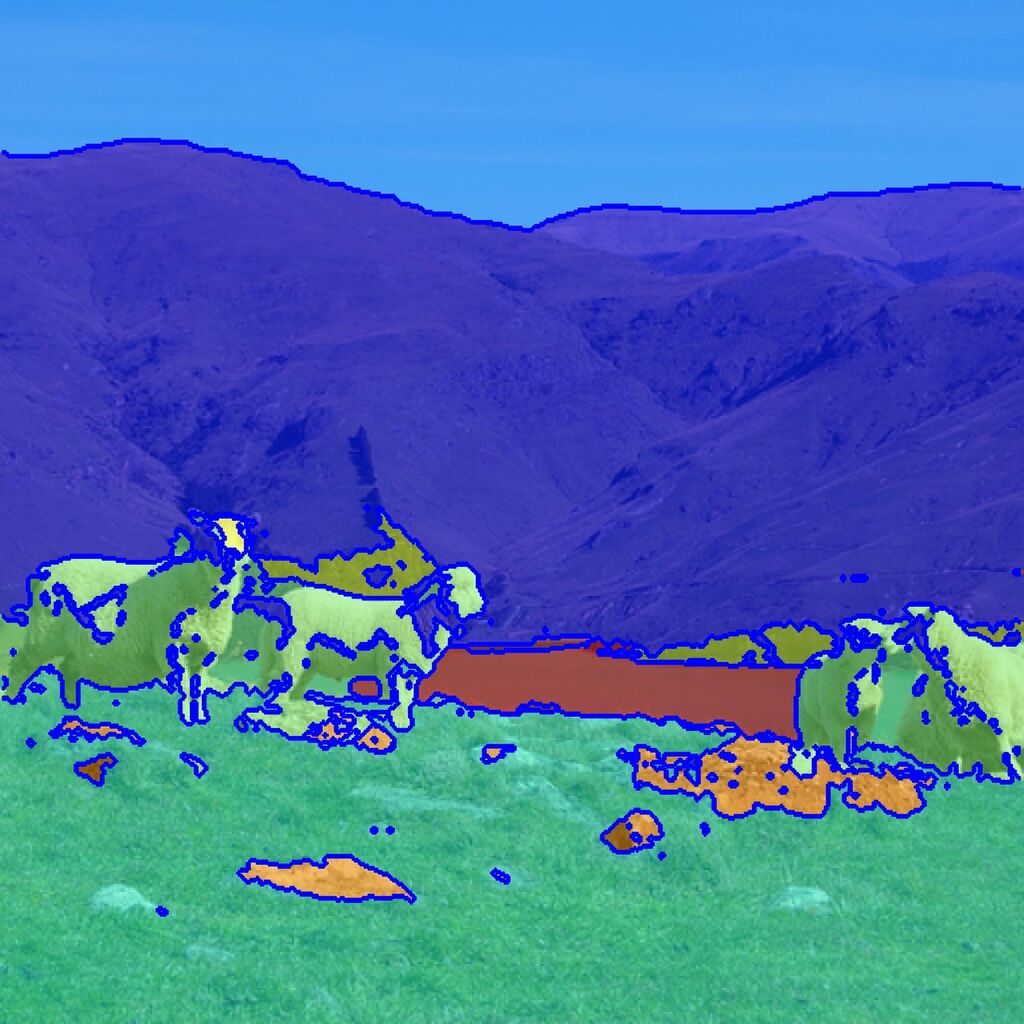}
    \end{minipage}%
    \begin{minipage}[c]{0.105\textwidth}
        \includegraphics[width=\textwidth]{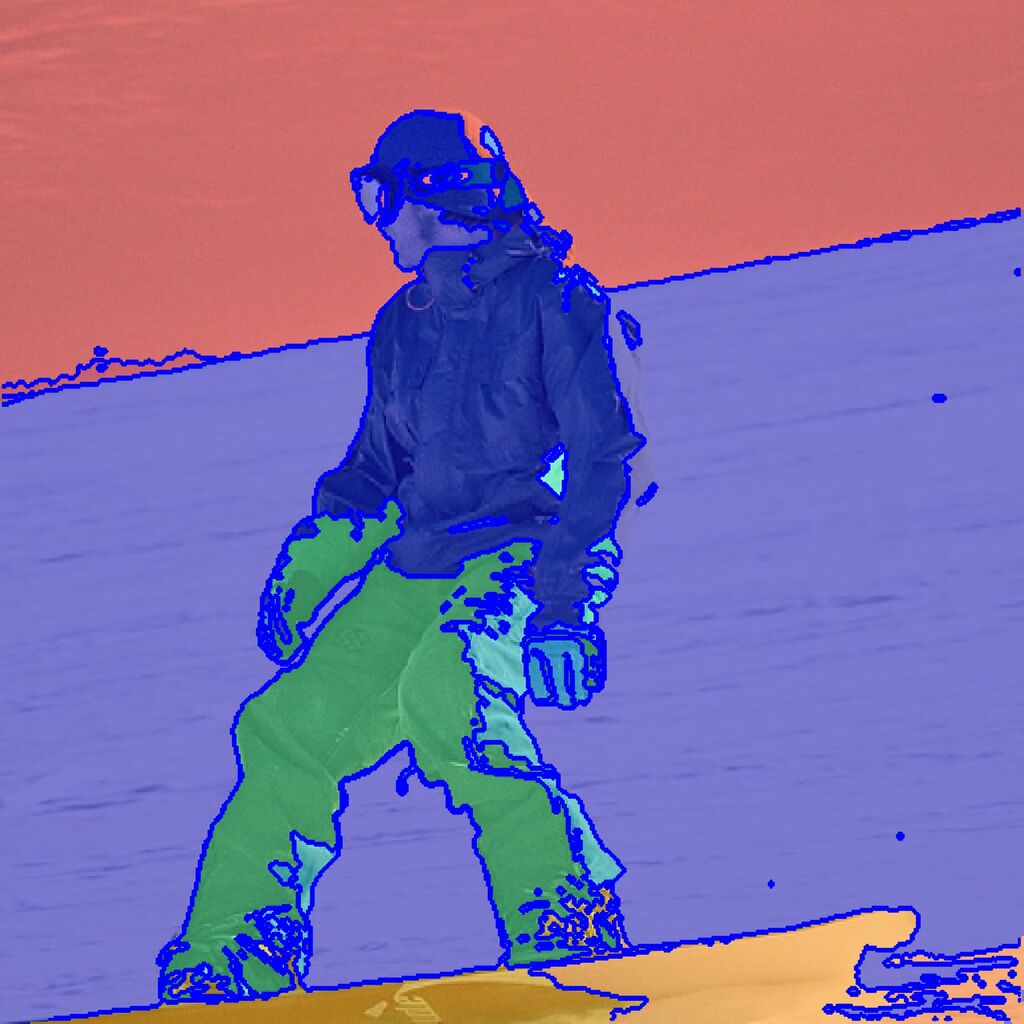}
    \end{minipage}%
    \begin{minipage}[c]{0.105\textwidth}
        \includegraphics[width=\textwidth]{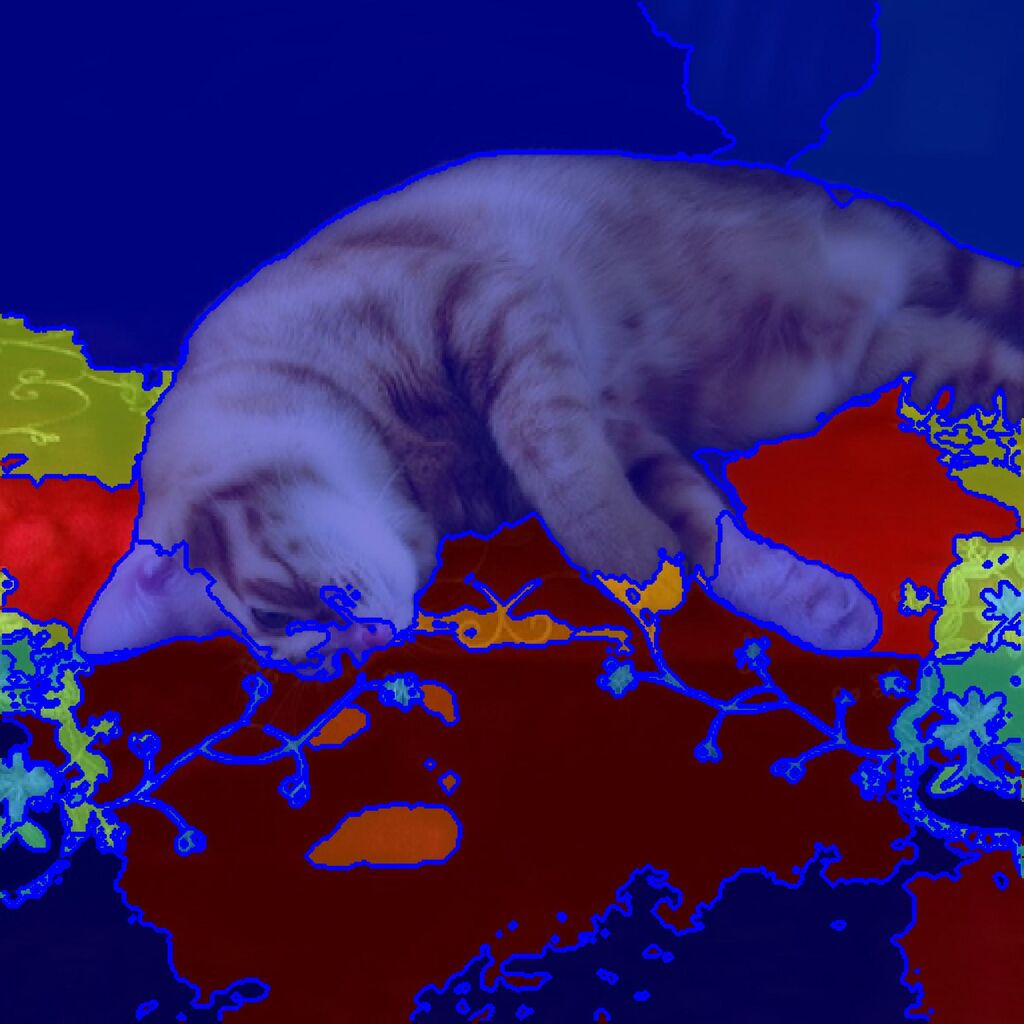}
    \end{minipage}%
    \begin{minipage}[c]{0.105\textwidth}
        \includegraphics[width=\textwidth]{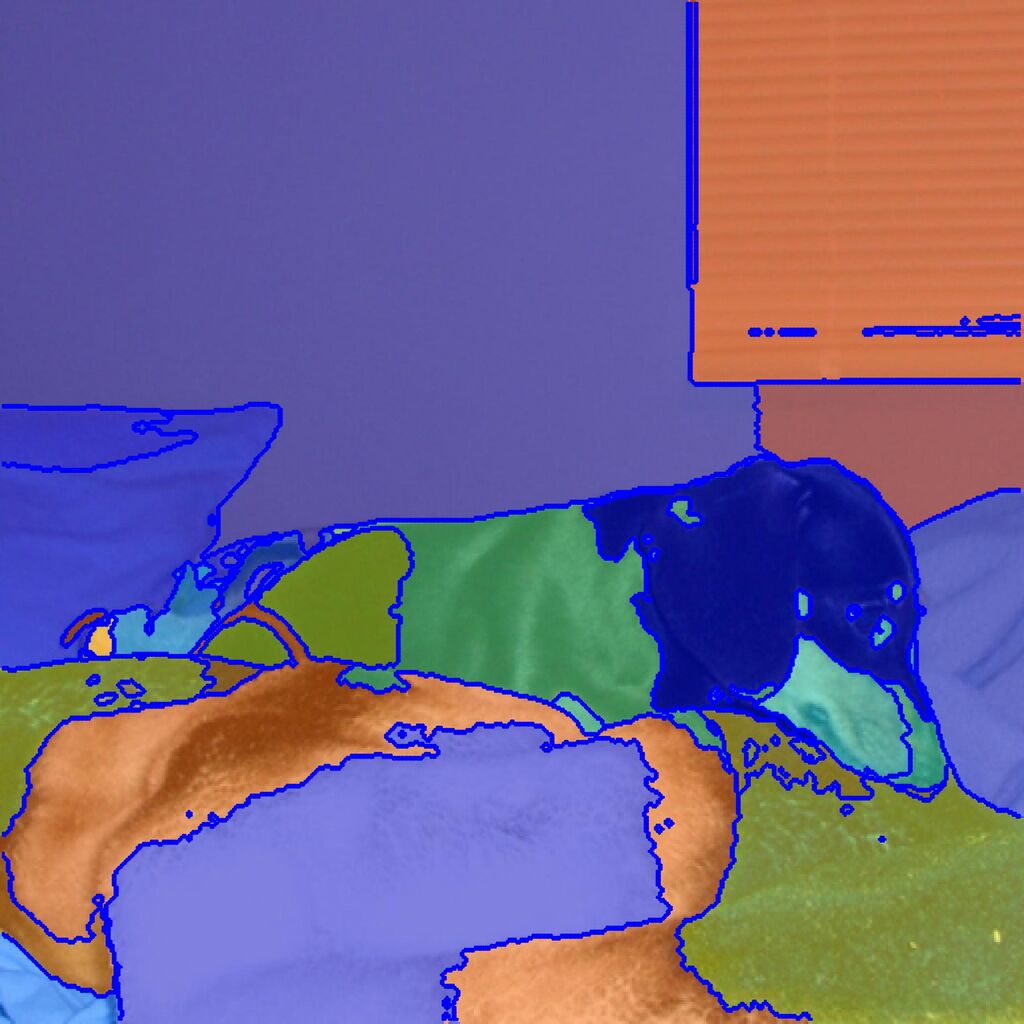}
    \end{minipage}%
    \begin{minipage}[c]{0.105\textwidth}
        \includegraphics[width=\textwidth]{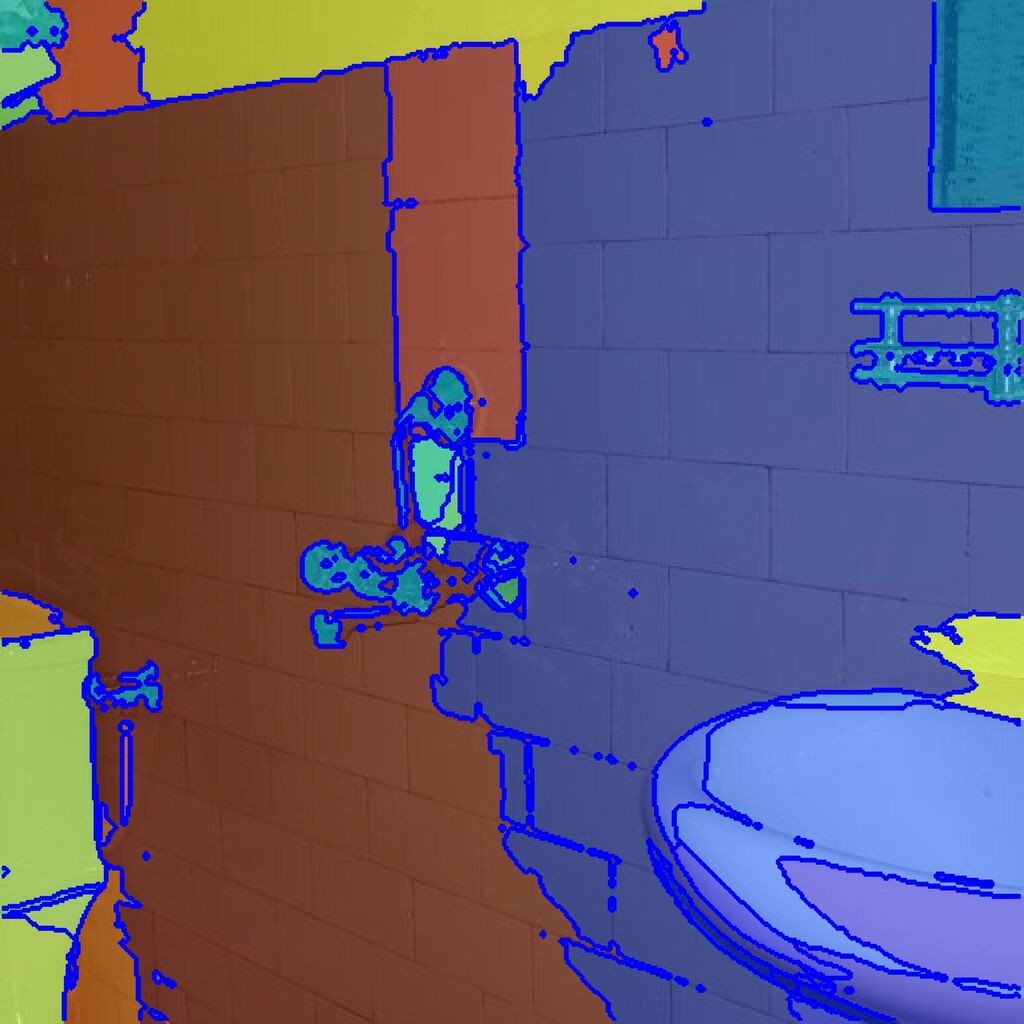}
    \end{minipage}%
    \begin{minipage}[c]{0.105\textwidth}
        \includegraphics[width=\textwidth]{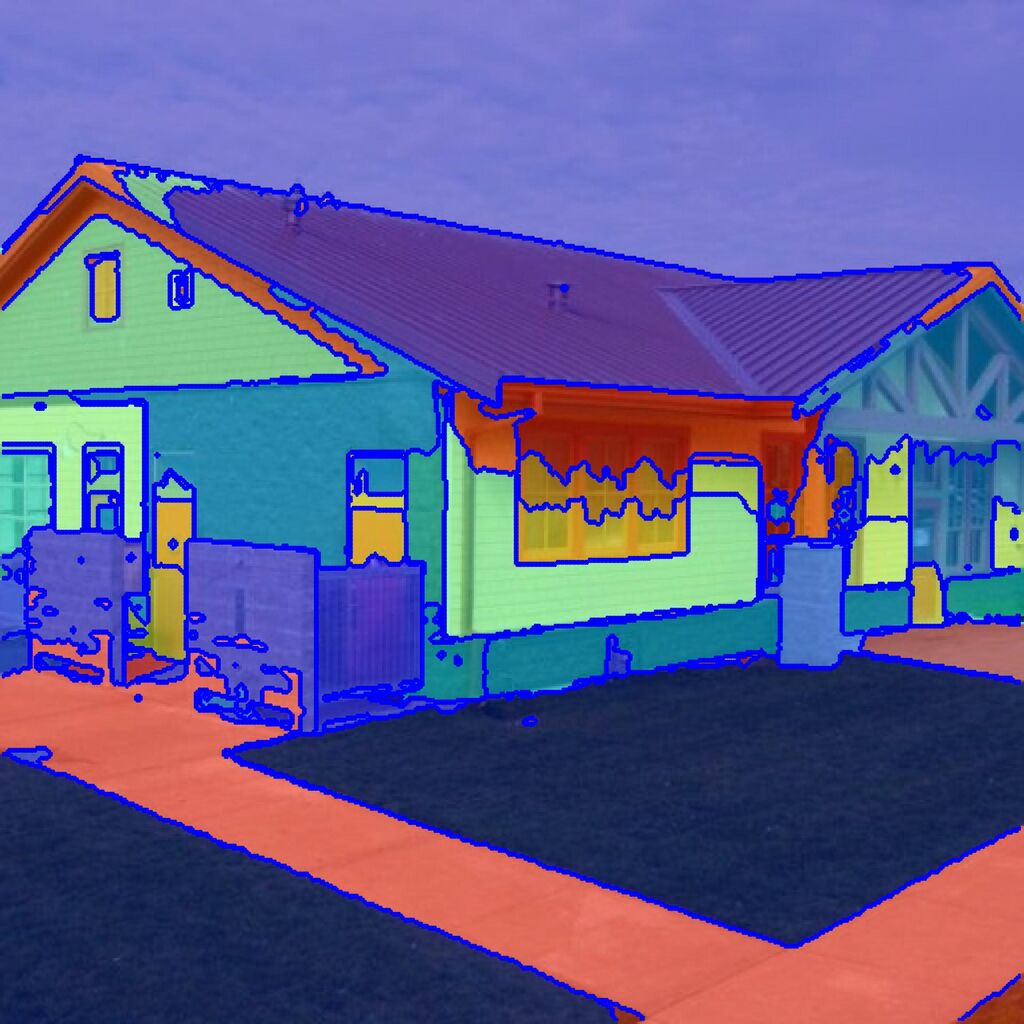}
    \end{minipage}%
    \begin{minipage}[c]{0.105\textwidth}
        \includegraphics[width=\textwidth]{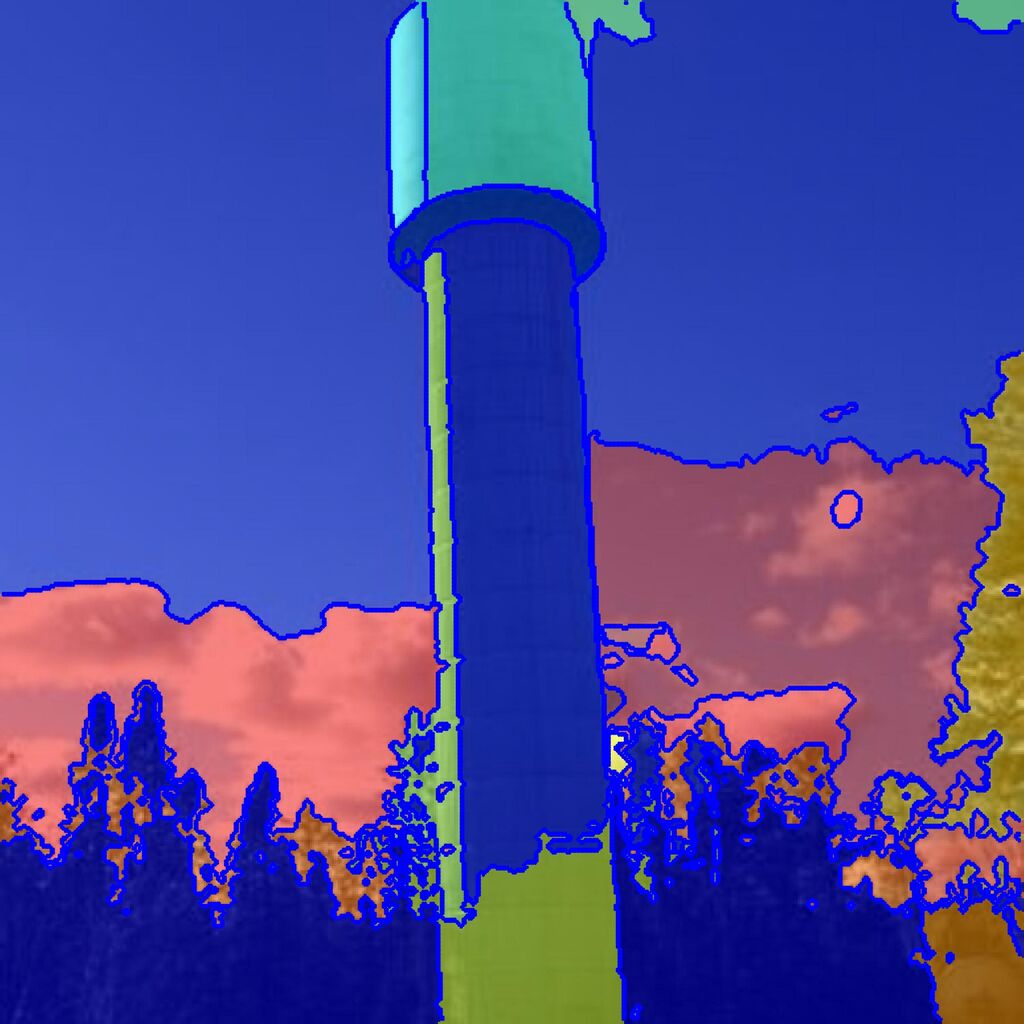}
    \end{minipage}%
    \begin{minipage}[c]{0.105\textwidth}
        \includegraphics[width=\textwidth]{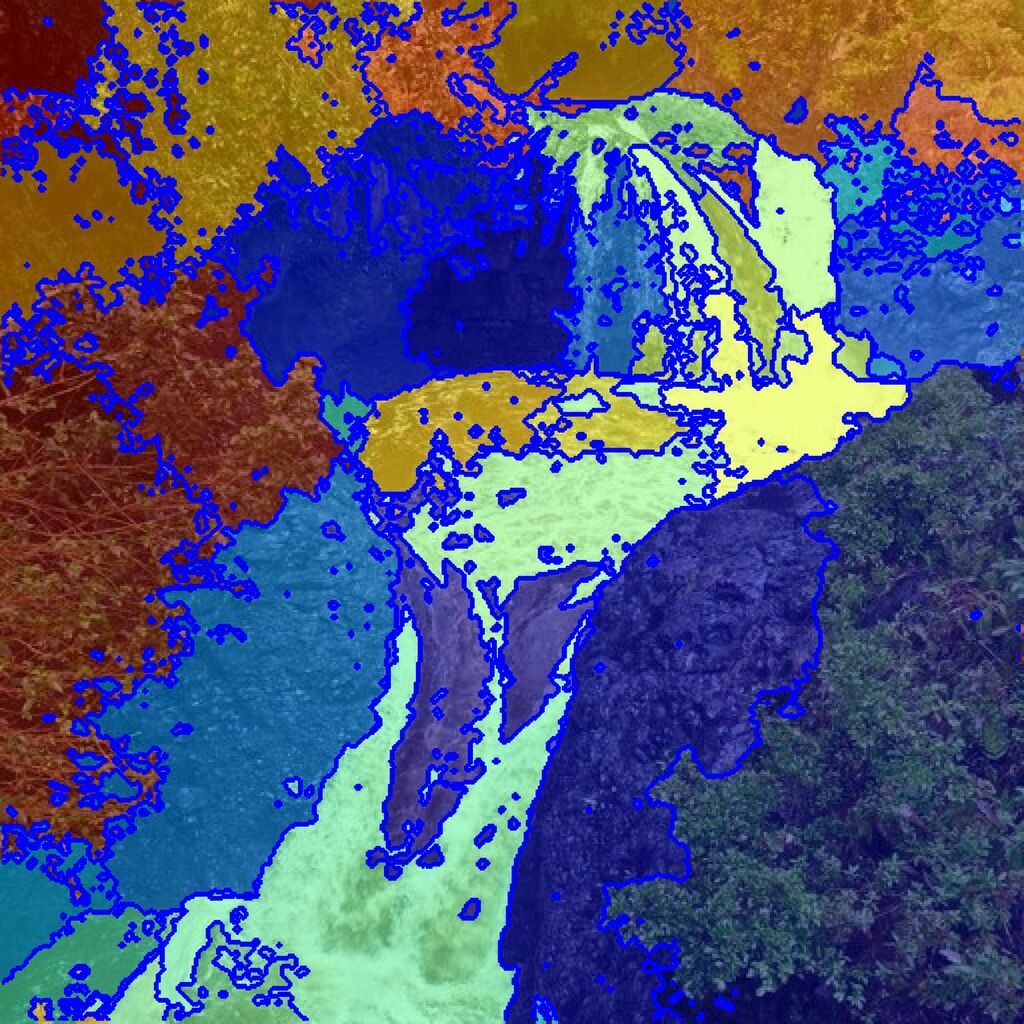}
    \end{minipage}%

    \vspace{-0.5em}

    \begin{minipage}[c]{0.05\textwidth}
        \rotatebox{90}{\textbf{DiffSeg}}
    \end{minipage}%
    \begin{minipage}[c]{0.105\textwidth}
        \includegraphics[width=\textwidth]{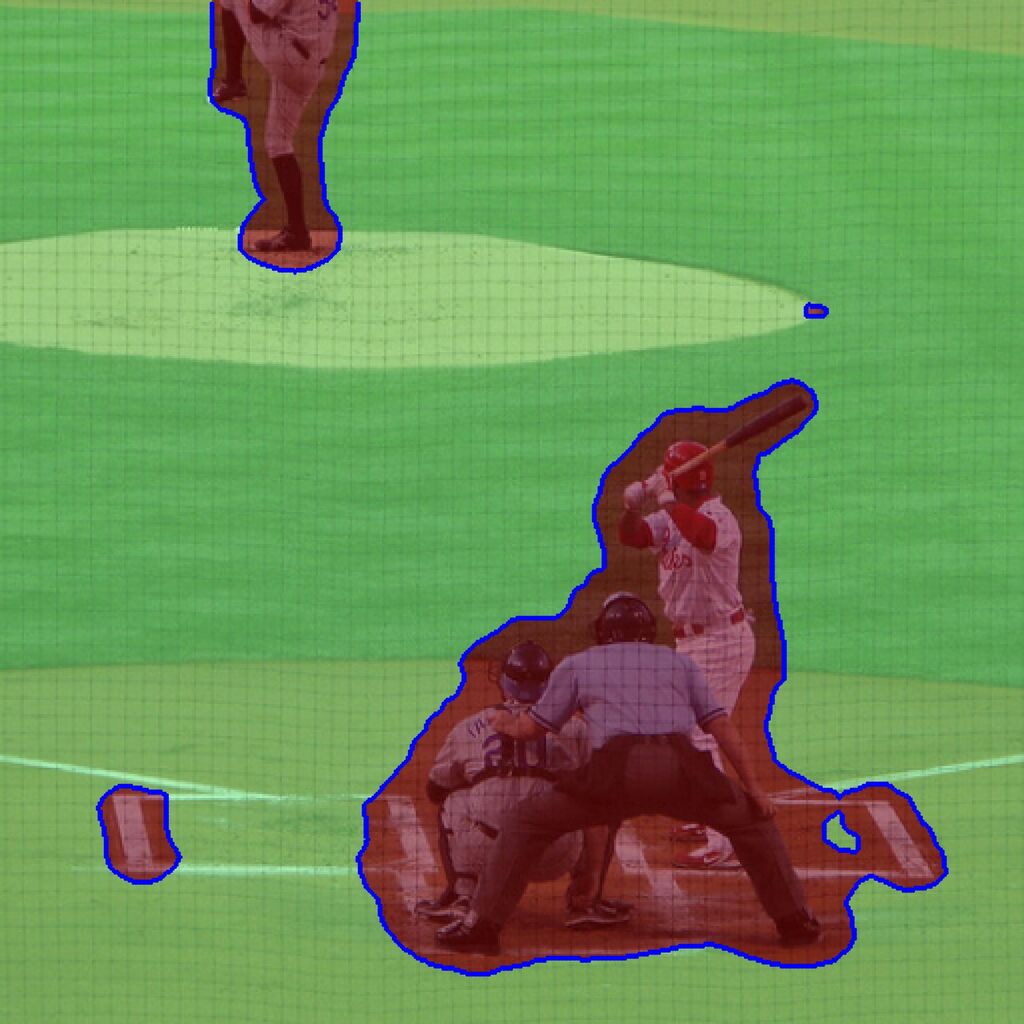}
    \end{minipage}%
    \begin{minipage}[c]{0.105\textwidth}
        \includegraphics[width=\textwidth]{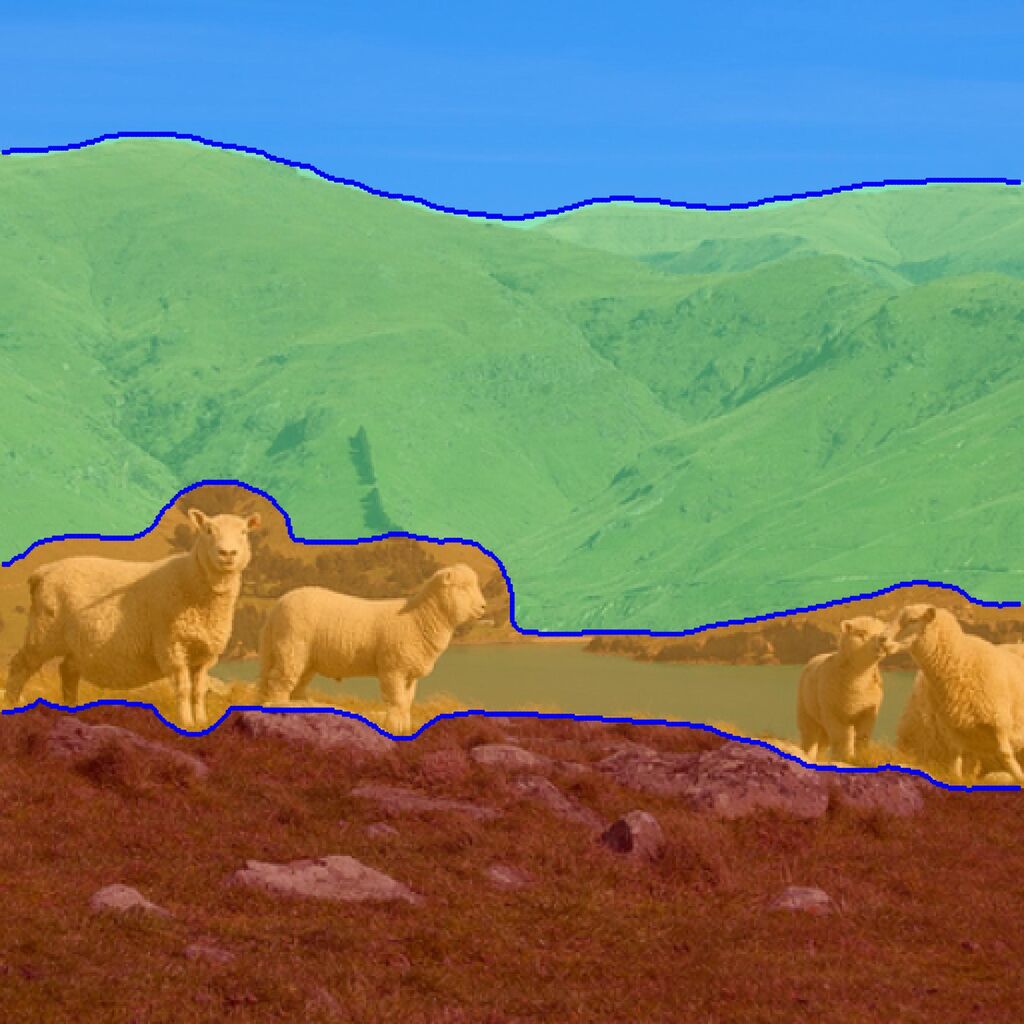}
    \end{minipage}%
    \begin{minipage}[c]{0.105\textwidth}
        \includegraphics[width=\textwidth]{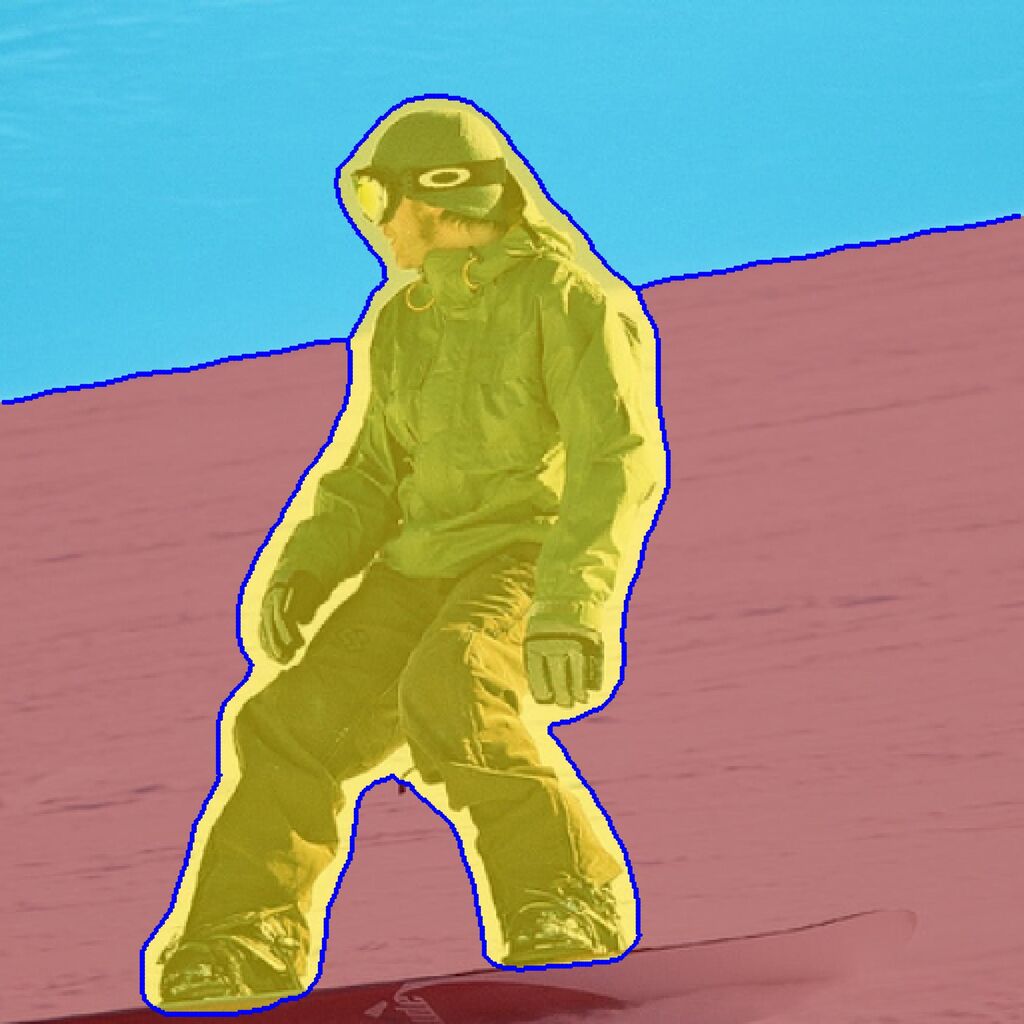}
    \end{minipage}%
    \begin{minipage}[c]{0.105\textwidth}
        \includegraphics[width=\textwidth]{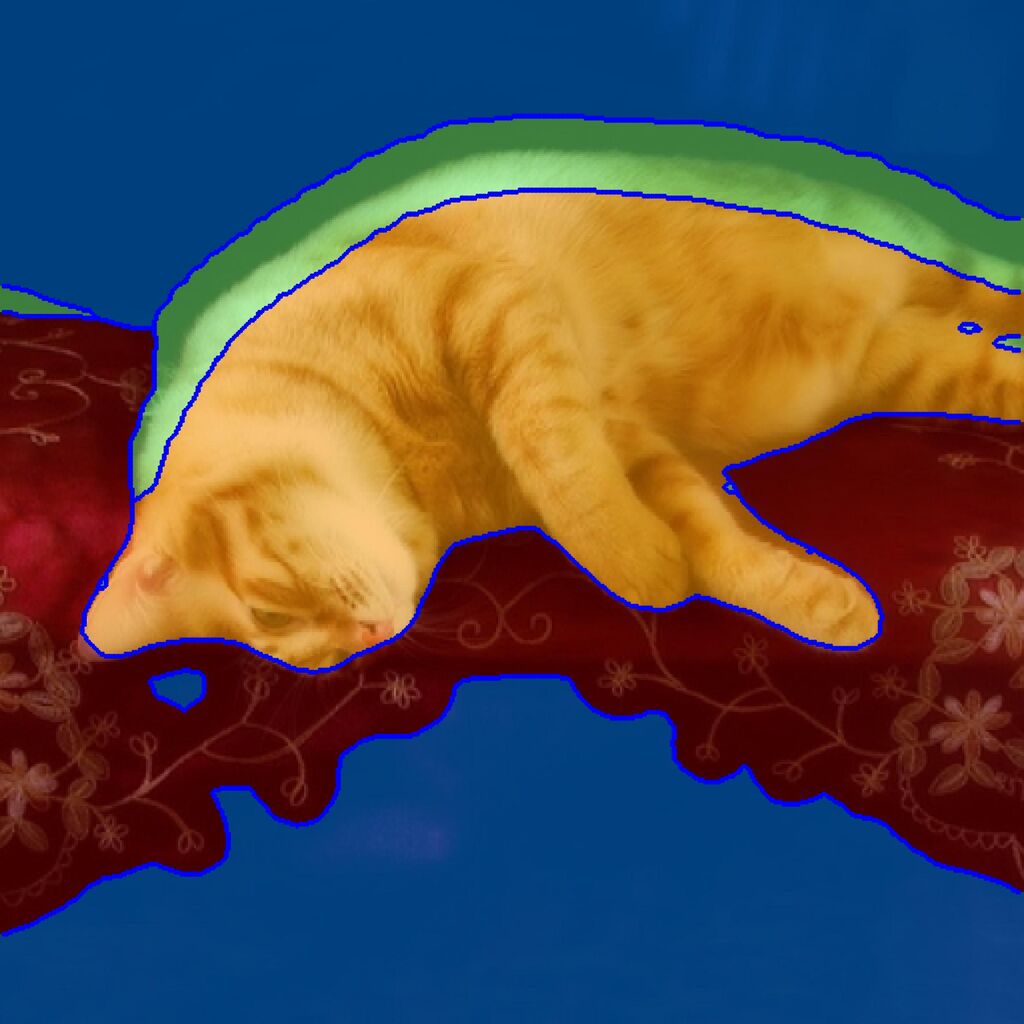}
    \end{minipage}%
    \begin{minipage}[c]{0.105\textwidth}
        \includegraphics[width=\textwidth]{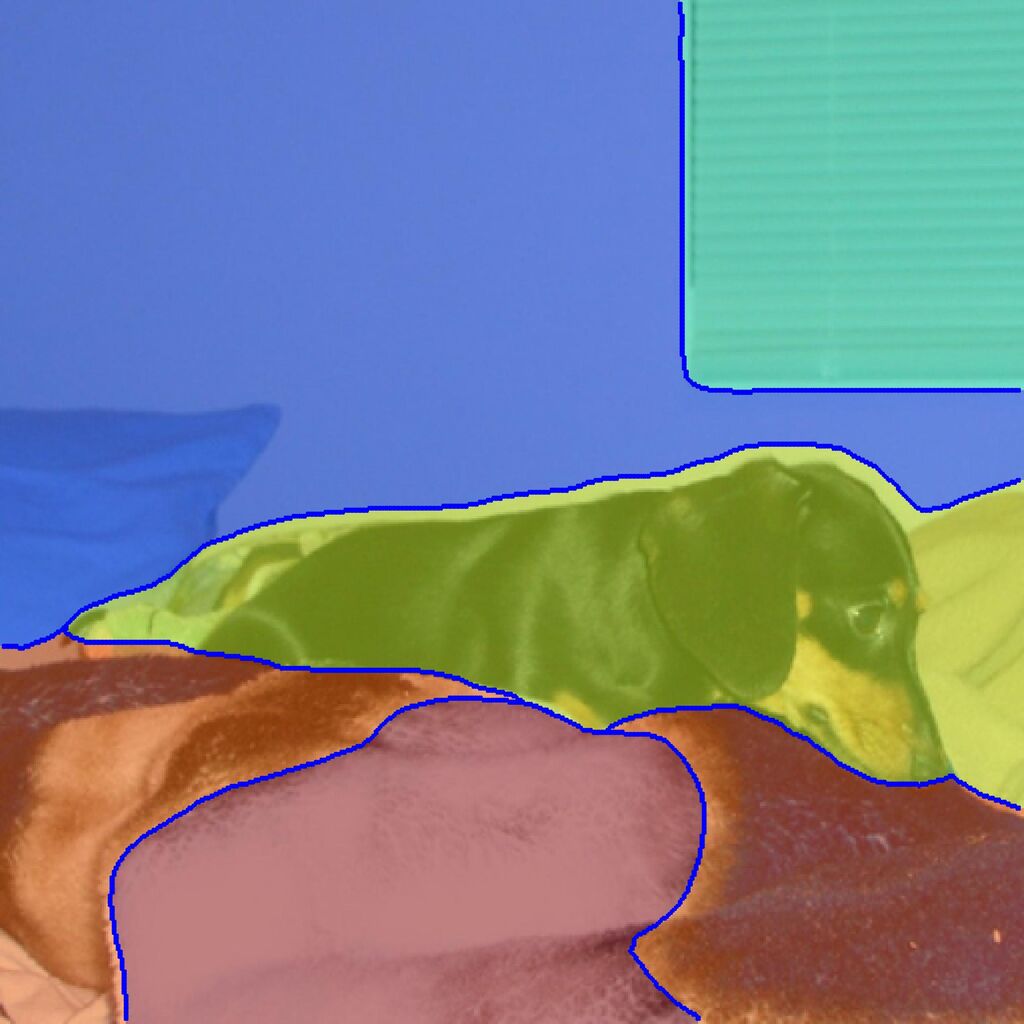}
    \end{minipage}%
    \begin{minipage}[c]{0.105\textwidth}
        \includegraphics[width=\textwidth]{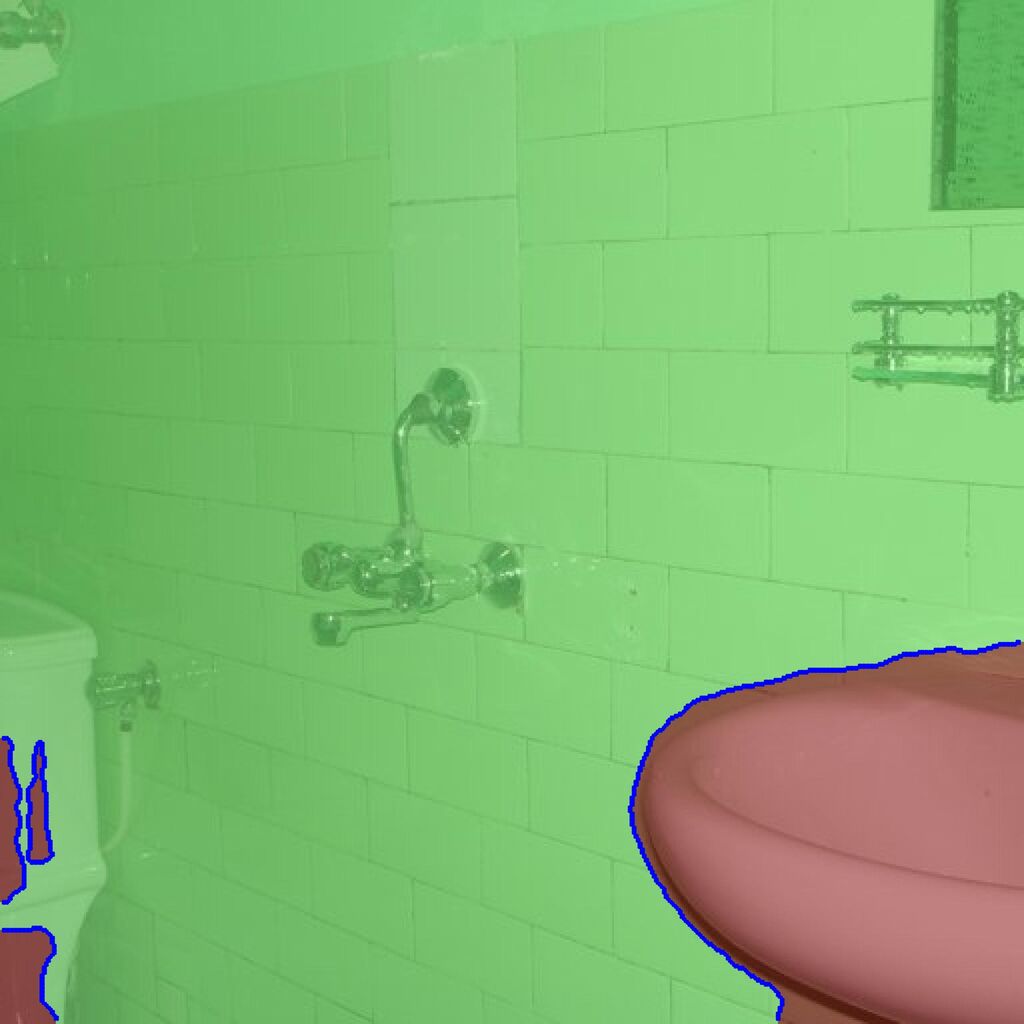}
    \end{minipage}%
    \begin{minipage}[c]{0.105\textwidth}
        \includegraphics[width=\textwidth]{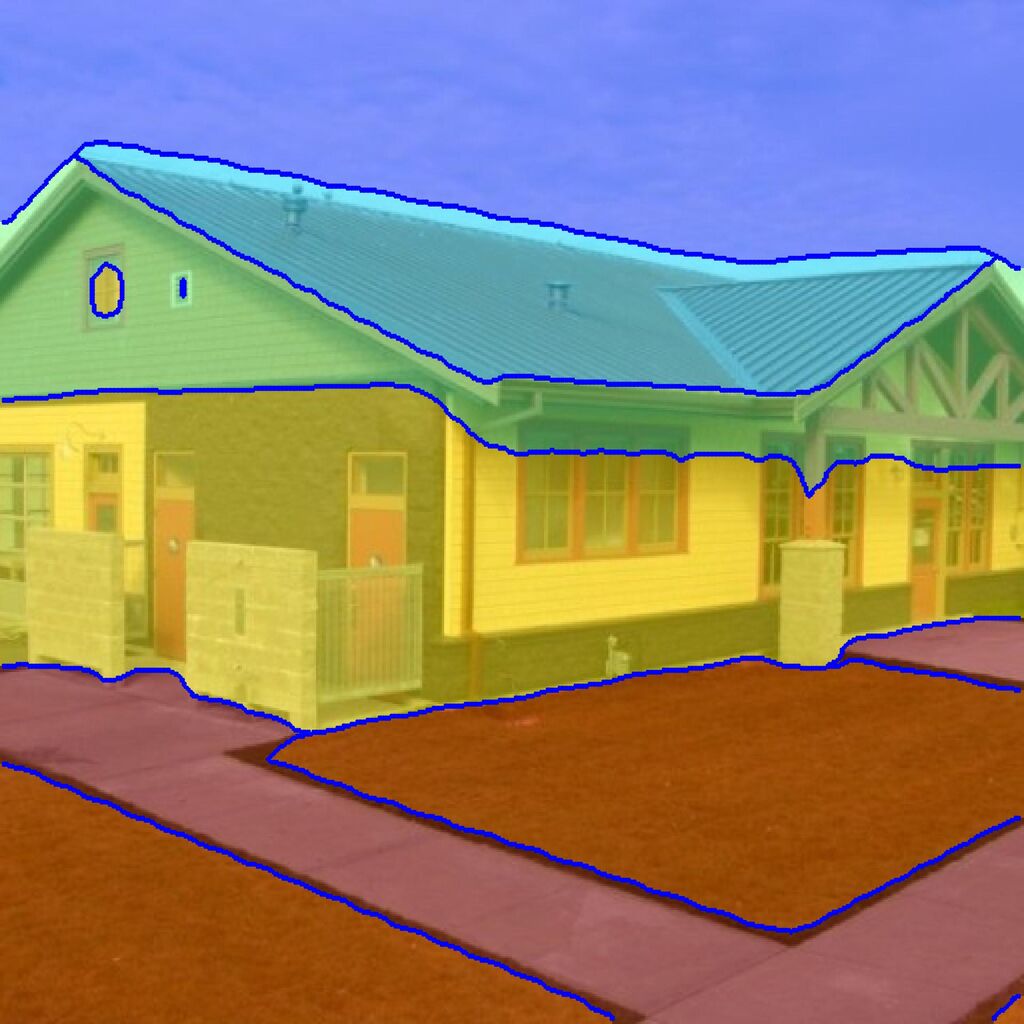}
    \end{minipage}%
    \begin{minipage}[c]{0.105\textwidth}
        \includegraphics[width=\textwidth]{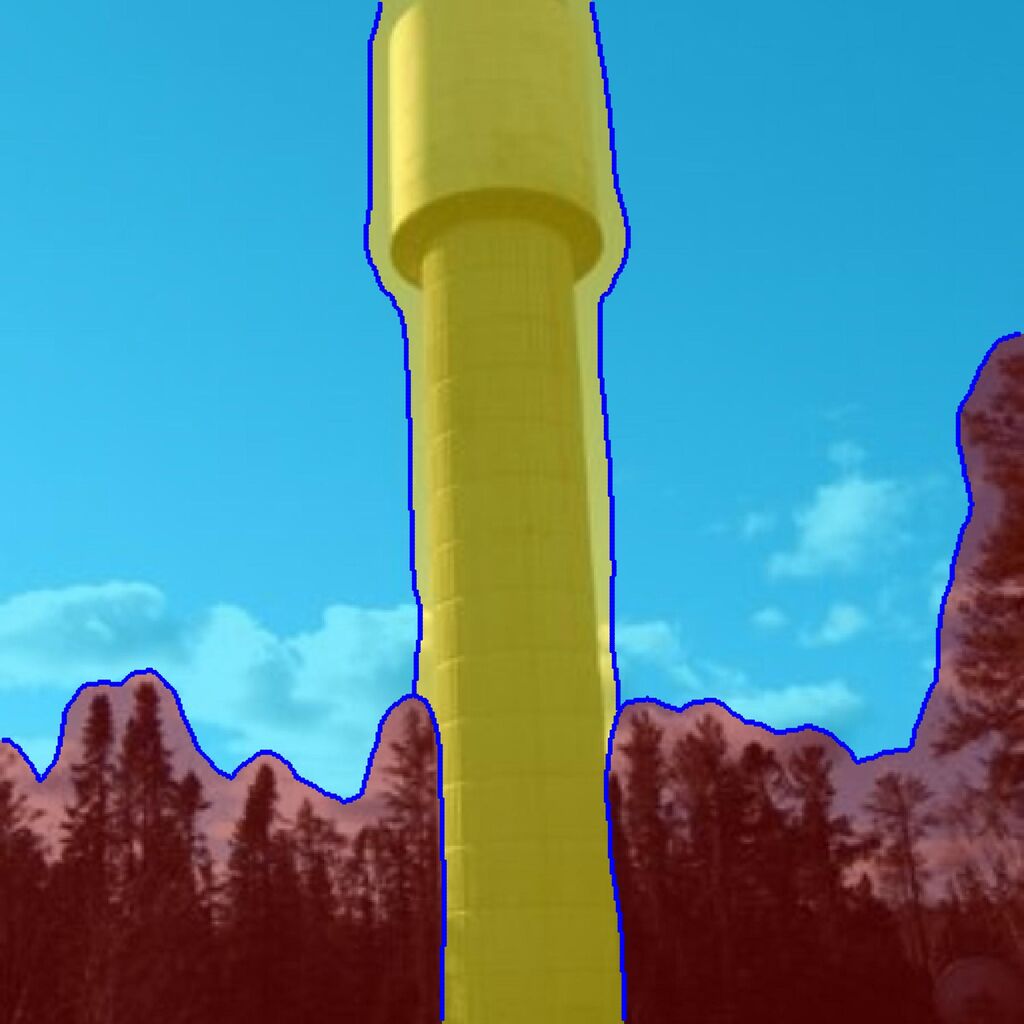}
    \end{minipage}%
    \begin{minipage}[c]{0.105\textwidth}
        \includegraphics[width=\textwidth]{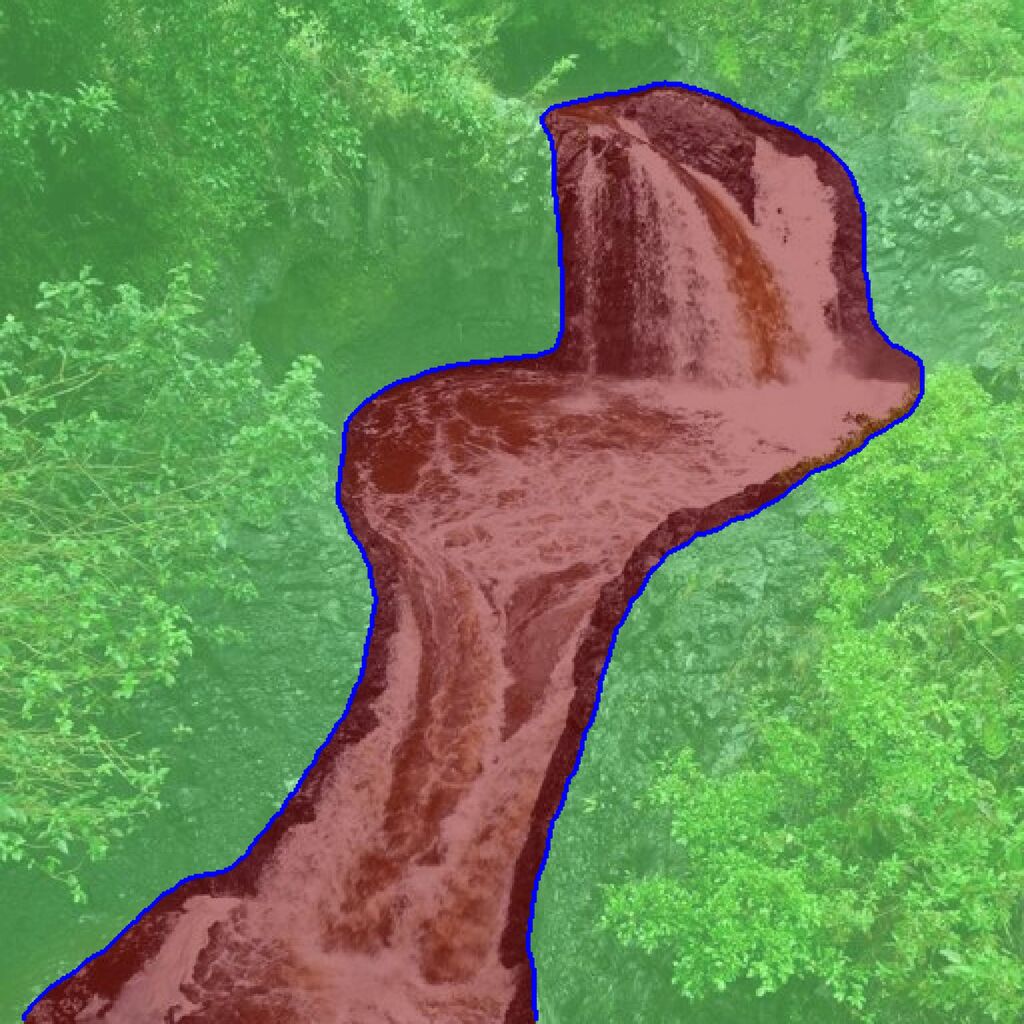}
    \end{minipage}%
    
    \vspace{-0.5em}

    \begin{minipage}[c]{0.05\textwidth}
        \rotatebox{90}{\textbf{Ours}}
    \end{minipage}%
    \begin{minipage}[c]{0.105\textwidth}
        \includegraphics[width=\textwidth]{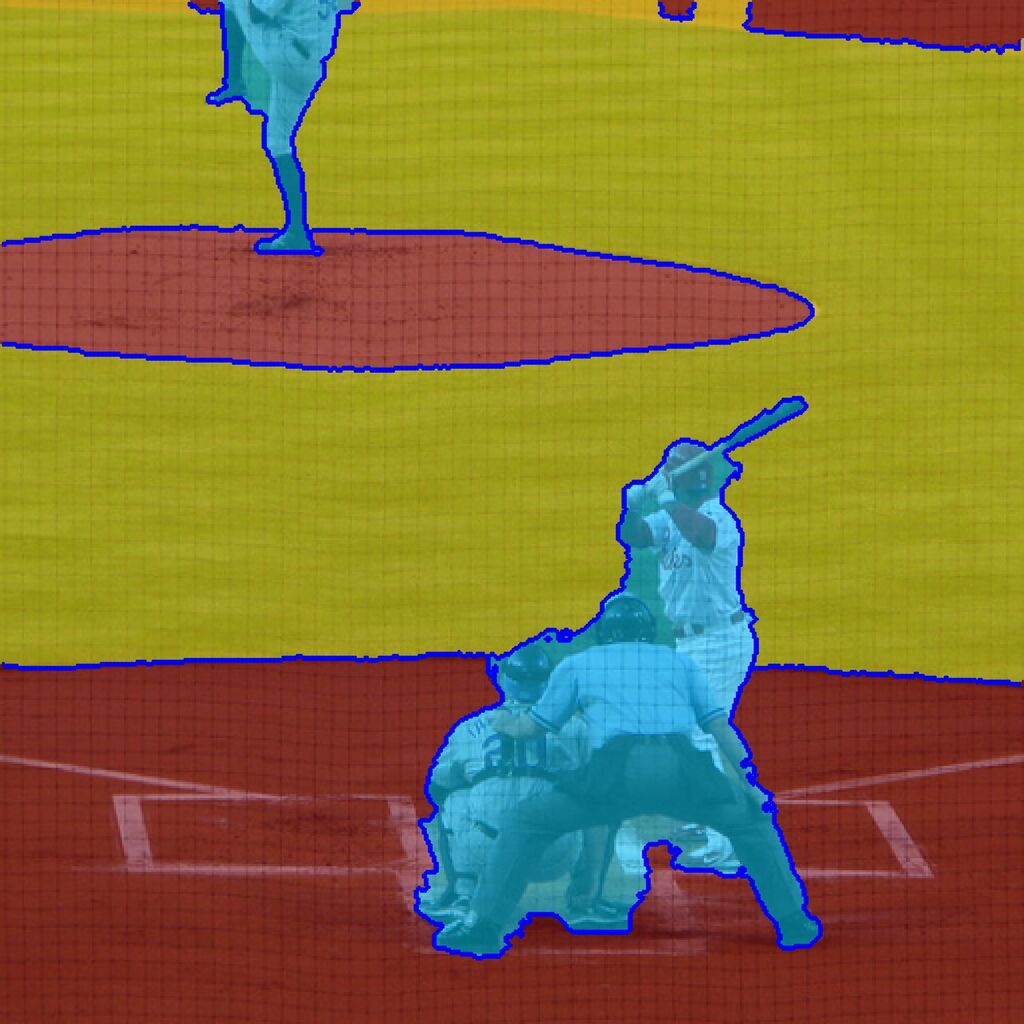}
    \end{minipage}%
    \begin{minipage}[c]{0.105\textwidth}
        \includegraphics[width=\textwidth]{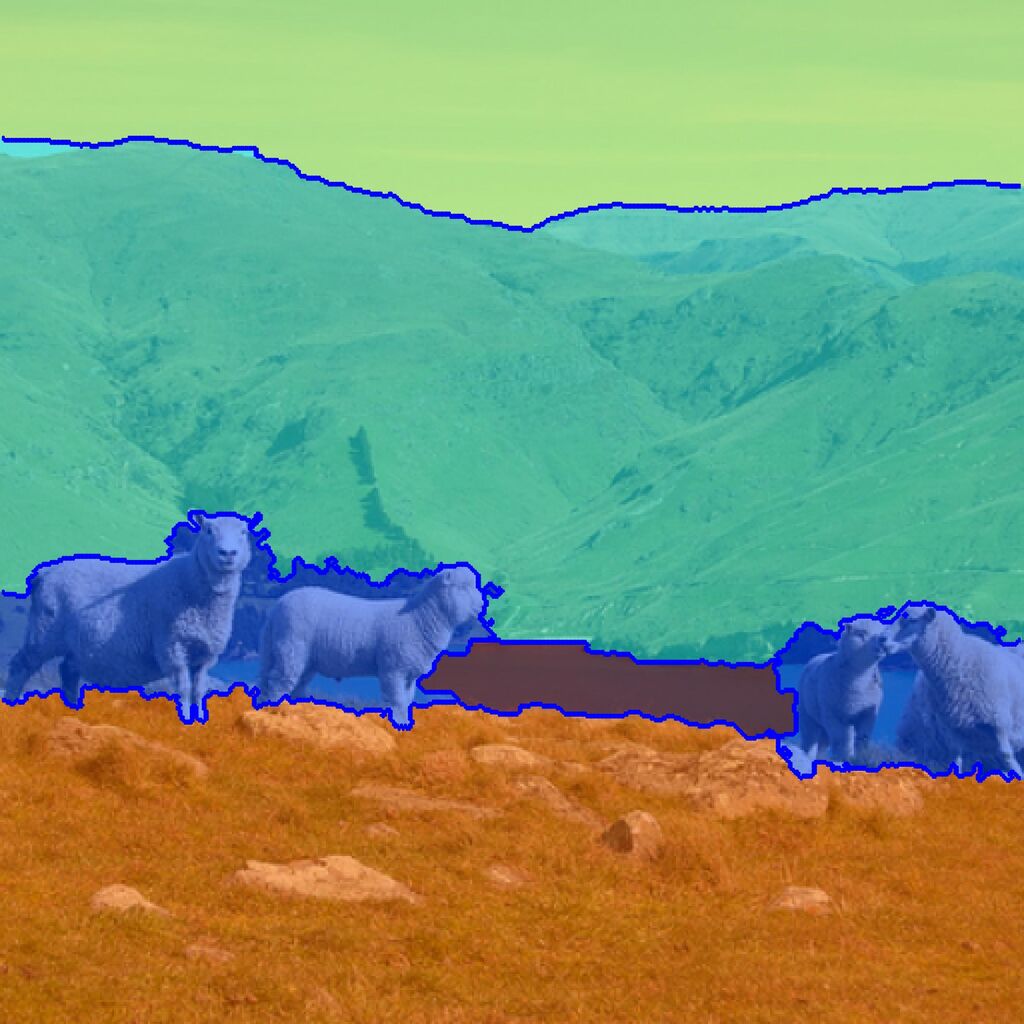}
    \end{minipage}%
    \begin{minipage}[c]{0.105\textwidth}
        \includegraphics[width=\textwidth]{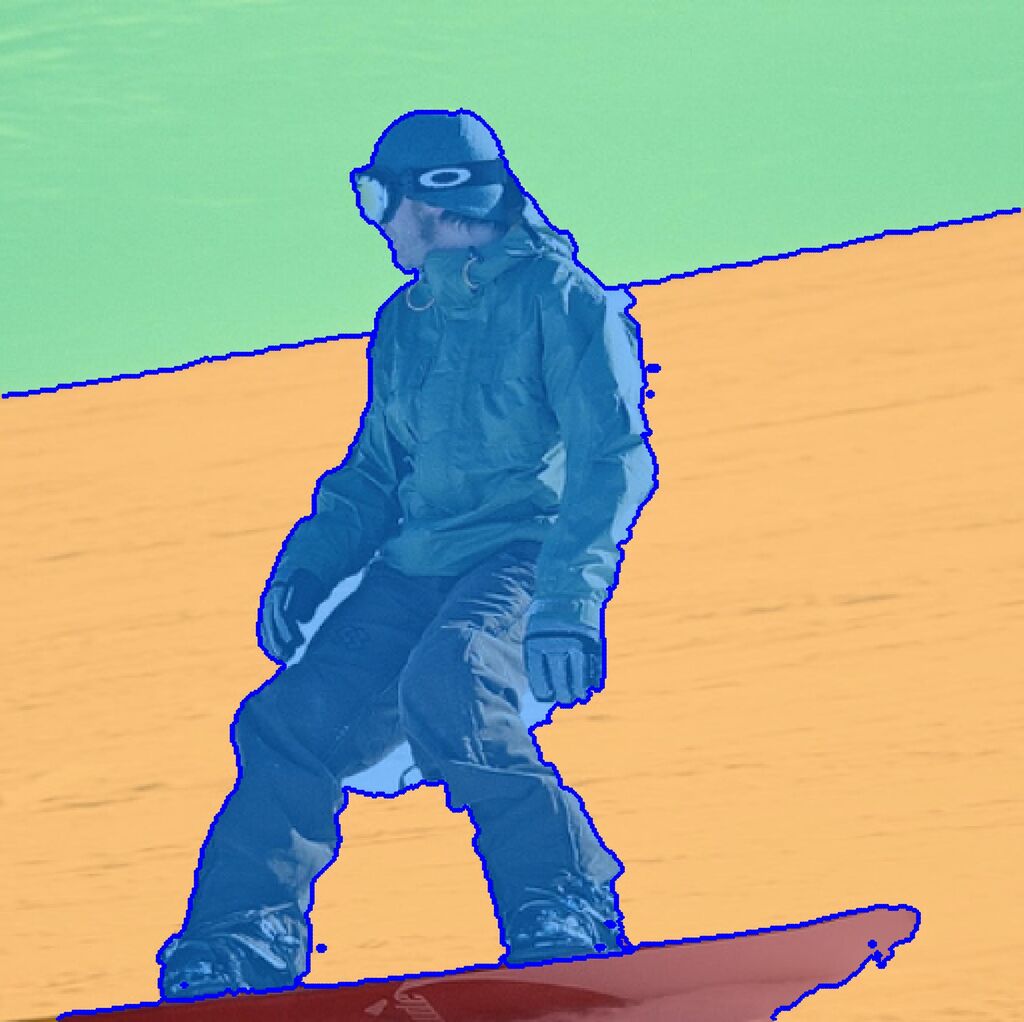}
    \end{minipage}%
    \begin{minipage}[c]{0.105\textwidth}
        \includegraphics[width=\textwidth]{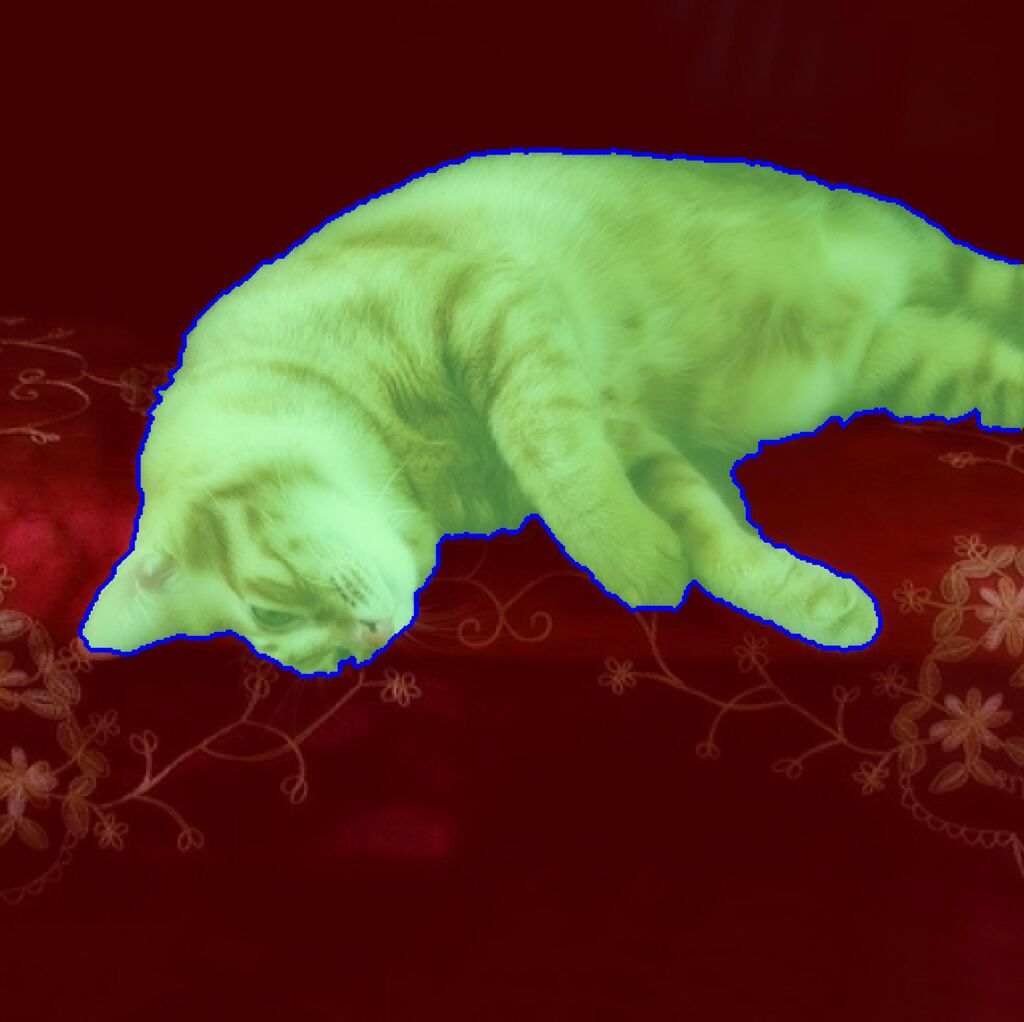}
    \end{minipage}%
    \begin{minipage}[c]{0.105\textwidth}
        \includegraphics[width=\textwidth]{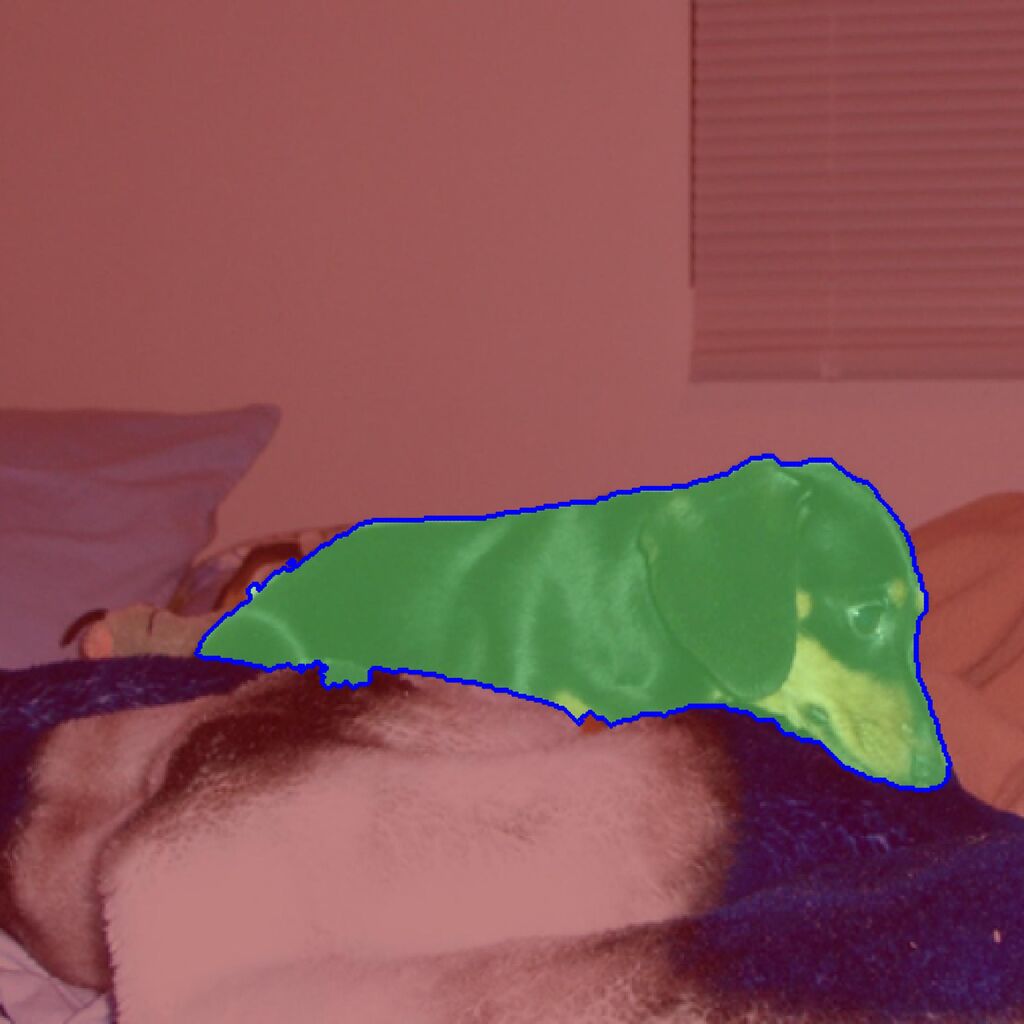}
    \end{minipage}%
    \begin{minipage}[c]{0.105\textwidth}
        \includegraphics[width=\textwidth]{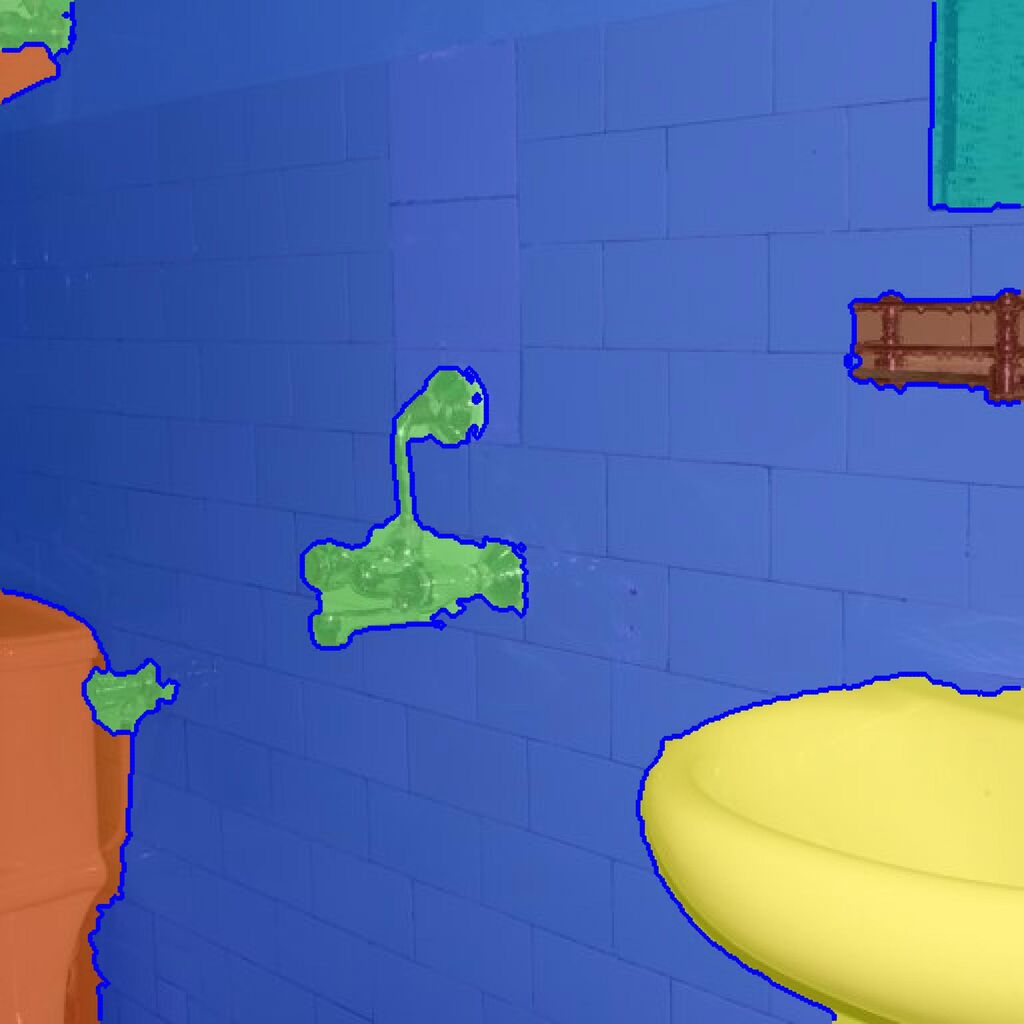}
    \end{minipage}%
    \begin{minipage}[c]{0.105\textwidth}
        \includegraphics[width=\textwidth]{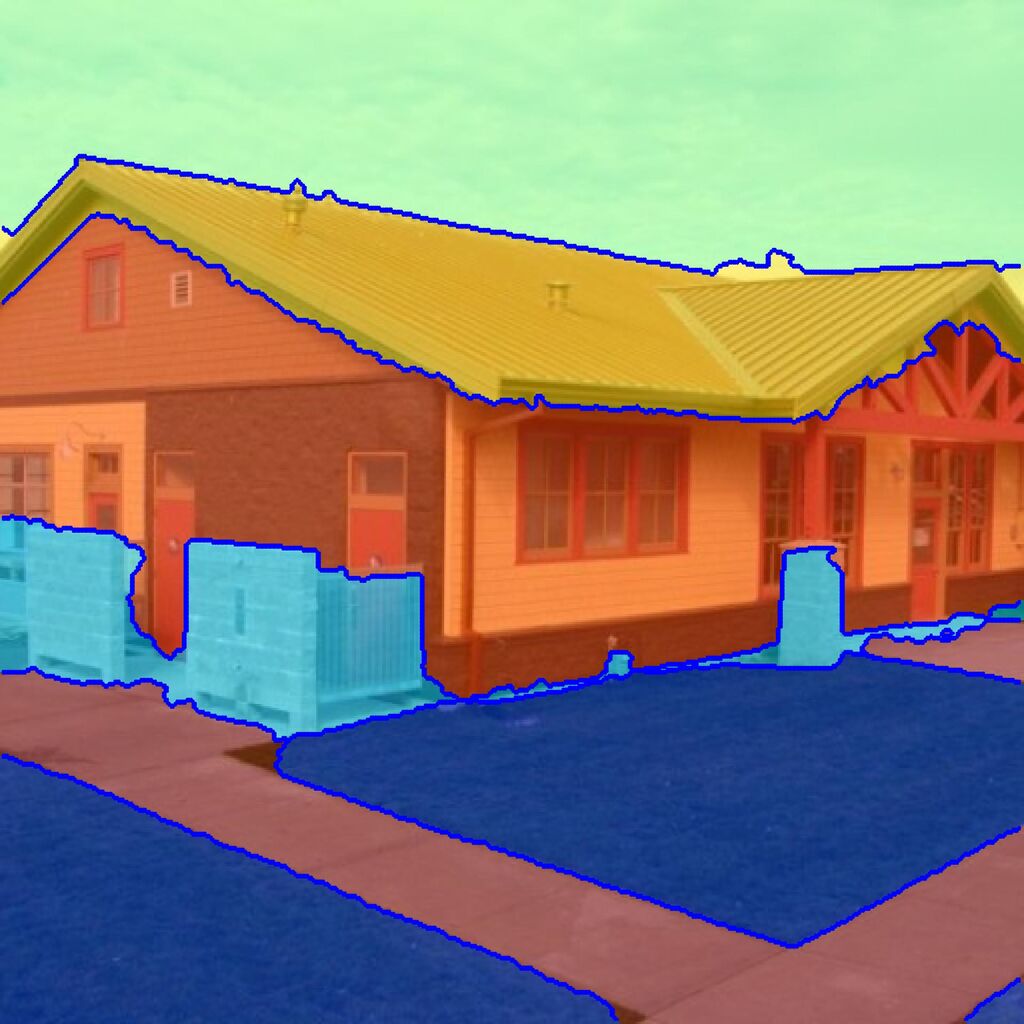}
    \end{minipage}%
    \begin{minipage}[c]{0.105\textwidth}
        \includegraphics[width=\textwidth]{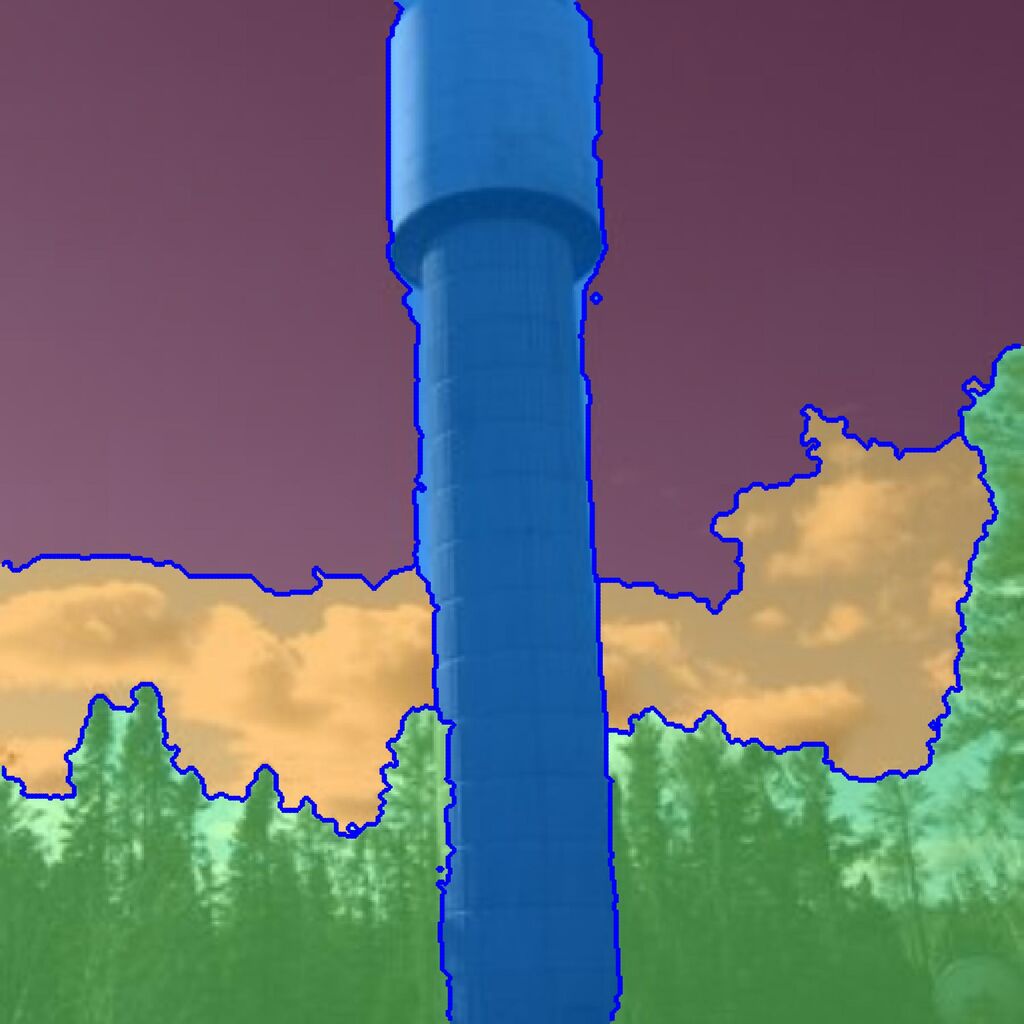}
    \end{minipage}%
    \begin{minipage}[c]{0.105\textwidth}
        \includegraphics[width=\textwidth]{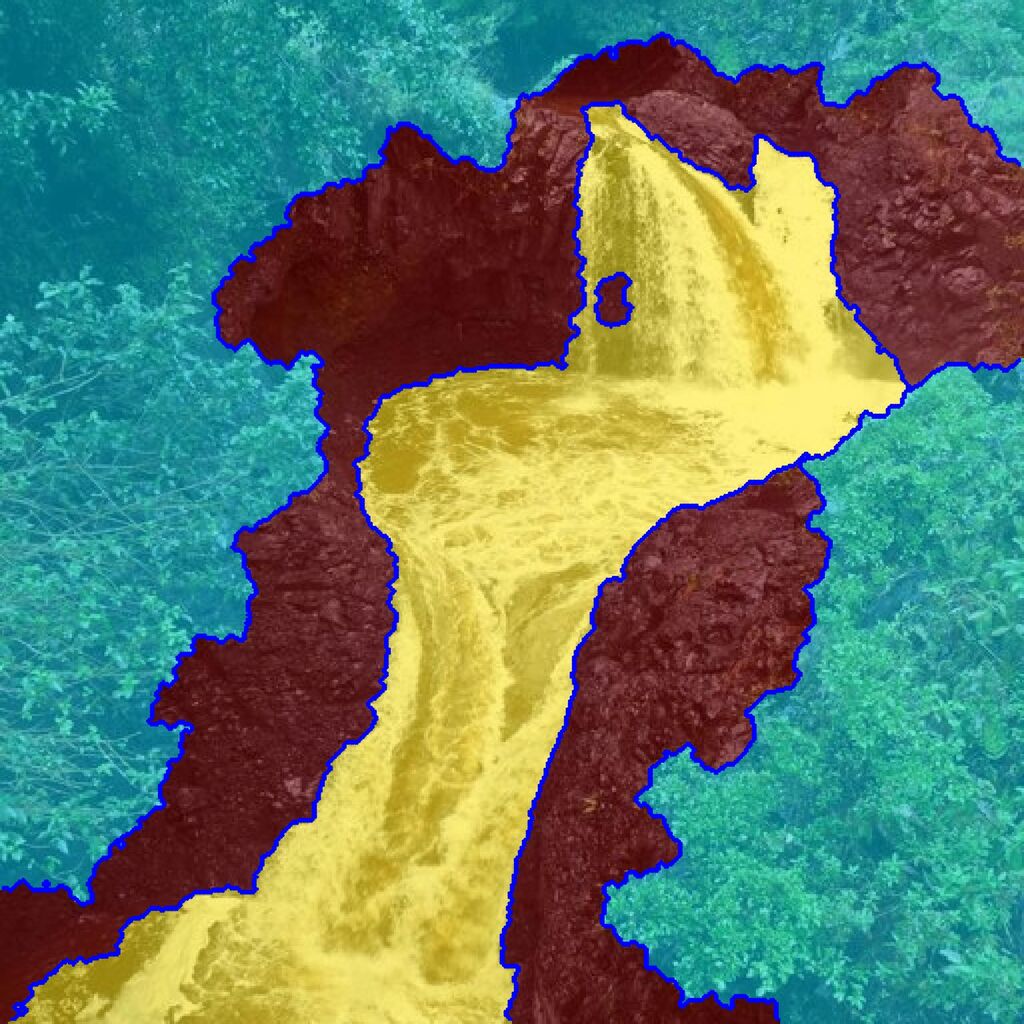}
    \end{minipage}%
    
    \caption{Qualitative comparison of the proposed method against DiffSeg~\cite{tian2024diffuse} and DiffCut~\cite{couairon2024diffcut}.}
    \label{fig:baseline_qualitative_comparison}
\end{figure*}

\begin{abstract}
Zero-shot segmentation has recently shown notable improvement by leveraging the rich visual priors in large-scale text-to-image diffusion models, such as Stable Diffusion. However, current diffusion-based methods often face limitations due to the trade-off between spatial resolution and contextual information\replaced{, as well as their reliance on a single static timestep for feature extraction.}{. Furthermore, their reliance on static feature extraction from a single denoising timestep overlooks the dynamic evolution of semantic granularity inherent in the diffusion process.} To overcome these\added{ challenges}, our work introduces two key advancements. First, our Contextual Similarity Maps fuse high-resolution attention maps with rich U-Net encoder features, providing both fine-grained and robust per-pixel representations. Second, \replaced{we identify an emergent hierarchical semantic progression within the denoising process of various diffusion models: representations transition from part-level abstractions at earlier timesteps to object-level abstractions at later stages. Leveraging this insight,}{recognizing that these maps capture hierarchical semantic abstractions across the denoising process,} we introduce a mechanism to adaptively select the optimal timestep for each pixel. \replaced{Extensive}{Our extensive} experiments demonstrate that our method consistently outperforms \deleted{all} existing zero-shot segmentation baselines, validating the efficacy of combining contextual features with dynamic, hierarchical timestep selection.
\end{abstract}    
\section{Introduction}
\label{sec:intro}

The ability to precisely delineate semantic regions at the pixel level is fundamental to a myriad of computer vision applications, from autonomous navigation~\cite{Chen2017DeepLabSP, cordts2016cityscapes} and augmented reality~\cite{lalonde2018deep} to medical image analysis~\cite{ronneberger2015u, jiang2025advancing}. Deep learning has propelled semantic segmentation to unprecedented performance, largely driven by the availability of meticulously annotated, large-scale datasets~\cite{Chen2017DeepLabSP, Cheng_2022_CVPR, cordts2016cityscapes, Long2015FullyCN, lin2014microsoft, zhou2019semantic}. However, despite the proliferation of such resources, the inherent reliance on extensive labeled data poses a critical bottleneck for real-world adaptability and scalability. Supervised models, while powerful, demonstrate limited generalization to novel environments, different lighting conditions, or entirely unseen object categories~\cite{gao2024generalize, benigmim2024collaborating, kim2023texture, jia2024dginstyle}. The fixed semantic taxonomies of existing benchmarks often fail to capture the nuanced or domain-specific semantic concepts required by emergent applications. This continuous demand for costly and labor-intensive re-annotation or fine-tuning for every new domain or granular semantic level fundamentally limits widespread deployment~\cite{fobi2020learning, zhou2017scene, grad2024benchmarking, qiu2024aligndiff}. This challenge has accelerated research into zero-shot segmentation, a pivotal paradigm that seeks to achieve fine-grained understanding without reliance on \deleted{any} class-specific training examples or dense pixel annotations~\cite{tian2024diffuse, couairon2024diffcut}.

The recent advent of large-scale text-to-image diffusion models, exemplified by Stable Diffusion~\cite{rombach2022latent}, has opened unprecedented avenues for addressing this zero-shot challenge. Trained on vast, web-scale image-text data~\cite{schuhmann2022laion}, these models internalize a rich and implicit understanding of visual semantics~\cite{rombach2022latent, ramesh2021zero, tang2023emergent}. Pioneering works~\cite{couairon2024diffcut, tian2024diffuse} have compellingly demonstrated that the internal representations, particularly the self-attention maps and U-Net features within the denoising process, can serve as powerful priors for unsupervised semantic segmentation. These methods leverage the inherent ability of diffusion models to group pixels into semantically coherent regions based on the learned visual-semantic correspondences.

Despite their promising success, existing diffusion-based zero-shot segmentation methods grapple with two critical limitations. First, they typically rely on either self-attention maps or the encoder features from the U-Net architecture, leading to an unavoidable trade-off between resolution and contextual information. Approaches utilizing high-resolution self-attention maps excel in capturing fine boundary details but often lack the comprehensive contextual understanding necessary for robust semantic grouping~\cite{tian2024diffuse}. Conversely, methods employing features from deeper encoder layers offer rich contextual representations but at the cost of significantly reduced spatial resolution, leading to coarse segmentation boundaries~\cite{couairon2024diffcut}. This dichotomy compromises either the boundary precision or the semantic coherence of the generated masks. Second, and more fundamentally, these methods extract features from a single, fixed timestep within the diffusion process. This static approach overlooks the dynamic evolution of \deleted{semantic} information throughout the denoising process~\cite{ho2020denoising, luo2023diffusion}. As a diffusion model iteratively refines a noisy latent into a coherent image, different levels of \deleted{semantic} granularity emerge and solidify at varying timesteps. Relying on a single timestep inherently limits the model's ability to capture the full spectrum of \deleted{semantic} cues required for accurate and nuanced segmentation.

In this work, we propose a novel approach that overcomes these limitations by fully harnessing the multi-faceted semantic knowledge encoded within the denoising process of pre-trained \replaced{diffusion models}{Stable Diffusion}. Our core insight is twofold: (1) to combine the complementary strengths of high-resolution attention and contextual encoder features into a unified contextual similarity map, thereby achieving both fine boundary delineation and robust semantic coherence; and (2) our investigations reveal a hierarchical progress of the semantic information in \deleted{the contextual similarity map of} the denoising process of each pixel in the latent space, allowing us to move beyond static timestep selection for feature extraction. Our method adaptively selects the most semantically informative timestep for each pixel individually, allowing for a more precise and comprehensive understanding of image semantics.

Our contributions can be summarized as follows:
\begin{itemize}
    \item We introduce a novel approach for zero-shot semantic segmentation that effectively leverages the contextual internal representations of pre-trained \replaced{diffusion}{Stable Diffusion} models.
    \item We propose a method to generate contextual similarity maps by synergically fusing high-resolution attention information with contextually rich U-Net features, thereby mitigating the resolution-context trade-off.
    \item We reveal a hierarchical progress of semantic information within diffusion models across timesteps and propose a mechanism to exploit this observation to find the optimal timestep for feature extraction along the denoising process.
    \item We present extensive experimental results demonstrating that the proposed method achieves superior performance on various zero-shot semantic segmentation benchmarks, outperforming existing diffusion-based approaches.
\end{itemize}

\begin{figure*}[h!]
  \centering
  
  \includegraphics[width=\textwidth]{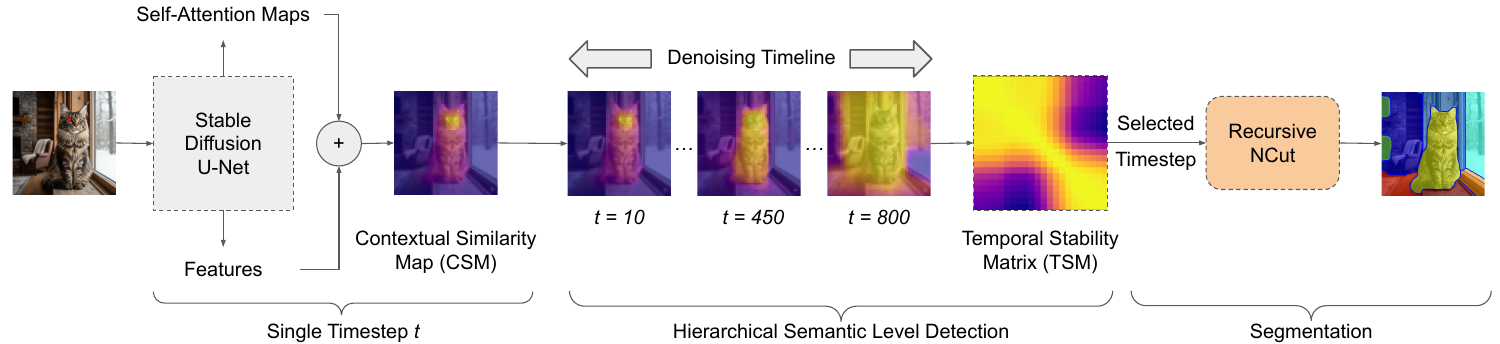}
  \caption{Method Overview. At each timestep $t$, a Contextual Similarity Map (CSM) is generated using the self-attention maps and the features from a Stable Diffusion U-Net. A Temporal Stability Matrix (TSM) is then computed by measuring the similarity between CSMs at different timesteps. The resulting TSM is used to identify the optimal timestep for each pixel. The corresponding CSM is then used to generate the final segmentation mask.}
  \label{fig:graphical_abstract}
\end{figure*}

\section{Related work}
\label{sec:related_work}

\subsection{Semantic segmentation}

Semantic segmentation evolved from hand-crafted features~\cite{Shotton2006TextonBoostFJ,Lafferty2001ConditionalRF} to deep learning CNNs~\cite{Long2015FullyCN}. Encoder-decoder architectures like U-Net~\cite{Ronneberger2015UNetCN} used skip connections to recover spatial detail. Subsequent work such as the DeepLab family~\cite{Chen2017DeepLabSP,Chen2018EncoderDecoderWA} improved feature representation and context aggregation, while Vision Transformers~\cite{Dosovitskiy2021AnII,Zheng2021RethinkingSS,Xie2021SegFormerSA} and promptable models like SAM~\cite{kirillov2023segment,ravi2024sam} have shown strong performance. Many recent works also adopt vision-language models for this purpose~\cite{yang2025hsrdiff,wang2025parables,bi2024learning,nguyen2025calico,lai2024lisa,ulger2025auto}.

\subsection{Unsupervised segmentation}
Unsupervised segmentation has expanded from clustering~\cite{Jabri2019UnsupervisedOW} to self-supervised methods~\cite{cho2021picie, Hamilton2022UnsupervisedSS, oconnor2021unsupervised}. Recent works adopt foundation models, enabling unified frameworks~\cite{niu2024unsupervised}, unsupervised SAM~\cite{wang2024segment}, object-centric semantics~\cite{kim2024eagle}, and DINO upsampling for segmentation~\cite{docherty2024upsampling}. Concurrently, the zero-shot efforts have been able to enhance CLIP~\cite{zhang2024exploring}, mitigate misalignments~\cite{ge2024alignzeg}, and use prompts~\cite{li2024promerge, qu2024vcpclipvisualcontextprompting, sal2024eccv}, while also addressing domain adaptation~\cite{mata2025coptunsuperviseddomainadaptive, li2025towards}. Current works focus on scalability with high-res panoptic labels~\cite{hahn2025scene}, synthetic captions~\cite{lebailly2025synthetic}, and zero-shot matting~\cite{kim2025zim}, alongside new techniques like spectral tuning, quantization, and iterative refinement~\cite{xu2024spectral, kim2024expand, wang2025iterprime, huang2025zori, li2024infusing}. Despite this progress, these methods often lag supervised techniques and struggle with fine-grained or rare categories, limited by the semantic capacity of their underlying foundation models.


\subsection{Segmentation with diffusion models}

The rich, pixel-level representations from generative diffusion models~\cite{ho2020denoising,SohlDickstein2015DeepUL} are now widely repurposed for a variety of discriminative tasks~\cite{mukhopadhyay2024text, wang2025diffusion, zbinden2023stochastic, zhao2023unleashing}. These include depth prediction~\cite{ke2024repurposing,xu2025pixel}, super-resolution~\cite{chen2025adversarial,esmaeilzehi2025zfusion}, semantic correspondence~\cite{tang2023emergent,fundel2025distillation,zhang2023tale}, and segmentation~\cite{zhang2024three,peng2023diffusion,barsellotti2024training,wu2023diffumask,amit2021segdiff}. Much of these works rely on fine-tuning, adapting models for label-efficient~\cite{baranchuk2021label}, weakly-supervised~\cite{yoon2024diffusion}, or few-shot~\cite{zhu2024unleashing} learning, and tackling challenges from 3D point clouds~\cite{qu2025end, citation-0} to domain generalization and other applications~\cite{jia2024dginstyle, li2024exploring, xia2024unsupervised, park2025seediff, rahman2023ambiguous, wu2023datasetdm}. Several recent works focus on open-vocabulary segmentation~\cite{marcos2024open, karazija2024diffusion, kim2025seg4diff, zhu2024open, xu2023open}, zero-shot instance segmentation~\cite{ulmer2025conditional}, and zero-shot video segmentation~\cite{wang2024zero}.

Another line of work utilizes internal features of diffusion models for zero-shot segmentation of images. For instance, DiffSeg~\cite{tian2024diffuse} aggregates multi-scale self-attention maps from a Stable Diffusion U-Net, then applies iterative KL-divergence–based merging and non-maximum suppression to extract segmentation masks. Similarly, DiffCut~\cite{couairon2024diffcut} exclusively uses features from \replaced{the last layer of U-Net's encoder}{the final self-attention block of a diffusion U-Net} to build an affinity graph. It then employs recursive Normalized Cut to produce semantically coherent segments. While these methods prove to be effectively outperforming prior zero-shot methods, they both rely on either self-attention map or U-Net encoding features, with the former losing contextual representation and the latter limiting to low-resolution feature map. Additionally, both methods utilize a pre-defined timestep for feature extraction, leading to a suboptimal solution.

\section{Method}


We present a novel approach for high-quality segmentation that harnesses a pre-trained \replaced{diffusion}{Stable Diffusion} model~\cite{rombach2022latent}. Fig.~\ref{fig:graphical_abstract} shows the overview of the proposed approach. Our method is predicated on the key insight that a hierarchical semantic progression emerges throughout the model’s iterative denoising process. To capture this phenomenon, we introduce the concept of Contextual Similarity Map, which is used to identify a series of candidate timesteps representing distinct levels of semantic abstraction. This hierarchical information is then systematically aggregated to produce high-quality segmentations. \replaced{We elaborate on the details of our methodology in the subsequent sections.}{The subsequent sections will detail our methodology.}

\begin{figure}[t!]
    \centering
    \captionsetup{skip=2pt}
    
    \begin{subfigure}[t]{0.18\linewidth} 
        \centering
        \includegraphics[width=\linewidth]{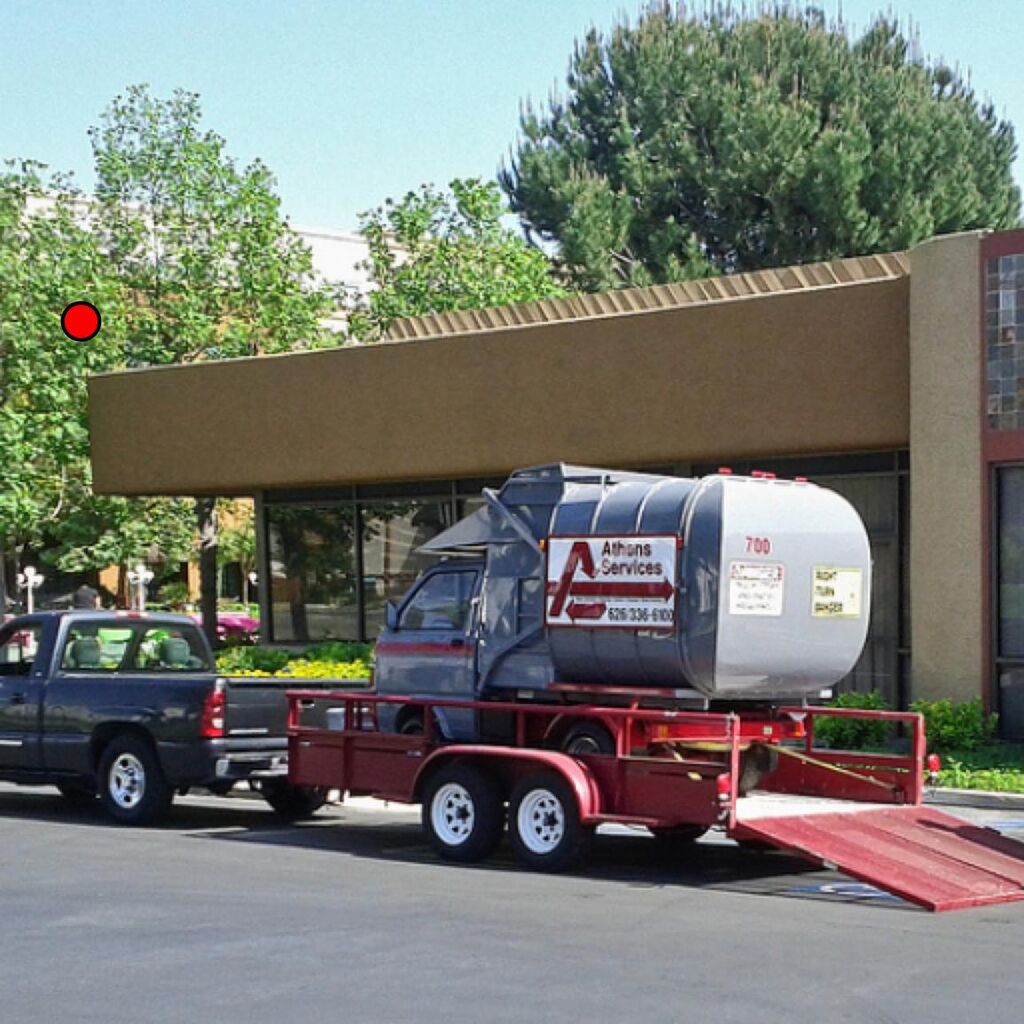}
        \caption{Image}
        \label{fig:comp_features_image}
    \end{subfigure}
    \hspace{5pt} 
    \begin{subfigure}[t]{0.18\linewidth}
        \centering
        \includegraphics[width=\linewidth]{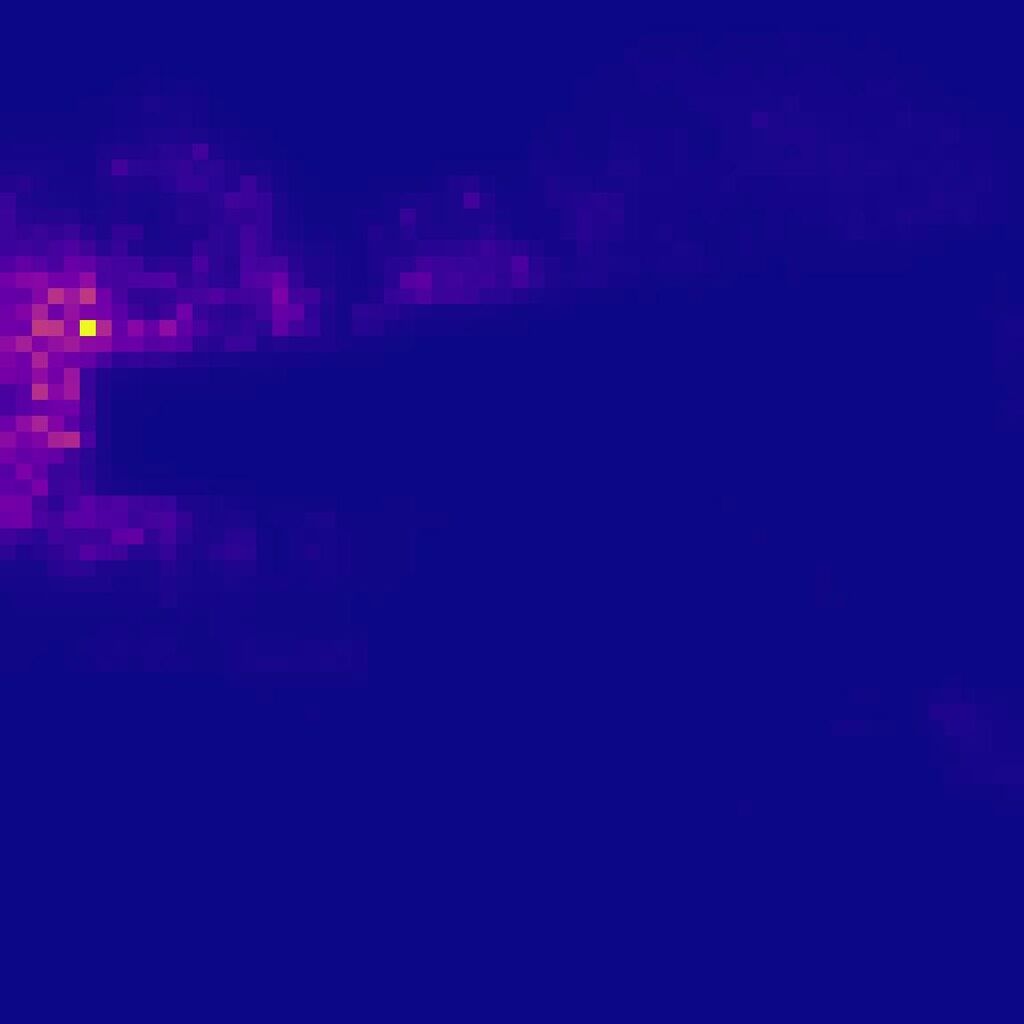}
        \caption{Attention}
        \label{fig:comp_features_attention}
    \end{subfigure}
    \hspace{5pt}
    \begin{subfigure}[t]{0.18\linewidth}
        \centering
        \includegraphics[width=\linewidth]{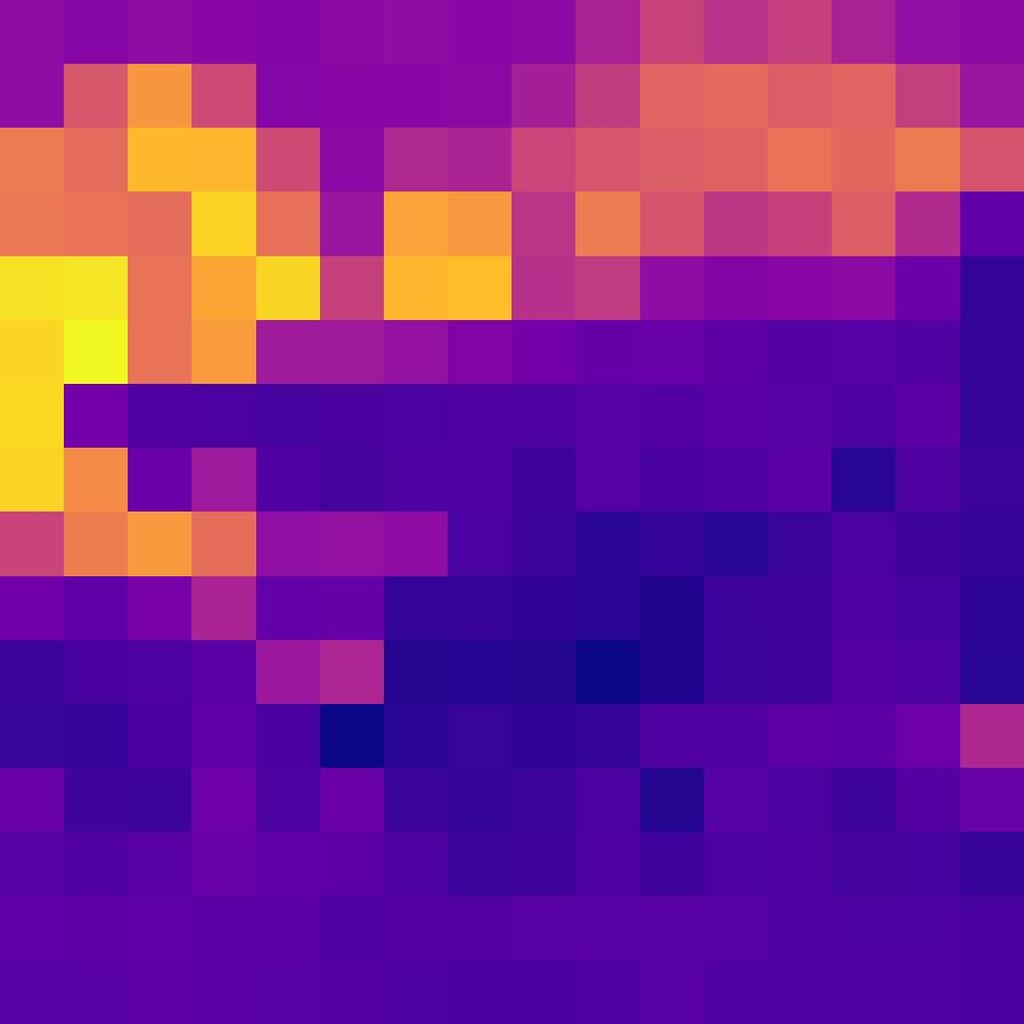}
        \caption{Feature}
        \label{fig:comp_features_feature}
    \end{subfigure}
    \hspace{5pt}
    \begin{subfigure}[t]{0.18\linewidth}
        \centering
        \includegraphics[width=\linewidth]{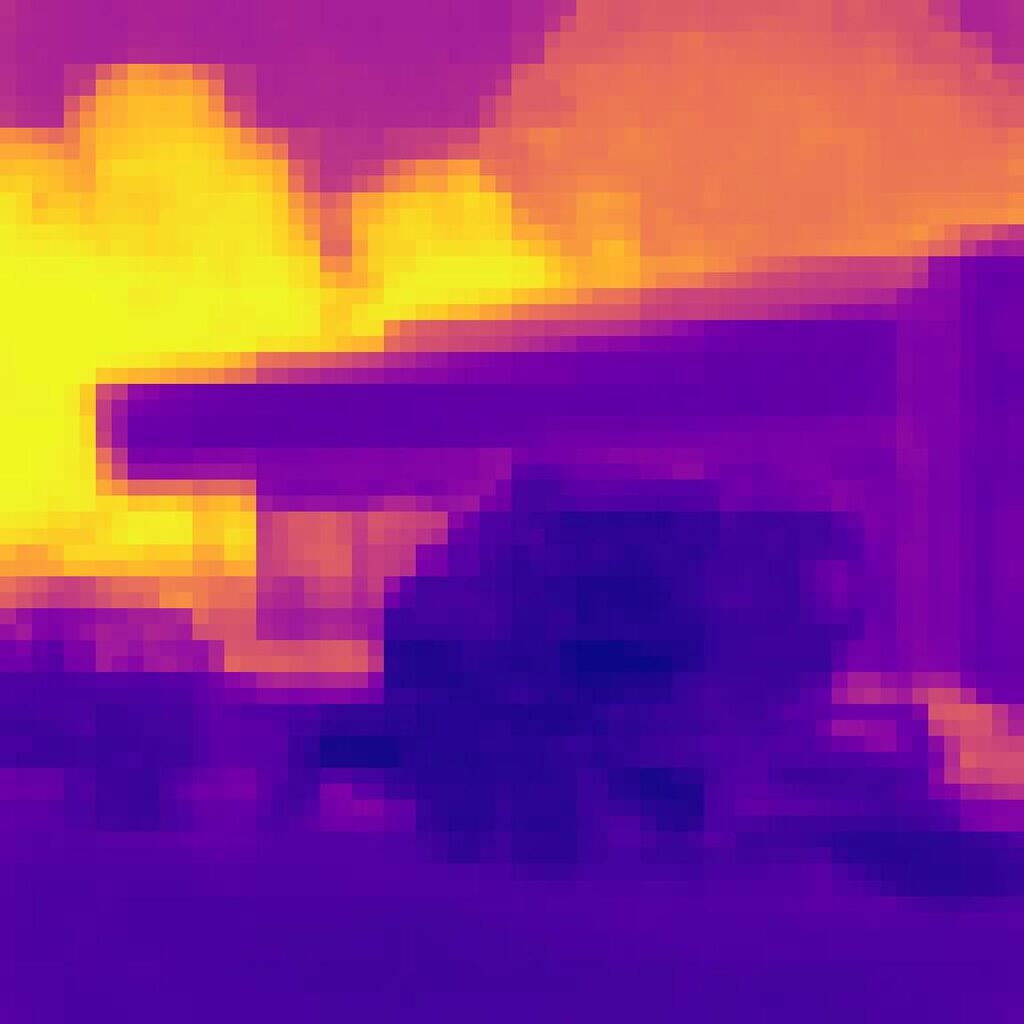}
        \caption{CSM}
        \label{fig:comp_features_similarity}
    \end{subfigure}

    \caption{Comparison between self-attention maps (b), encoder features (c), and the proposed Contextual Similarity Map (d) in determining semantic similarities between pixels. The red dot on image (a) denotes the query point. Bright and dark colors denote the high and low semantic similarities, respectively.}
    \label{fig:comp_features}
\end{figure}

\subsection{Diffusion model overview}

Stable Diffusion is a prominent Latent Diffusion Model (LDM) built upon the principles of Denoising Diffusion Models~\cite{ho2020denoising,song2020denoising}, performing an iterative denoising process within a latent space. This space is accessed via the encoder of a pre-trained variational autoencoder (VAE)~\cite{kingma2013auto} which encodes an input image $\mathbf{I} \in \mathbb{R}^{H \times W \times 3}$ into a compact latent representation $\mathbf{z_0} \in \mathbb{R}^{h \times w \times 4}$, where $h = H/8$ and $w = W/8$. 
For clarity in the following sections, we will refer to the pixels in the latent $\mathbf{z_0}$ simply as ``pixels,'' acknowledging that they represent \replaced{receptive}{perceptive} field regions in $\mathbf{I}$.

During inference, the Stable Diffusion model generates an image by an iterative Markov chain process. Starting from a pure noise latent $\mathbf{z_T} \sim \mathcal{N}(\mathbf{0}, \mathbf{1})$, the model progressively removes the noise to obtain $\mathbf{z_0}$. At each step $t \in [0, T]$, the model predicts the noise component $\epsilon_\theta(\mathbf{z_t}, t, \mathbf{c})$ in $\mathbf{z_t}$, where $\mathbf{c}$ is a condition signal. The predicted noise is then used to estimate $\mathbf{z}_{t-1}$. 
This iterative process continues until the final latent $\mathbf{z_0}$ is obtained, which is then passed through the VAE decoder to generate the high-resolution image.

The core of Stable Diffusion's denoising mechanism is a U-Net~\cite{ronneberger2015u} that estimates the noise component $\epsilon_\theta(\mathbf{z_t}, t, \mathbf{c})$. While this U-Net is primarily used for generating new images from noise, it implicitly captures a semantic understanding of the latent of any arbitrary input image. This key insight allows it to be repurposed for tasks beyond generation, such as segmentation~\cite{xu2023open,kawano2024maskdiffusion, tian2024diffuse}\added{, which is achieved by passing the noise-augmented latent representation of the image through the U-Net at a specific timestep $t$}. For instance, the self-attention maps collected from U-Net have been shown to indicate semantic dependencies within the image~\cite{tian2024diffuse}. On the other hand, DiffCut~\cite{couairon2024diffcut} demonstrates that features from the last layer of the U-Net's encoder can also be used for this purpose.


The aforementioned methods\added{, however,} have two key limitations. First, while self-attention maps provide high-resolution details, they lack rich contextual representations and highlight fine-grained structural similarities rather than the abstract semantic context (see Fig.~\ref{fig:comp_features_attention}). Conversely, U-Net encoder features offer robust contextual representations but at a low resolution (see Fig.~\ref{fig:comp_features_feature}). Second, both methods rely on a single, pre-defined timestep to extract features. \replaced{Nonetheless}{However}, our investigations reveal that a hierarchical semantic structure emerges for each pixel throughout the denoising process. This means that a single timestep is insufficient \added{for feature extraction} as each pixel achieves its optimal level of semantic abstraction at a different point in the process.

\begin{figure*}[htbp]
    \noindent
    \centering
    \begin{tabular}{@{}p{0.22\textwidth}@{\hfill}p{0.55\textwidth}@{\hfill}p{0.22\textwidth}@{}}
        \centering
        \begin{minipage}{\linewidth}
            \centering
            \begin{subfigure}[b]{\linewidth}
                \includegraphics[width=0.95\linewidth, keepaspectratio=true]{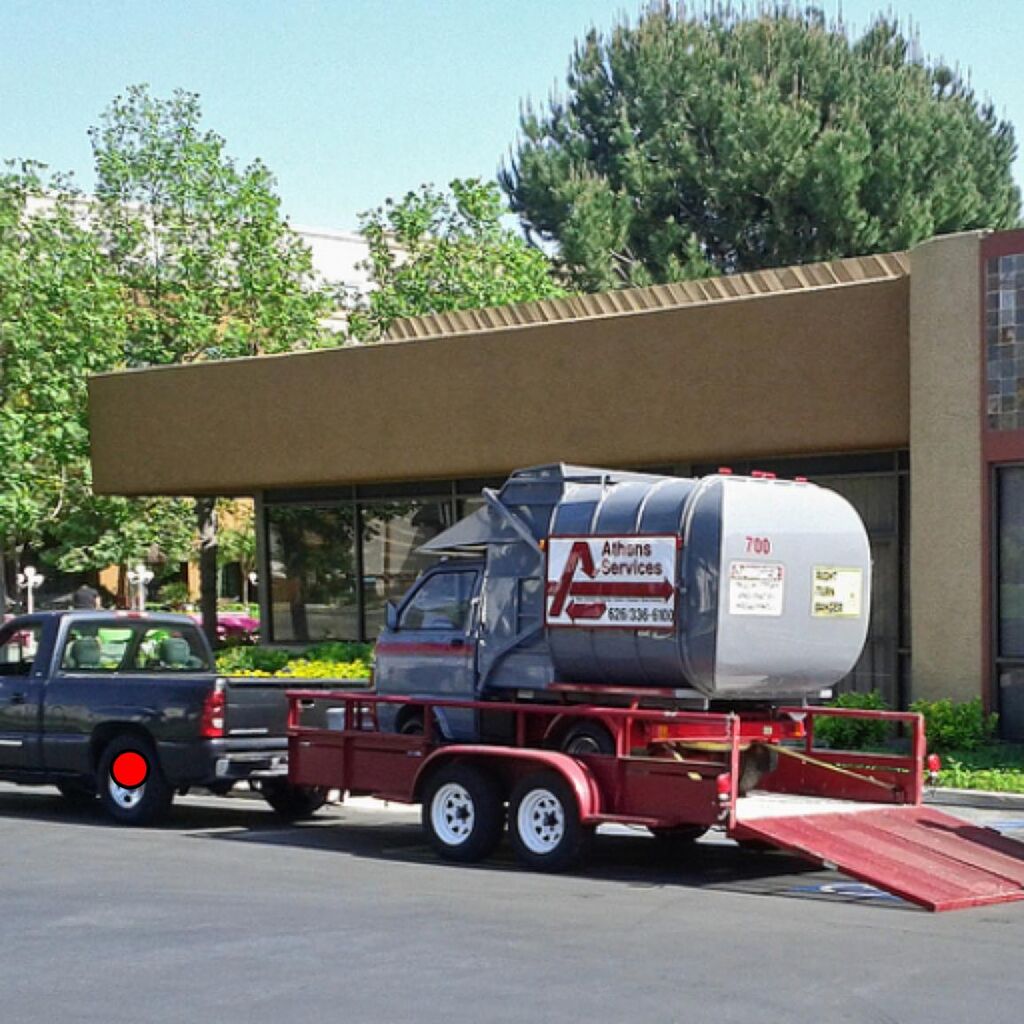}
                \caption{Input Image}
                \label{fig:hierarchical_progress_input_image}
            \end{subfigure}
        \end{minipage} &
        
        \begin{minipage}{\linewidth}
            \centering
            \begin{subfigure}[b]{0.24\linewidth}
                \centering
                \includegraphics[width=0.95\linewidth, keepaspectratio=true]{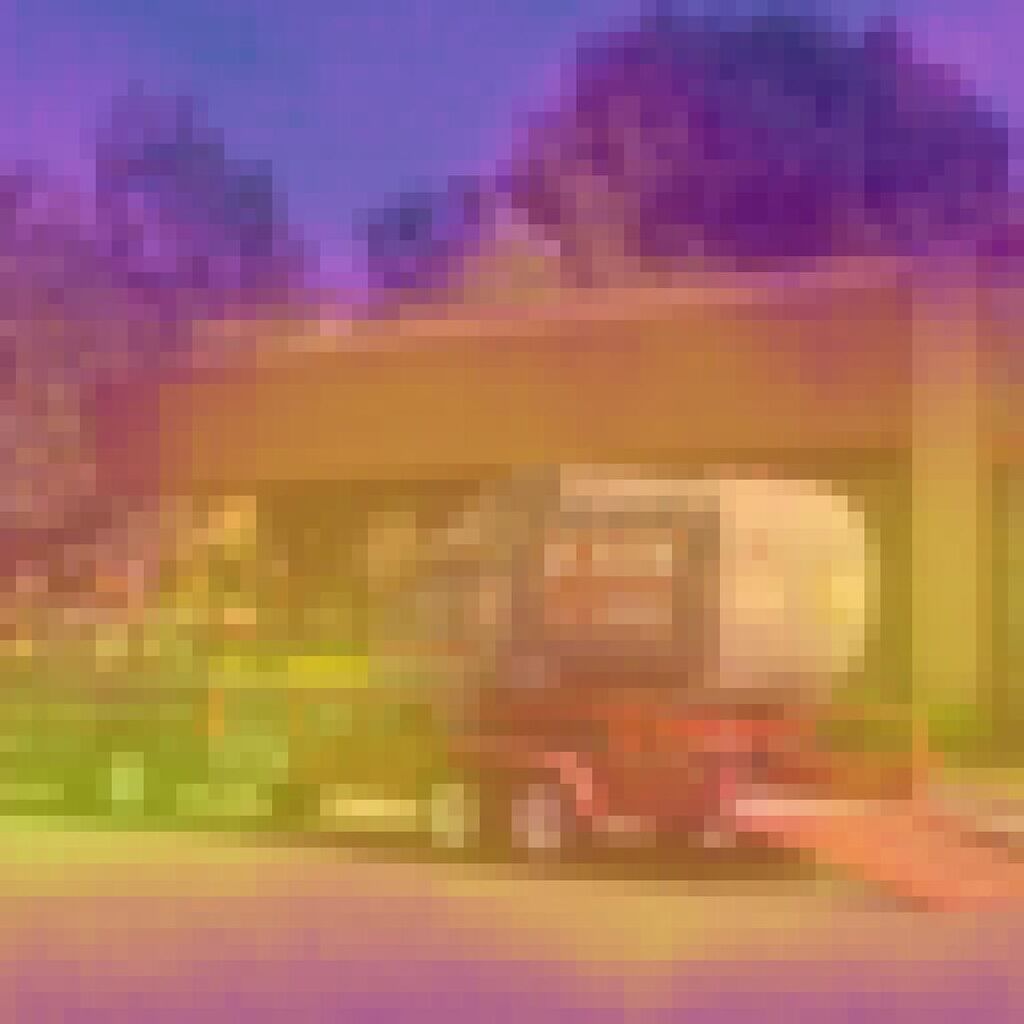}
                \caption{$t=800$}
                \label{fig:hierarchical_progress_timestep_800}
            \end{subfigure}\hfill
            \begin{subfigure}[b]{0.24\linewidth}
                \centering
                \includegraphics[width=0.95\linewidth, keepaspectratio=true]{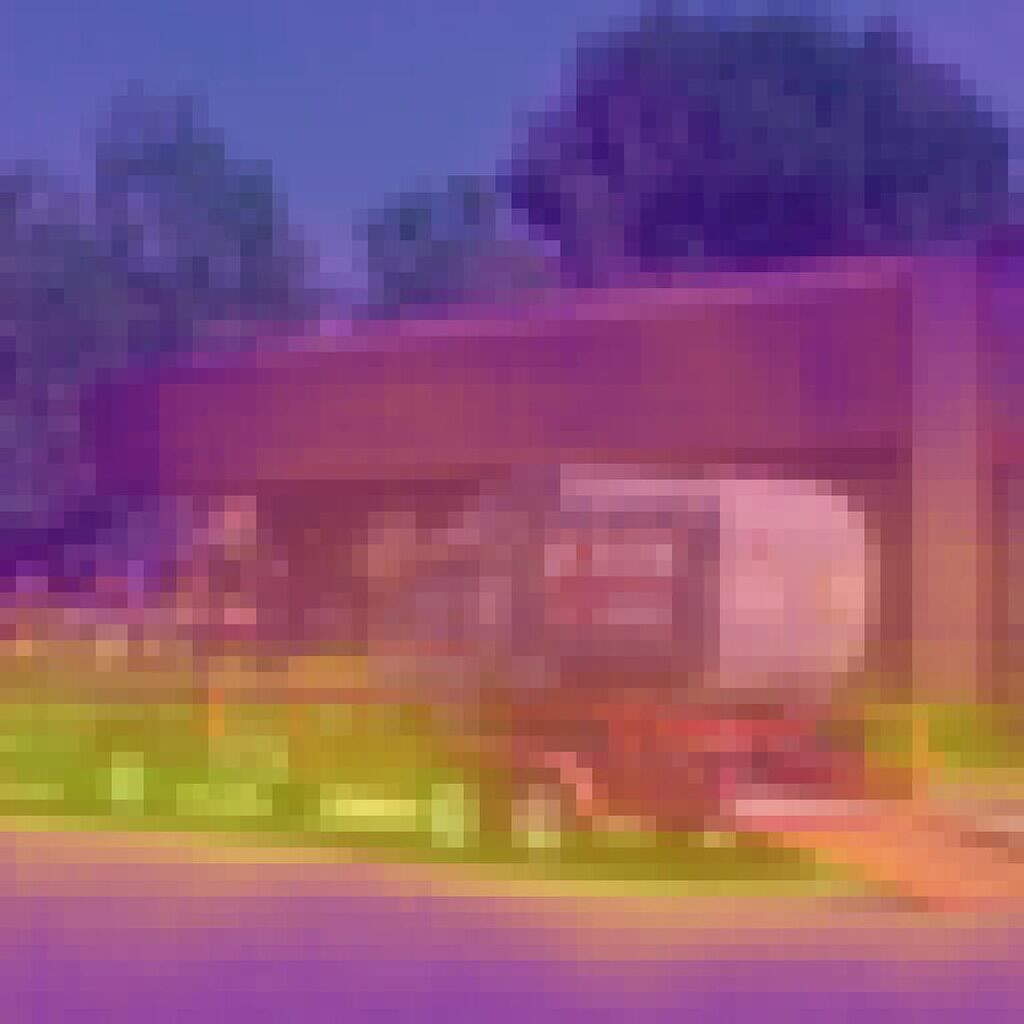}
                \caption{$t=700$}
                \label{fig:hierarchical_progress_timestep_700}
            \end{subfigure}\hfill
            \begin{subfigure}[b]{0.24\linewidth}
                \centering
                \includegraphics[width=0.95\linewidth, keepaspectratio=true]{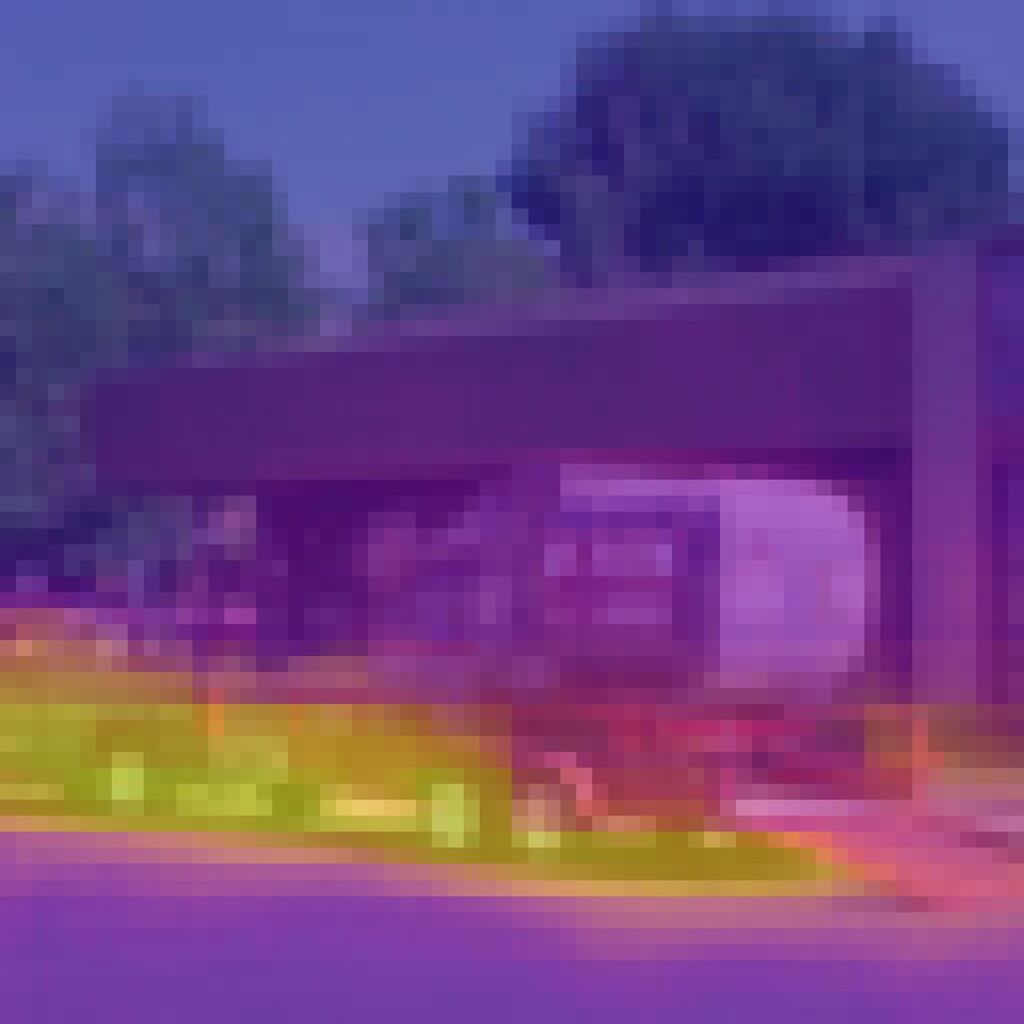}
                \caption{$t=600$}
                \label{fig:hierarchical_progress_timestep_600}
            \end{subfigure}\hfill
            \begin{subfigure}[b]{0.24\linewidth}
                \centering
                \includegraphics[width=0.95\linewidth, keepaspectratio=true]{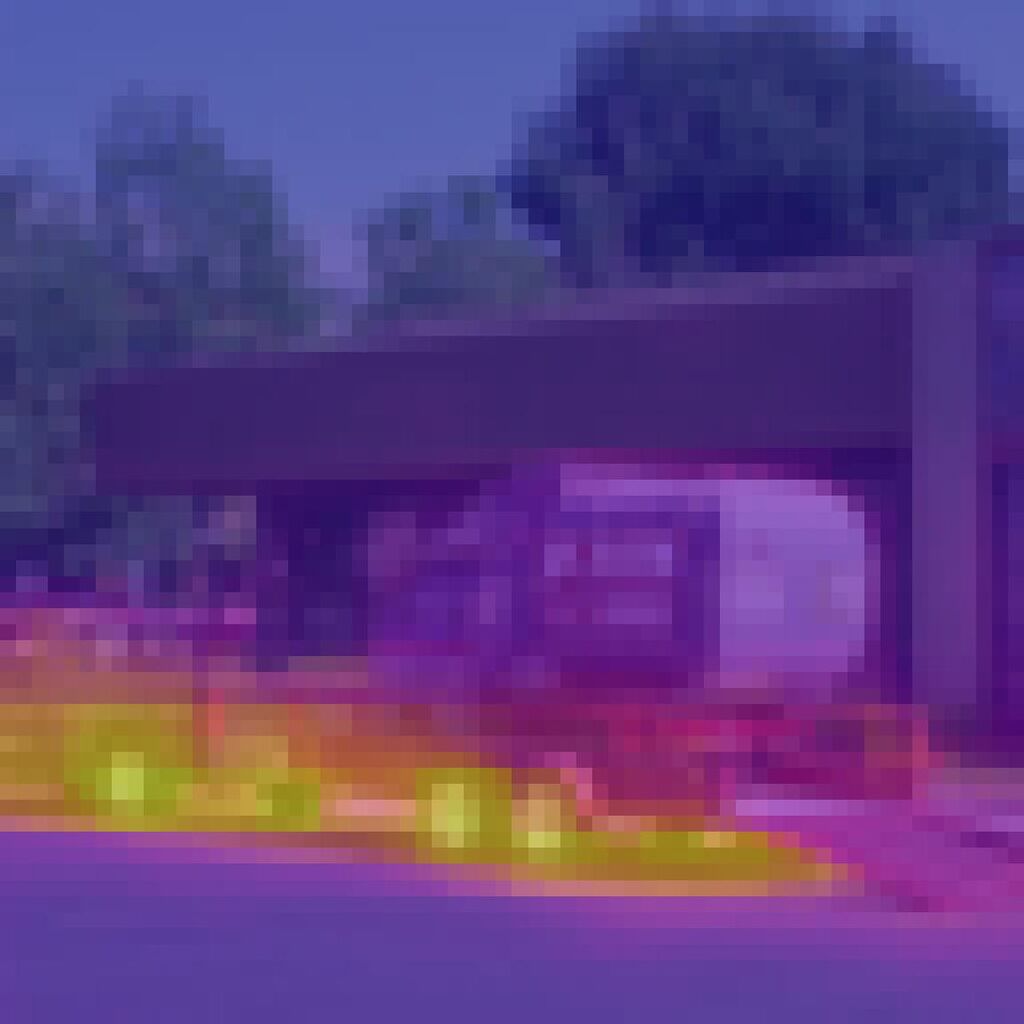}
                \caption{$t=500$}
                \label{fig:hierarchical_progress_timestep_500}
            \end{subfigure}
            \vspace{0.5em}
            \begin{subfigure}[b]{0.24\linewidth}
                \includegraphics[width=0.95\linewidth, keepaspectratio=true]{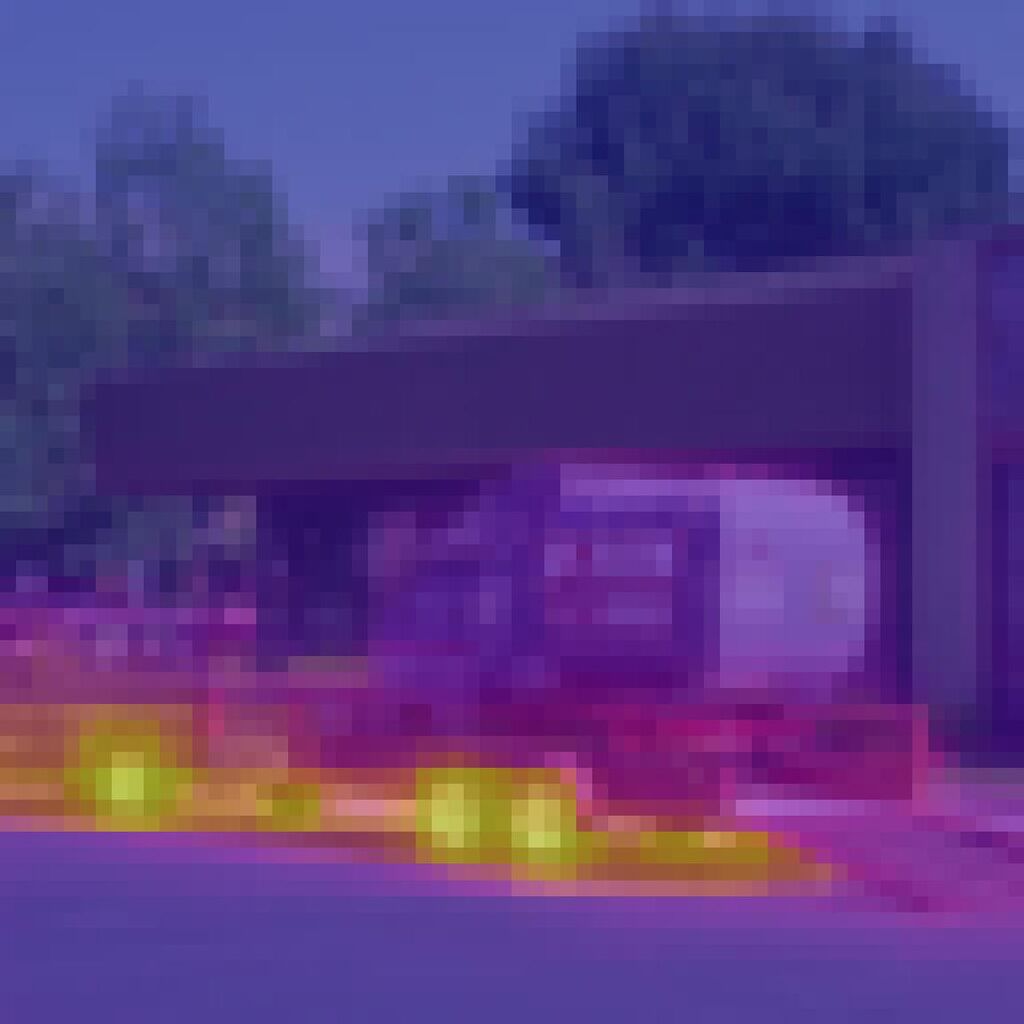}
                \caption{$t=400$}
                \label{fig:hierarchical_progress_timestep_400}
            \end{subfigure}\hfill
            \begin{subfigure}[b]{0.24\linewidth}
                \includegraphics[width=0.95\linewidth, keepaspectratio=true]{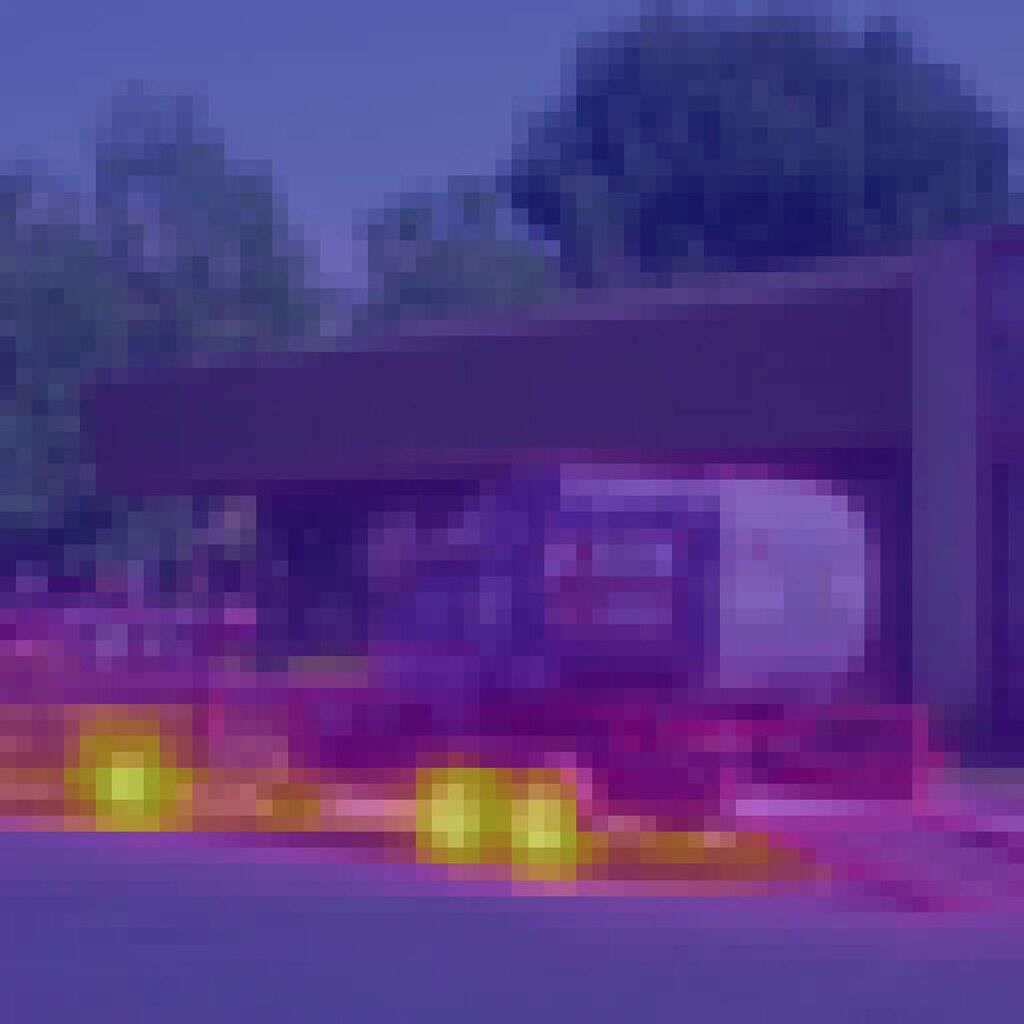}
                \caption{$t=300$}
                \label{fig:hierarchical_progress_timestep_300}
            \end{subfigure}\hfill
            \begin{subfigure}{0.24\linewidth}
                \includegraphics[width=0.95\linewidth, keepaspectratio=true]{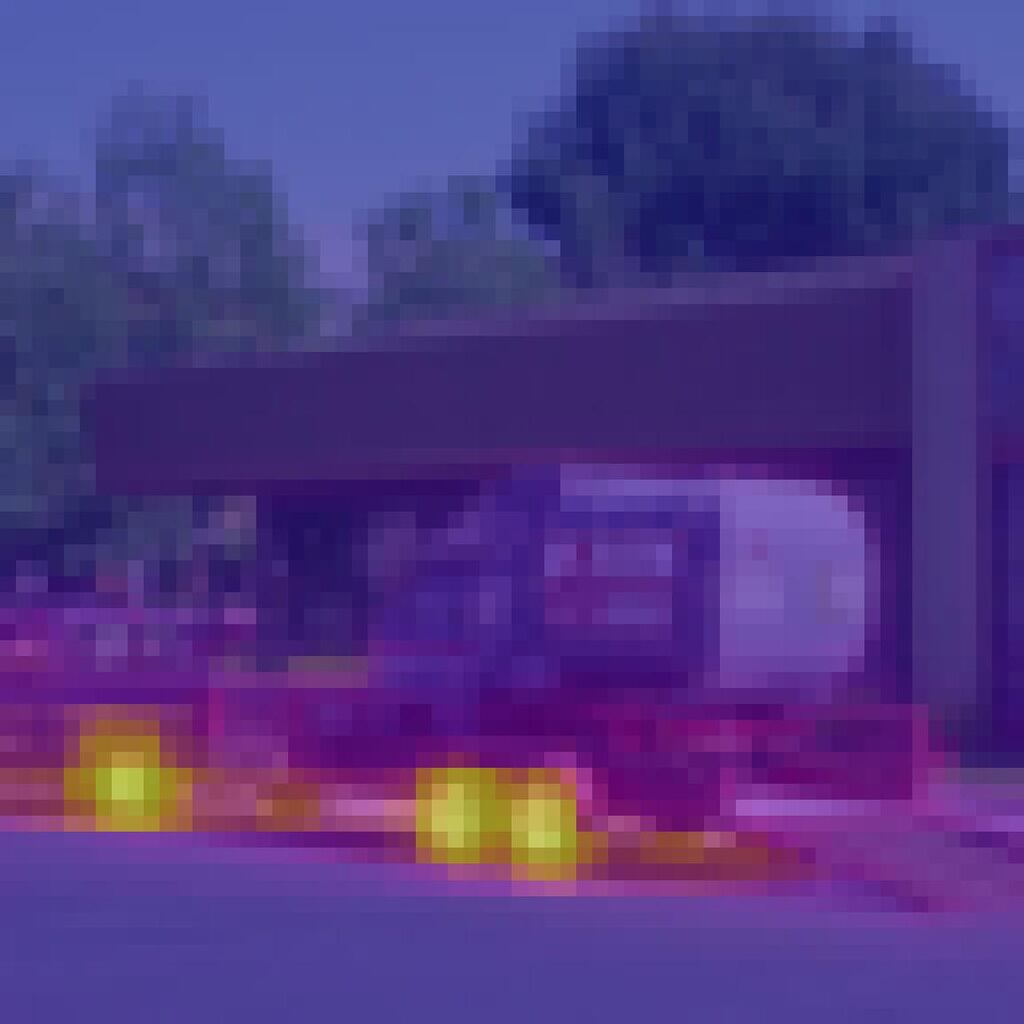}
                \caption{$t=200$}
                \label{fig:hierarchical_progress_timestep_200}
            \end{subfigure}\hfill
            \begin{subfigure}{0.24\linewidth}
                \includegraphics[width=0.95\linewidth, keepaspectratio=true]{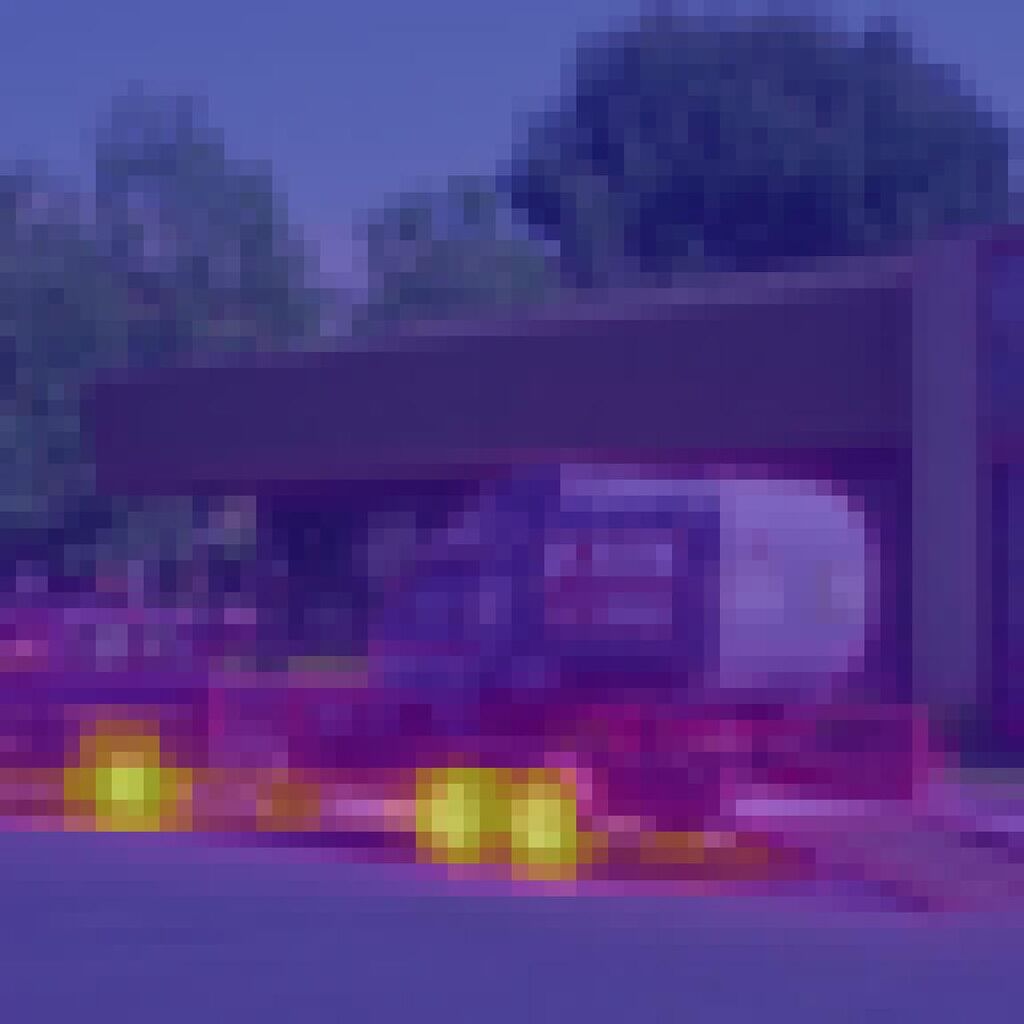}
                \caption{$t=100$}
                \label{fig:hierarchical_progress_timestep_100}
            \end{subfigure}
        \end{minipage} &
        
        \begin{minipage}{\linewidth}
            \centering
            \begin{subfigure}[b]{\linewidth}
                \includegraphics[width=0.95\linewidth, keepaspectratio=true]{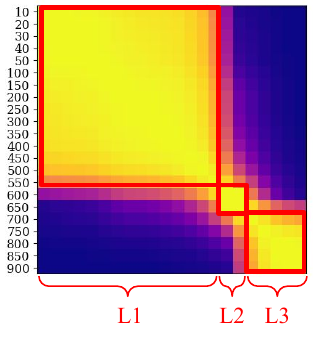}
                \caption{TSM}
                \label{fig:hierarchical_progress_consistency_matrix}
            \end{subfigure}
        \end{minipage}

    \end{tabular}
    \caption{
    Contextual Similarity Maps (CSMs) reveal the hierarchical progression of semantic information across different denoising timesteps. (a) Input image with a query pixel (red). (b-i) CSMs across timesteps: higher timesteps capture high-level semantics, while lower timesteps resolve into detailed regions. (j) The TSM for the query pixel reveals three distinct blocks representing different levels of semantic abstraction across time.}
    \label{fig:hierarchical_progress}
\end{figure*}

\subsection{Contextual similarity map}
\label{sec:contextual_similarity_map}

To address the above limitations, we propose a Contextual Similarity Map (CSM) combining U-Net's self-attention and features to create a more comprehensive semantic context. CSM is a per-pixel representation that harmoniously blends the high-resolution power of self-attention with the deep semantic understanding of U-Net encoder features.

To construct the CSM, we follow a two-stage process. For any given timestep $t$, we first aggregate the multi-resolution self-attention maps across different layers of the U-Net into a unified matrix, $\mathbf{M_t} \in \mathbb{R}^{h \cdot w \times h \cdot w}$, where each entry $i, j$ represents the aggregated self-attention value from pixel $i$ to pixel $j$. Concurrently, we extract the U-Net's semantically rich but low-resolution encoder features and upsample them to match the latent dimensions, forming a feature matrix $\mathbf{F_t} \in \mathbb{R}^{h \cdot w \times 1280}$. Next, we leverage the high-resolution relationships encoded in the self-attention matrix $\mathbf{M_t}$ to guide a weighted aggregation of the upsampled encoder features, as shown in the following equation:
\begin{equation}
\label{eq:contextual_vector}
\mathbf{V_t} = \mathbf{M_t} \times \mathbf{F_t},
\end{equation}
where row $i$ in $\mathbf{V_t}$, denoted as $\mathbf{v_i}$, holds the combined feature representation for pixel $i$. The contextual similarity map at timestep $t$, $\mathbf{S_t} \in \mathbb{R}^{h \cdot w \times h \cdot w}$, is then defined as the pairwise cosine similarity between these feature representations. Formally, each element $(i, j)$ of $\mathbf{S_t}$ is computed as:

\begin{equation}
\mathbf{S_t}(i,j) = \frac{\mathbf{v_i} \cdot \mathbf{v_j}}{\|\mathbf{v_i}\|_2 \|\mathbf{v_j}\|_2}.
\end{equation}



Fig.~\ref{fig:comp_features_similarity} shows the comparison of CSM with its constituent parts, illustrating that it effectively resolves the contextual limitations of self-attention maps and the resolution constraints of U-Net features.

\subsection{Hierarchical concepts}
\label{sec:hierarchical_concepts}

Our investigation on the CSM reveals the hierarchical progression of semantic information across different denoising timesteps $t$. This is a natural consequence of the iterative denoising process, which refines an image's representation from coarse (global structures) to fine-grained (local details).

Fig.~\ref{fig:hierarchical_progress} illustrates an example of this property. At high denoising timesteps (like in Fig.~\ref{fig:hierarchical_progress_timestep_800}), the CSM prioritizes broad structural elements. Precisely, given a query pixel on the truck's tire, CSM highlights major components such as the truck and the surrounding building, reflecting a global understanding of the scene. As denoising progresses to lower timesteps (like in Fig.~\ref{fig:hierarchical_progress_timestep_100}), the CSM's focus sharpens on fine-grained semantic details such as the individual wheels of the truck, highlighting regions that are closely related to the query pixel in detailed semantics. 

The shift in the semantic hierarchy effectively reveals distinct ``semantic levels", each representing a different degree of abstraction. We define a ``semantic boundary" as the timestep where the similarity map transitions from one level of abstraction to another.

Crucially, this observed semantic hierarchy is a per-pixel property, meaning that each pixel has its own unique semantic levels and boundaries. This is an important observation as even pixels with similar semantics can exhibit distinct patterns (please see Supp.~\cref{fig:supp_hierarchical_progress_sample1,fig:supp_hierarchical_progress_sample3,fig:supp_hierarchical_progress_sample5,fig:supp_hierarchical_progress_sample7,fig:supp_hierarchical_progress_sample14} for more examples).

\subsection{Per-pixel semantic boundary detection}
\label{sec:semantic_boundary_detection}

As part of our observations, we noticed that within a single semantic level, CSM tends to remain stable across different timesteps (see Fig.~\ref{fig:hierarchical_progress} and Supp.~\cref{fig:supp_hierarchical_progress_sample1,fig:supp_hierarchical_progress_sample3,fig:supp_hierarchical_progress_sample5,fig:supp_hierarchical_progress_sample7,fig:supp_hierarchical_progress_sample14}). This insight led to a central hypothesis: semantic boundaries can be identified as the points in time when a pixel's CSM undergoes a significant change. To formalize this, we introduce the concept of Temporal Stability Matrix (TSM), representing the stability of the CSM across different denoising timesteps for each pixel.

For a given pixel $k$, let $\mathbf{S_t}(k) \in \mathbb{R}^{hw}$ be the $k$-th row of the CSM at timestep $t$, representing the similarities between pixel $k$ and all the other pixels in the image at timestep $t$. The TSM \replaced{for pixel}{in} $k$ is defined as the pairwise cosine similarity between the pixel's CSMs for every possible pair of timesteps. Precisely, the TSM value between $t_1$ and $t_2$ is defined as:

\begin{equation}
\mathbf{C_k}(t_1, t_2) = \frac{\mathbf{S_{t_1}}(k) \cdot \mathbf{S_{t_2}}(k)}{\|\mathbf{S_{t_1}}(k)\| \|\mathbf{S_{t_2}}(k)\|}.
\end{equation}
A high value of $\mathbf{C_k}(t_1, t_2)$ indicates a stable semantic representation for pixel \replaced{$k$}{$p_k$} between timesteps $t_1$ and $t_2$. Conversely, a low value signals a divergence in the CSM pattern, marking a potential semantic change.

Fig.~\ref{fig:hierarchical_progress_consistency_matrix} shows an example of TSM for the query pixel. This TSM reveals distinct, high-similarity blocks of timesteps, with each block corresponding to a particular semantic level. For instance, an early block captures fine-grained details, such as the tires of a truck (L1), while the subsequent block represents the broader concept of the entire truck (L2), and the third block encompasses the foreground objects (L3). The boundaries between these blocks visualize the moments when the model transitions its focus from a semantic level to the other (i.e., the semantic \replaced{boundaries}{boundries}).

Given this observation, the task of detecting semantic levels for each pixel now translates to identifying these high-similarity blocks within the TSM. We approach this task as a graph partitioning problem by treating the TSM as a weighted adjacency matrix, where each timestep is a node. We then leverage techniques from spectral graph theory to analyze the connectivity of this graph.

For each pixel $k$, we construct the corresponding graph Laplacian matrix, $\mathbf{L}_k = \mathbf{D}_k - \mathbf{C}_k$, where $\mathbf{D}_k(i, i) = \sum_j \mathbf{C}_k (i, j)$ is the degree matrix. 
The Fiedler vector~\cite{fiedler1989laplacian}, defined as the eigenvector corresponding to the second smallest eigenvalue of $\mathbf{L}_k$, provides an optimal one-dimensional embedding that captures the graph's connectivity. In our context, sudden changes in the values of the Fiedler vector indicate the semantic boundaries.

To precisely pinpoint these transitioning timesteps for each pixel, we apply the Pruned Exact Linear Time (PELT) algorithm~\cite{killick2012optimal} to the Fiedler vector of the corresponding TSM. 
The detected change points serve as candidate timesteps, allowing us to selectively extract features that represent the model's semantic understanding at each level for a given pixel.

\subsection{Segmentation pipeline}

Our segmentation pipeline begins by generating CSMs for a range of timesteps for each pixel of a given input image, as detailed in Sec.~\ref{sec:contextual_similarity_map}. We then apply our semantic boundary detection algorithm (Sec.~\ref{sec:semantic_boundary_detection}) to identify the semantic levels for each pixel.

Next, a semantic level is chosen for each pixel independently as the target level for segmentation. We found that the first semantic level typically captures highly localized details, whereas the second level encapsulates higher-level semantic information, making it an ideal candidate for our use case. This observation is empirically supported by the visualizations in Fig.~\ref{fig:hierarchical_progress} and experimentally evaluated in Sec.~\ref{sec:semantic_level_distribution}.
We refer to the timestep corresponding to the second semantic boundary as the target timestep, and the second semantic level as the target level, respectively.

\replaced{Low}{Early} timesteps provide highly localized CSMs, while \replaced{later}{higher} ones contain broader semantic information at the cost of granularity. To address this trade-off, we employ a two-phase aggregation process. First, for each pixel in the latent space, we create an intermediate representation by averaging its CSMs across all timesteps up to target timestep. This phase ensures that the intermediate representation is robustly anchored in low-level and granular details. 
\replaced{We construct a composite matrix of $\bar{\mathbf{S}} \in \mathbb{R}^{h \cdot w \times h \cdot w}$, where each row $j$ represents the averaged CSM for pixel $j$ up to its target timestep $t^{*}_j$, denoted as $\bar{\mathbf{S}}(j) = \text{mean}\{\mathbf{S}_t(j) \mid t < t^{*}_j\}$.}
{If $t^{*}_k$ is the target timestep for pixel $k$, this averaged CSM is denoted as $\bar{\mathbf{S}} = \text{mean}\{\mathbf{S_t}(k) | t < t^{*}_k\}$.}

Second, the final segmentation is computed by a weighted averaging of \replaced{$\bar{\mathbf{S}}$}{the CSMs} for each pixel $k$, where the weights are derived from the normalized CSM \replaced{for pixel}{in} $k$ at the target timestep $t^{*}_k$. This two-phase aggregation is crucial for creating a comprehensive feature vector that effectively combines granular details with abstract semantics from the target level. Formally, this is expressed by the equation:
\begin{equation}
\label{eq:hierarchical_aggregation}
\mathbf{Q}_{k} = \text{softmax}\left(\frac{\mathbf{S}_{t_k^*}(k)}{\tau_s}\right) \times \bar{\mathbf{S}}.
\end{equation}
Here, \deleted{$\bar{\mathbf{S}} \in \mathbb{R}^{h \cdot w \times h \cdot w}$ is the averaged CSM  across timesteps up to $t^{*}_k$.} $\mathbf{S}_{t_k^*}(k) \in \mathbb{R}^{1 \times h \cdot w}$ is the CSM at $t^{*}_k$ for pixel $k$, and $\tau_s$ is a temperature parameter that controls the sharpness of the softmax distribution. Once the feature vector $\mathbf{Q}_{k}$ is computed\added{ for every pixel}, they are passed to a recursive Normalized Cut (Ncut) algorithm~\cite{shi2000normalized} to generate the final segmentation mask.

\section{Results}

\textbf{Datasets.} We \replaced{utilized}{evaluated our method on} six semantic segmentation datasets\added{ for our experiments}: Pascal VOC~\cite{everingham2010pascal} (20 foreground classes and a background class), Pascal Context~\cite{mottaghi2014role} (59 foreground classes and a background class), COCO-Object~\cite{lin2014microsoft} (80 foreground classes and a background class), COCO-Stuff-27 (27 foreground classes), Cityscapes~\cite{cordts2016cityscapes} (27 foreground classes), and ADE20K~\cite{zhou2017scene} (150 foreground classes). We tested our method and the baselines on the validation sets of these datasets. For COCO, we utilized a specific validation split that has been used by previous research~\cite{tian2024diffuse,couairon2024diffcut}.

\textbf{Metrics.} Our primary evaluation metric is the mean Intersection over Union (mIoU). Due to the unsupervised nature of our method and the lack of semantic labels, we performed a Hungarian Matching~\cite{kuhn1955hungarian} with the ground truth labels prior to the mIoU calculation~\cite{couairon2024diffcut, tian2024diffuse}. For the datasets with a background class, the unmatched classes were all assigned to the background class.


\textbf{Implementation details.} Please see Supp. Sec.~\ref{sec:supp_implementation_details} for implementation details.

\subsection{Empirical evidence of hierarchical semantic progression}
\label{sec:results_empirical_evidence_of_hierarchical}


To quantitatively assess the progression of hierarchical semantics during the diffusion process, we conducted a series of experiments across various \replaced{diffusion}{Stable Diffusion} backbones using the Pascal VOC dataset. This dataset is particularly suited for this analysis as it provides annotations at two distinct granularities: 1) object-level, capturing the entire object (e.g., car, human), and 2) part-level, isolating specific parts (e.g., door, tire, hood). At each timestep, we extracted the attention maps for individual pixels, computed the soft-IoU against their corresponding ground-truth annotation masks, averaged these scores across the entire image (see Supp. Sec.~\ref{sec:supp_object_vs_part} for details). As illustrated in Fig.~\ref{fig:hierarchy_evolution}, increasing the timestep consistently leads to the emergence of higher-level semantics that align more closely with object-level annotations, transitioning away from localized part-level details, while reducing the timestep results in higher soft-IoU in part-level segmentation. This uniform trend across all evaluated backbones provides concrete, quantitative evidence of semantic information progression along the denoising trajectory of diffusion models.

\subsection{Zero-shot segmentation}
\label{sec:results_zero_shot_segmentation}

\begin{table*}[h!]
\centering
\small
\caption{Zero-shot semantic segmentation results (mIoU). Best method in \textbf{bold}, second is \underline{underlined}. Results of non-diffusion methods were taken from~\cite{tian2024diffuse,couairon2024diffcut,cha2023learning,kim2024eagle}
, while those using diffusion-based methods were recomputed using their official codebase with SDv1.4 as backbone and $512\times512$ as input resolution.
}

\label{tab:segmentation_miou}
\resizebox{\textwidth}{!}{
\begin{tabular}{lcccccc ccc}
\toprule
\textbf{Method} & \textbf{LD} & \textbf{AX} & \textbf{UA} & \textbf{COCO-Stuff} & \textbf{COCO-Object}& \textbf{CityScapes} & \textbf{ADE20K} & \textbf{VOC} & \textbf{Context} \\
\midrule
IIC \cite{ji2019invariant} &  &  & \checkmark & 6.7 & - & 6.4 & - & 9.8 & - \\
MDC \cite{caron2018deep} &  &  & \checkmark & 9.8 & - & 7.1 & - & - & - \\
PiCIE \cite{cho2021picie} & &  & \checkmark & 13.8 & - & 12.3 & - & - & - \\
PiCIE+H \cite{cho2021picie} & & & \checkmark & 14.4 & - & - & - & - & - \\
CuVLER \cite{arica2024cuvler} & & & \checkmark & 30.0 & 16.0 & 3.2 & 13.3 & 19.4 & 25.6 \\
STEGO \cite{Hamilton2022UnsupervisedSS} & \checkmark & & \checkmark & 28.2 & - & 21.0 & - & - & - \\
ACSeg \cite{li2023acseg} & \checkmark & & \checkmark & 28.1 & - & - & - & \underline{53.9} & - \\
Eagle \cite{kim2024eagle} & & & \checkmark & 27.2 & - & 22.1 & - & - & - \\
ReCO \cite{shin2022reco} & \checkmark & \checkmark & & 26.3 & 15.7 & 19.3 & 11.2 & 25.1 & 19.9 \\
MaskCLIP \cite{zhou2022extract} & \checkmark & & & 19.6 & 20.6 & 10.0 & 9.8 & 38.8 & 23.6 \\
MaskCut \cite{simeoni2023unsupervised} &  & &  & 41.7 & 30.1 & 18.7 & \underline{35.7} & 53.8 & 43.4 \\
ProMerge \cite{li2024promerge} & & & & 35.6 & 30.2 & 11.8 & 23.0 & 35.4 & 31.8 \\
DINOv3 + Recursive-NCut \cite{simeoni2025cijo} & & & & 40.2 & 27.5 & 9.0 & 16.4 & 38.3 & 37.2 \\
DiffSeg \cite{tian2024diffuse} & & & & \underline{43.7} & - & \underline{22.2} & 35.0 & 47.6 & \underline{47.6} \\
DiffCut \cite{couairon2024diffcut} & & & & 19.7 & \textbf{31.2} & 13.4 & 21.1 & 36.0 & 40.8 \\
\midrule
Ours & & & & \textbf{45.0} & \underline{30.8} & \textbf{27.8} & \textbf{45.3} & \textbf{58.2} & \textbf{55.4} \\
\bottomrule
\end{tabular}
}
\end{table*}


To assess the zero-shot performance of our approach, we compared it against several state-of-the-art methods, including IIC~\cite{ji2019invariant}, MDC~\cite{caron2018deep}, PiCIE and PiCIE+H~\cite{cho2021picie}, STEGO~\cite{Hamilton2022UnsupervisedSS}, ACSeg~\cite{li2023acseg}, ReCO~\cite{shin2022reco}, MaskCLIP~\cite{zhou2022extract}, MaskCut~\cite{simeoni2023unsupervised}, CuVLER~\cite{arica2024cuvler}, ProMerge~\cite{li2024promerge}, DiffSeg~\cite{tian2024diffuse}, DiffCut~\cite{couairon2024diffcut}, and DINOv3~\cite{simeoni2025cijo} with Recursive-NCut. Some \added{of these} methods required additional dependencies: unsupervised adaptation (UA), language dependency (LD), and auxiliary image requirement (AX). UA indicates that the method required unsupervised training; methods without this requirement are considered zero-shot. LD means the method relied on textual input to aid segmentation. AX refers to the use of an additional set of references or synthetic images.

Quantitative results are summarized in Tab.~\ref{tab:segmentation_miou}. 
To ensure a fair comparison, results from DiffSeg and DiffCut were recomputed using SDv1.4 and input resolution $512 \times 512$. \deleted{See supplementary material (}Supp. Sec.~\ref{sec:supp_backbone_comparison}\deleted{) for} \added{also shows} \deleted{an} evaluation on \deleted{the effects of the} different backbones \added{with various architectures (e.g., DiT, flow-based)}\deleted{on the segmentation performance}.
As evident from Tab.~\ref{tab:segmentation_miou}, our method consistently outperforms its counterparts across nearly all evaluated datasets. Specifically, our method obtained the highest mIoU on COCO-Stuff, CityScapes, ADE20K, VOC, and the Context dataset. While DiffCut showed competitive results on COCO-Object (31.2\% vs. our 30.8\%), our method demonstrated superior overall robustness and adaptability across a broader spectrum of benchmarks. This strong performance highlighted the effectiveness of our approach in leveraging more comprehensive semantic understanding through CSM and hierarchical semantic levels.



\textbf{Qualitative comparison:} Fig.~\ref{fig:baseline_qualitative_comparison} shows some segmentations generated by the proposed approach compared against DiffSeg and DiffCut. These examples demonstrate that our approach coherently identifies semantics across different object types and environments (e.g., object-centric or scene-based) and is able to segment small objects in scenes. More examples can be found in the supplementary material (see Supp.~\cref{fig:gallery,fig:baseline_qualitative_comparison_1,fig:baseline_qualitative_comparison_2,fig:baseline_qualitative_comparison_3}). 

\subsection{Ablation studies}
\label{sec:results_ablation_studies}

\begin{figure}[t]
    \centering
    \begin{minipage}[t]{0.55\textwidth}
        \vspace{0pt} 
        \centering
        \includegraphics[width=\linewidth]{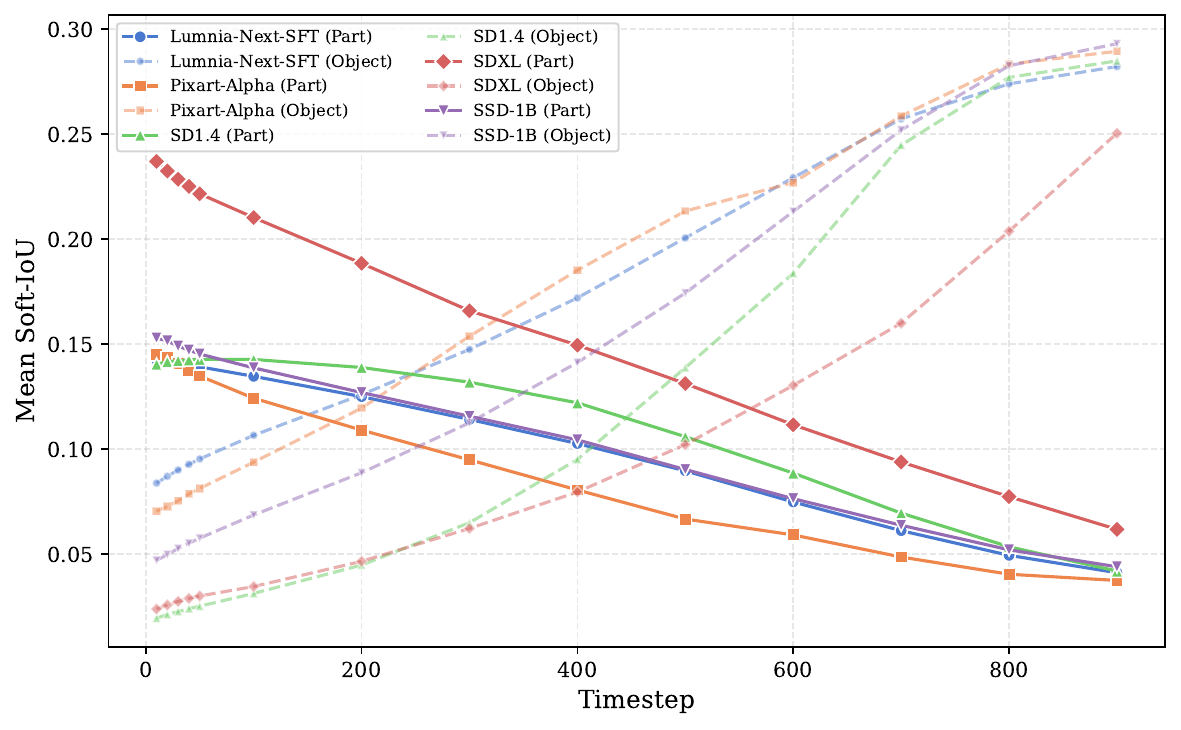}
        \caption{Hierarchical semantic evolution in diffusion trajectory. Mean Soft-IoU across various diffusion backbones shows that higher timesteps consistently yield better object-level accuracy, whereas lower timesteps favor part-level details.}
        \label{fig:hierarchy_evolution}
    \end{minipage}
    \hfill
    \begin{minipage}[t]{0.41\textwidth}
        \vspace{0pt} 
        \centering
        \small
        \captionof{table}{Ablation study results on the CityScapes dataset.}
        \vspace{2mm} 
        \resizebox{\linewidth}{!}{%
        \begin{tabular}{lll}
        \toprule
         & \textbf{Ablated Feature} & \textbf{mIoU} \\
        \midrule
        1 & CSM w/ timestep sel.  (our approach)   &   \textbf{27.8} \\
        \midrule
          & no CSM w/ timestep sel. & \\
        2 & \ \ \ \  Feature Similarity & 18.7 \\
        3 & \ \ \ \ Attention Map Similarity  & 17.7 \\
        \midrule
          & no CSM w/o timestep sel. & \\
        4 & \ \ \ \ Feature Similarity (DiffCut) & 13.4  \\ 
        5 & \ \ \ \ Attention Map - based (DiffSeg) &  22.2   \\
        \midrule
        6 & CSM w/o timestep sel.  & 25.6 \\
        \midrule
          & CSM w/ timestep sel.   & \\
        7 & \ \ \ \ Decoder Layers & 20.9 \\
        8 & \ \ \ \ Encoder-Decoder Layers & 21.2 \\
        9 & \ \ \ \ Encoder Layers & 24.6 \\
        10 & \ \ \ \ Two-phase Aggregation  & 26.4 \\
        11 & \ \ \ \ \added{w/o} Added Noise  & 27.2 \\
        \bottomrule
        \end{tabular}%
        }
        \label{tab:ablation_results}
    \end{minipage}
\end{figure}

We conducted a series of ablation studies to systematically evaluate the contribution of each component, as shown in Tab.~\ref{tab:ablation_results}. We noticed a significant drop in performance as we replaced the CSM with the similarity map obtained from either the encoder features (row 2) or the self-attention maps (row 3), demonstrating that combining features with the high-resolution self-attention maps plays a pivotal role in capturing more accurate semantic information. Removing both the dynamic timestep selection and the CSMs results in a method equivalent to either DiffCut (row 4) or DiffSeg (row 5), both yielding lower performance compared to ours (row 1). Replacing our dynamic timestep selection approach with a fixed timestep results in reduced performance (row 6), validating our hypothesis that each pixel in the latent space requires a different timestep for the optimal performance. With respect to encoder features, we observed noticeable performance drops as we used features from all layers of decoder (row 7), encoder + decoder (row 8), or encoder (row 9), instead of the features obtained from the last layer of U-Net's encoder (row 1). \replaced{Using CSM at the target timestep instead}{the Elimination} of the feature aggregation presented in Eq.~\ref{eq:hierarchical_aggregation} led to reduced performance (row 10), demonstrating the importance of the trade-off between granular and abstract semantic features. While in DiffSeg, the original image is supplied to the network without adding any noise, row 11 shows that adding noise helps with the model's performance because it provides an input closer to what the model has seen during training. 


\subsection{Semantic level distribution \& selection}
\label{sec:semantic_level_distribution}

We analyzed the distribution of the semantic levels within different datasets (please refer to Supp. Fig.~\ref{fig:semantic_level_average_per_dataset} for details).
Our analysis showed that majority of pixels have between 3 to 5 semantic levels, with an average of 4. We also conducted an analysis to investigate the impact of selecting different semantic levels on the performance of the model.   
Tab.~\ref{tab:ablation_semantic_levels} shows that selecting the second semantic level consistently provides the largest performance compared to other semantic levels.


\subsection{Open-vocabulary segmentation}
We also expanded our model to an open-vocabulary setting, where the identified masks were assigned to an arbitrary set of textual semantic labels (see Supp. Sec.~\ref{sec:supp_open_vocabulary} for details). Tab.~\ref{tab:open_vocabulary_results} compares our model against DiffCut, showing a better performance across different datasets. 

\begin{table}[h!]
    \centering
    \begin{minipage}{0.5\textwidth}
        \centering
        \caption{Ablation of selected target level on the performance in the CityScapes dataset.}
        \label{tab:ablation_semantic_levels}
        \small
        \begin{tabular}{lccccc}
            \toprule
            \textbf{Method} & \textbf{L1} & \textbf{L2} & \textbf{L3} & \textbf{L4} & \textbf{L5} \\
            \midrule
            mIOU & 25.5 & \textbf{27.8} & 25.3 & 17.5 & 12.4 \\
            \bottomrule
        \end{tabular}
    \end{minipage}
    \hfill 
    \begin{minipage}{0.45\textwidth}
        \centering
        \caption{Open-voc segmentation (mIoU).}
        \label{tab:open_vocabulary_results}
        \small
        \begin{tabular}{lcccc}
            \toprule
            \textbf{Model} & \textbf{VOC} & \textbf{Context} & \textbf{COCO} \\
            \midrule
            DiffCut & 36.8 & 20.6 & 19.1 \\
            Ours & \textbf{44.0} & \textbf{22.4} & \textbf{20.6} \\
            \bottomrule
        \end{tabular}
    \end{minipage}
\end{table}

\section{Discussion}

\deleted{Existing zero-shot segmentation methods often face limitations due to a trade-off between spatial resolution and contextual information, and their reliance on feature extraction from a single denoising timestep. This overlooks the dynamic evolution of semantic granularity within diffusion models.}

\deleted{To overcome these, our work introduces two key advancements. First, our Contextual Similarity Maps synergistically fuse high-resolution attention maps with rich U-Net encoder features, providing both fine-grained and robust representations (Sec.~\ref{sec:contextual_similarity_map}). Second, recognizing that these maps capture hierarchical semantics across the denoising trajectory, we developed a mechanism to adaptively identify and utilize the optimal timestep for each pixel in the latent space (Sec.~\ref{sec:semantic_boundary_detection}).
Our experiments demonstrate that our method consistently outperforms existing zero-shot segmentation baselines, validating the efficacy of our contextual similarity maps with dynamic timestep selection.}

\added{In this work, our empirical analysis reveals that a hierarchical semantic progression naturally emerges across the denoising trajectory of diffusion models, transitioning from part-level details to object-level abstractions. This dynamic evolution exposes the fundamental drawback of relying on a single, static extraction timestep, a limitation that we overcome through adaptive, per-pixel timestep selection. Concurrently, our investigations reveal an inherent trade-off between spatial resolution and contextual awareness in prior work. To address this, we introduce Contextual Similarity Maps to synergize high-resolution attention with contextual feature representations. Our experiments demonstrate that the proposed method founded on these insights consistently outperforms existing zero-shot segmentation baselines, validating the efficacy of our contextual similarity maps with dynamic timestep selection.}

Our findings open several promising research avenues. The concept of dynamic semantic levels naturally extends to interactive segmentation, enabling user-controlled abstraction, as well as to hierarchical vision tasks like instruction-based image editing. Beyond vision, adapting our core principles to recent diffusion-based language models could provide a mechanism for finer semantic understanding and control. Finally, investigating how domain-adaptive fine-tuning impacts the specialization of these internal representations remains an important area for future study.


\bibliographystyle{plainnat}
\bibliography{main}

\clearpage

\appendix

\section{Limitations and Future Directions}
\label{sec:supp_limitation_and_future_directions}

While our work successfully demonstrates the hierarchical progression of semantic information within the denoising process, we identify some valuable avenues for future investigations.

\textit{Expansion to diffusion language models:} In this work, we demonstrated the emergence of hierarchical semantics progression in vision-based generative diffusion models. This emergence is primarily based on self-attention maps and the features of diffusion backbones which are the core parts of the denoising process. With the rapid development and extensive interest in diffusion-based language models, a promising future direction is to investigate the same idea within this scope to assist with finer understanding and generation in these models. 

\textit{Computational Trade-offs:} Our current approach analyzes features across multiple denoising steps. Consequently, the runtime scales with the number of timesteps evaluated, presenting a trade-off between resolution and computational cost. While we have conducted a series of experiments to analyze the computational footprint of the model, we believe an important future direction is to investigate the temporal dynamics of this hierarchy more deeply. Such a study could reveal methods to capture the full progression from a sparser, optimized set of timesteps, enhancing efficiency.

\textit{Feature Access and Efficiency:} To access the rich, detailed self-attention maps, our method (akin to DiffSeg \cite{tian2024diffuse}) requires computation of the full U-Net. This methodological choice is necessary for the depth of our analysis but results in higher computational and memory requirements compared to approaches like DiffCut and DiffSeg. This highlights a clear opportunity for investigation of characteristics such as dimensionality reduction on the extracted features for more efficiency.

\textit{Generalization of Analysis:} Like other deep learning methods, Stable Diffusion models are naturally influenced by their training distribution, which can affect performance on out-of-domain data. In this work, our analysis is grounded in the internal representations of Stable Diffusion with in-domain data. Investigating the robustness and generalization of these observed semantic hierarchies to out-of-domain data such as medical images, satellite imagery, etc. is an interesting next step, representing a broader challenge in understanding these foundation models.

\begin{sidewaysfigure*}[htb!] 
    \centering
    \begin{tabular}{ccc}
        \includegraphics[width=0.2\linewidth]{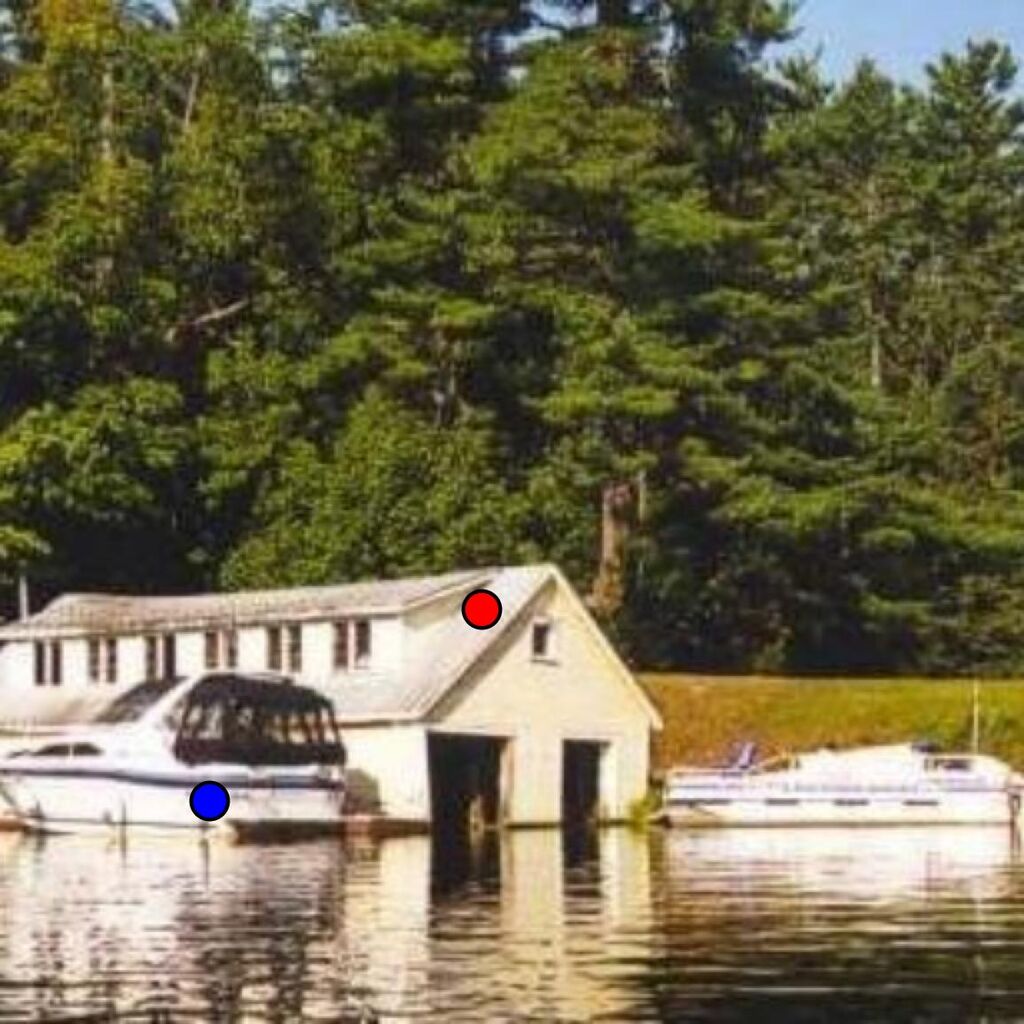}  & \includegraphics[width=0.2\linewidth]{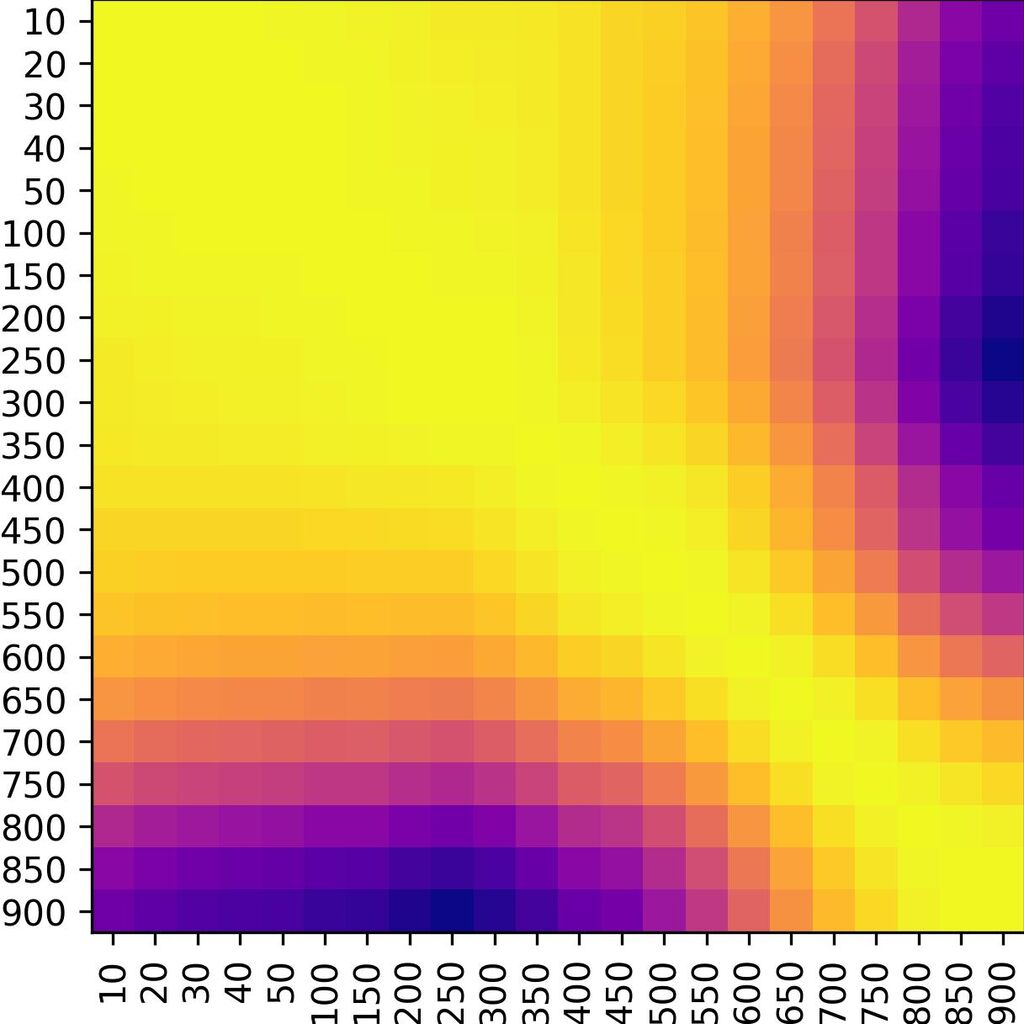} &
        \includegraphics[width=0.2\linewidth]{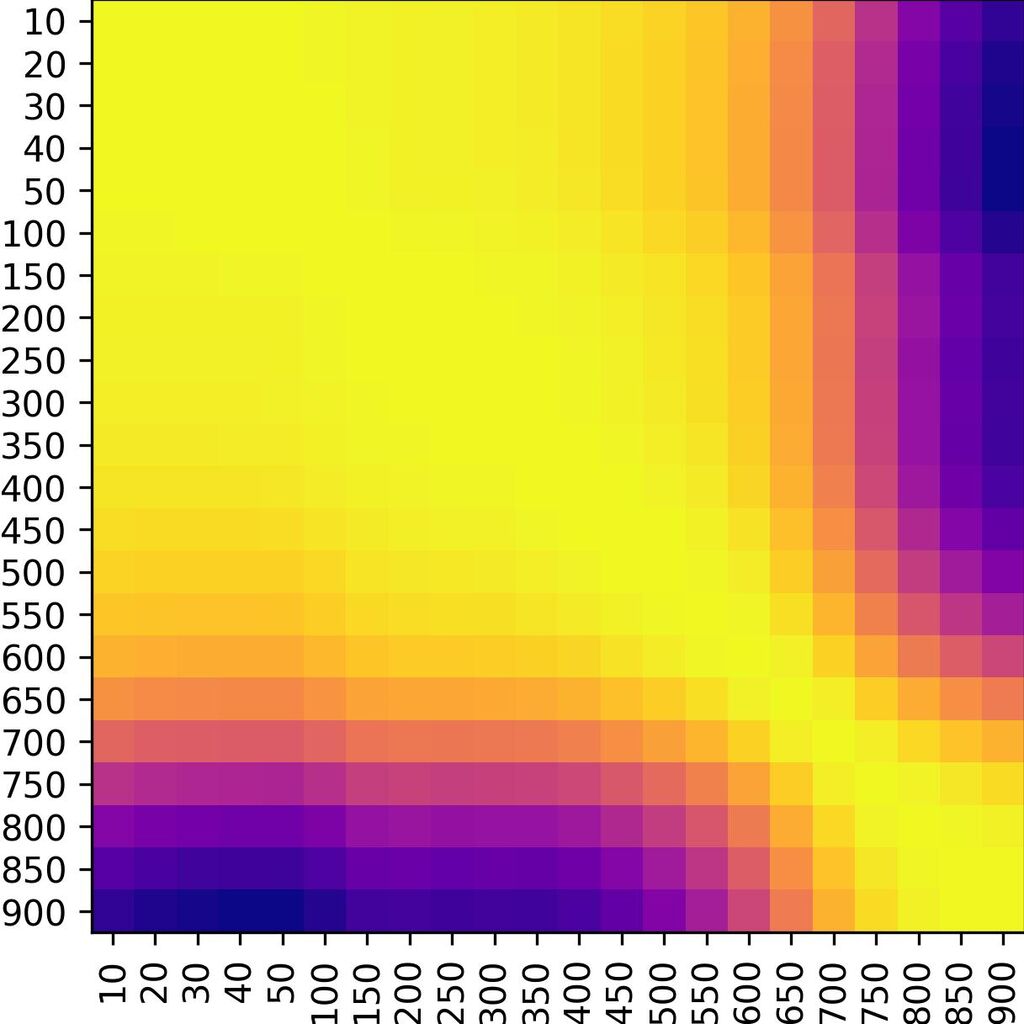}\\
        Input Image & Red point TSM & Blue point TSM
    \end{tabular}
    \vspace{1em} 

    \setlength{\tabcolsep}{2pt} 
    \begin{tabularx}{\linewidth}{ *{11}{>{\centering\arraybackslash}p{0.0845\linewidth}} } 
    
    \multicolumn{11}{c}{\textbf{Red point CSMs}} \\
    
    \fcolorbox{red}{white}{\includegraphics[width=0.95\linewidth]{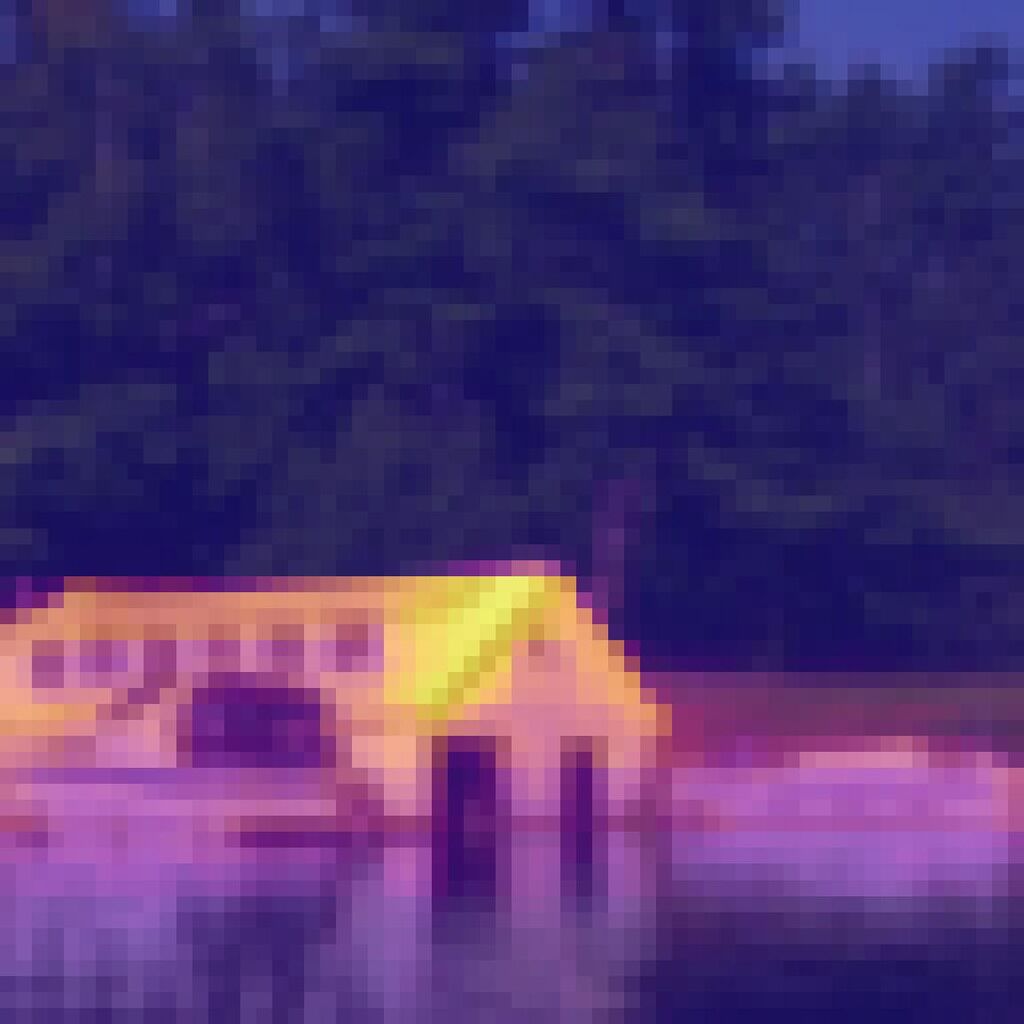}} &
    \fcolorbox{red}{white}{\includegraphics[width=0.95\linewidth]{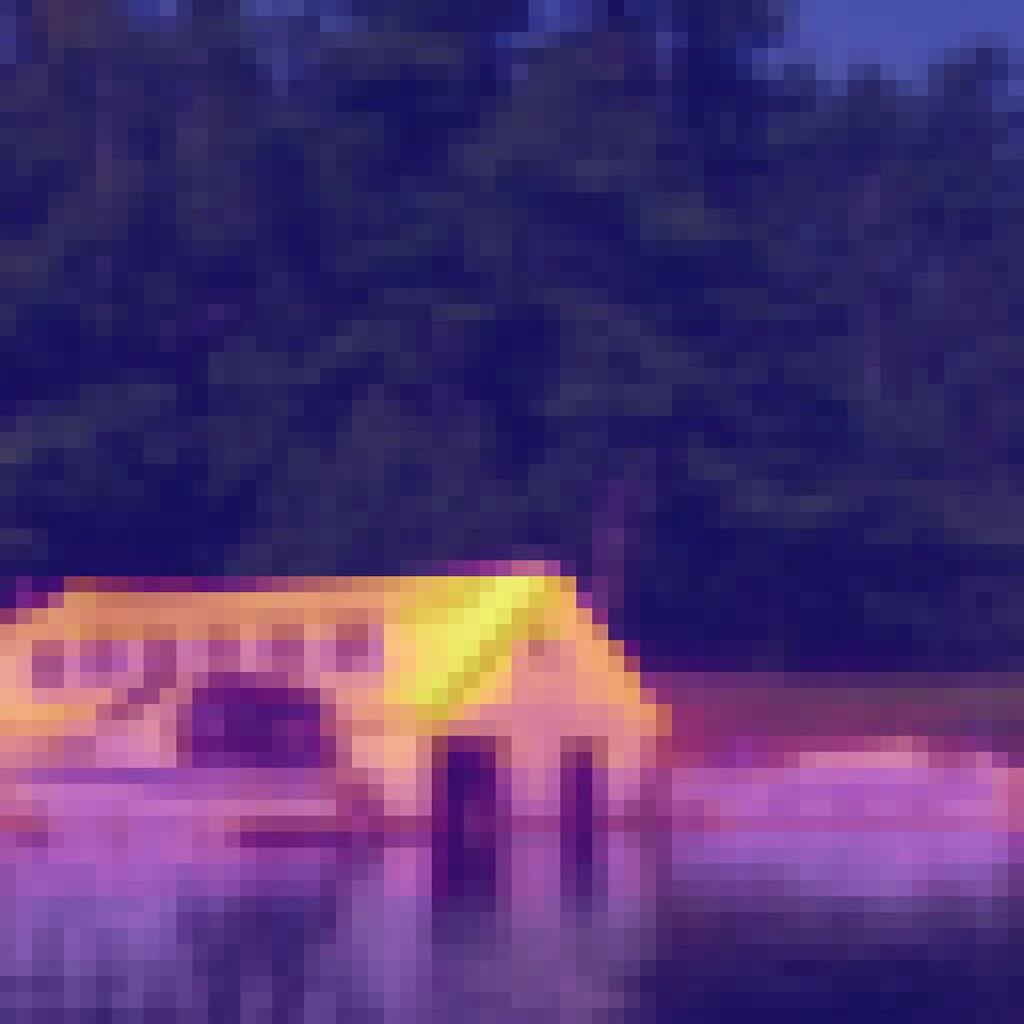}} &
    \fcolorbox{red}{white}{\includegraphics[width=0.95\linewidth]{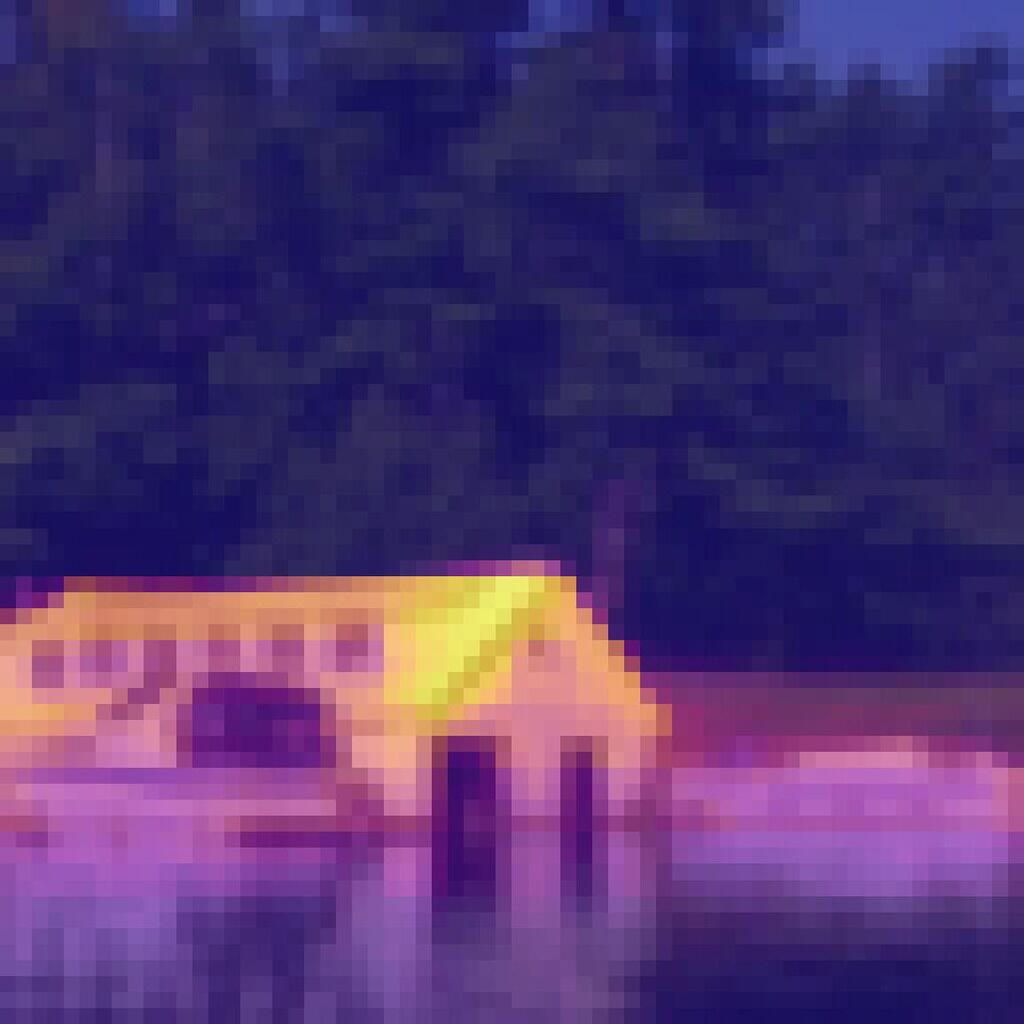}} &
    \fcolorbox{red}{white}{\includegraphics[width=0.95\linewidth]{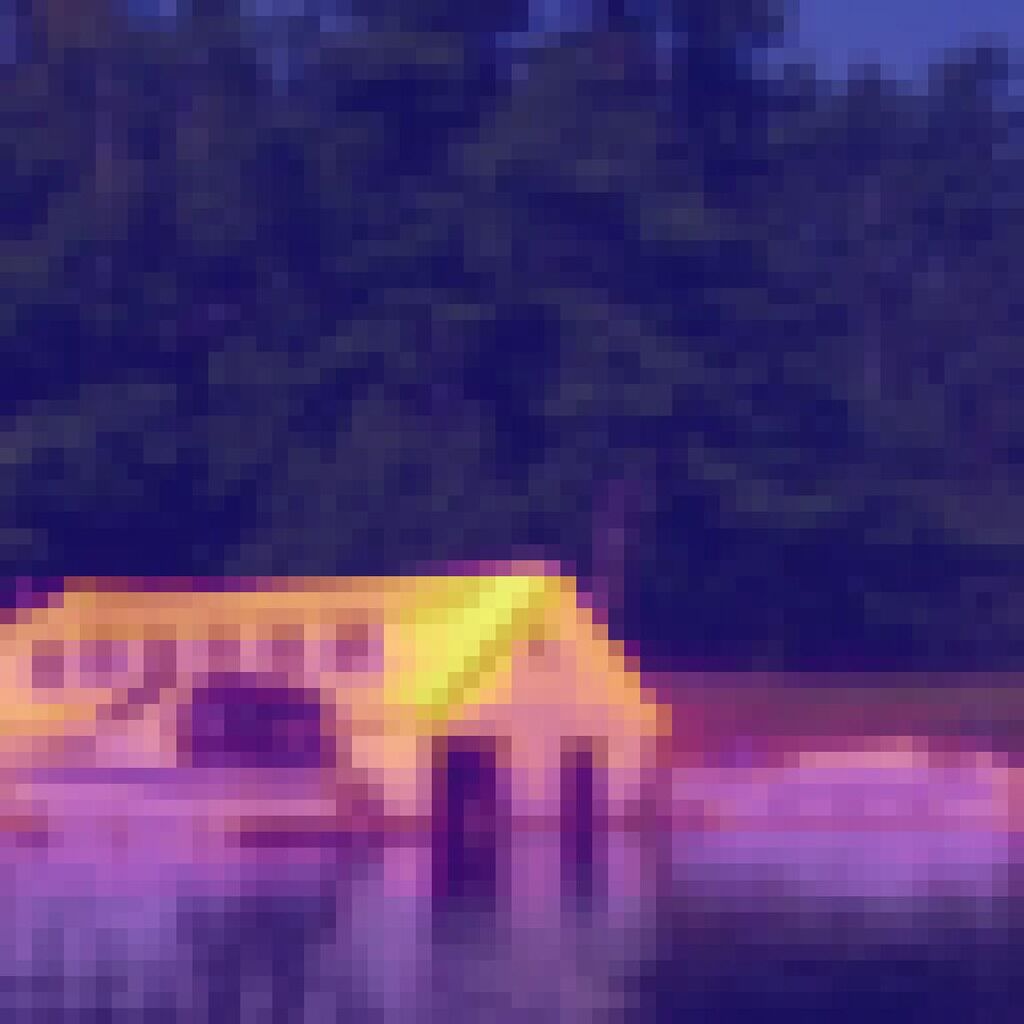}} &
    \fcolorbox{red}{white}{\includegraphics[width=0.95\linewidth]{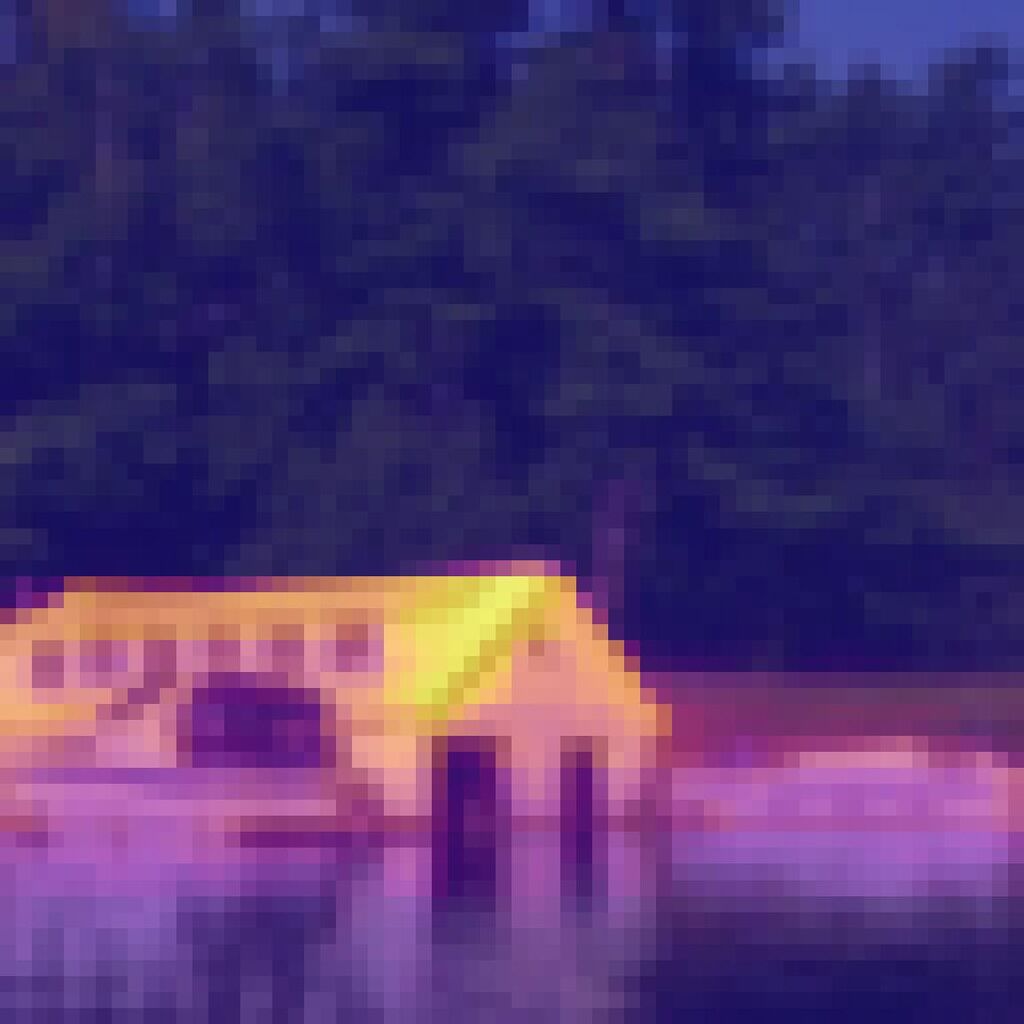}} &
    \fcolorbox{red}{white}{\includegraphics[width=0.95\linewidth]{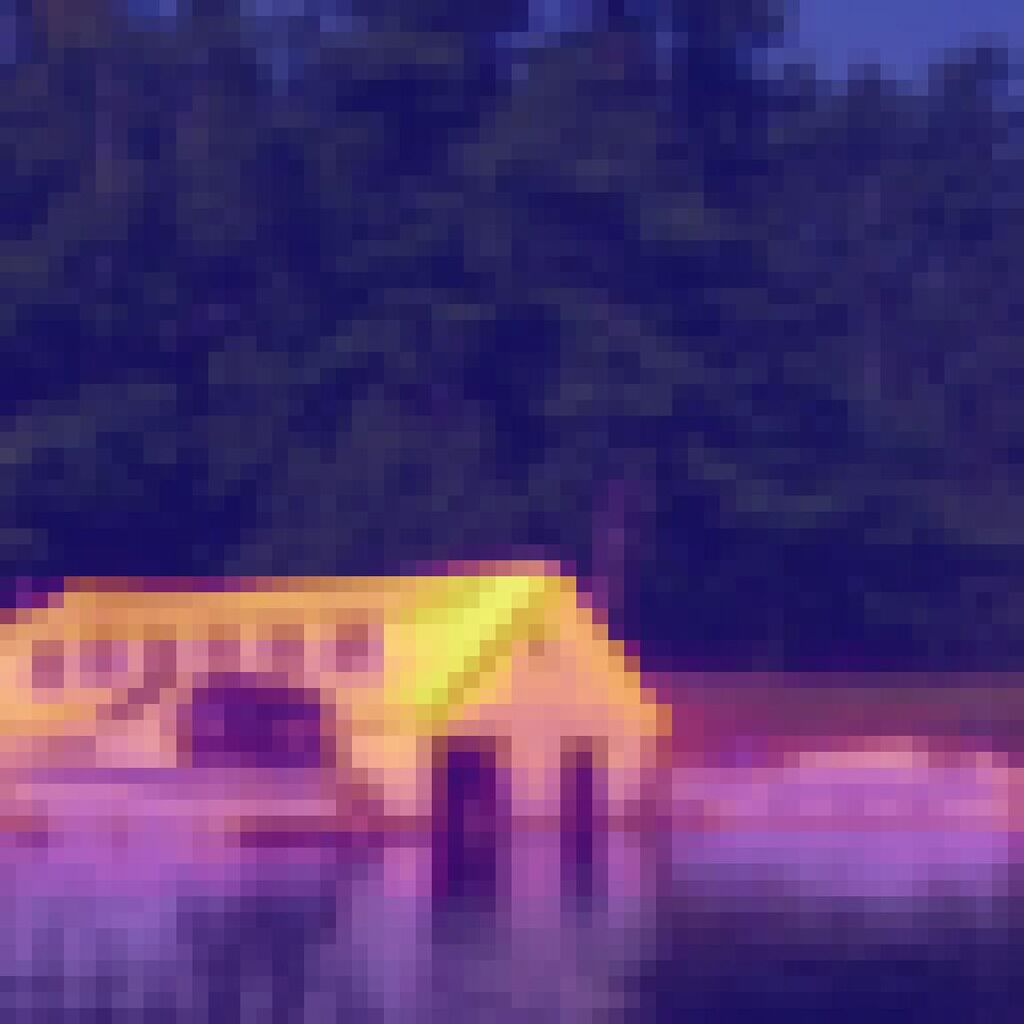}} &
    \fcolorbox{red}{white}{\includegraphics[width=0.95\linewidth]{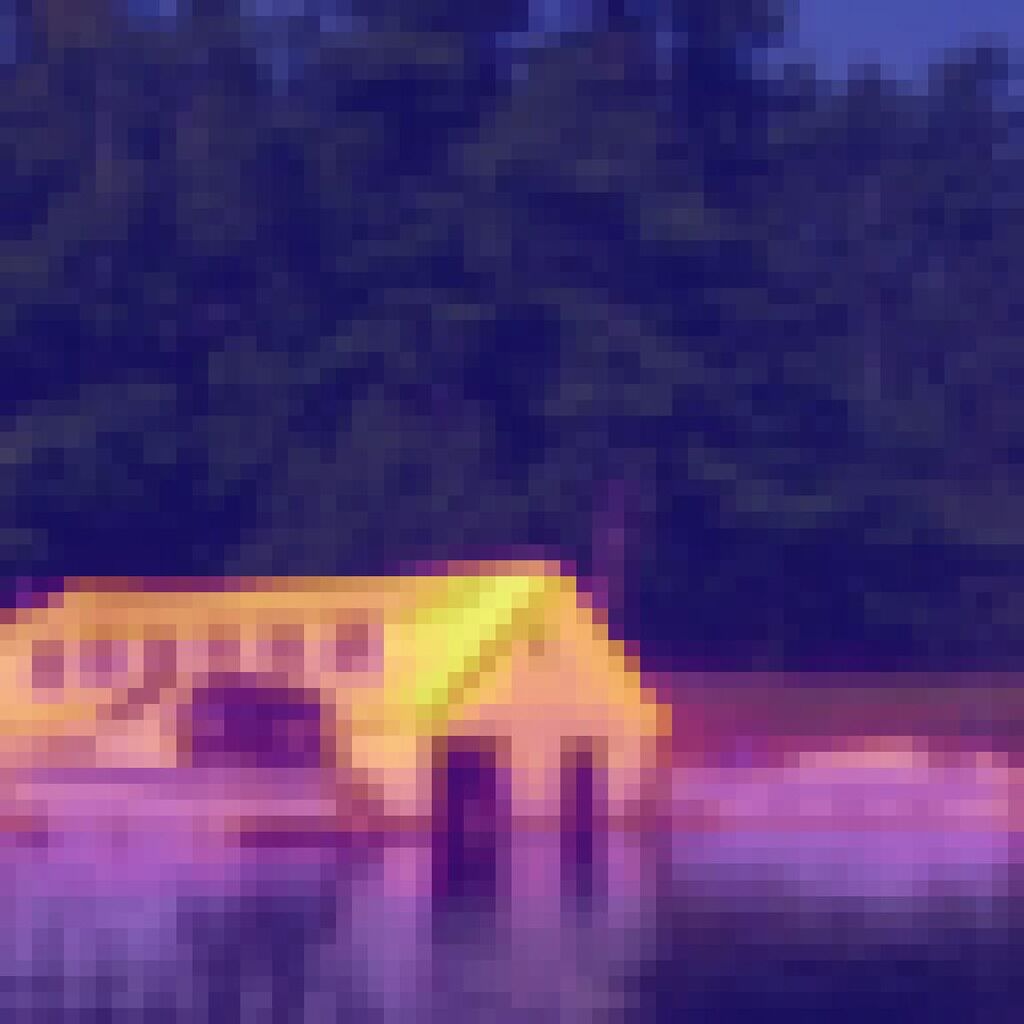}} &
    \fcolorbox{red}{white}{\includegraphics[width=0.95\linewidth]{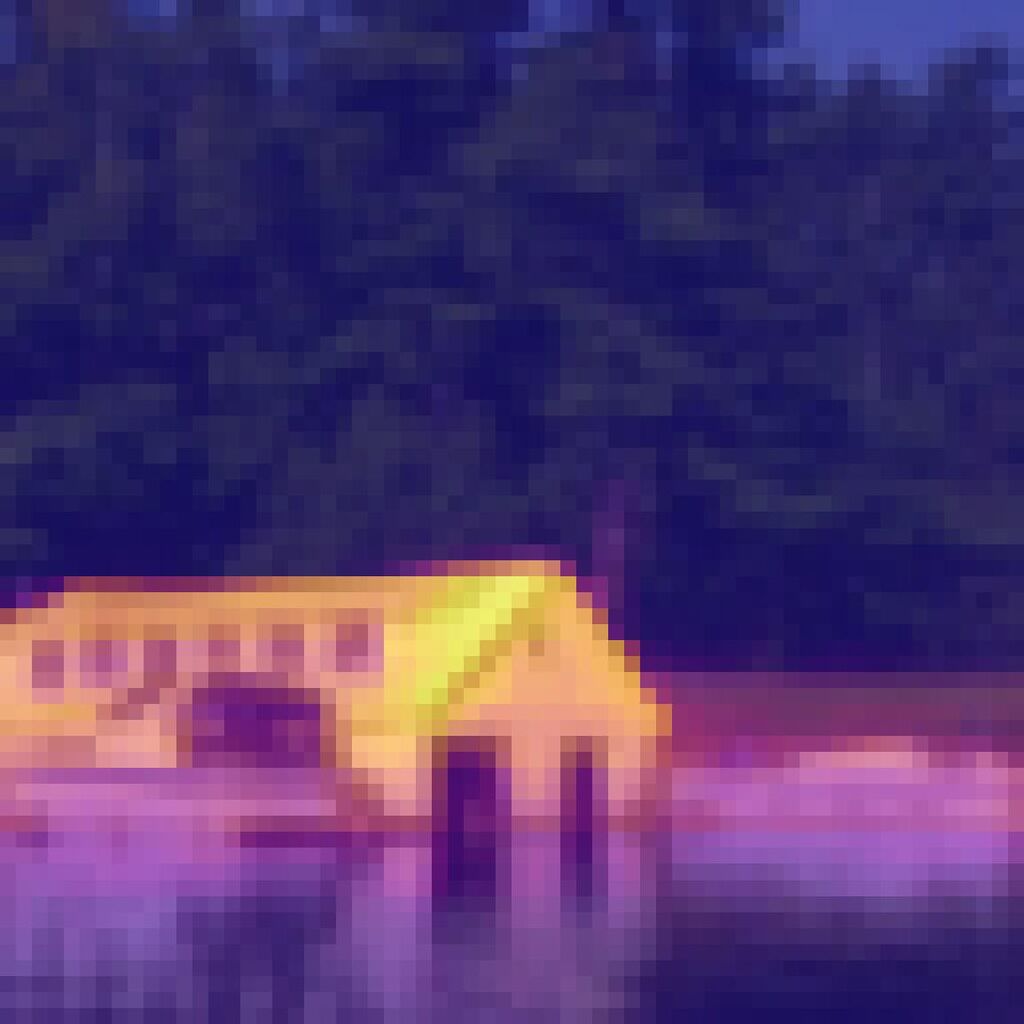}} &
    \fcolorbox{red}{white}{\includegraphics[width=0.95\linewidth]{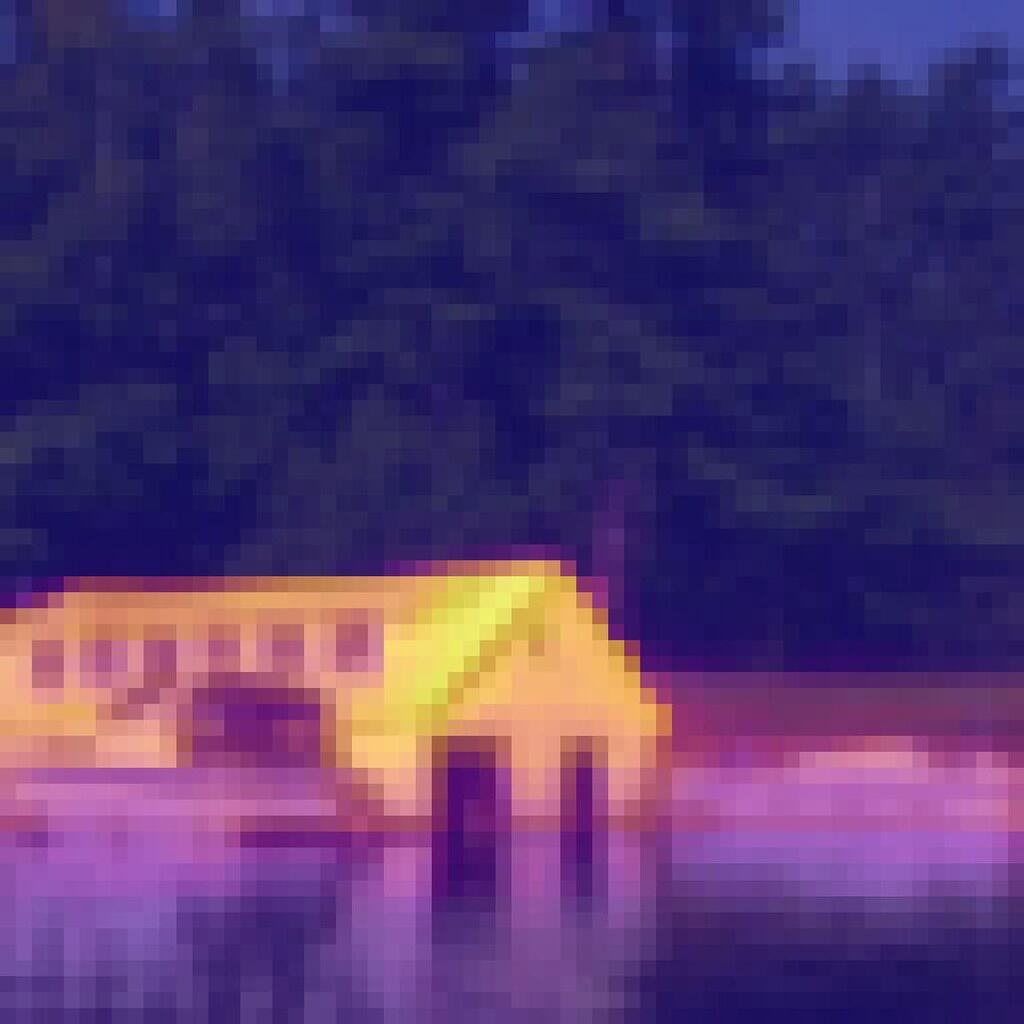}} &
    \fcolorbox{red}{white}{\includegraphics[width=0.95\linewidth]{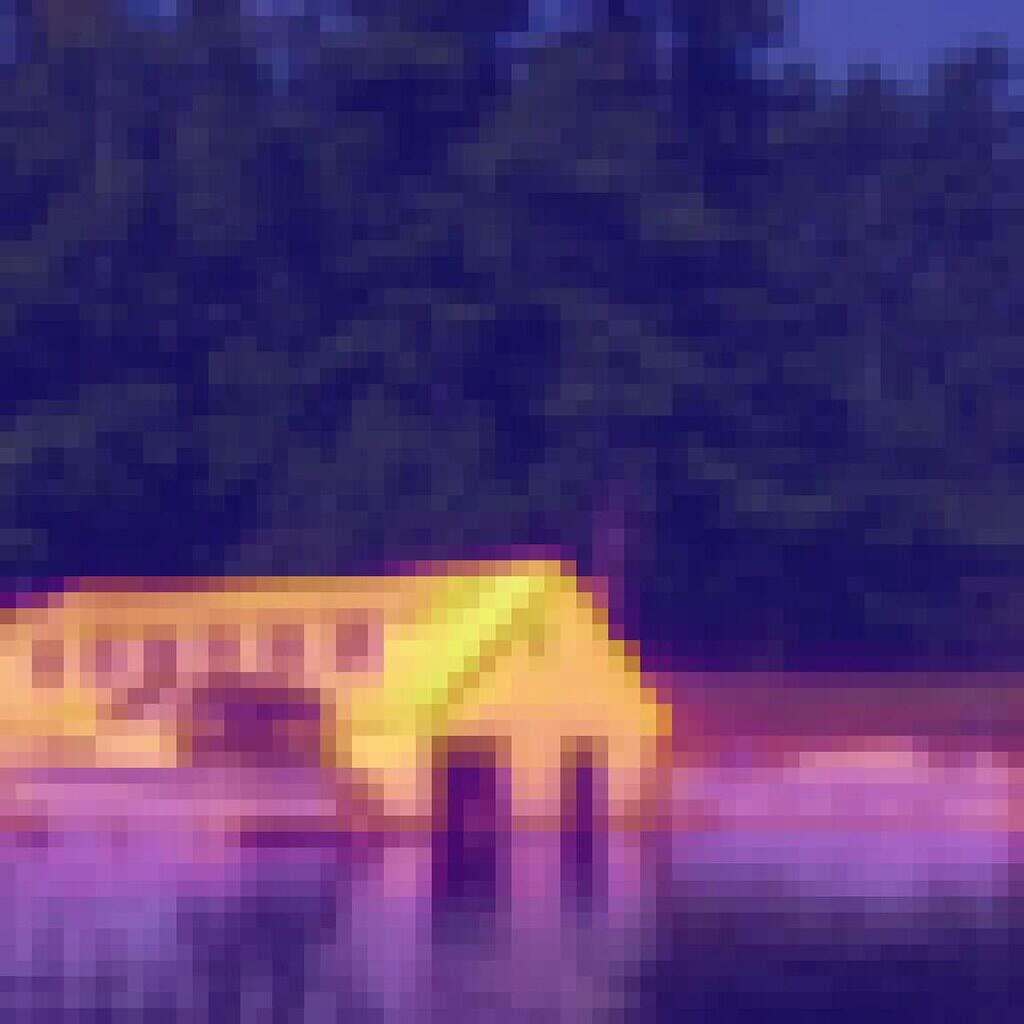}} &
    \fcolorbox{red}{white}{\includegraphics[width=0.95\linewidth]{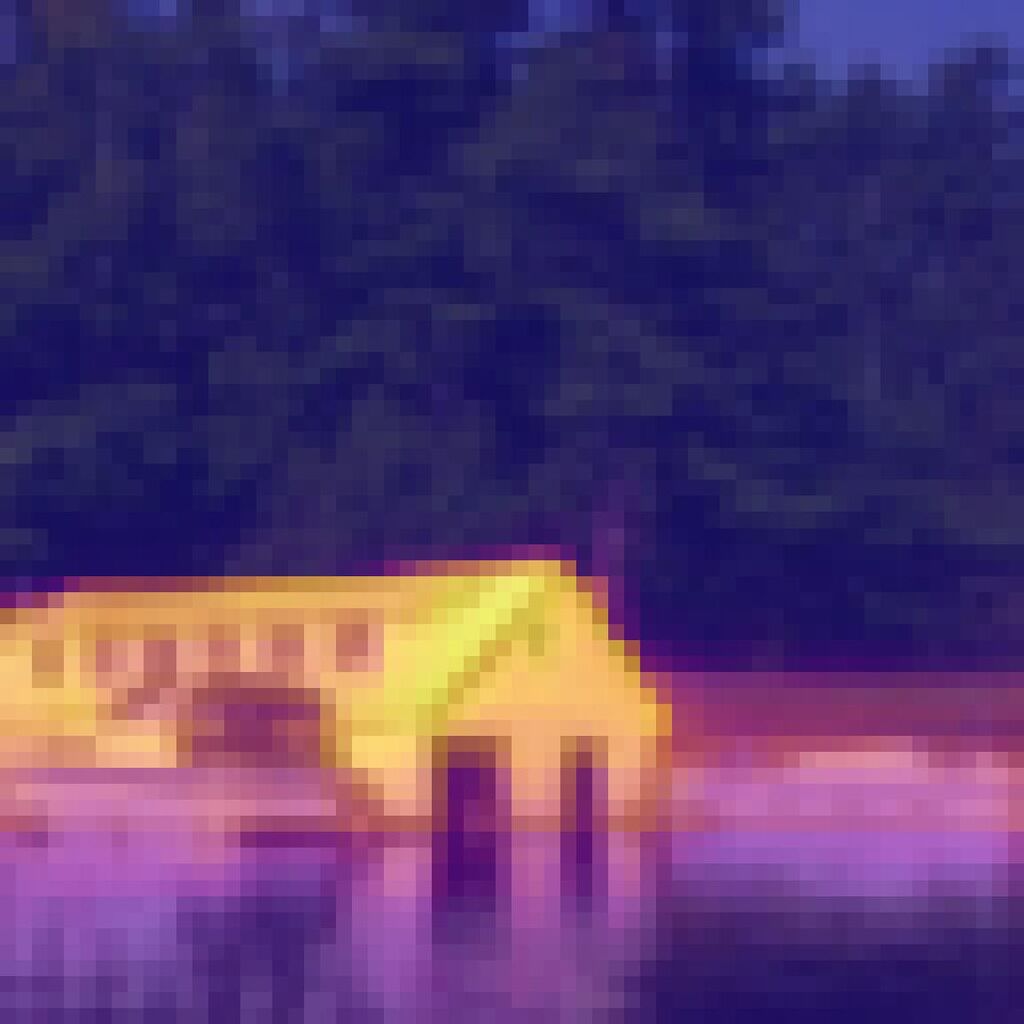}} \\
    
    t=10 & t=20 & t=30 & t=40 & t=50 & t=100 & t=150 & t=200 & t=250 & t=300 & t=350 \\[1em]
    
    \fcolorbox{red}{white}{\includegraphics[width=0.95\linewidth]{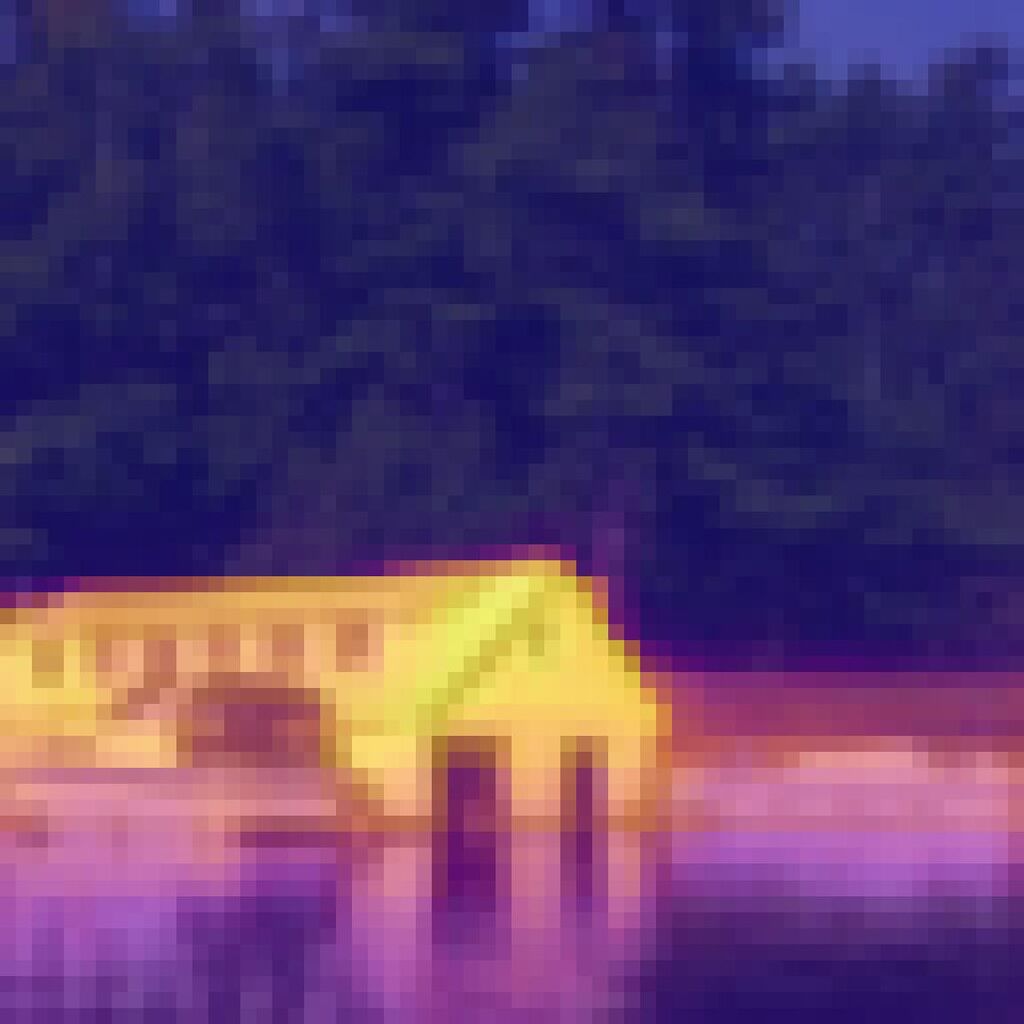}} &
    \fcolorbox{red}{white}{\includegraphics[width=0.95\linewidth]{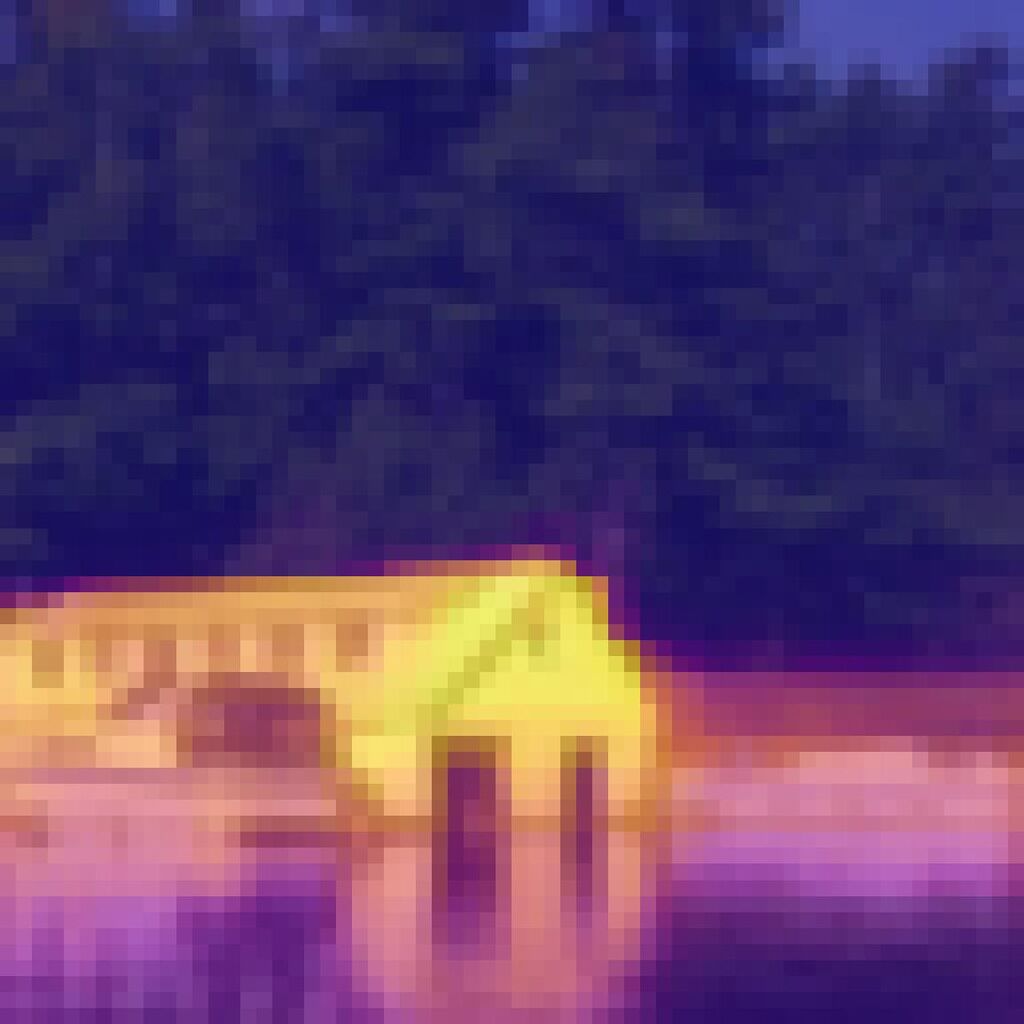}} &
    \fcolorbox{red}{white}{\includegraphics[width=0.95\linewidth]{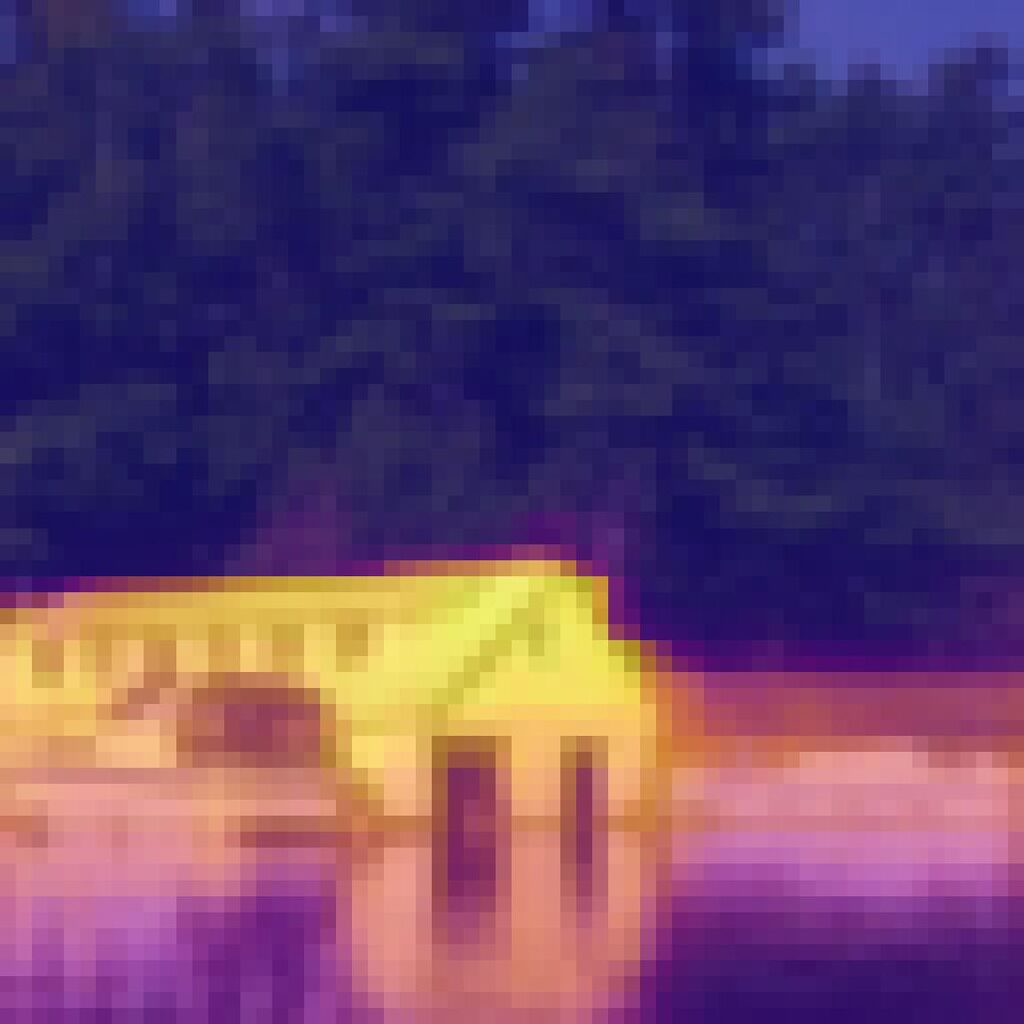}} &
    \fcolorbox{red}{white}{\includegraphics[width=0.95\linewidth]{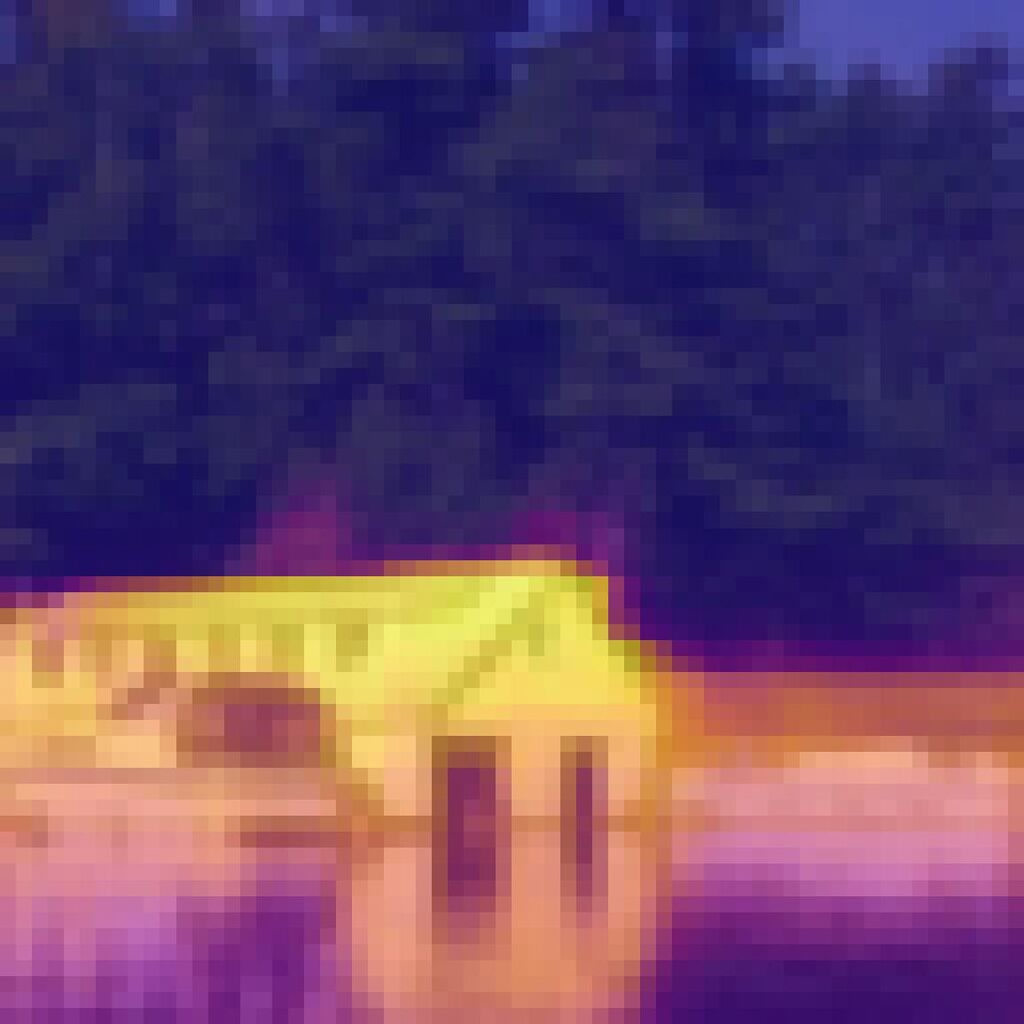}} &
    \fcolorbox{red}{white}{\includegraphics[width=0.95\linewidth]{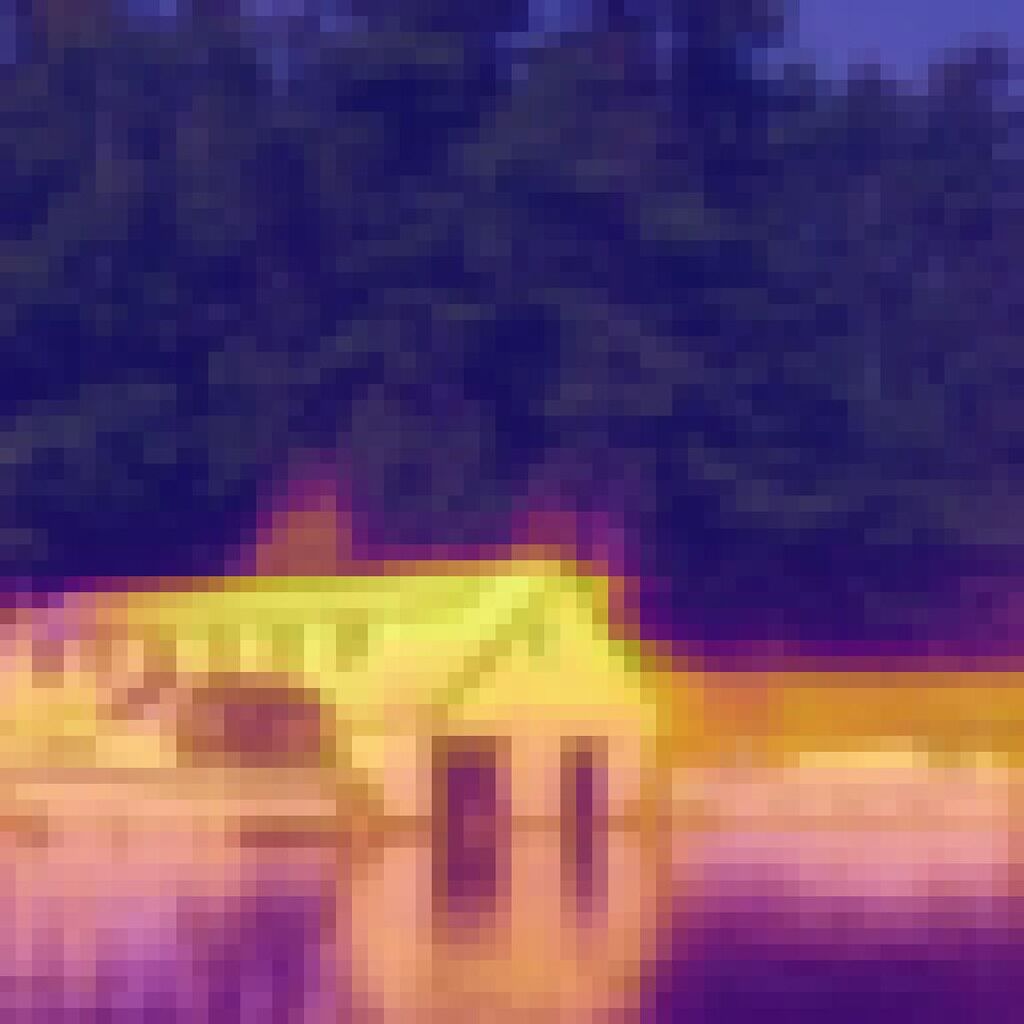}} &
    \fcolorbox{red}{white}{\includegraphics[width=0.95\linewidth]{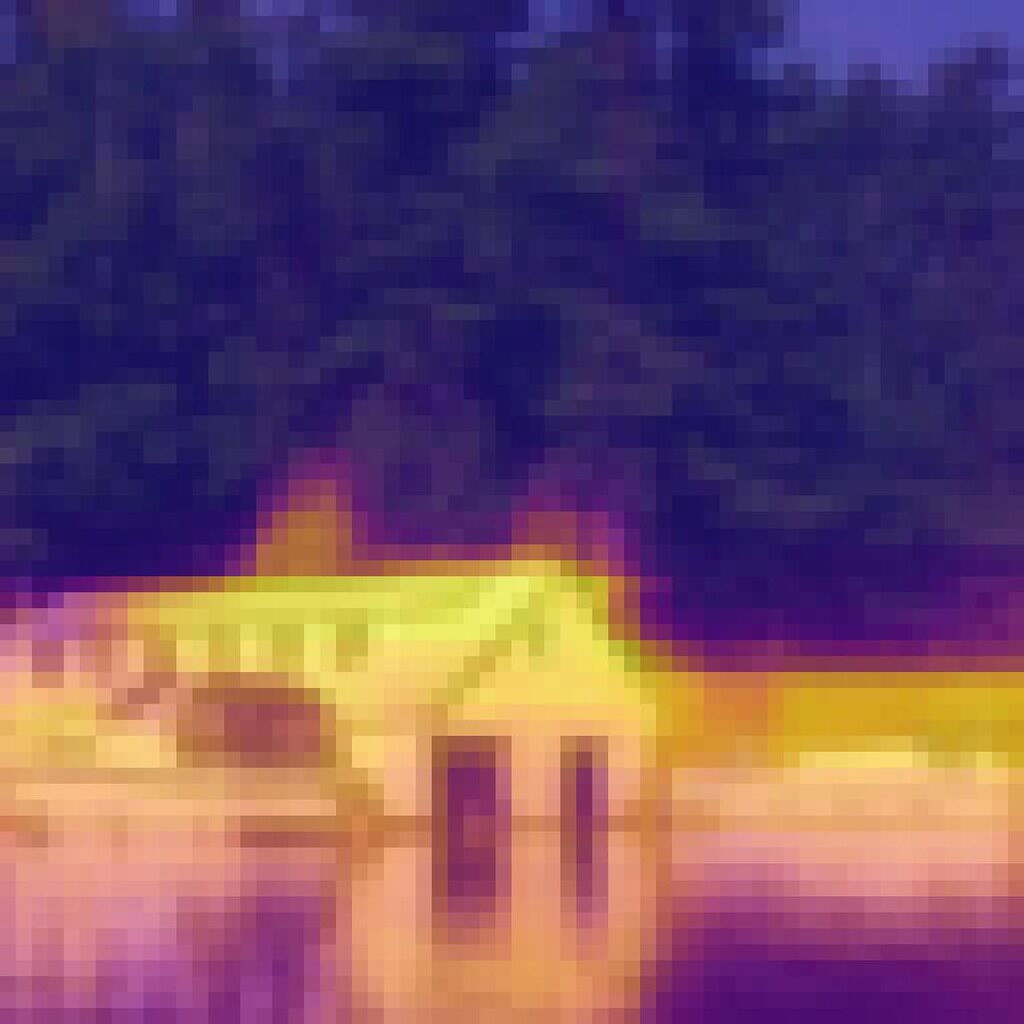}} &
    \fcolorbox{red}{white}{\includegraphics[width=0.95\linewidth]{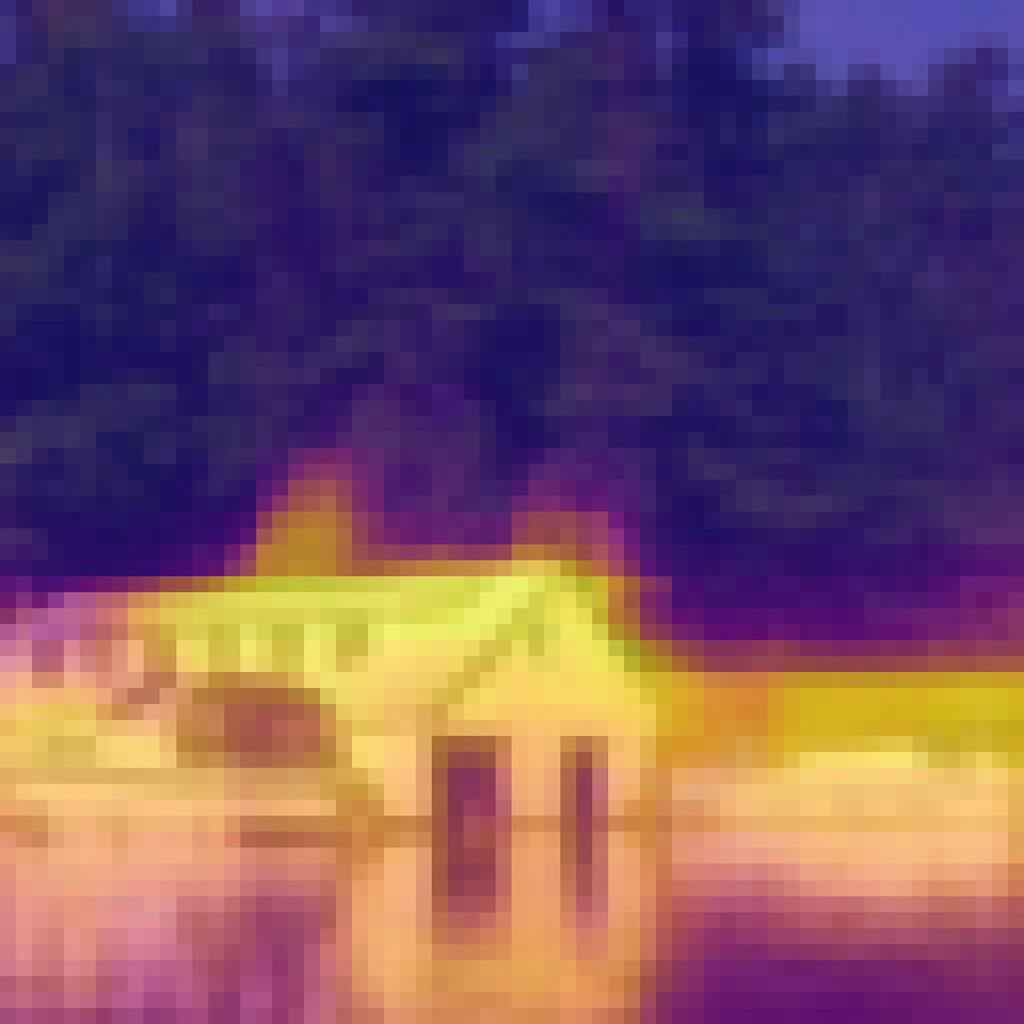}} &
    \fcolorbox{red}{white}{\includegraphics[width=0.95\linewidth]{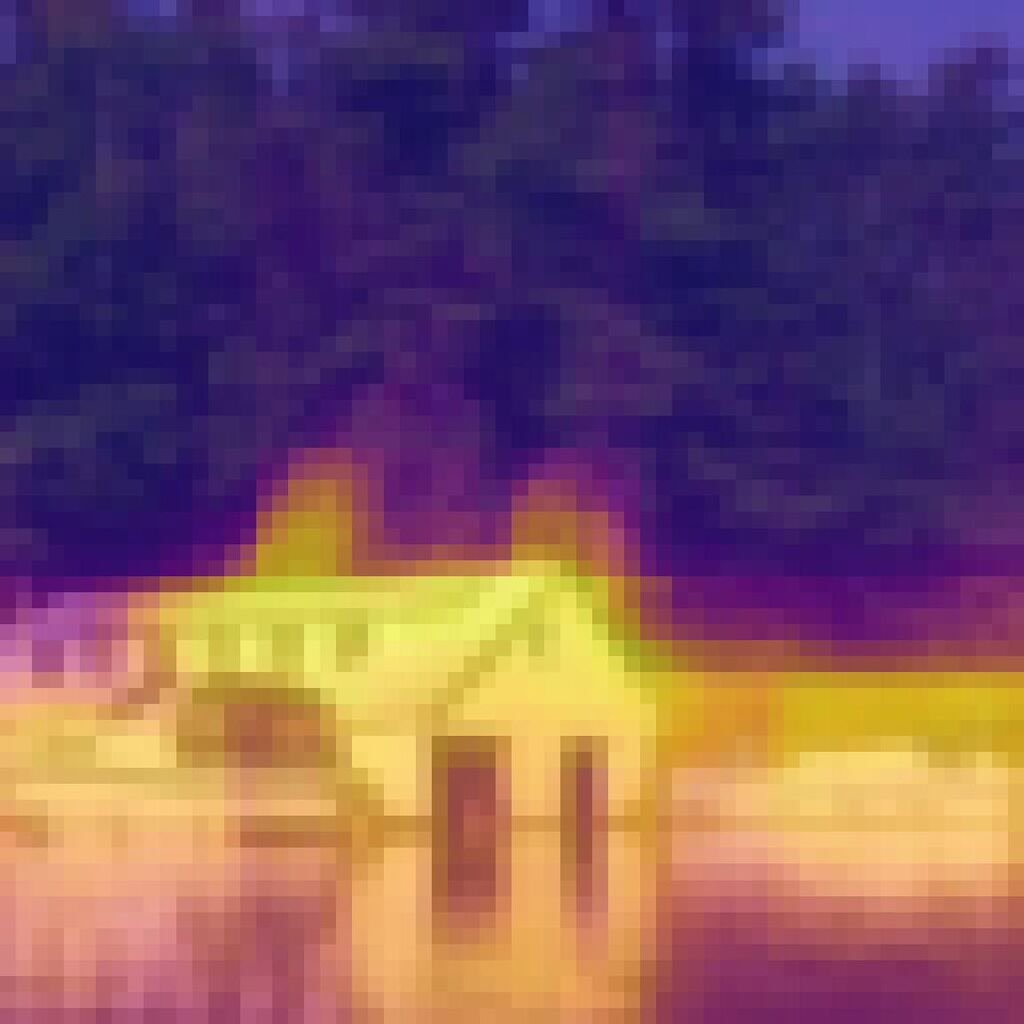}} &
    \fcolorbox{red}{white}{\includegraphics[width=0.95\linewidth]{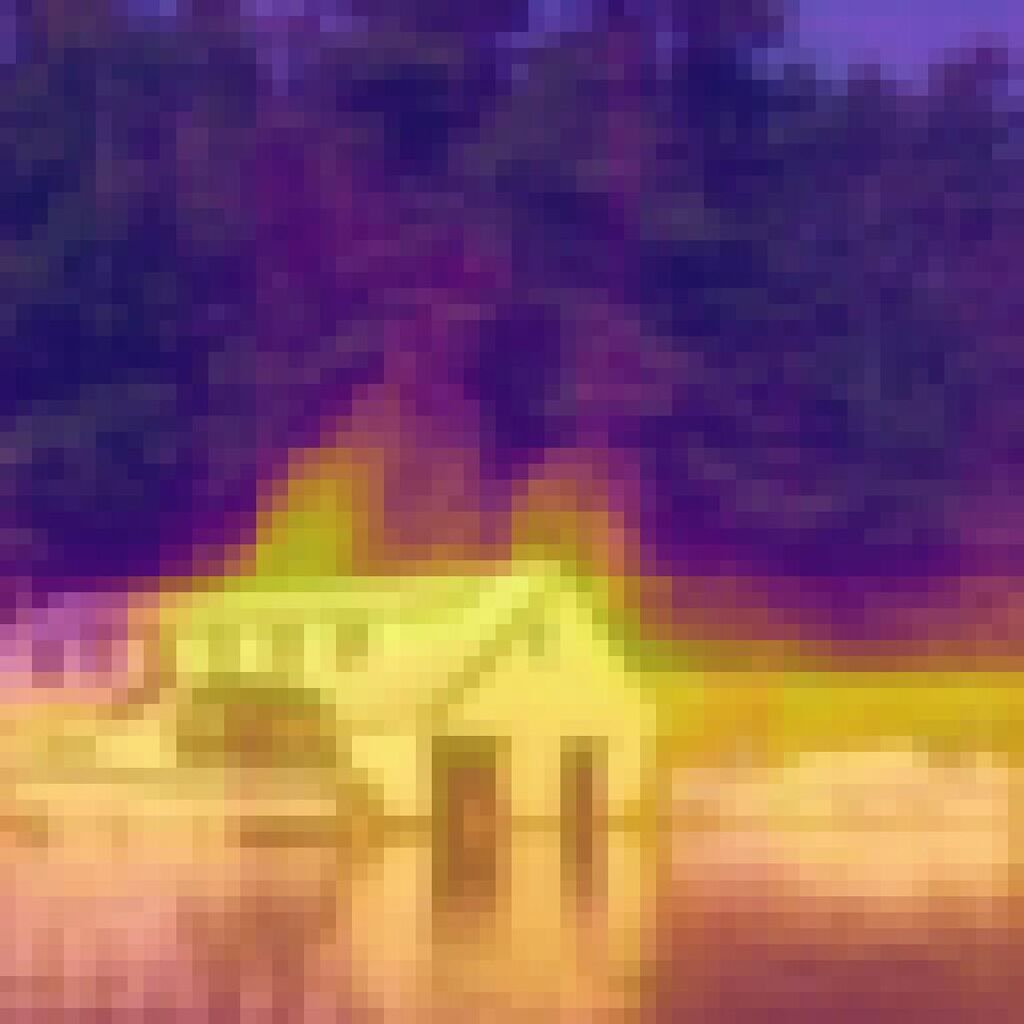}} &
    \fcolorbox{red}{white}{\includegraphics[width=0.95\linewidth]{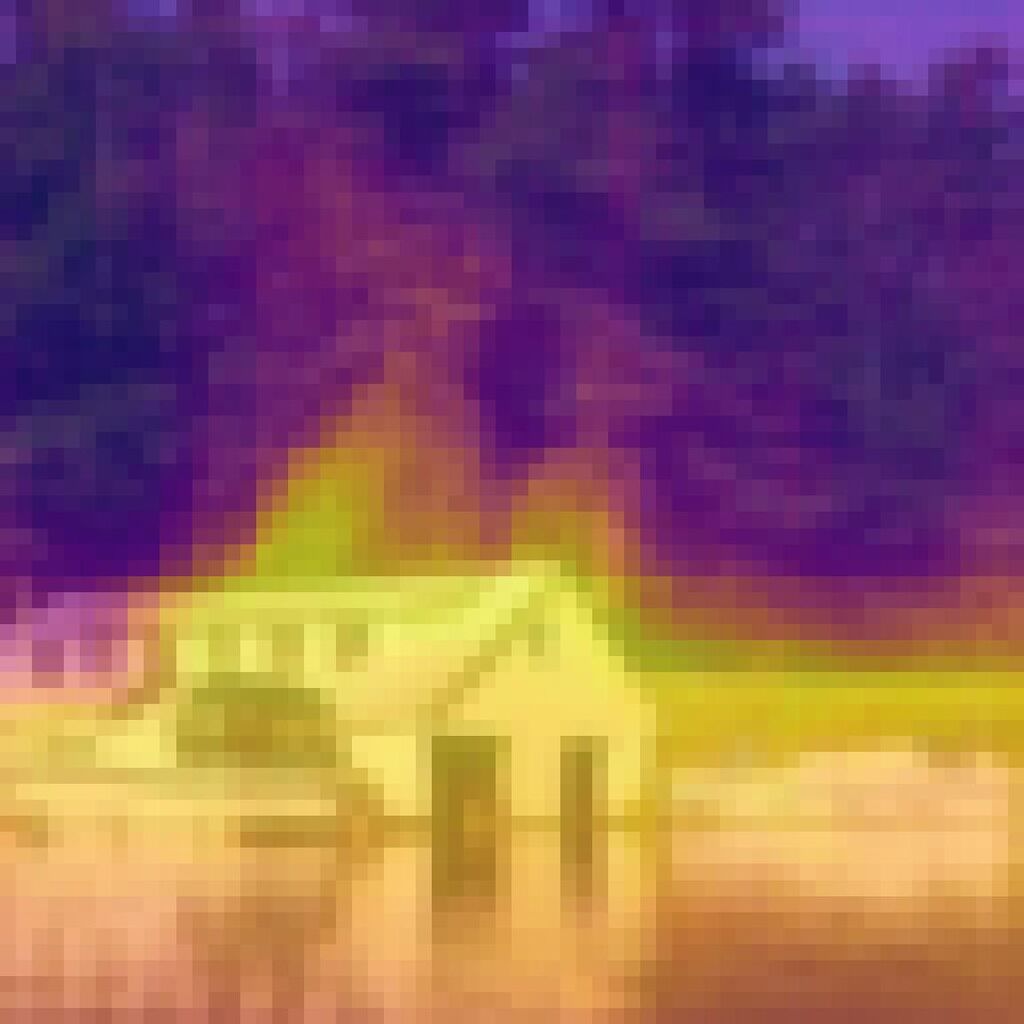}} &
    \fcolorbox{red}{white}{\includegraphics[width=0.95\linewidth]{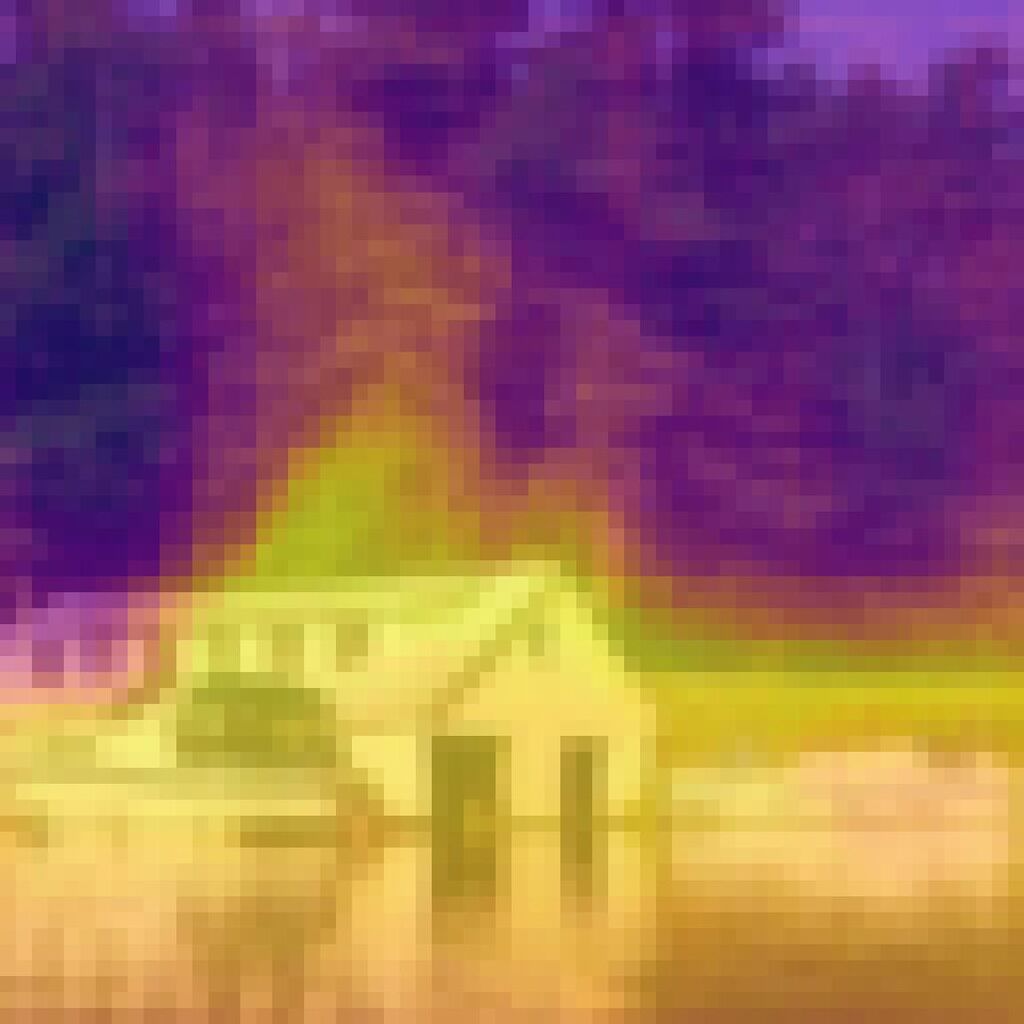}} \\
    
    t=400 & t=450 & t=500 & t=550 & t=600 & t=650 & t=700 & t=750 & t=800 & t=850 & t=900 \\
    
    \multicolumn{11}{c}{\vspace{0.5em}} \\ 

    \multicolumn{11}{c}{\textbf{Blue point CSMs}} \\
    
    \fcolorbox{blue}{white}{\includegraphics[width=0.95\linewidth]{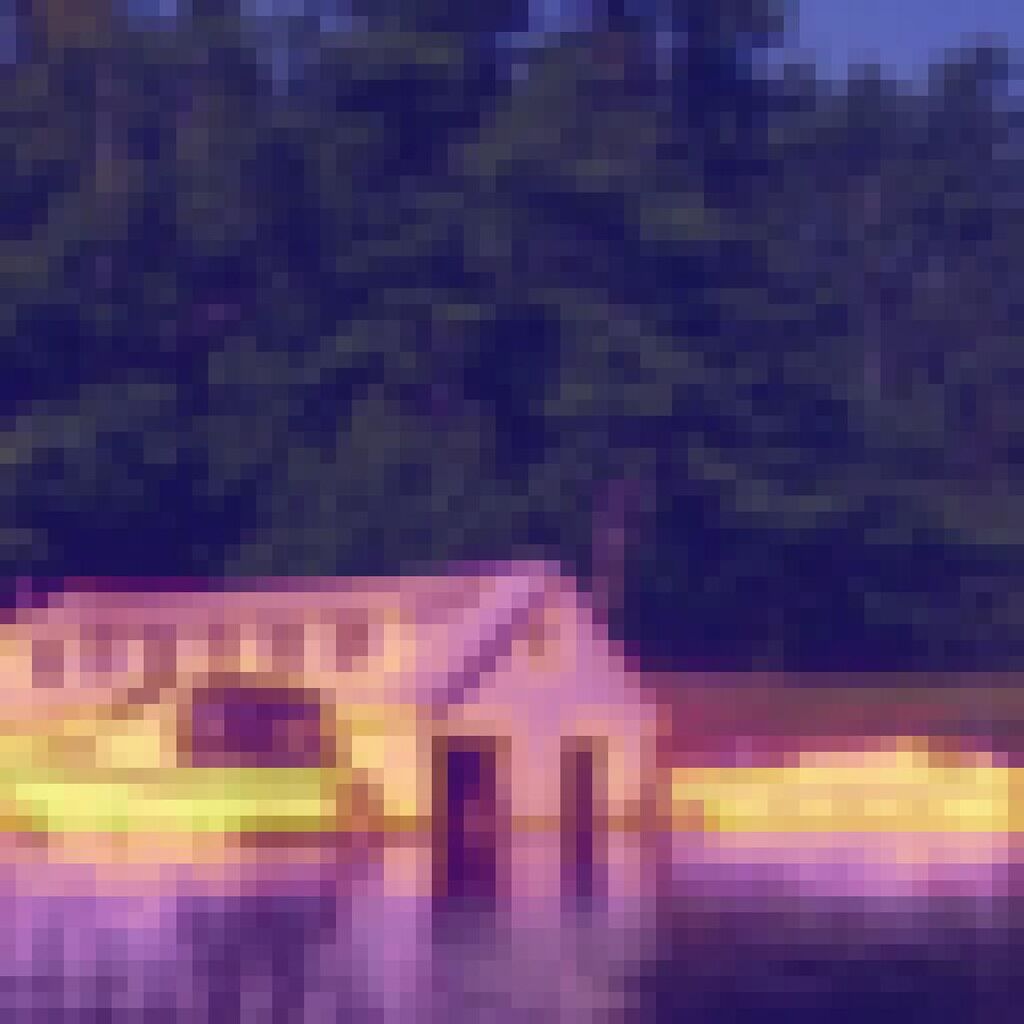}} &
    \fcolorbox{blue}{white}{\includegraphics[width=0.95\linewidth]{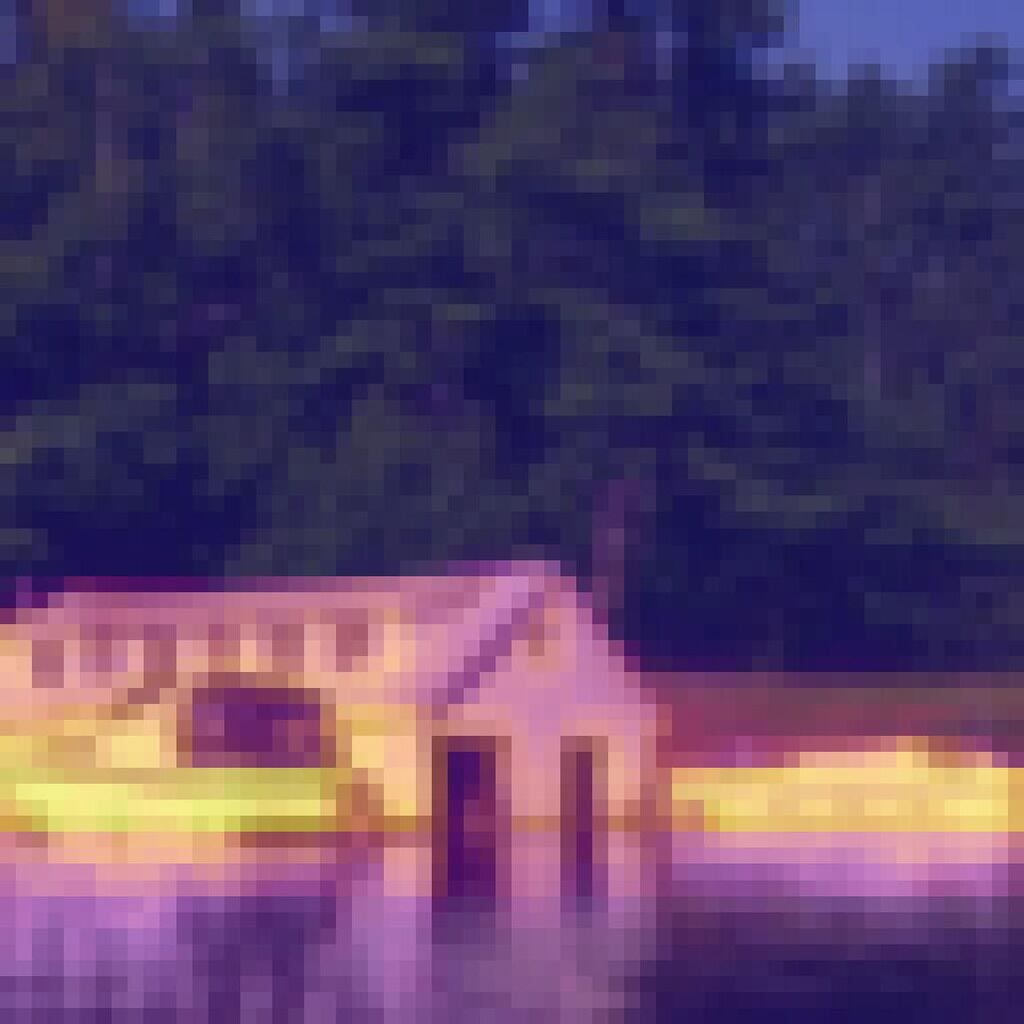}} &
    \fcolorbox{blue}{white}{\includegraphics[width=0.95\linewidth]{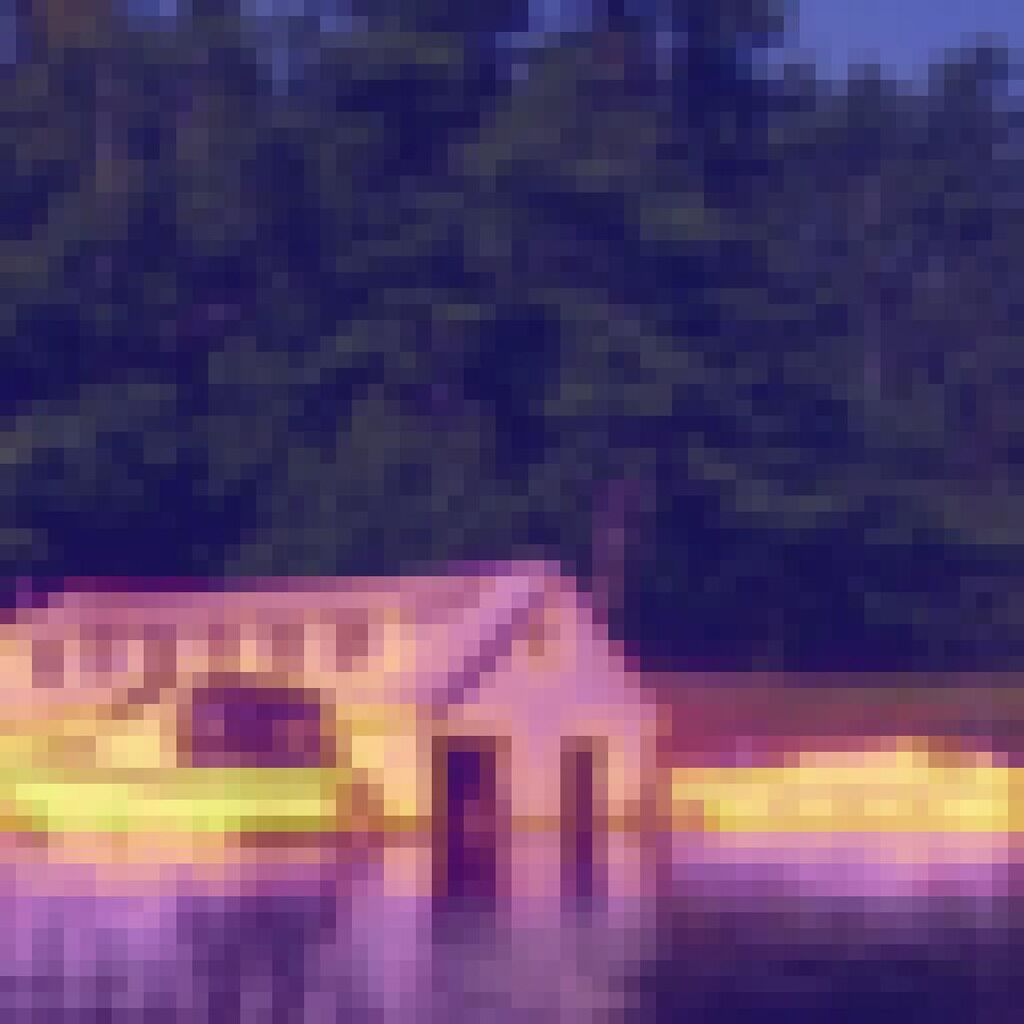}} &
    \fcolorbox{blue}{white}{\includegraphics[width=0.95\linewidth]{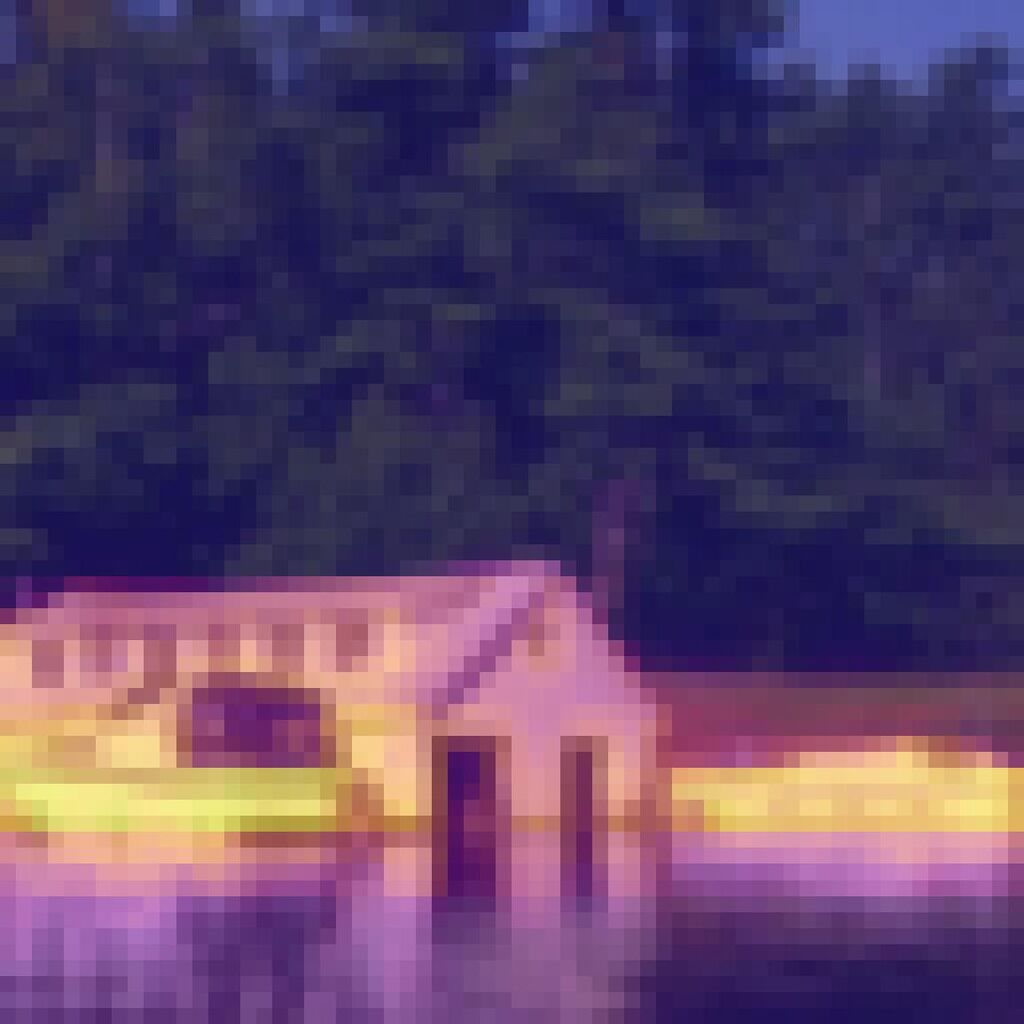}} &
    \fcolorbox{blue}{white}{\includegraphics[width=0.95\linewidth]{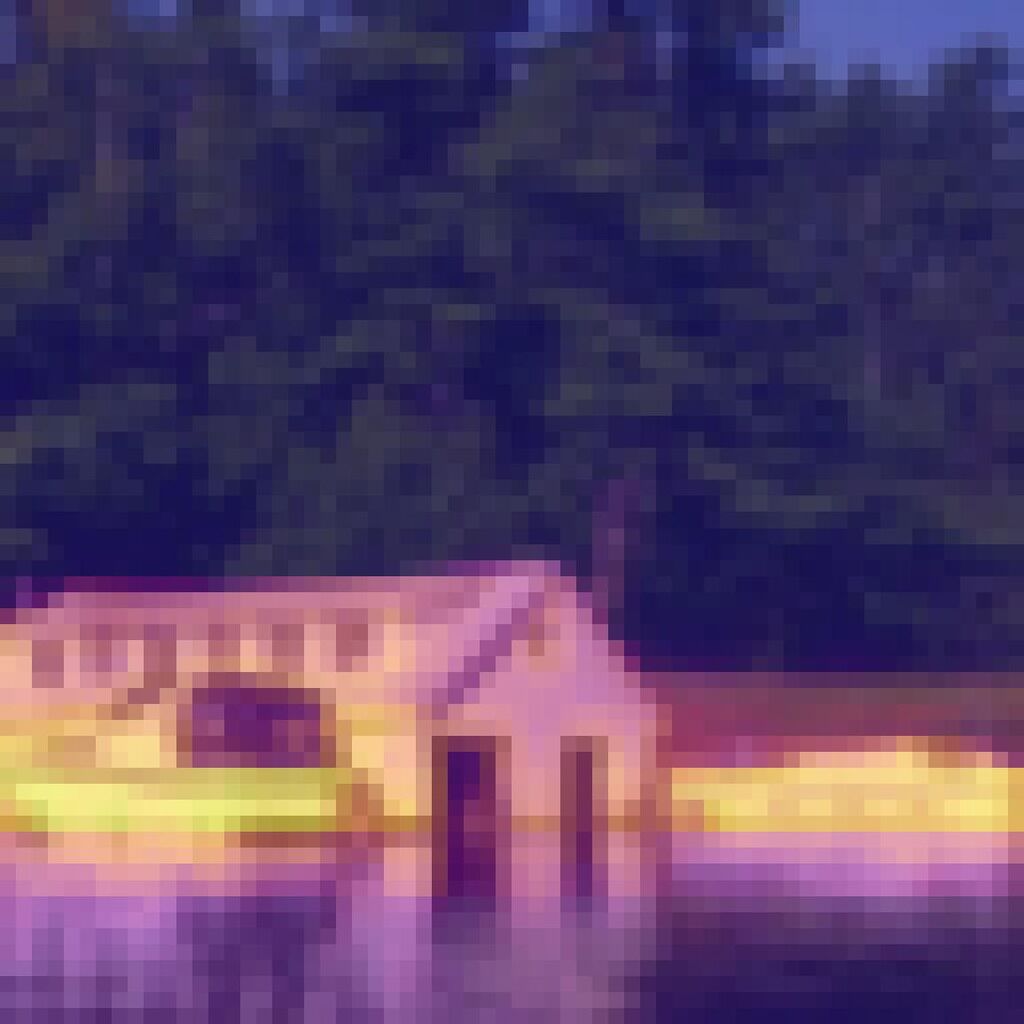}} &
    \fcolorbox{blue}{white}{\includegraphics[width=0.95\linewidth]{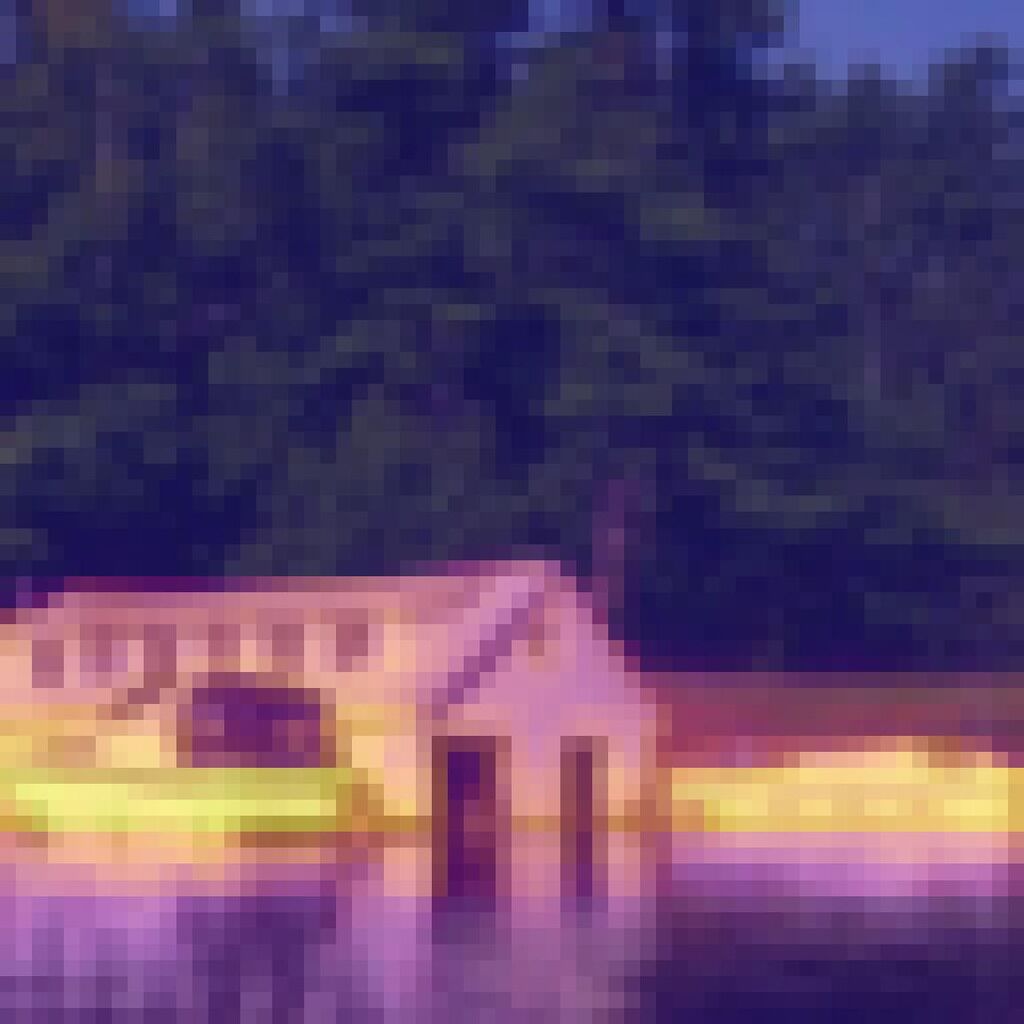}} &
    \fcolorbox{blue}{white}{\includegraphics[width=0.95\linewidth]{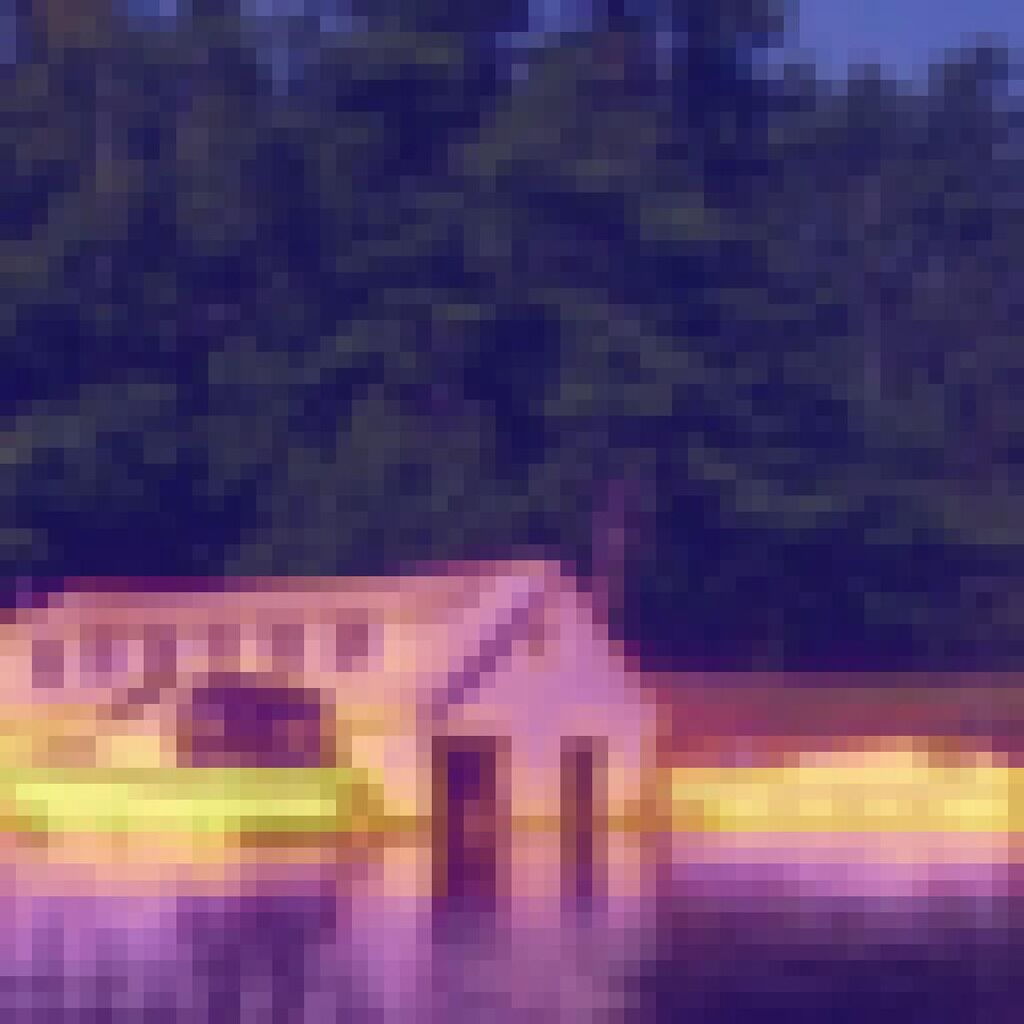}} &
    \fcolorbox{blue}{white}{\includegraphics[width=0.95\linewidth]{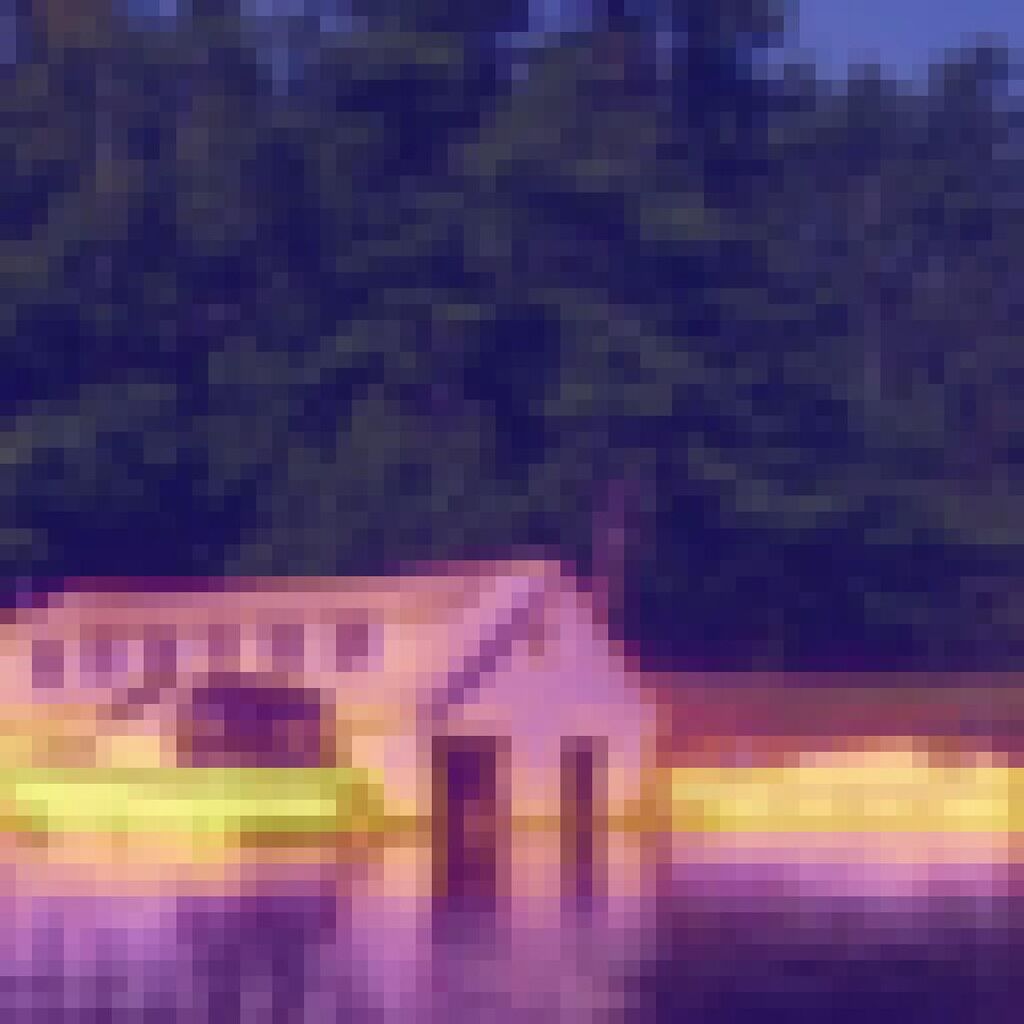}} &
    \fcolorbox{blue}{white}{\includegraphics[width=0.95\linewidth]{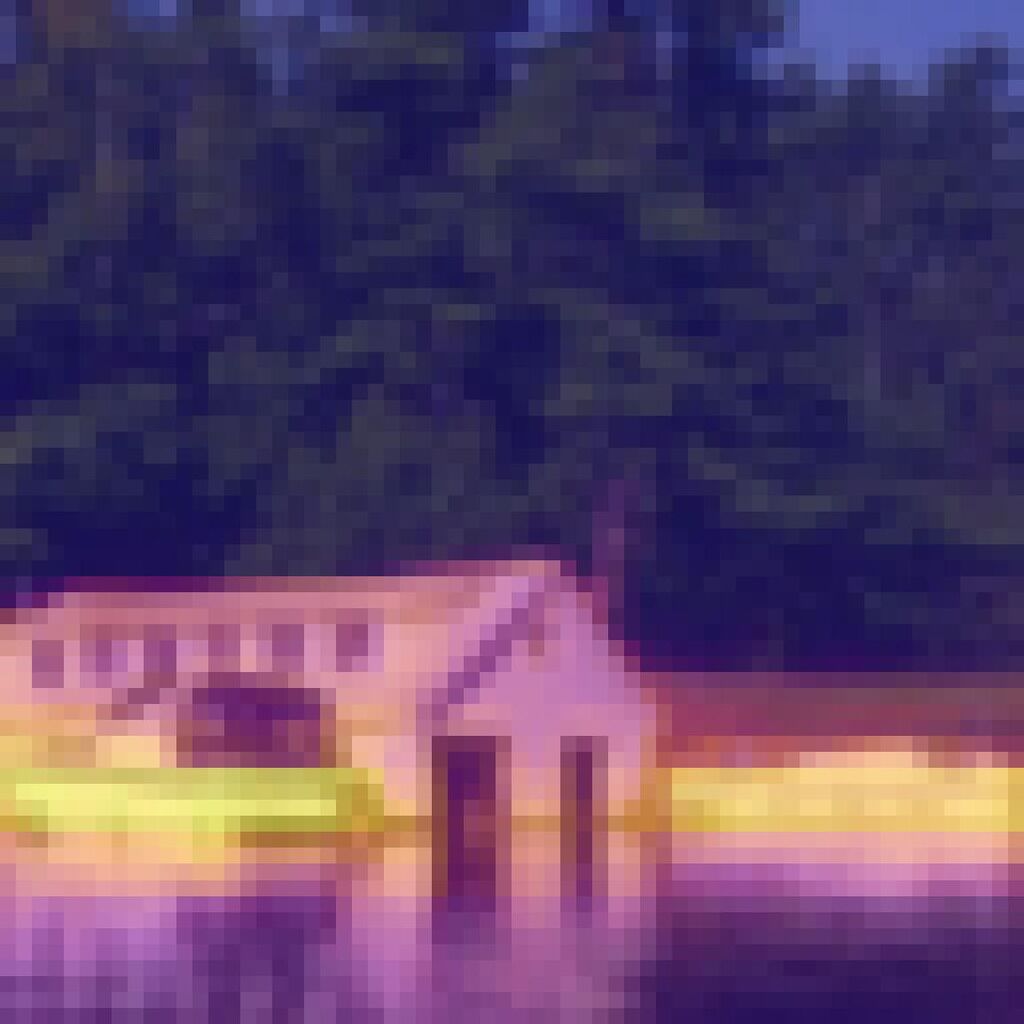}} &
    \fcolorbox{blue}{white}{\includegraphics[width=0.95\linewidth]{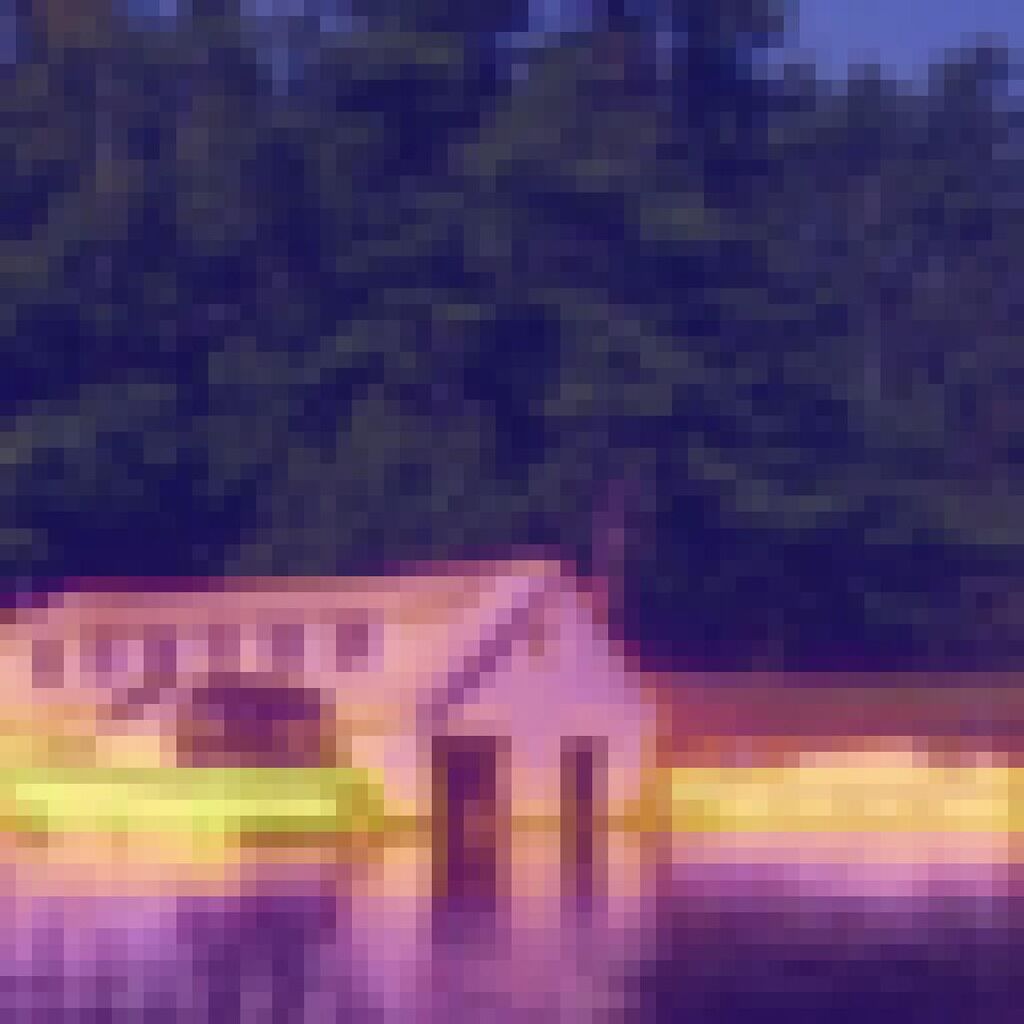}} &
    \fcolorbox{blue}{white}{\includegraphics[width=0.95\linewidth]{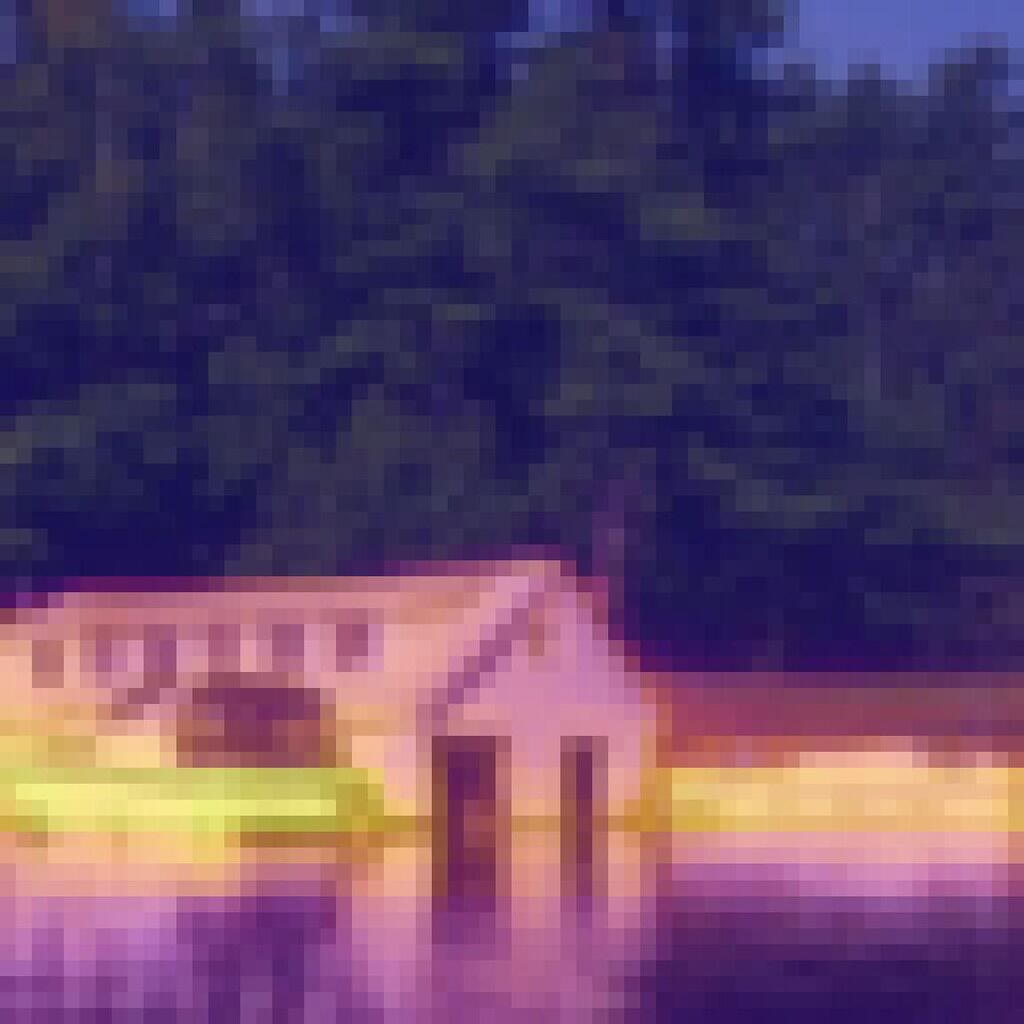}} \\

    t=10 & t=20 & t=30 & t=40 & t=50 & t=100 & t=150 & t=200 & t=250 & t=300 & t=350 \\[1em]
    
    \fcolorbox{blue}{white}{\includegraphics[width=0.95\linewidth]{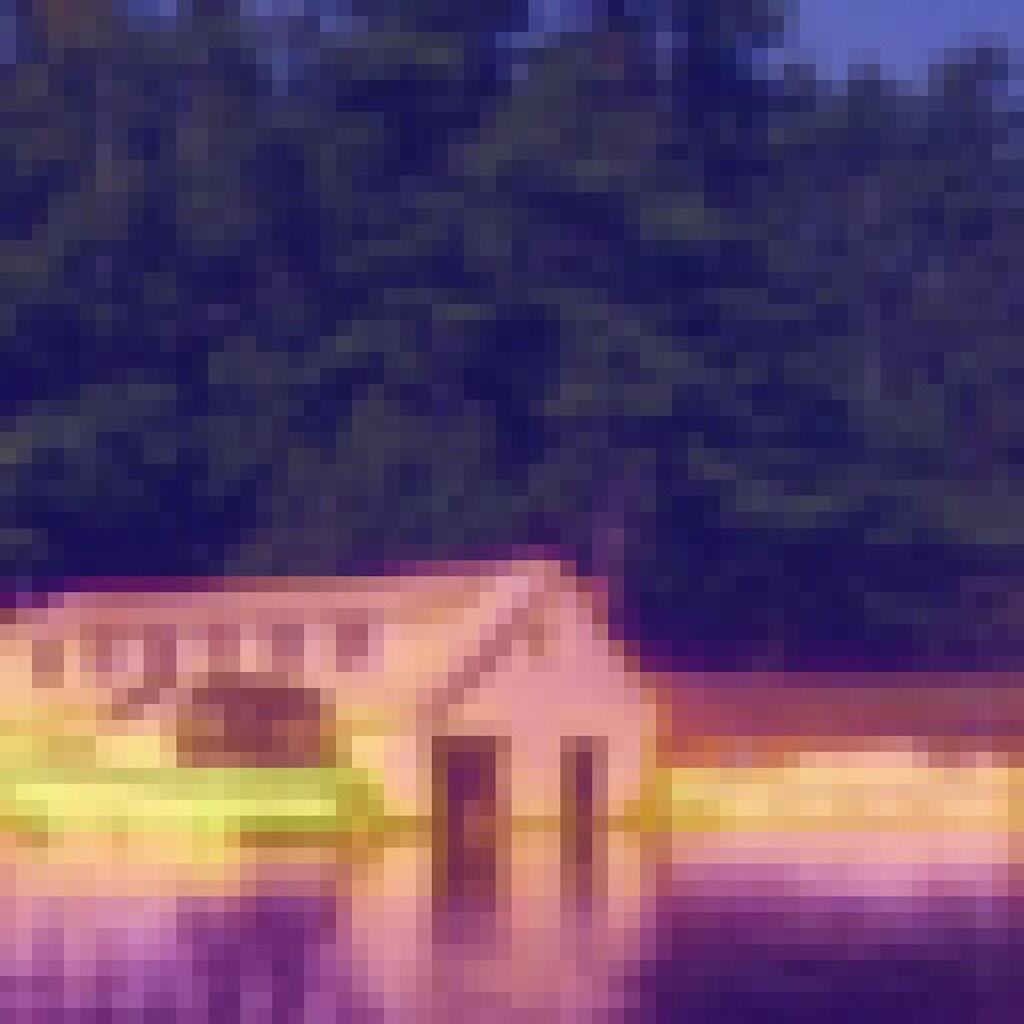}} &
    \fcolorbox{blue}{white}{\includegraphics[width=0.95\linewidth]{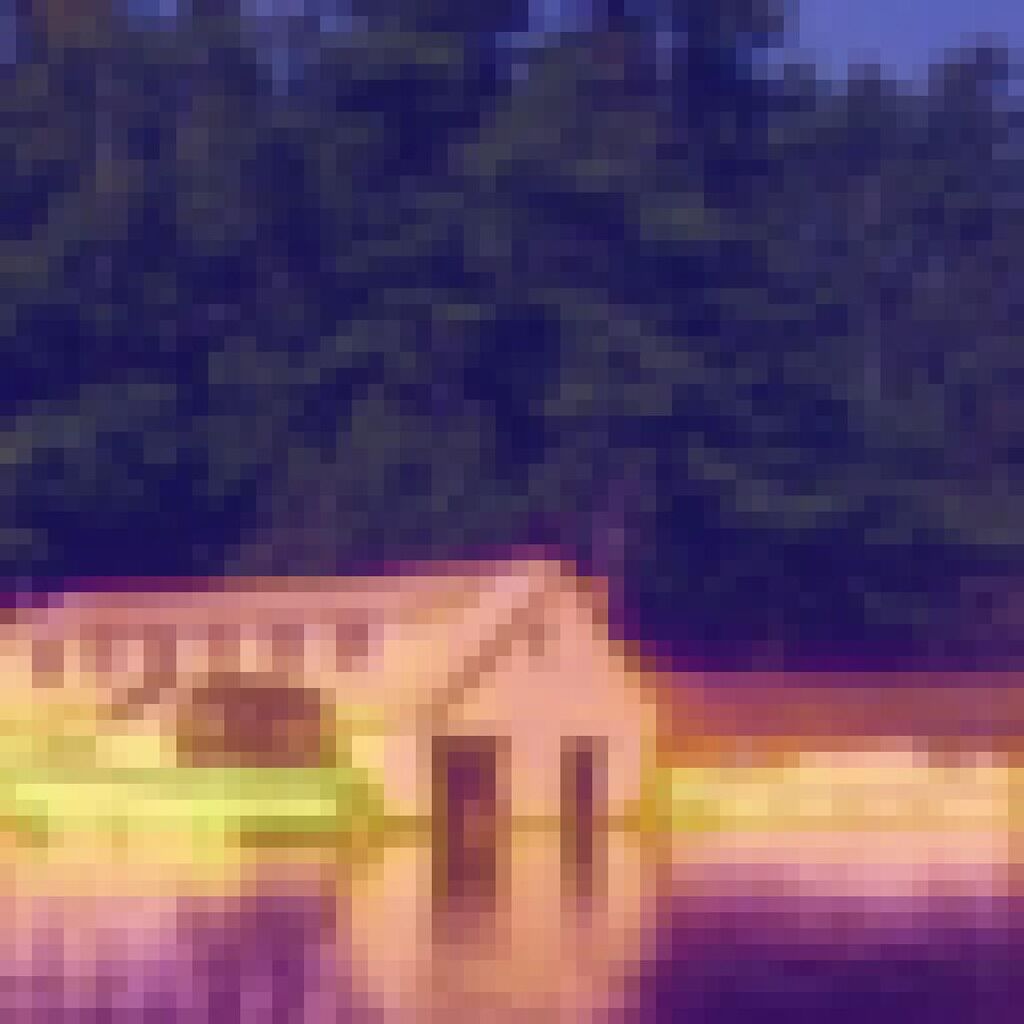}} &
    \fcolorbox{blue}{white}{\includegraphics[width=0.95\linewidth]{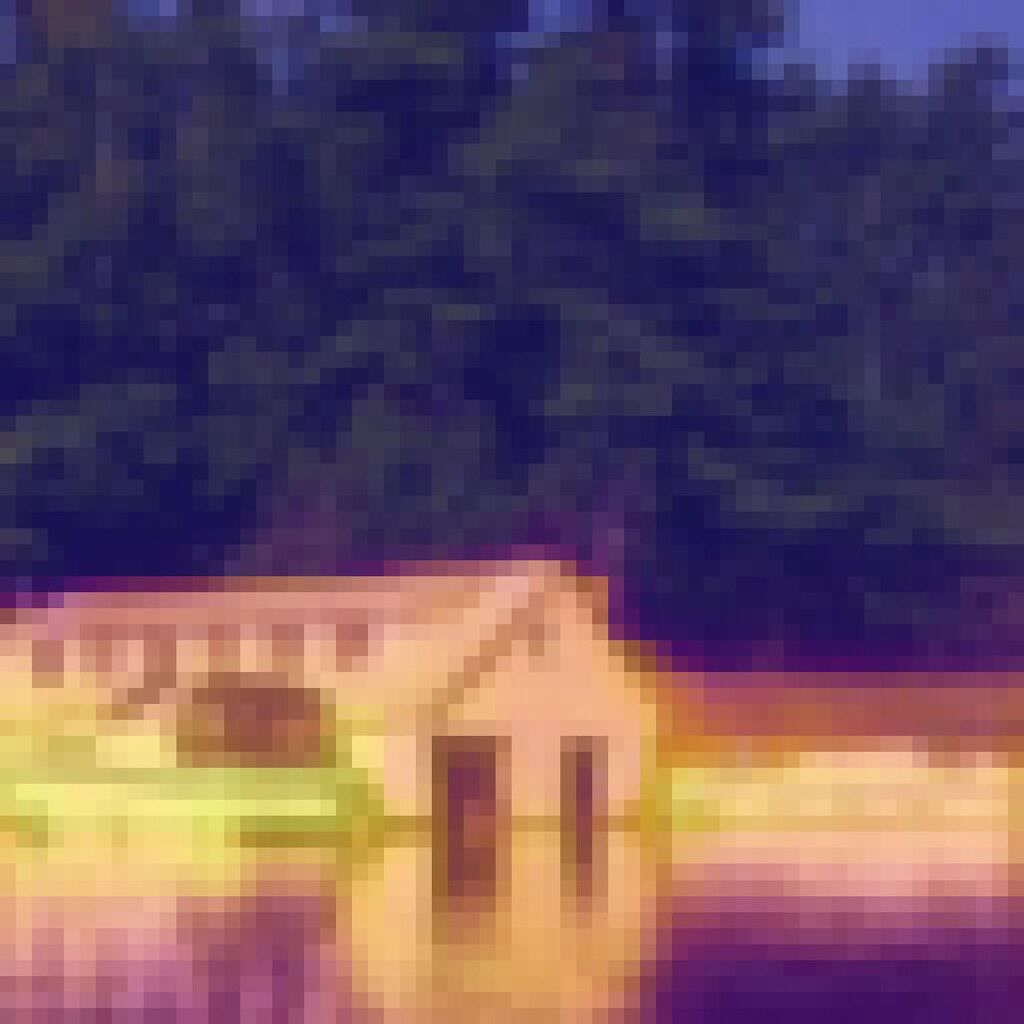}} &
    \fcolorbox{blue}{white}{\includegraphics[width=0.95\linewidth]{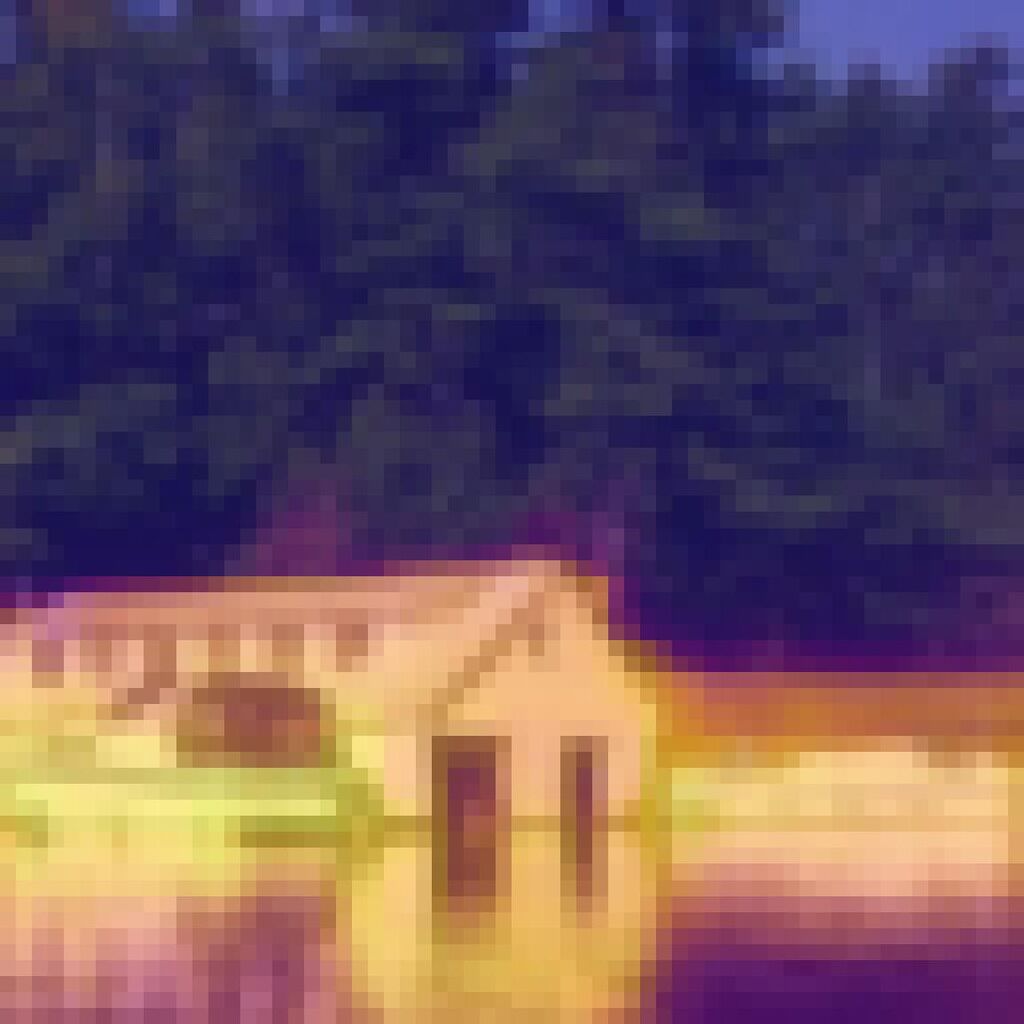}} &
    \fcolorbox{blue}{white}{\includegraphics[width=0.95\linewidth]{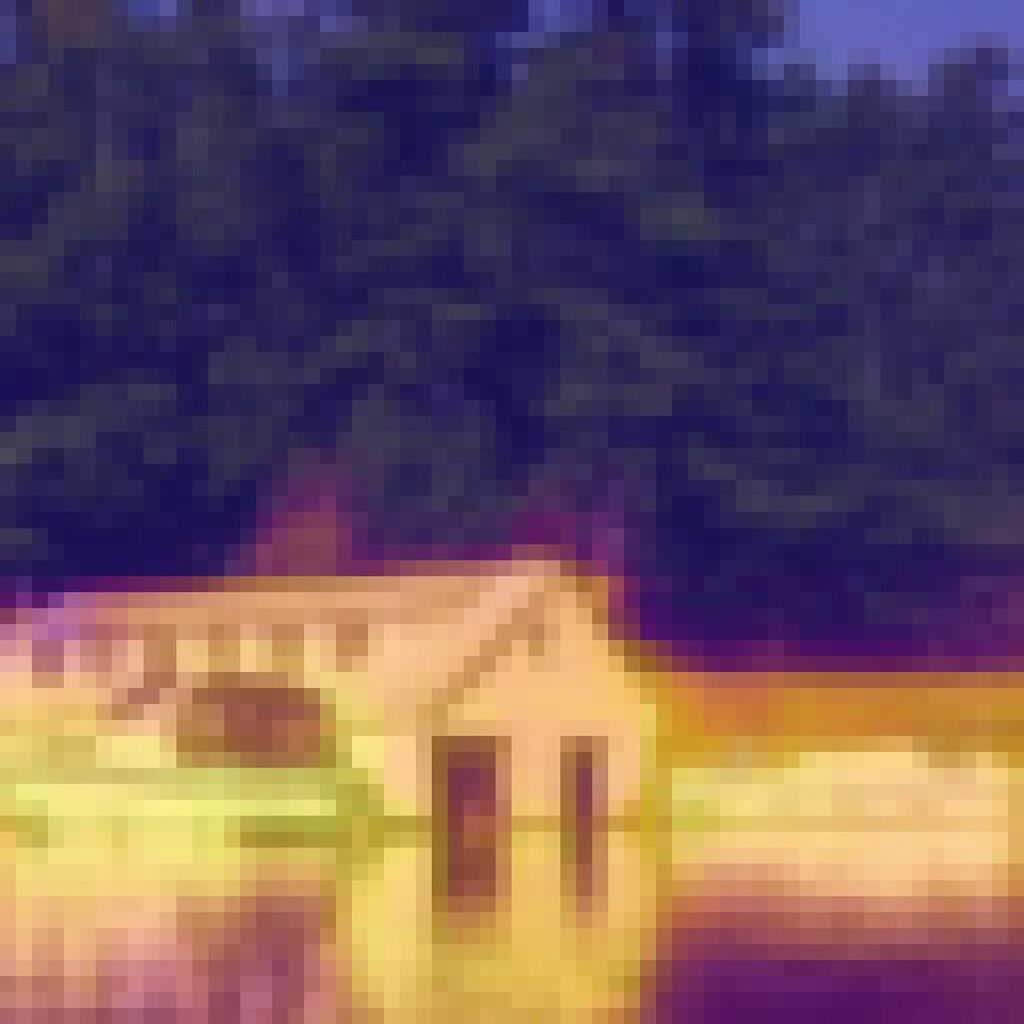}} &
    \fcolorbox{blue}{white}{\includegraphics[width=0.95\linewidth]{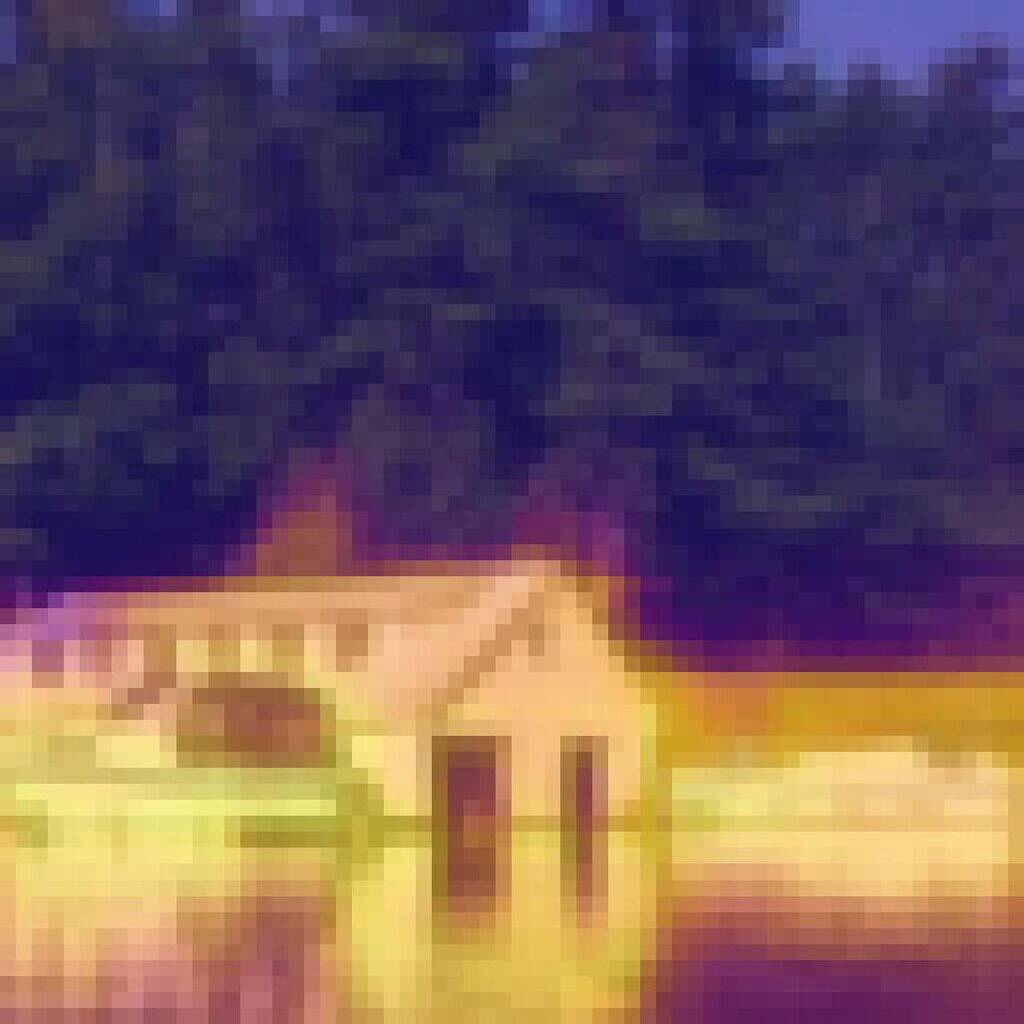}} &
    \fcolorbox{blue}{white}{\includegraphics[width=0.95\linewidth]{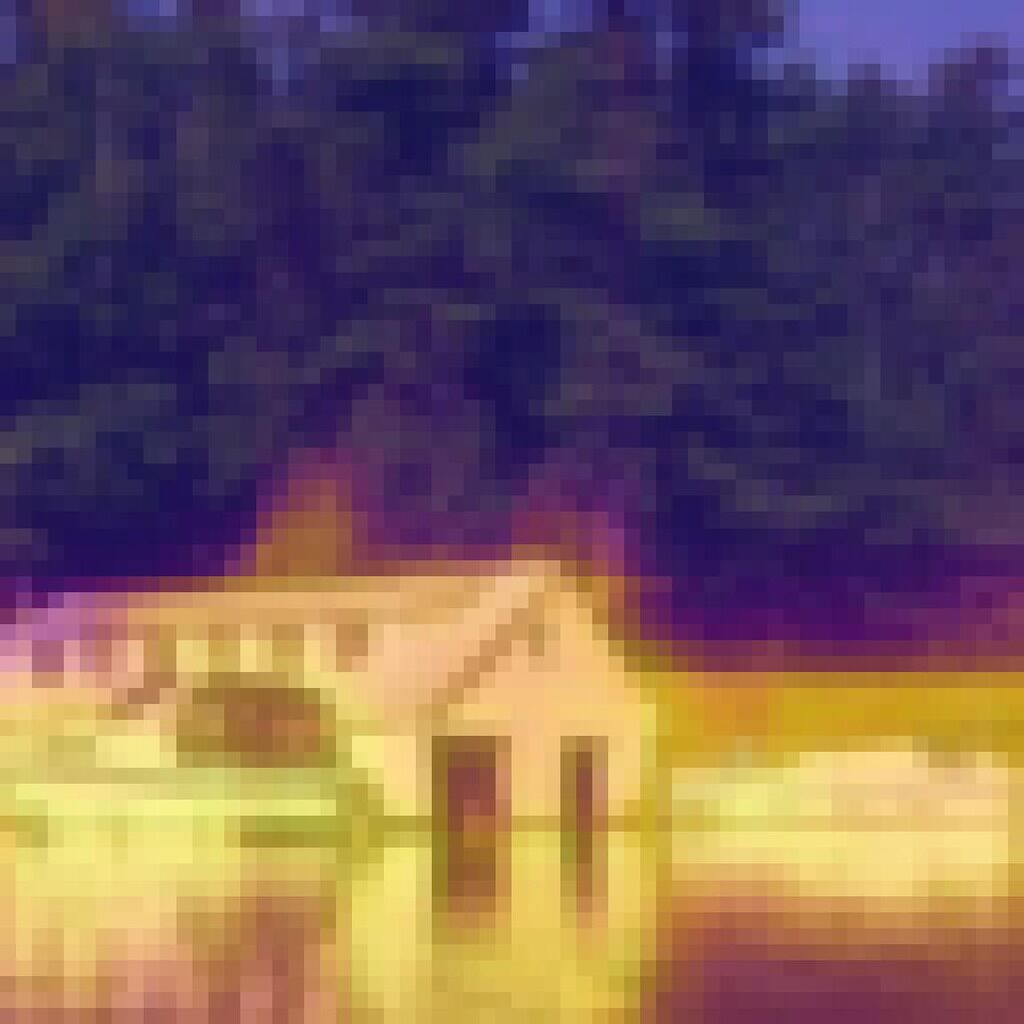}} &
    \fcolorbox{blue}{white}{\includegraphics[width=0.95\linewidth]{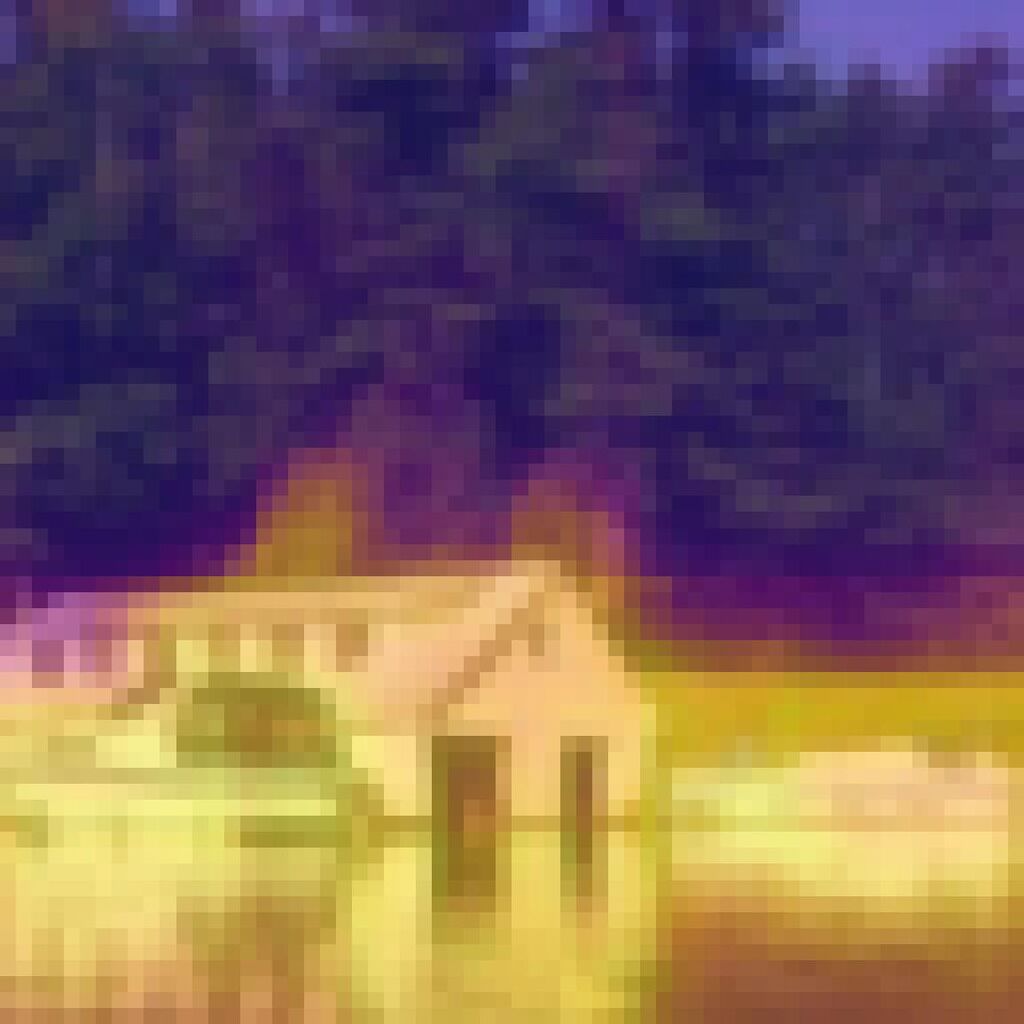}} &
    \fcolorbox{blue}{white}{\includegraphics[width=0.95\linewidth]{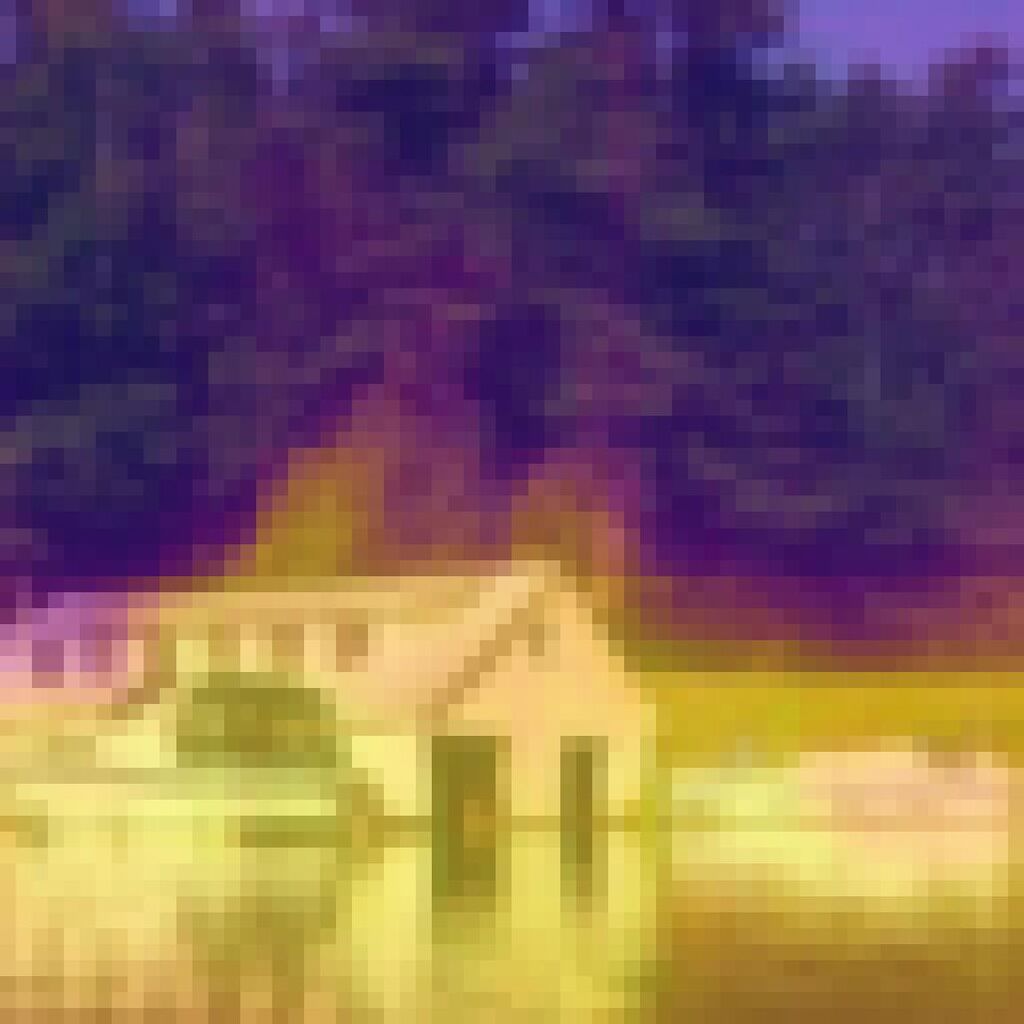}} &
    \fcolorbox{blue}{white}{\includegraphics[width=0.95\linewidth]{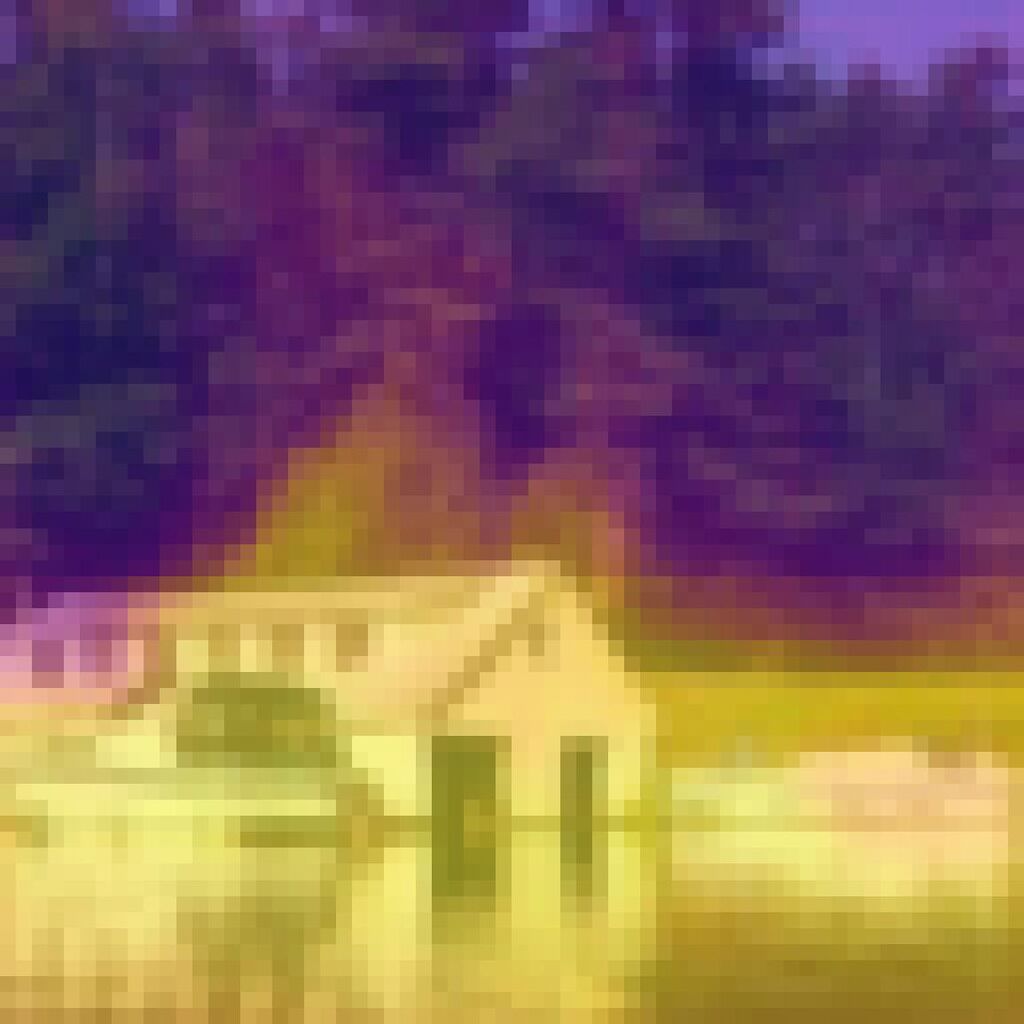}} &
    \fcolorbox{blue}{white}{\includegraphics[width=0.95\linewidth]{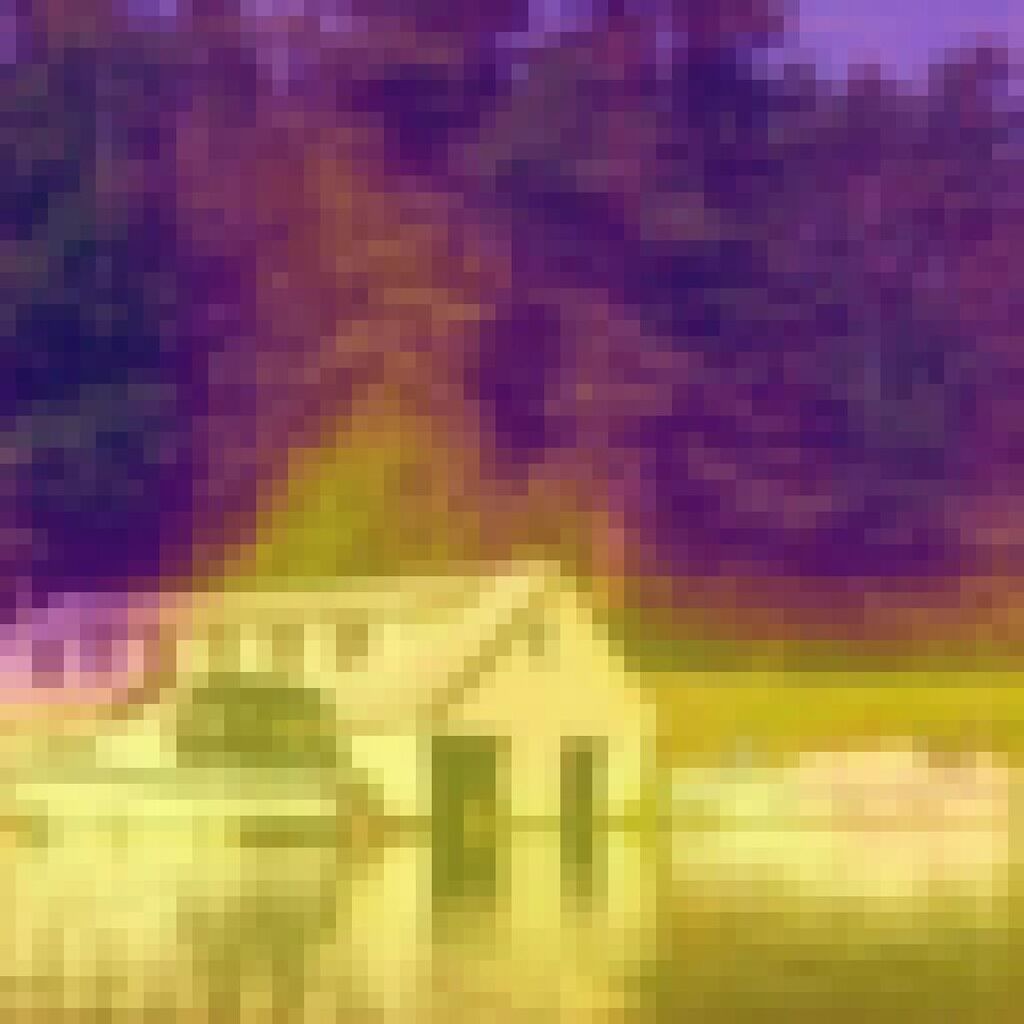}} \\

    t=400 & t=450 & t=500 & t=550 & t=600 & t=650 & t=700 & t=750 & t=800 & t=850 & t=900 \\
    
    \end{tabularx}
    
    \caption{The top row displays the original input image and the corresponding Temporal Stability Matrices (TSMs) for both red and blue query point. The panels below show the evolution of their corresponding Contextual Similarity Maps (CSMs).}

    \label{fig:supp_hierarchical_progress_sample3}
\end{sidewaysfigure*}
\begin{sidewaysfigure*}[htb!] 
    \centering
    \begin{tabular}{ccc}
        \includegraphics[width=0.2\linewidth]{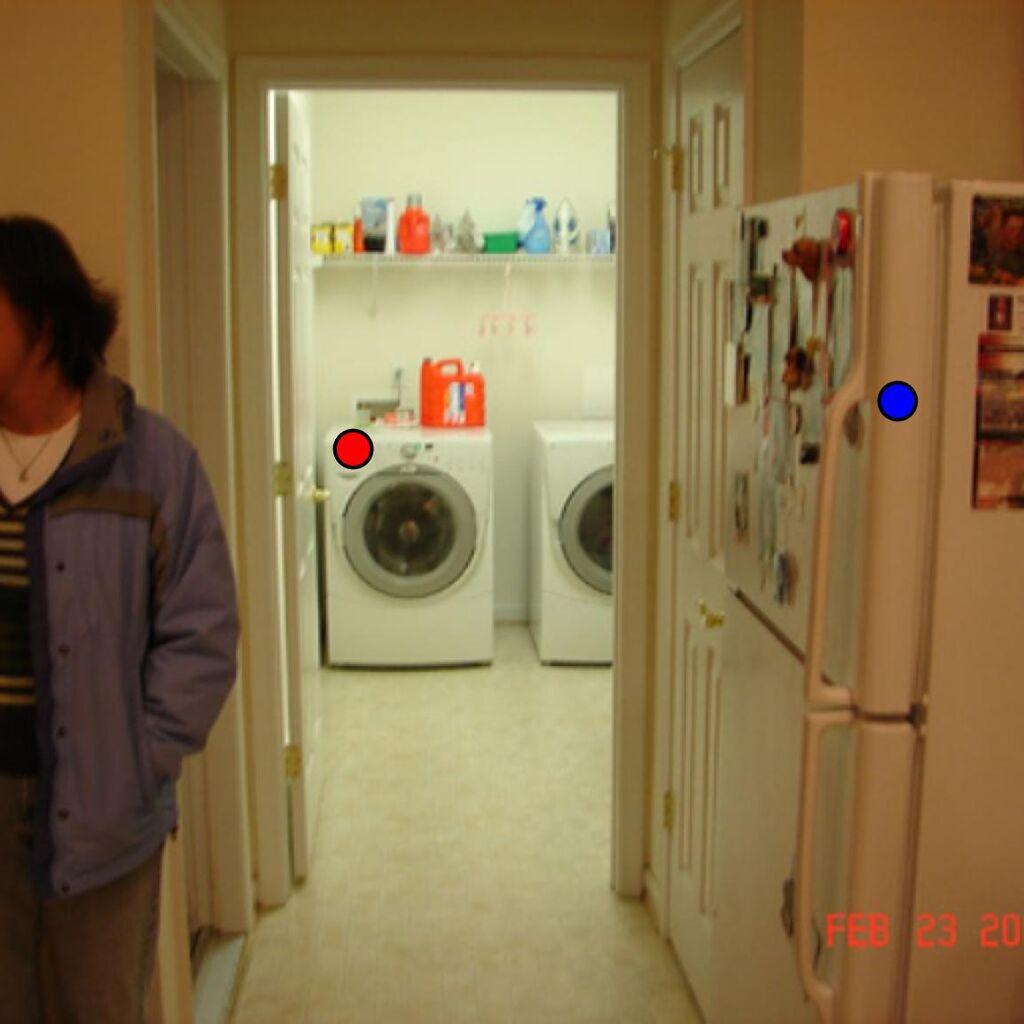}  & \includegraphics[width=0.2\linewidth]{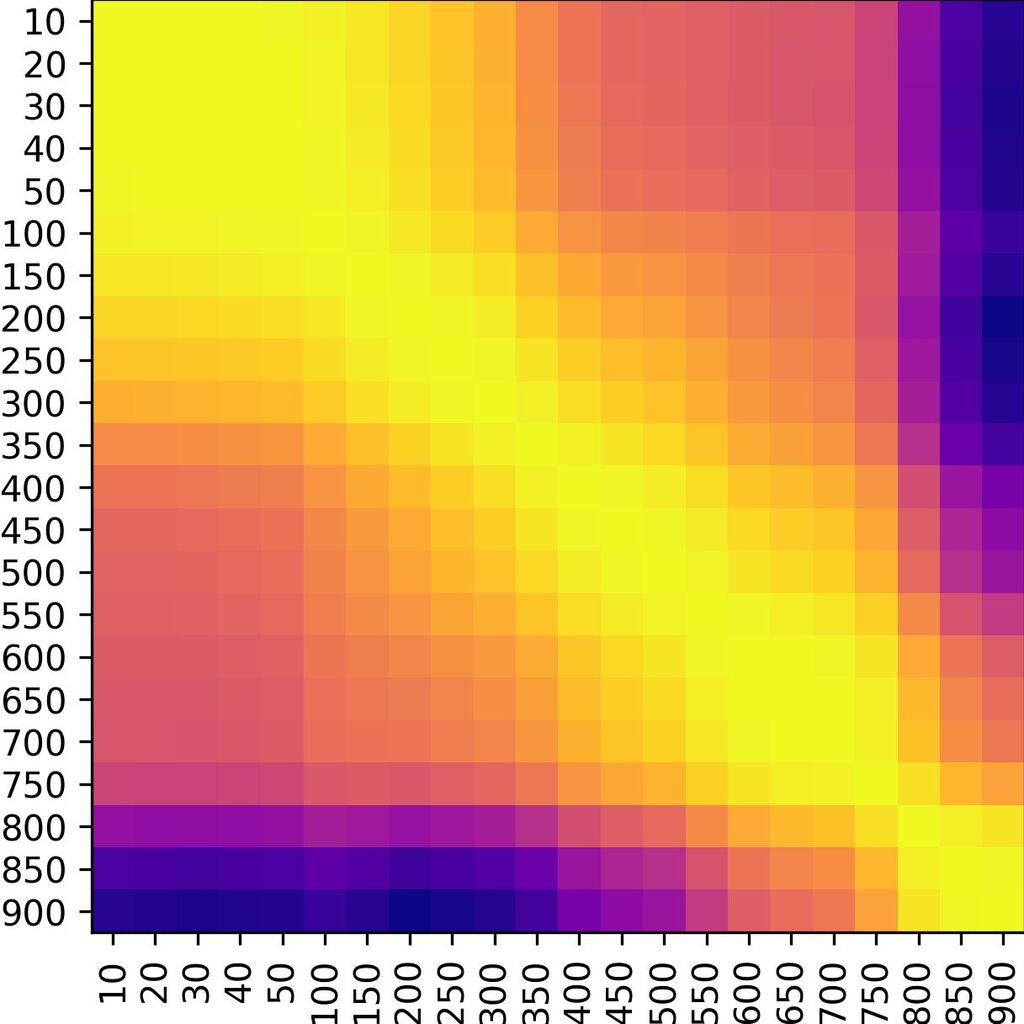} &
        \includegraphics[width=0.2\linewidth]{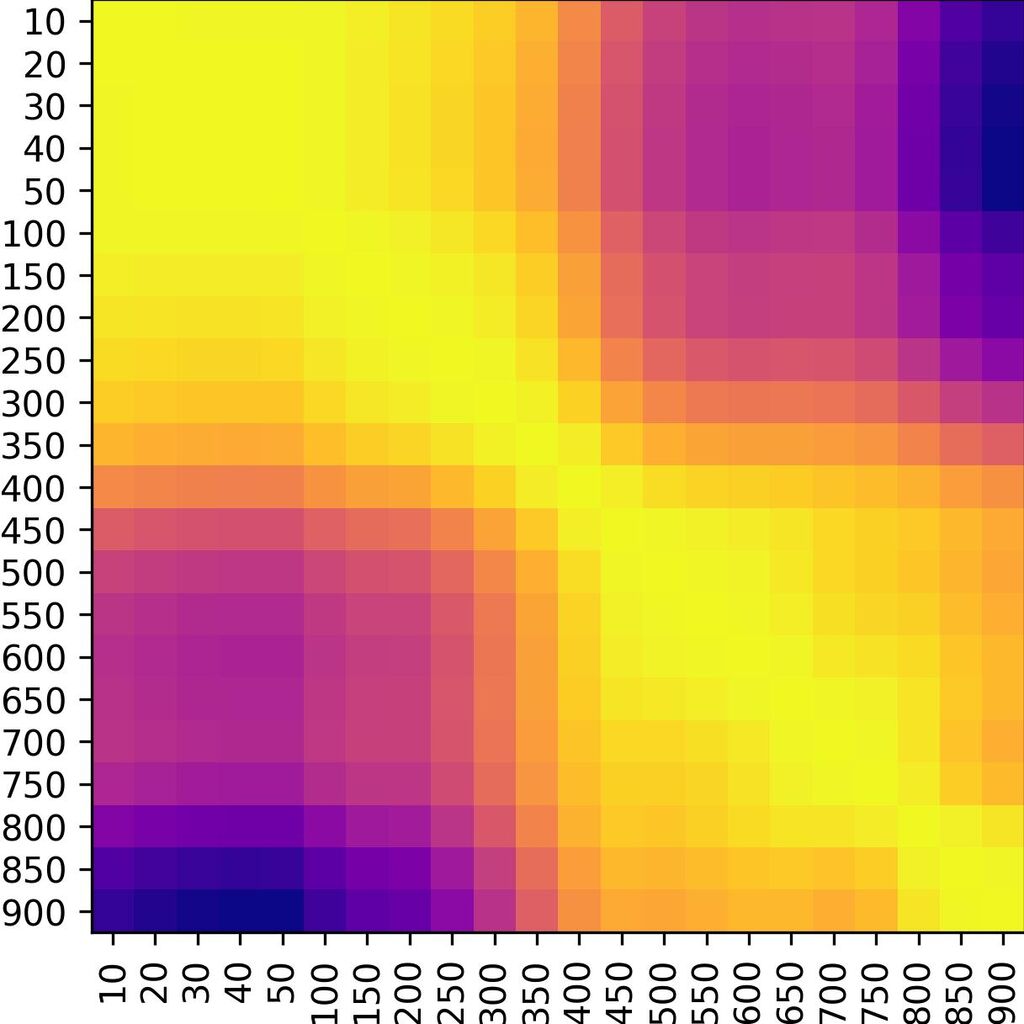}\\
        Input Image & Red Point TSM & Blue Point TSM
    \end{tabular}
    \vspace{1em} 

    \setlength{\tabcolsep}{2pt} 
    \begin{tabularx}{\linewidth}{ *{11}{>{\centering\arraybackslash}p{0.0845\linewidth}} } 
    
    \multicolumn{11}{c}{\textbf{Red Point CSMs}} \\
    
    \fcolorbox{red}{white}{\includegraphics[width=0.95\linewidth]{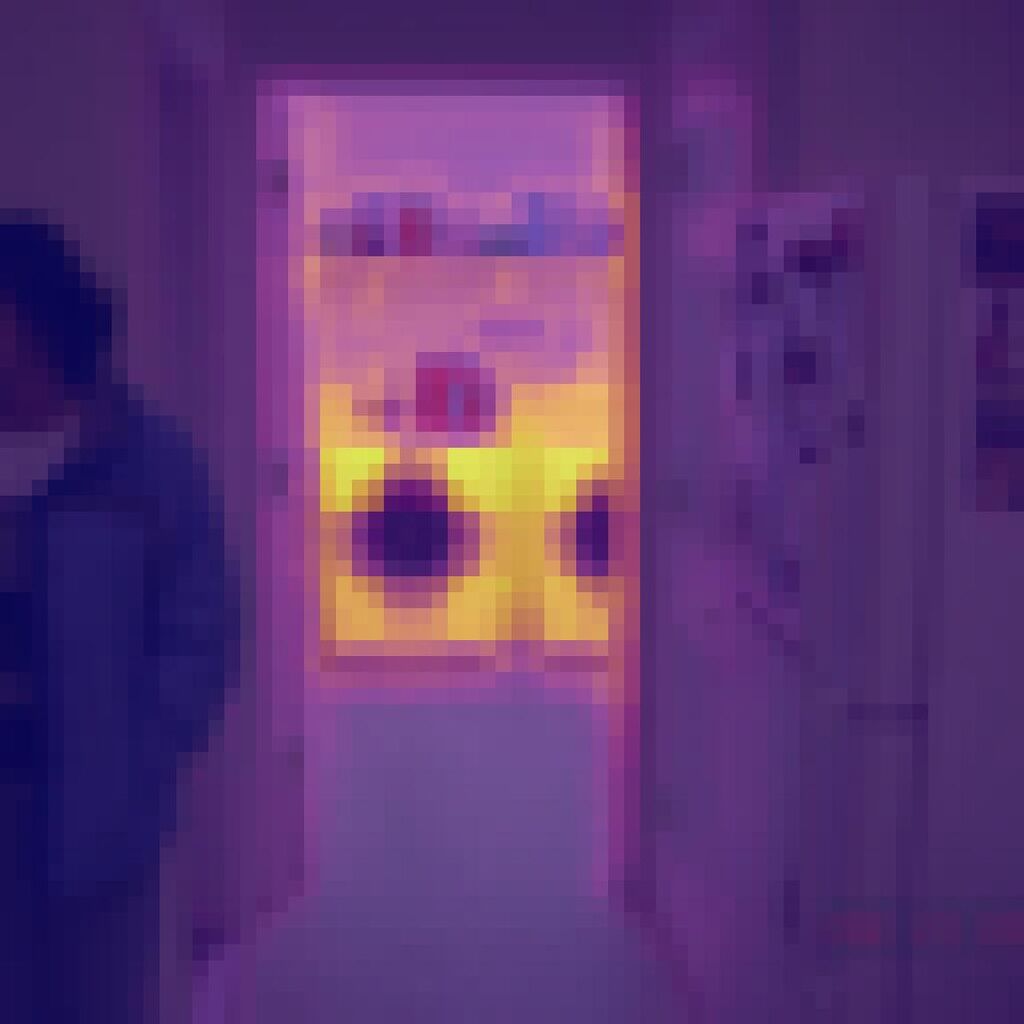}} &
    \fcolorbox{red}{white}{\includegraphics[width=0.95\linewidth]{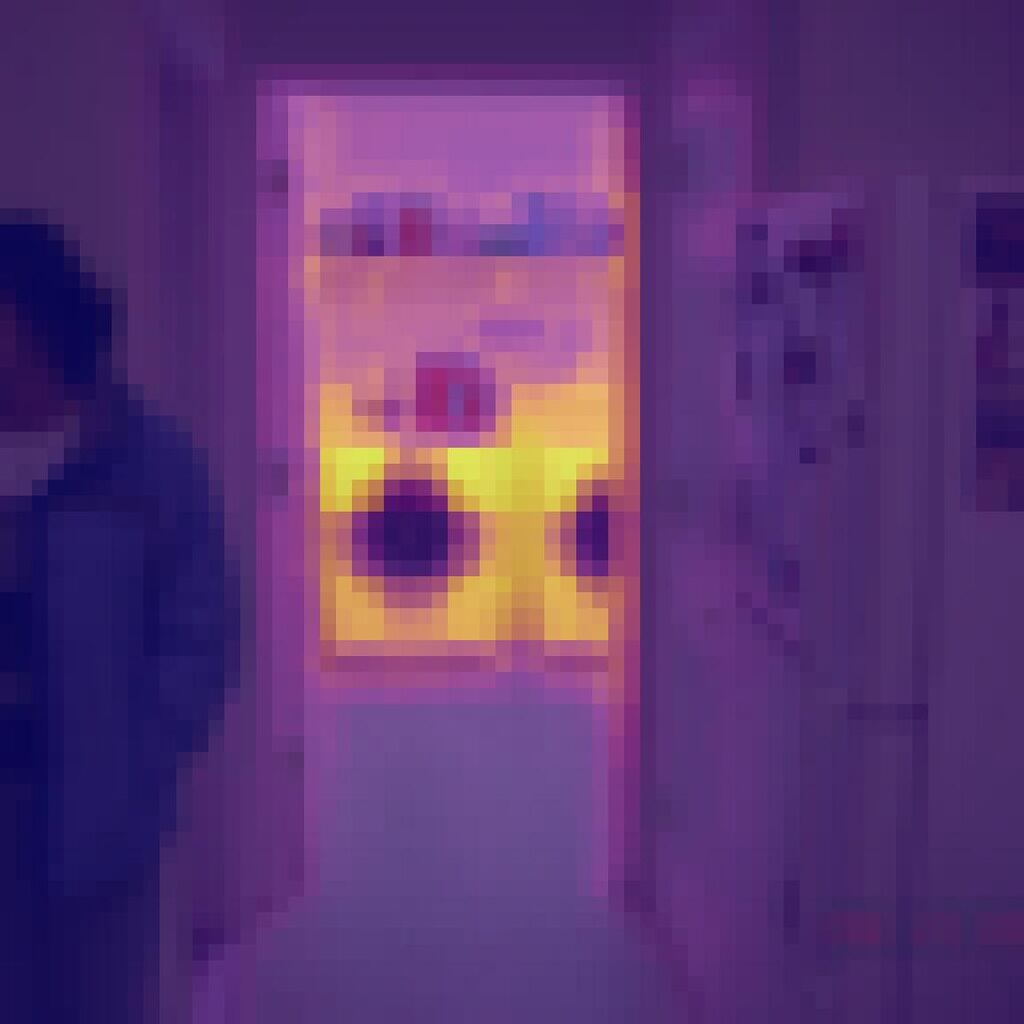}} &
    \fcolorbox{red}{white}{\includegraphics[width=0.95\linewidth]{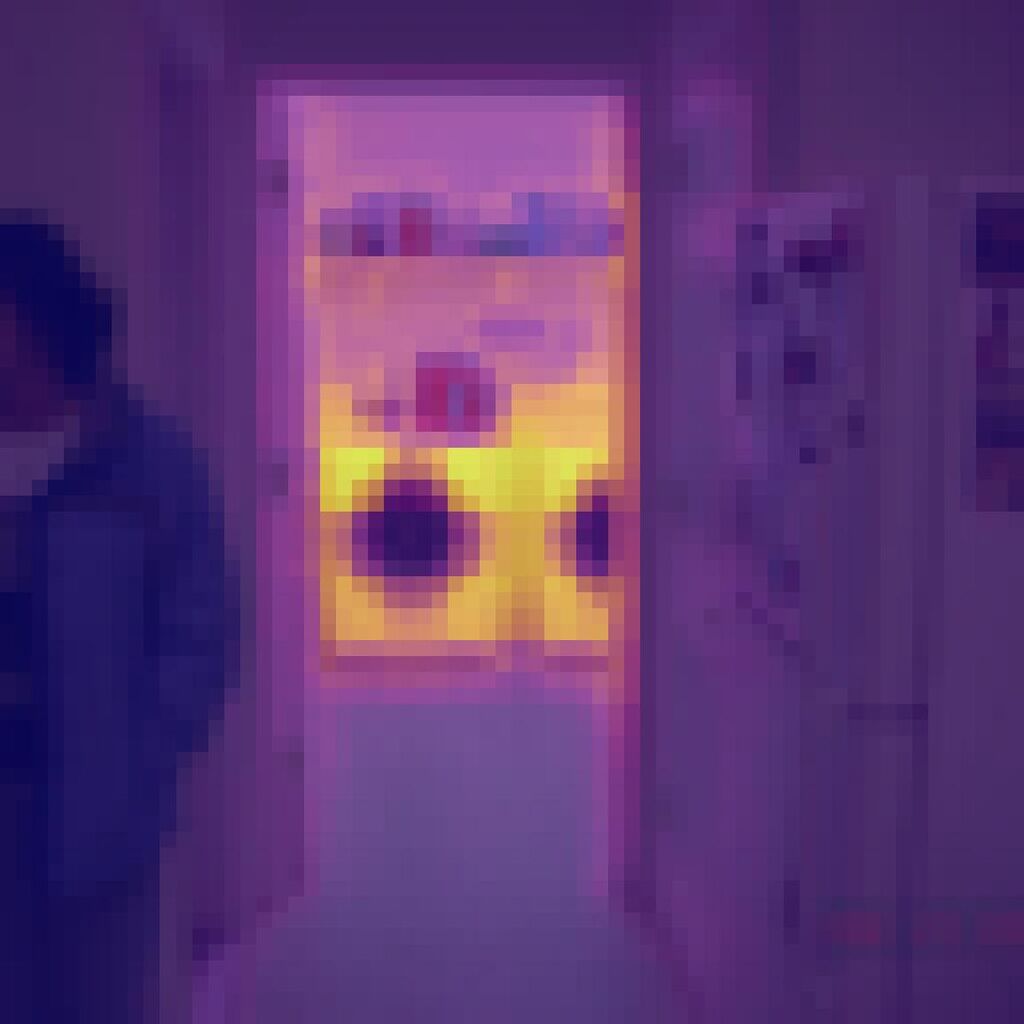}} &
    \fcolorbox{red}{white}{\includegraphics[width=0.95\linewidth]{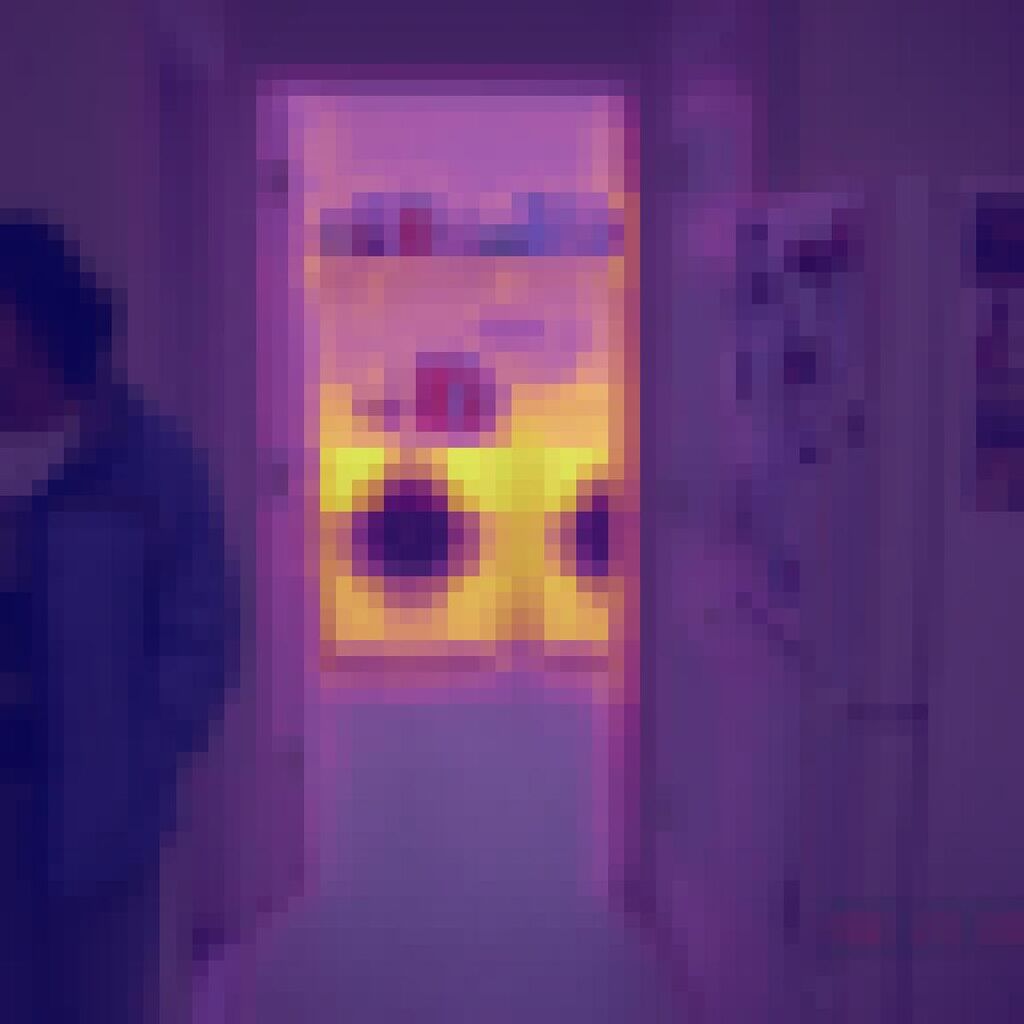}} &
    \fcolorbox{red}{white}{\includegraphics[width=0.95\linewidth]{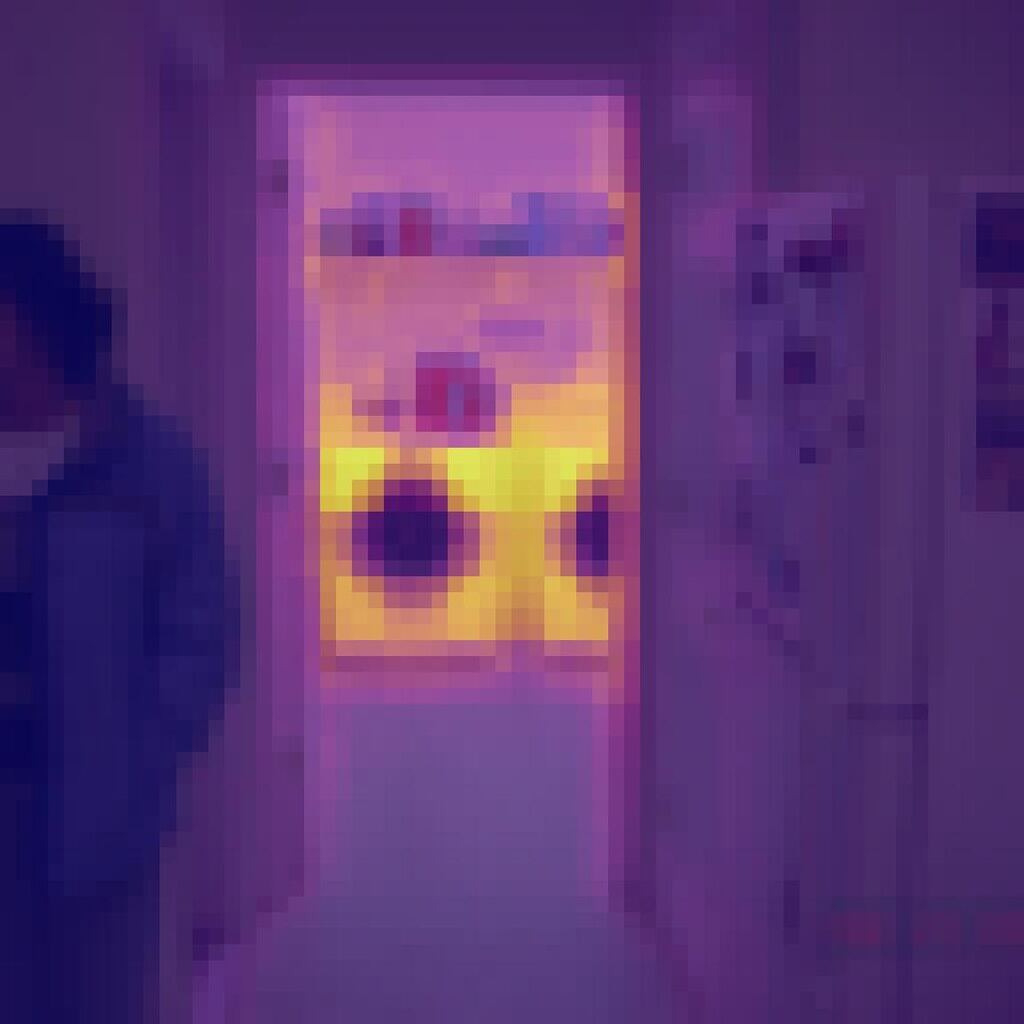}} &
    \fcolorbox{red}{white}{\includegraphics[width=0.95\linewidth]{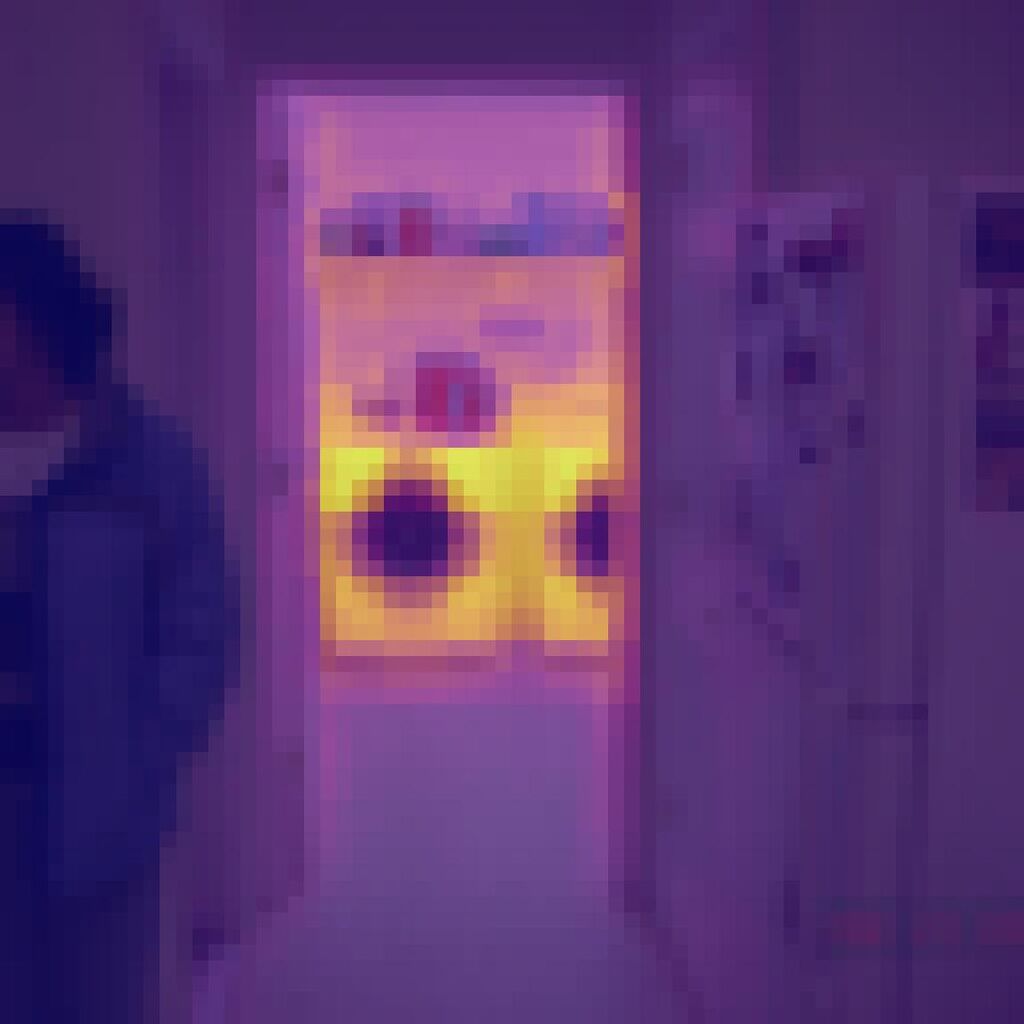}} &
    \fcolorbox{red}{white}{\includegraphics[width=0.95\linewidth]{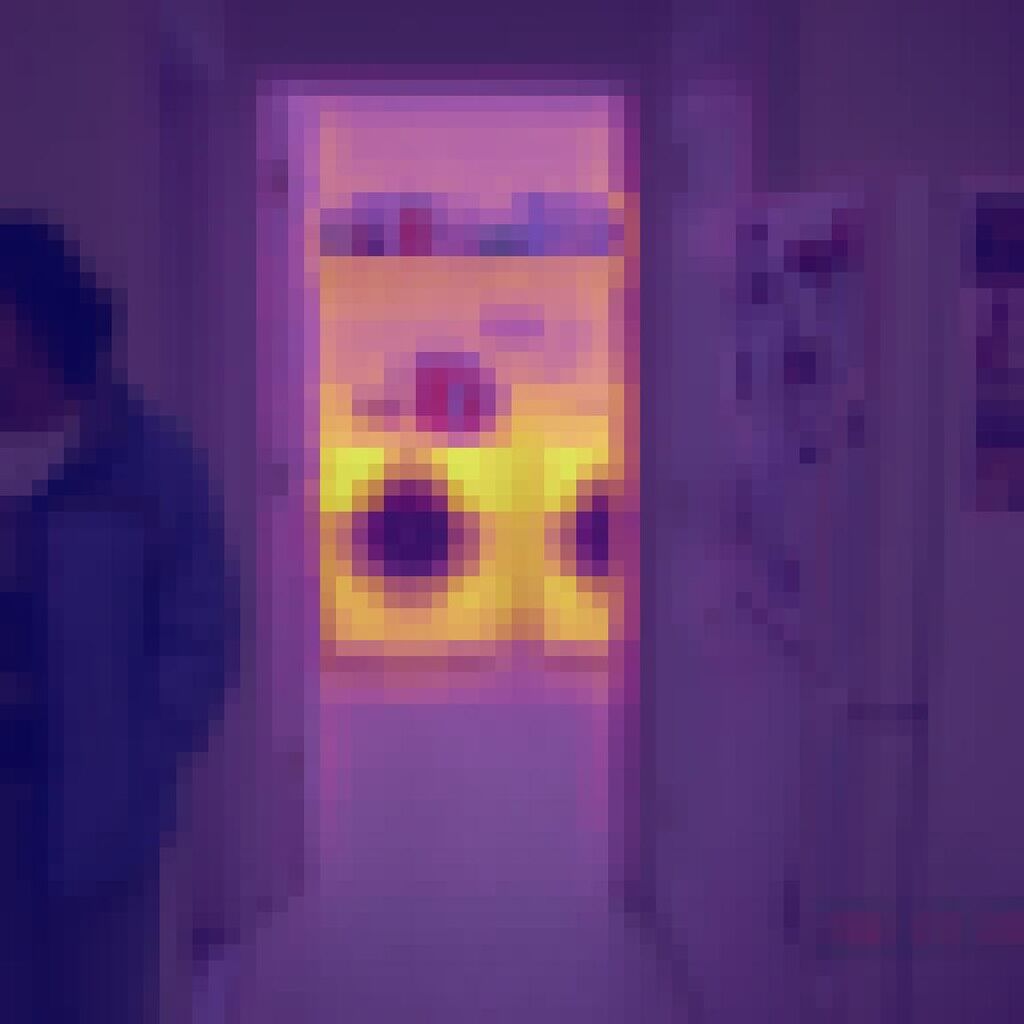}} &
    \fcolorbox{red}{white}{\includegraphics[width=0.95\linewidth]{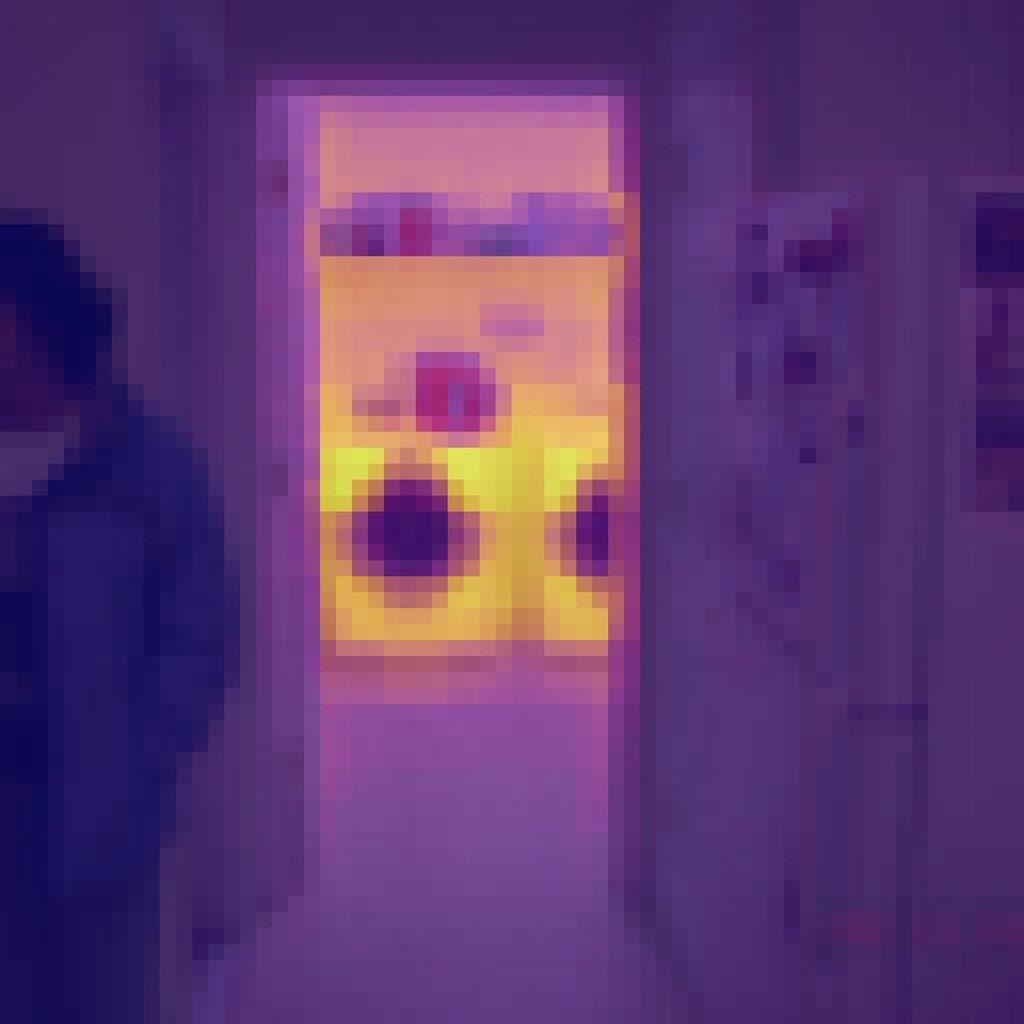}} &
    \fcolorbox{red}{white}{\includegraphics[width=0.95\linewidth]{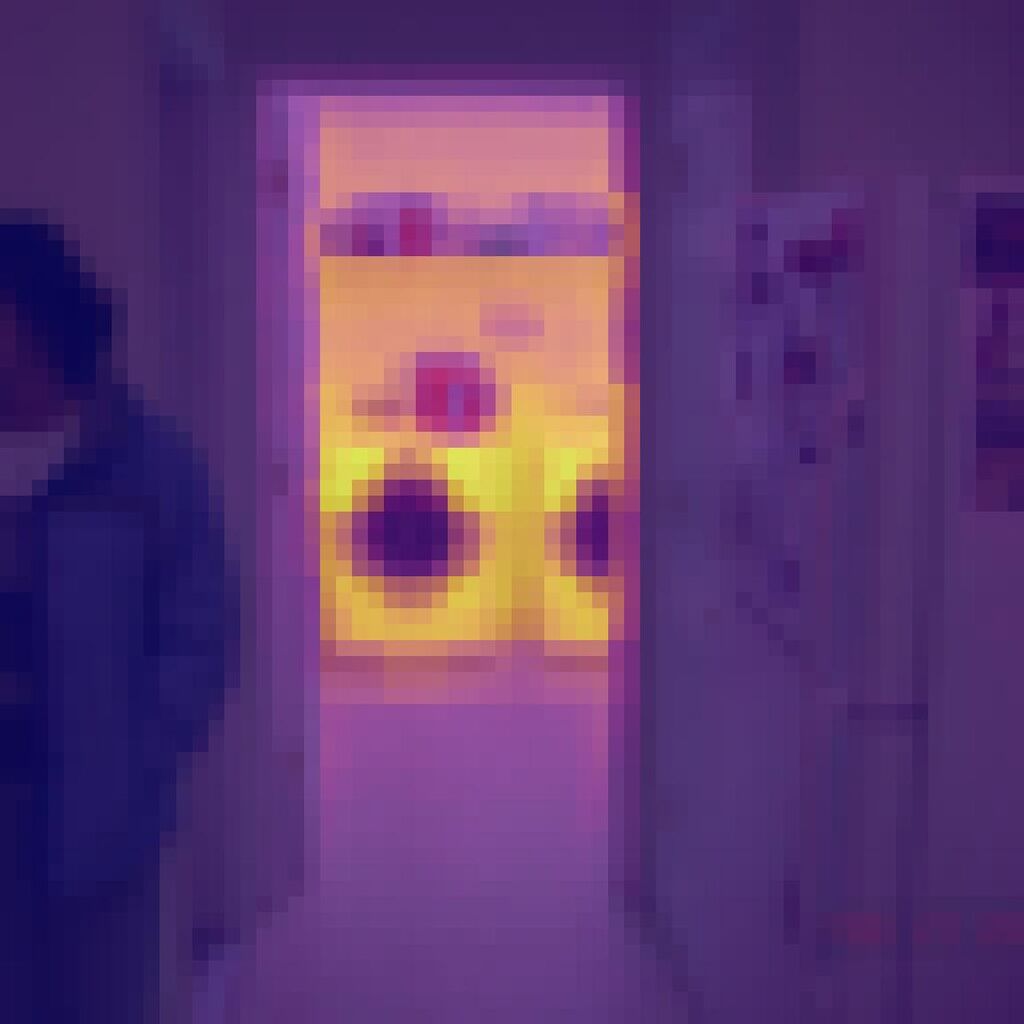}} &
    \fcolorbox{red}{white}{\includegraphics[width=0.95\linewidth]{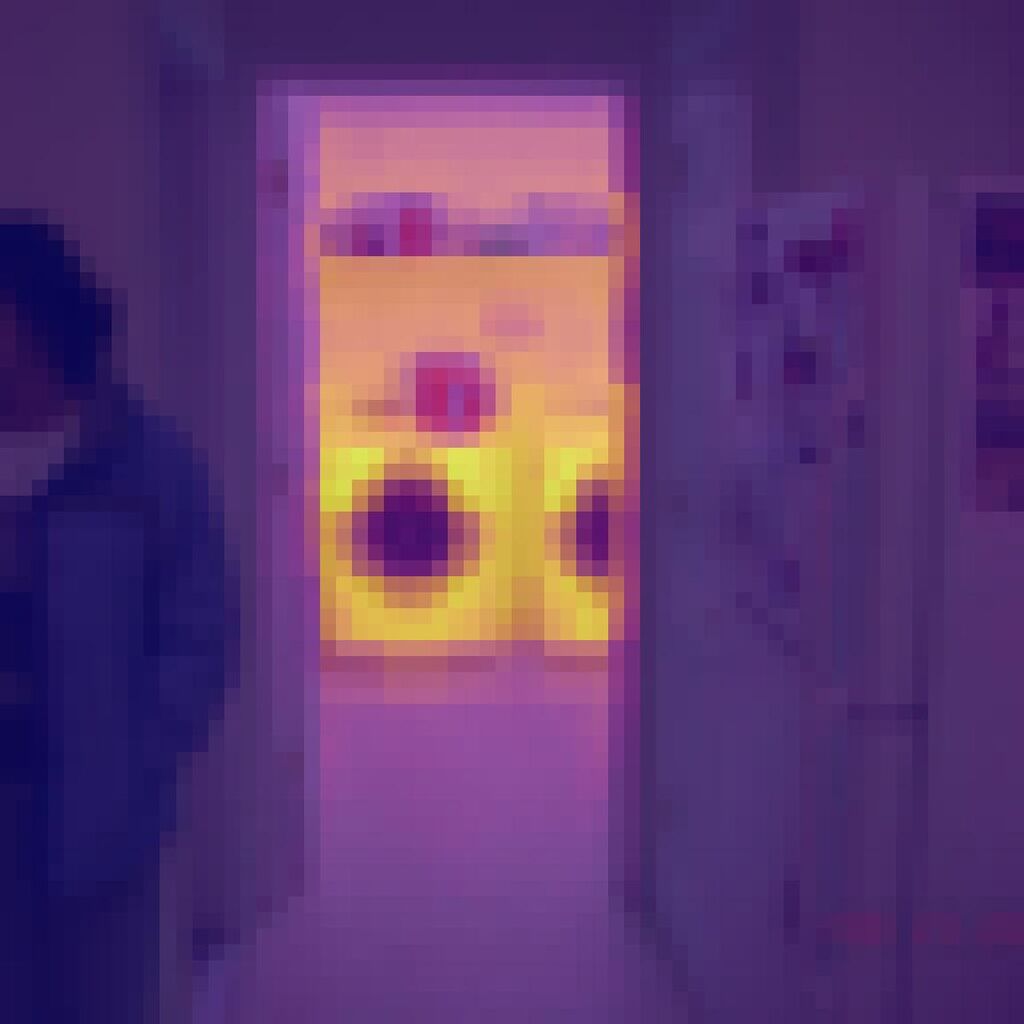}} &
    \fcolorbox{red}{white}{\includegraphics[width=0.95\linewidth]{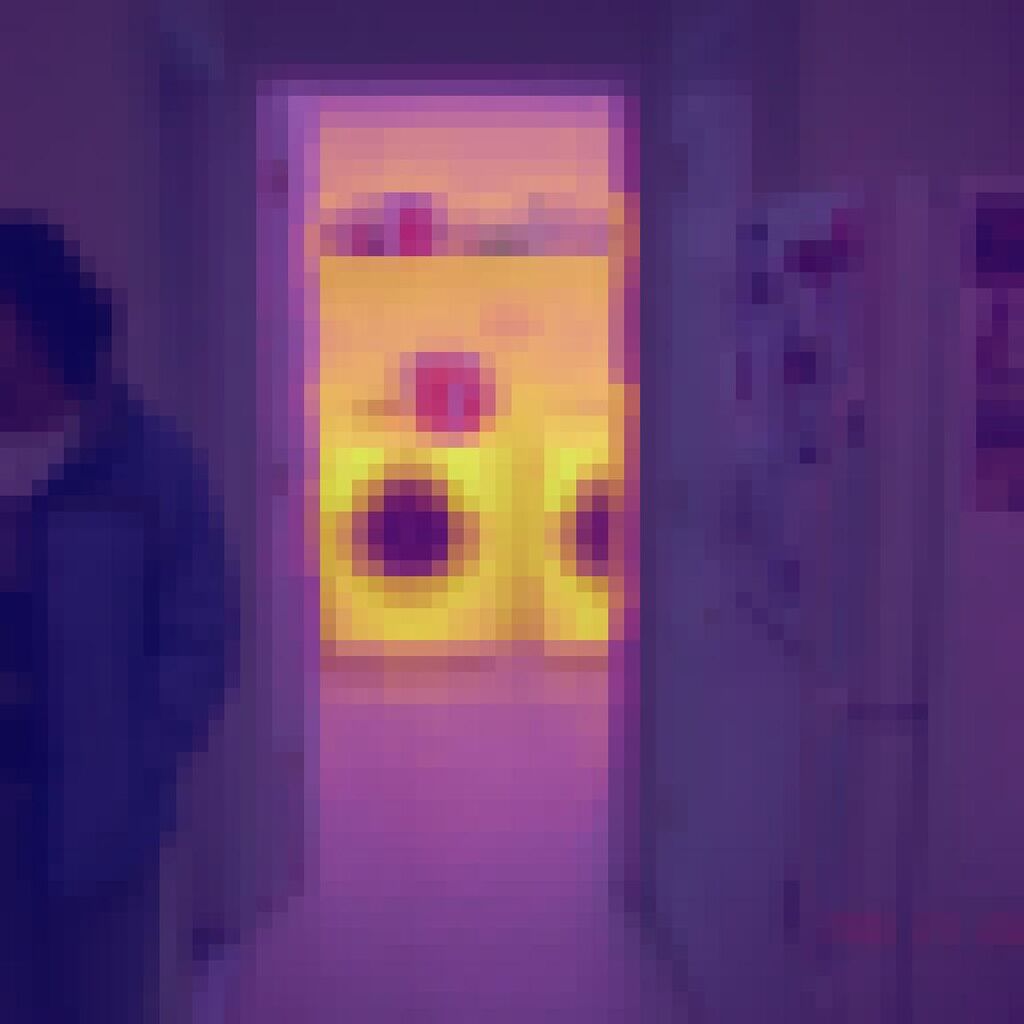}} \\

    t=10 & t=20 & t=30 & t=40 & t=50 & t=100 & t=150 & t=200 & t=250 & t=300 & t=350 \\[1em]
    
    \fcolorbox{red}{white}{\includegraphics[width=0.95\linewidth]{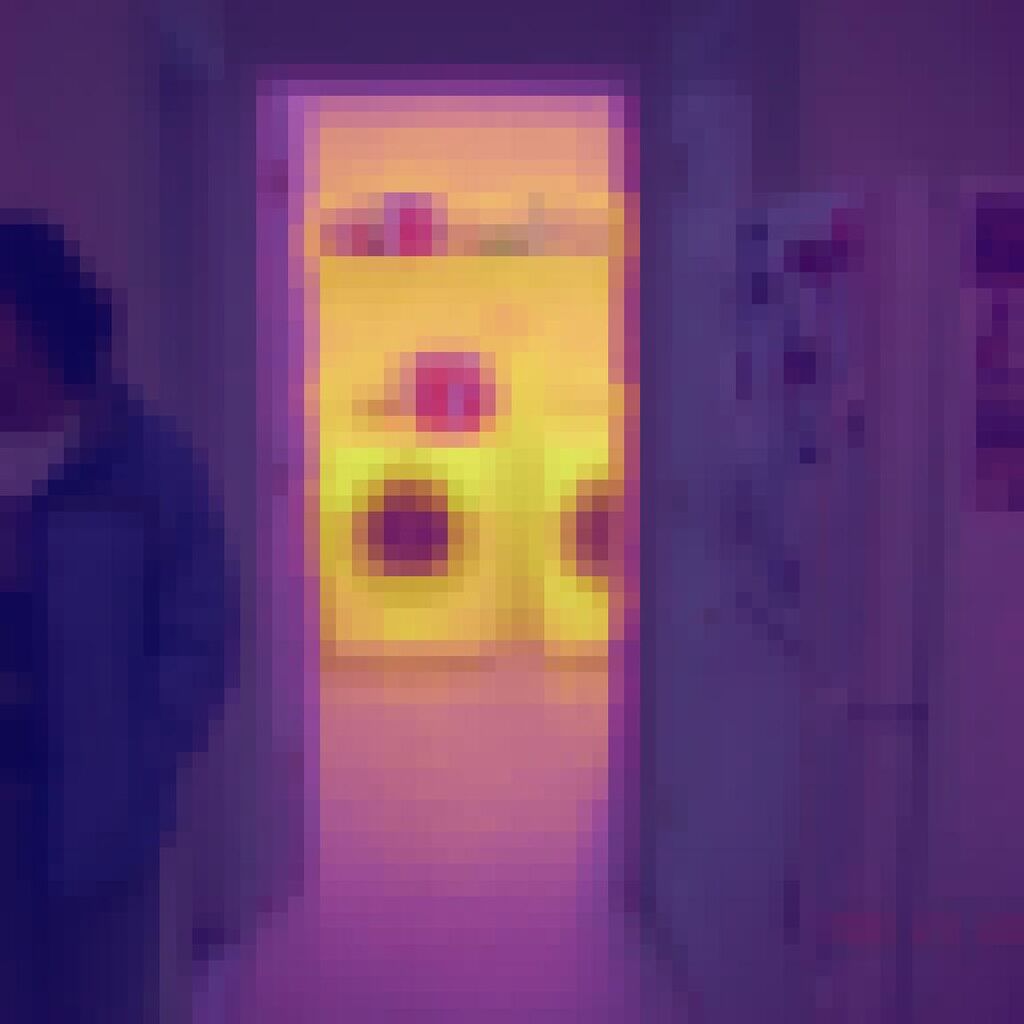}} &
    \fcolorbox{red}{white}{\includegraphics[width=0.95\linewidth]{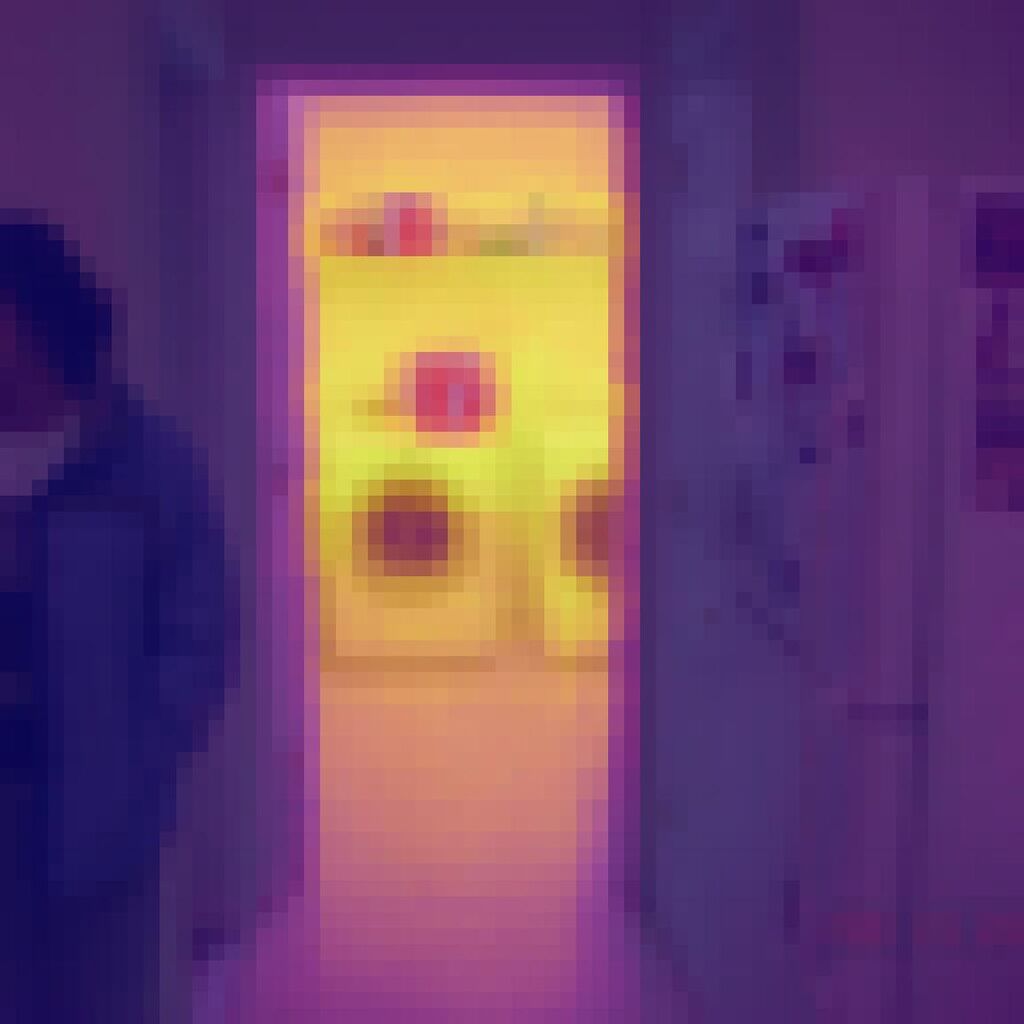}} &
    \fcolorbox{red}{white}{\includegraphics[width=0.95\linewidth]{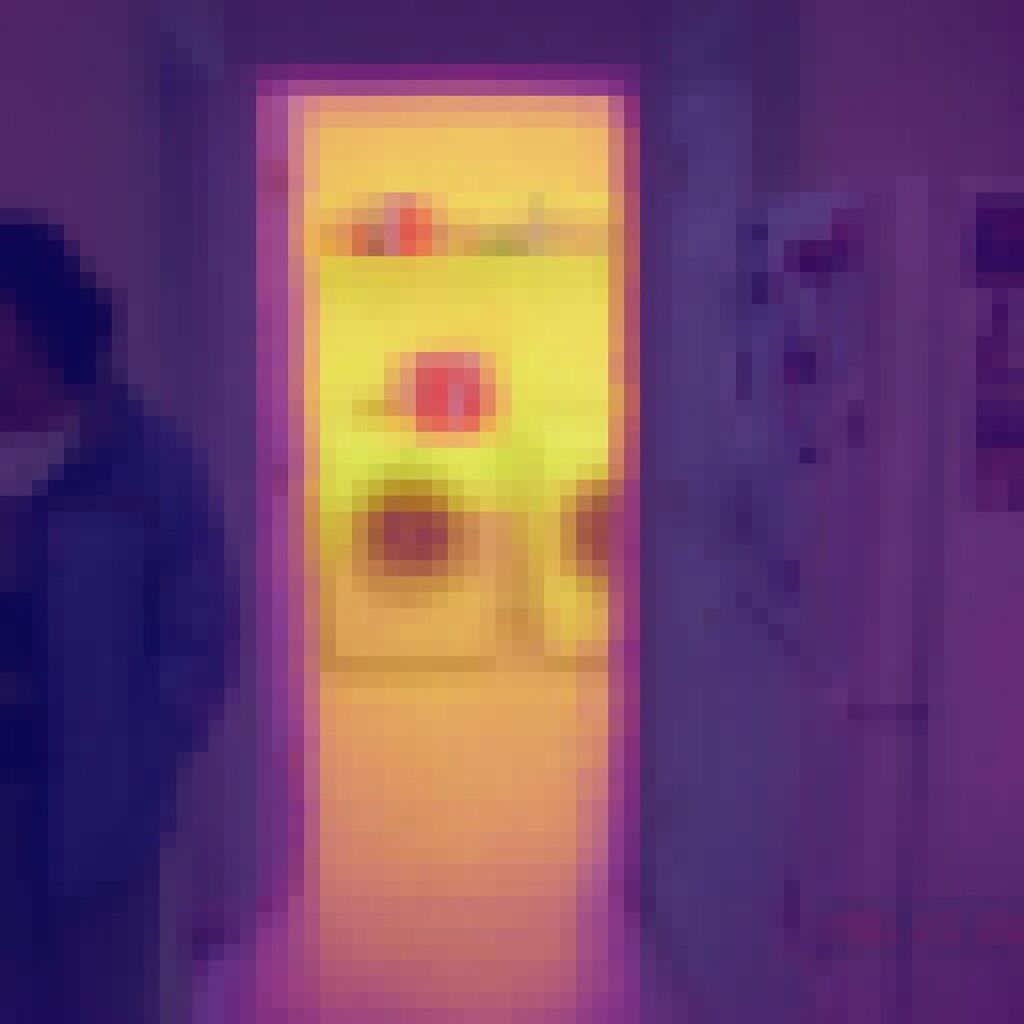}} &
    \fcolorbox{red}{white}{\includegraphics[width=0.95\linewidth]{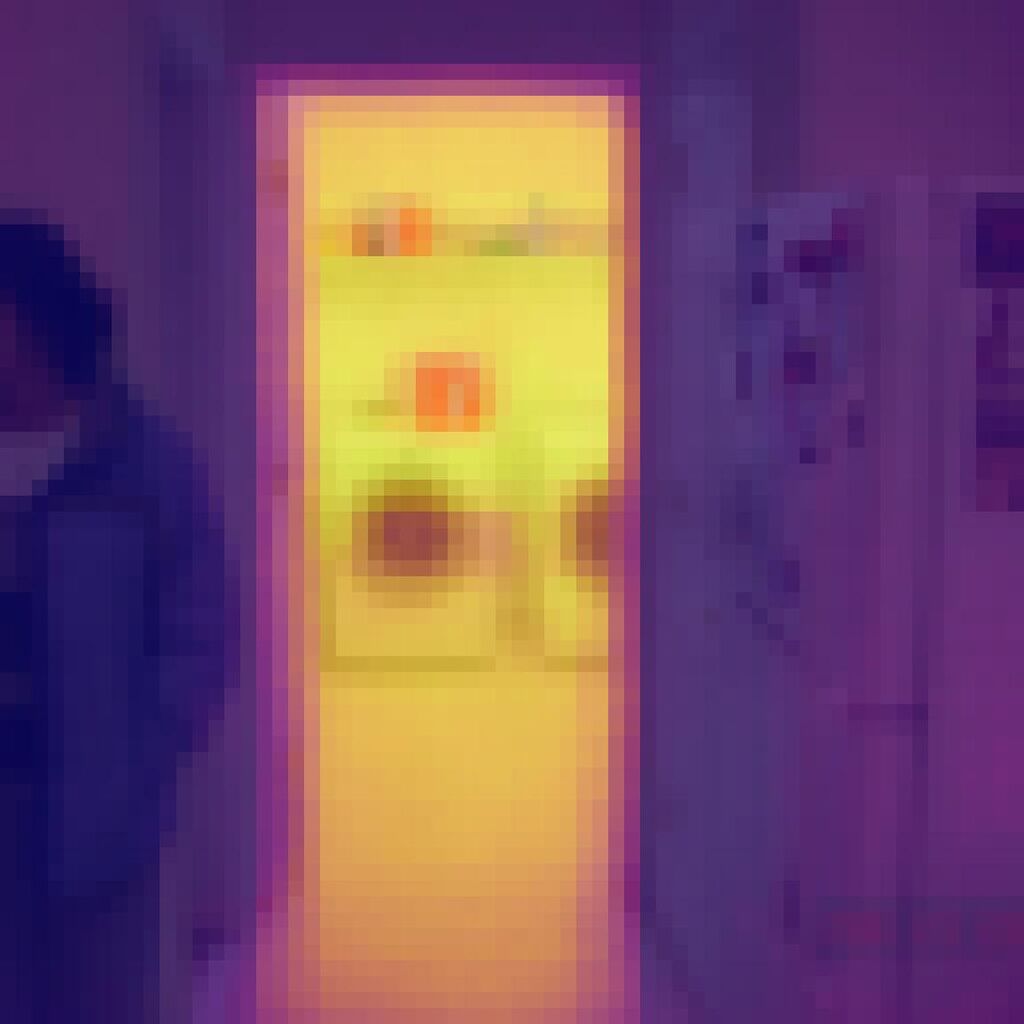}} &
    \fcolorbox{red}{white}{\includegraphics[width=0.95\linewidth]{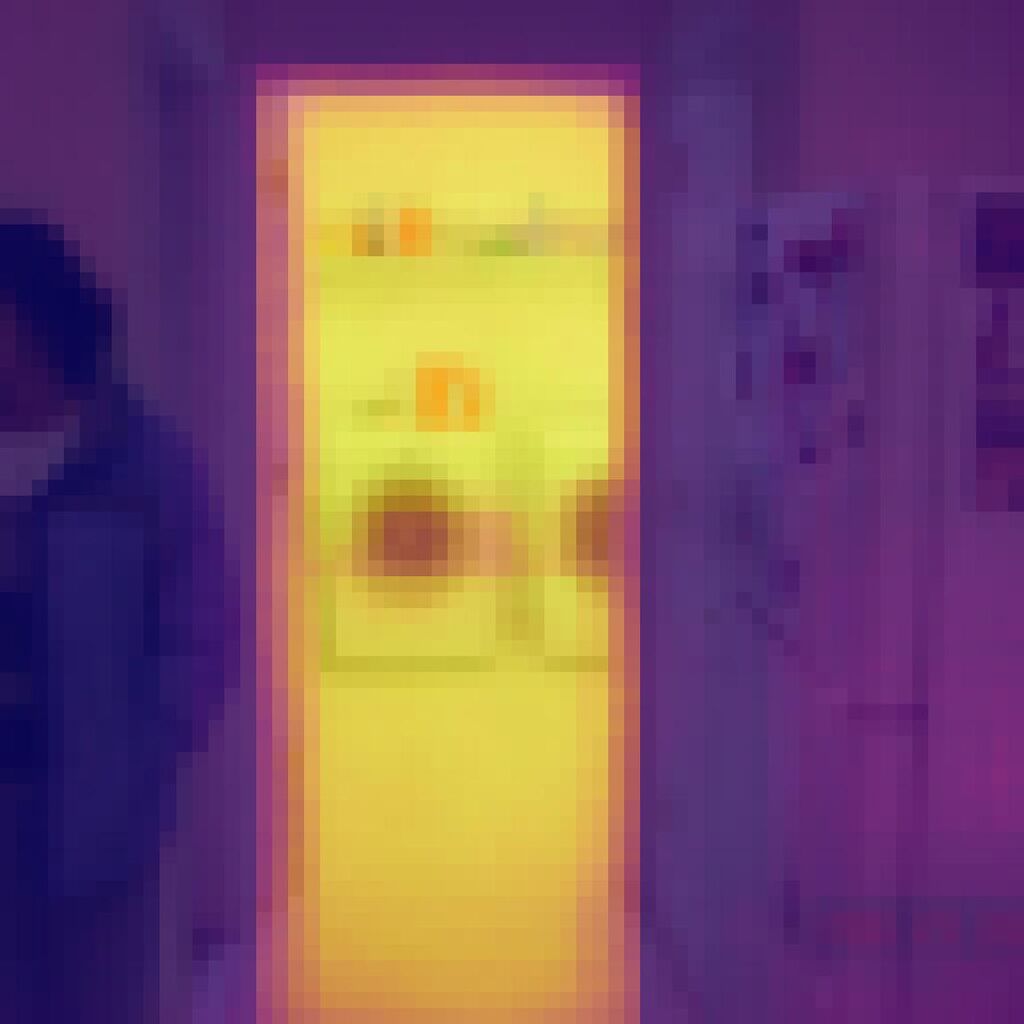}} &
    \fcolorbox{red}{white}{\includegraphics[width=0.95\linewidth]{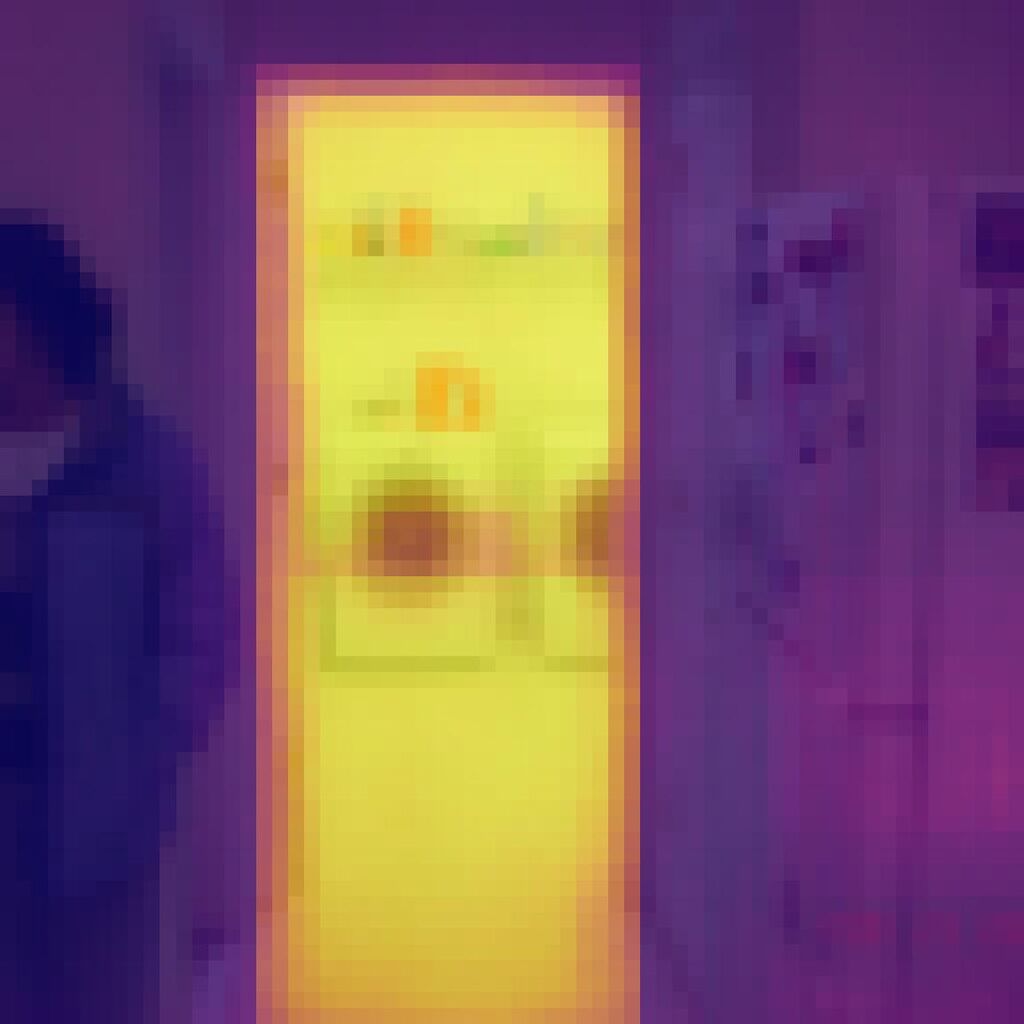}} &
    \fcolorbox{red}{white}{\includegraphics[width=0.95\linewidth]{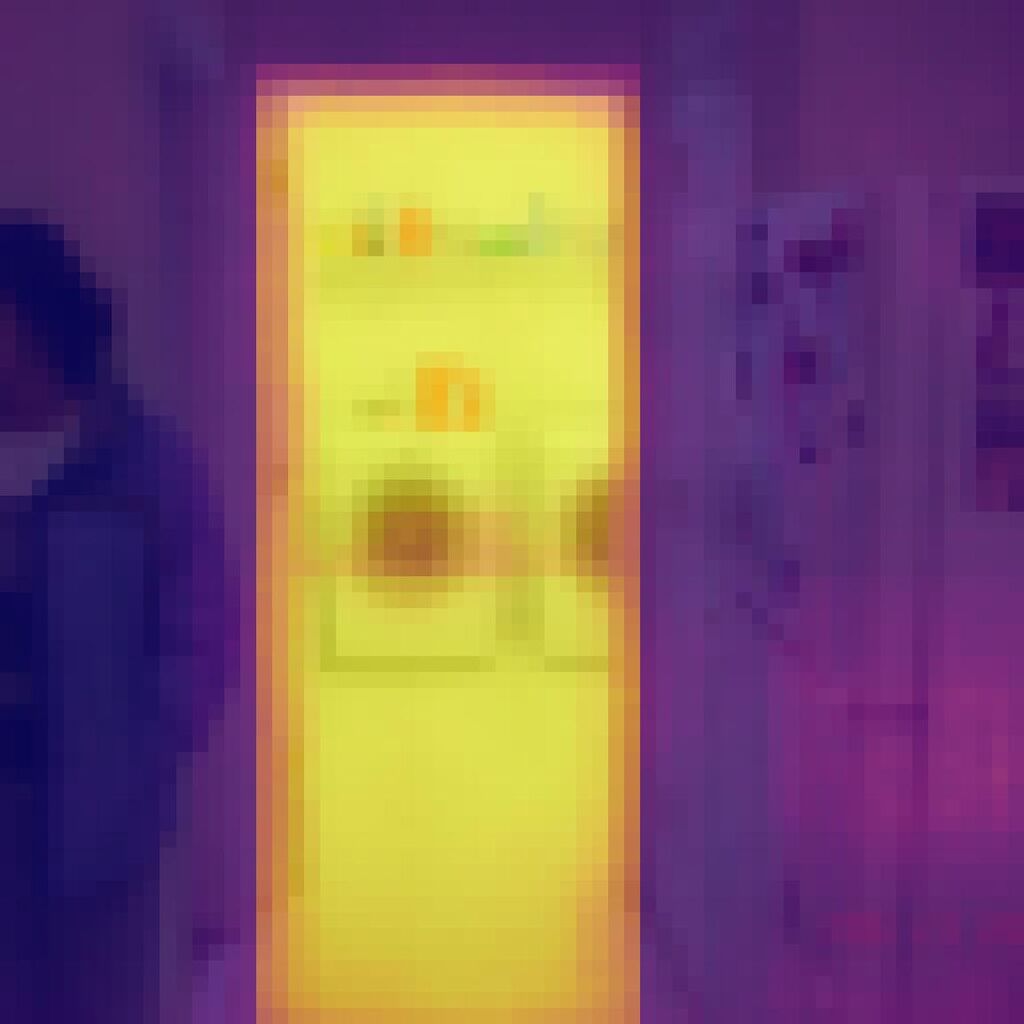}} &
    \fcolorbox{red}{white}{\includegraphics[width=0.95\linewidth]{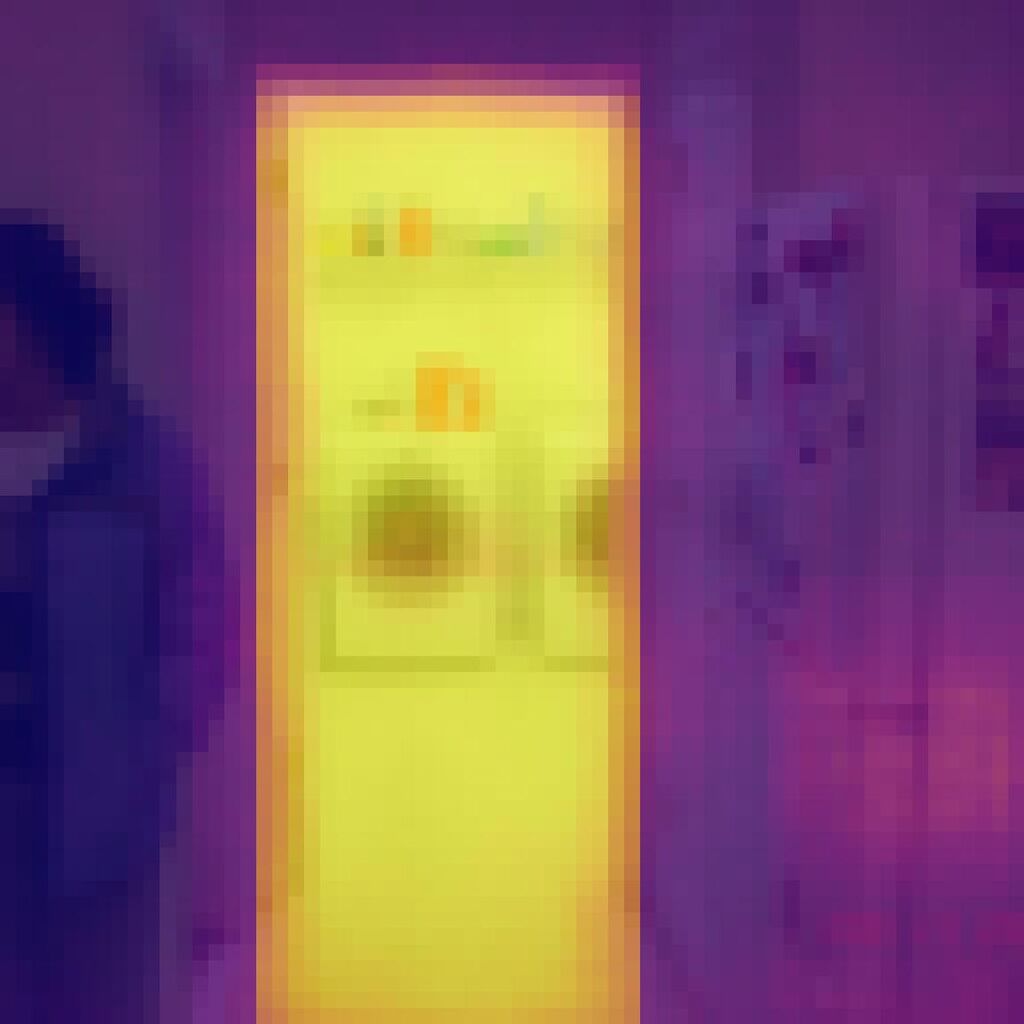}} &
    \fcolorbox{red}{white}{\includegraphics[width=0.95\linewidth]{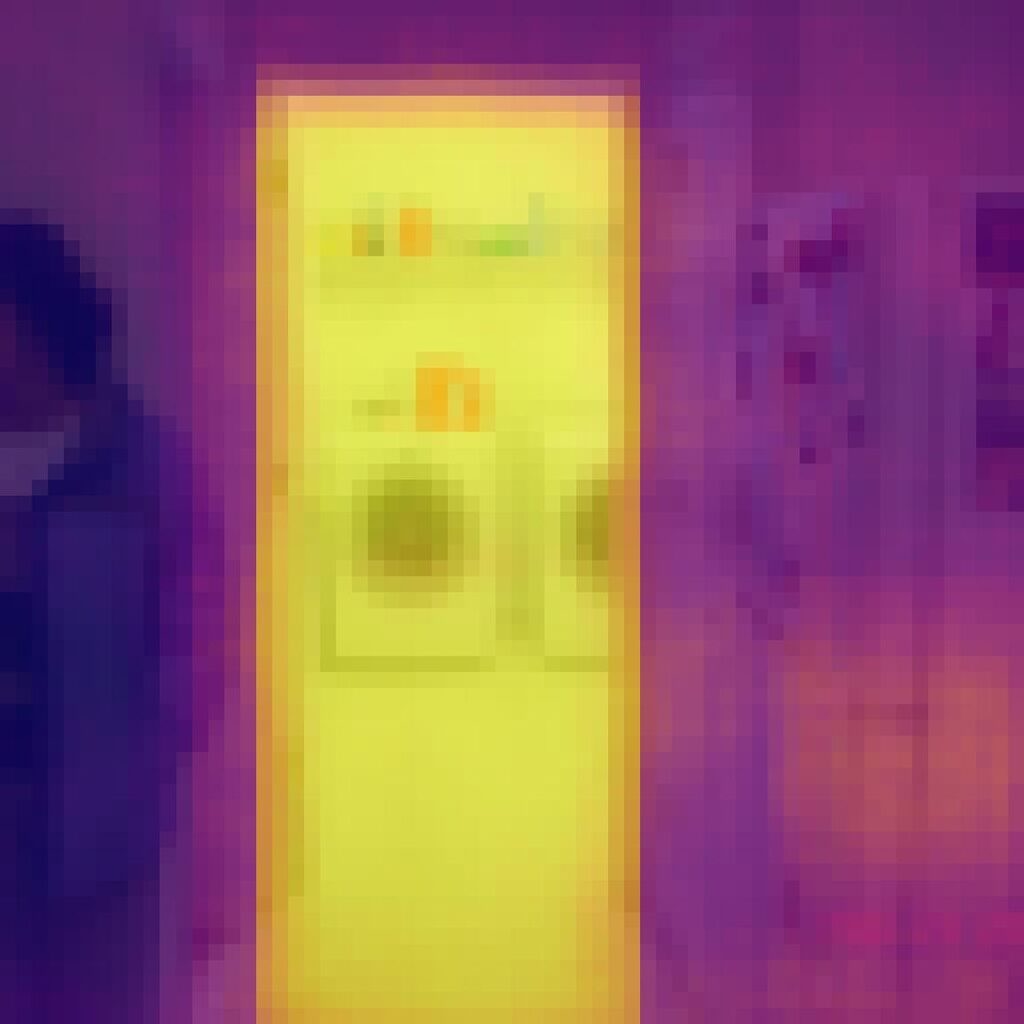}} &
    \fcolorbox{red}{white}{\includegraphics[width=0.95\linewidth]{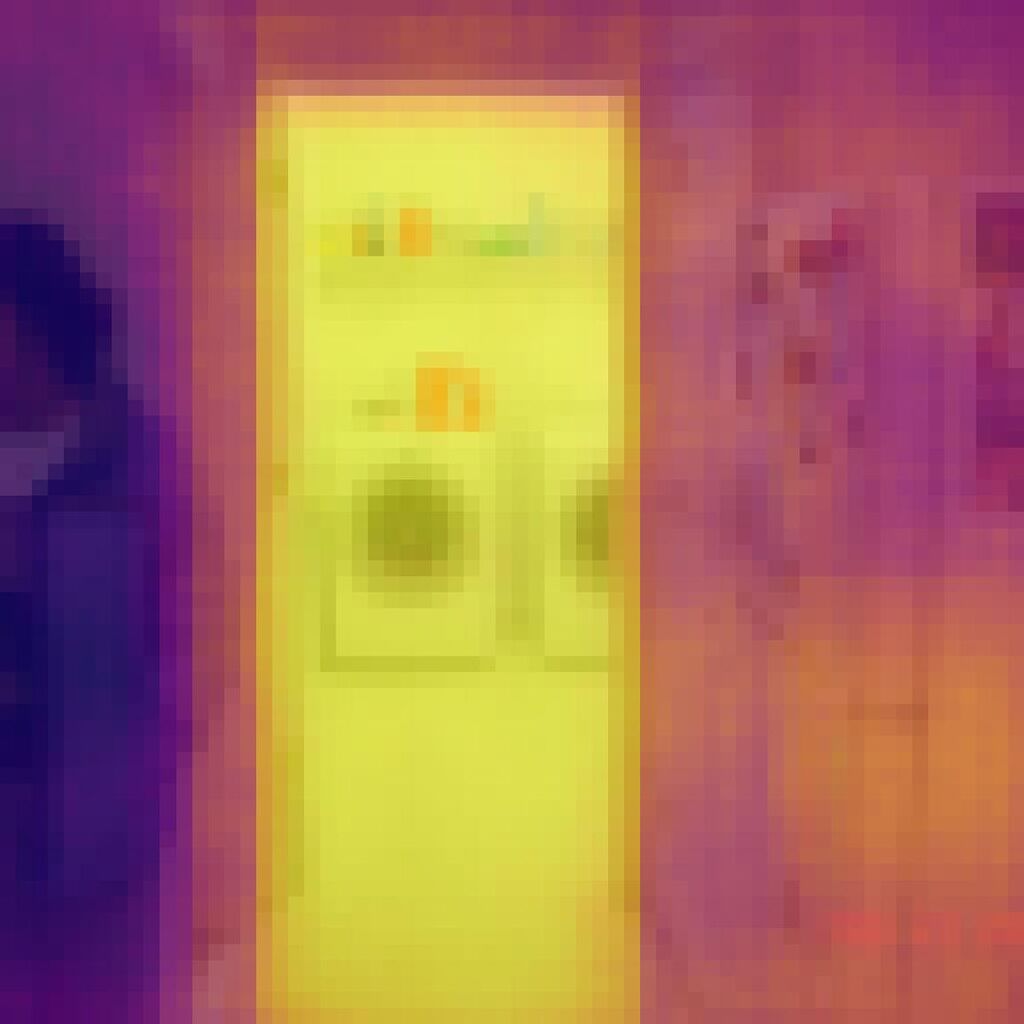}} &
    \fcolorbox{red}{white}{\includegraphics[width=0.95\linewidth]{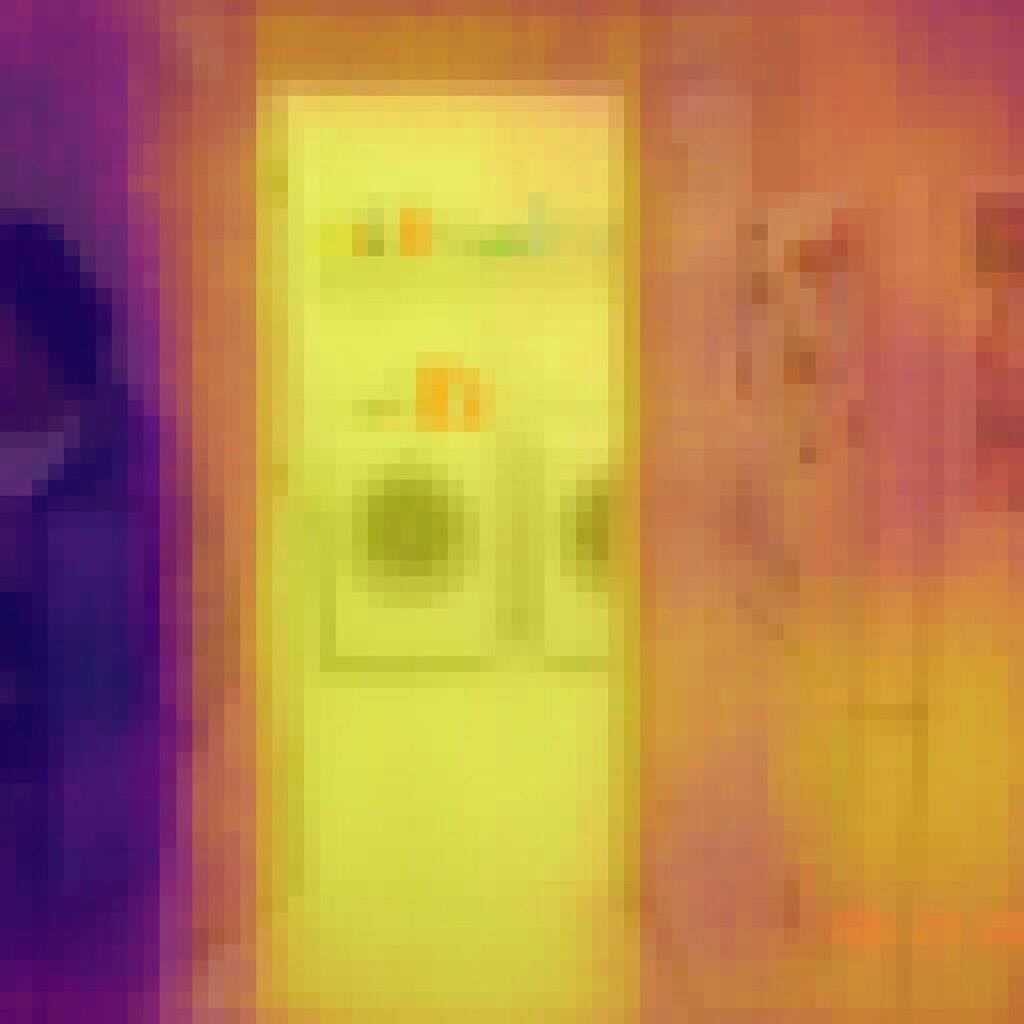}} \\
    
    t=400 & t=450 & t=500 & t=550 & t=600 & t=650 & t=700 & t=750 & t=800 & t=850 & t=900 \\
    
    \multicolumn{11}{c}{\vspace{0.5em}} \\ 
    
    \multicolumn{11}{c}{\textbf{Blue Point CSMs}} \\
    
    \fcolorbox{blue}{white}{\includegraphics[width=0.95\linewidth]{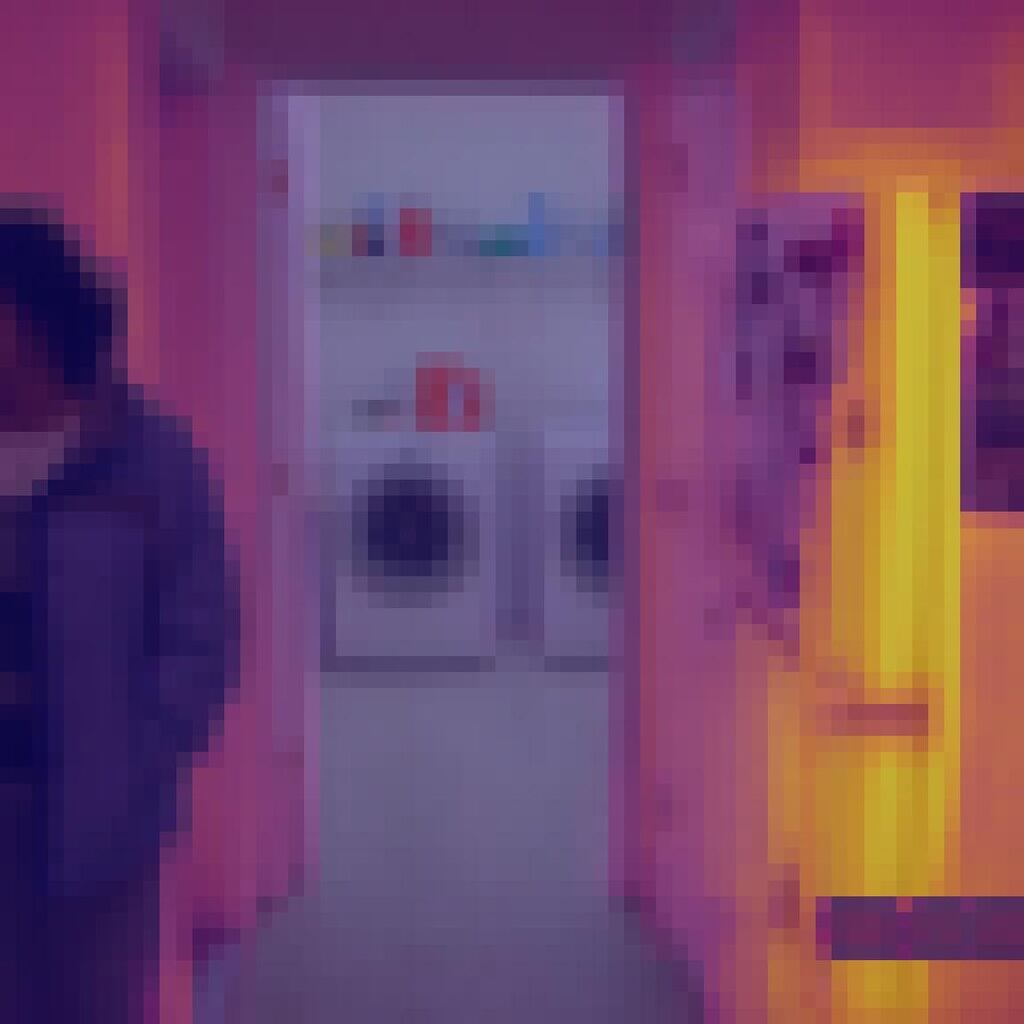}} &
    \fcolorbox{blue}{white}{\includegraphics[width=0.95\linewidth]{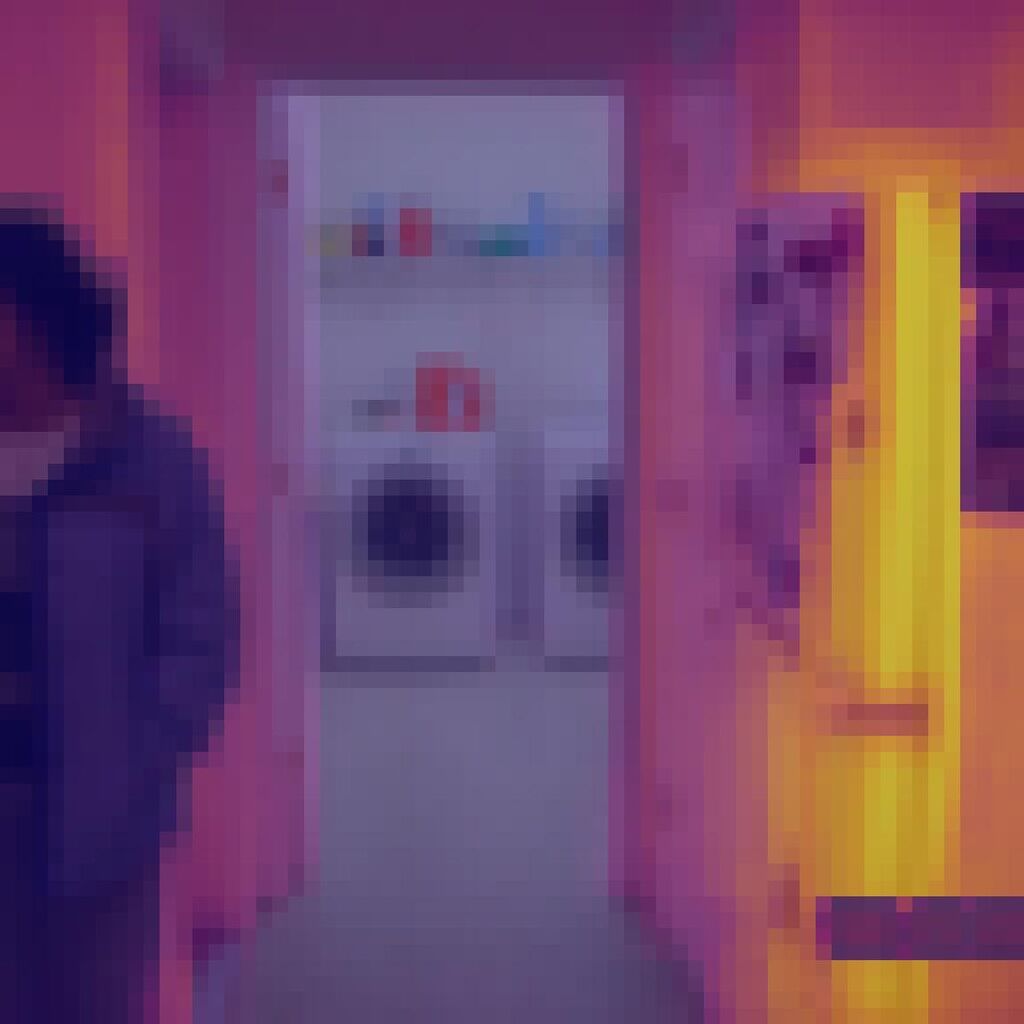}} &
    \fcolorbox{blue}{white}{\includegraphics[width=0.95\linewidth]{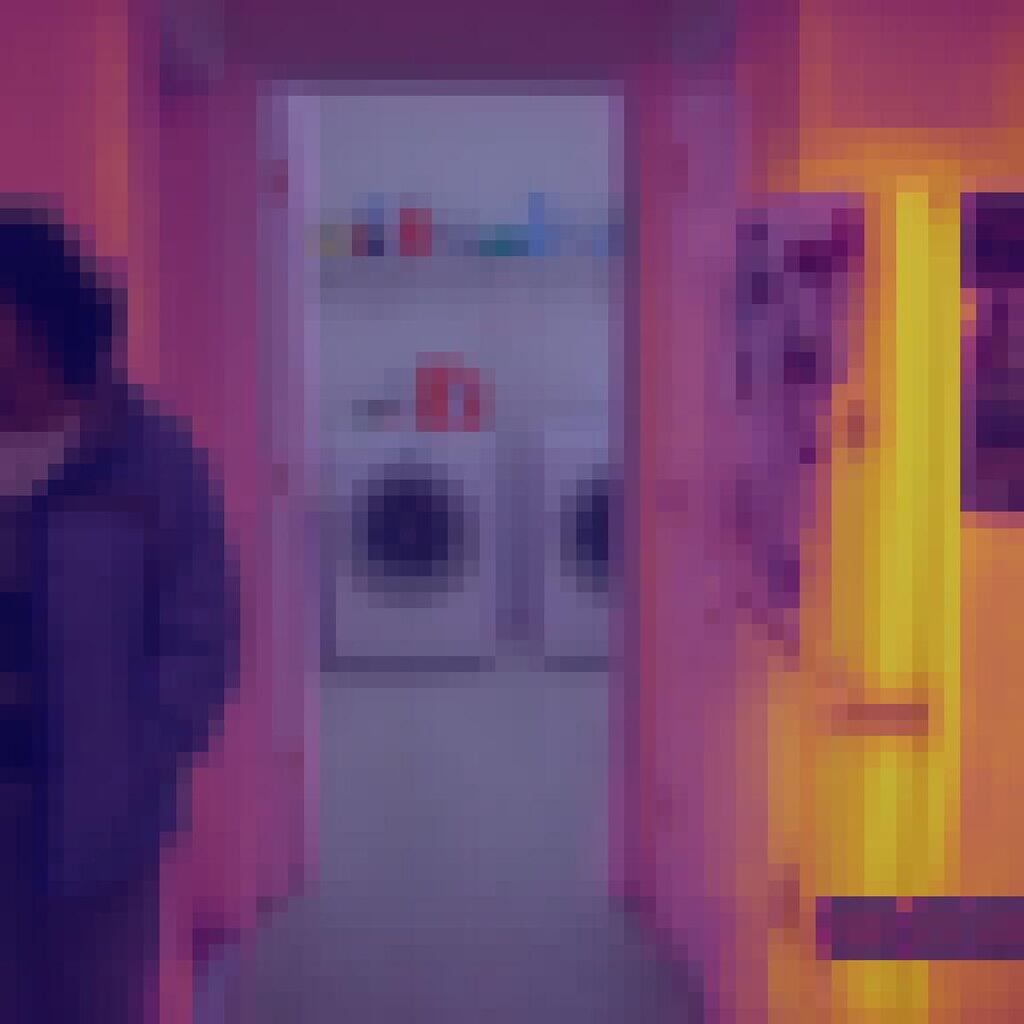}} &
    \fcolorbox{blue}{white}{\includegraphics[width=0.95\linewidth]{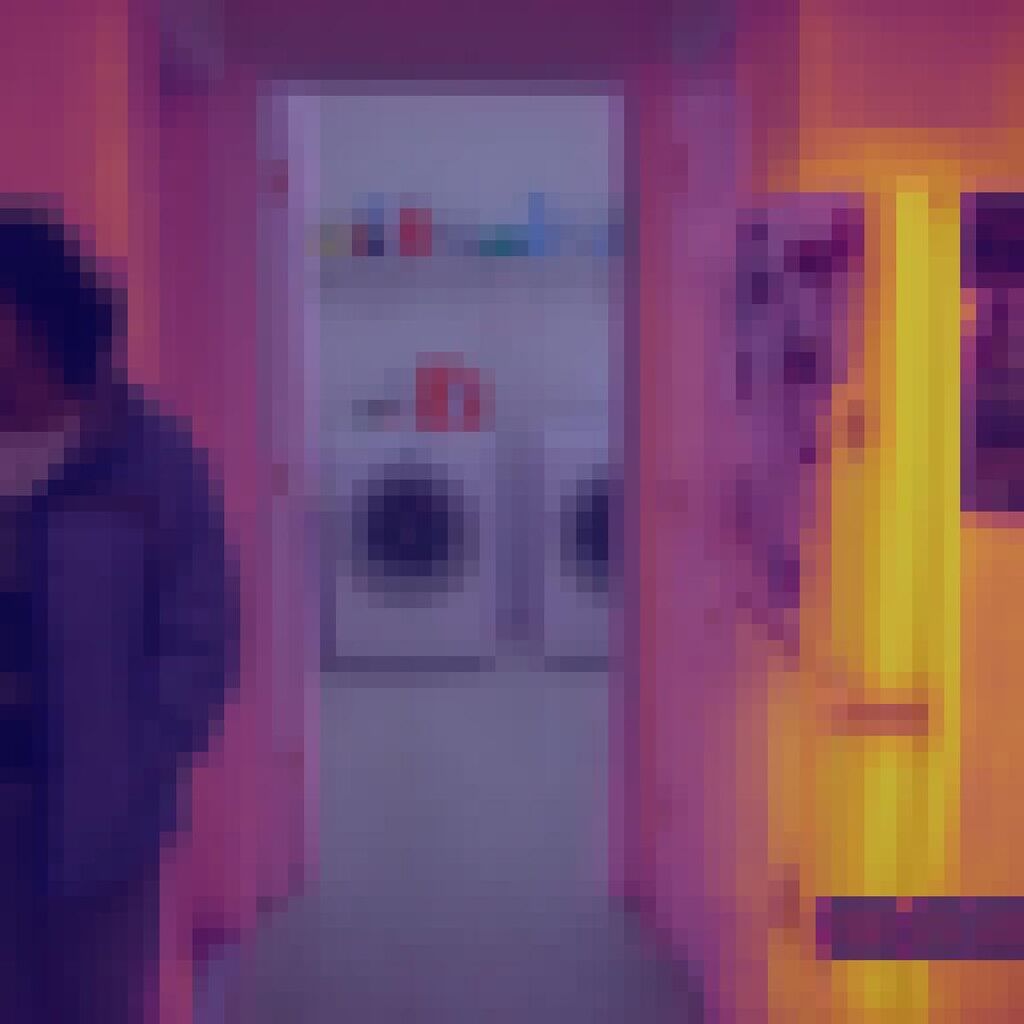}} &
    \fcolorbox{blue}{white}{\includegraphics[width=0.95\linewidth]{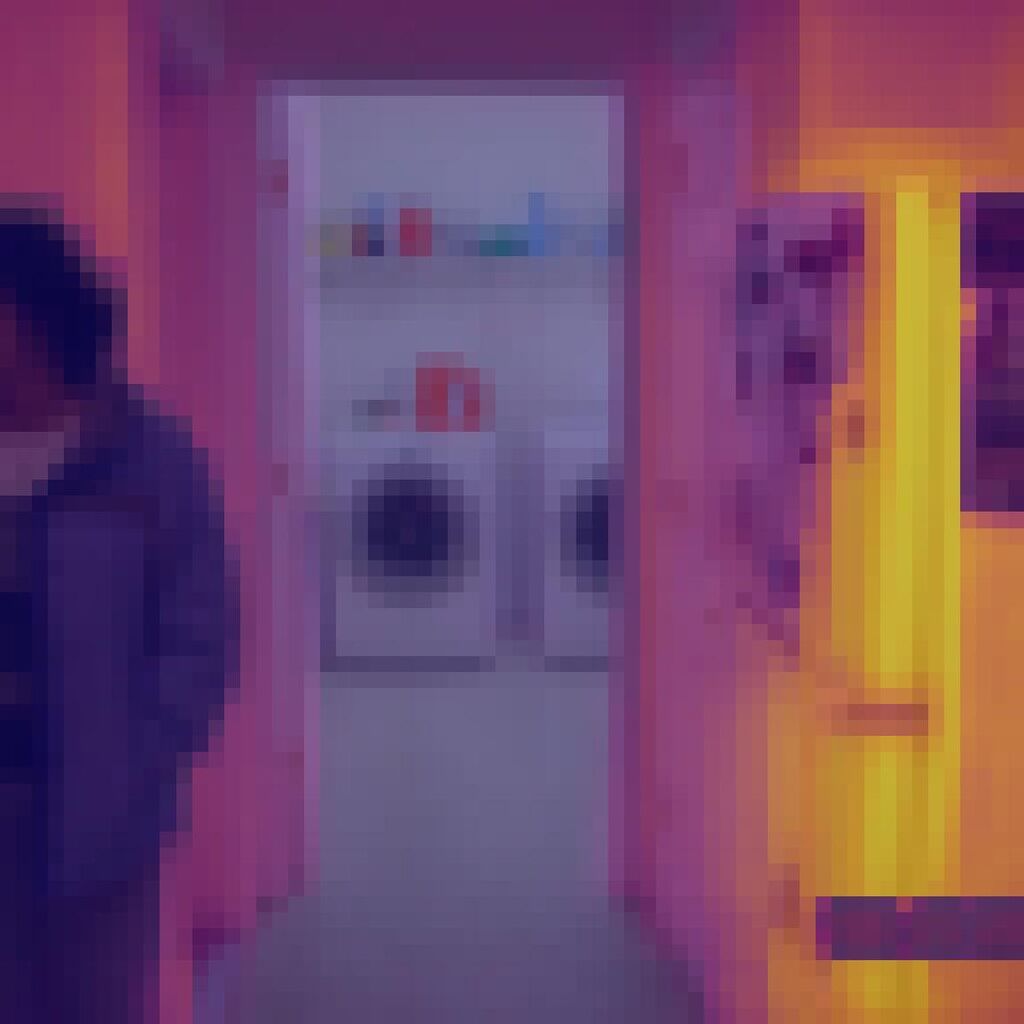}} &
    \fcolorbox{blue}{white}{\includegraphics[width=0.95\linewidth]{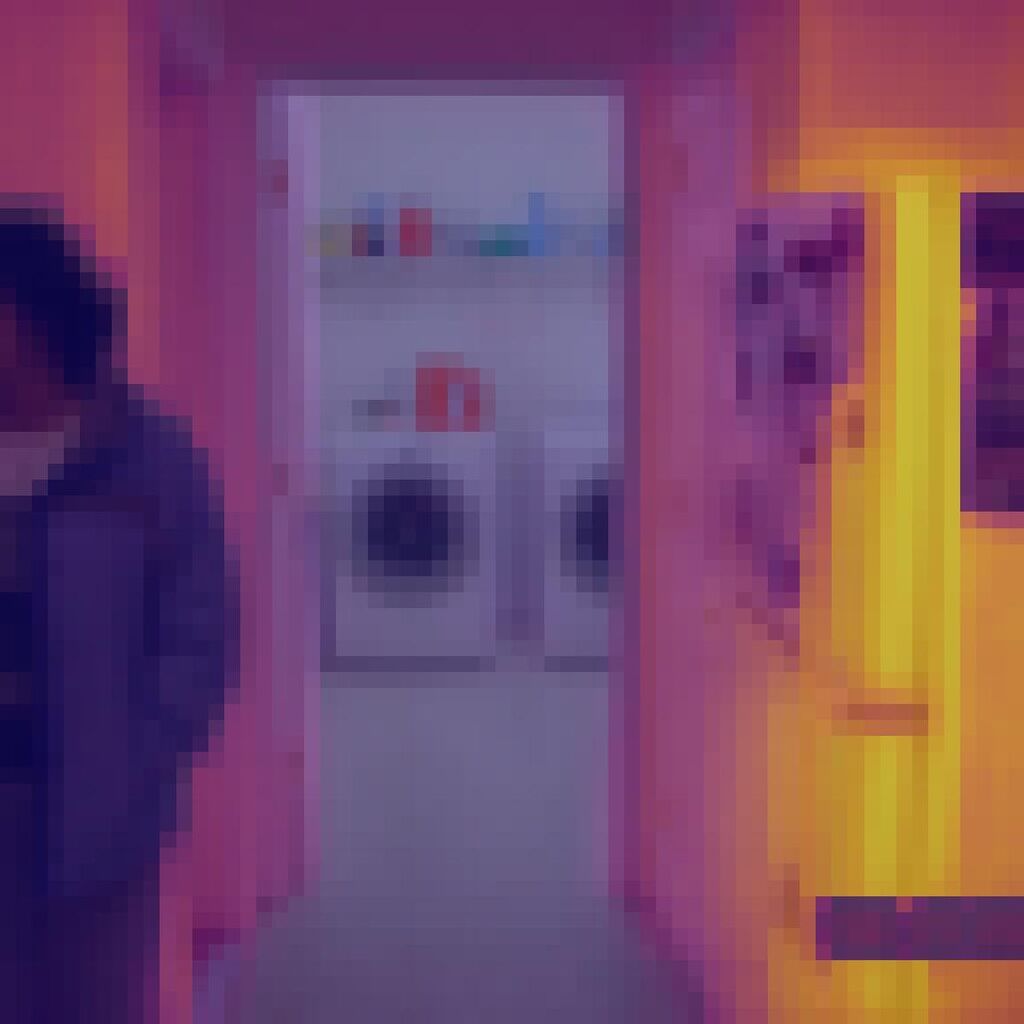}} &
    \fcolorbox{blue}{white}{\includegraphics[width=0.95\linewidth]{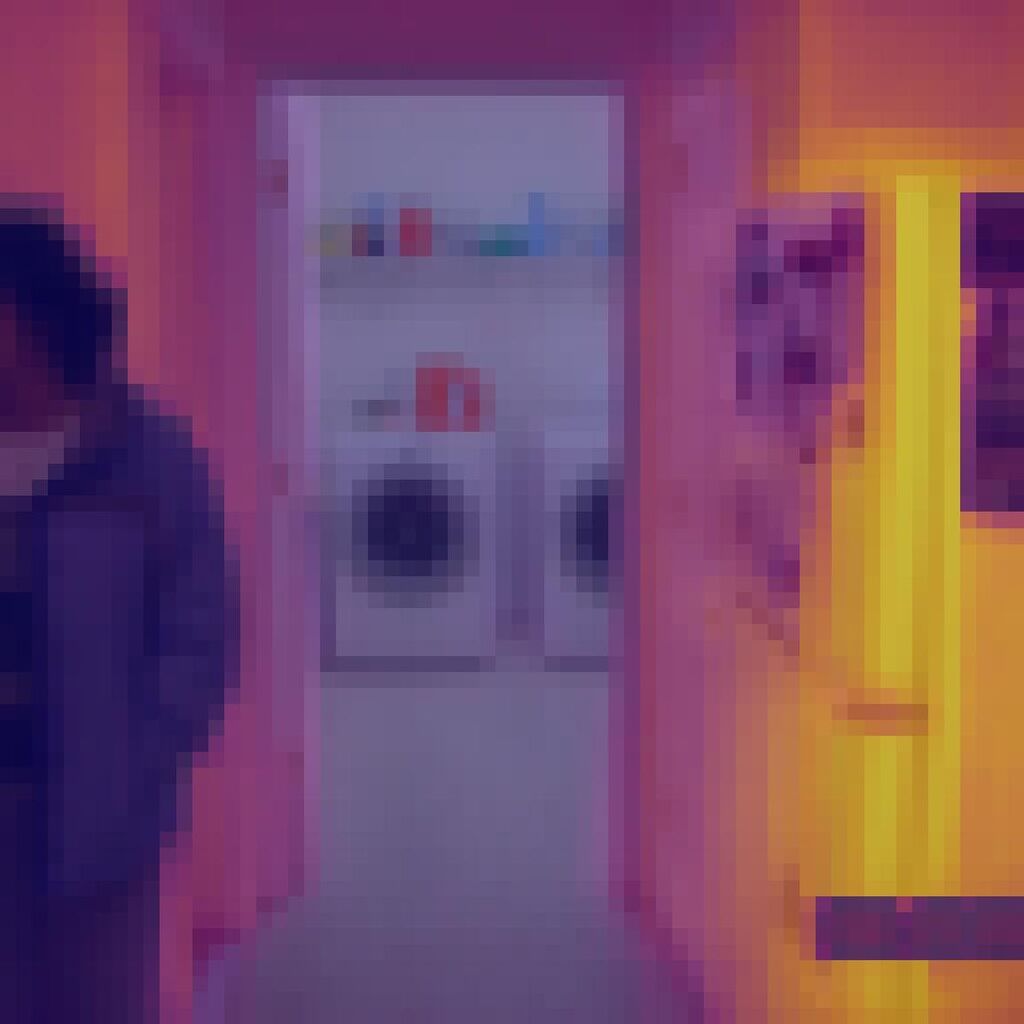}} &
    \fcolorbox{blue}{white}{\includegraphics[width=0.95\linewidth]{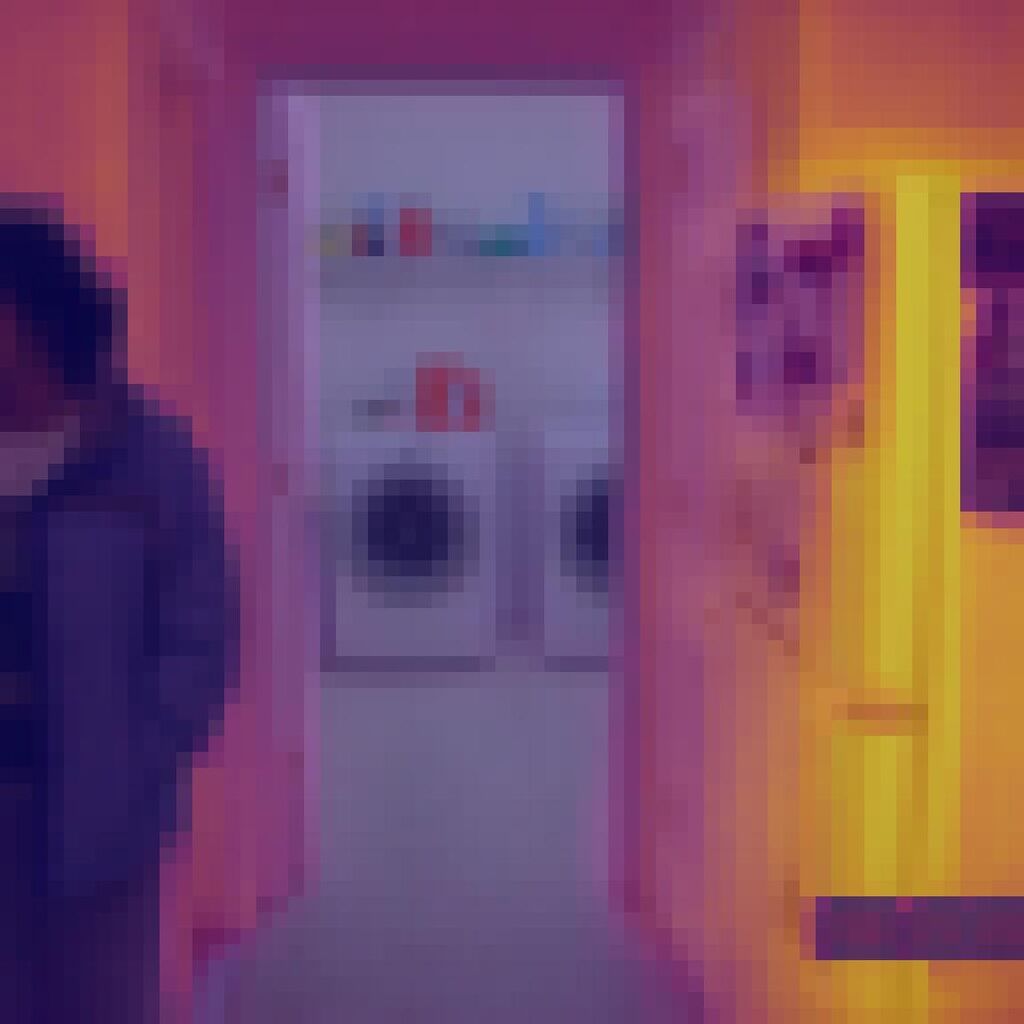}} &
    \fcolorbox{blue}{white}{\includegraphics[width=0.95\linewidth]{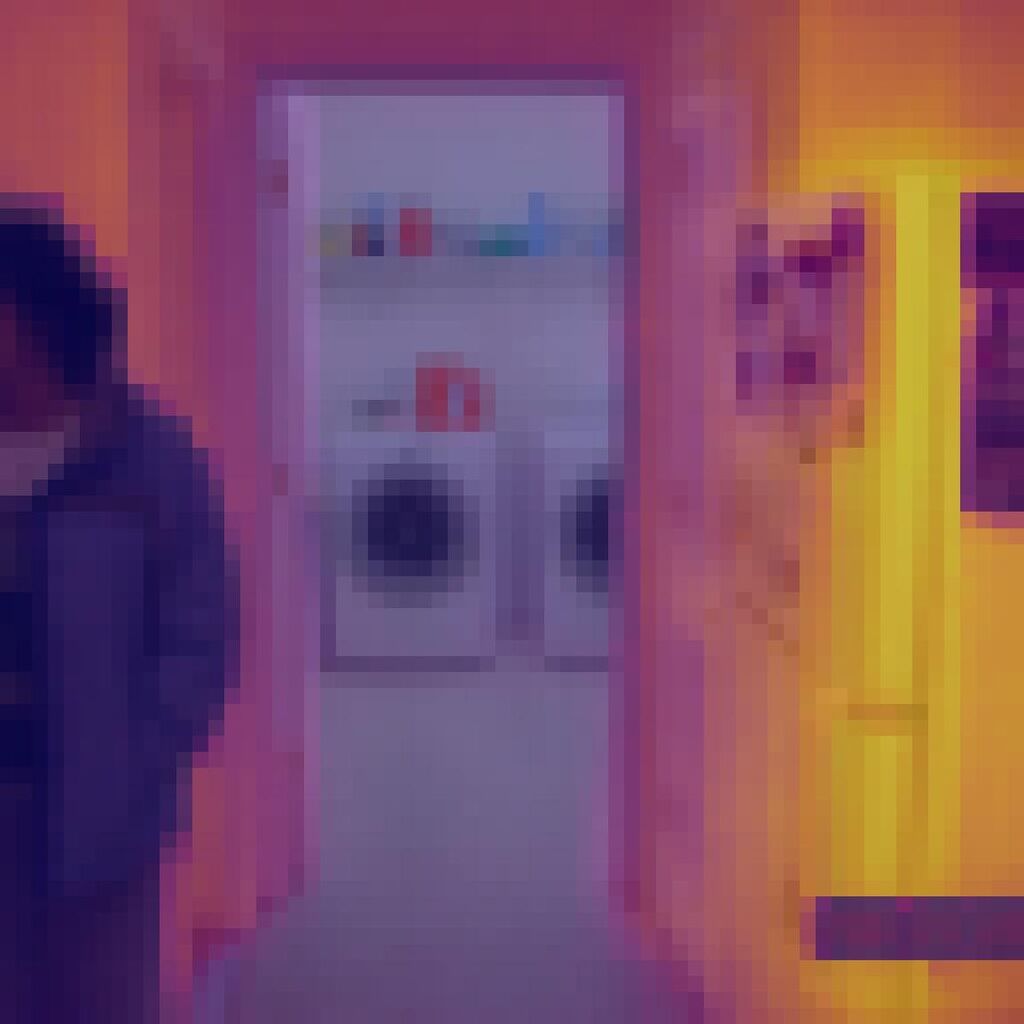}} &
    \fcolorbox{blue}{white}{\includegraphics[width=0.95\linewidth]{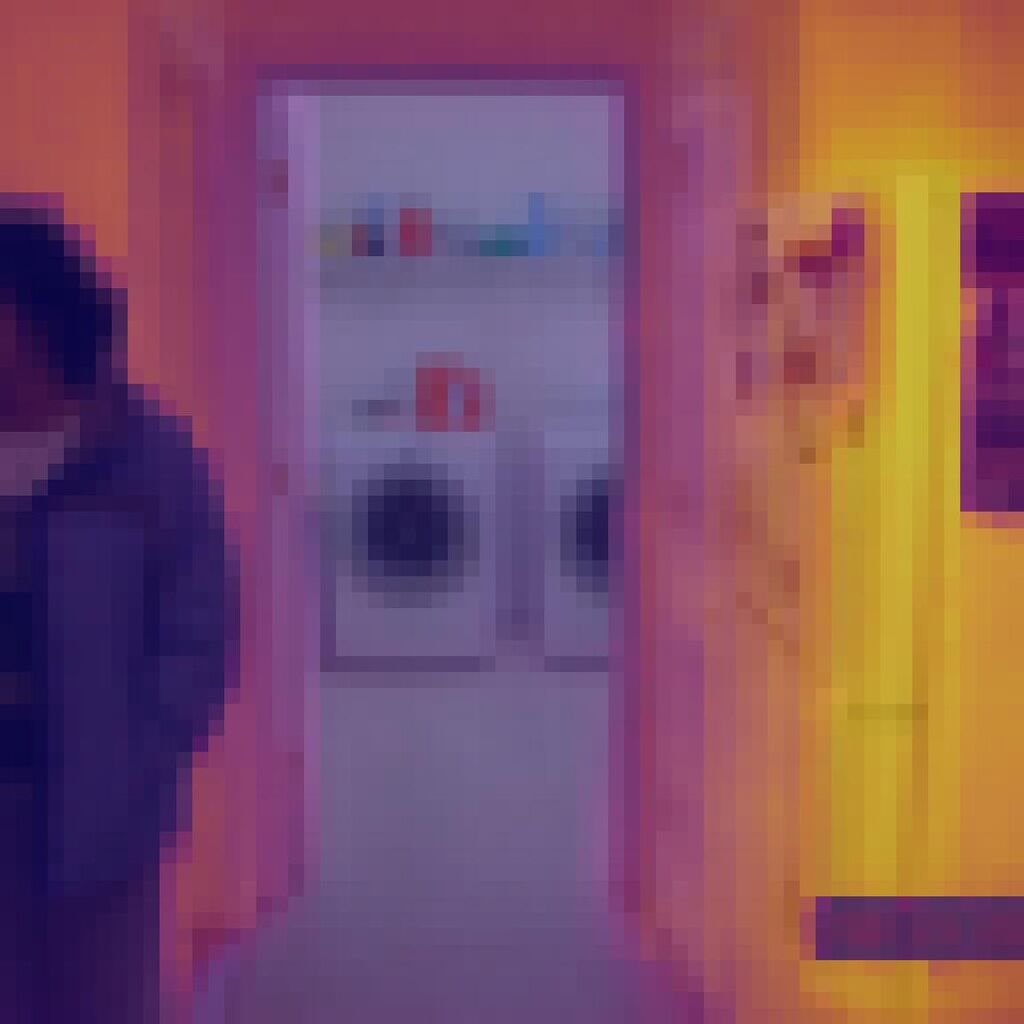}} &
    \fcolorbox{blue}{white}{\includegraphics[width=0.95\linewidth]{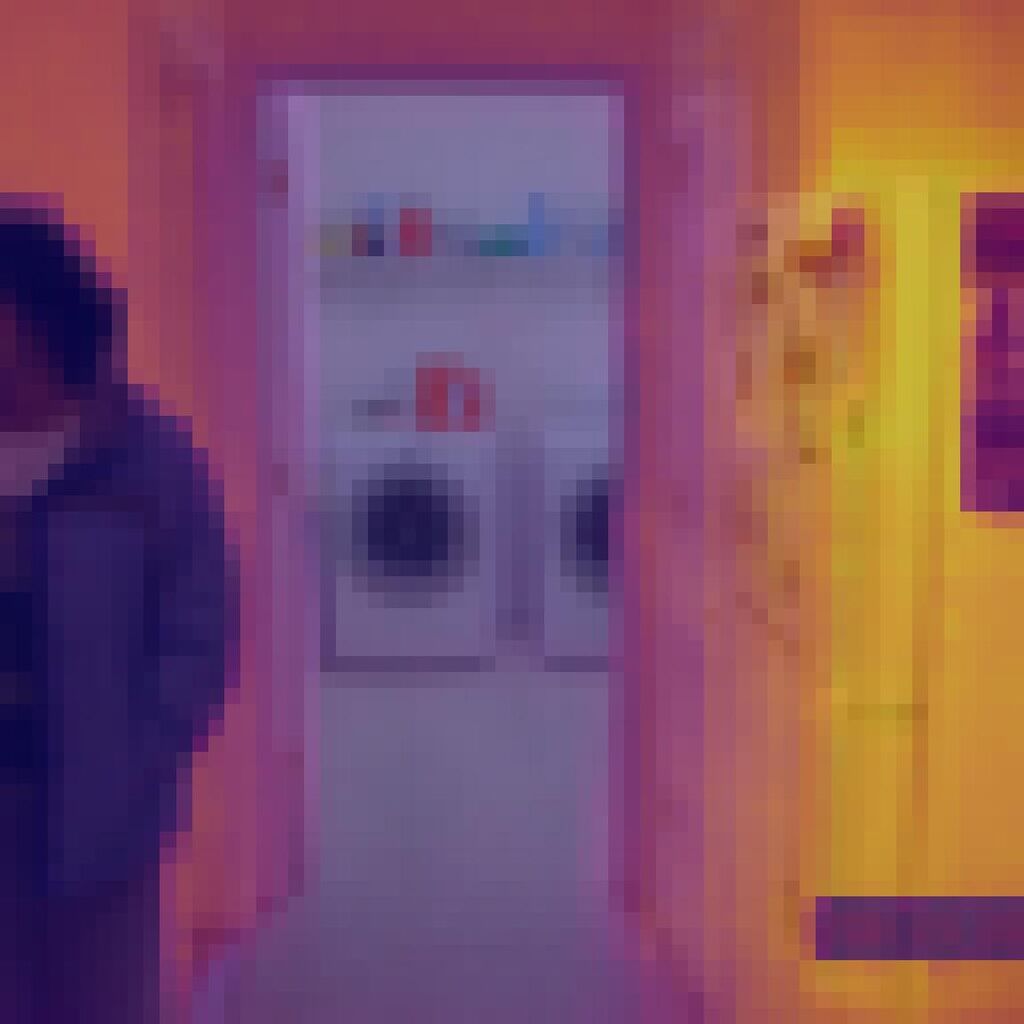}} \\

    t=10 & t=20 & t=30 & t=40 & t=50 & t=100 & t=150 & t=200 & t=250 & t=300 & t=350 \\[1em]
    
    \fcolorbox{blue}{white}{\includegraphics[width=0.95\linewidth]{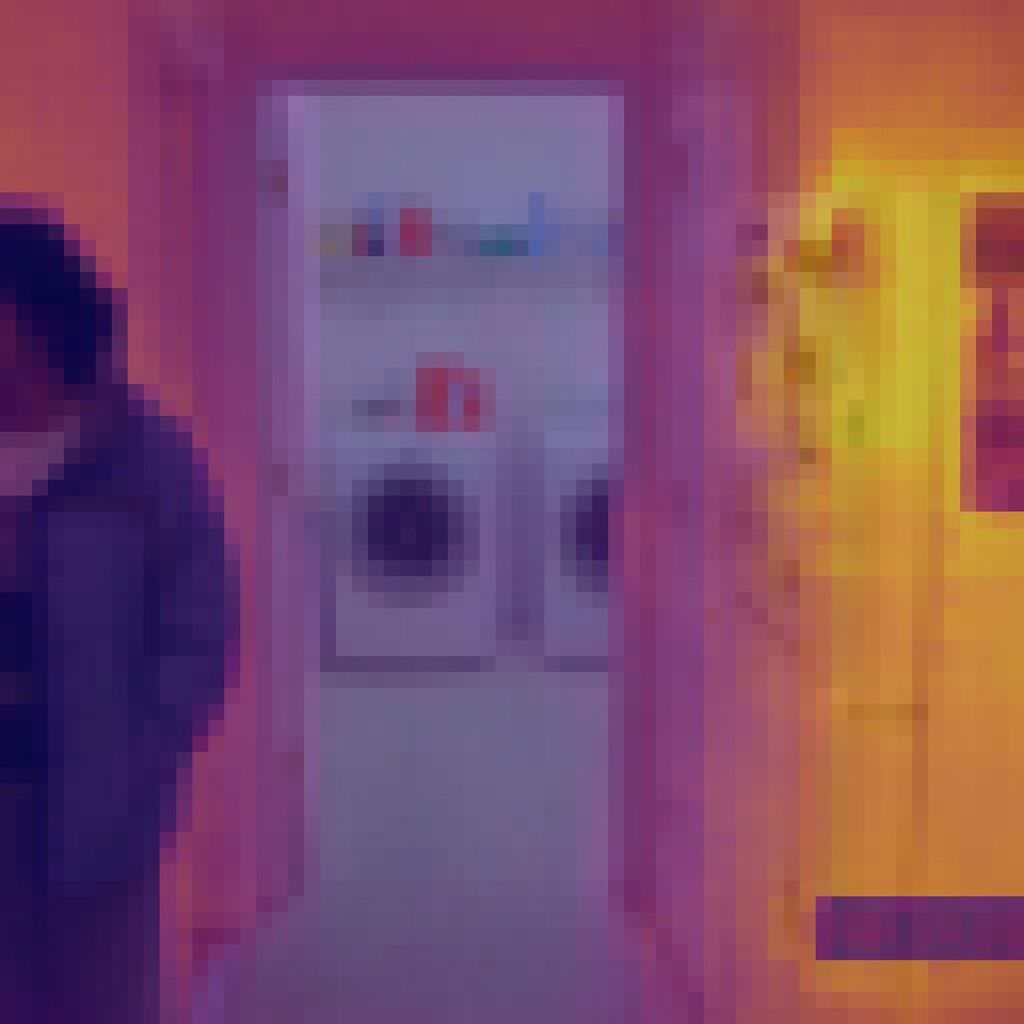}} &
    \fcolorbox{blue}{white}{\includegraphics[width=0.95\linewidth]{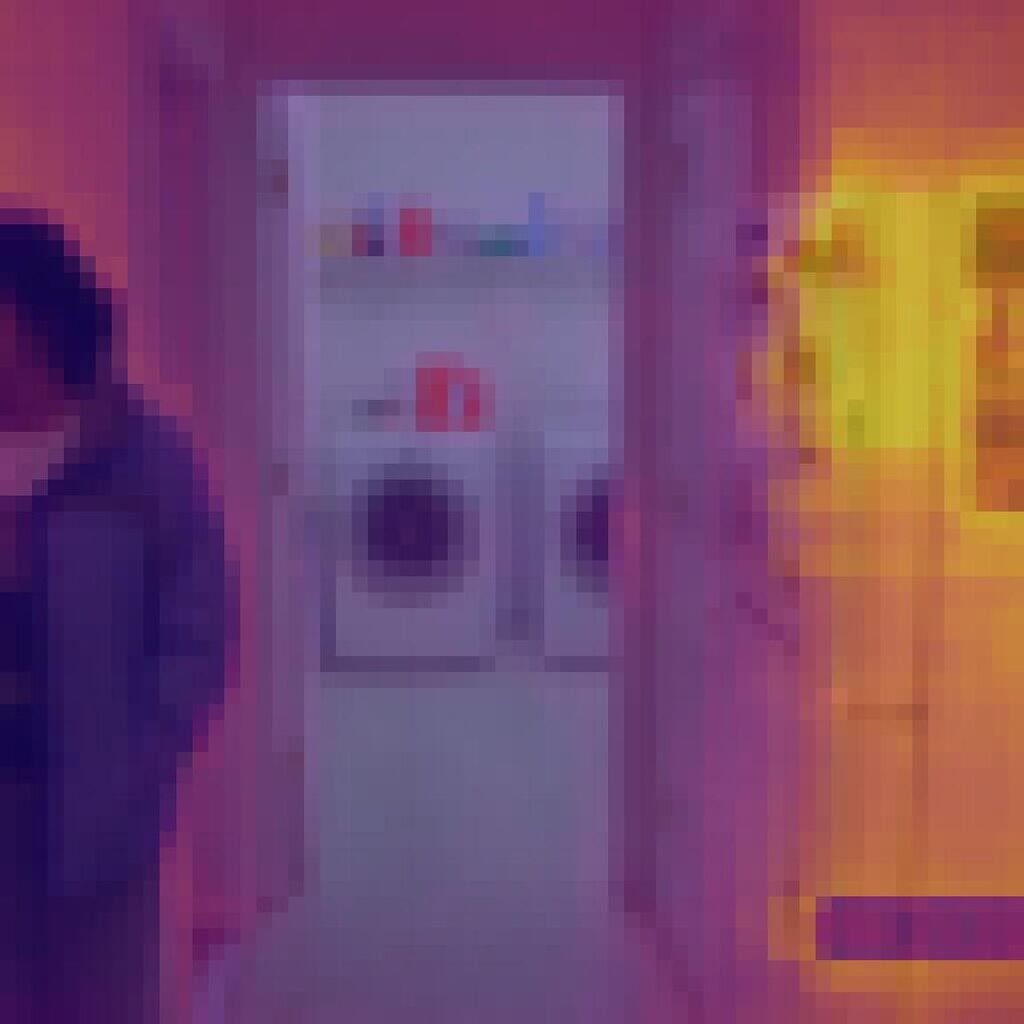}} &
    \fcolorbox{blue}{white}{\includegraphics[width=0.95\linewidth]{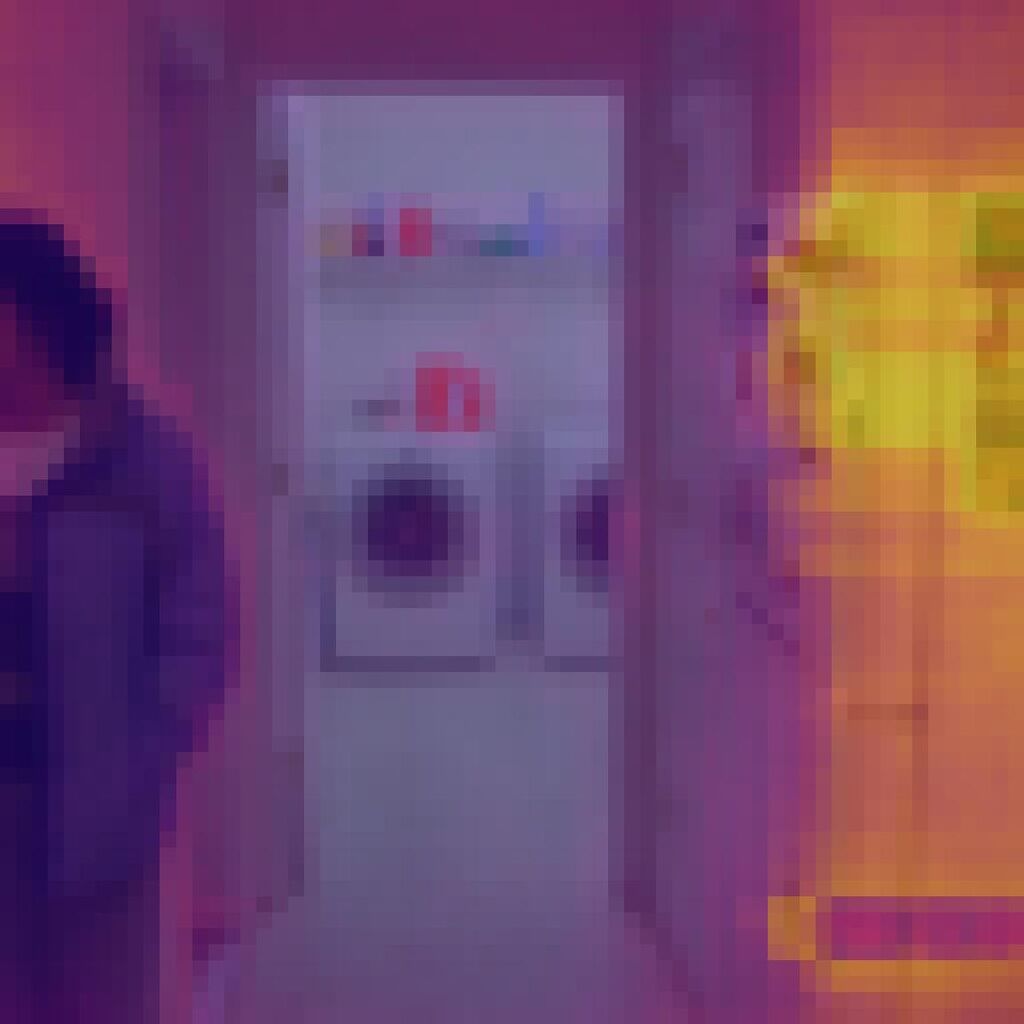}} &
    \fcolorbox{blue}{white}{\includegraphics[width=0.95\linewidth]{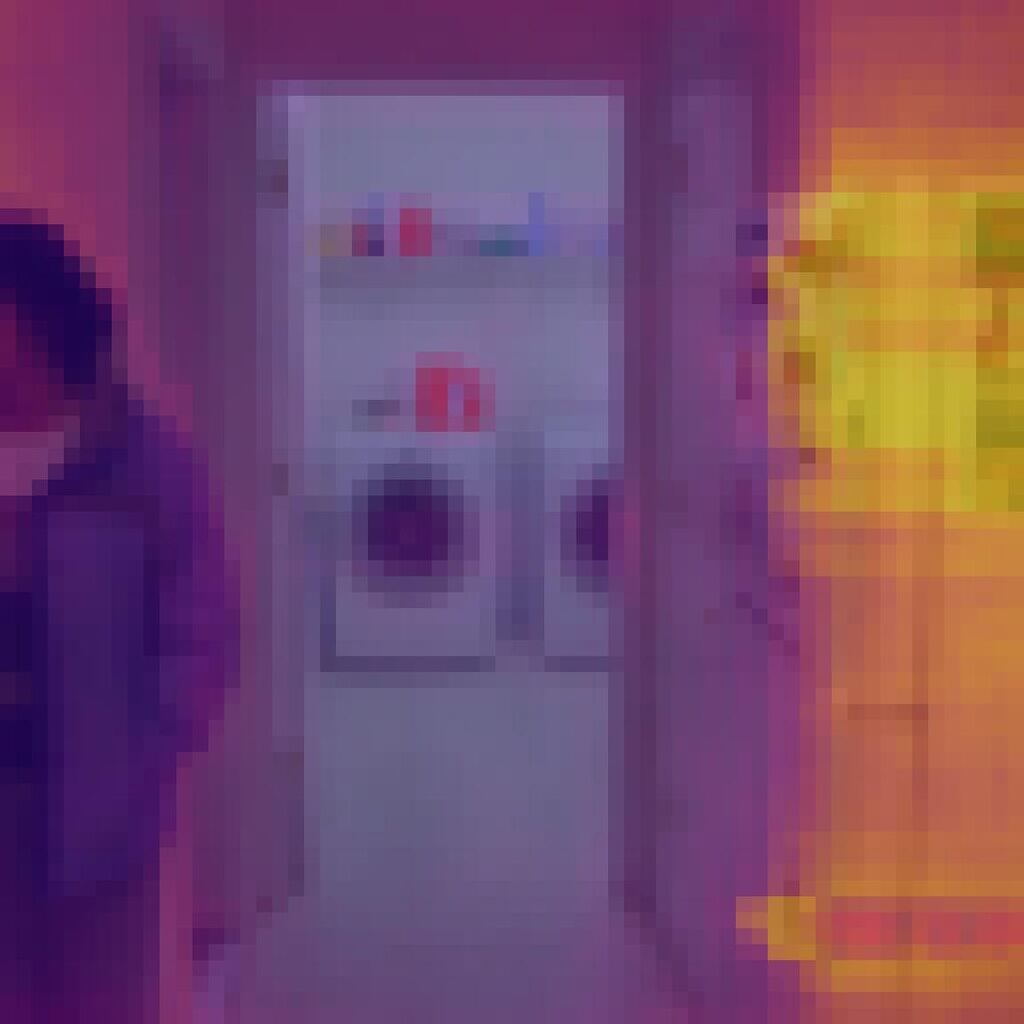}} &
    \fcolorbox{blue}{white}{\includegraphics[width=0.95\linewidth]{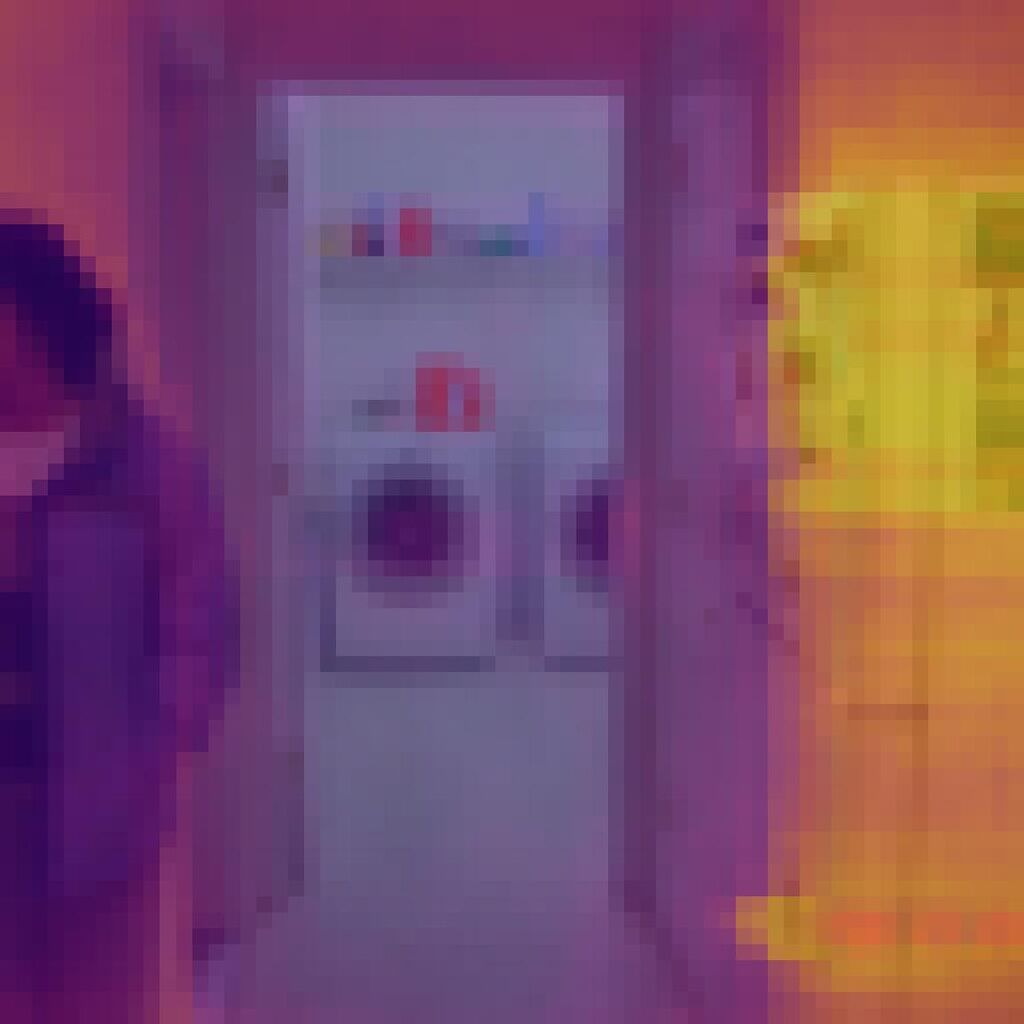}} &
    \fcolorbox{blue}{white}{\includegraphics[width=0.95\linewidth]{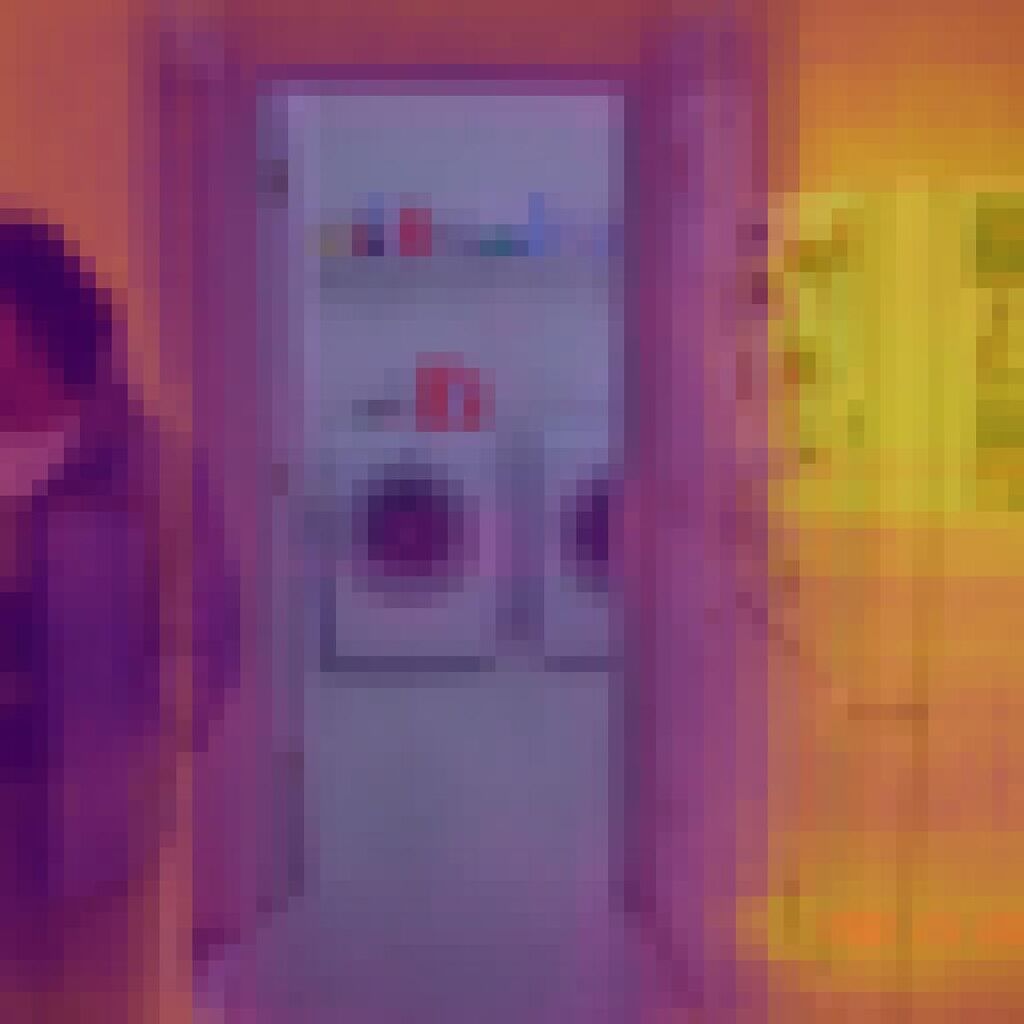}} &
    \fcolorbox{blue}{white}{\includegraphics[width=0.95\linewidth]{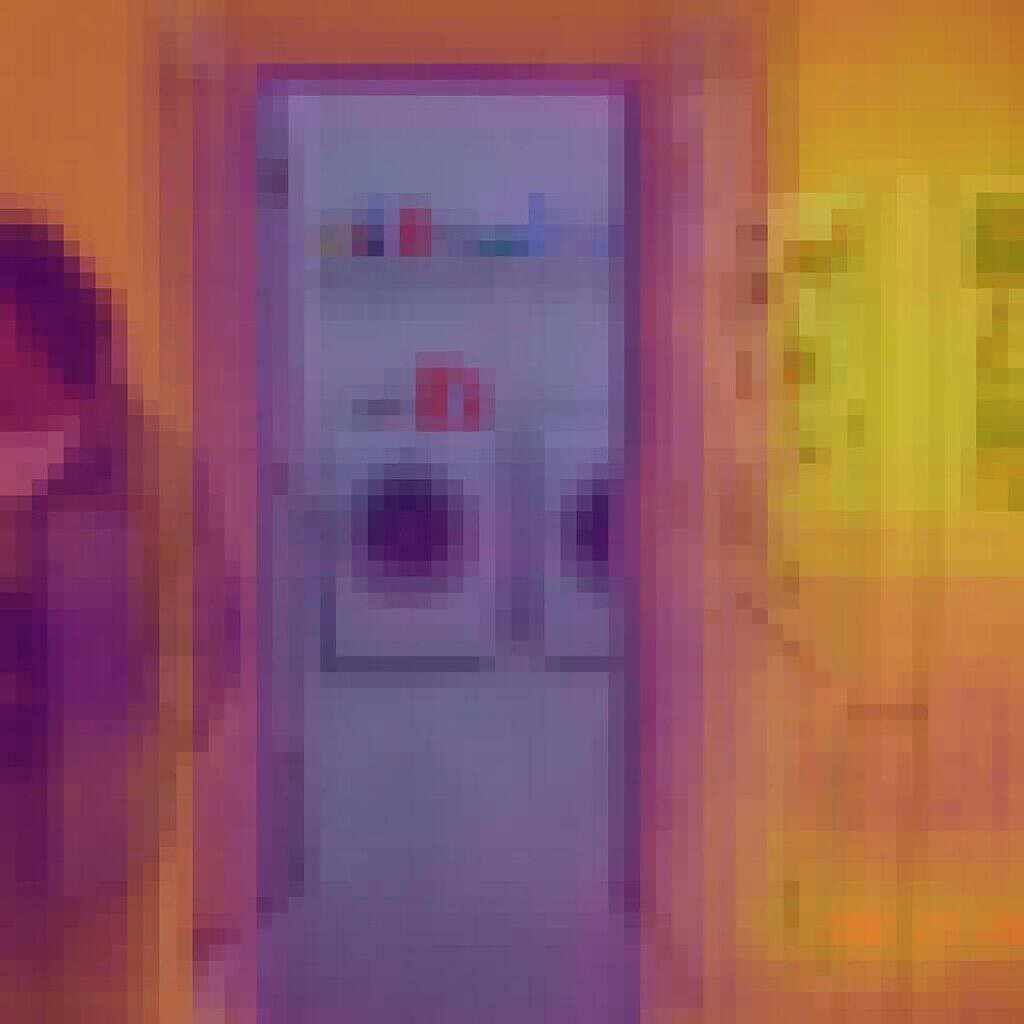}} &
    \fcolorbox{blue}{white}{\includegraphics[width=0.95\linewidth]{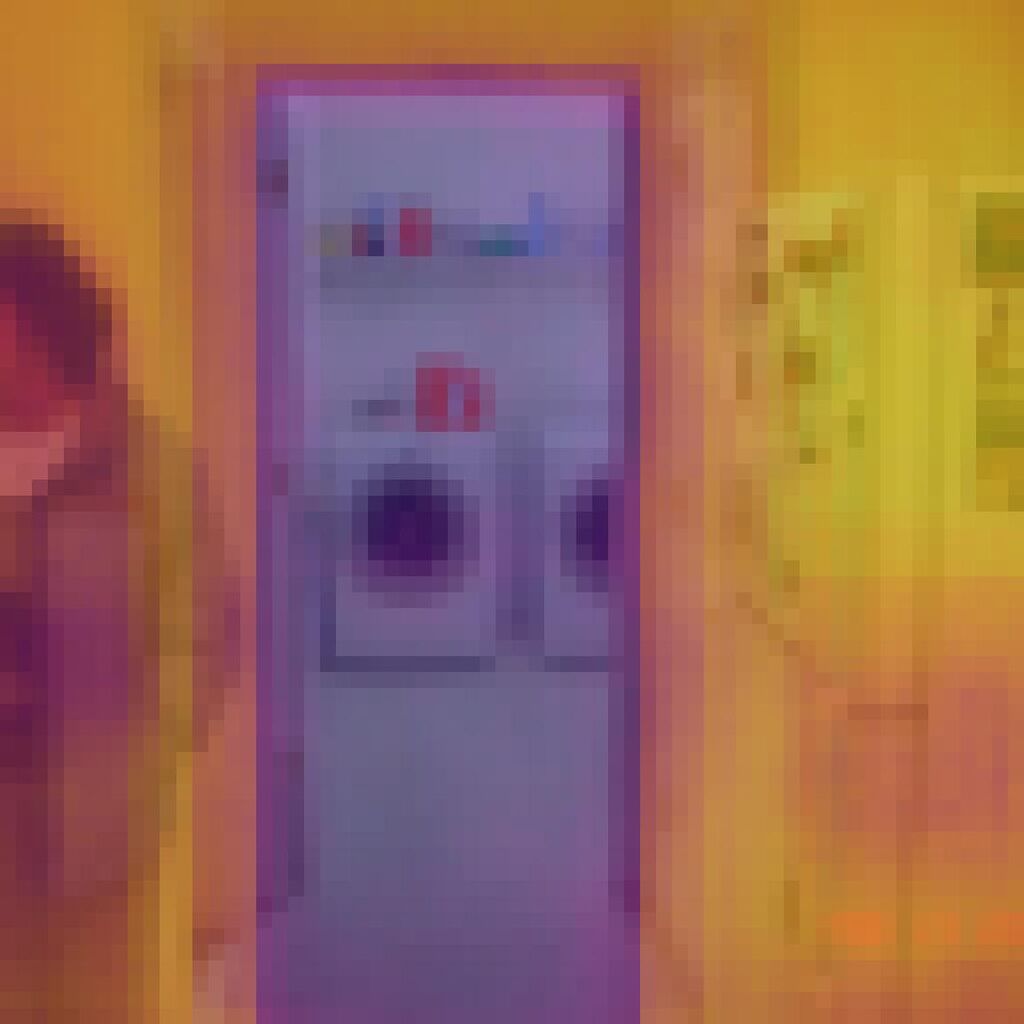}} &
    \fcolorbox{blue}{white}{\includegraphics[width=0.95\linewidth]{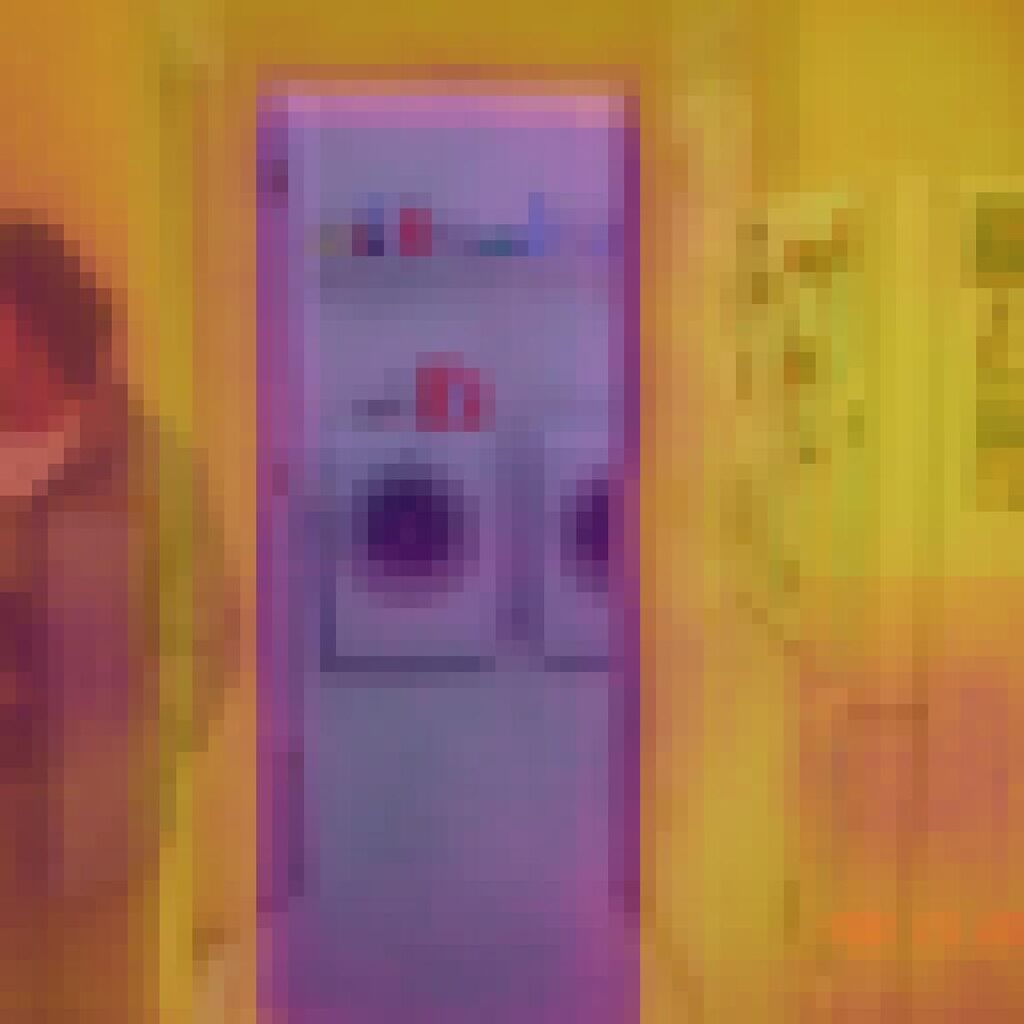}} &
    \fcolorbox{blue}{white}{\includegraphics[width=0.95\linewidth]{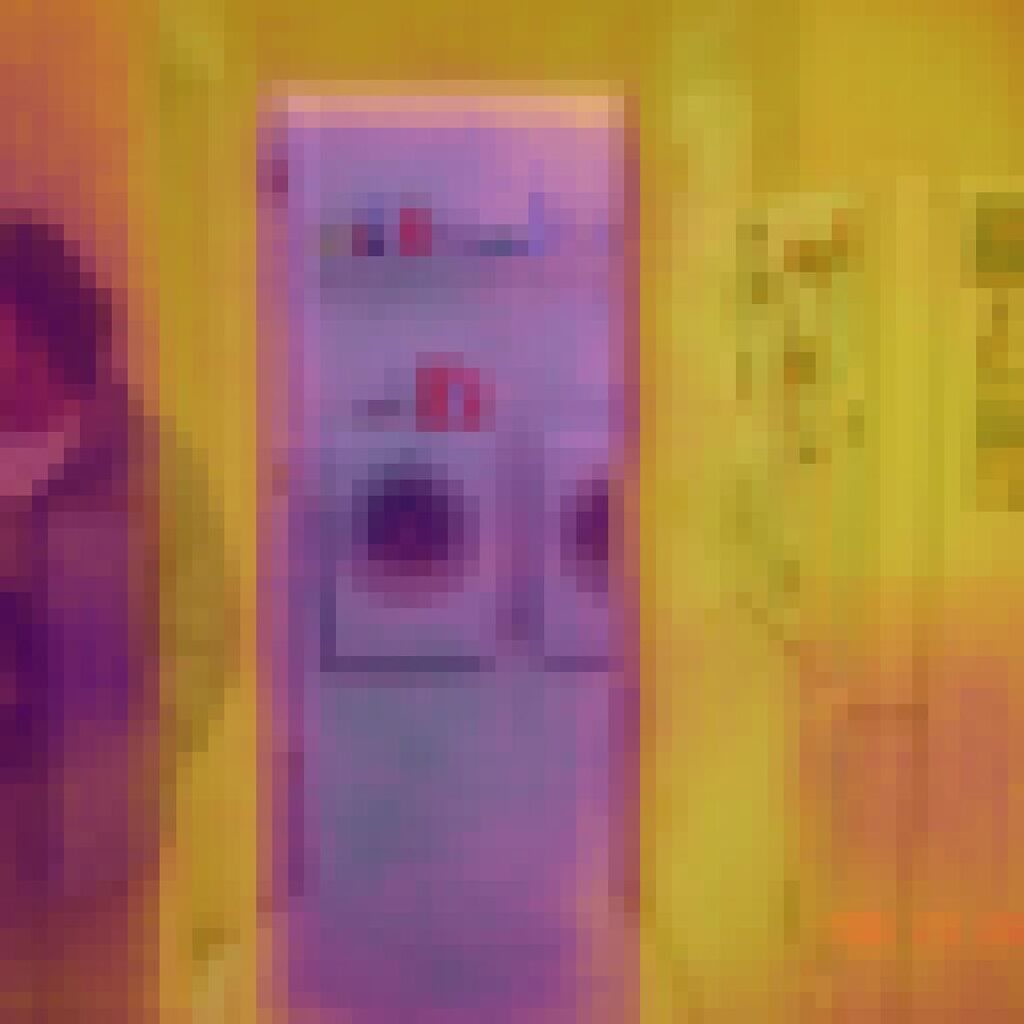}} &
    \fcolorbox{blue}{white}{\includegraphics[width=0.95\linewidth]{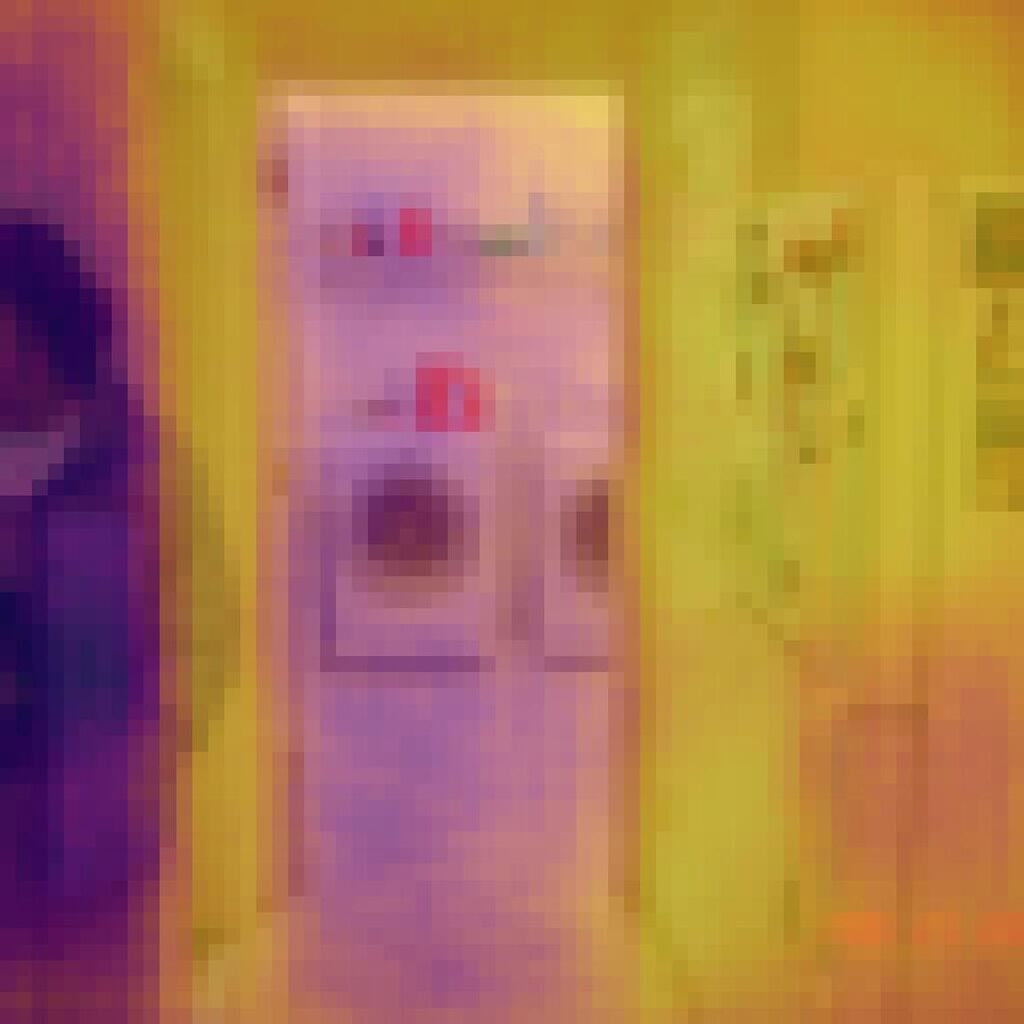}} \\

    t=400 & t=450 & t=500 & t=550 & t=600 & t=650 & t=700 & t=750 & t=800 & t=850 & t=900 \\
    
    \end{tabularx}
    
    \caption{The top row displays the original input image and the corresponding Temporal Stability Matrices (TSMs) for both red and blue query point. The panels below show the evolution of their corresponding Contextual Similarity Maps (CSMs).}

    \label{fig:supp_hierarchical_progress_sample5}
\end{sidewaysfigure*}

\begin{sidewaysfigure*}[htb!] 
    \centering
    
    \begin{tabular}{cc}
        \includegraphics[width=0.2\linewidth]{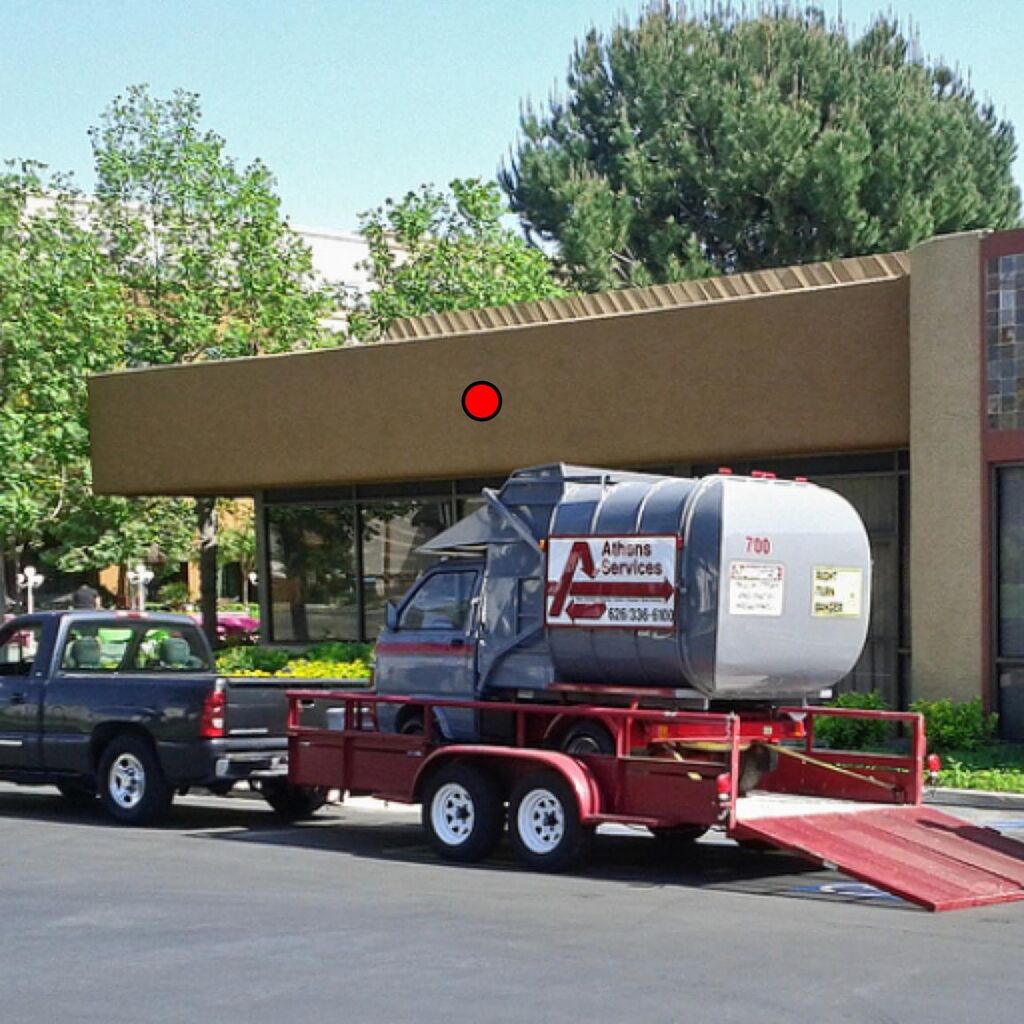} & 
        \includegraphics[width=0.2\linewidth]{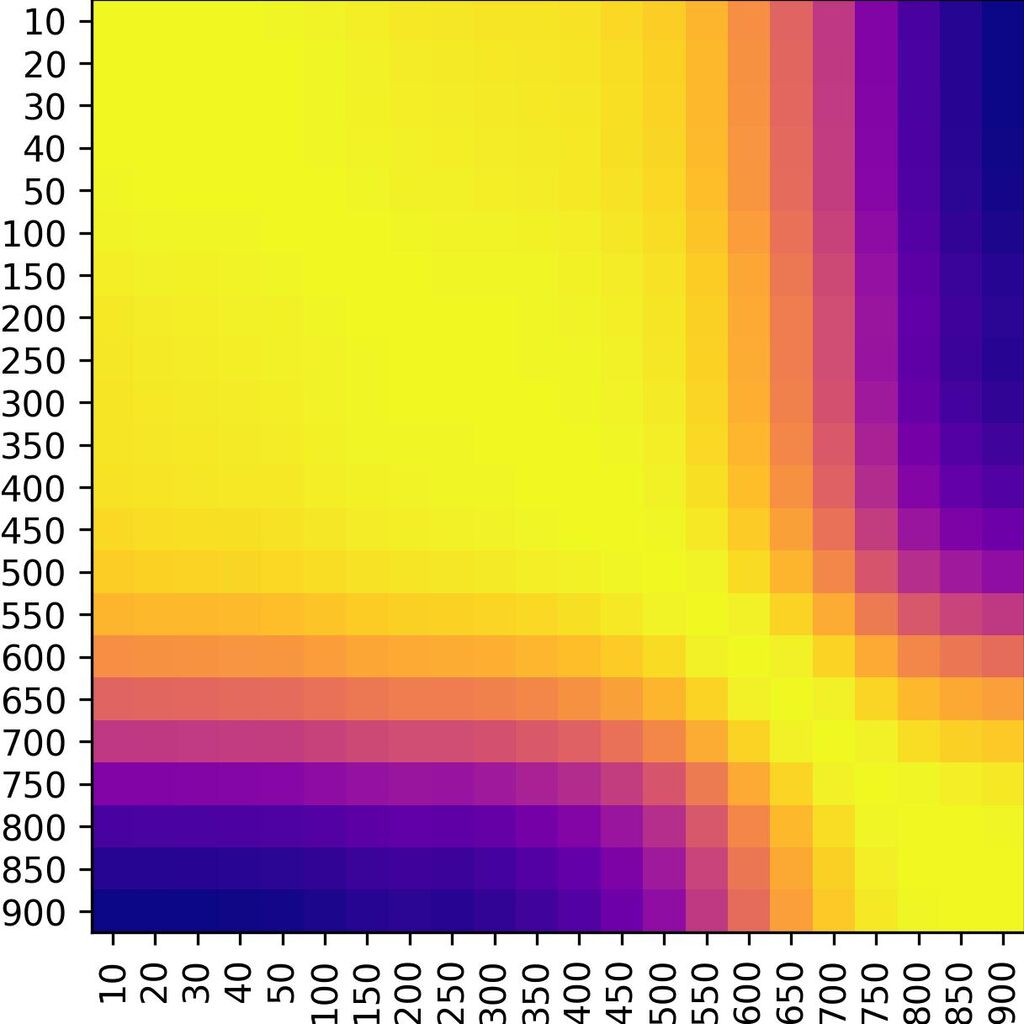} \\
        Input Image & TSM
    \end{tabular}
    
    \vspace{1em} 

    \setlength{\tabcolsep}{2pt} 
    
    \begin{tabularx}{\linewidth}{ *{11}{>{\centering\arraybackslash}p{0.0845\linewidth}} } 
    
    \multicolumn{11}{c}{\textbf{CSMs across Timesteps}} \\
    \noalign{\vspace{0.5em}}

    \fcolorbox{red}{white}{\includegraphics[width=0.95\linewidth]{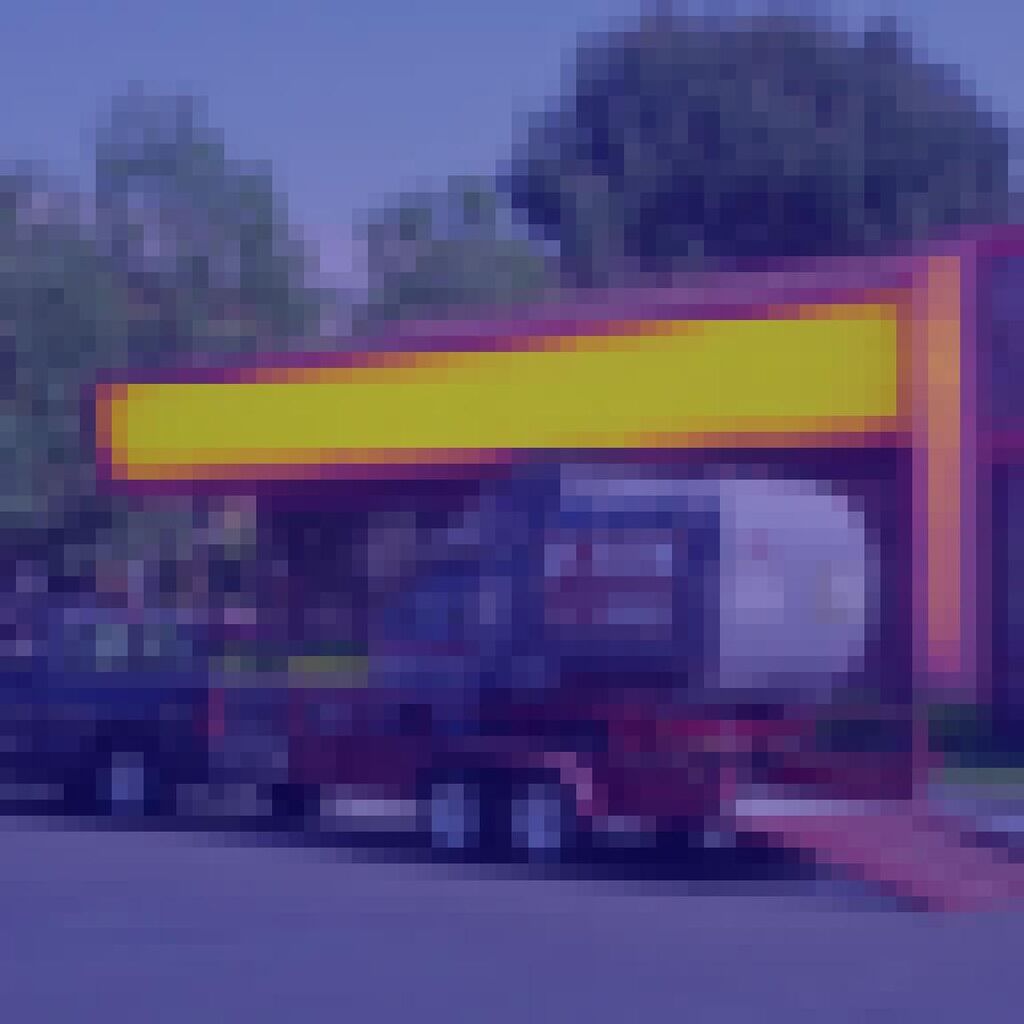}} &
    \fcolorbox{red}{white}{\includegraphics[width=0.95\linewidth]{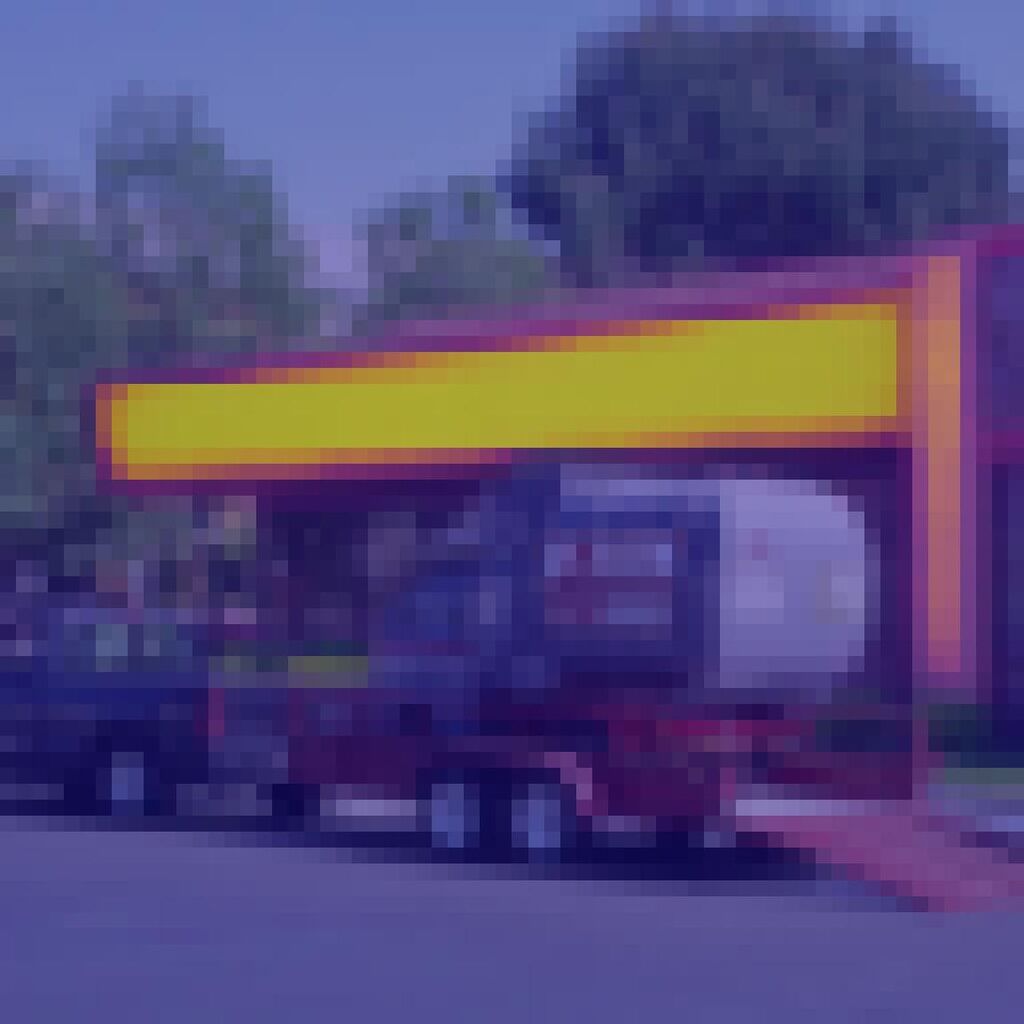}} &
    \fcolorbox{red}{white}{\includegraphics[width=0.95\linewidth]{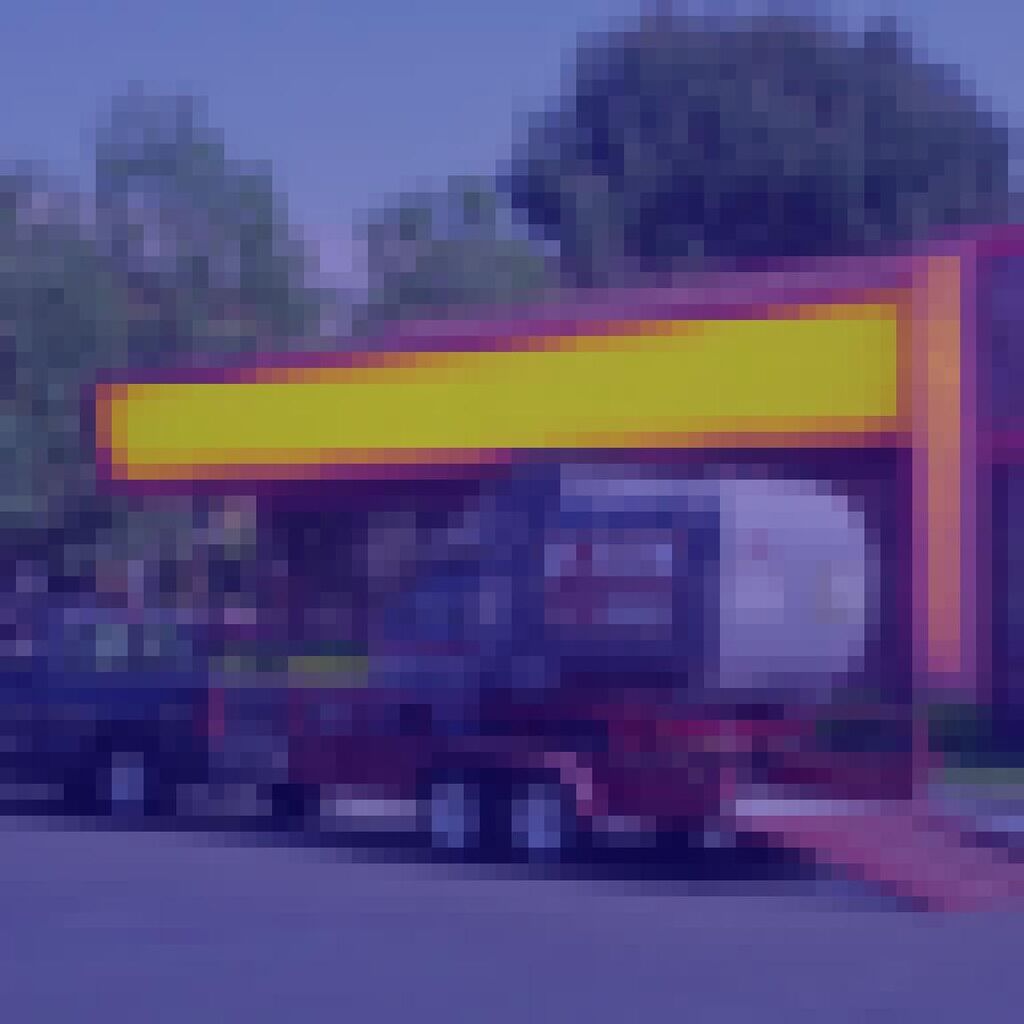}} &
    \fcolorbox{red}{white}{\includegraphics[width=0.95\linewidth]{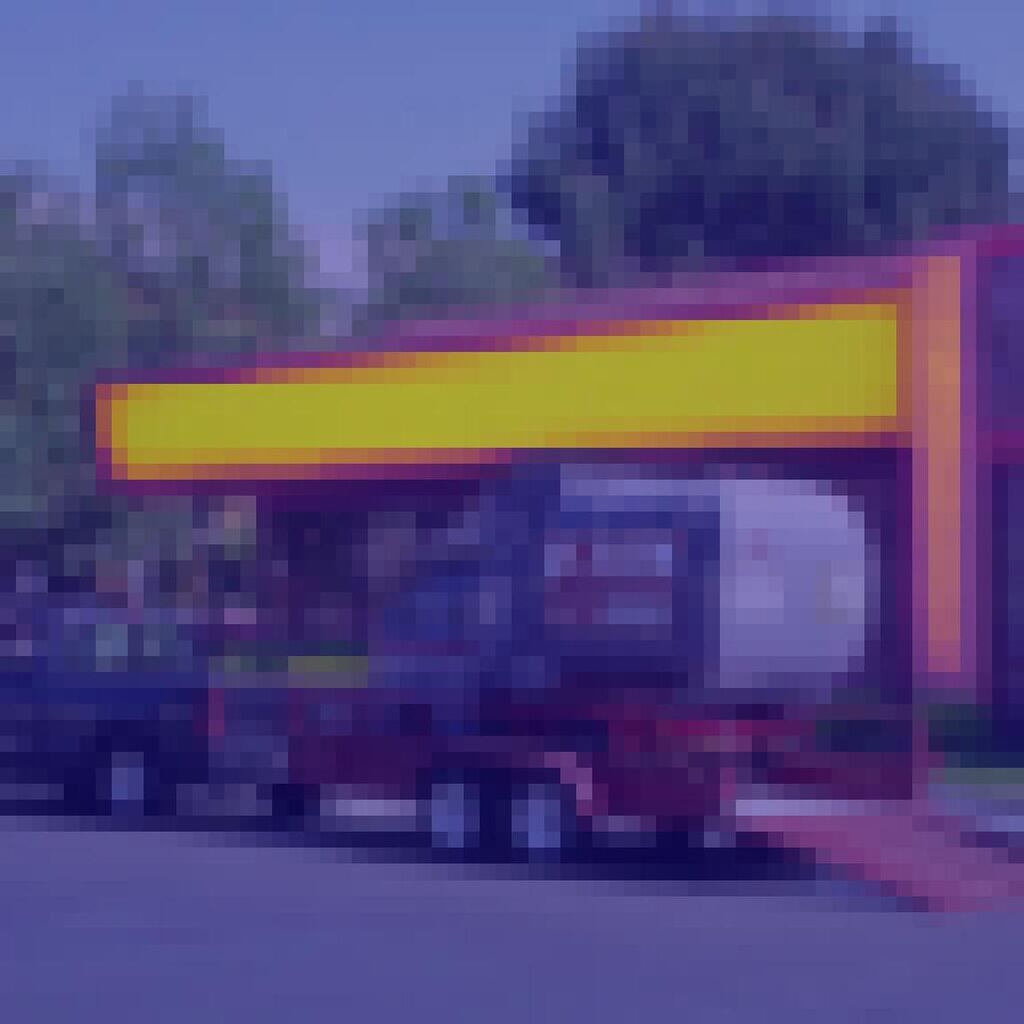}} &
    \fcolorbox{red}{white}{\includegraphics[width=0.95\linewidth]{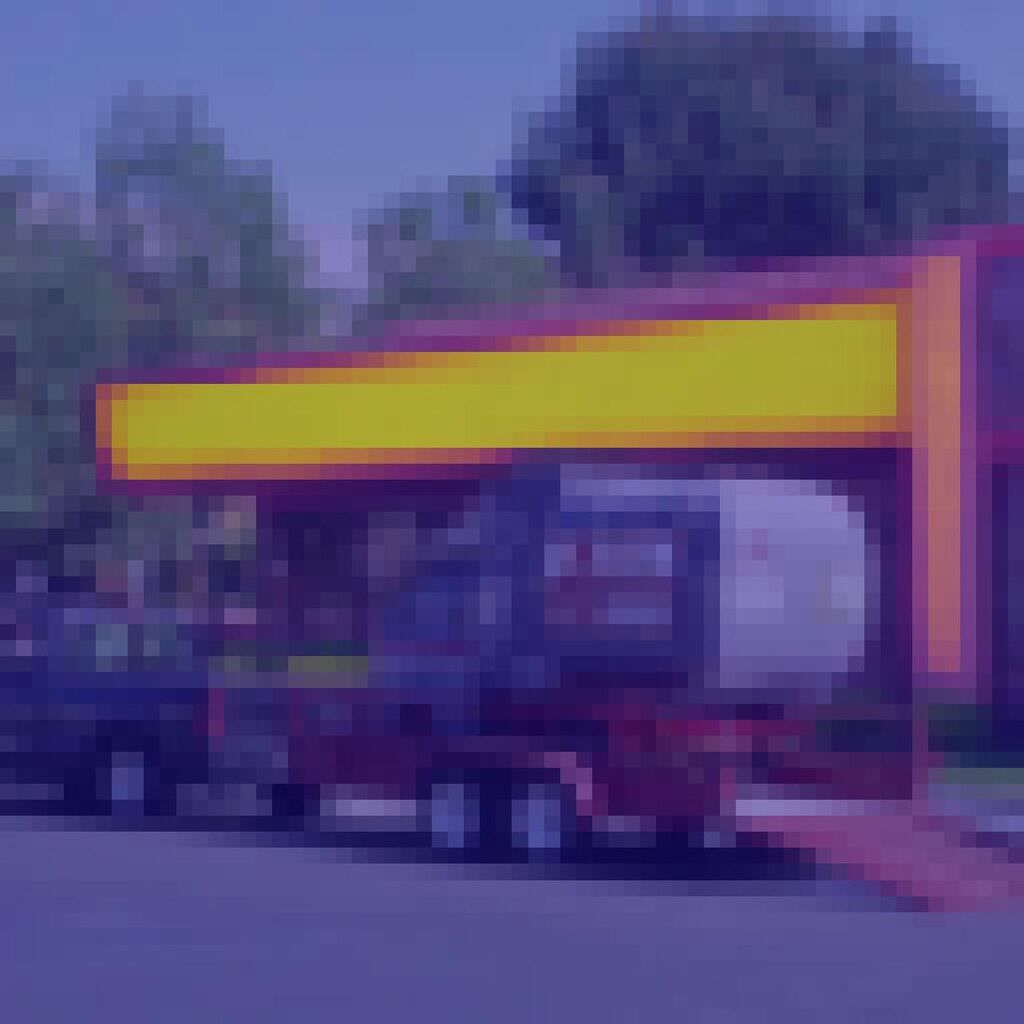}} &
    \fcolorbox{red}{white}{\includegraphics[width=0.95\linewidth]{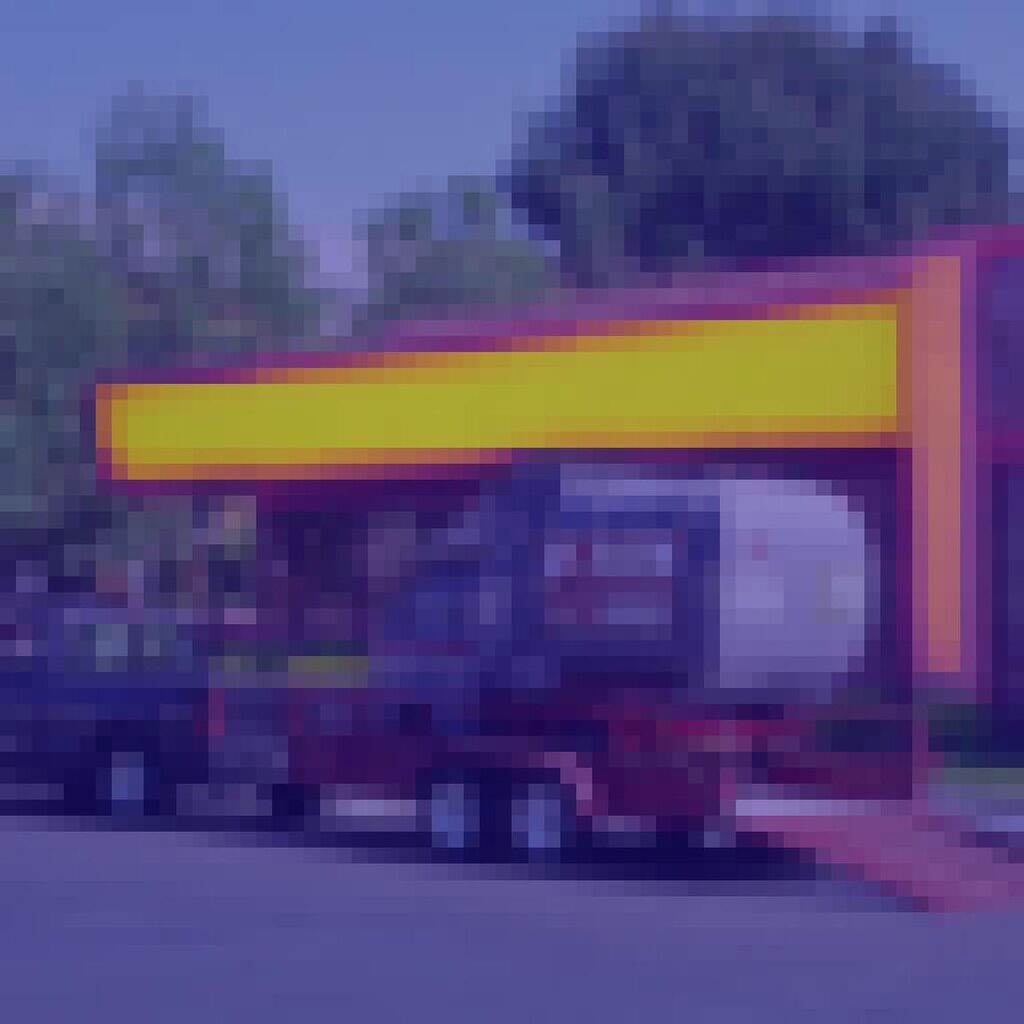}} &
    \fcolorbox{red}{white}{\includegraphics[width=0.95\linewidth]{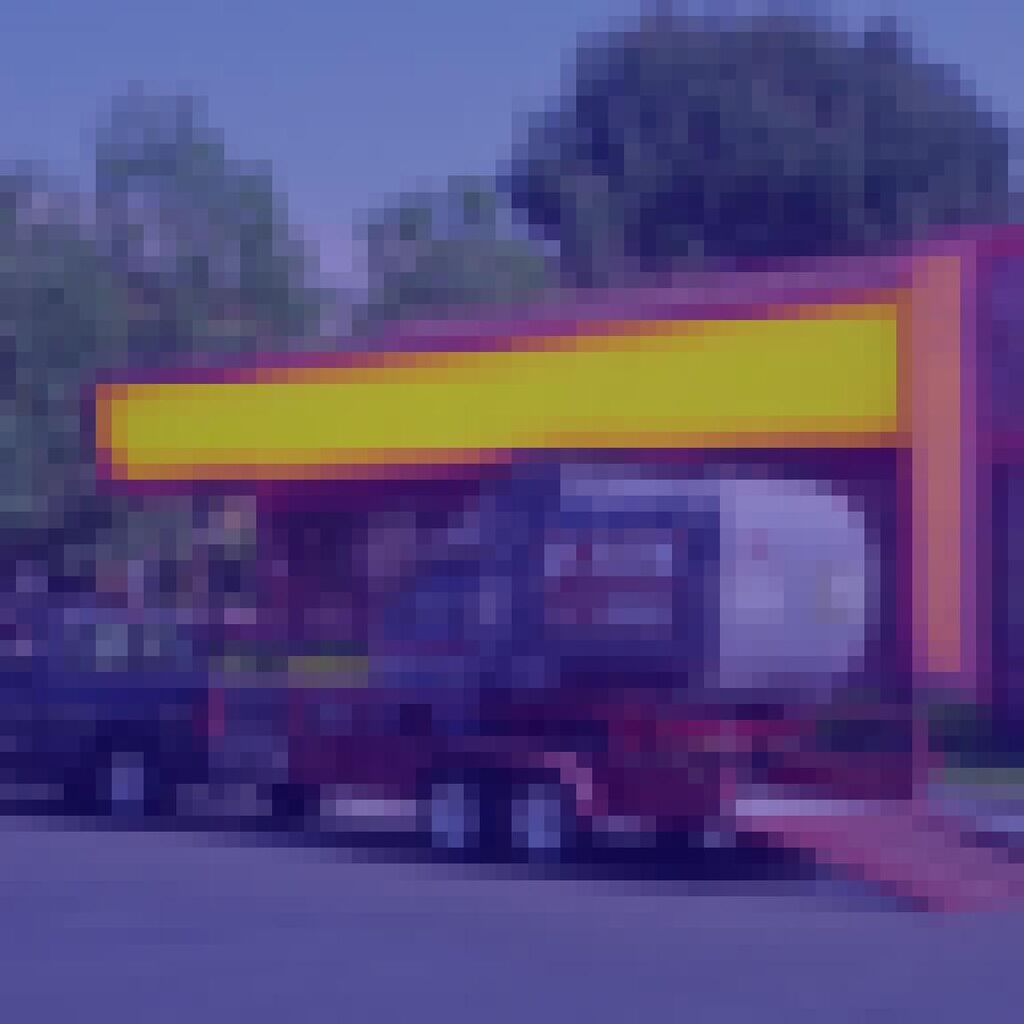}} &
    \fcolorbox{red}{white}{\includegraphics[width=0.95\linewidth]{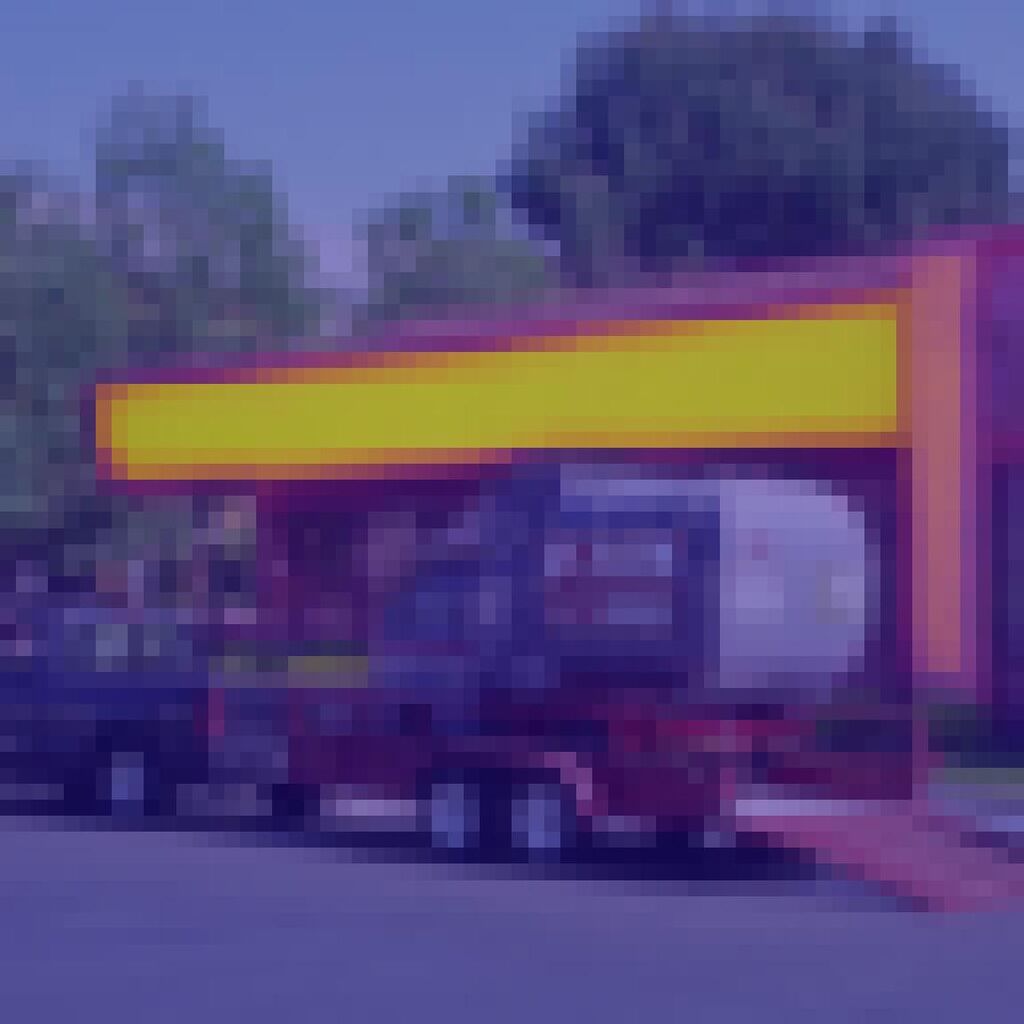}} &
    \fcolorbox{red}{white}{\includegraphics[width=0.95\linewidth]{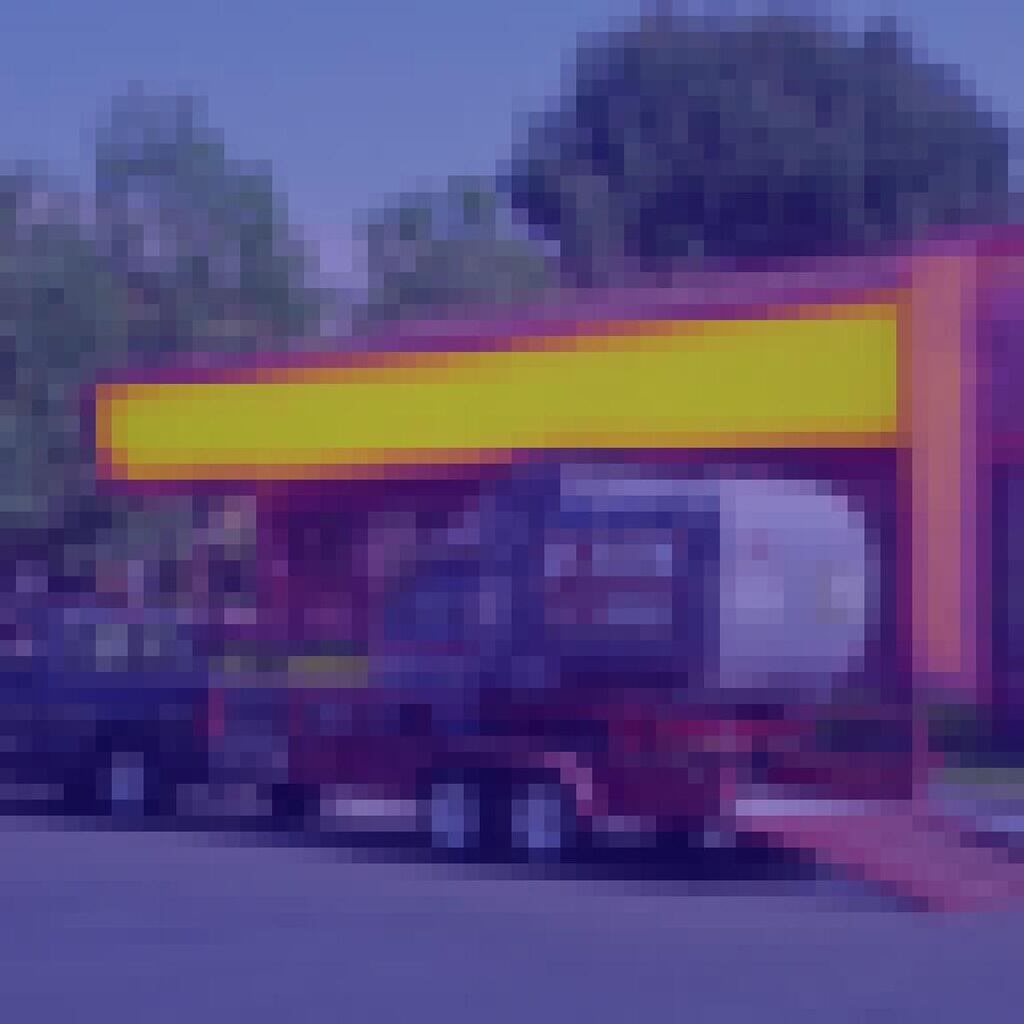}} &
    \fcolorbox{red}{white}{\includegraphics[width=0.95\linewidth]{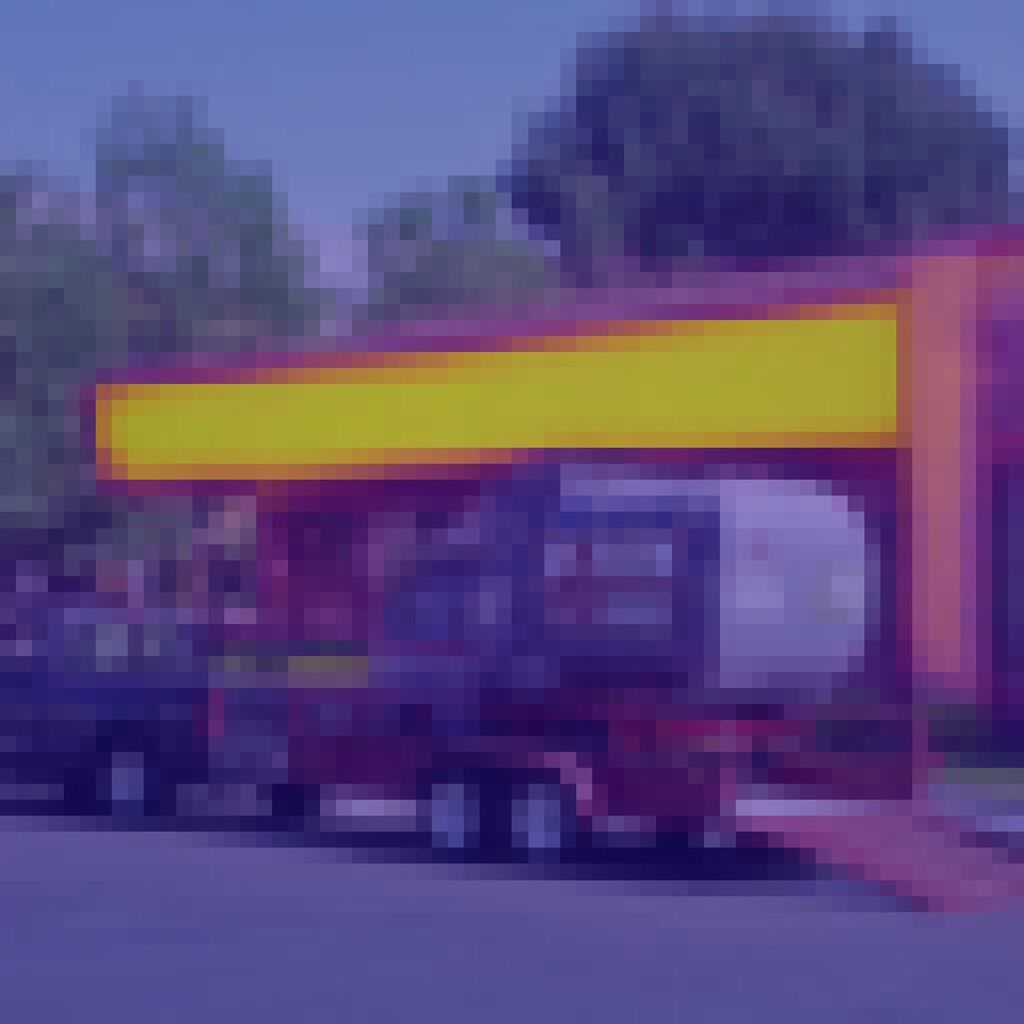}} &
    \fcolorbox{red}{white}{\includegraphics[width=0.95\linewidth]{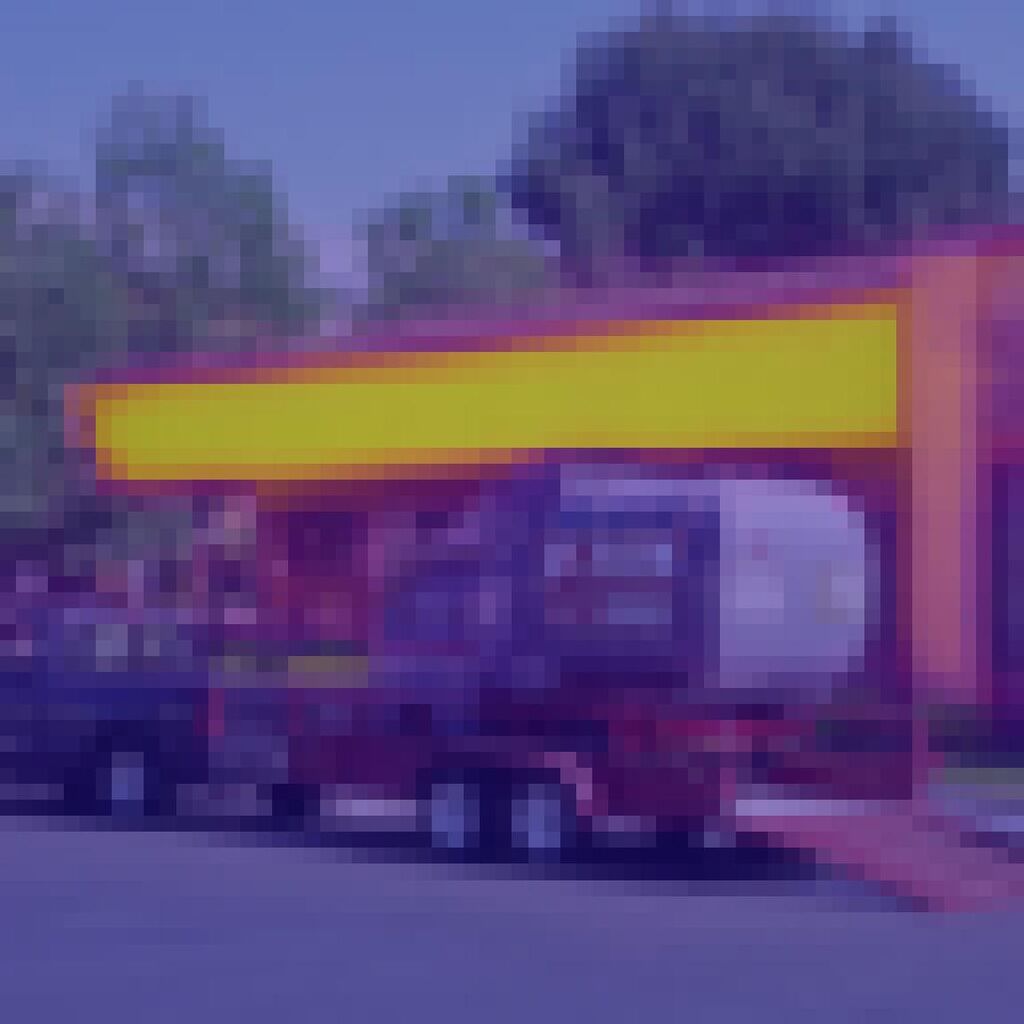}} \\
    
    t=10 & t=20 & t=30 & t=40 & t=50 & t=100 & t=150 & t=200 & t=250 & t=300 & t=350 \\[1em]
    
    \fcolorbox{red}{white}{\includegraphics[width=0.95\linewidth]{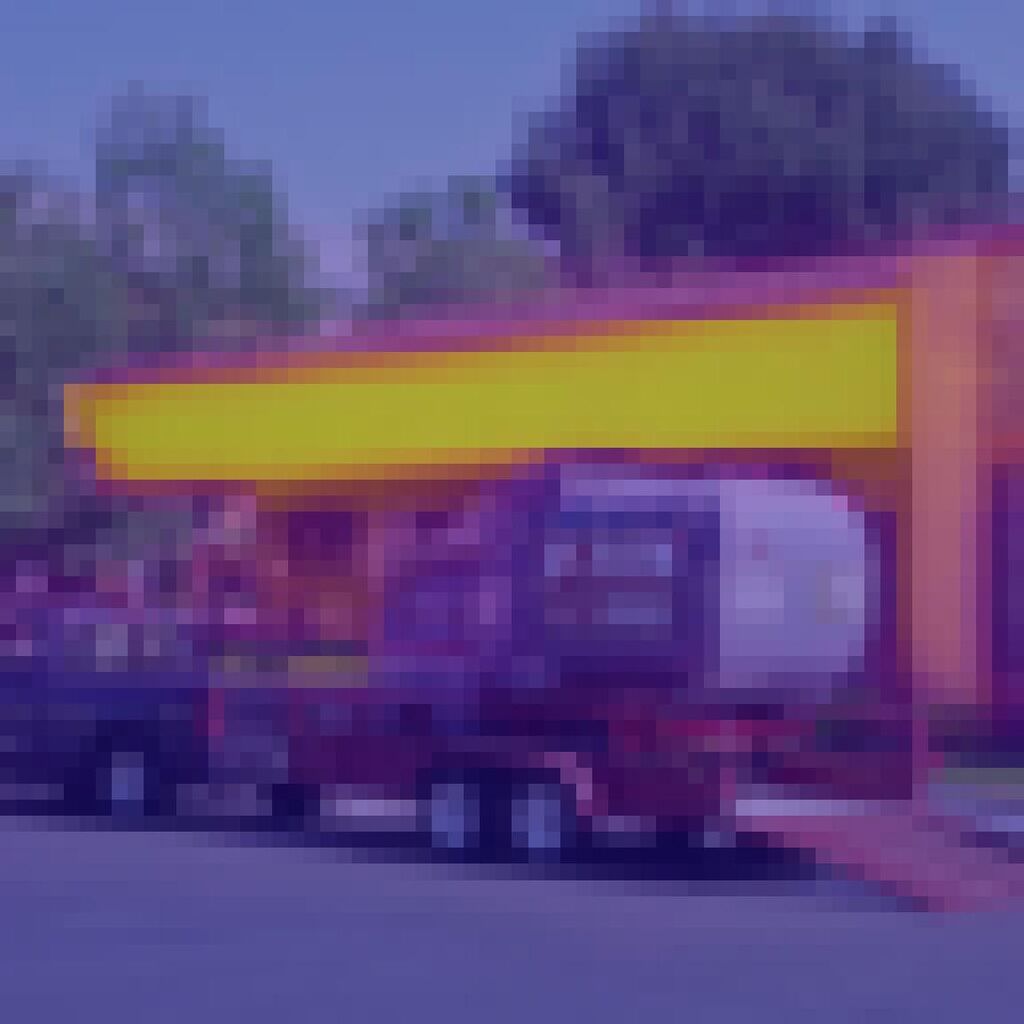}} &
    \fcolorbox{red}{white}{\includegraphics[width=0.95\linewidth]{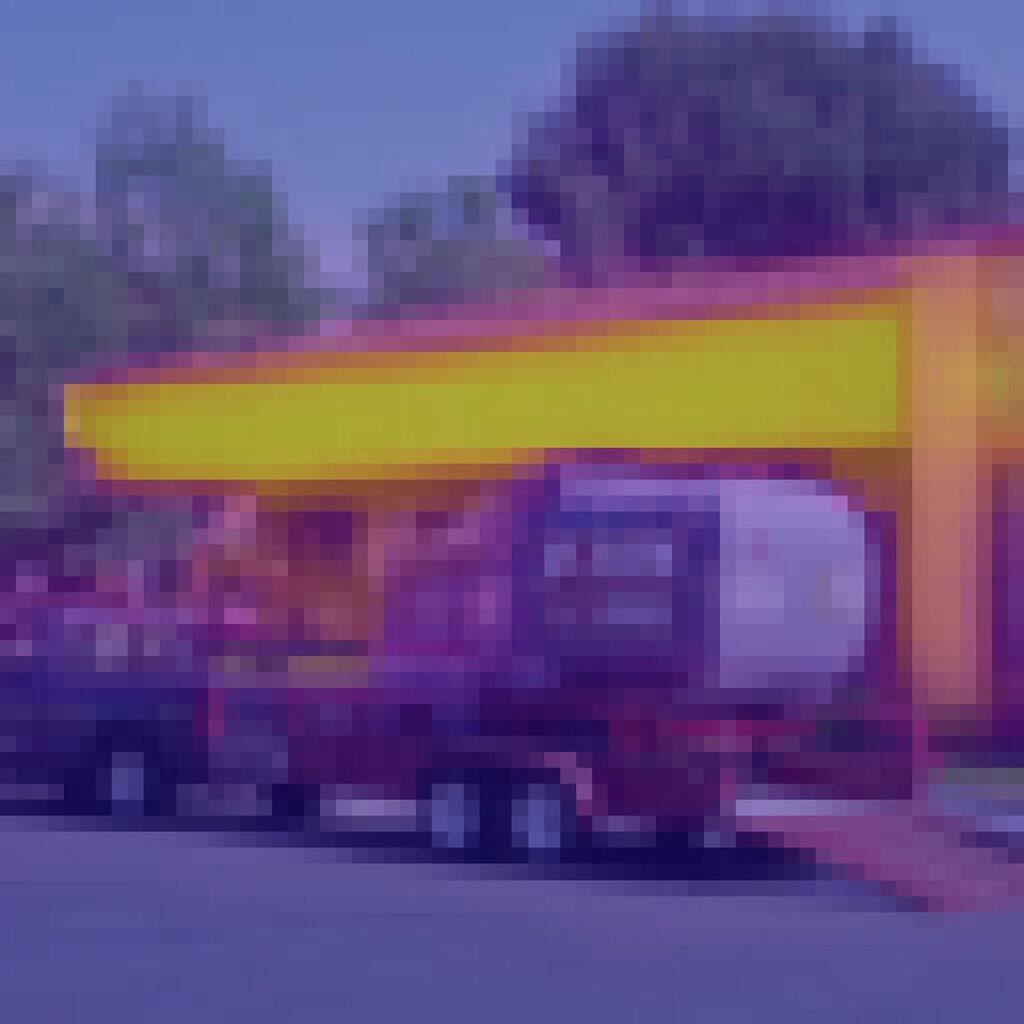}} &
    \fcolorbox{red}{white}{\includegraphics[width=0.95\linewidth]{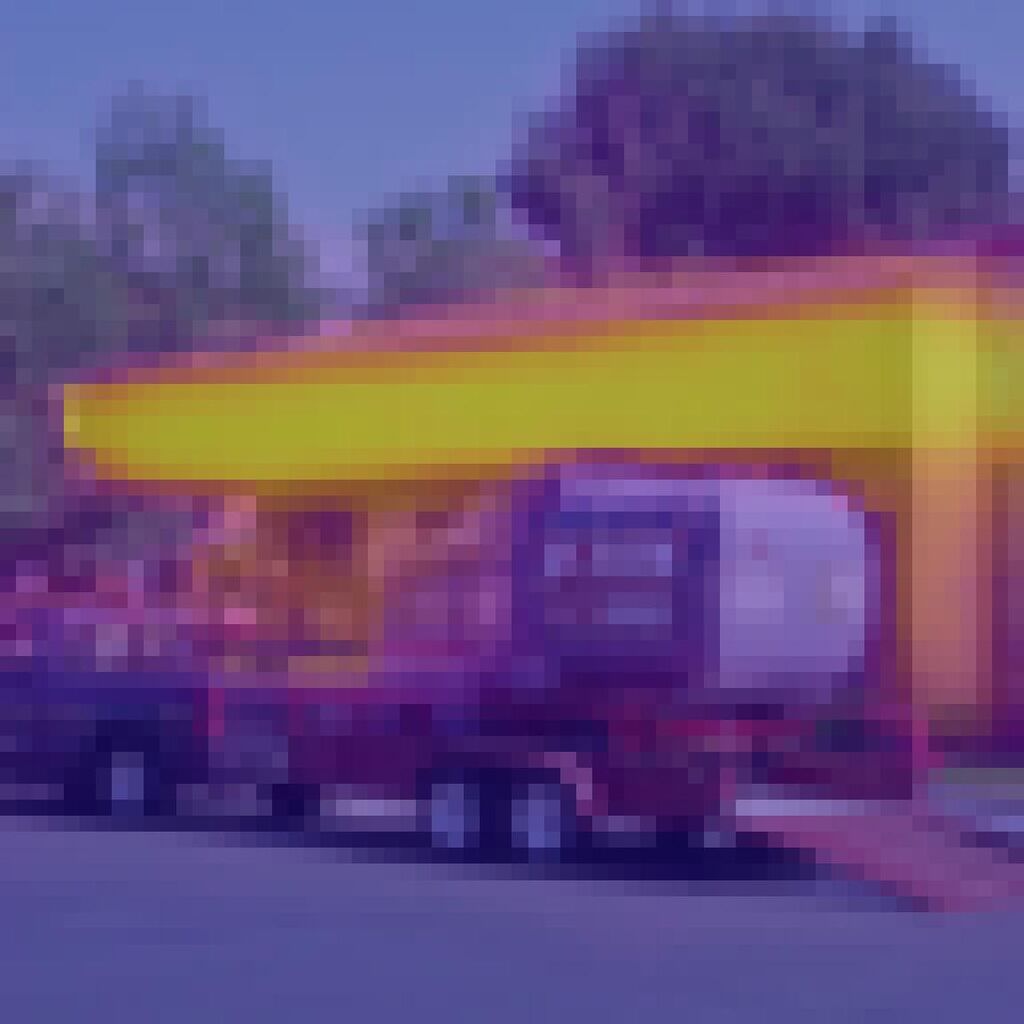}} &
    \fcolorbox{red}{white}{\includegraphics[width=0.95\linewidth]{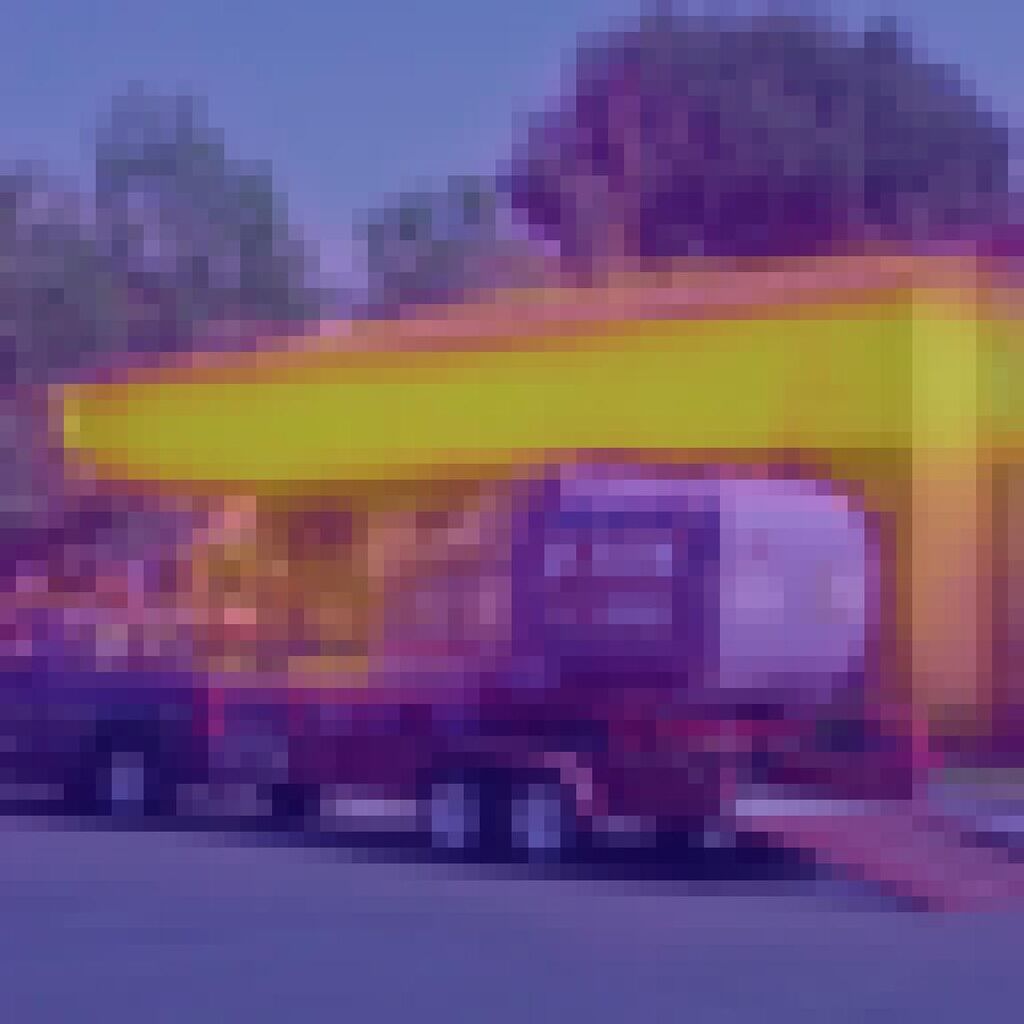}} &
    \fcolorbox{red}{white}{\includegraphics[width=0.95\linewidth]{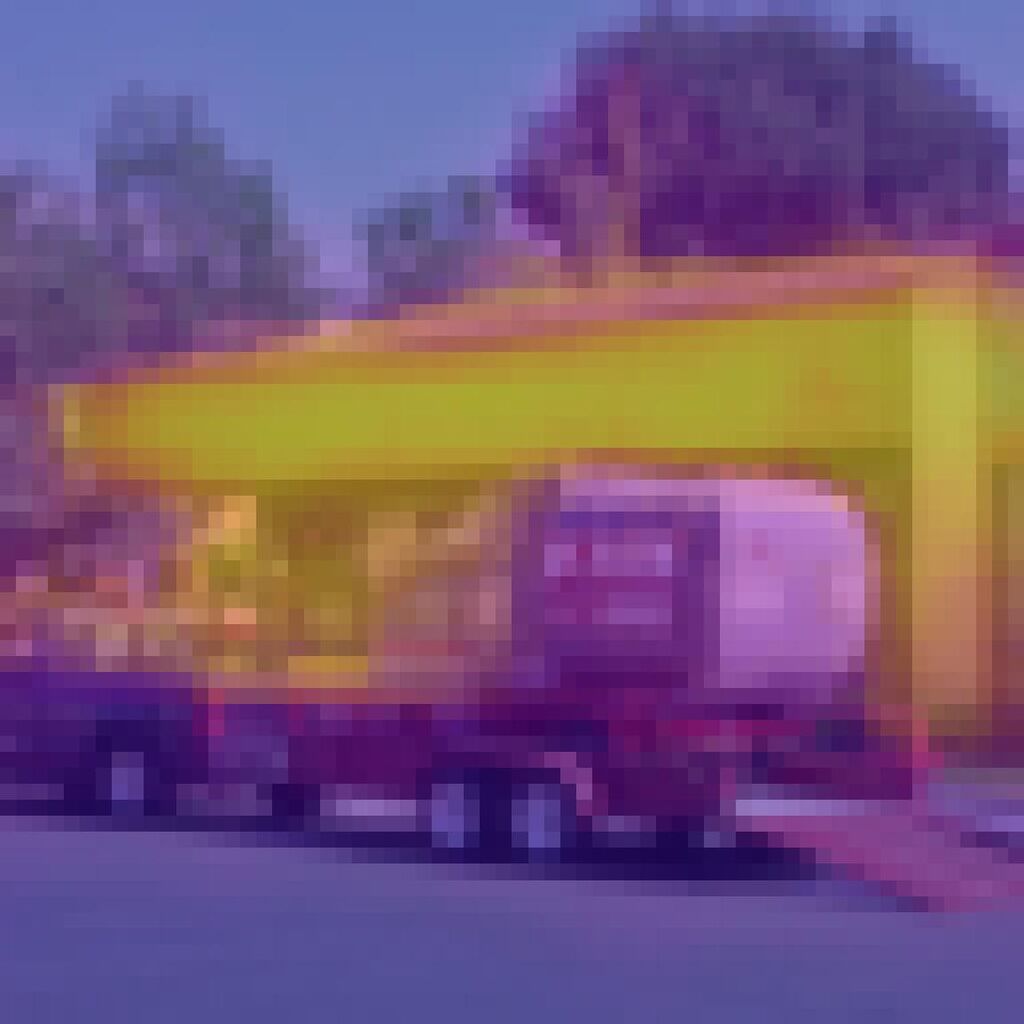}} &
    \fcolorbox{red}{white}{\includegraphics[width=0.95\linewidth]{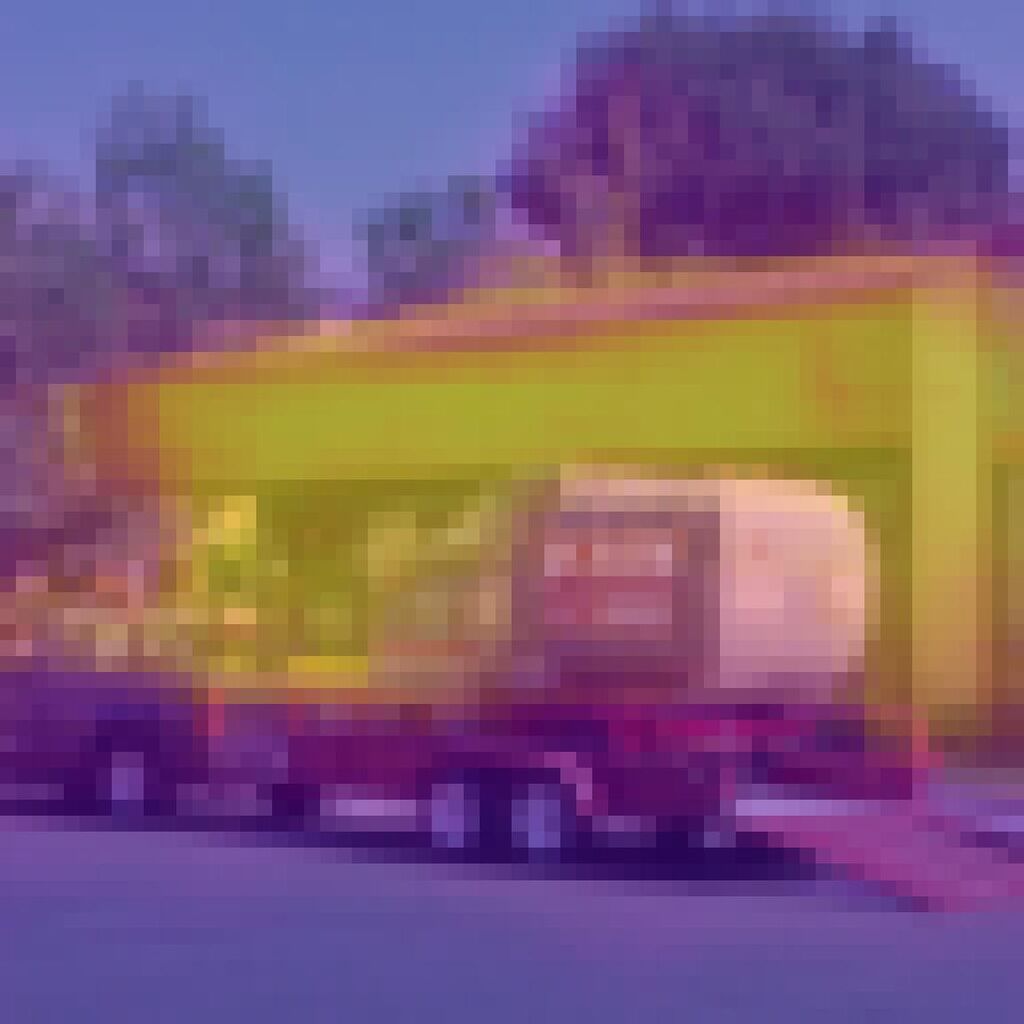}} &
    \fcolorbox{red}{white}{\includegraphics[width=0.95\linewidth]{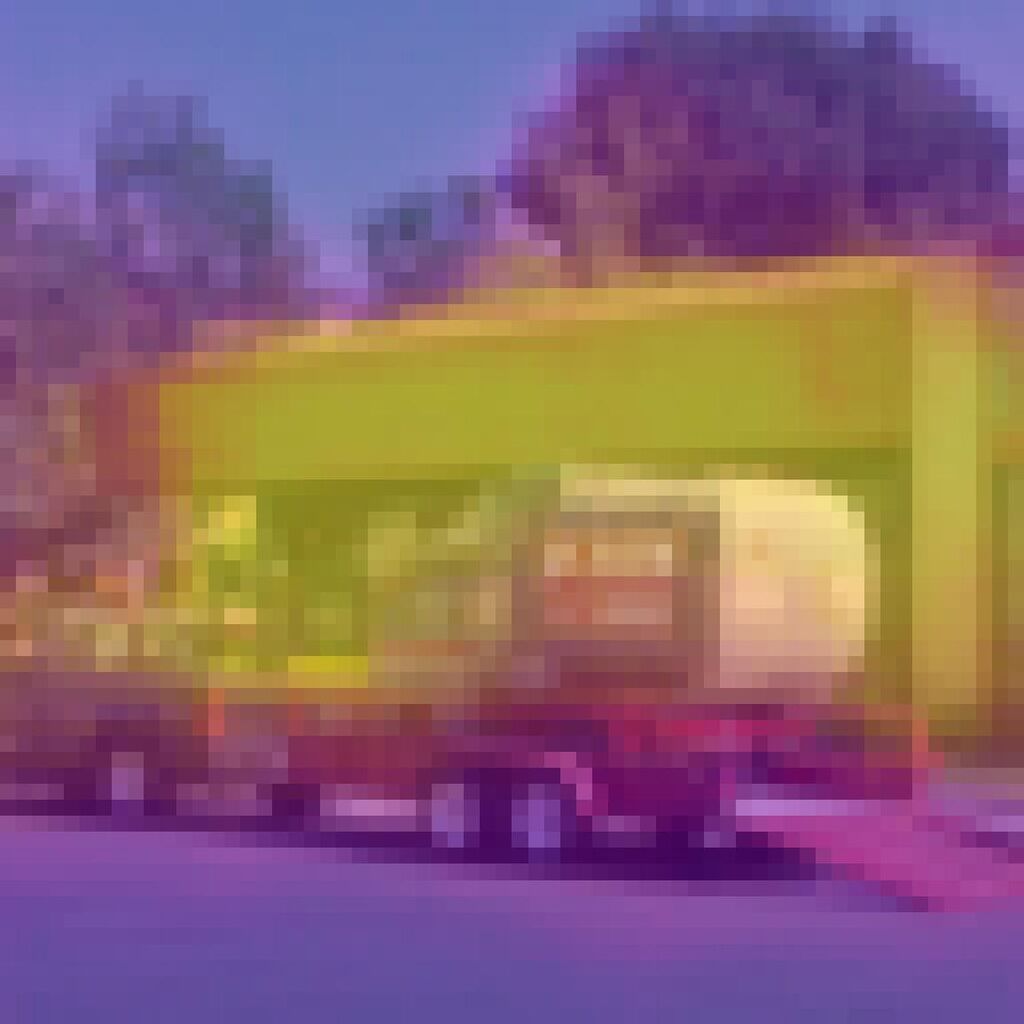}} &
    \fcolorbox{red}{white}{\includegraphics[width=0.95\linewidth]{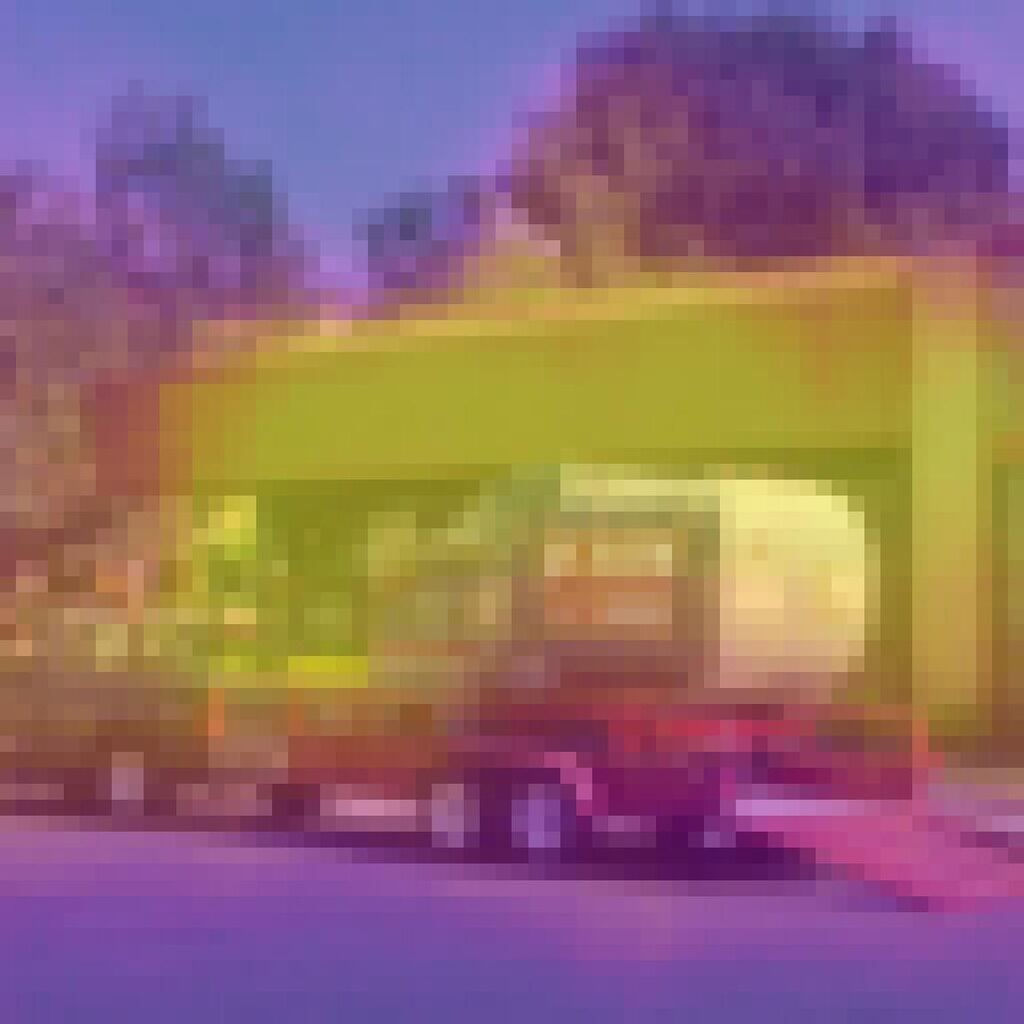}} &
    \fcolorbox{red}{white}{\includegraphics[width=0.95\linewidth]{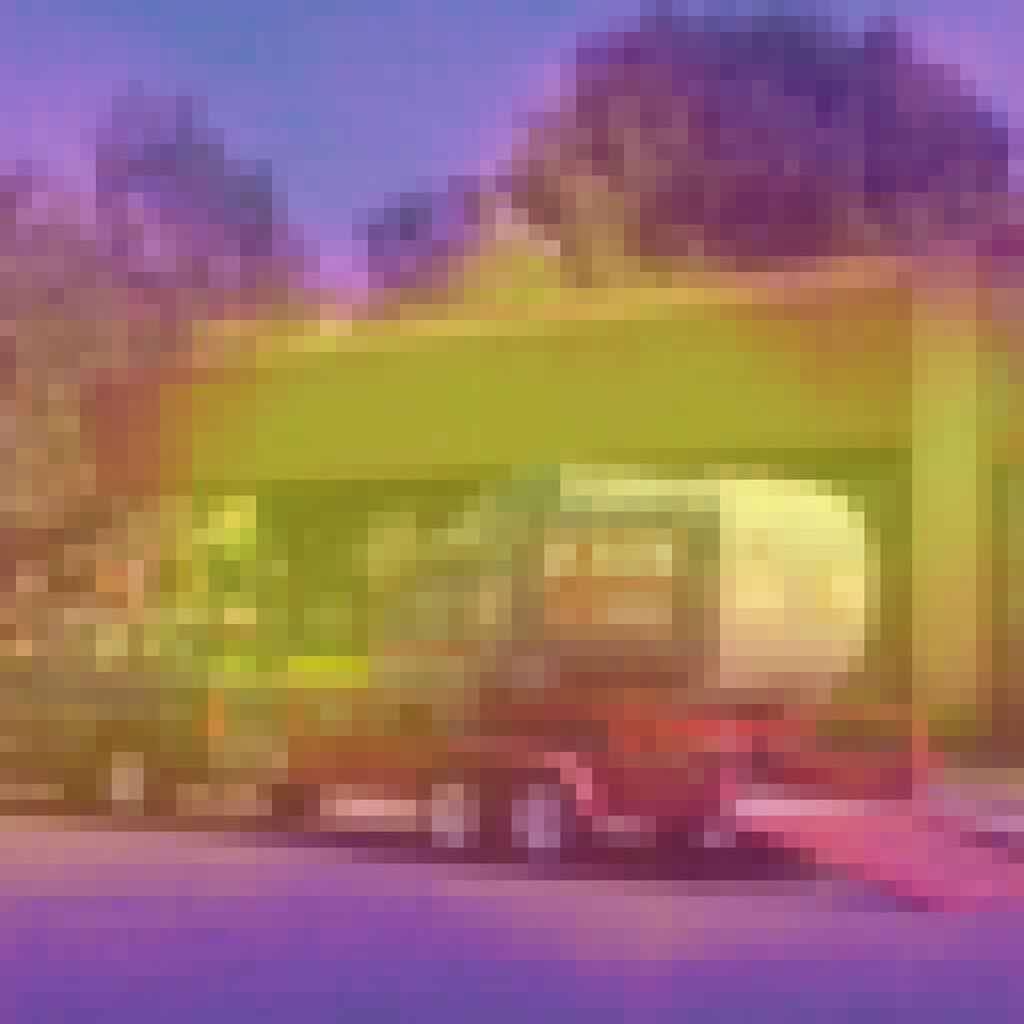}} &
    \fcolorbox{red}{white}{\includegraphics[width=0.95\linewidth]{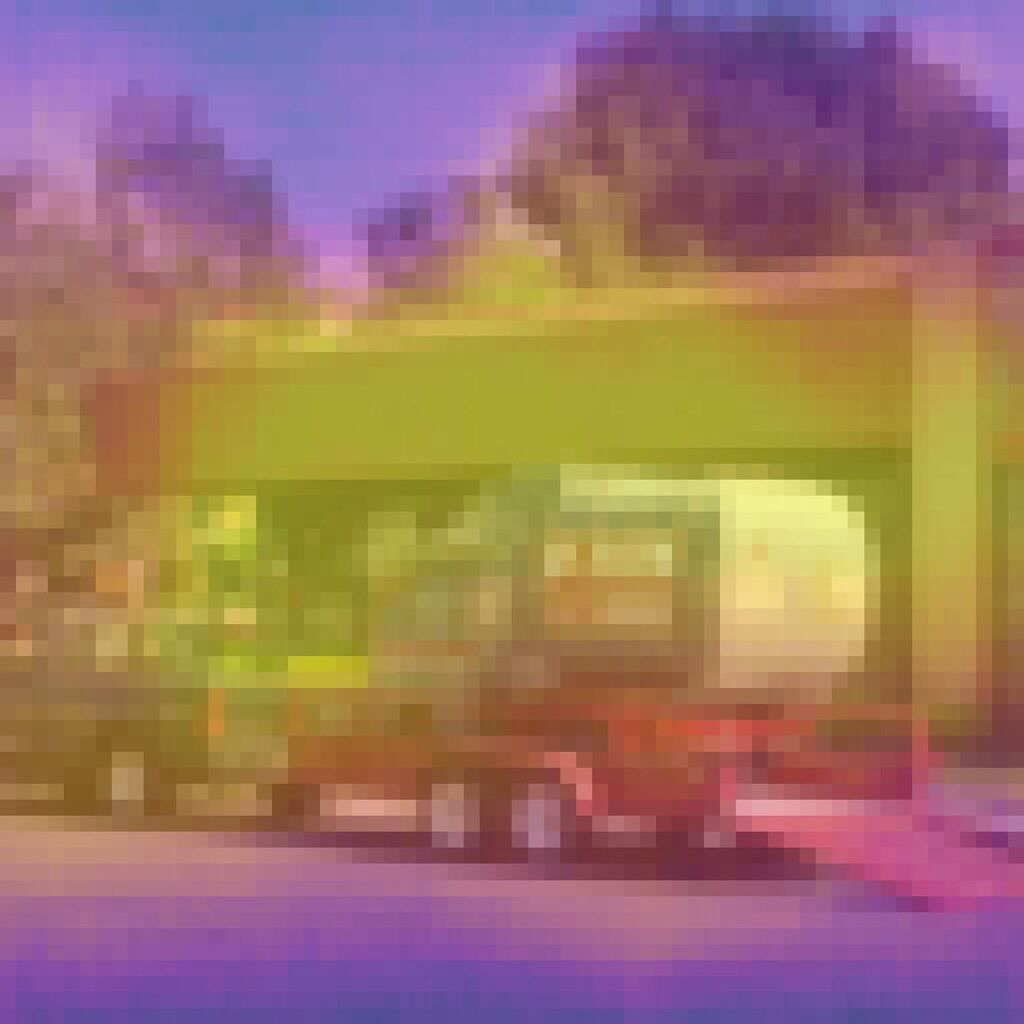}} &
    \fcolorbox{red}{white}{\includegraphics[width=0.95\linewidth]{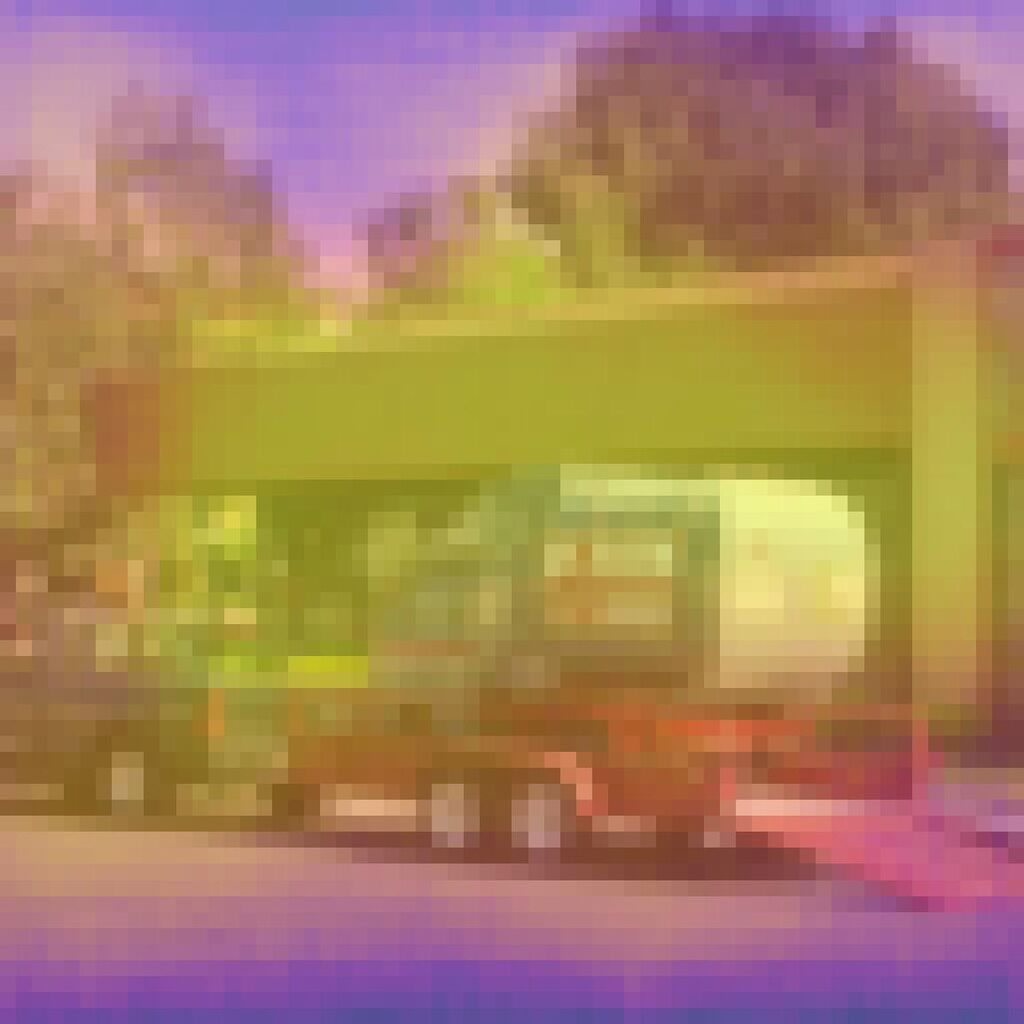}} \\
    
    t=400 & t=450 & t=500 & t=550 & t=600 & t=650 & t=700 & t=750 & t=800 & t=850 & t=900 \\
    
    \end{tabularx}

    \caption{The top row displays the original input image and the corresponding Temporal Stability Matrix (TSM) for the query point (in red). The panels below show the evolution of the Contextual Similarity Maps (CSMs) across the denoising process.}
    \label{fig:supp_hierarchical_progress_sample1}
\end{sidewaysfigure*}

\begin{sidewaysfigure*}[htb!] 
    \centering
    
    \begin{tabular}{cc}
        \includegraphics[width=0.2\linewidth]{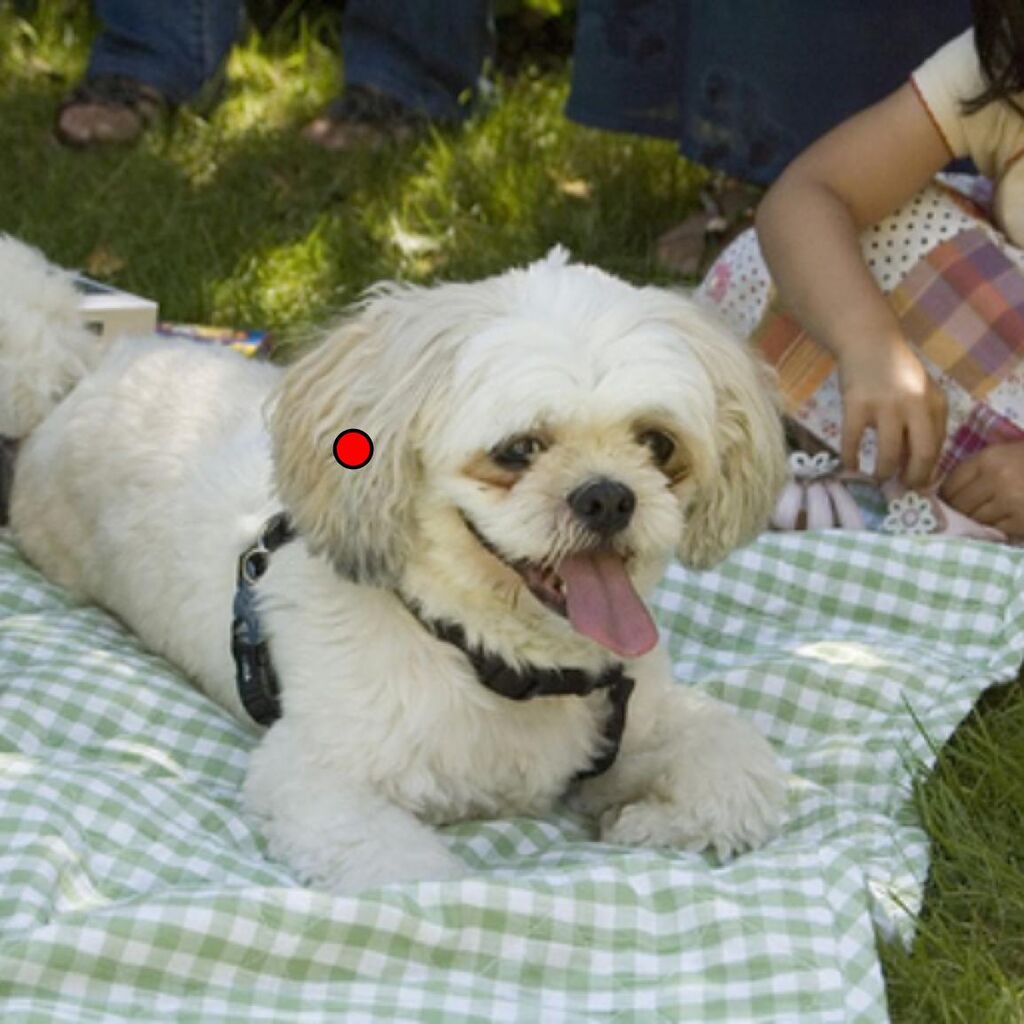} & 
        \includegraphics[width=0.2\linewidth]{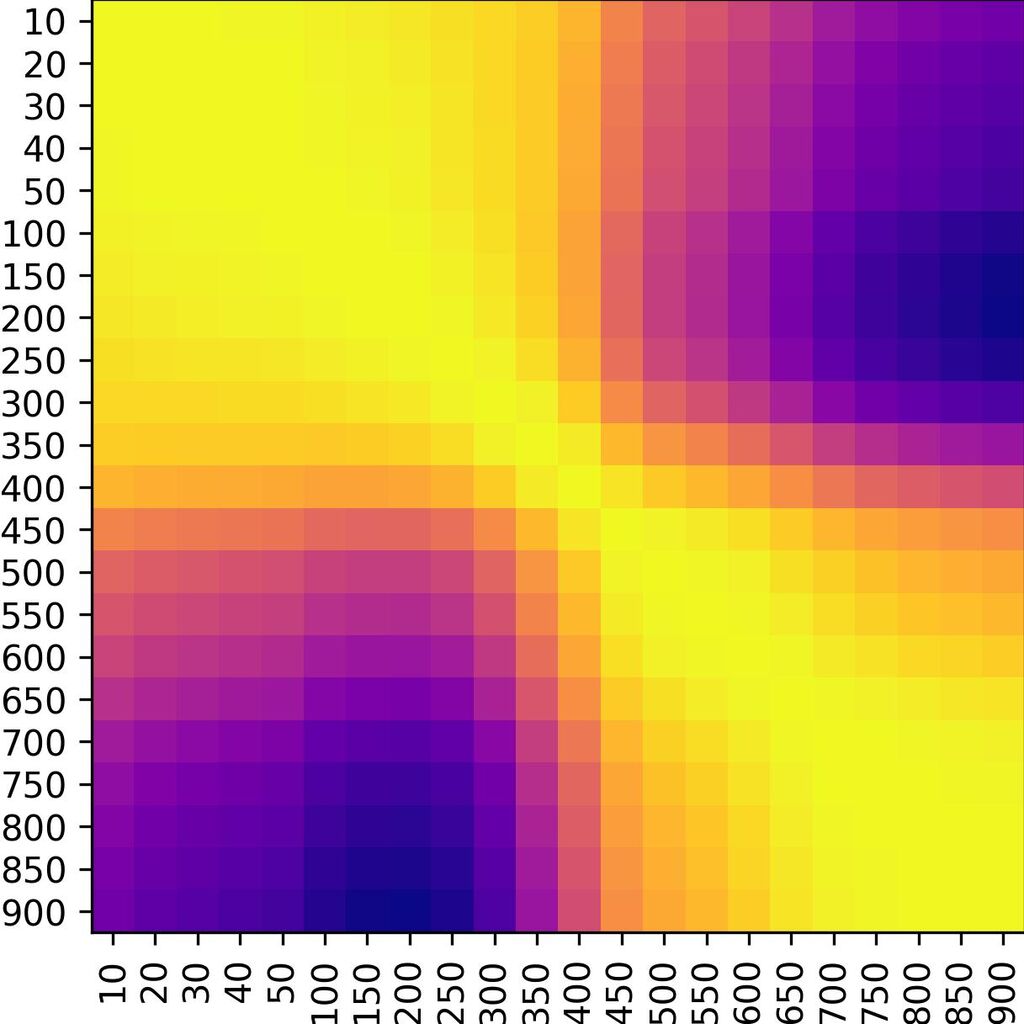} \\
        Input Image & TSM
    \end{tabular}
    
    \vspace{1em} 

    \setlength{\tabcolsep}{2pt} 
    
    \begin{tabularx}{\linewidth}{ *{11}{>{\centering\arraybackslash}p{0.0845\linewidth}} } 
    
    \multicolumn{11}{c}{\textbf{CSMs across Timesteps}} \\
    \noalign{\vspace{0.5em}}

    \fcolorbox{red}{white}{\includegraphics[width=0.95\linewidth]{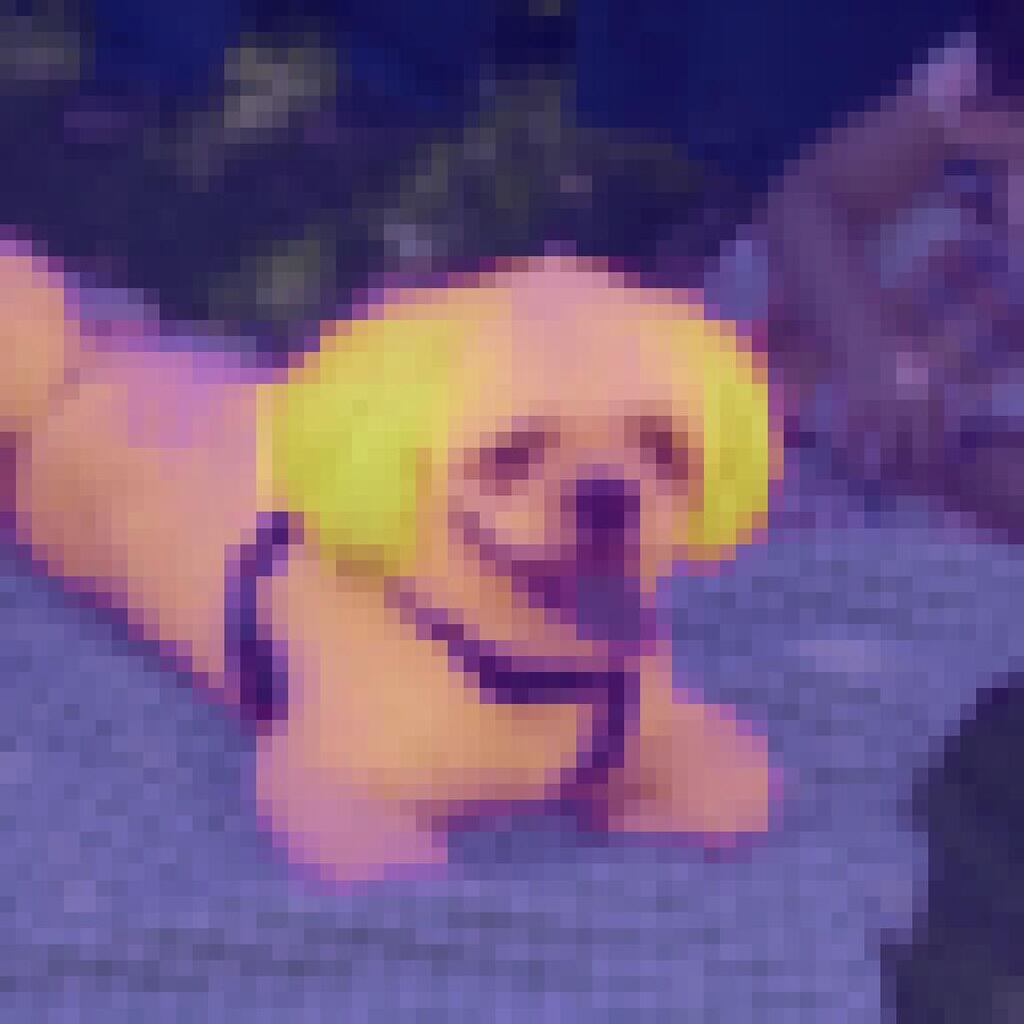}} &
    \fcolorbox{red}{white}{\includegraphics[width=0.95\linewidth]{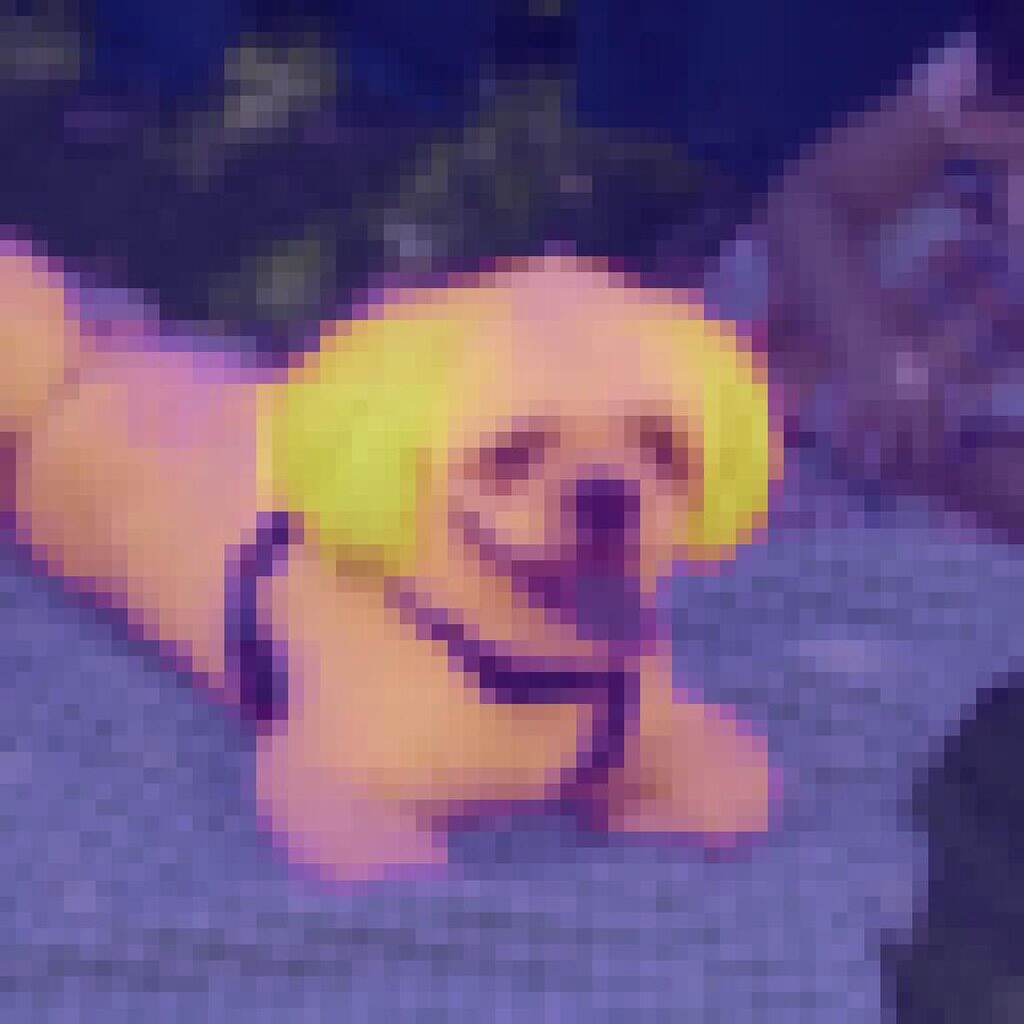}} &
    \fcolorbox{red}{white}{\includegraphics[width=0.95\linewidth]{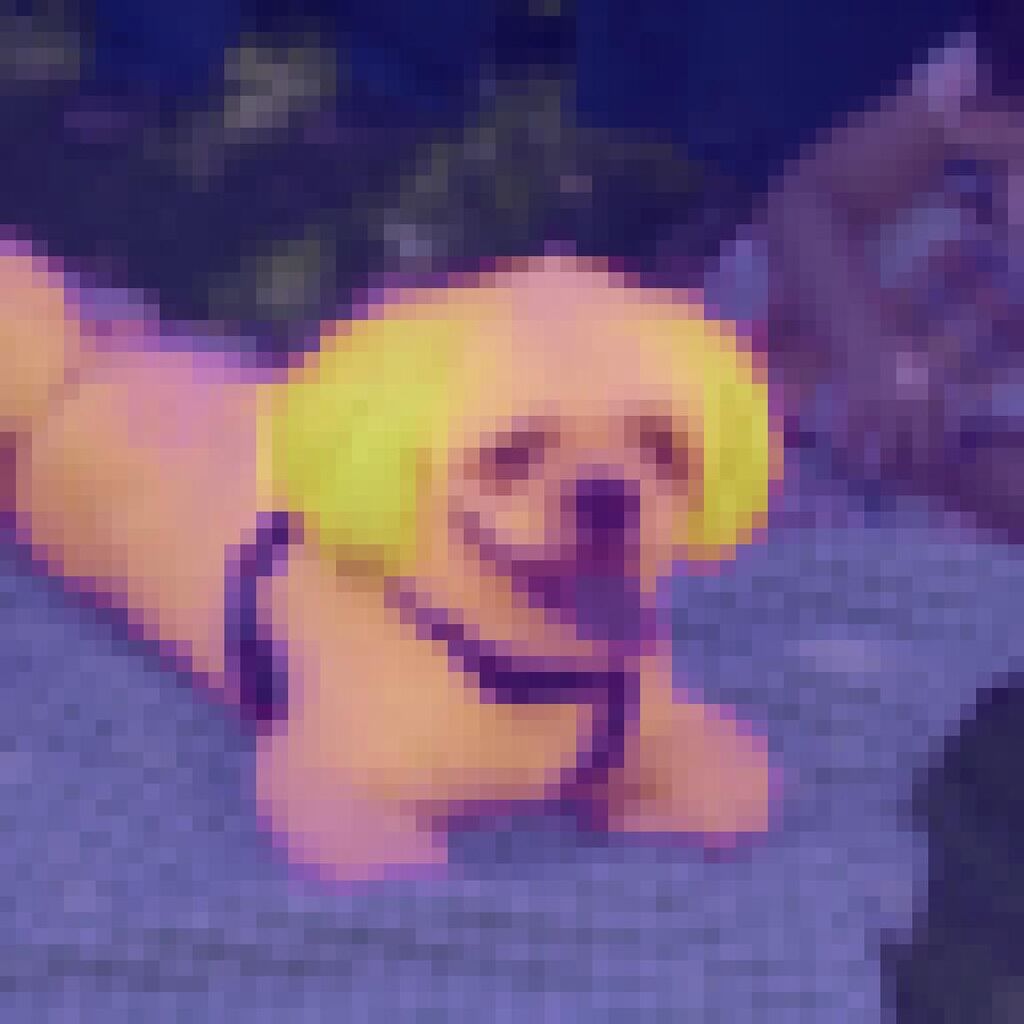}} &
    \fcolorbox{red}{white}{\includegraphics[width=0.95\linewidth]{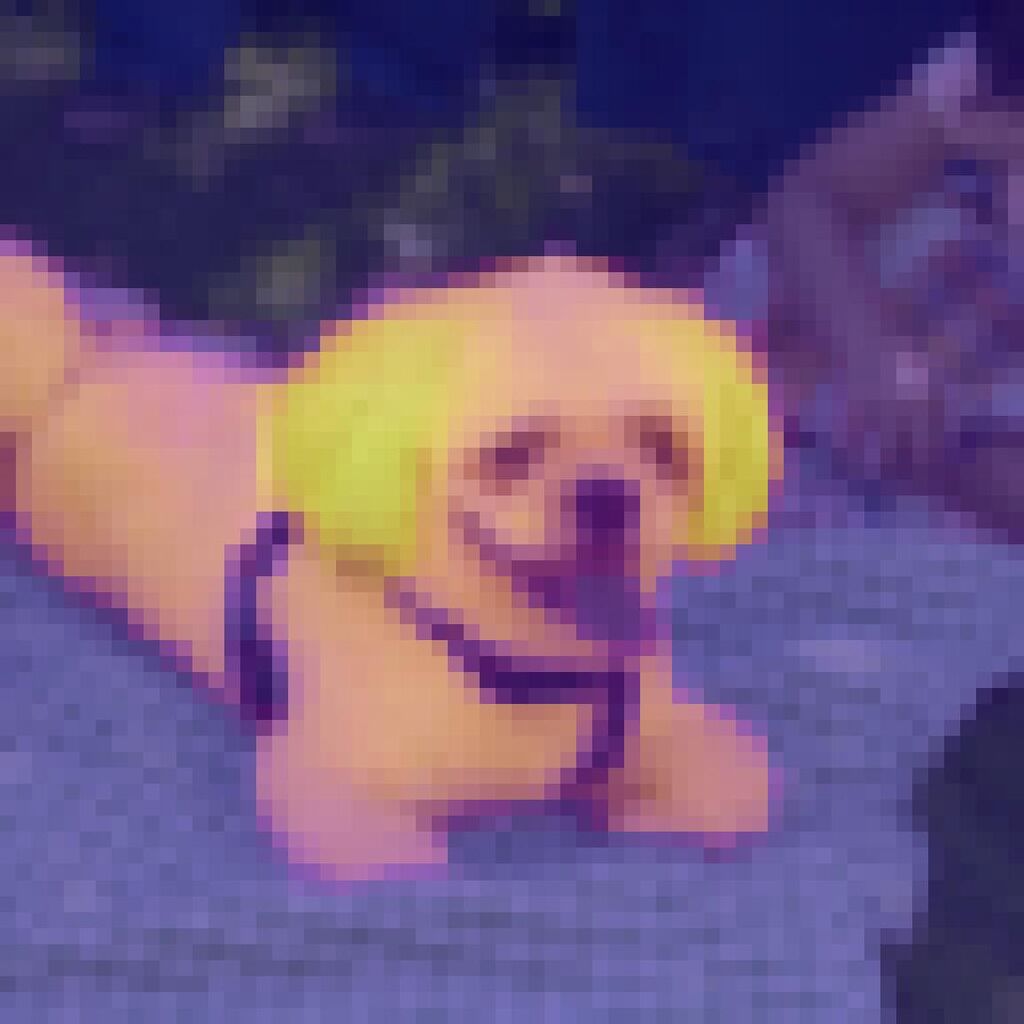}} &
    \fcolorbox{red}{white}{\includegraphics[width=0.95\linewidth]{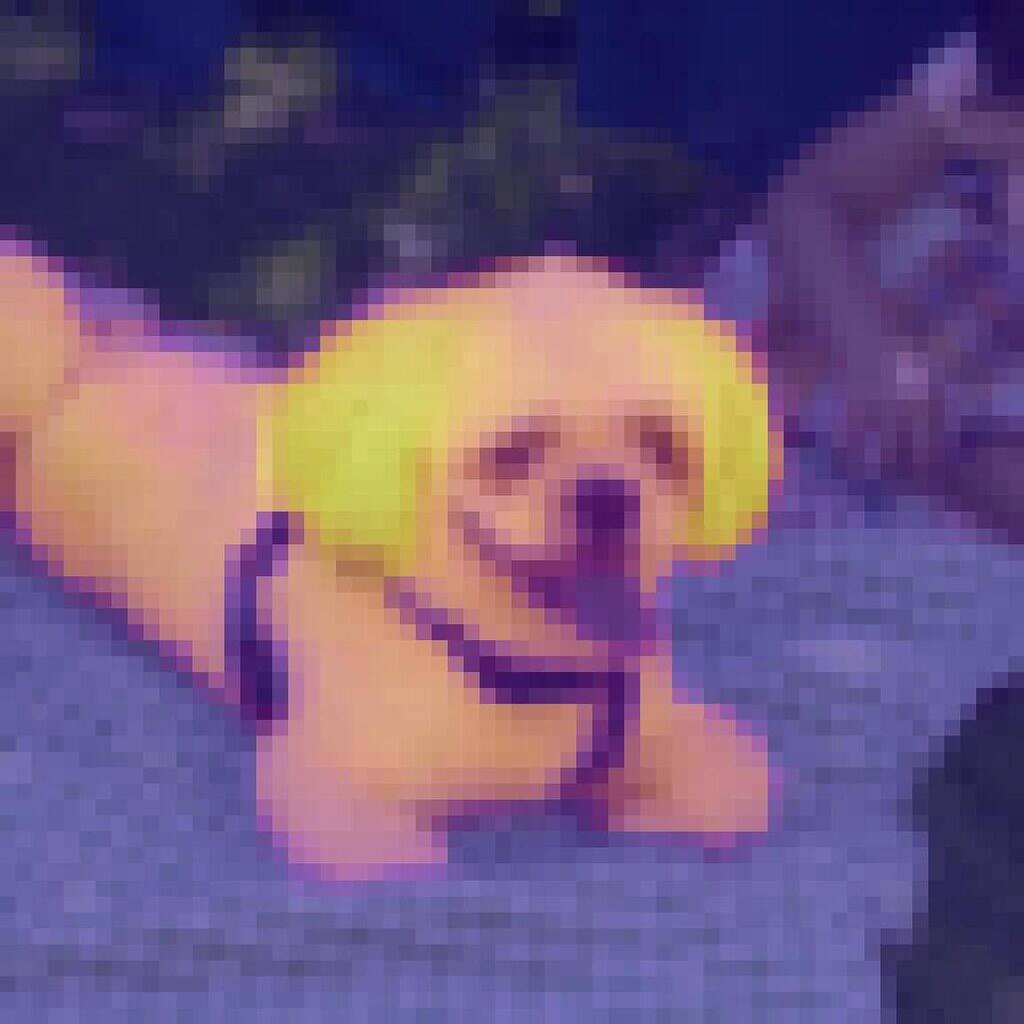}} &
    \fcolorbox{red}{white}{\includegraphics[width=0.95\linewidth]{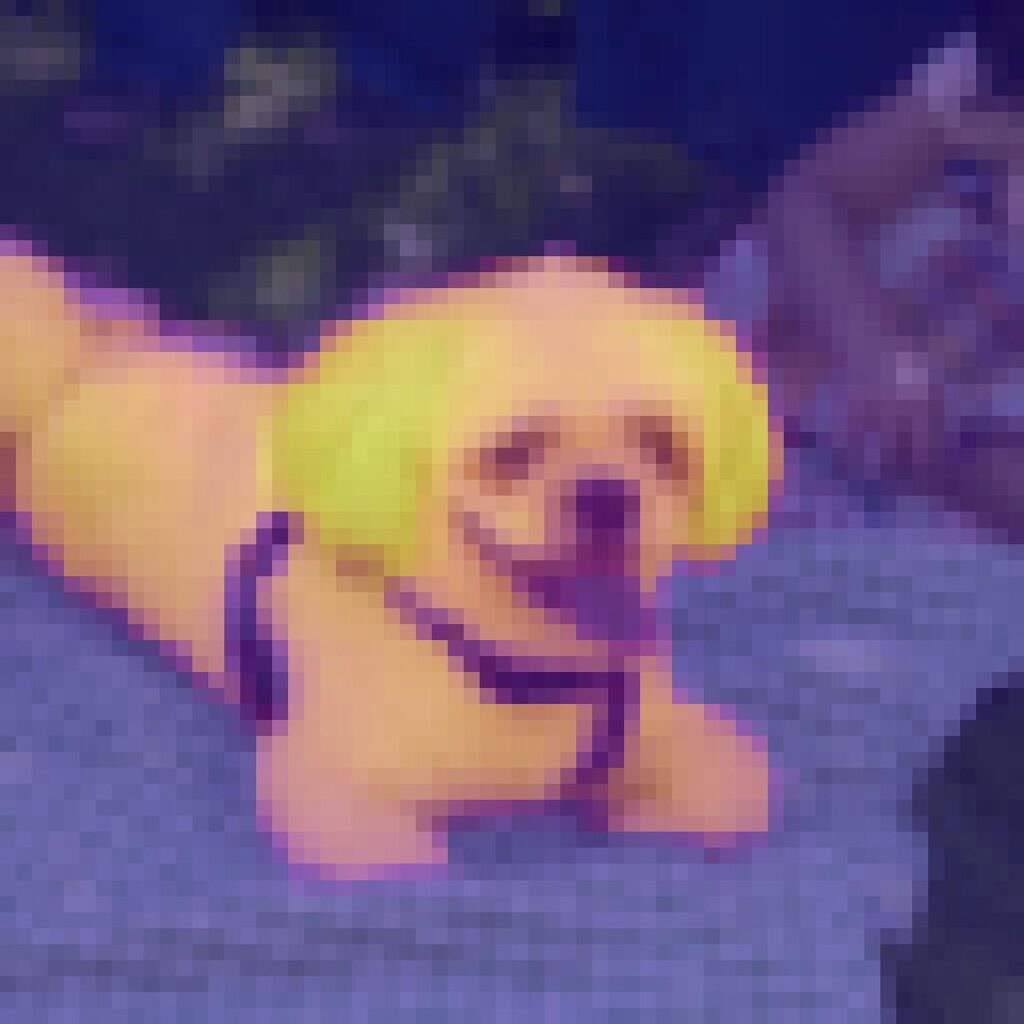}} &
    \fcolorbox{red}{white}{\includegraphics[width=0.95\linewidth]{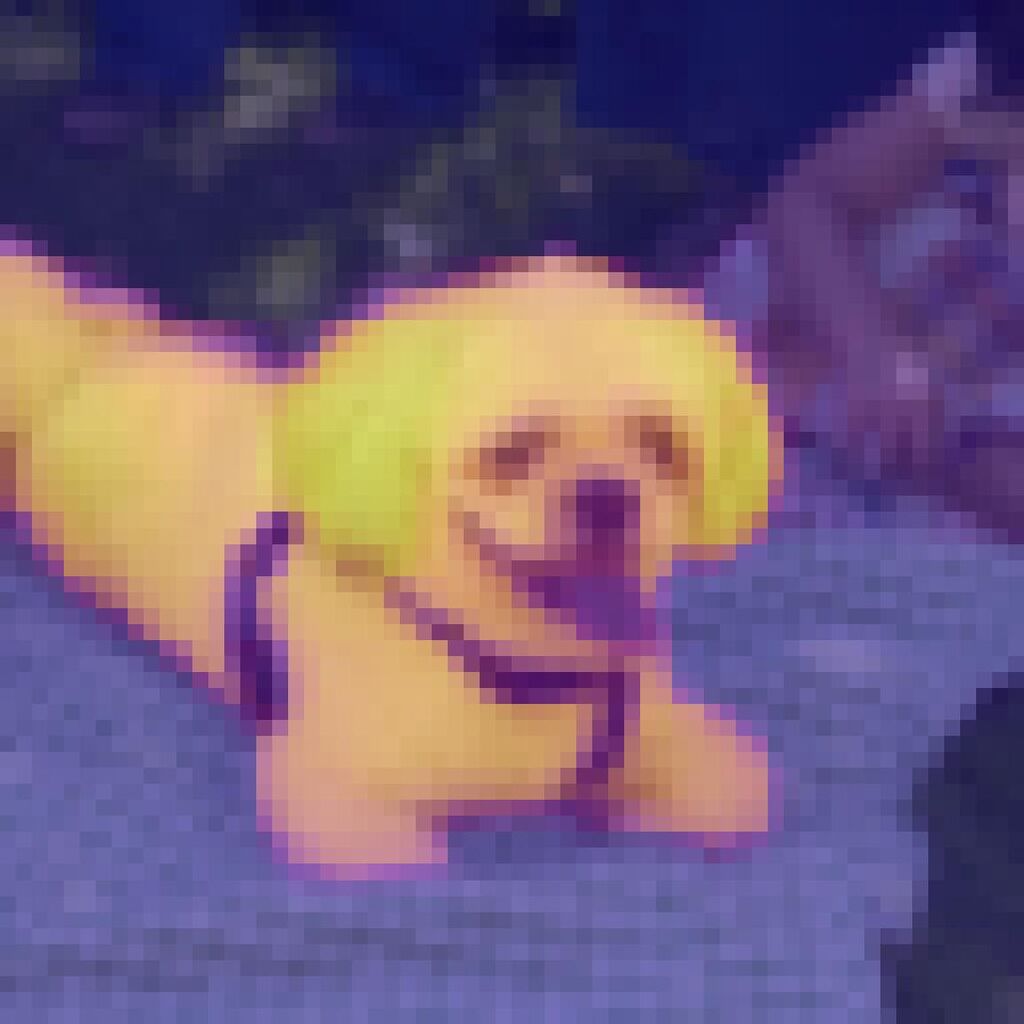}} &
    \fcolorbox{red}{white}{\includegraphics[width=0.95\linewidth]{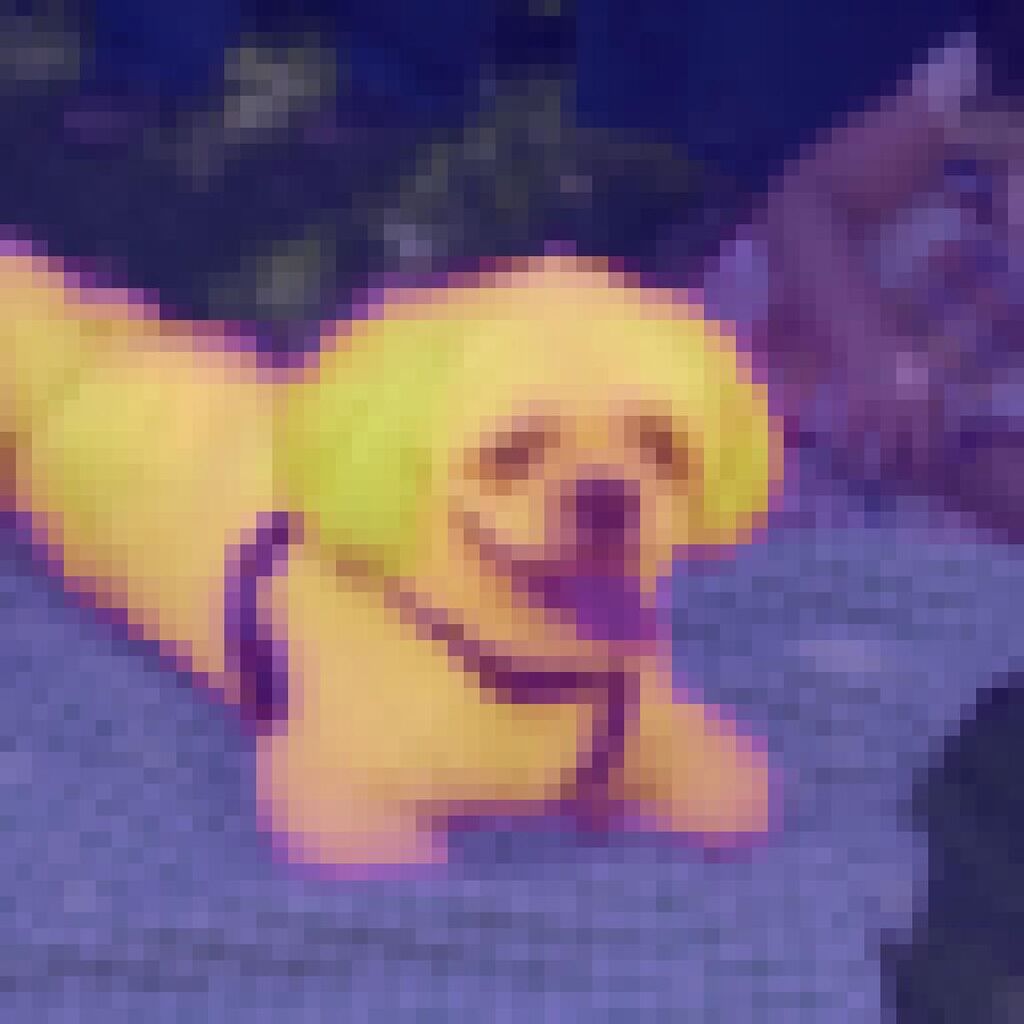}} &
    \fcolorbox{red}{white}{\includegraphics[width=0.95\linewidth]{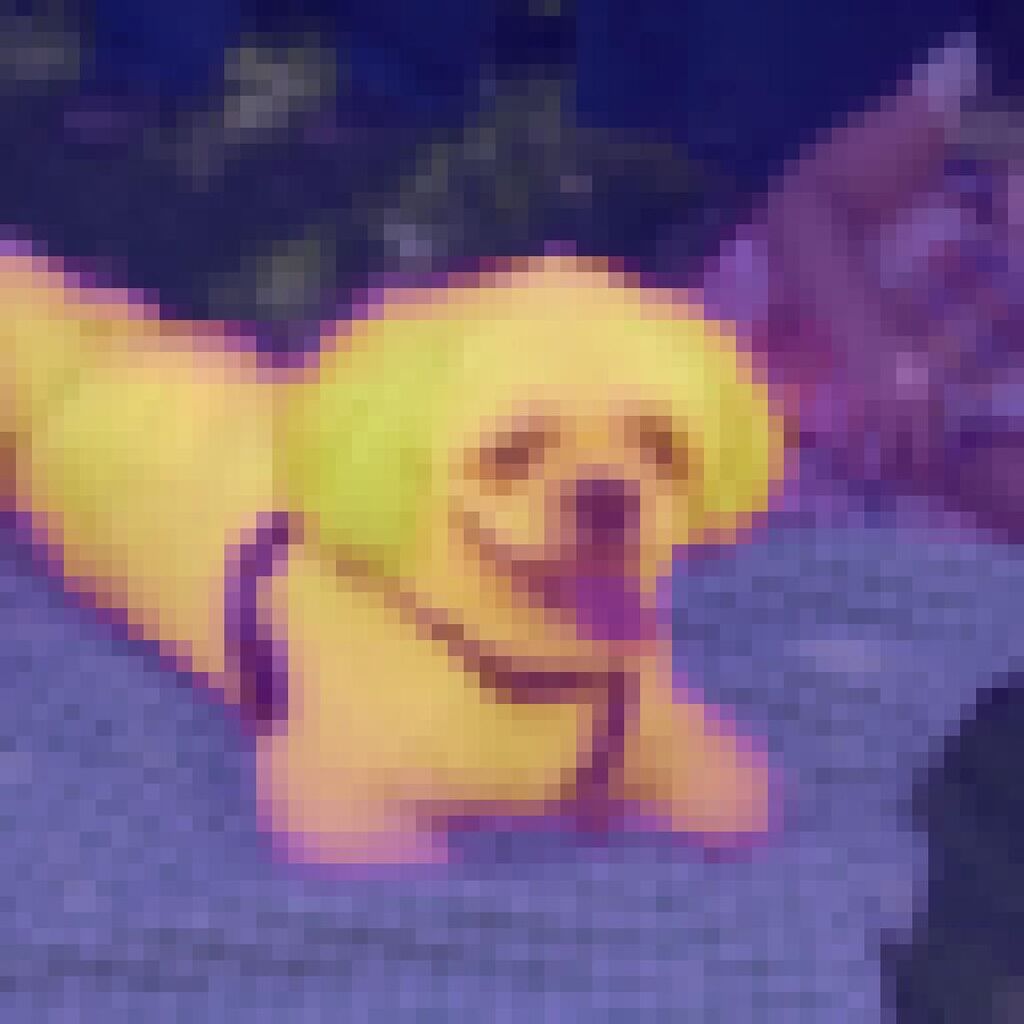}} &
    \fcolorbox{red}{white}{\includegraphics[width=0.95\linewidth]{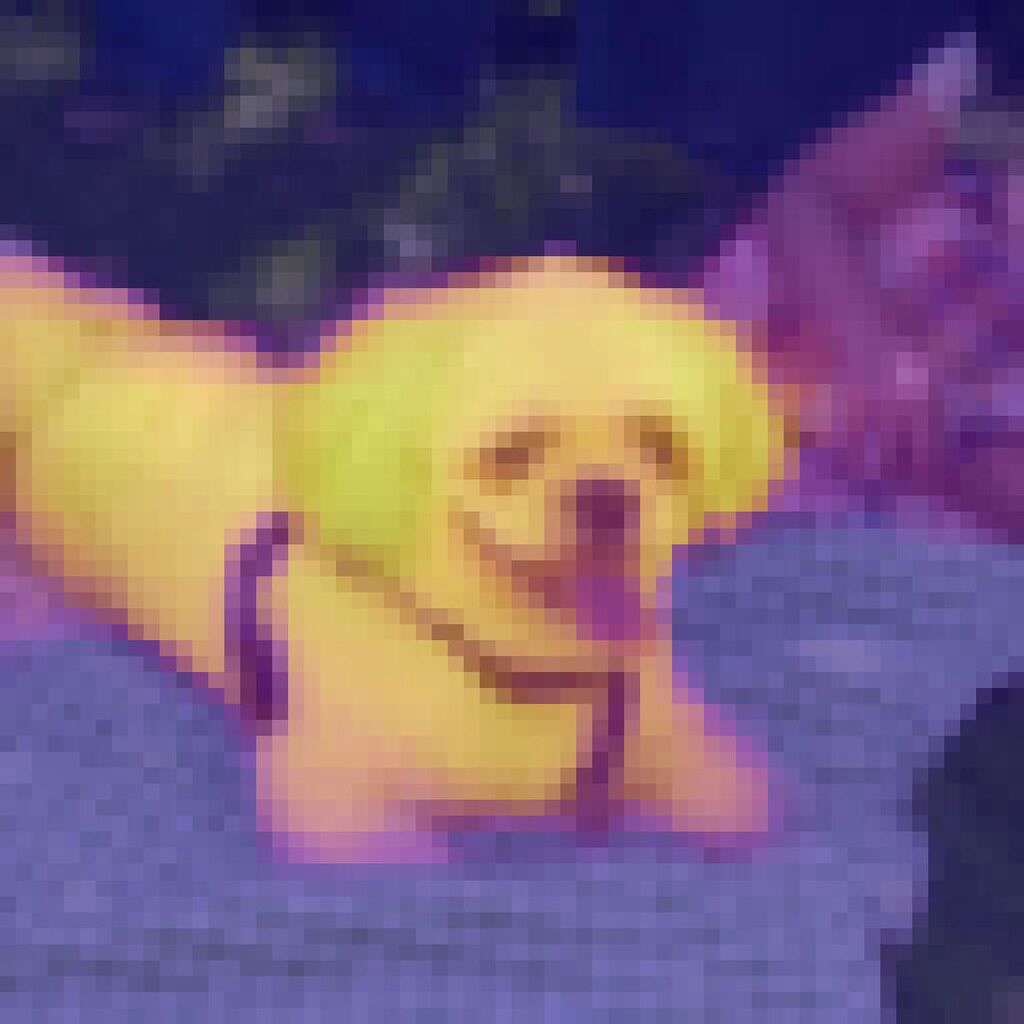}} &
    \fcolorbox{red}{white}{\includegraphics[width=0.95\linewidth]{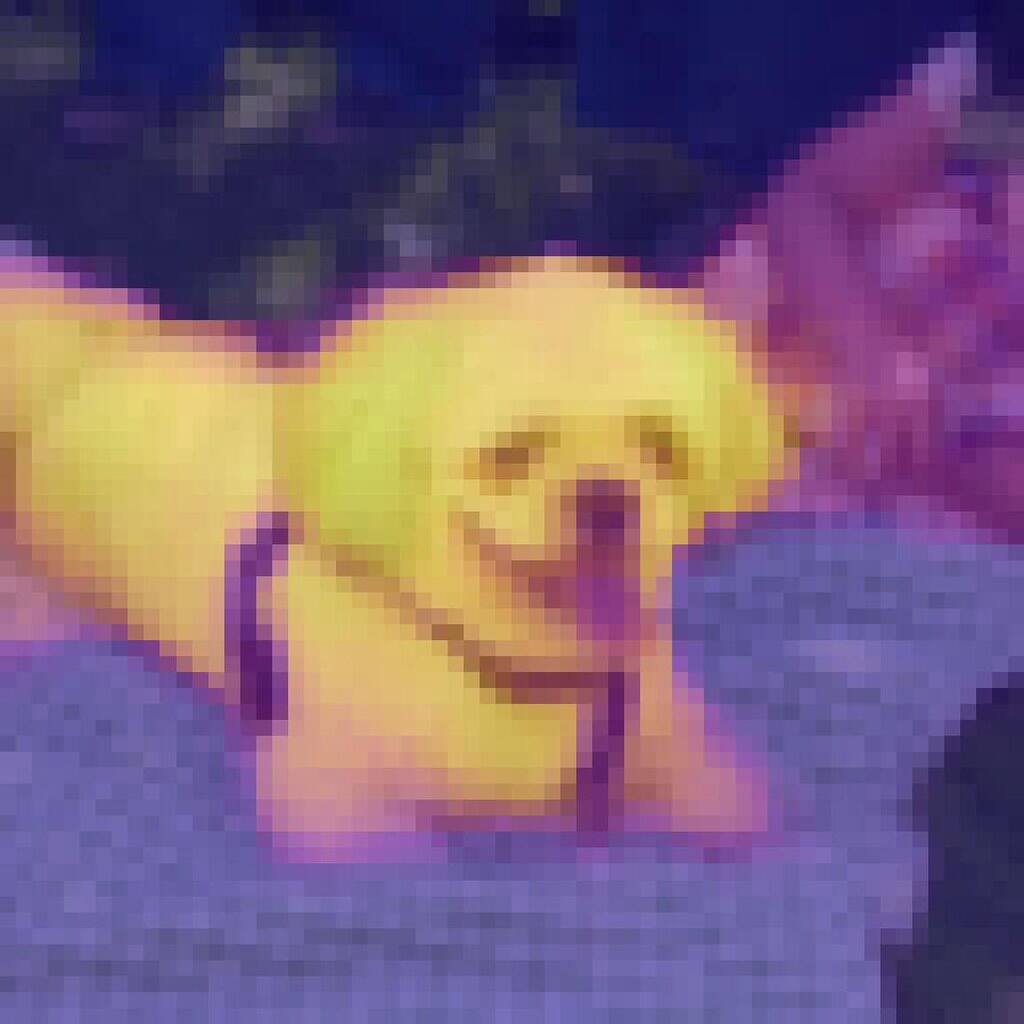}} \\
    
    t=10 & t=20 & t=30 & t=40 & t=50 & t=100 & t=150 & t=200 & t=250 & t=300 & t=350 \\[1em]
    
    \fcolorbox{red}{white}{\includegraphics[width=0.95\linewidth]{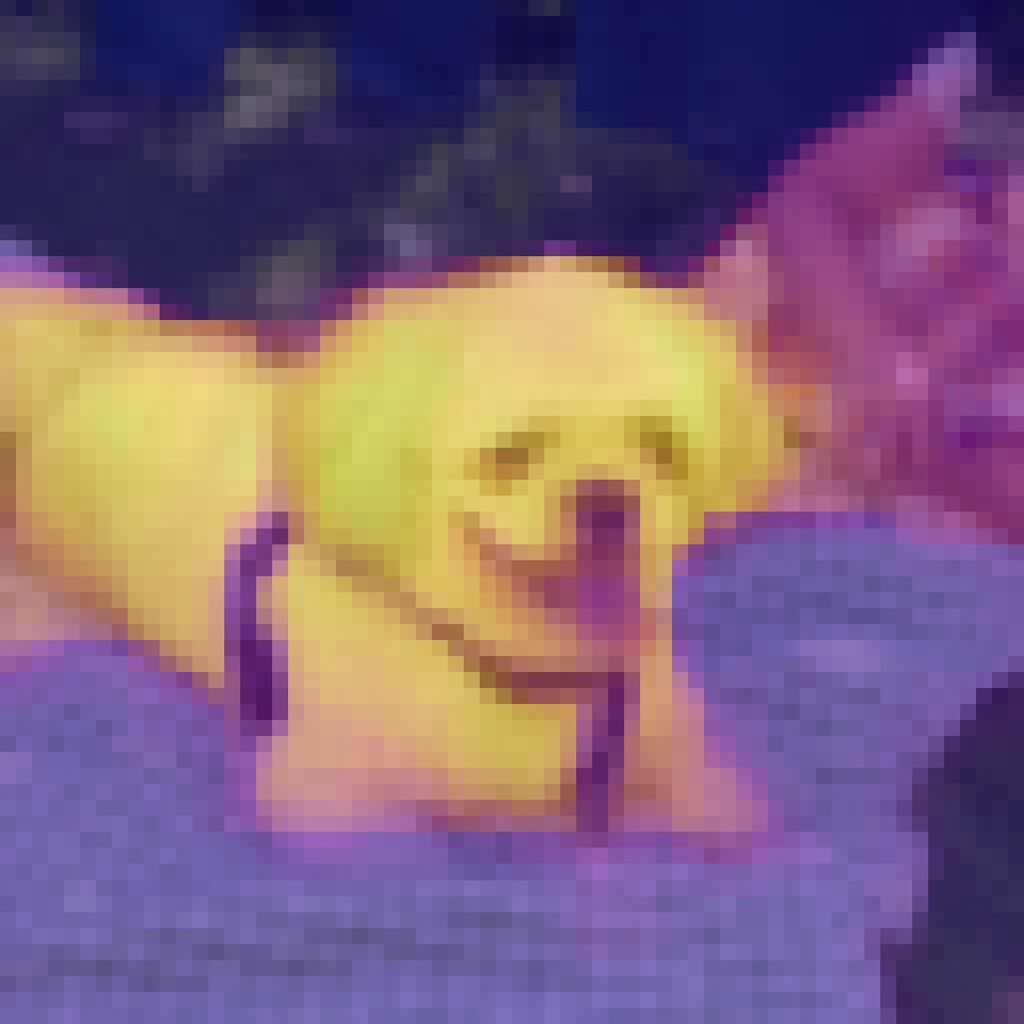}} &
    \fcolorbox{red}{white}{\includegraphics[width=0.95\linewidth]{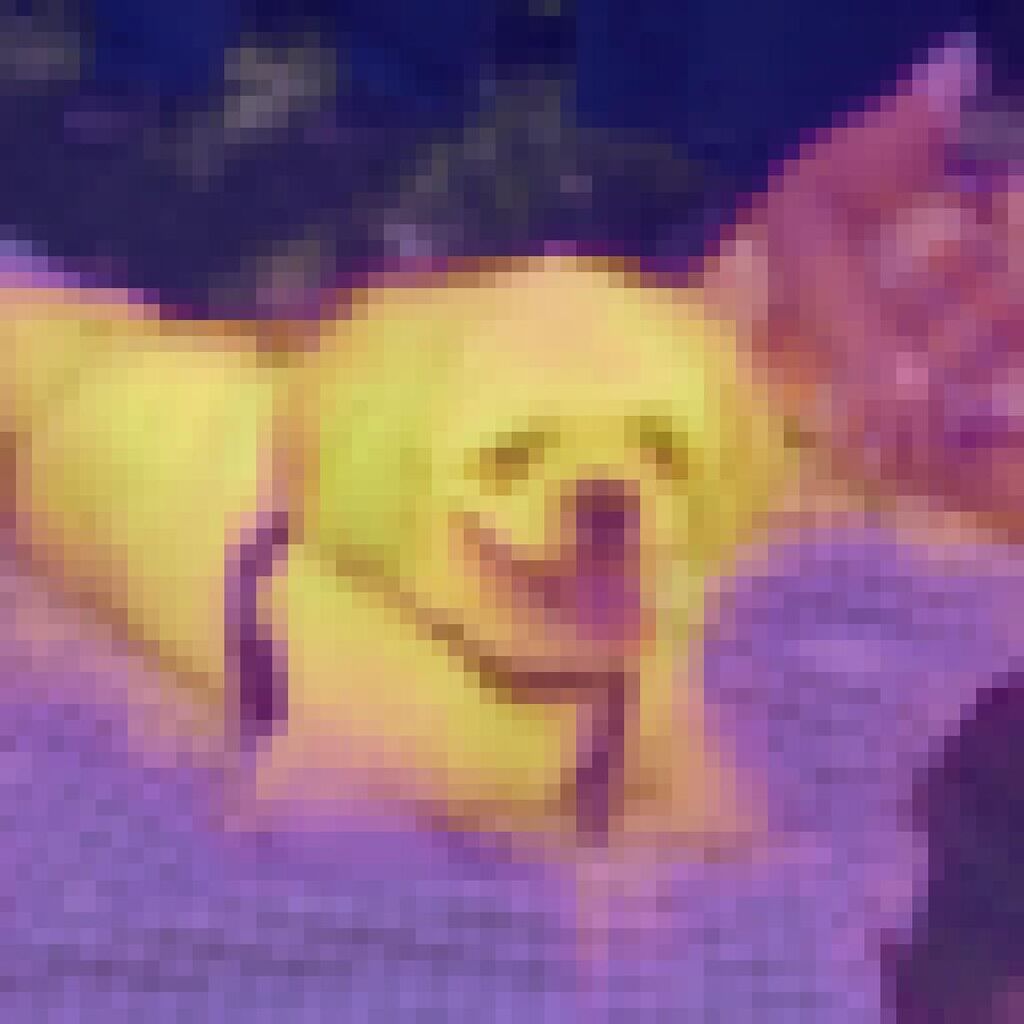}} &
    \fcolorbox{red}{white}{\includegraphics[width=0.95\linewidth]{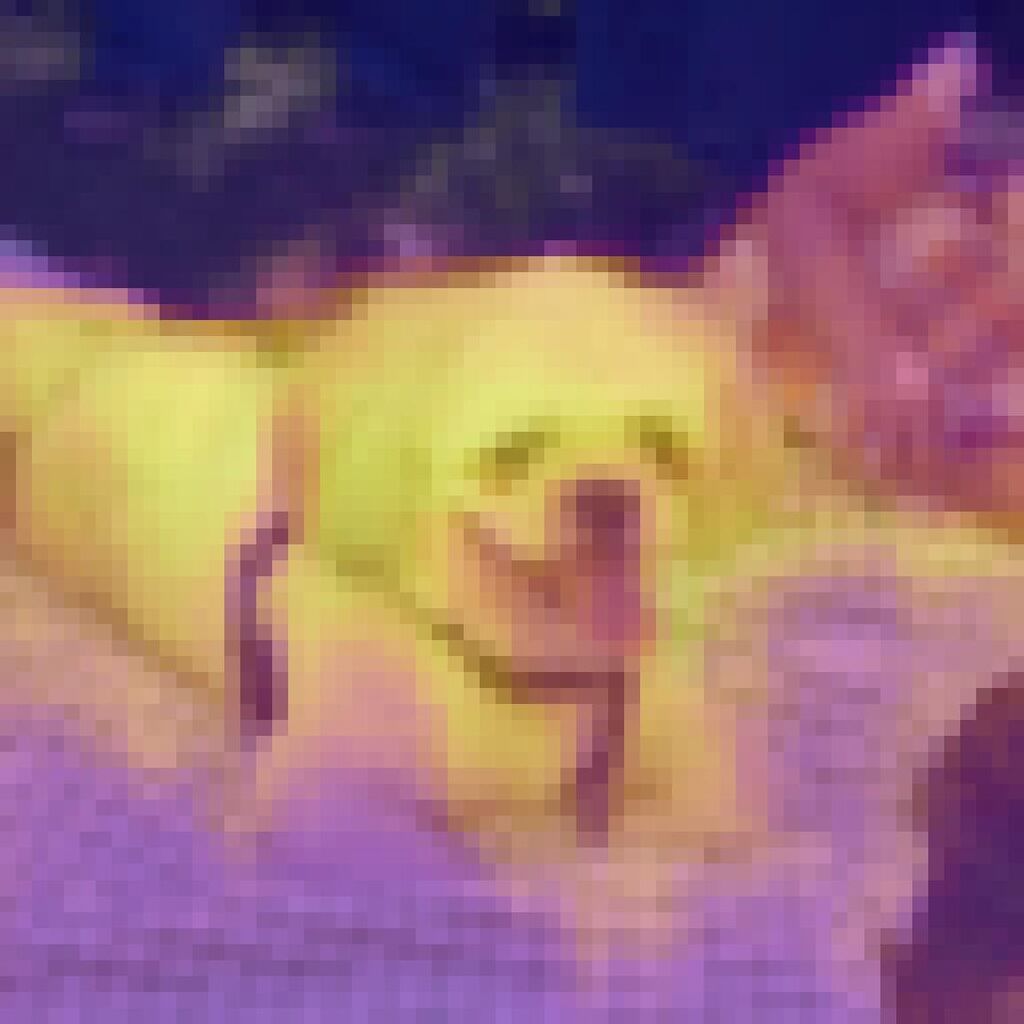}} &
    \fcolorbox{red}{white}{\includegraphics[width=0.95\linewidth]{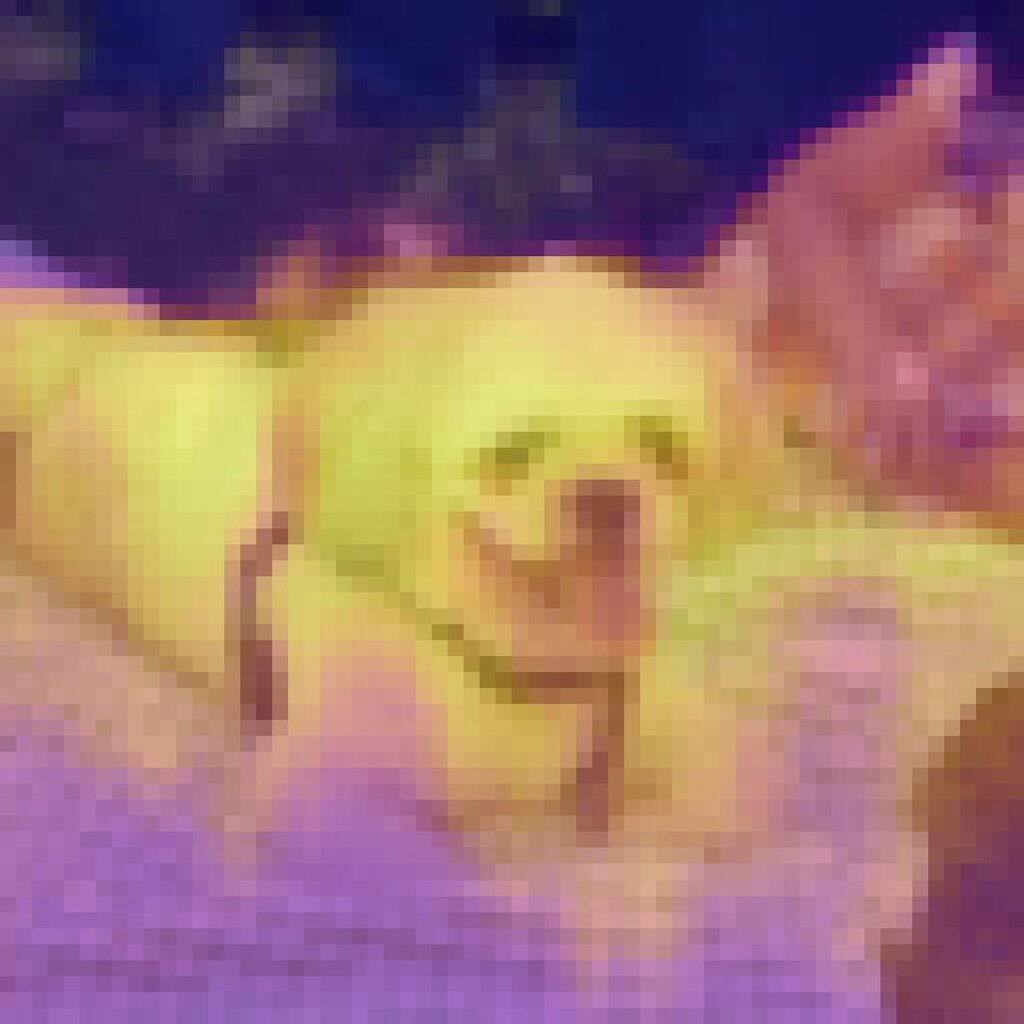}} &
    \fcolorbox{red}{white}{\includegraphics[width=0.95\linewidth]{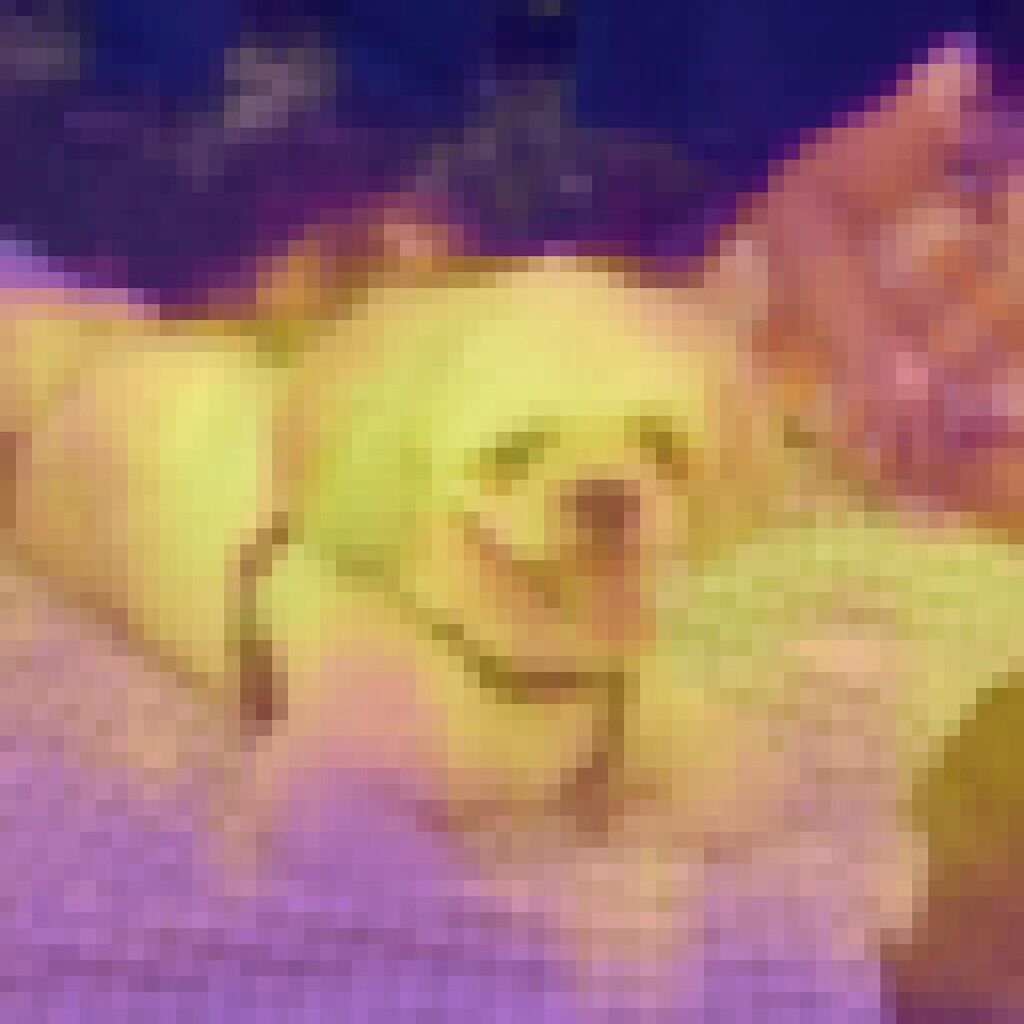}} &
    \fcolorbox{red}{white}{\includegraphics[width=0.95\linewidth]{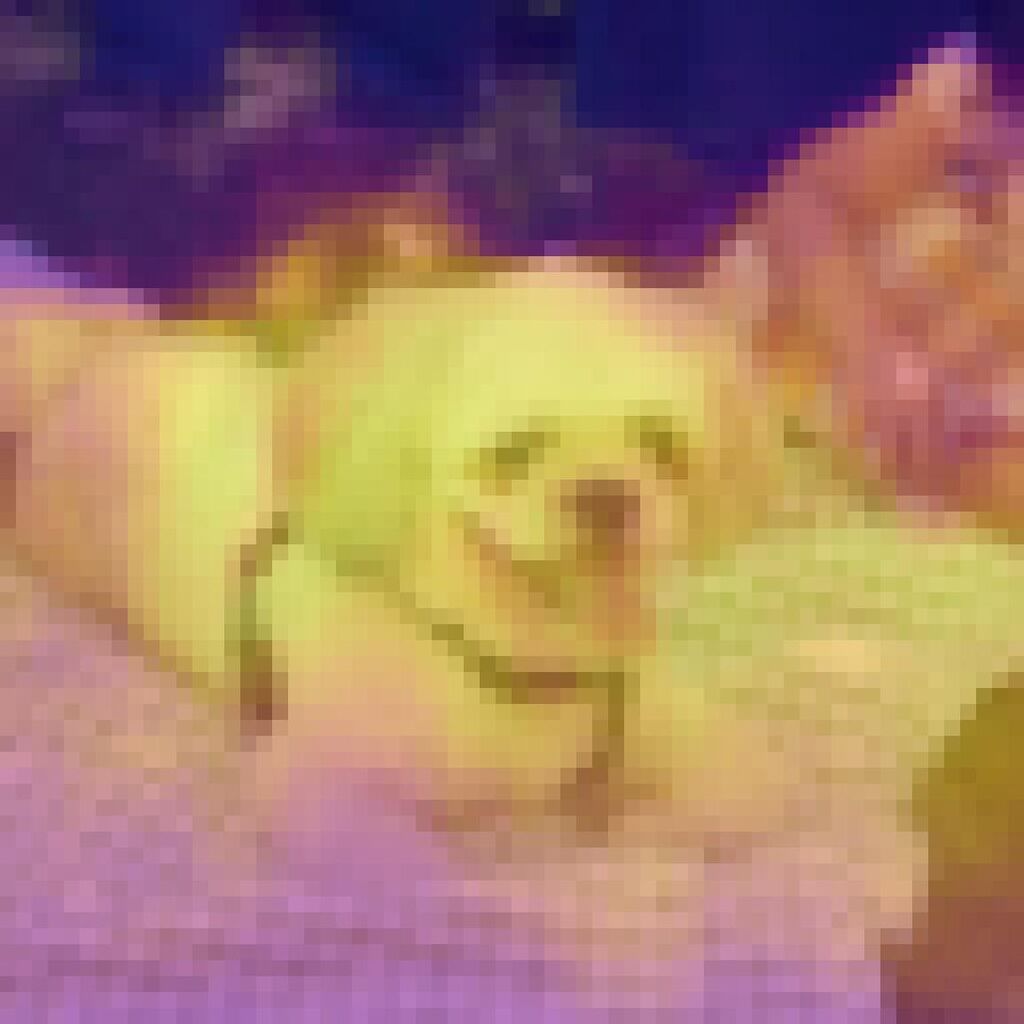}} &
    \fcolorbox{red}{white}{\includegraphics[width=0.95\linewidth]{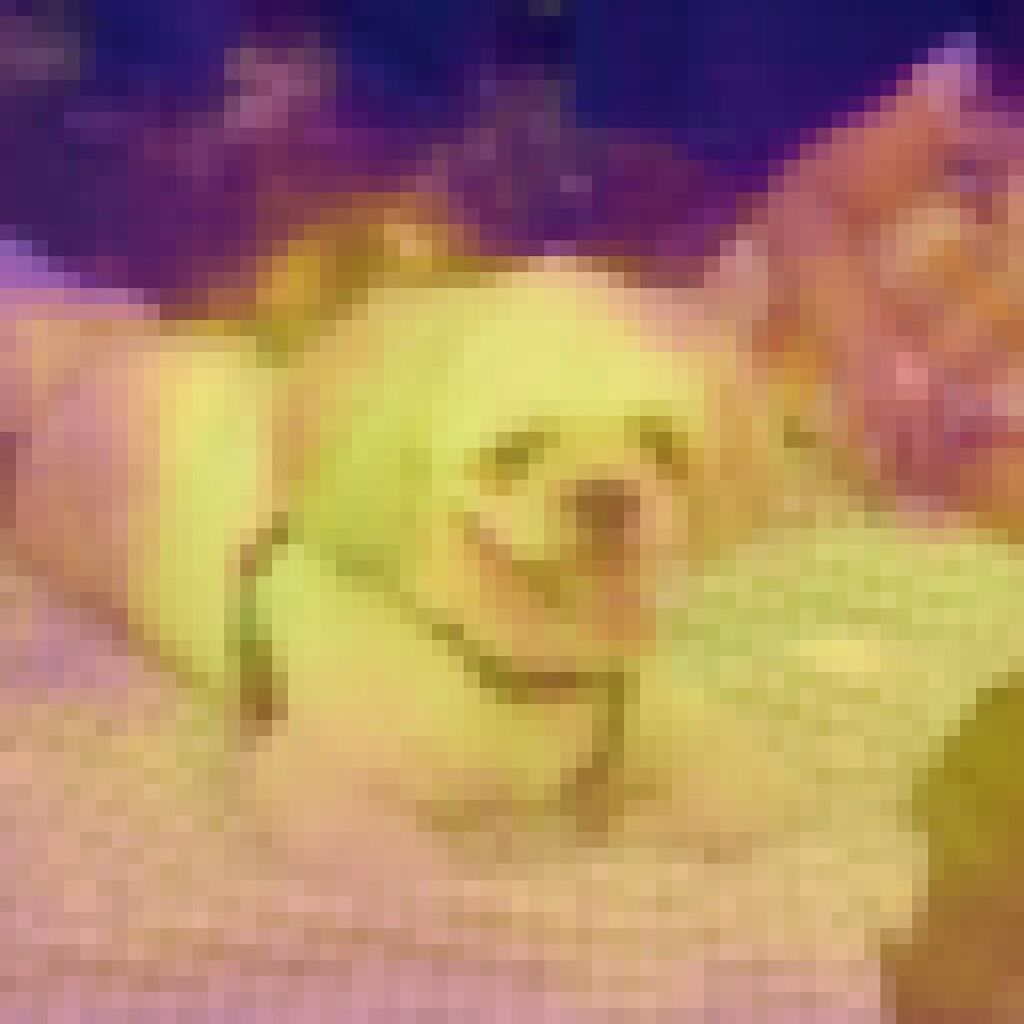}} &
    \fcolorbox{red}{white}{\includegraphics[width=0.95\linewidth]{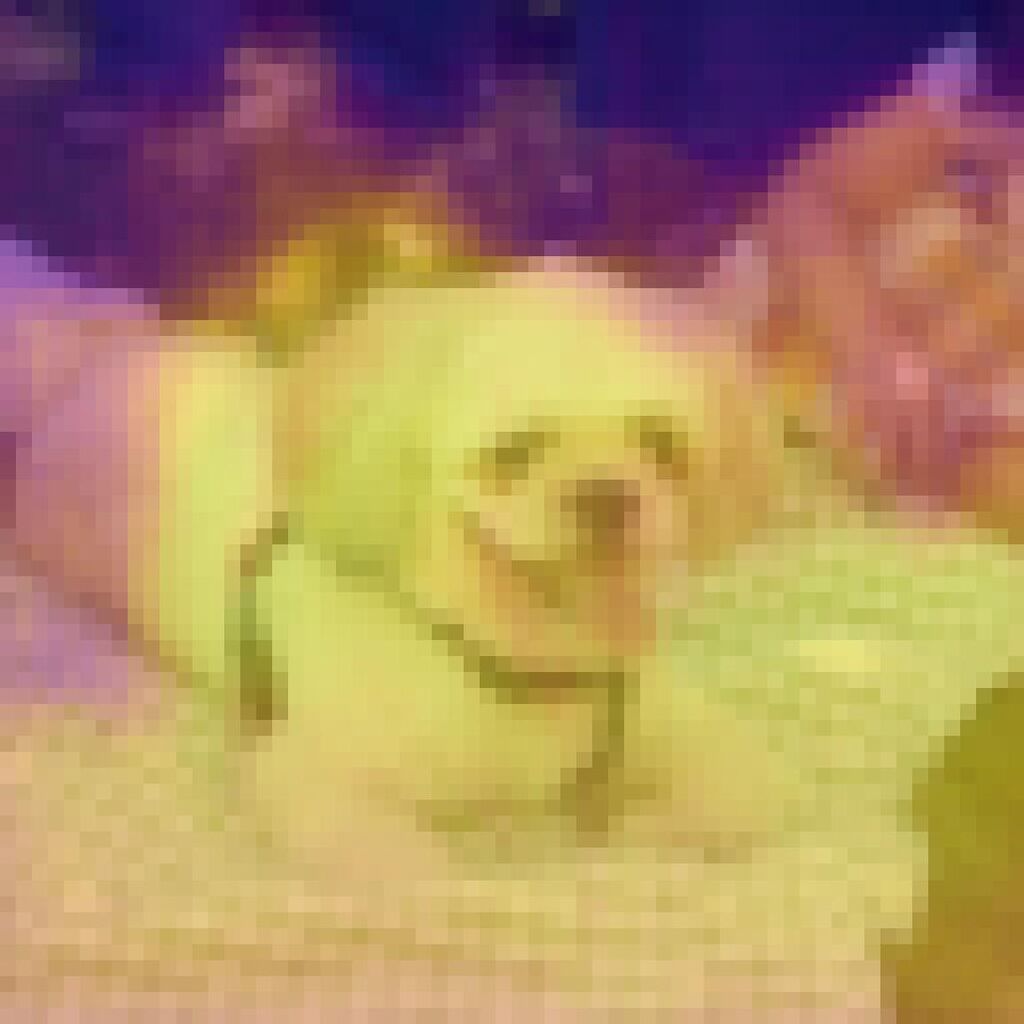}} &
    \fcolorbox{red}{white}{\includegraphics[width=0.95\linewidth]{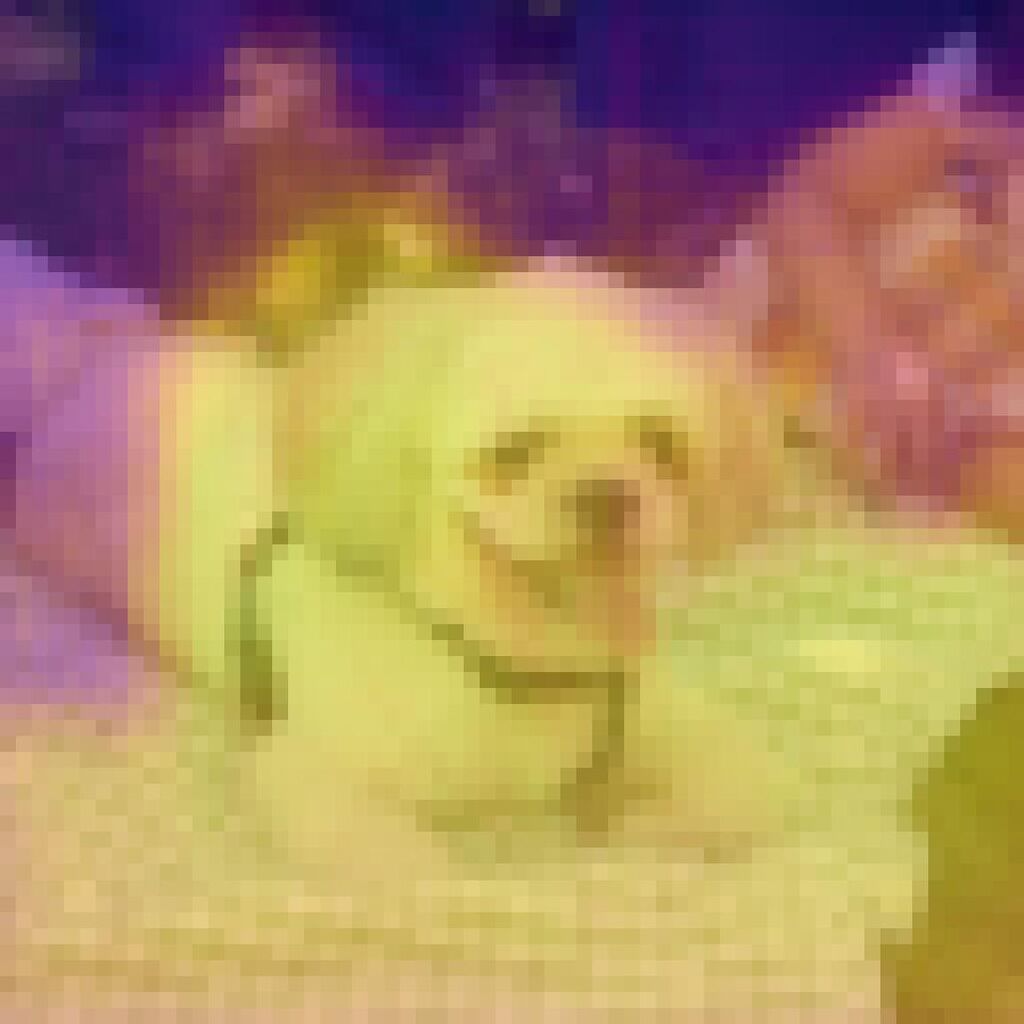}} &
    \fcolorbox{red}{white}{\includegraphics[width=0.95\linewidth]{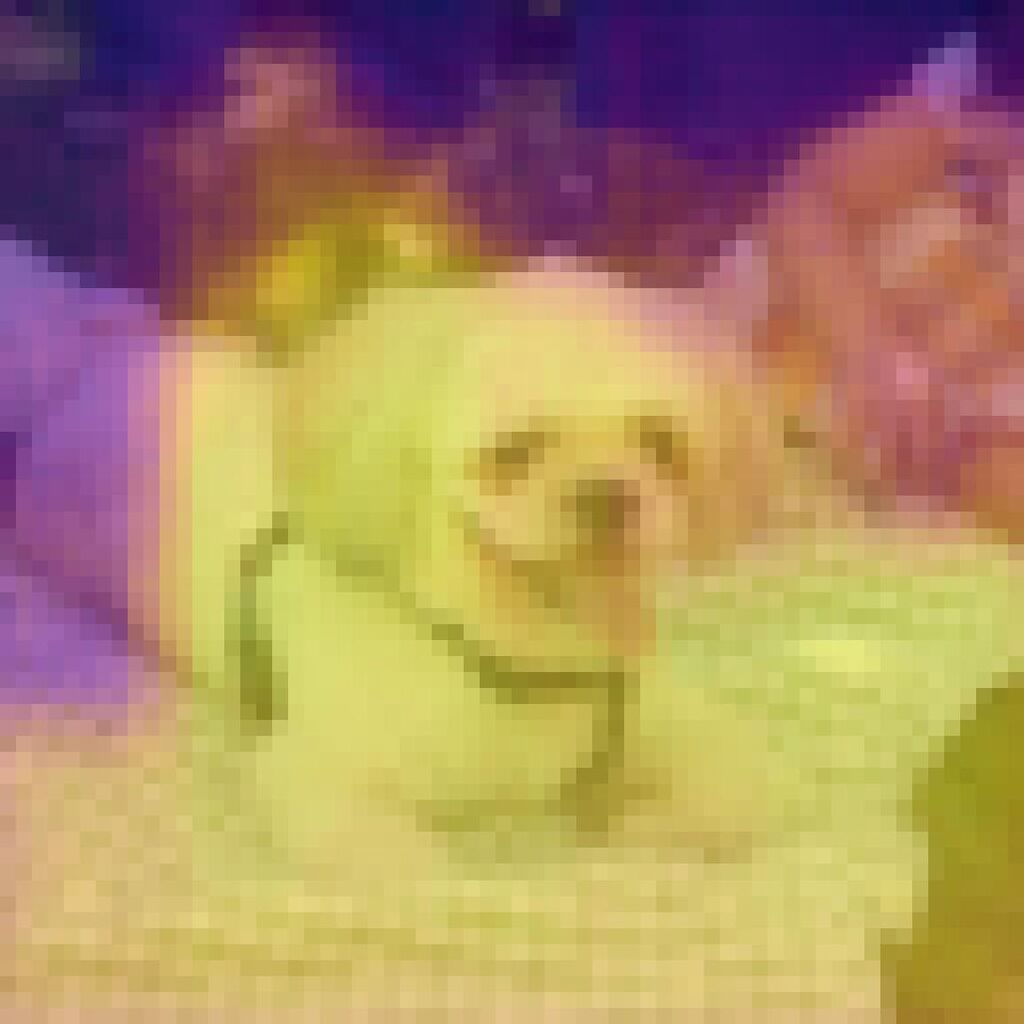}} &
    \fcolorbox{red}{white}{\includegraphics[width=0.95\linewidth]{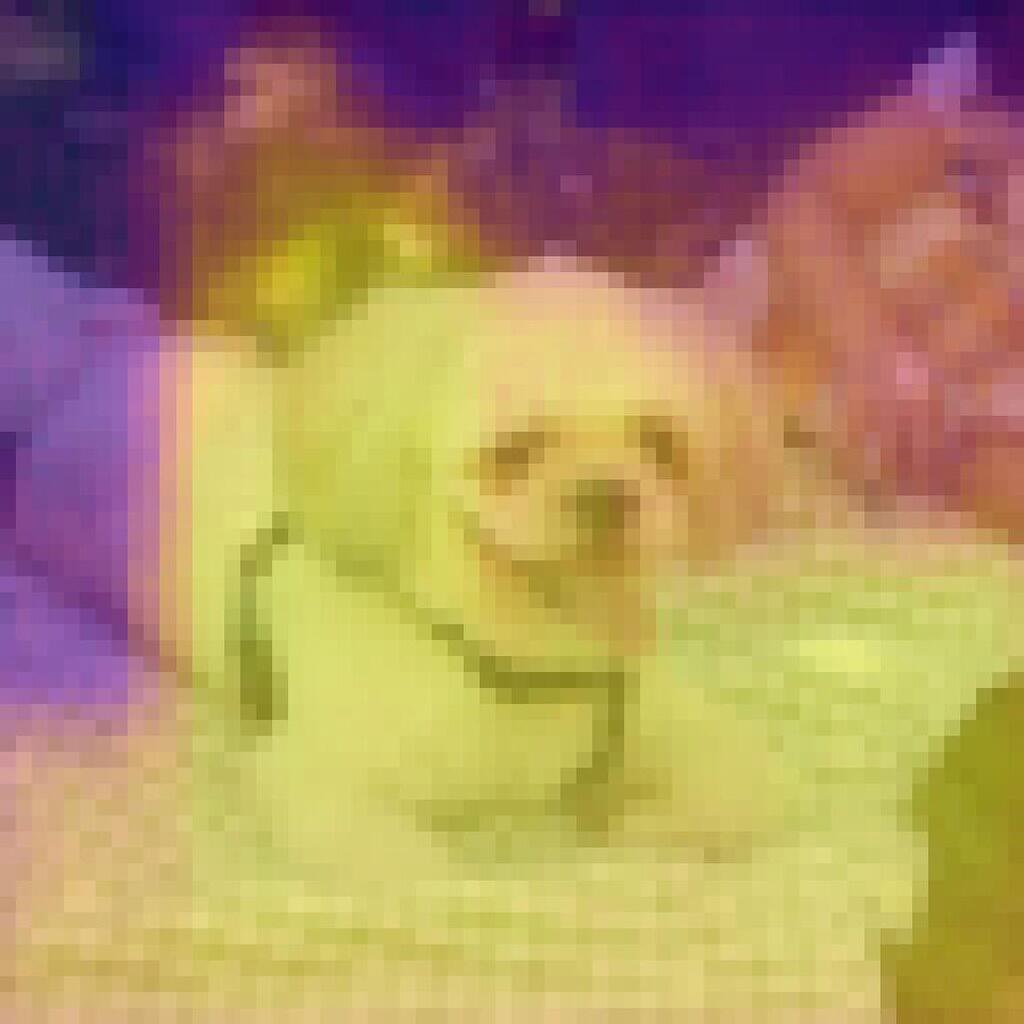}} \\
    
    t=400 & t=450 & t=500 & t=550 & t=600 & t=650 & t=700 & t=750 & t=800 & t=850 & t=900 \\
    
    \end{tabularx}

    \caption{The top row displays the original input image and the corresponding Temporal Stability Matrix (TSM) for the query point (in red). The panels below show the evolution of the Contextual Similarity Maps (CSMs) across the denoising process.}
    
    \label{fig:supp_hierarchical_progress_sample14}
\end{sidewaysfigure*}
\begin{sidewaysfigure*}[htb!] 
    \centering
    \begin{tabular}{ccc}
        \includegraphics[width=0.2\linewidth]{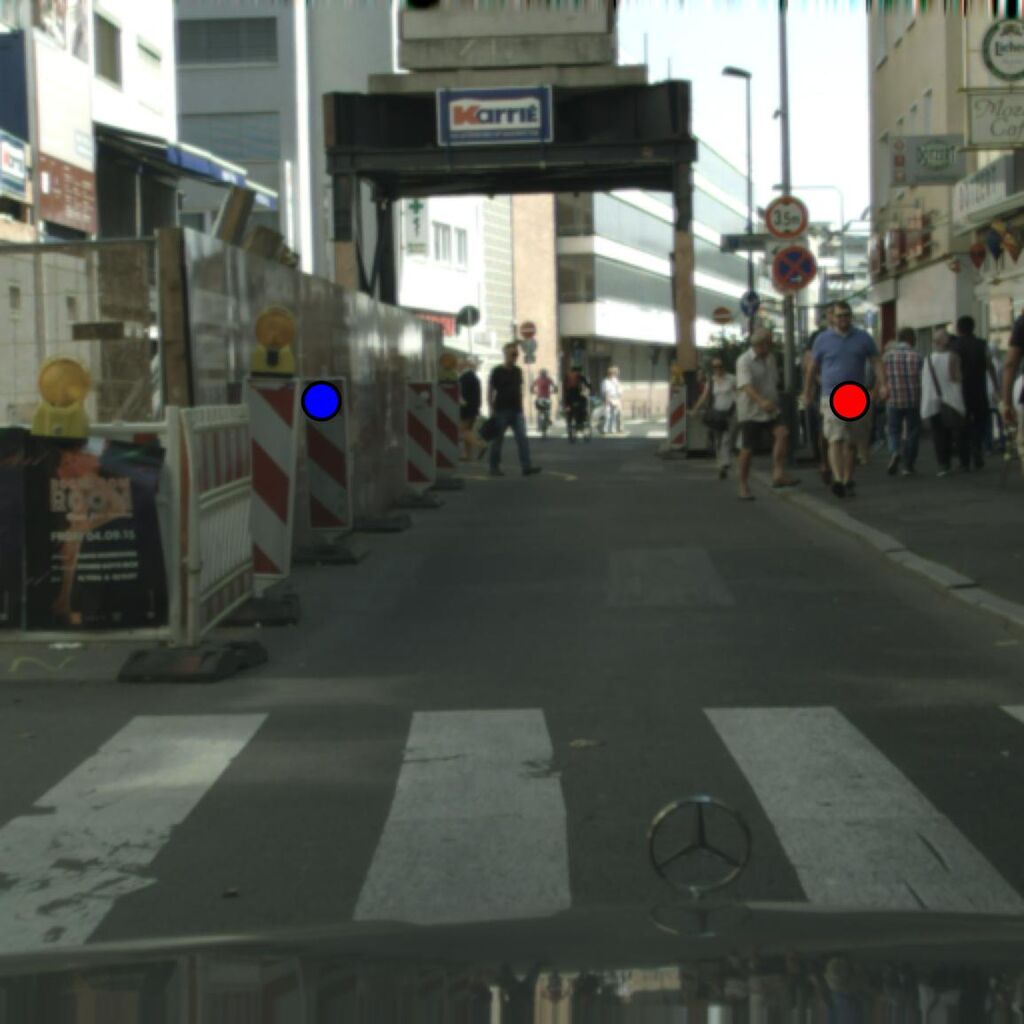}  & \includegraphics[width=0.2\linewidth]{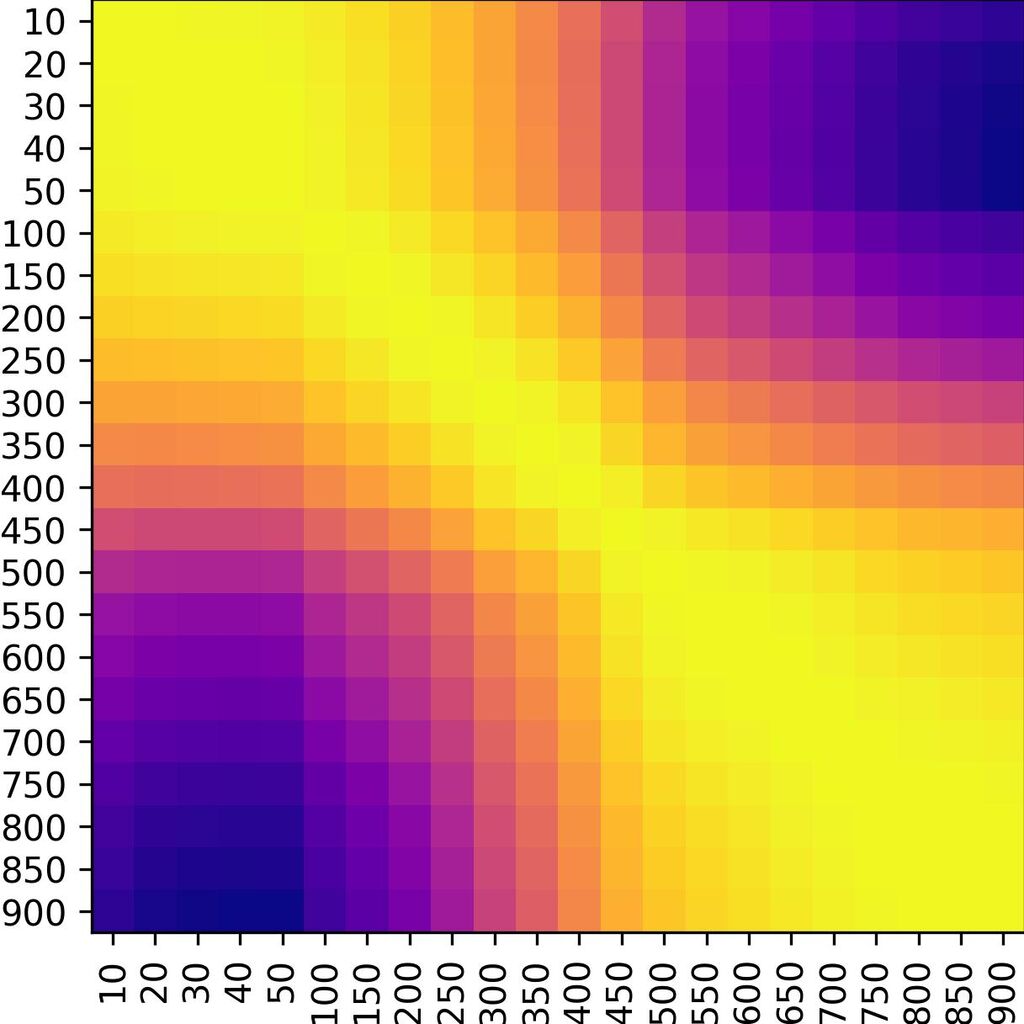} &
        \includegraphics[width=0.2\linewidth]{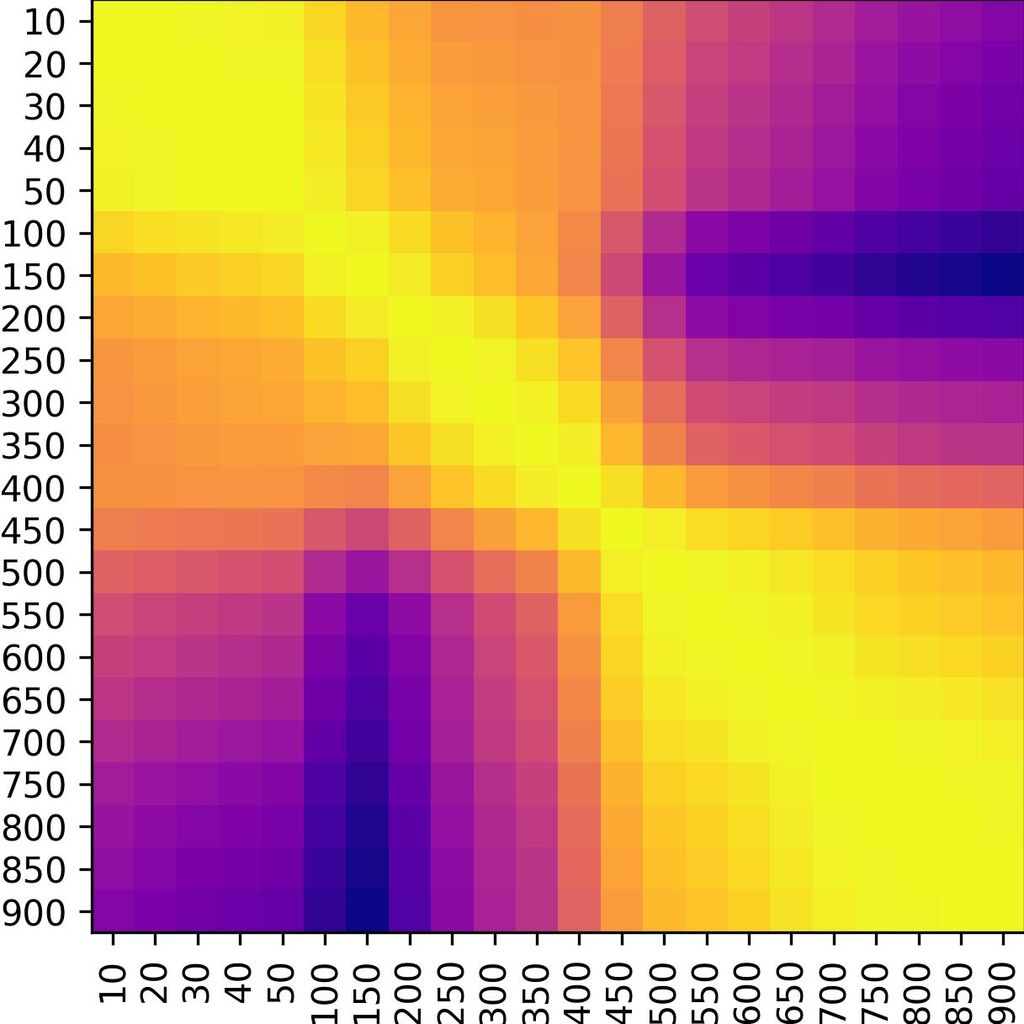}\\
        Input Image & Red Point TSM & Blue Point TSM
    \end{tabular}
    \vspace{1em} 

    \setlength{\tabcolsep}{2pt} 
    \begin{tabularx}{\linewidth}{ *{11}{>{\centering\arraybackslash}p{0.0845\linewidth}} } 
    
    \multicolumn{11}{c}{\textbf{Red Point CSM}} \\
    
    \fcolorbox{red}{white}{\includegraphics[width=0.95\linewidth]{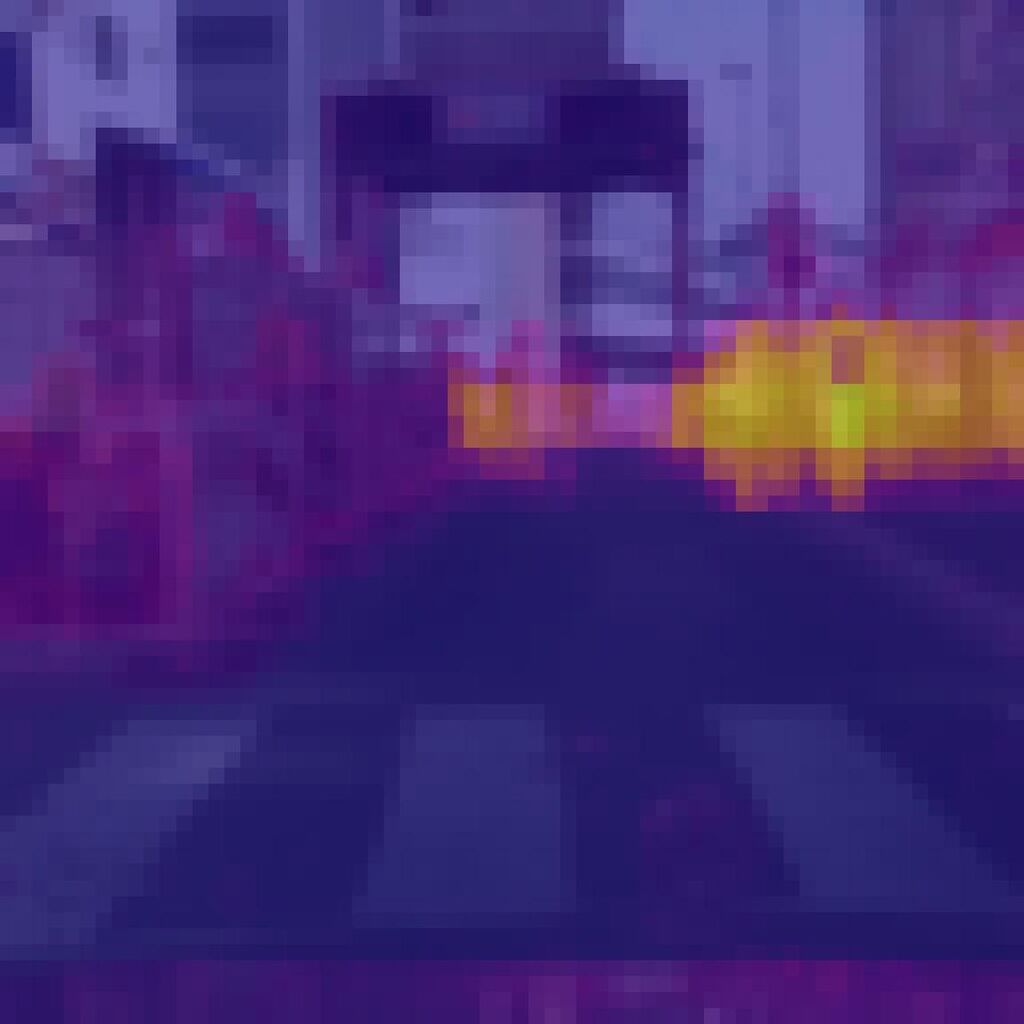}} &
    \fcolorbox{red}{white}{\includegraphics[width=0.95\linewidth]{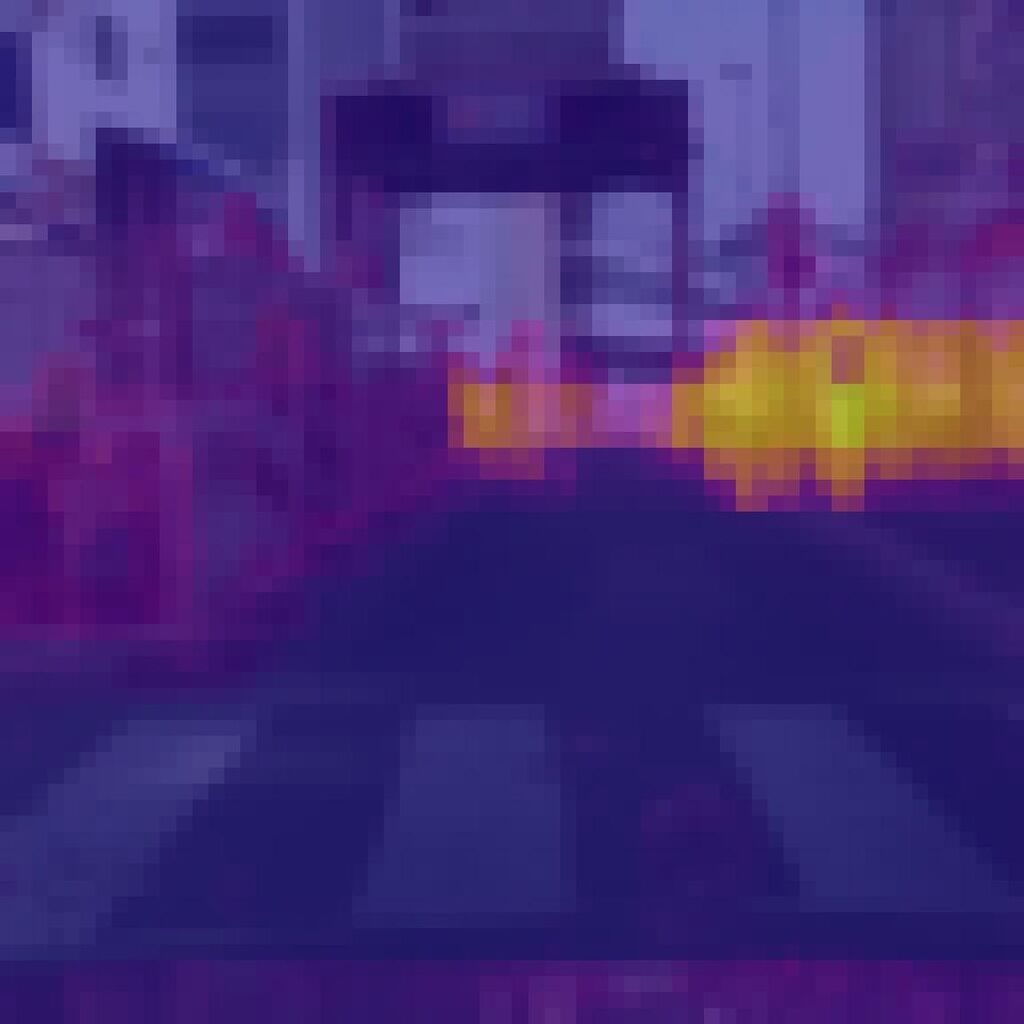}} &
    \fcolorbox{red}{white}{\includegraphics[width=0.95\linewidth]{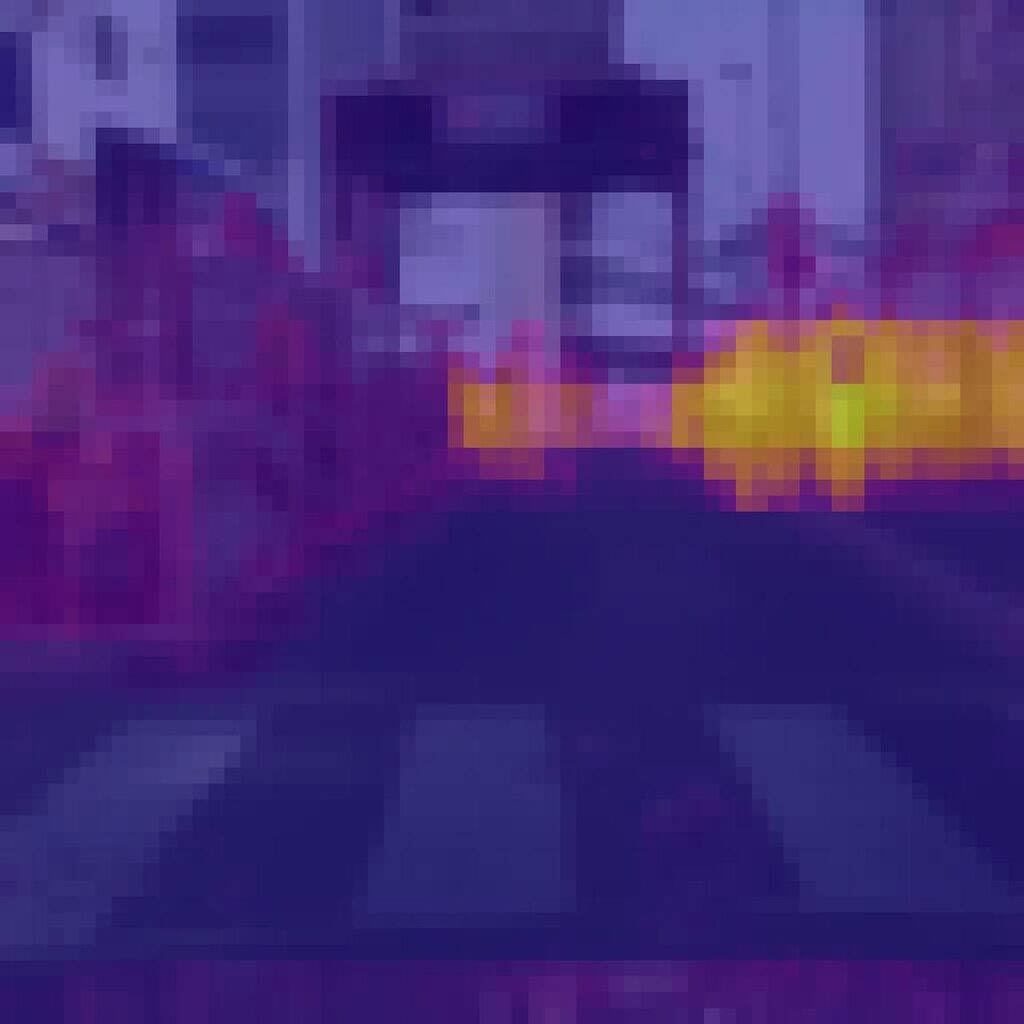}} &
    \fcolorbox{red}{white}{\includegraphics[width=0.95\linewidth]{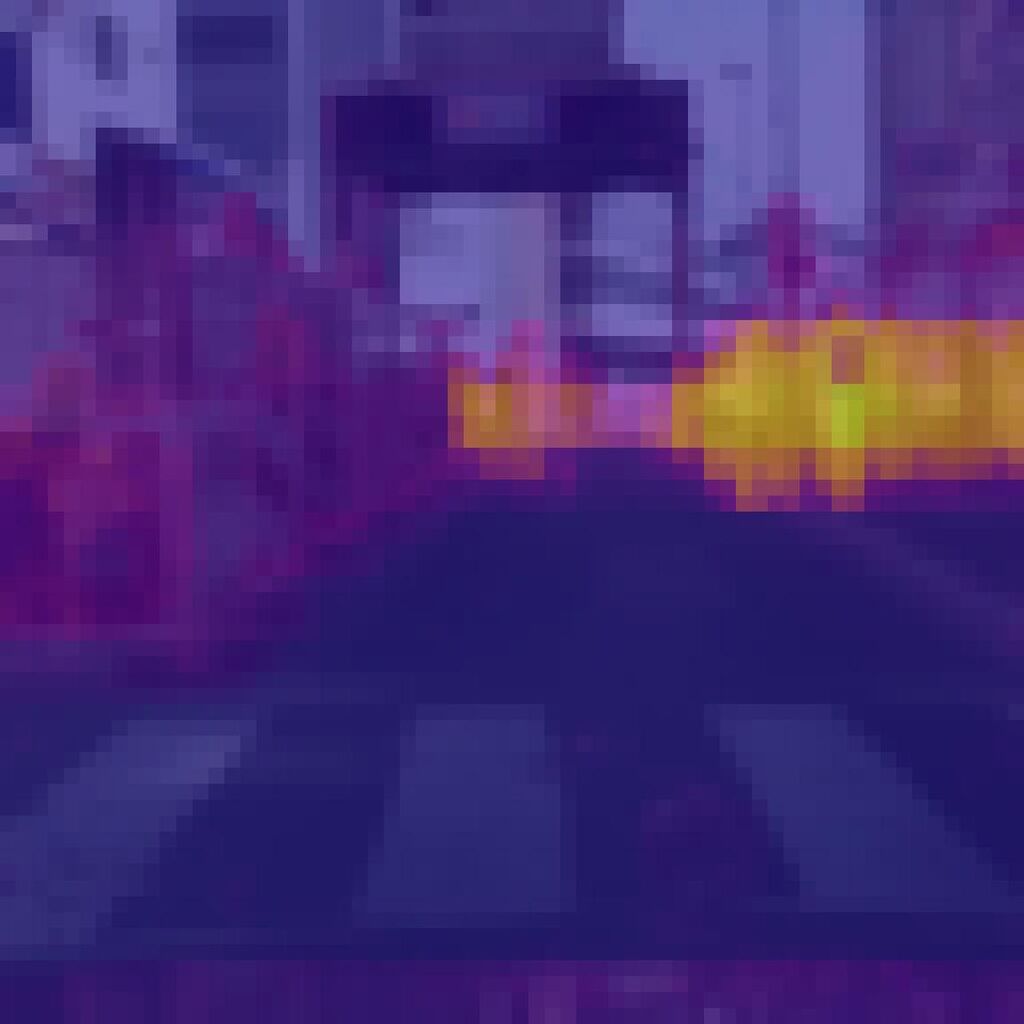}} &
    \fcolorbox{red}{white}{\includegraphics[width=0.95\linewidth]{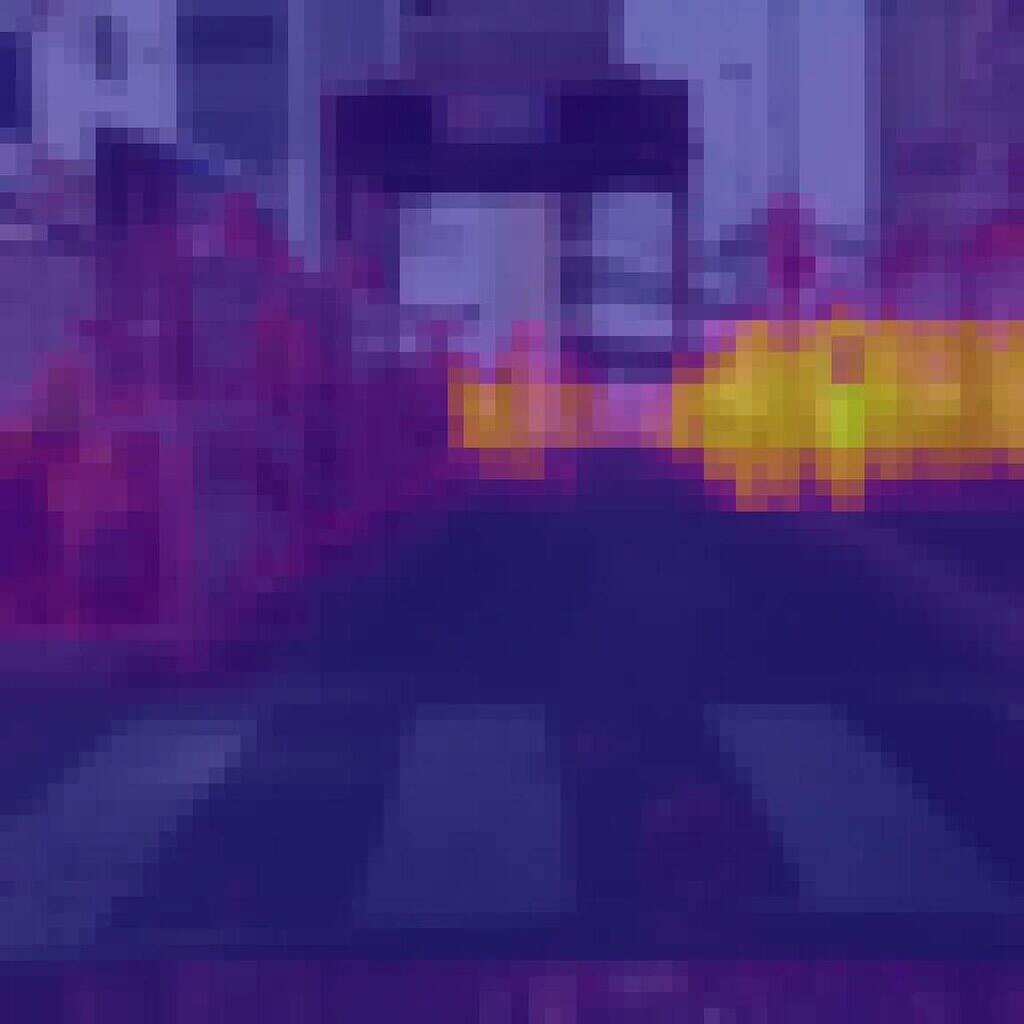}} &
    \fcolorbox{red}{white}{\includegraphics[width=0.95\linewidth]{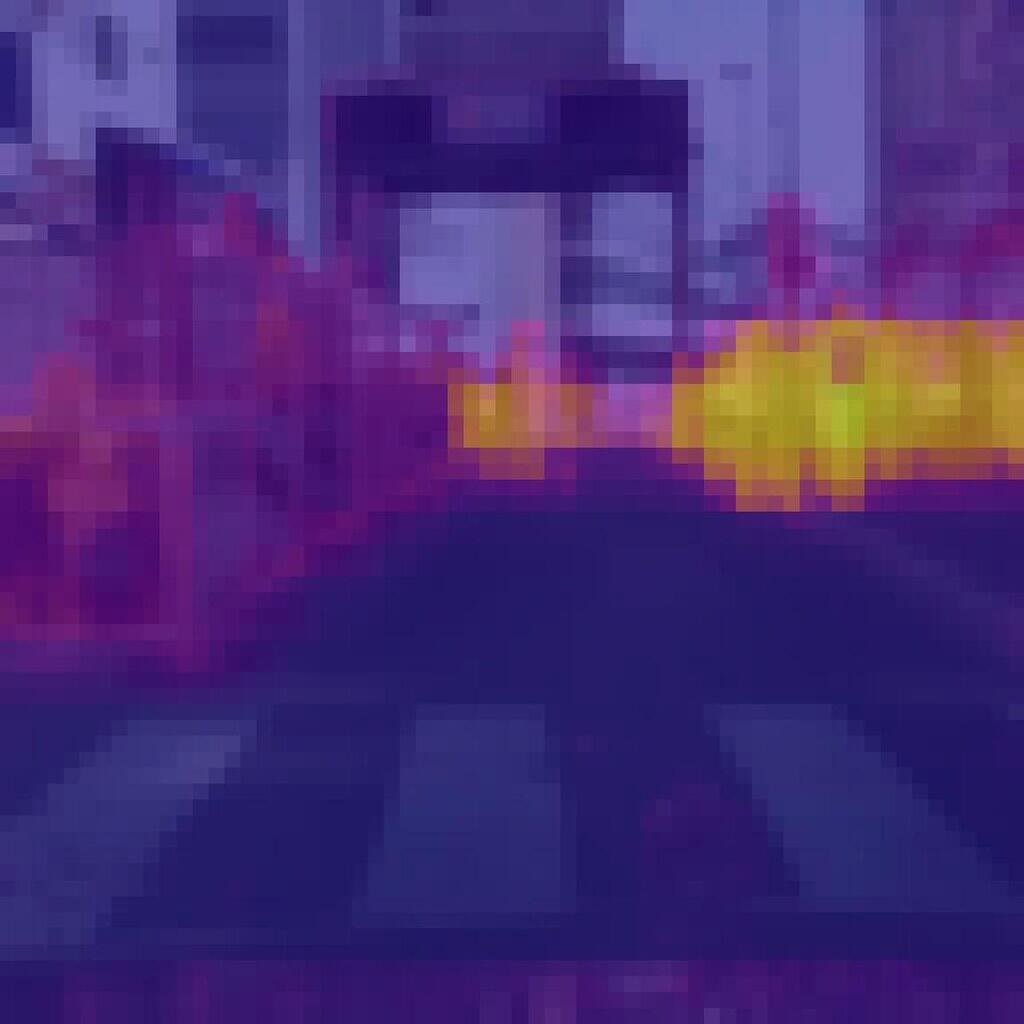}} &
    \fcolorbox{red}{white}{\includegraphics[width=0.95\linewidth]{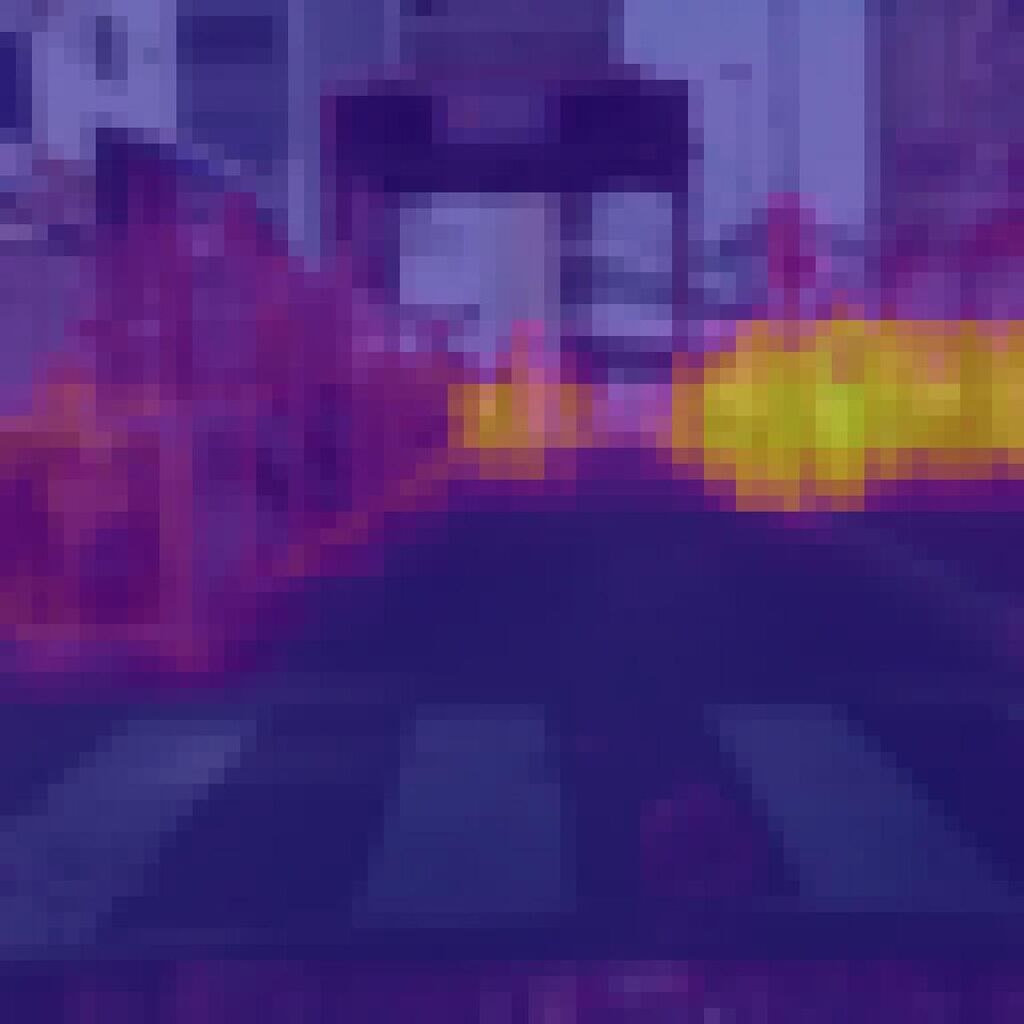}} &
    \fcolorbox{red}{white}{\includegraphics[width=0.95\linewidth]{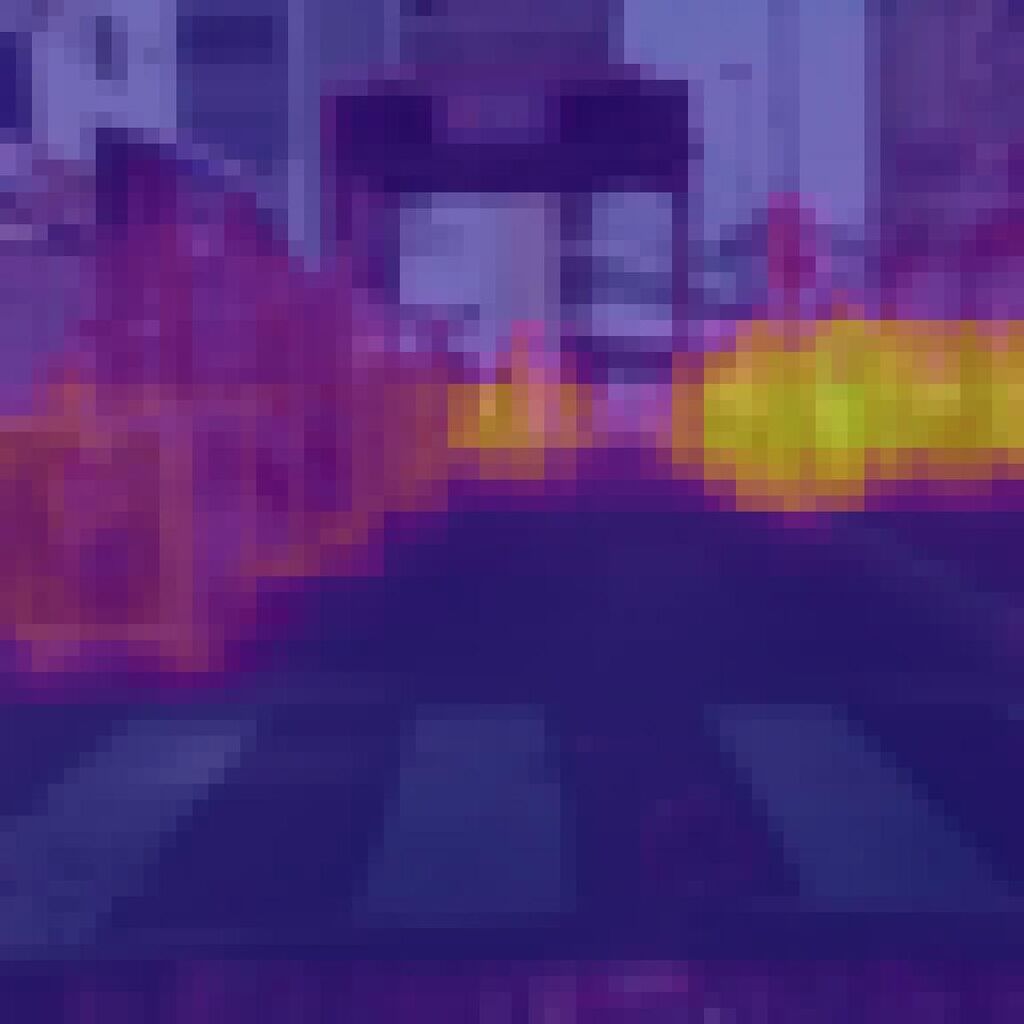}} &
    \fcolorbox{red}{white}{\includegraphics[width=0.95\linewidth]{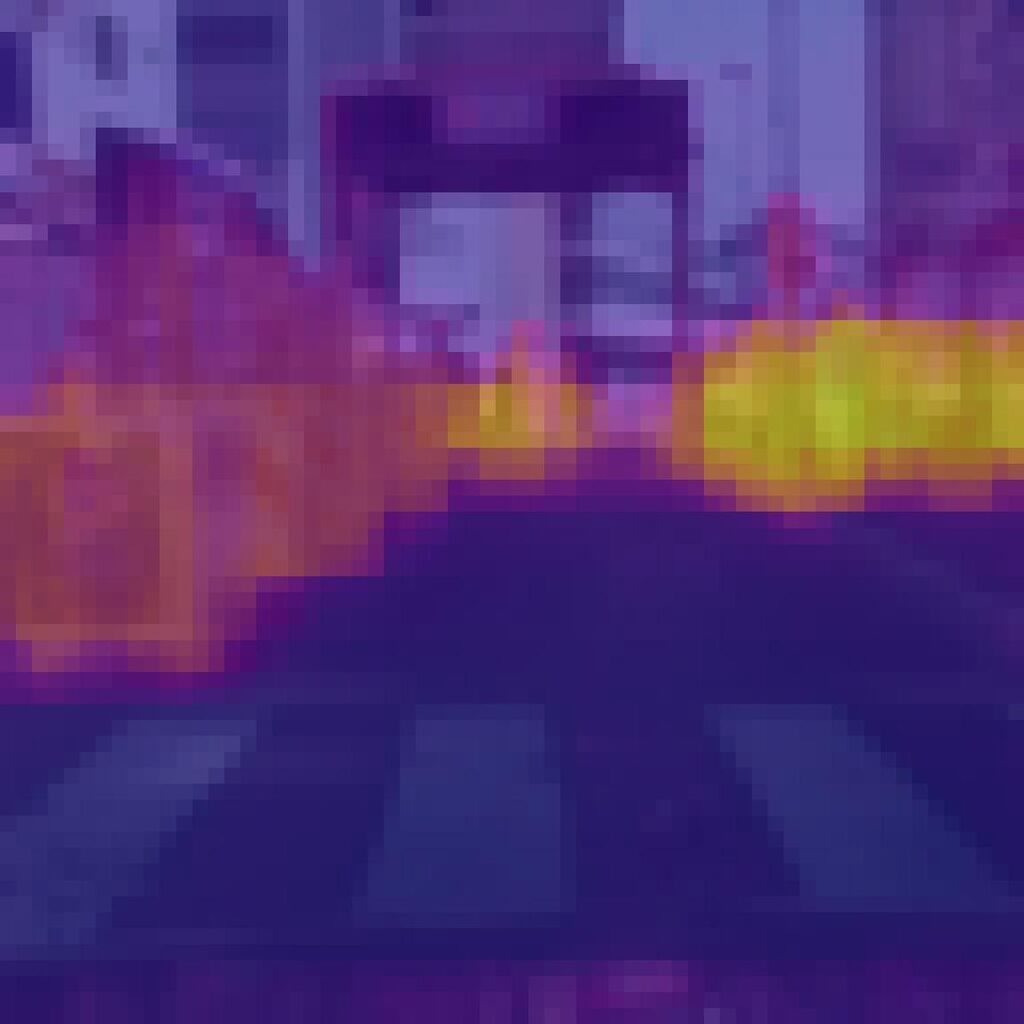}} &
    \fcolorbox{red}{white}{\includegraphics[width=0.95\linewidth]{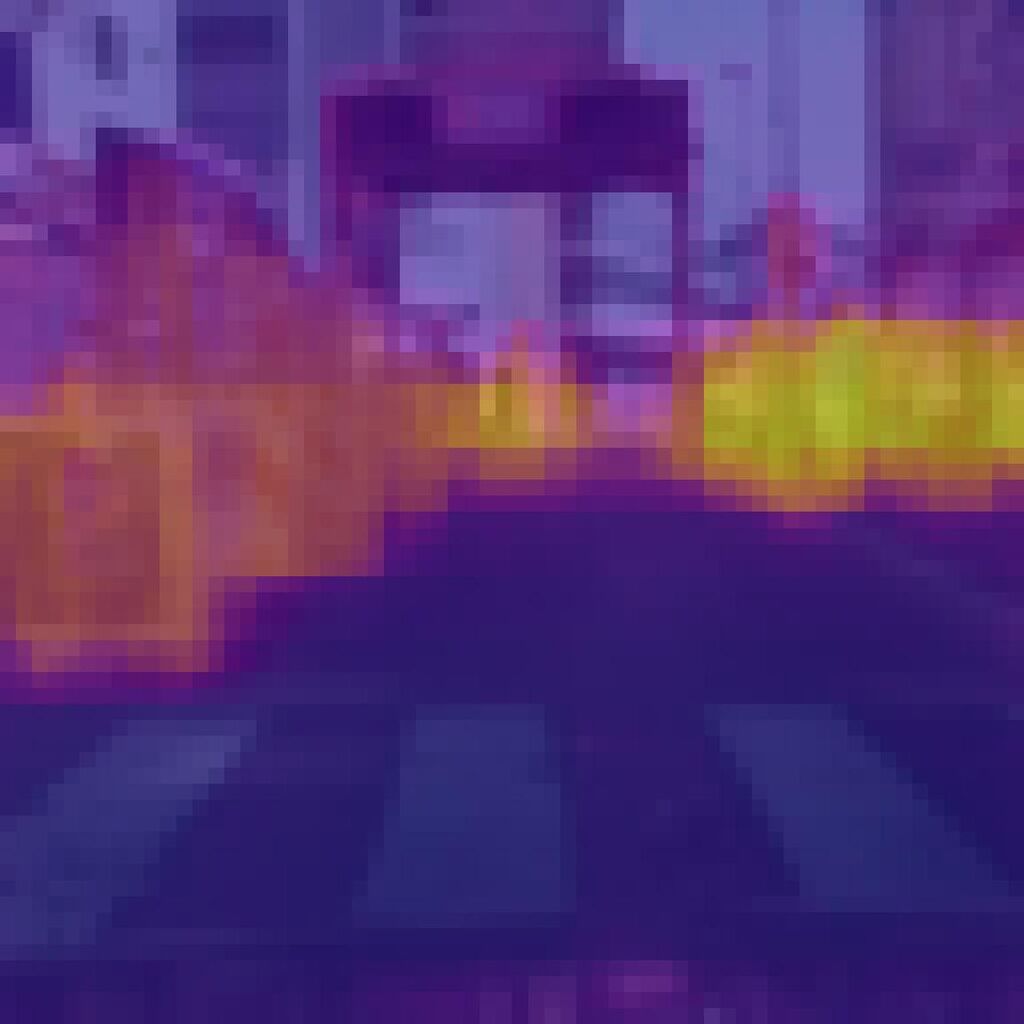}} &
    \fcolorbox{red}{white}{\includegraphics[width=0.95\linewidth]{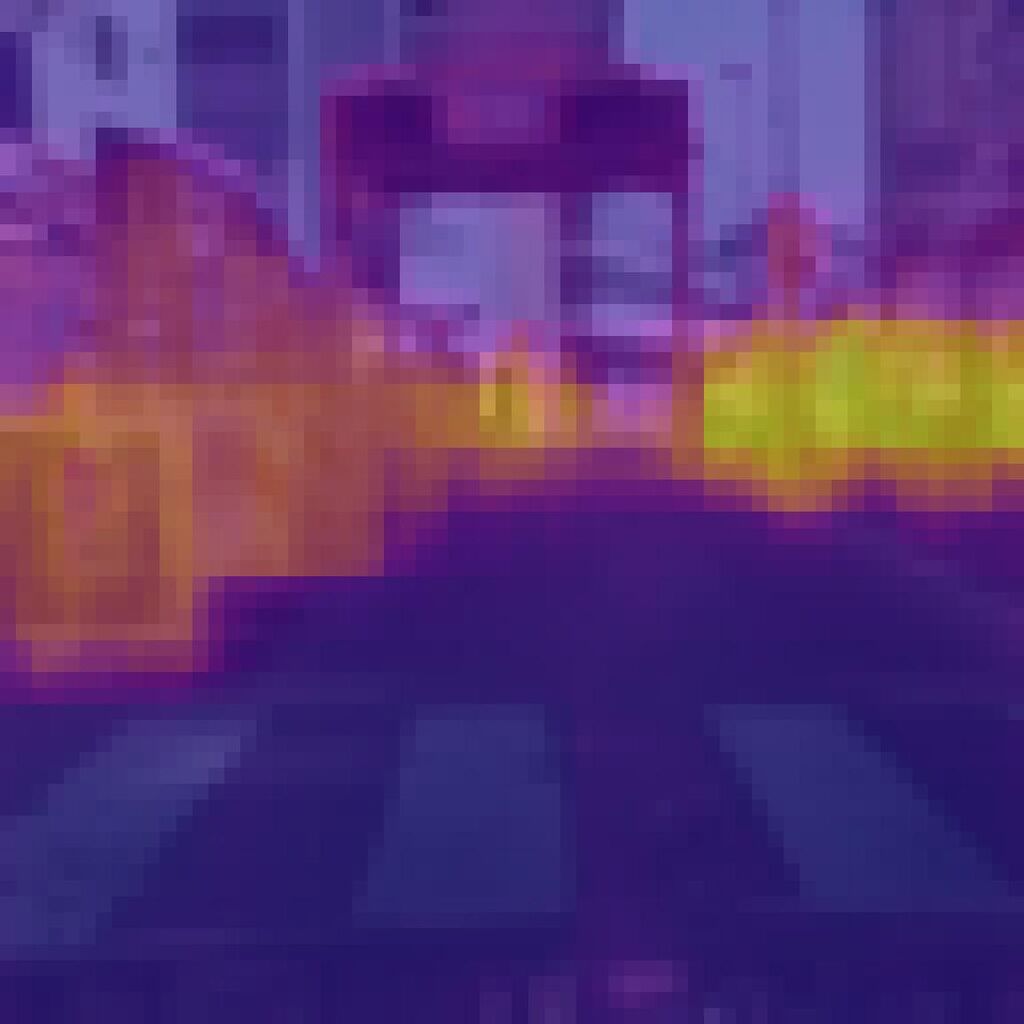}} \\

    t=10 & t=20 & t=30 & t=40 & t=50 & t=100 & t=150 & t=200 & t=250 & t=300 & t=350 \\[1em]
    
    \fcolorbox{red}{white}{\includegraphics[width=0.95\linewidth]{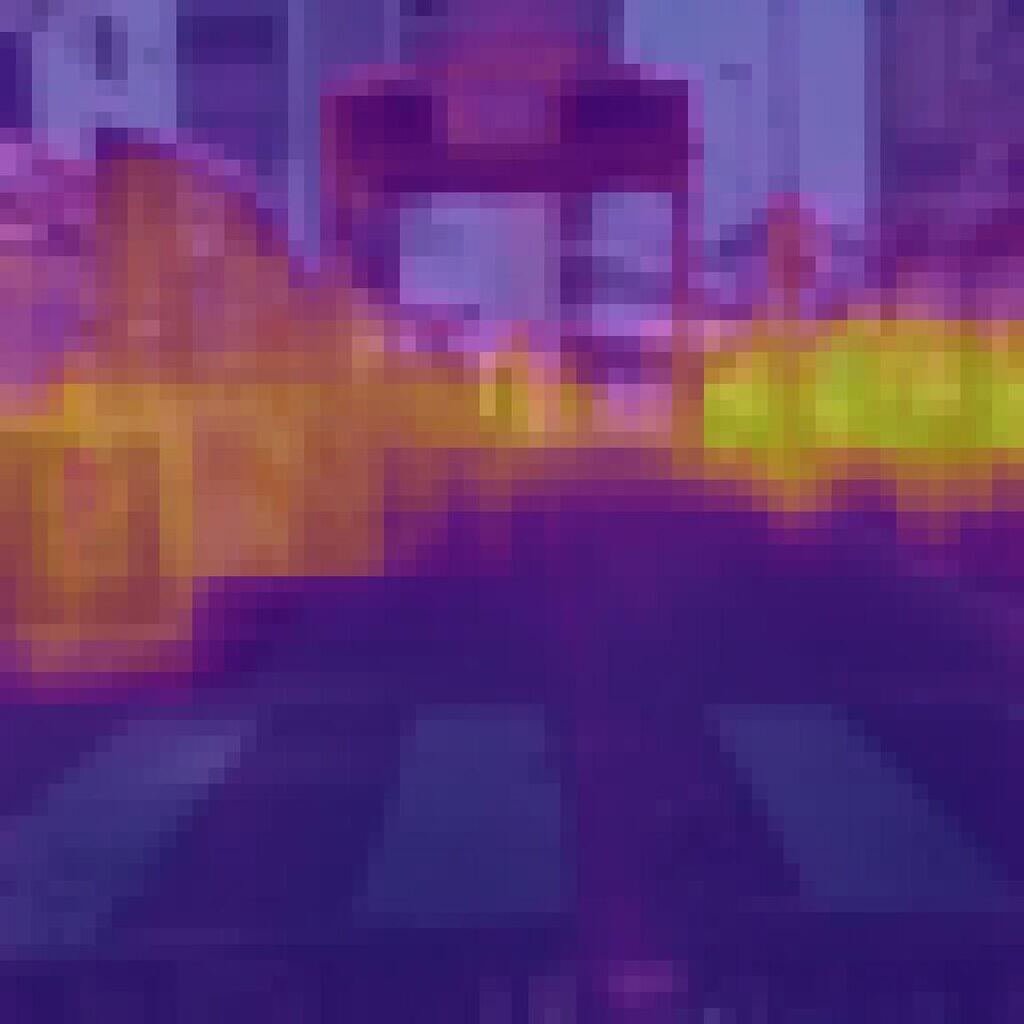}} &
    \fcolorbox{red}{white}{\includegraphics[width=0.95\linewidth]{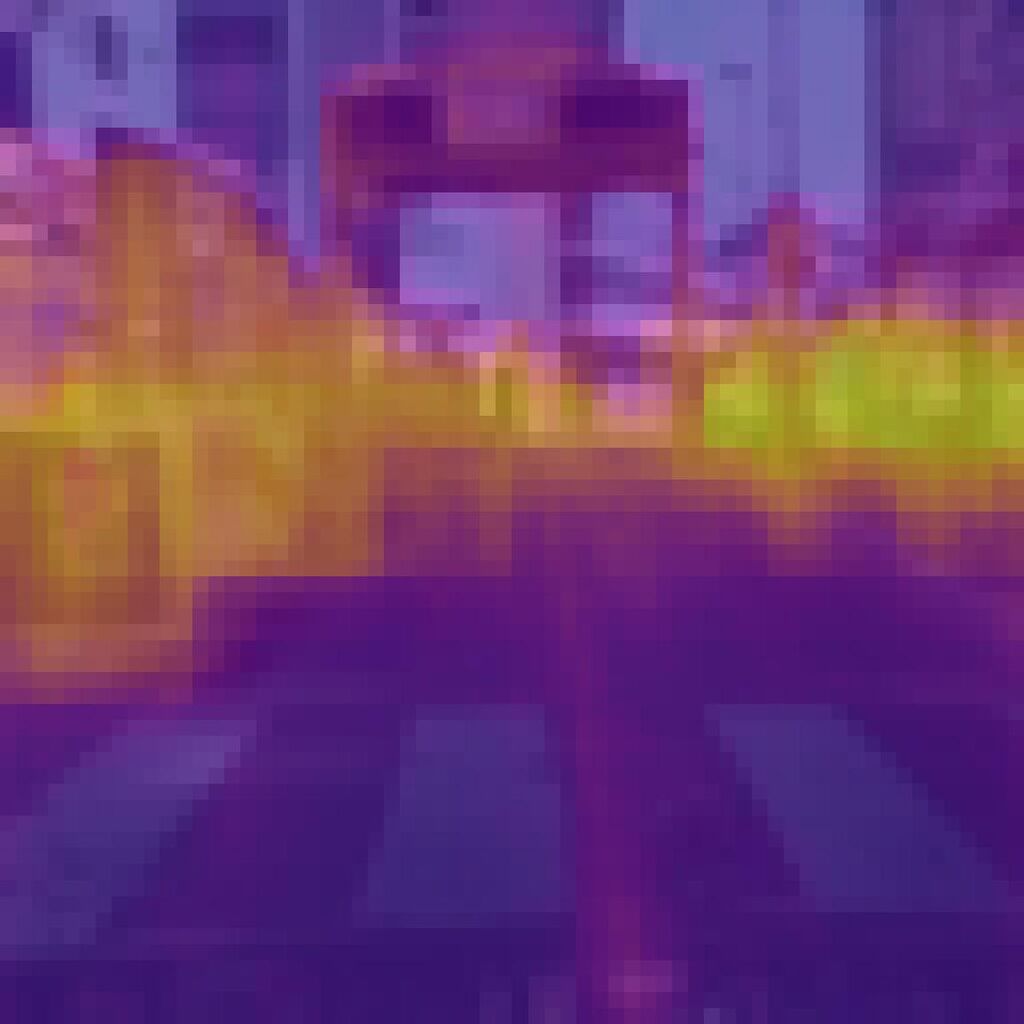}} &
    \fcolorbox{red}{white}{\includegraphics[width=0.95\linewidth]{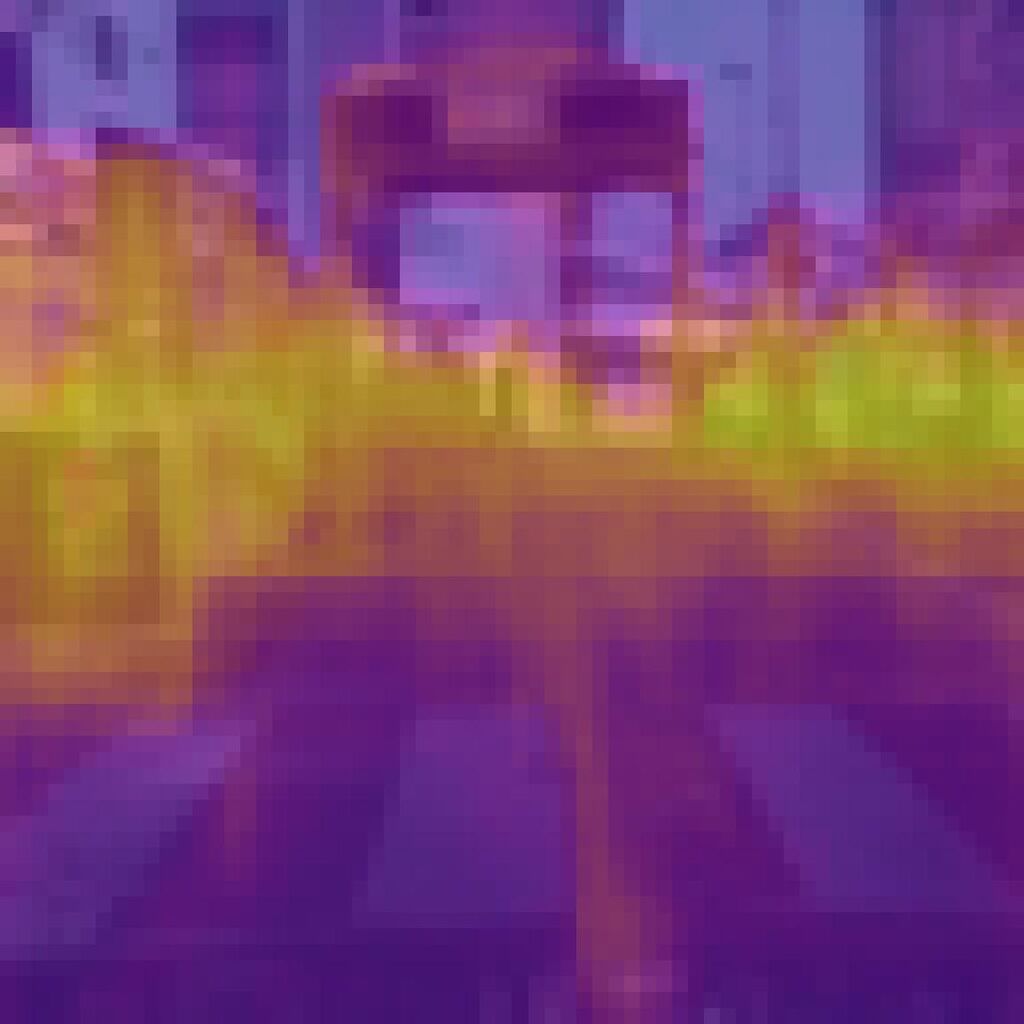}} &
    \fcolorbox{red}{white}{\includegraphics[width=0.95\linewidth]{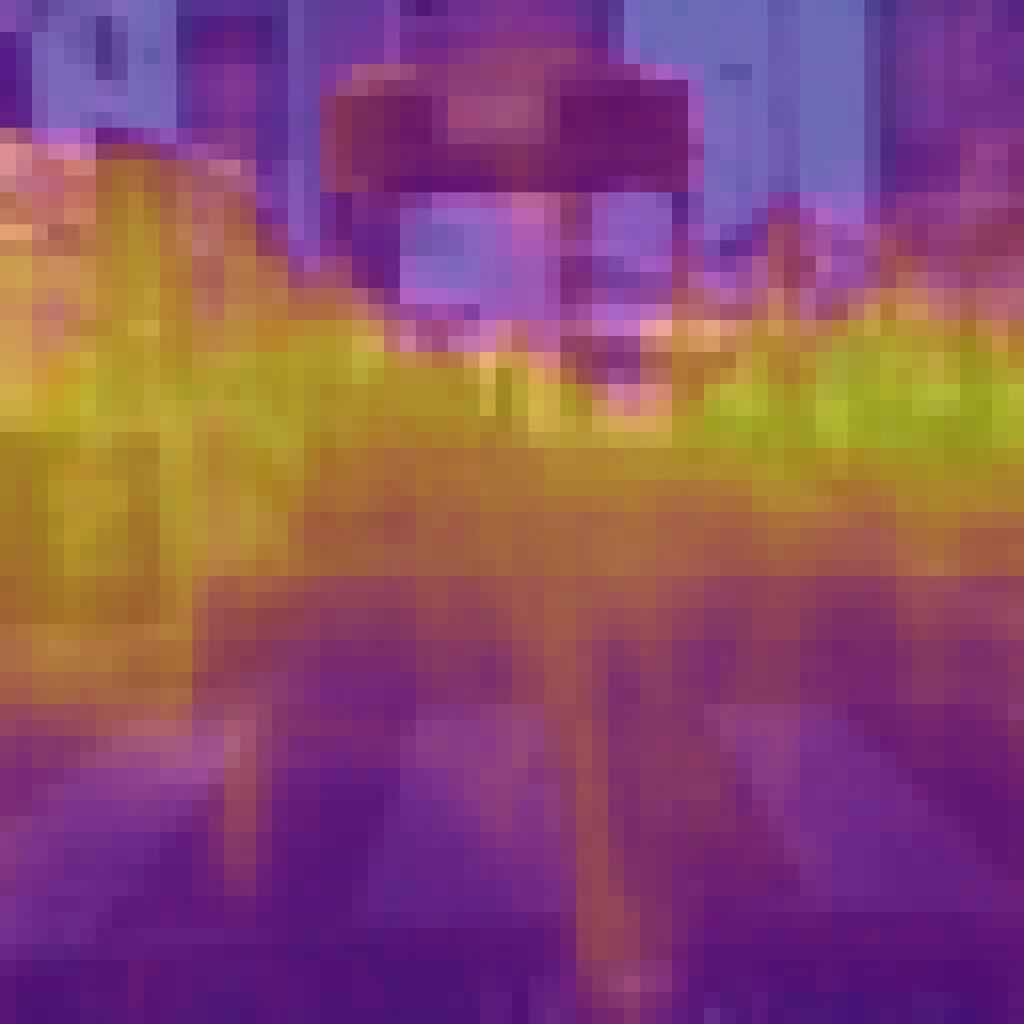}} &
    \fcolorbox{red}{white}{\includegraphics[width=0.95\linewidth]{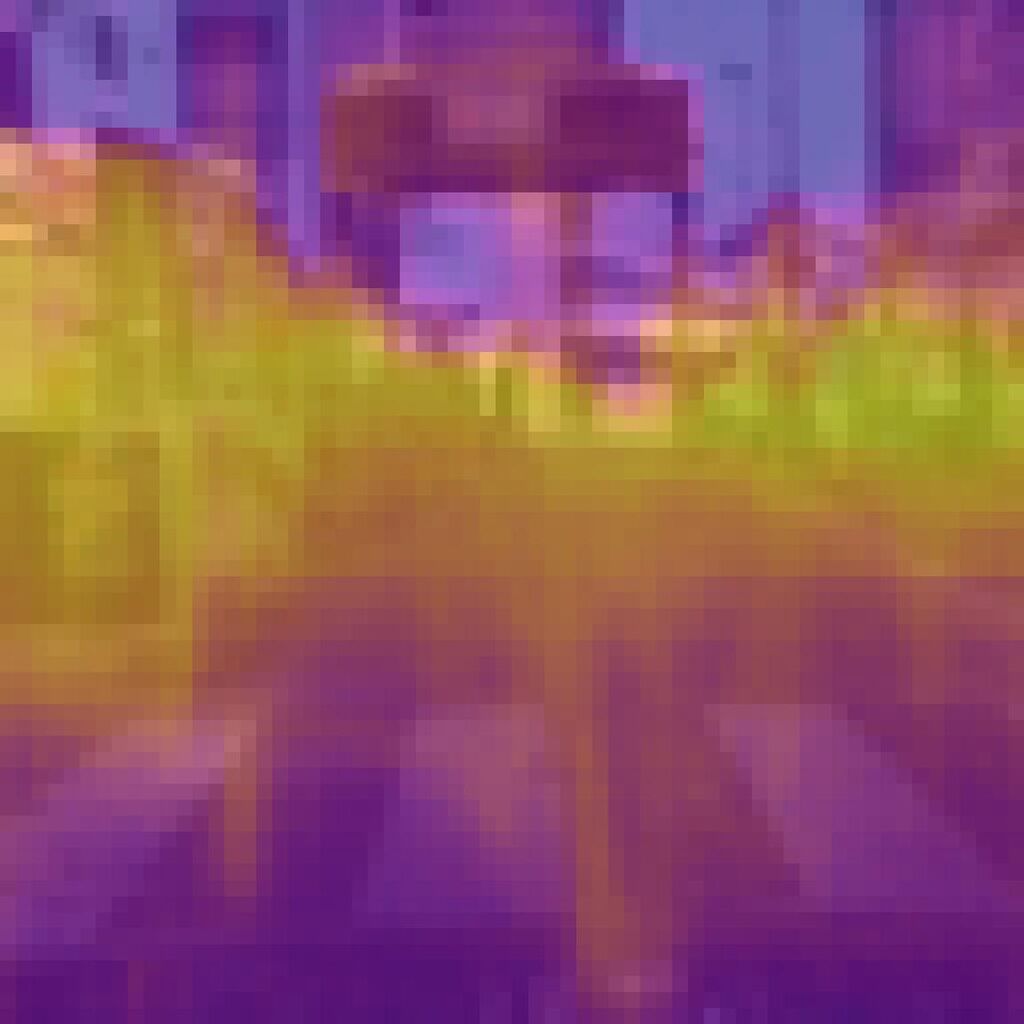}} &
    \fcolorbox{red}{white}{\includegraphics[width=0.95\linewidth]{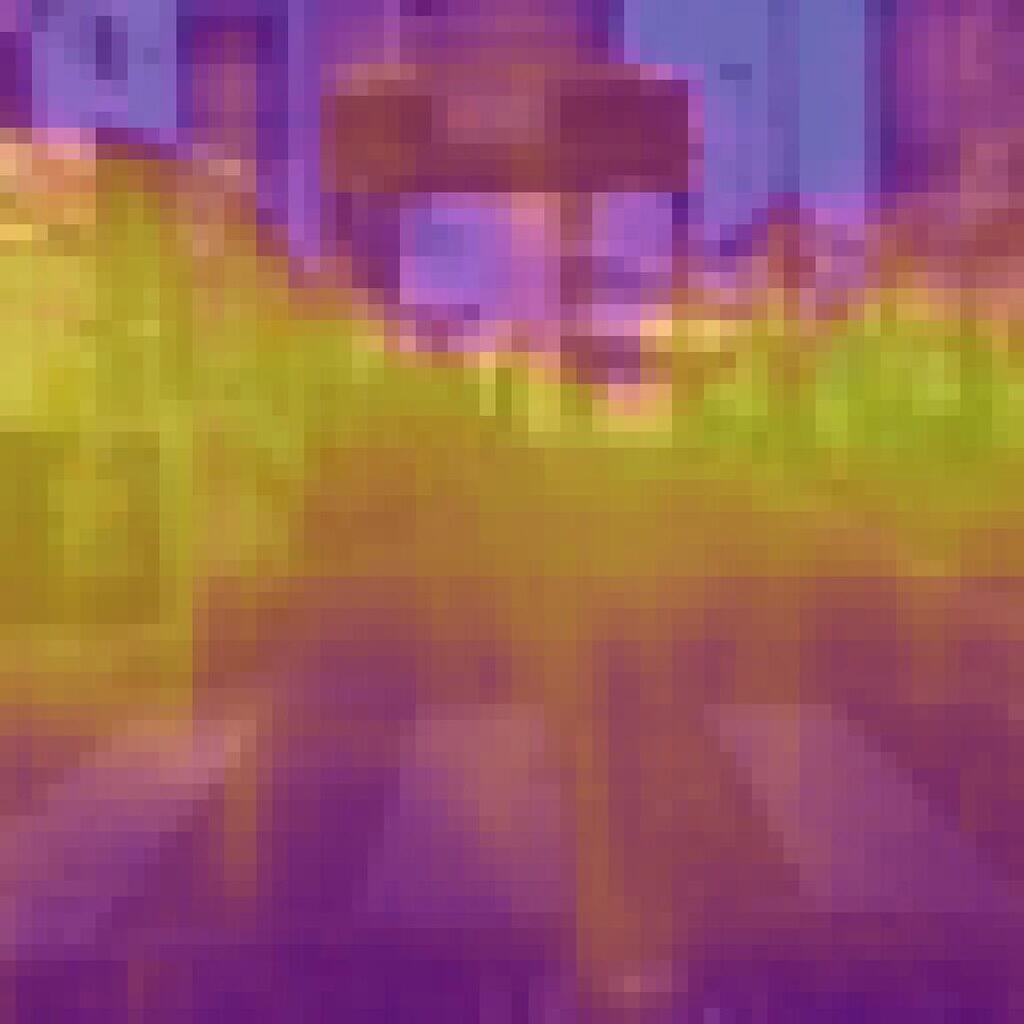}} &
    \fcolorbox{red}{white}{\includegraphics[width=0.95\linewidth]{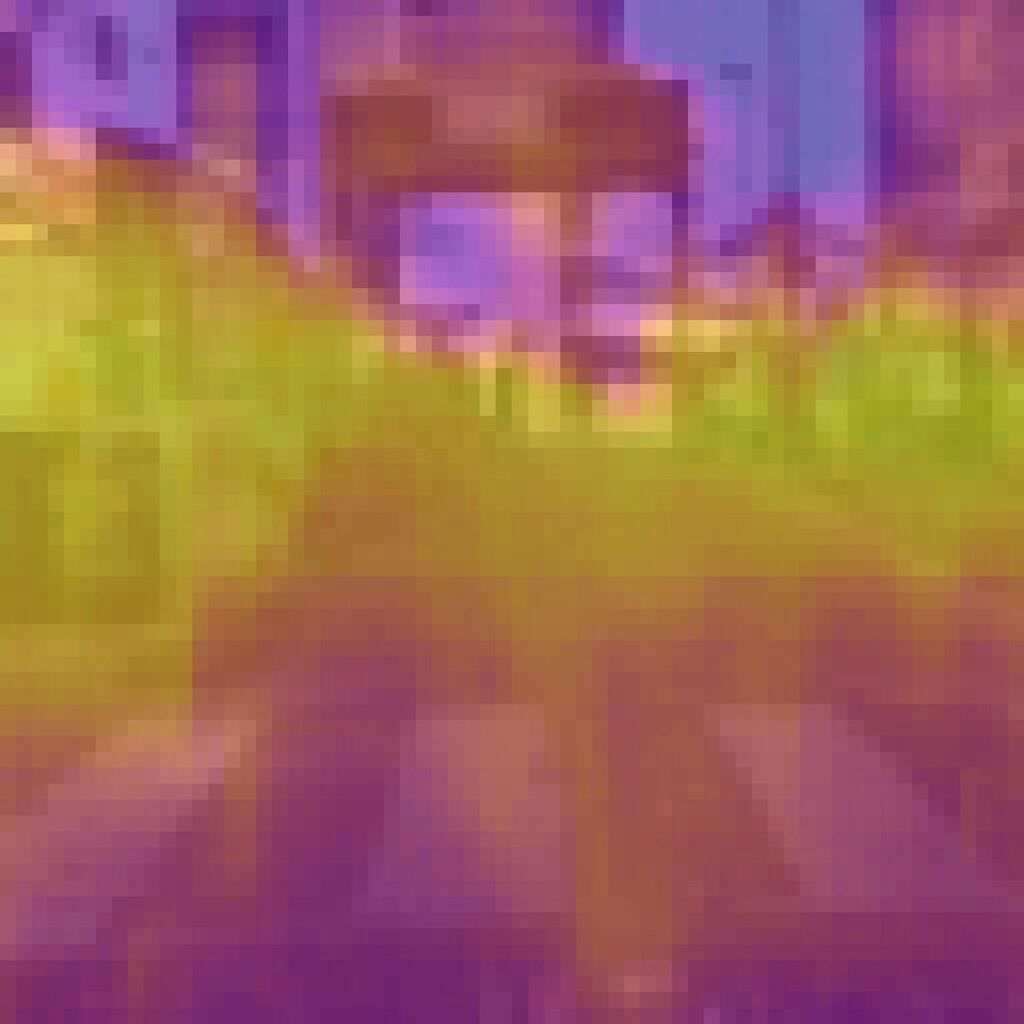}} &
    \fcolorbox{red}{white}{\includegraphics[width=0.95\linewidth]{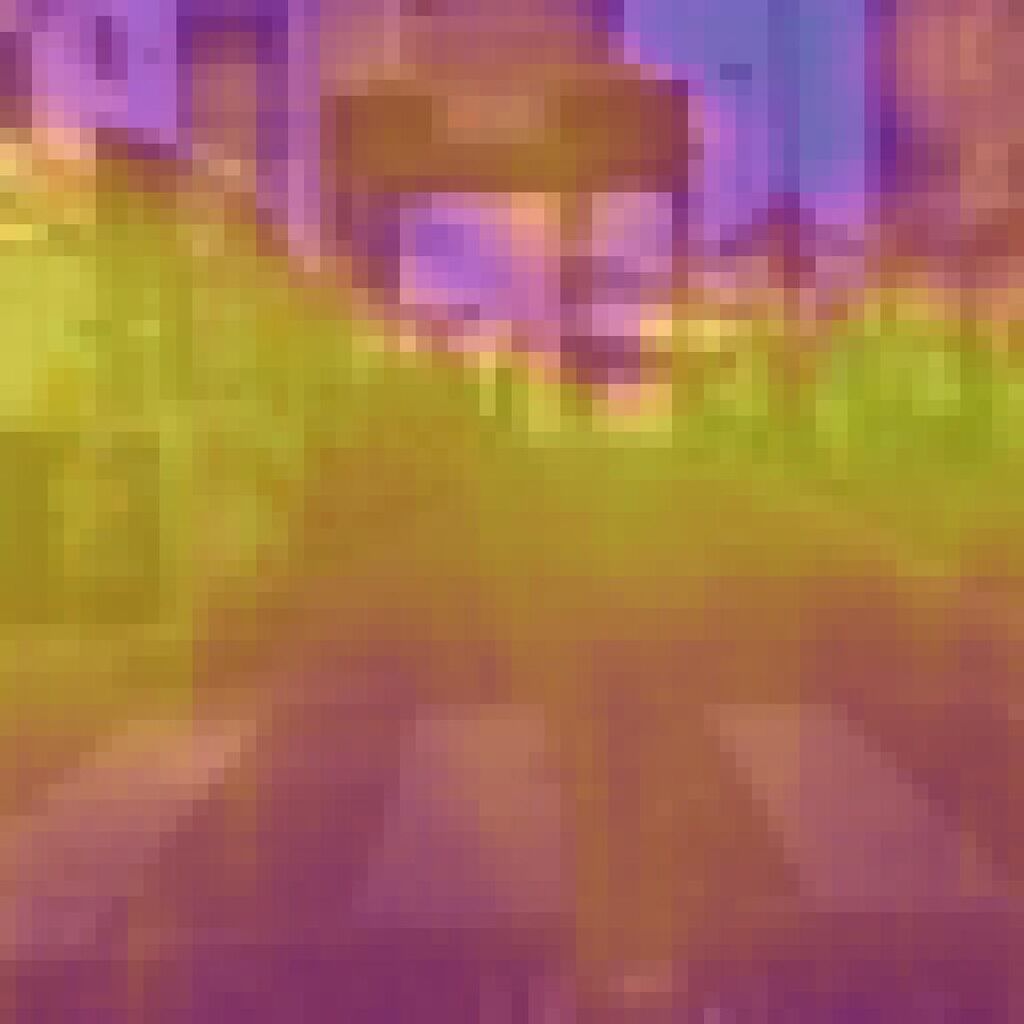}} &
    \fcolorbox{red}{white}{\includegraphics[width=0.95\linewidth]{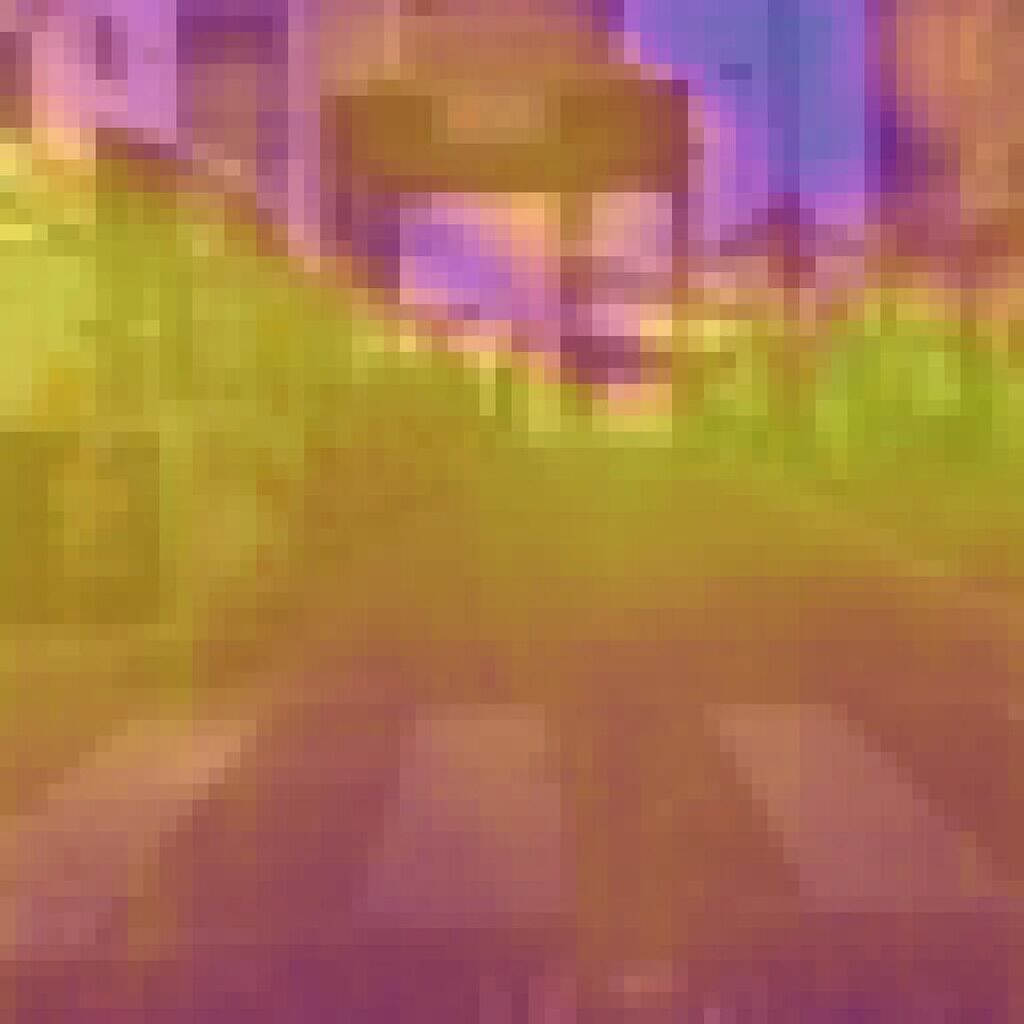}} &
    \fcolorbox{red}{white}{\includegraphics[width=0.95\linewidth]{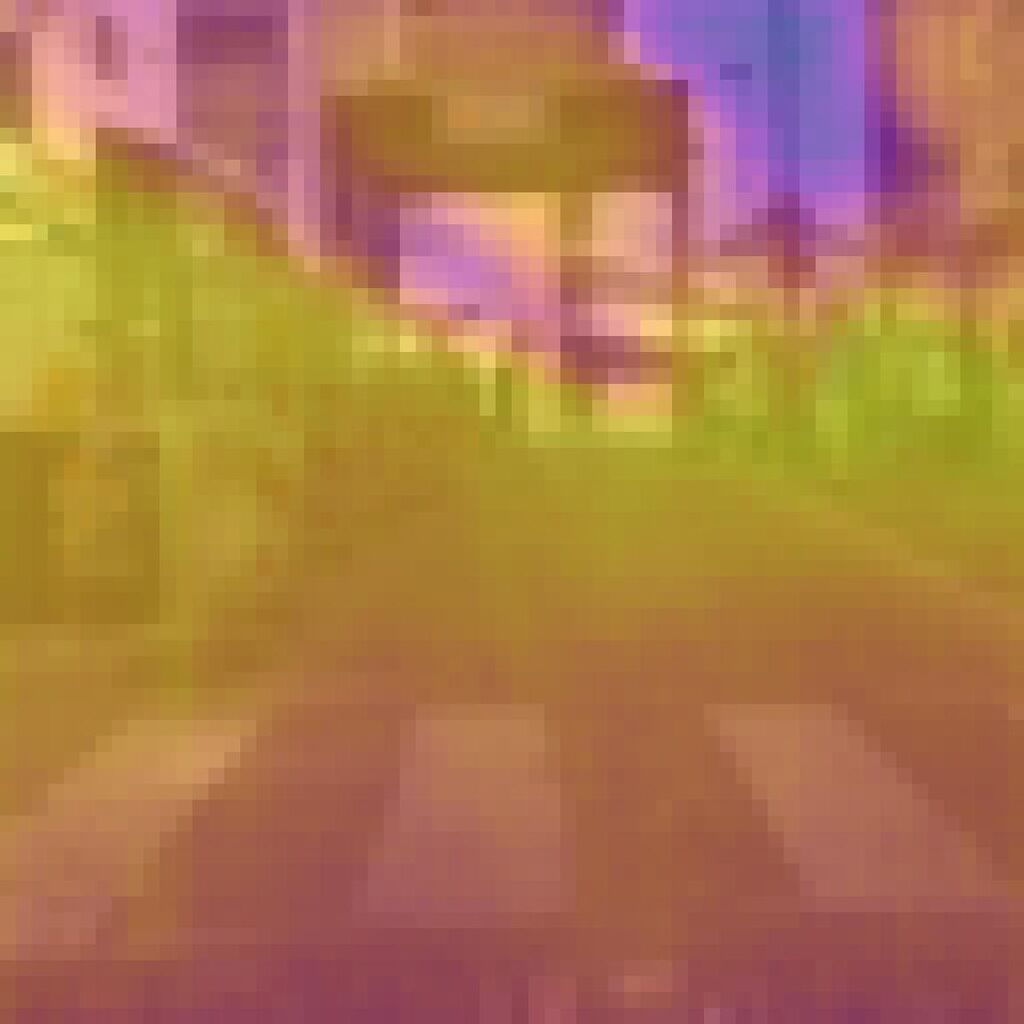}} &
    \fcolorbox{red}{white}{\includegraphics[width=0.95\linewidth]{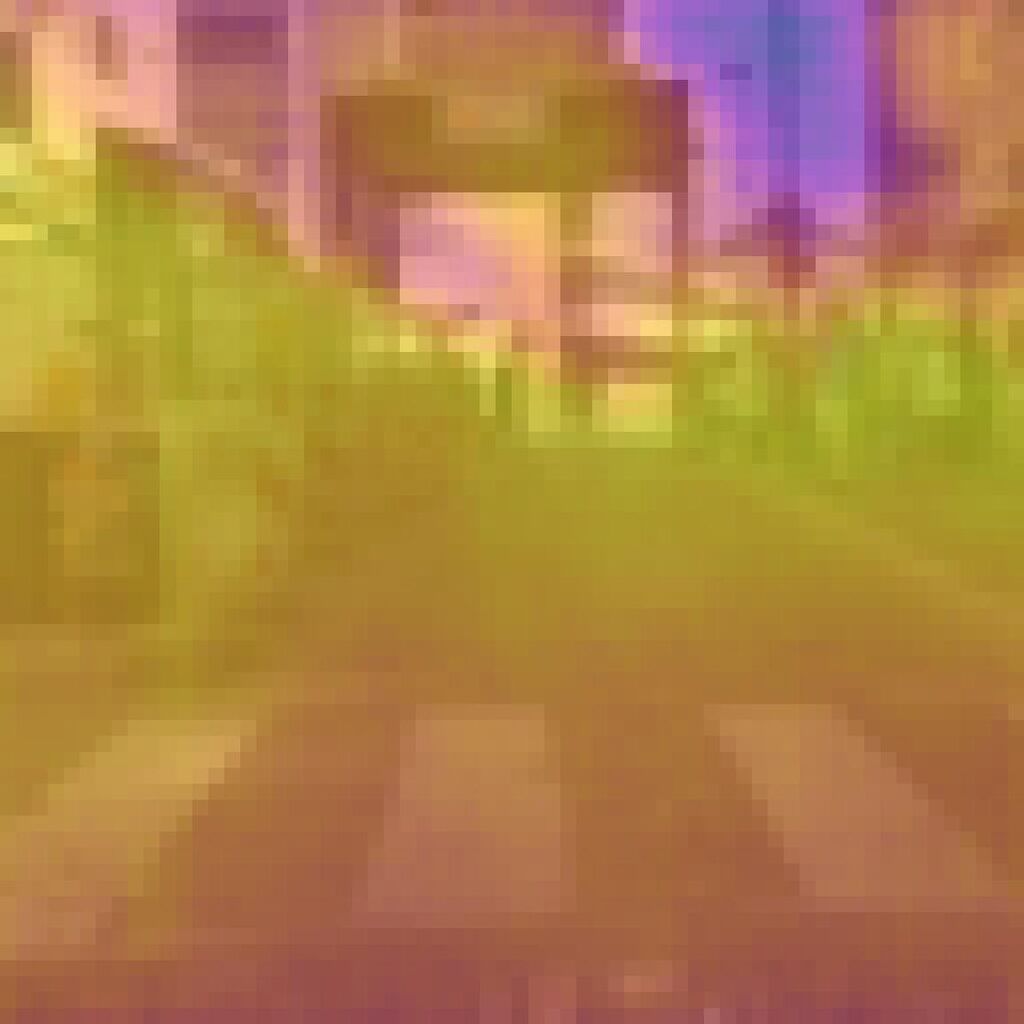}} \\

    t=400 & t=450 & t=500 & t=550 & t=600 & t=650 & t=700 & t=750 & t=800 & t=850 & t=900 \\
    
    \multicolumn{11}{c}{\vspace{0.5em}} \\ 
    
    \multicolumn{11}{c}{\textbf{Blue Point CSM}} \\
    
    \fcolorbox{blue}{white}{\includegraphics[width=0.95\linewidth]{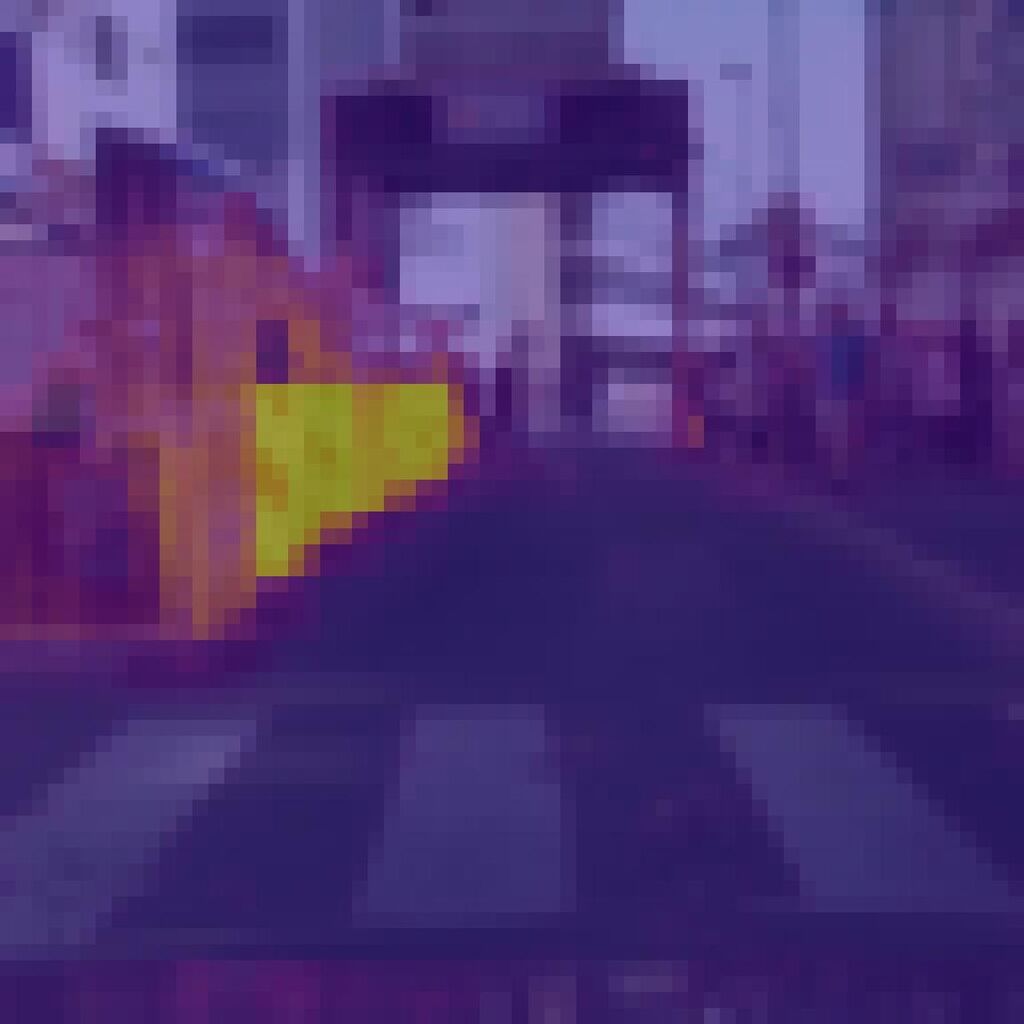}} &
    \fcolorbox{blue}{white}{\includegraphics[width=0.95\linewidth]{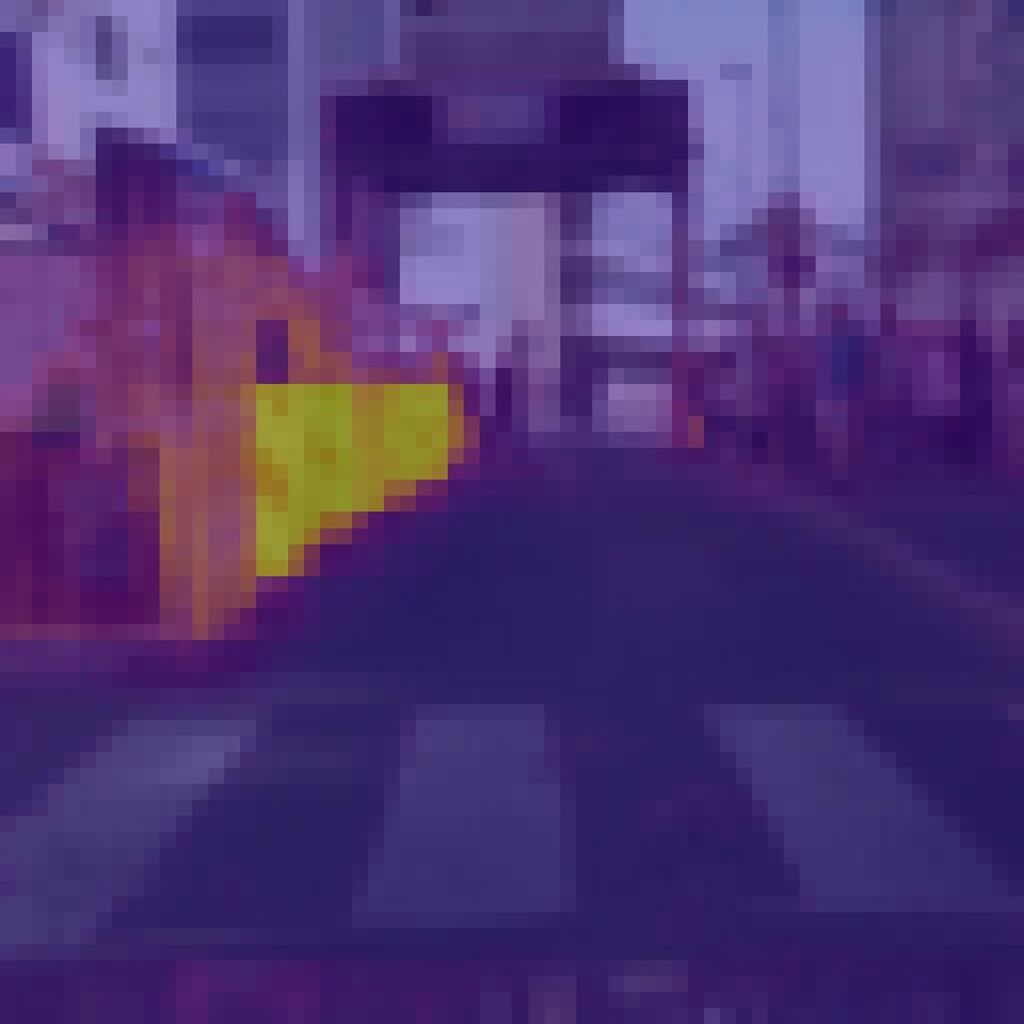}} &
    \fcolorbox{blue}{white}{\includegraphics[width=0.95\linewidth]{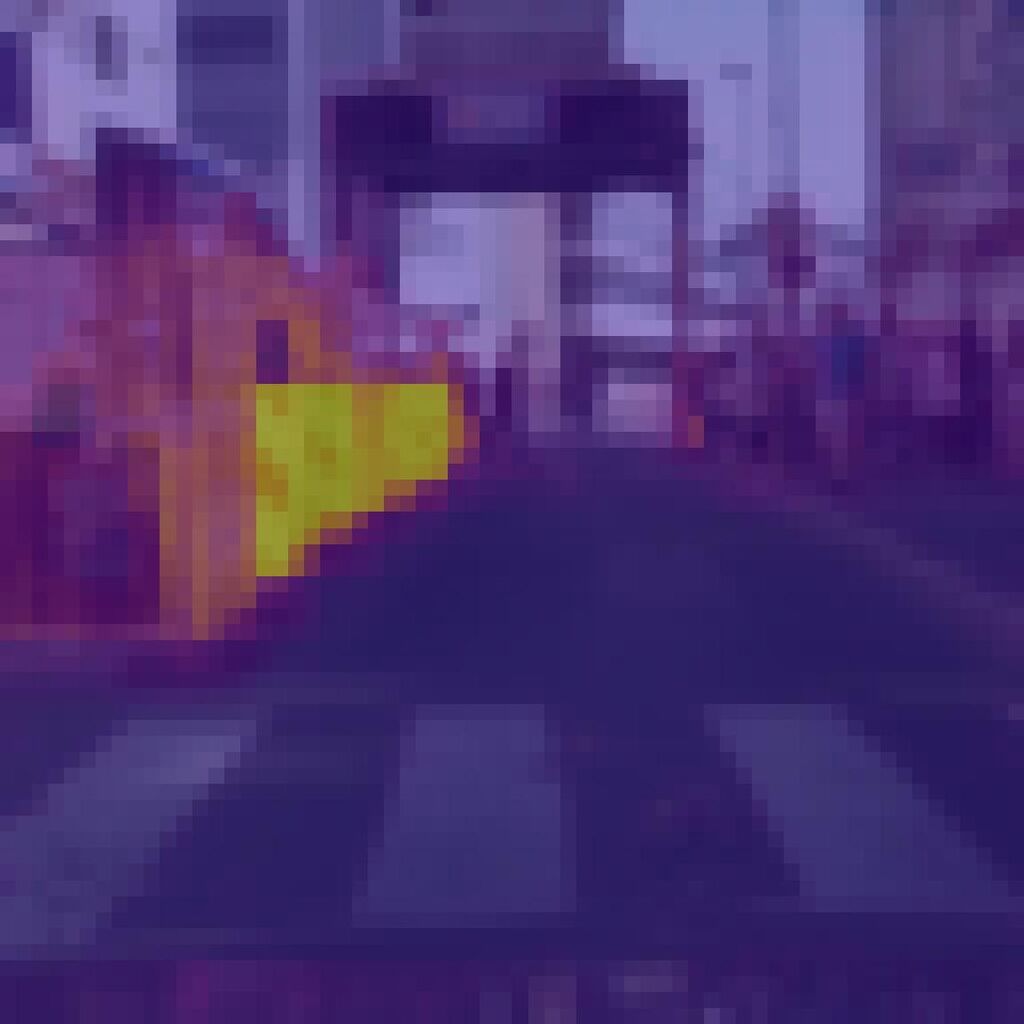}} &
    \fcolorbox{blue}{white}{\includegraphics[width=0.95\linewidth]{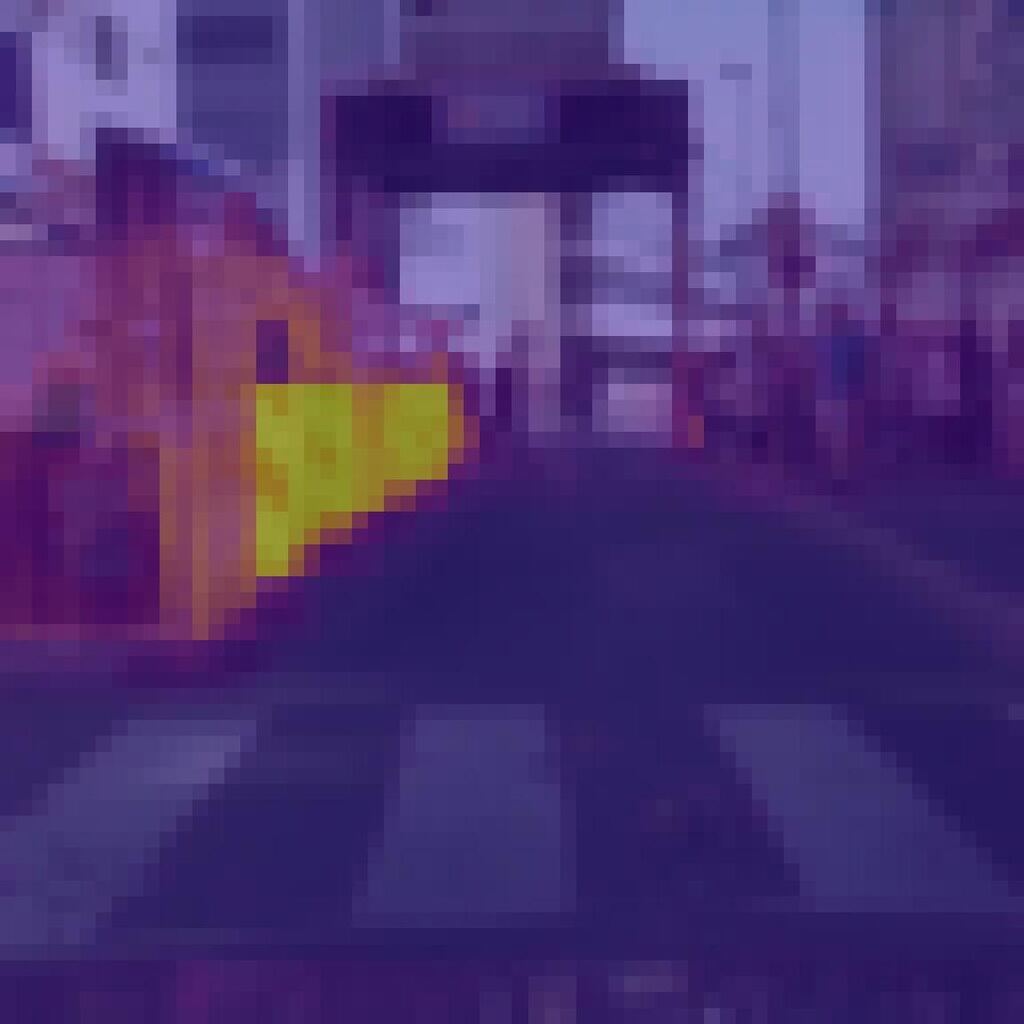}} &
    \fcolorbox{blue}{white}{\includegraphics[width=0.95\linewidth]{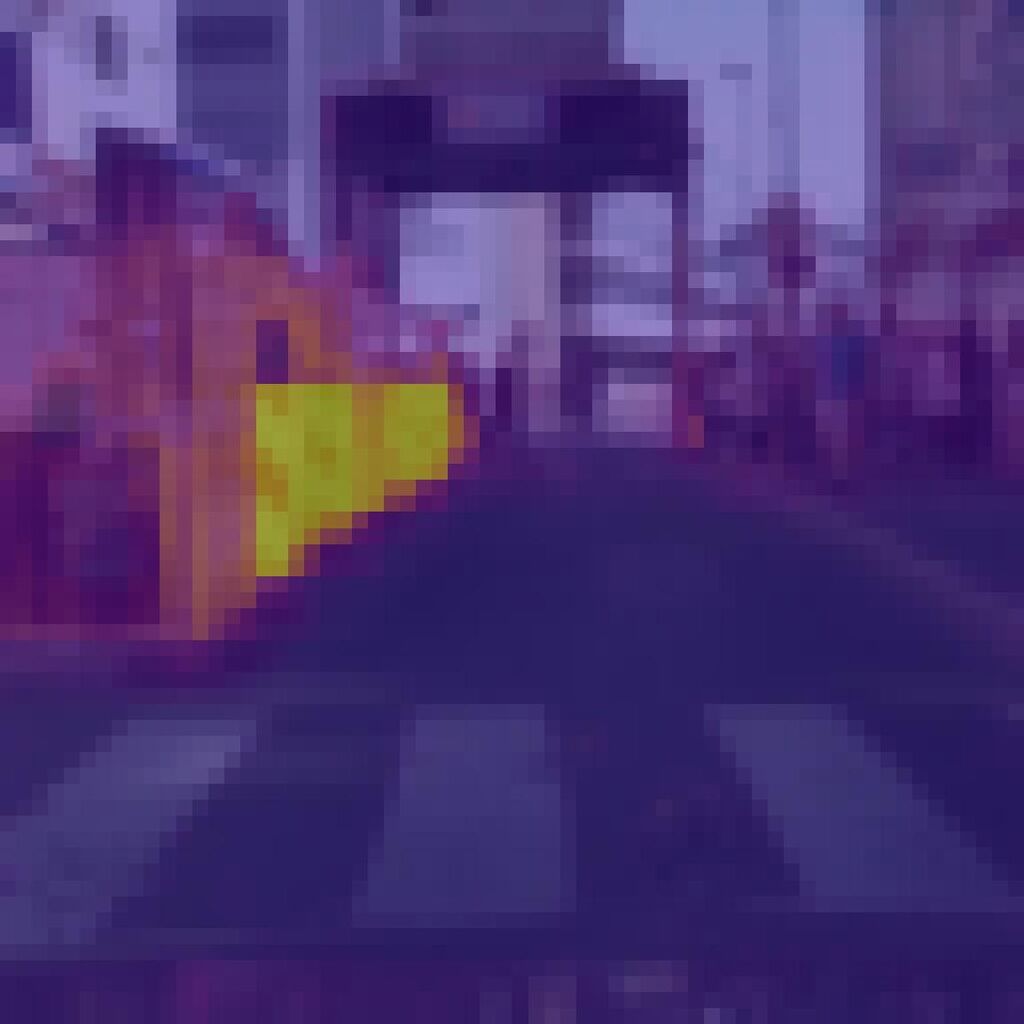}} &
    \fcolorbox{blue}{white}{\includegraphics[width=0.95\linewidth]{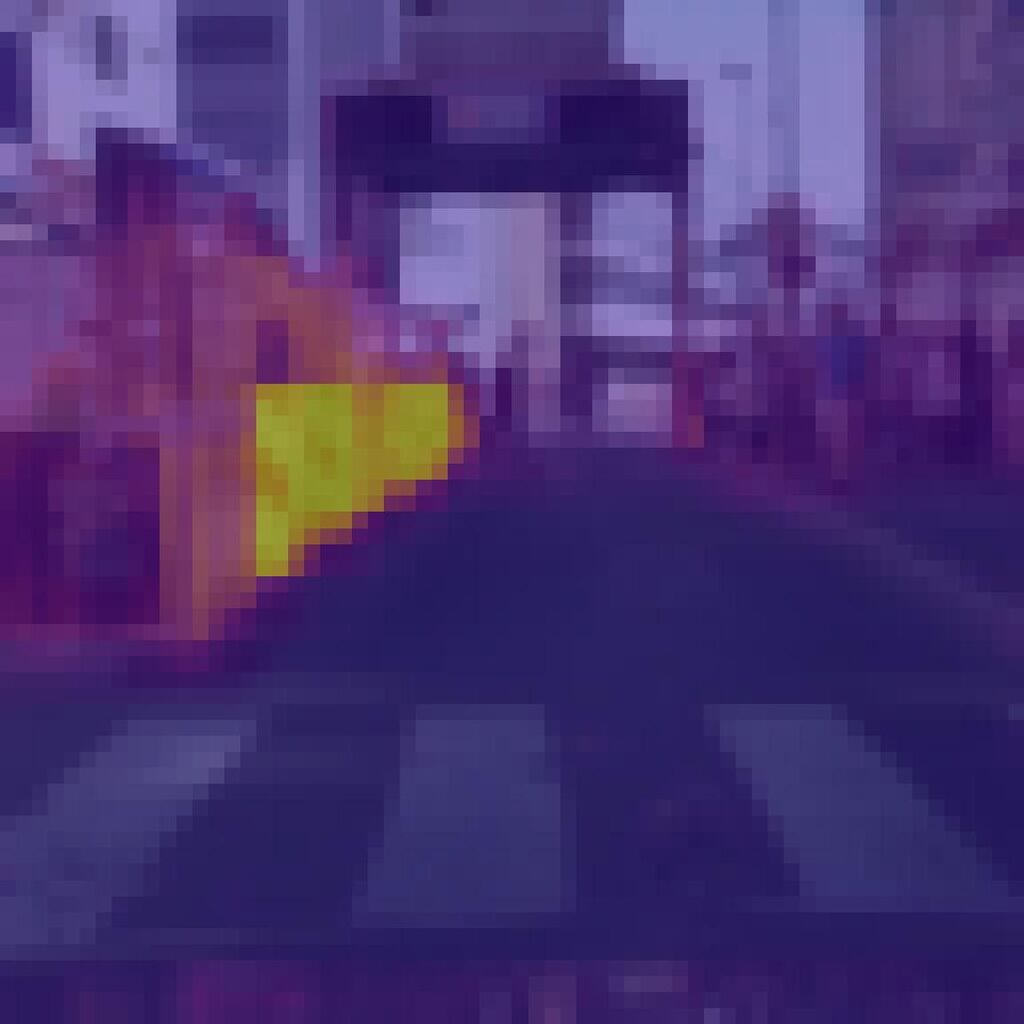}} &
    \fcolorbox{blue}{white}{\includegraphics[width=0.95\linewidth]{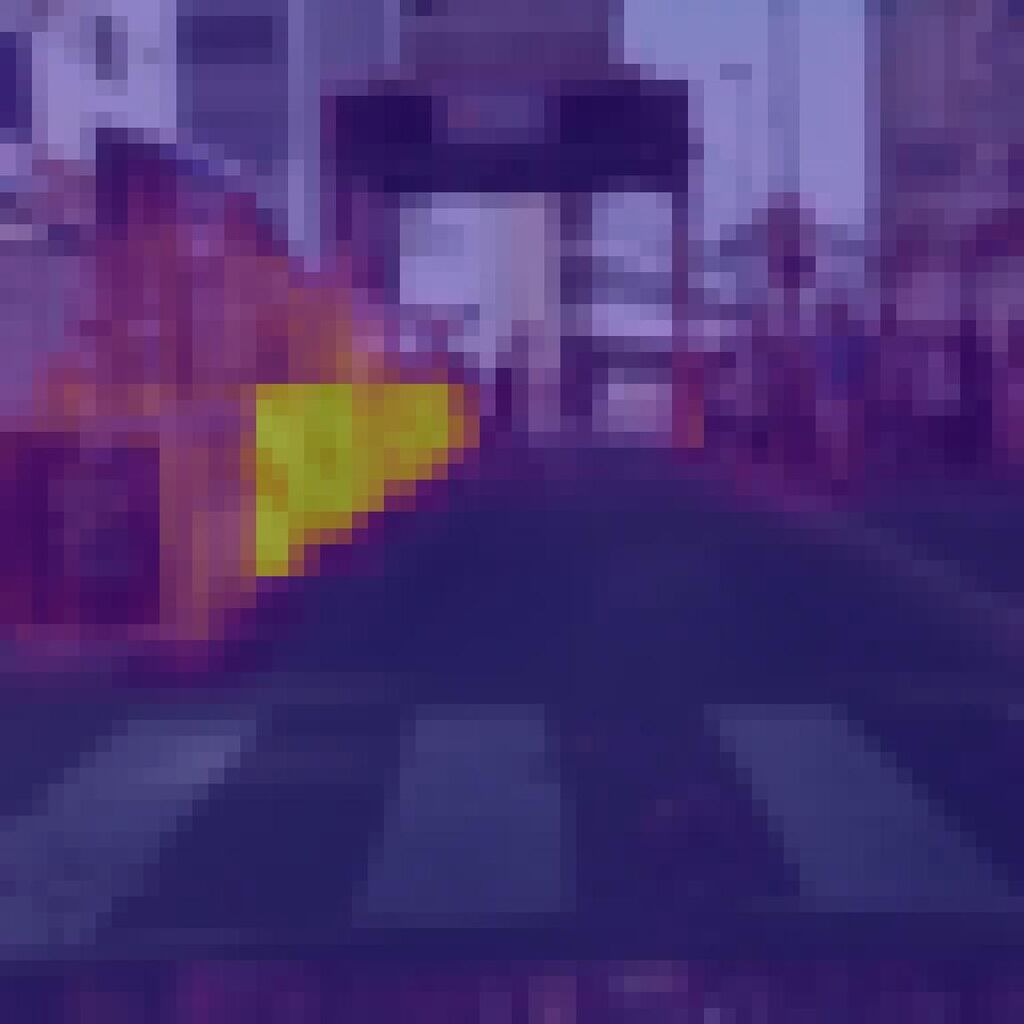}} &
    \fcolorbox{blue}{white}{\includegraphics[width=0.95\linewidth]{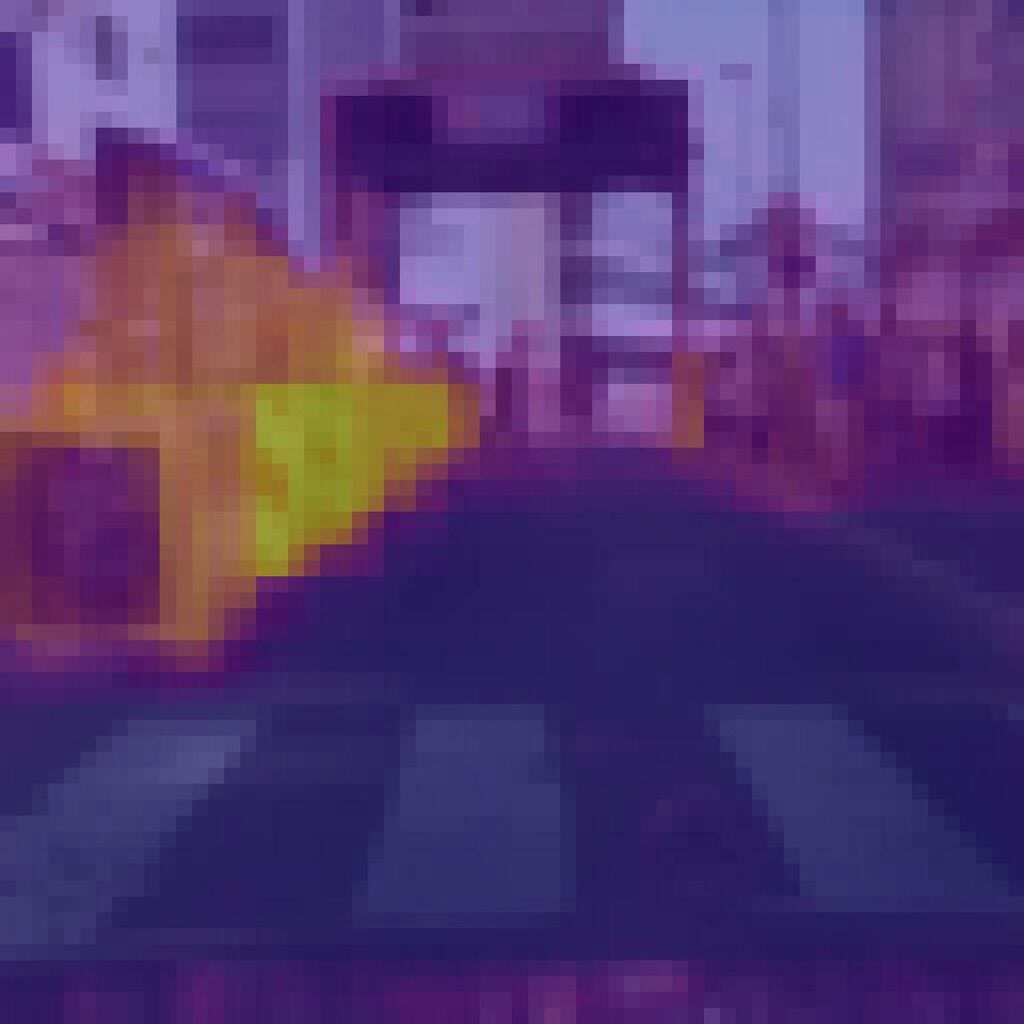}} &
    \fcolorbox{blue}{white}{\includegraphics[width=0.95\linewidth]{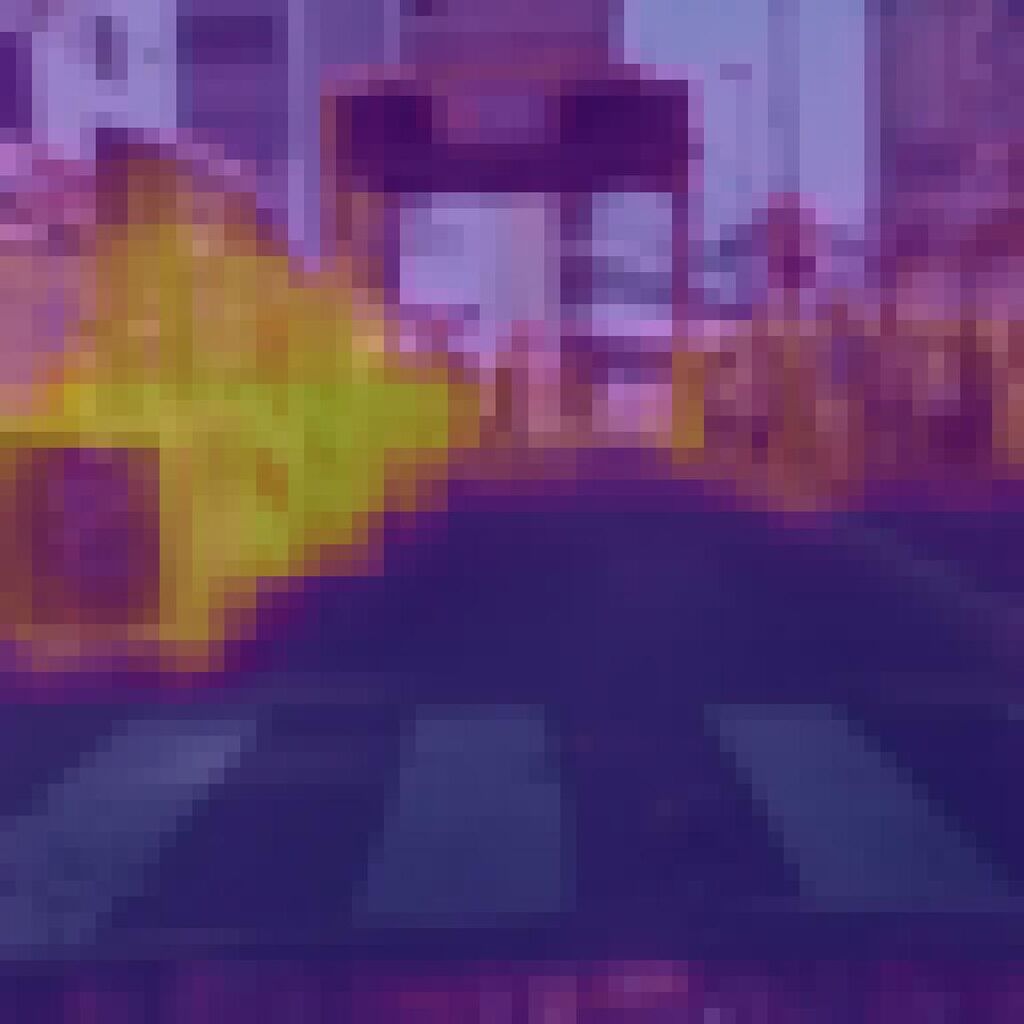}} &
    \fcolorbox{blue}{white}{\includegraphics[width=0.95\linewidth]{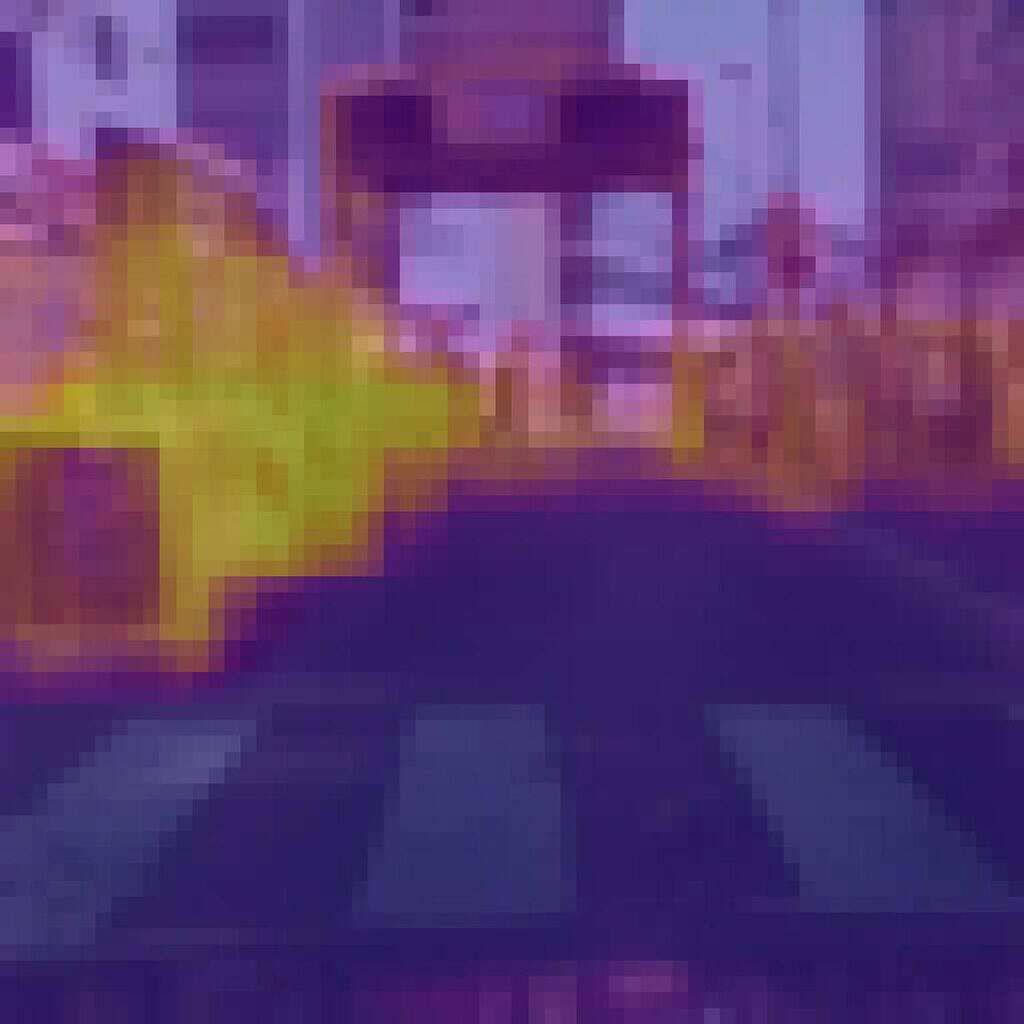}} &
    \fcolorbox{blue}{white}{\includegraphics[width=0.95\linewidth]{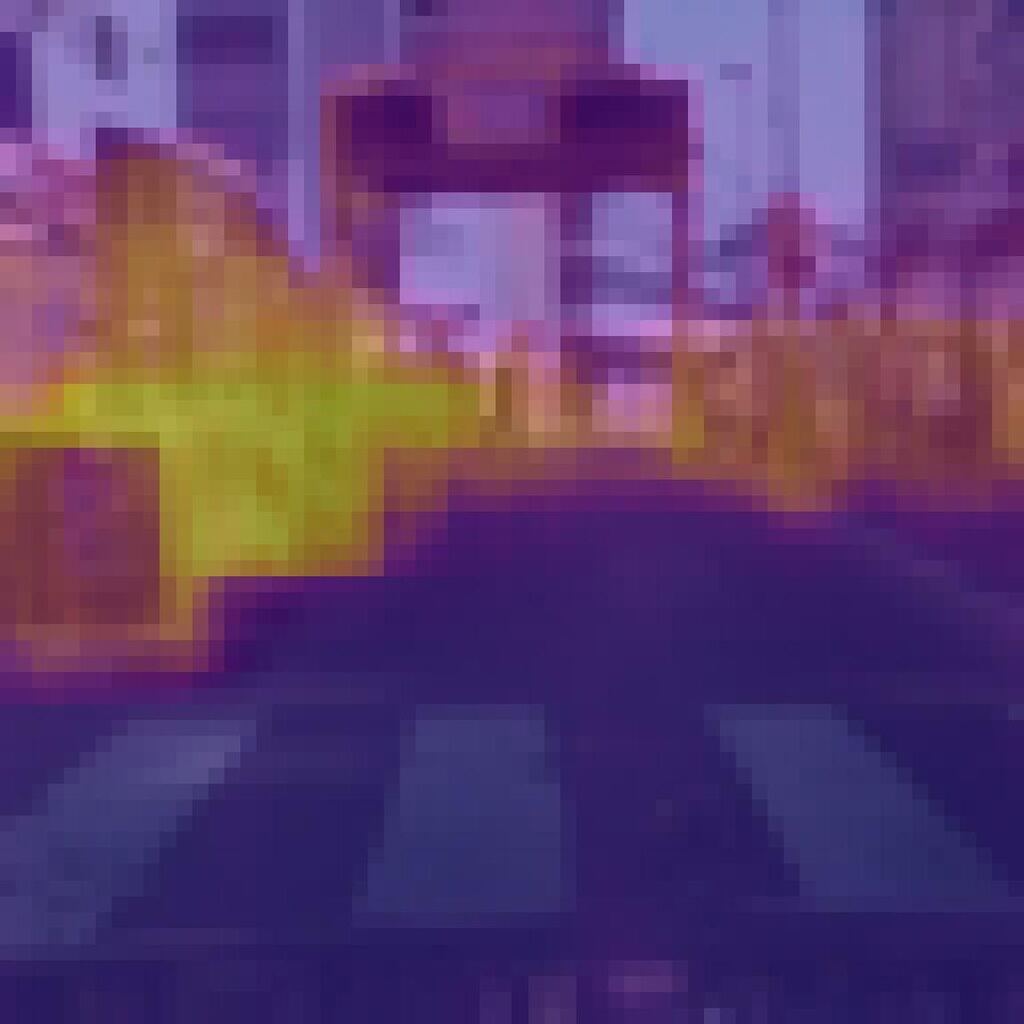}} \\

    t=10 & t=20 & t=30 & t=40 & t=50 & t=100 & t=150 & t=200 & t=250 & t=300 & t=350 \\[1em]
    
    \fcolorbox{blue}{white}{\includegraphics[width=0.95\linewidth]{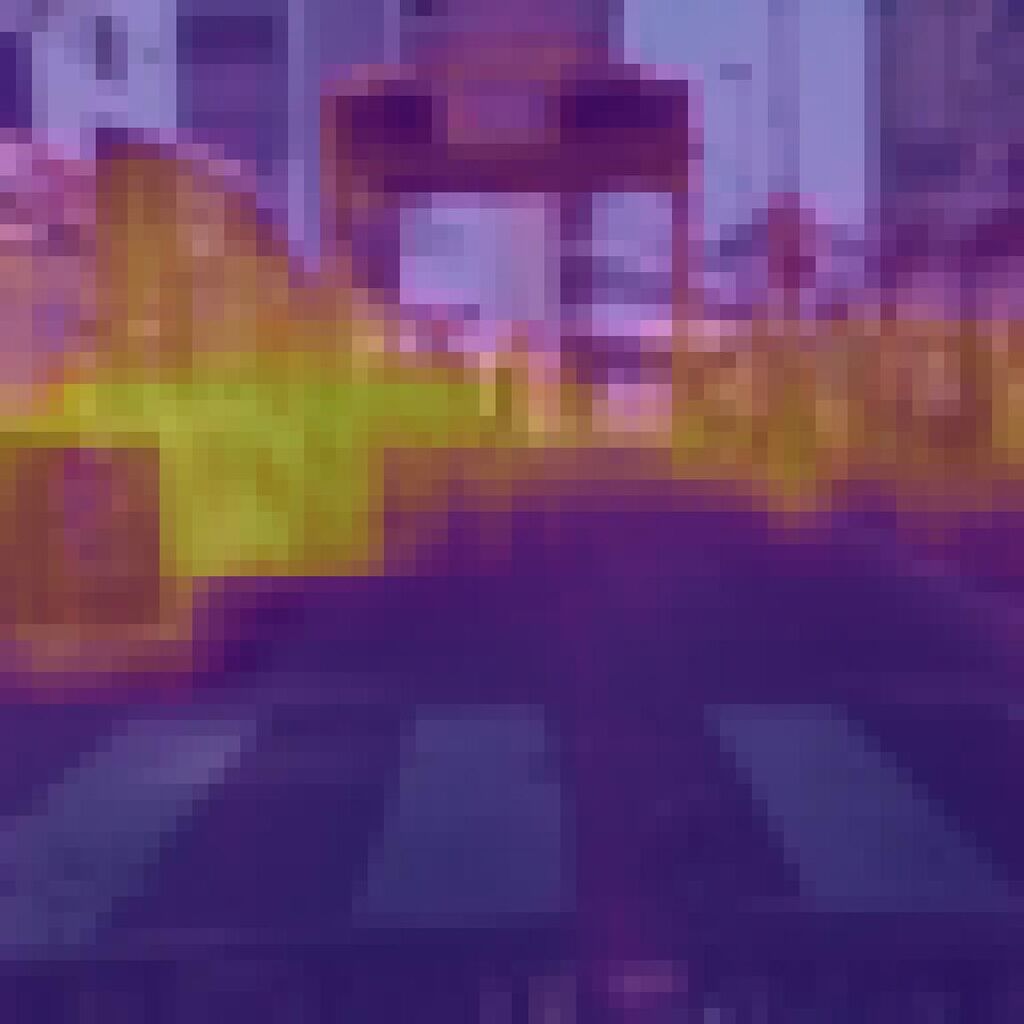}} &
    \fcolorbox{blue}{white}{\includegraphics[width=0.95\linewidth]{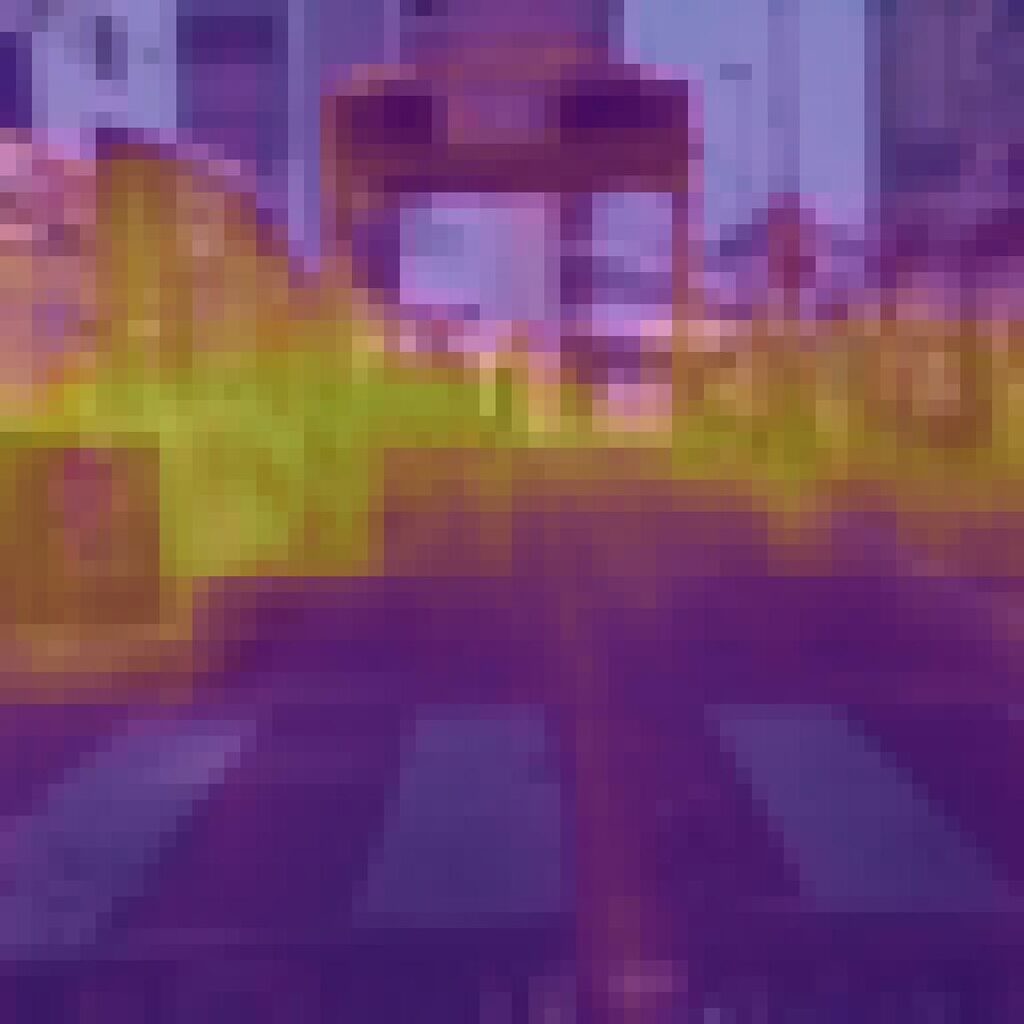}} &
    \fcolorbox{blue}{white}{\includegraphics[width=0.95\linewidth]{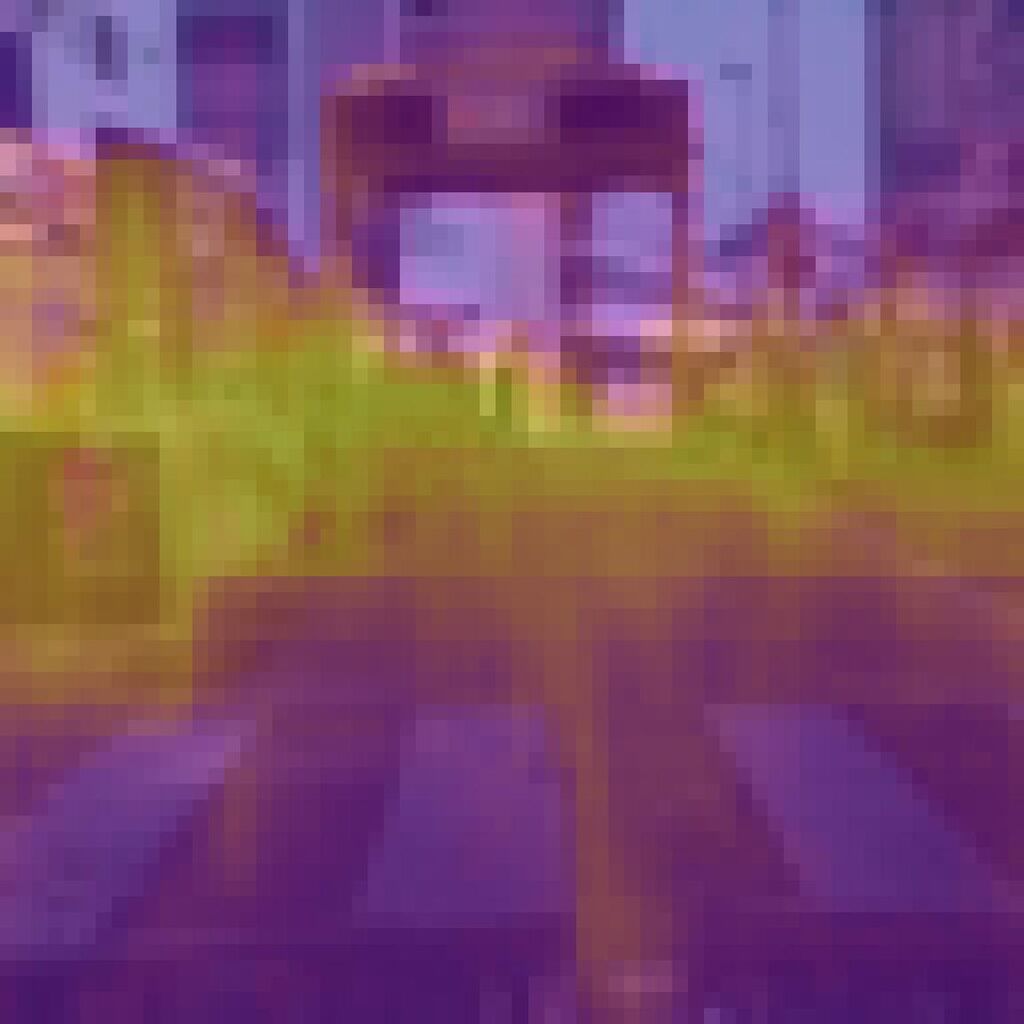}} &
    \fcolorbox{blue}{white}{\includegraphics[width=0.95\linewidth]{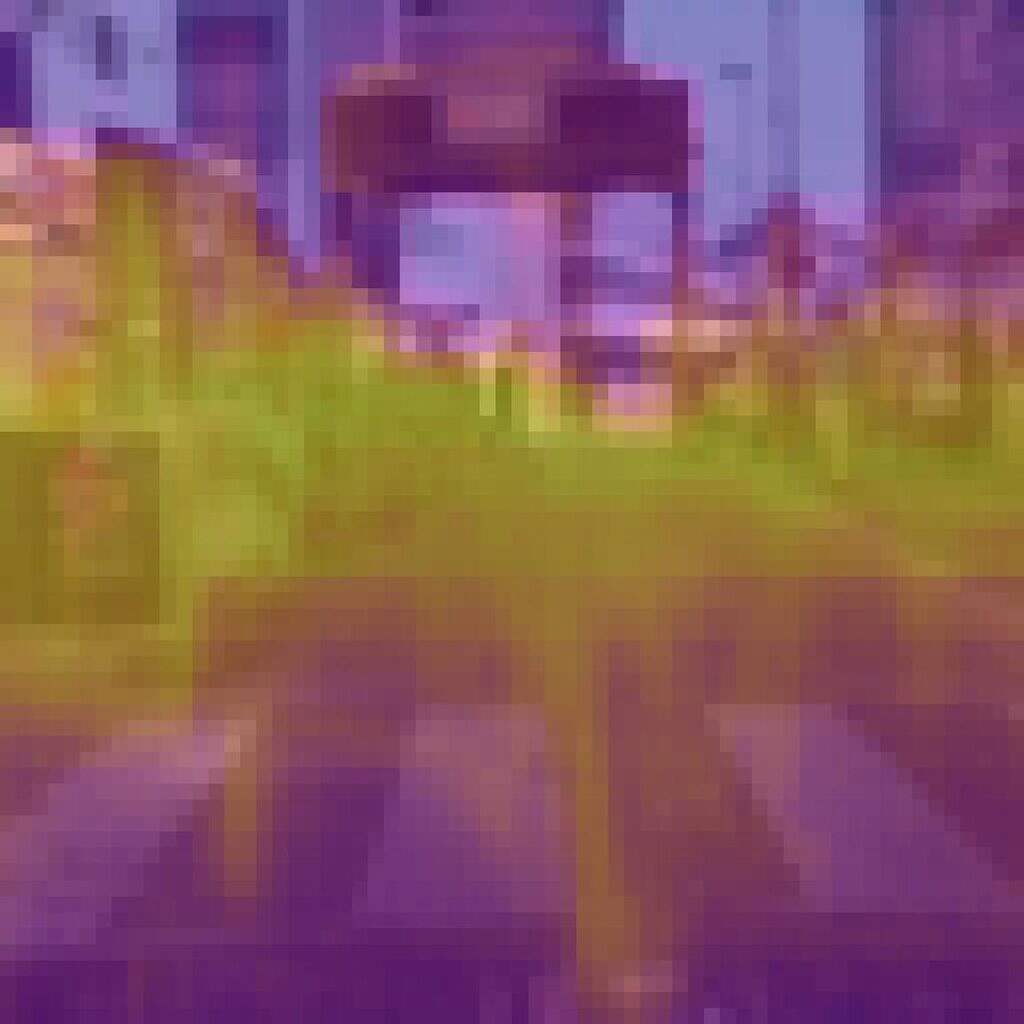}} &
    \fcolorbox{blue}{white}{\includegraphics[width=0.95\linewidth]{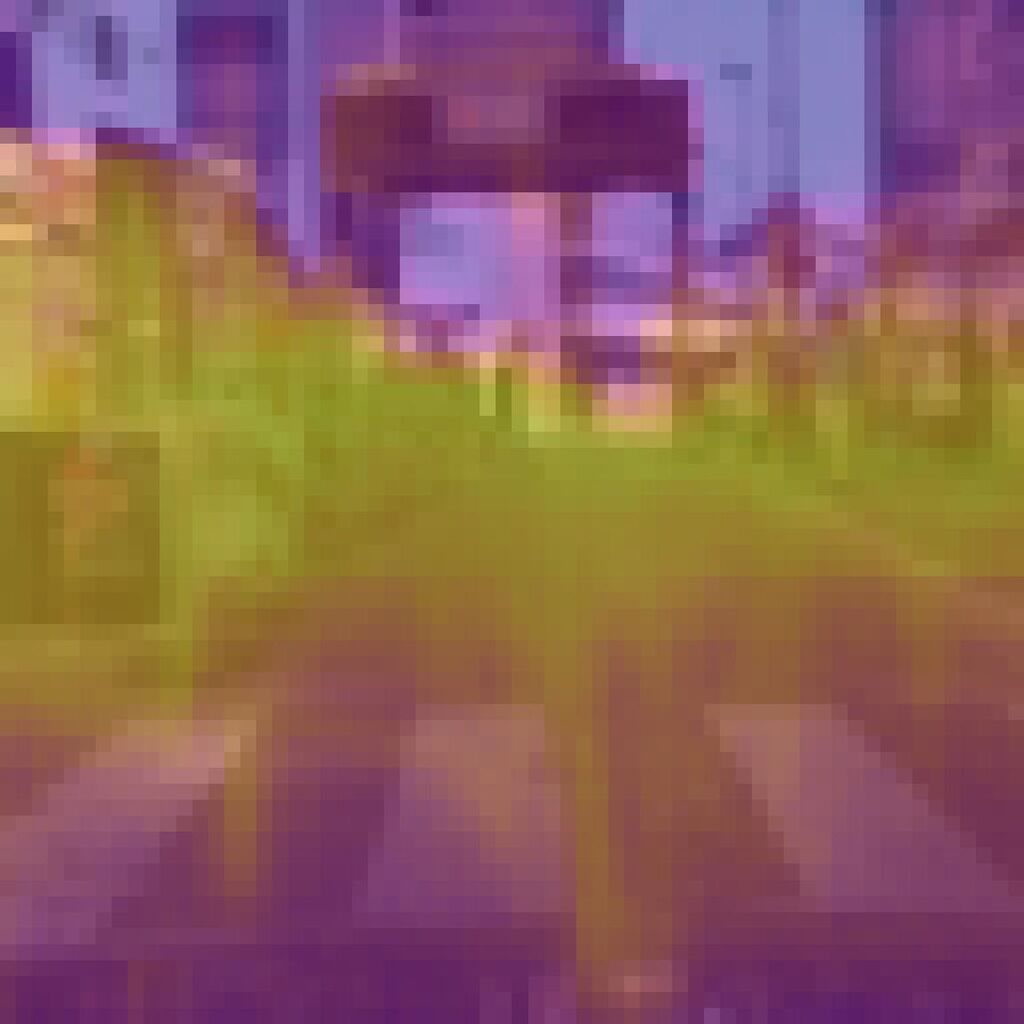}} &
    \fcolorbox{blue}{white}{\includegraphics[width=0.95\linewidth]{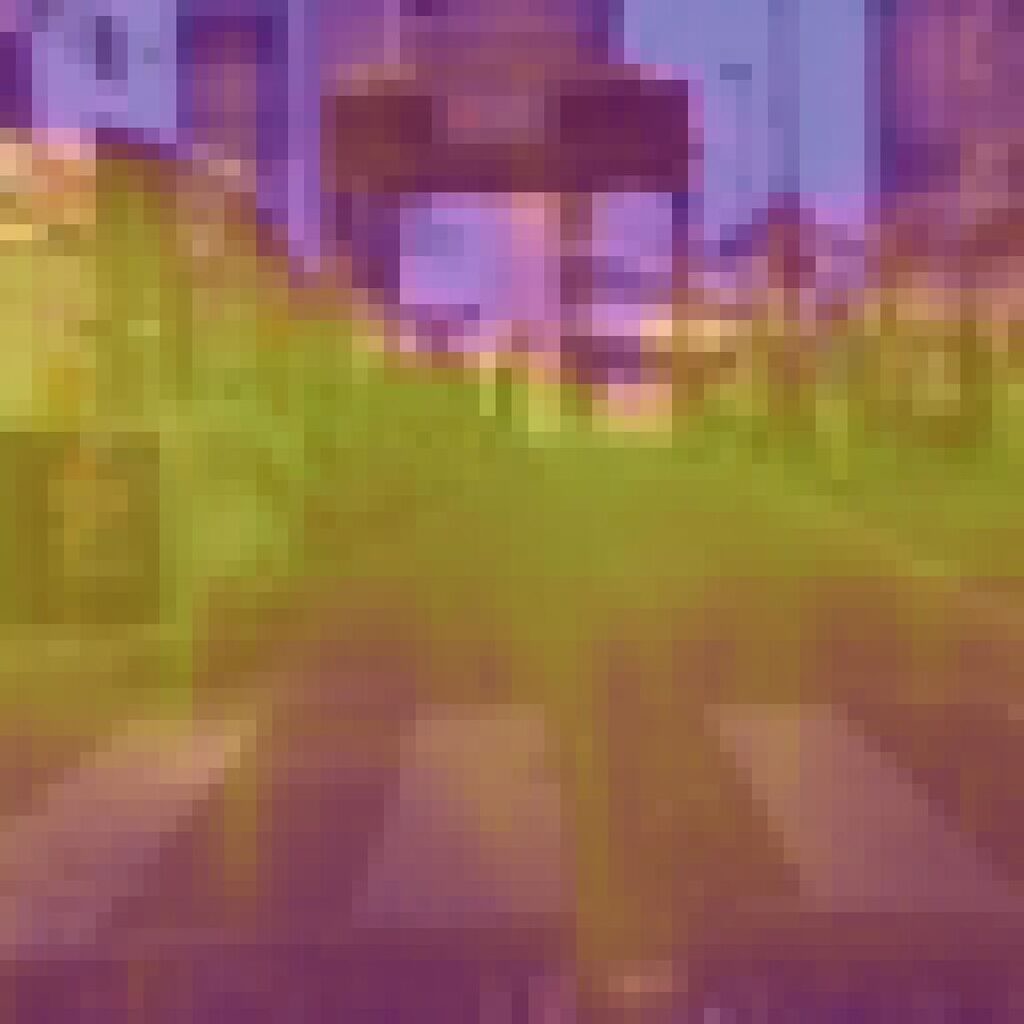}} &
    \fcolorbox{blue}{white}{\includegraphics[width=0.95\linewidth]{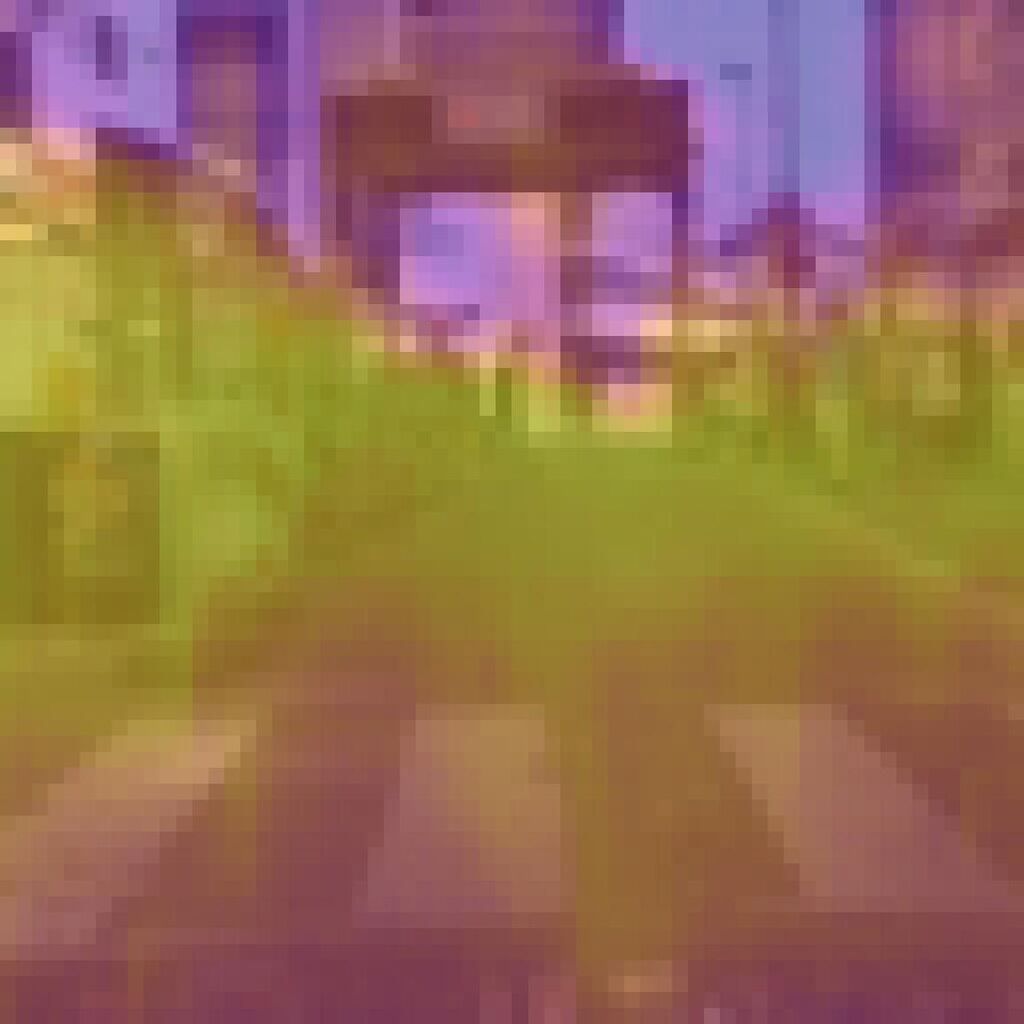}} &
    \fcolorbox{blue}{white}{\includegraphics[width=0.95\linewidth]{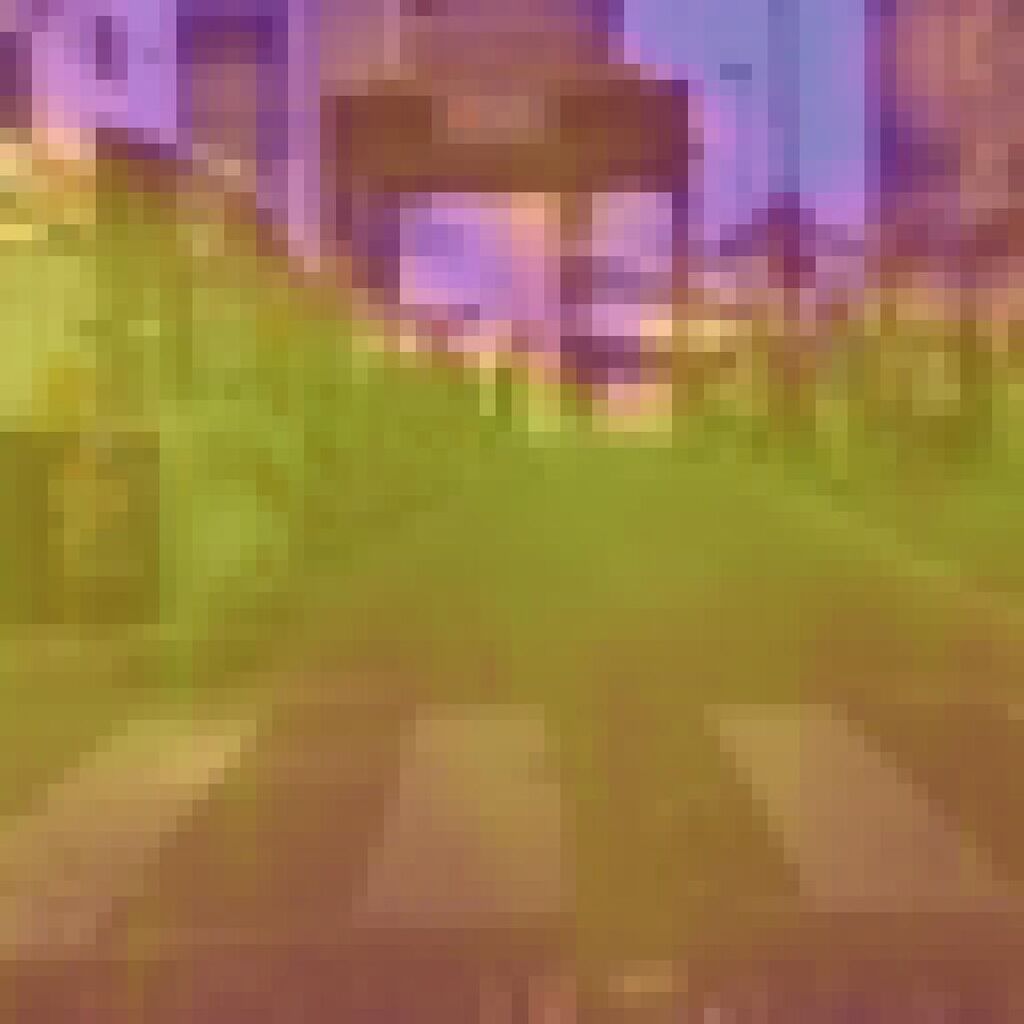}} &
    \fcolorbox{blue}{white}{\includegraphics[width=0.95\linewidth]{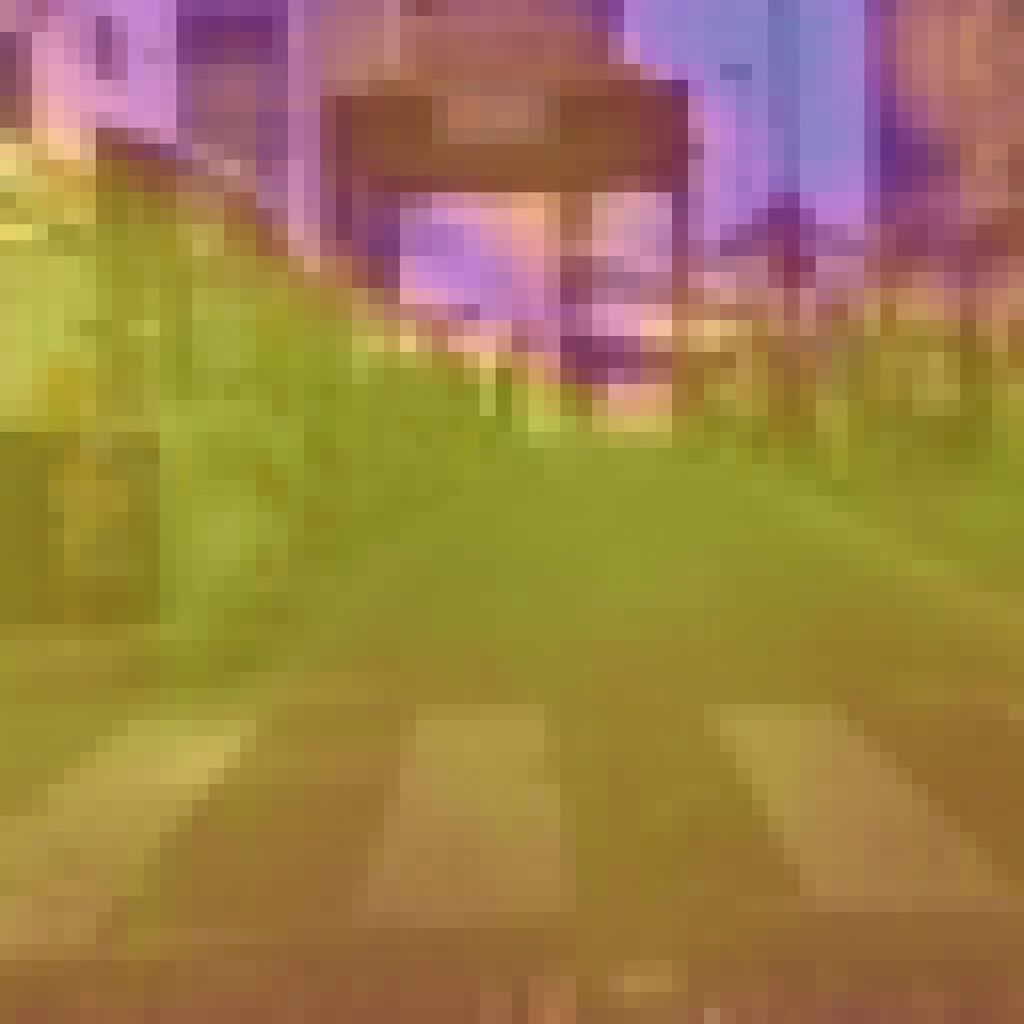}} &
    \fcolorbox{blue}{white}{\includegraphics[width=0.95\linewidth]{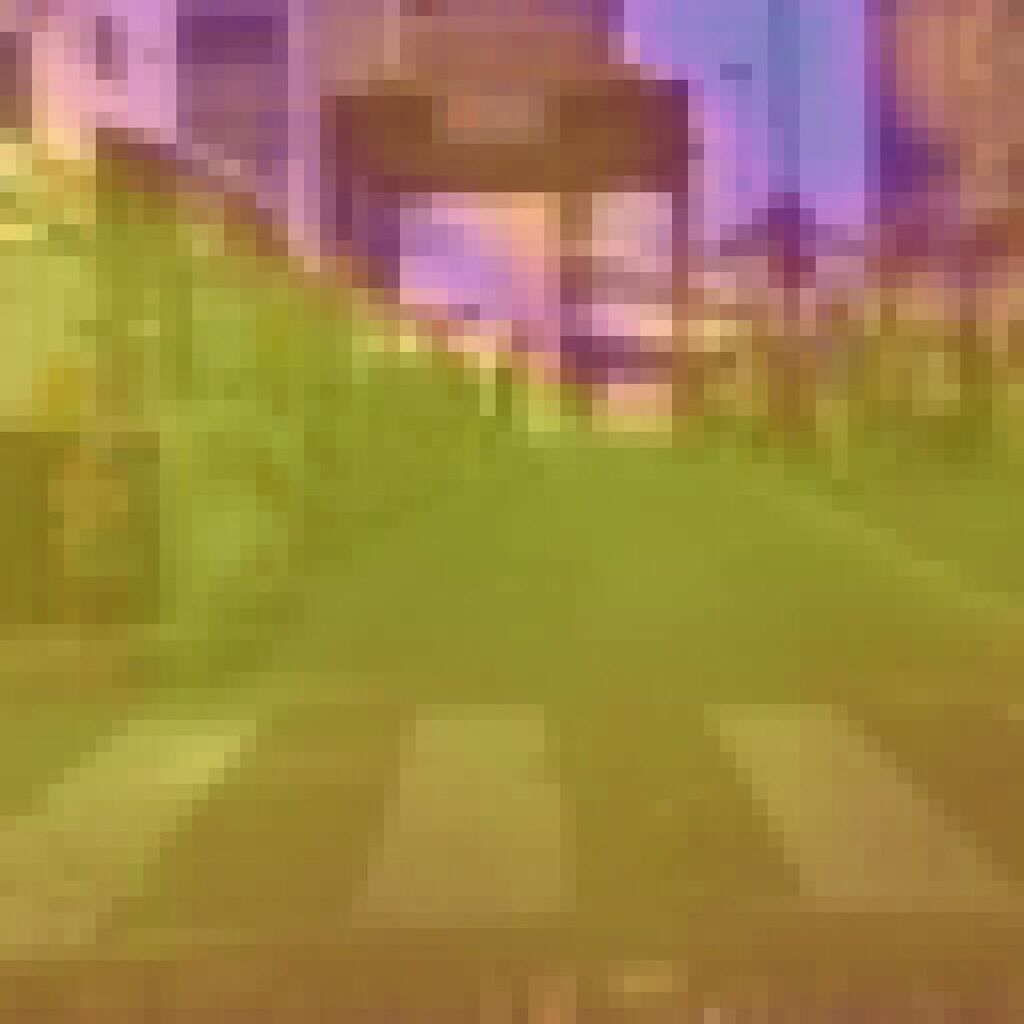}} &
    \fcolorbox{blue}{white}{\includegraphics[width=0.95\linewidth]{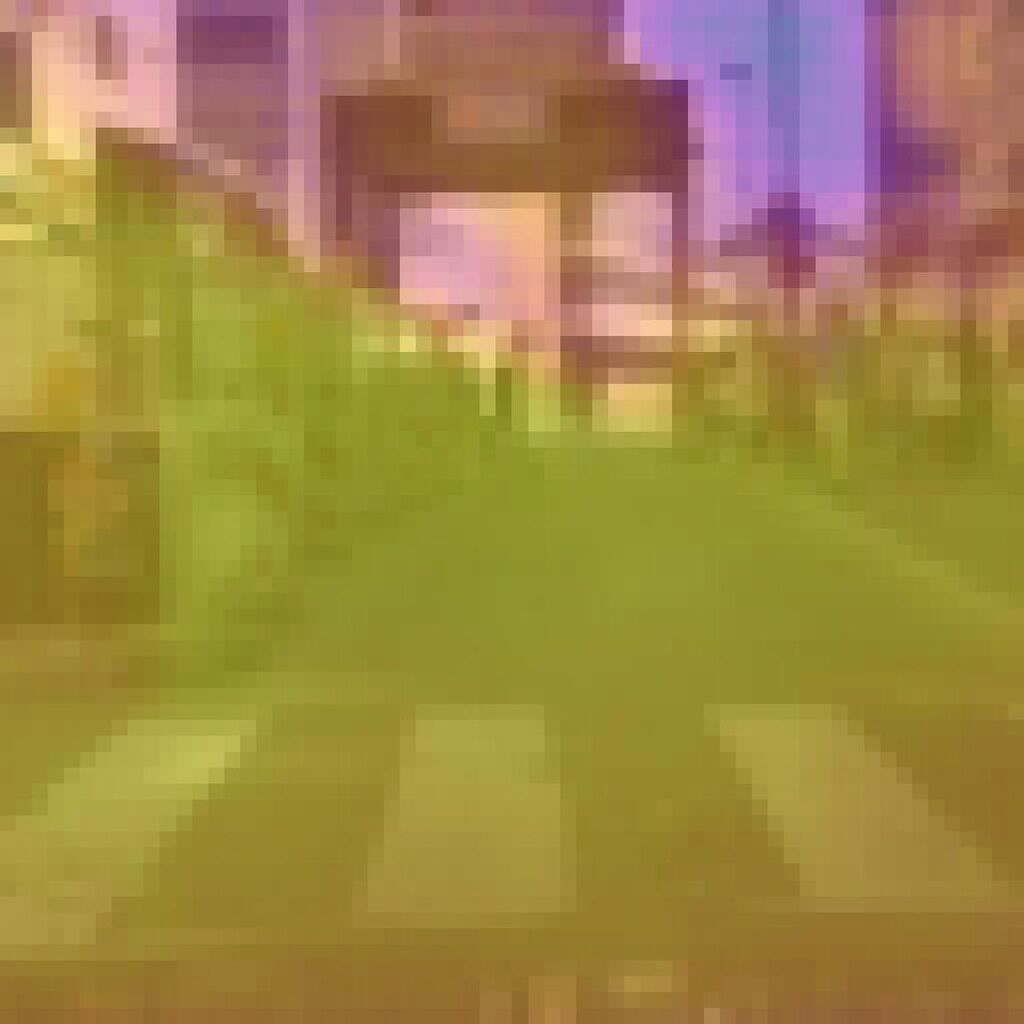}} \\

    t=400 & t=450 & t=500 & t=550 & t=600 & t=650 & t=700 & t=750 & t=800 & t=850 & t=900 \\
    
    \end{tabularx}
    
    \caption{The top row displays the original input image and the corresponding Temporal Stability Matrices (TSMs) for both red and blue query point. The panels below show the evolution of their corresponding Contextual Similarity Maps (CSMs).}

    \label{fig:supp_hierarchical_progress_sample7}
\end{sidewaysfigure*}

\section{Backbone Comparison}\label{sec:supp_backbone_comparison}


We evaluated the performance of our model across several SD backbones with varying architectures. Supp. Tab.~\ref{tab:rebut_results} shows the results. Notably, we see that the model is able to consistently outperform previous baselines in all settings and across all backbones (except for a few cases on COCO-Object dataset and Lumina-Next-SFT model). It is important to note that all experiments use the same hyperparameters, which are selected based on SDv1.4 for a fair comparison with the prior works. It can be seen that SDv2.1 results in the best performance since it has an architecture similar to SDv1.4, demonstrating that hyperparameter optimization can potentially result in better performance for the other backbones.

\begin{table}[h!]
\centering
\small
\caption{Segmentation results (mIoU) using our approach across different Stable Diffusion backbones. Best results are in \textbf{bold}.}
\label{tab:rebut_results}
\resizebox{\linewidth}{!}{
\begin{tabular}{lcccccc}
\toprule
\textbf{Backbone} & \textbf{COCO-Stuff} & \textbf{COCO-Object} & \textbf{CityScapes} & \textbf{ADE20K} & \textbf{VOC} & \textbf{Context} \\
\midrule
Pixart-Alpha (DiT) \cite{chen2023pixart} &  45.6 & {30.3} & 24.8 & 41.6 & 54.0 & 52.1 \\
Lumina-Next-SFT (DiT + Flow) \cite{gao2024lumina} &  {41.7} & {27.8} & {24.0} & 40.4 & {50.8} & 49.9 \\
SDv1.4 \cite{rombach2022latent} & 45.0 & {30.8} & 27.8 & 45.3 & 58.2 & 55.4 \\
SSD-1B \cite{gupta2024progressive} & 48.3 & {26.8} & 24.5 & 40.3 & 58.6 & 53.4 \\
SDXL \cite{podell2023sdxl} & 47.4 & {25.8} & 24.2 & 39.6 & 55.0 & 51.0 \\
SSD-Vega \cite{gupta2024progressive} & 45.9 & {27.3} & 25.6 & 43.2 & 54.8 & 51.8 \\
SDv2.1 \cite{rombach2022latent} & \textbf{49.7} & \textbf{31.1} & \textbf{28.4} & \textbf{46.7} & \textbf{61.7} & \textbf{58.6} \\
\bottomrule
\end{tabular}
}
\end{table}

\section{Computation Analysis}
\label{sec:supp_computation_analysis}

While the goal of this work is to demonstrate the emergence of hierarchical semantics during denoising, rather than efficient deployment, we performed several experiments to analyze the computational requirements of the model. Supp. Tab.~\ref{tab:rebut_compute} shows the time and computation analysis of our model against DiffSeg and DiffCut. It is important to note that while our model generates the CSM for several timesteps, the CSM for each timestep is independent and can be computed in-parallel.

\begin{table}[h!]
\centering
\small
\caption{Compute analysis on SDv1.4. $N$ is the number of pixels, and $T$ is the number of timesteps.}
\label{tab:rebut_compute}
\resizebox{\linewidth}{!}{
\begin{tabular}{lcccc}
\toprule
\textbf{Method} & \textbf{Time Complexity} & \textbf{Space Complexity} & \textbf{Latency (s)} & \textbf{VRAM (GB)} \\
\midrule
DiffCut (PyTorch) & $O(N^3)$ & $O(N^2)$ & $5.2$ & $4$ \\
DiffSeg (TensorFlow) & $O(N^2)$ & $O(N^2)$ & $4.7$ & $8.5$ \\
Ours (PyTorch) (limit VRAM) & $O(TN^2 + N^3)$ & $O(TN^2)$ & $45.6$ & $7.2$ \\
Ours (PyTorch) & $O(TN^2 + N^3)$ & $O(TN^2)$ & $15.2$ & $21.7$ \\
\bottomrule
\end{tabular}
}
\end{table}

We also evaluated the performance vs. time trade-off of the model by selecting different numbers of timesteps. Supp. Tab.~\ref{tab:rebut_time_acc_tradeoff} confirms that larger number of timesteps lead to better results due to more flexibility in adaptive timestep selection, while achieving similar performance with 11 timesteps and half of the computation time of the full 22 timesteps.

\begin{table}[h!]
\centering
\small
\caption{Time-Accuracy Trade-off on Pascal VOC. All experiments use a single seed. }
\label{tab:rebut_time_acc_tradeoff}
\begin{tabular}{lccccc}
\toprule
\textbf{\# Timesteps} & 22 & 15 & 11 & 8 & 3 \\
\midrule
\textbf{Latency} &  $15.2s$ & $10.2$s & $8.1$s & $6.5$s & $3.5$s \\
\textbf{mIoU} & $57.8$ & $57.0$ & $56.1$ & $54.4$ & $51.1$ \\
\bottomrule
\end{tabular}
\end{table}




\begin{figure*}[h]
    \centering
    \begin{subfigure}[b]{0.165\textwidth}
        \centering
        \includegraphics[width=\textwidth]{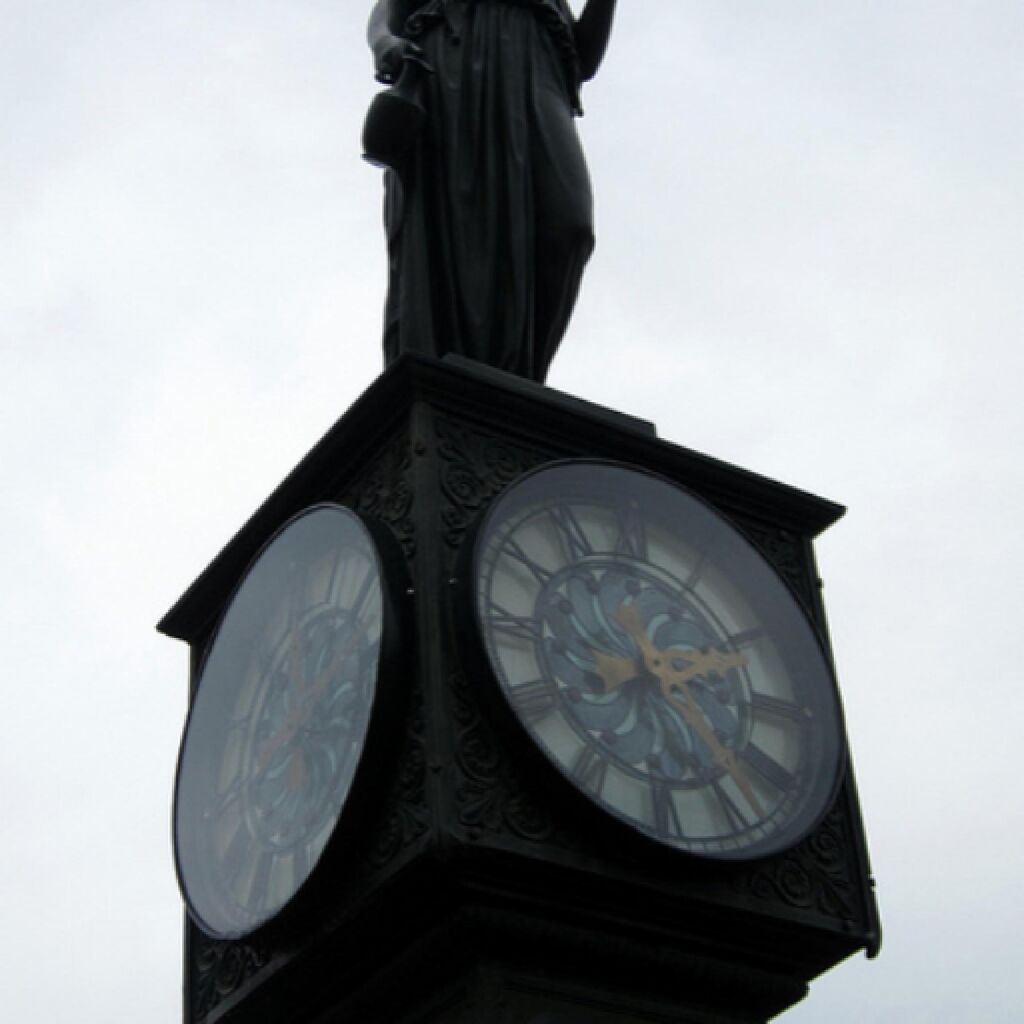}
    \end{subfigure}%
    \begin{subfigure}[b]{0.165\textwidth}
        \centering
        \includegraphics[width=\textwidth]{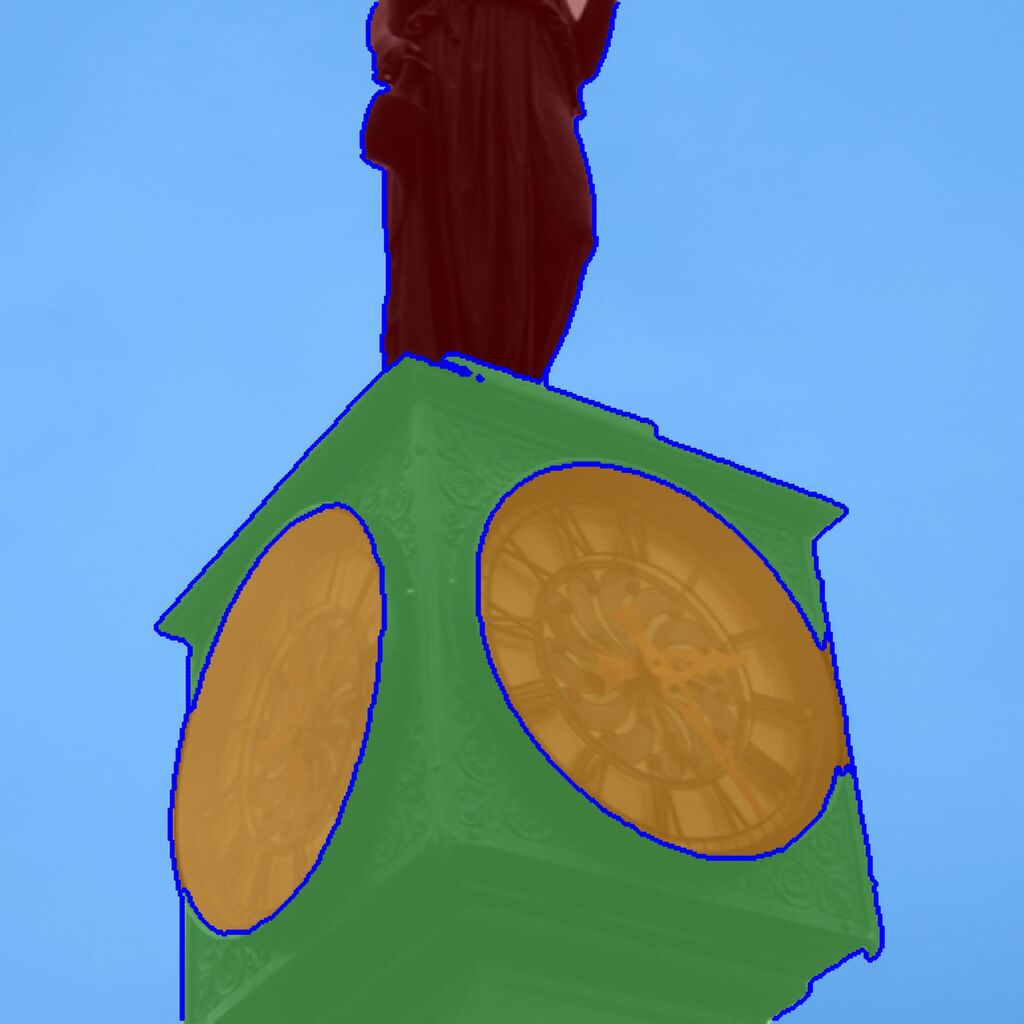}
    \end{subfigure}%
    \begin{subfigure}[b]{0.165\textwidth}
        \centering
        \includegraphics[width=\textwidth]{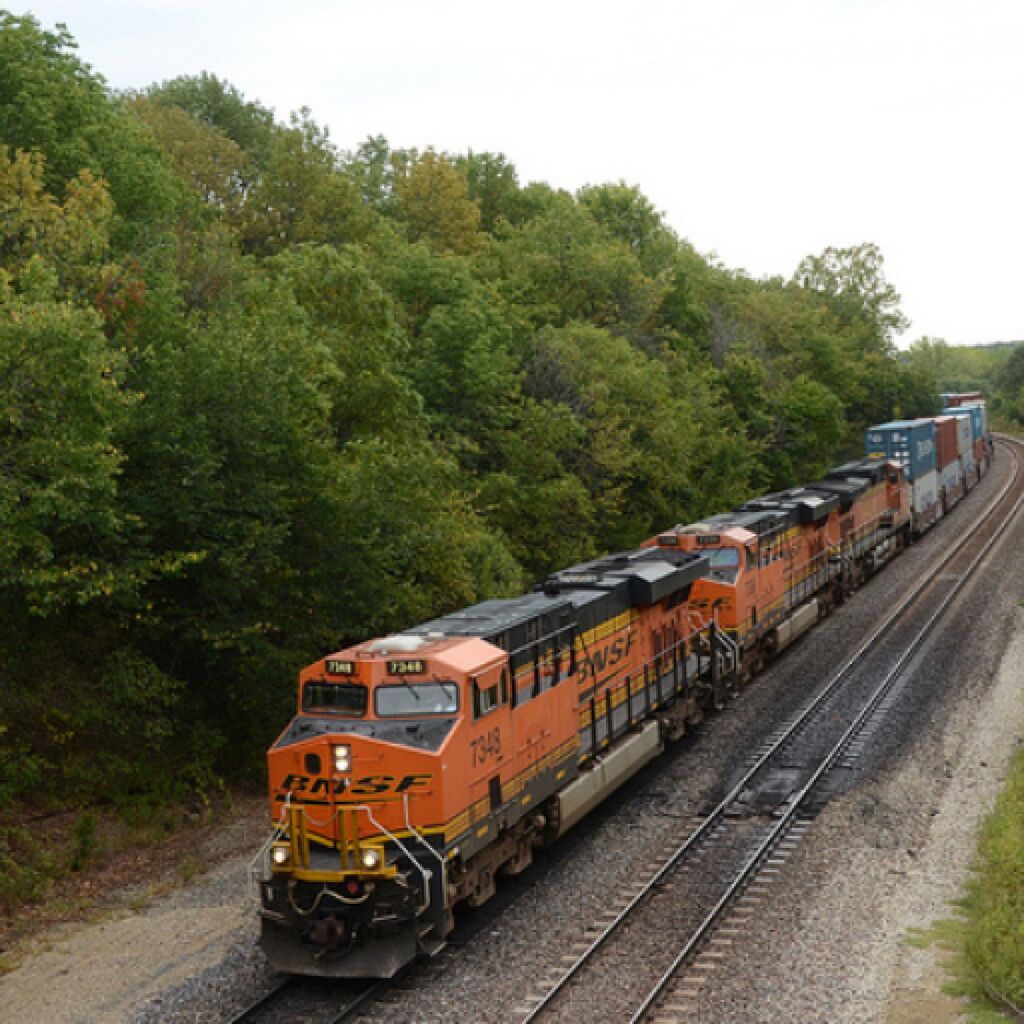}
    \end{subfigure}%
    \begin{subfigure}[b]{0.165\textwidth}
        \centering
        \includegraphics[width=\textwidth]{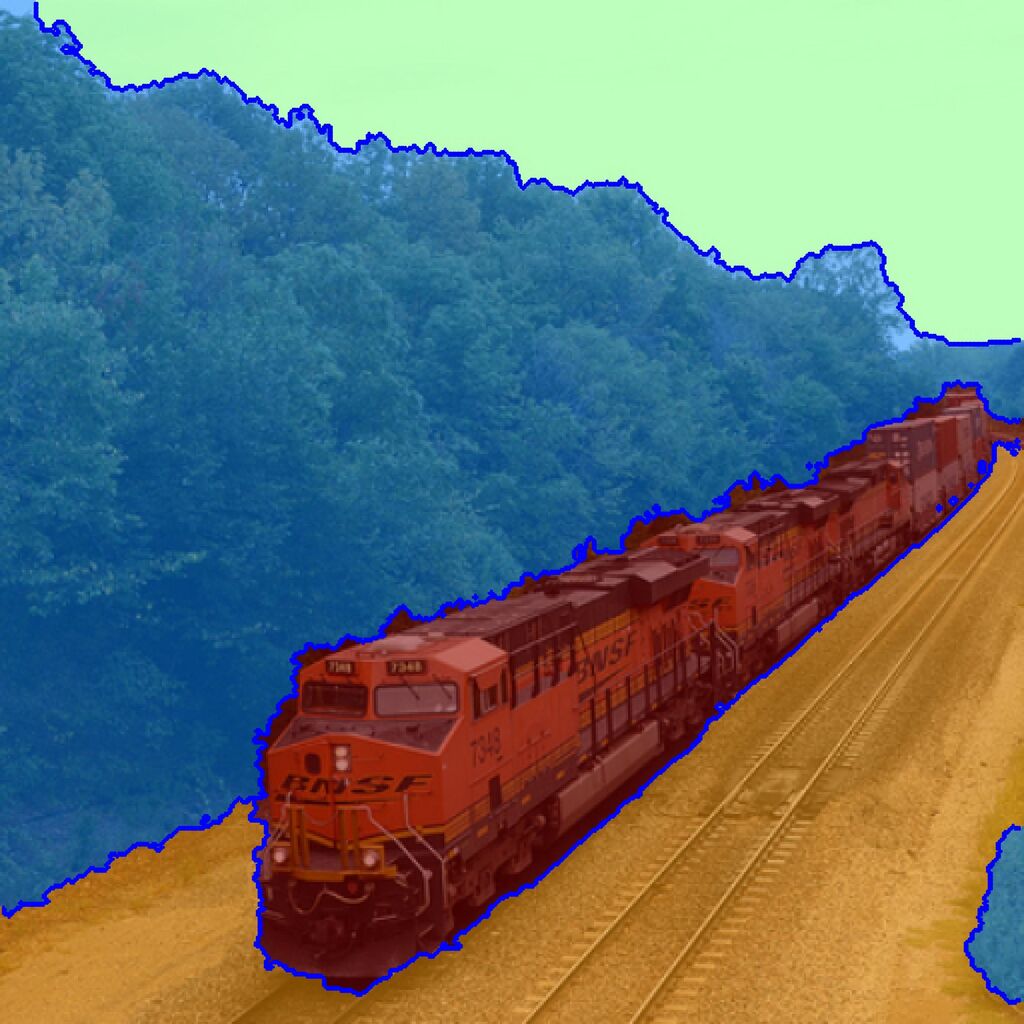}
    \end{subfigure}%
    \begin{subfigure}[b]{0.165\textwidth}
        \centering
        \includegraphics[width=\textwidth]{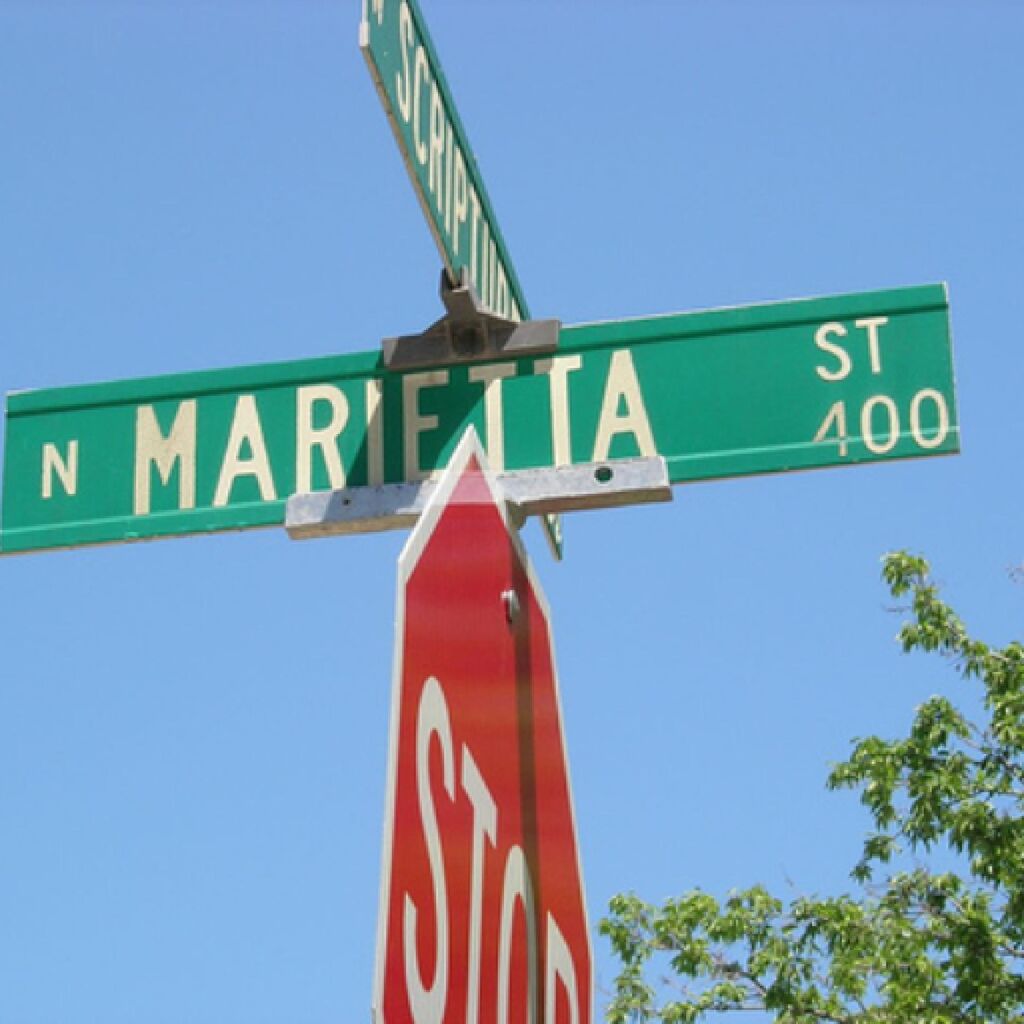}
    \end{subfigure}%
    \begin{subfigure}[b]{0.165\textwidth}
        \centering
        \includegraphics[width=\textwidth]{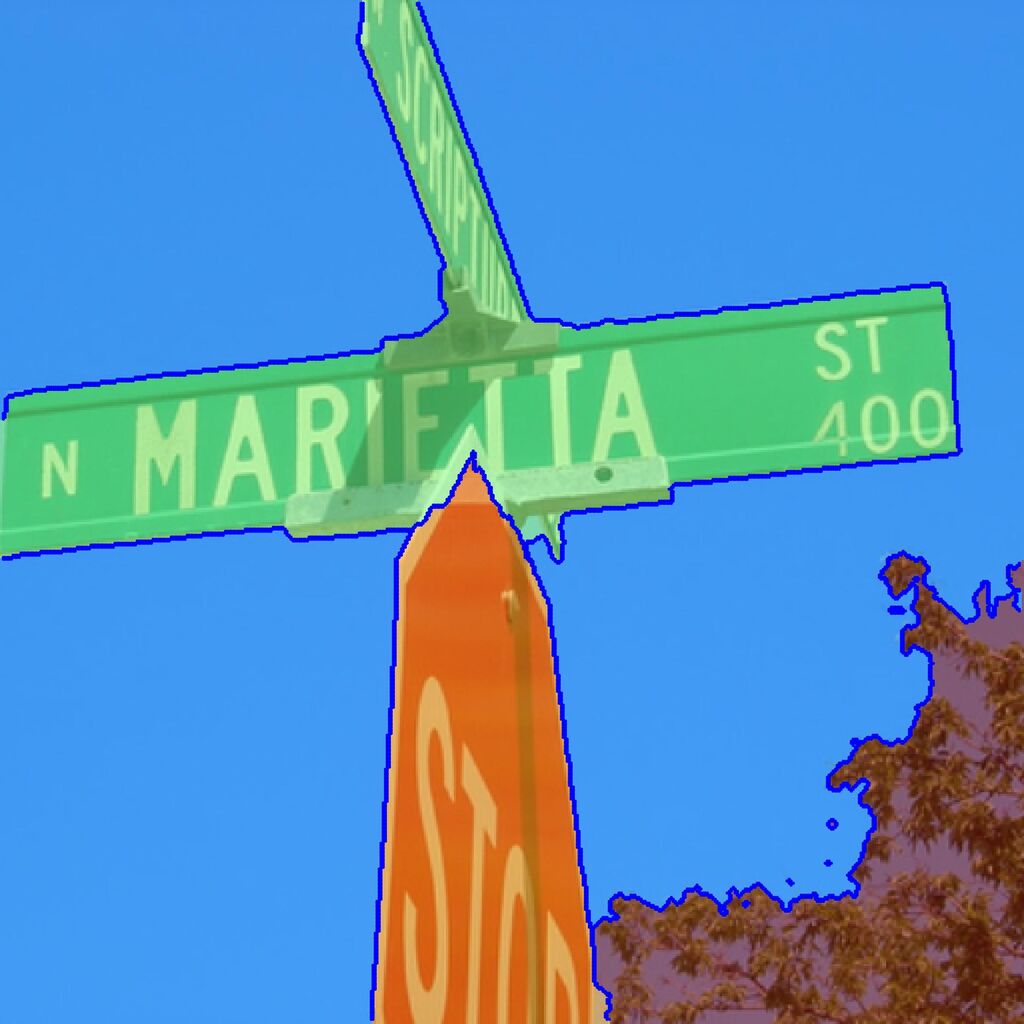}
    \end{subfigure}
    
    \vspace{-1em}
    
    \begin{subfigure}[b]{0.165\textwidth}
        \centering
        \includegraphics[width=\textwidth]{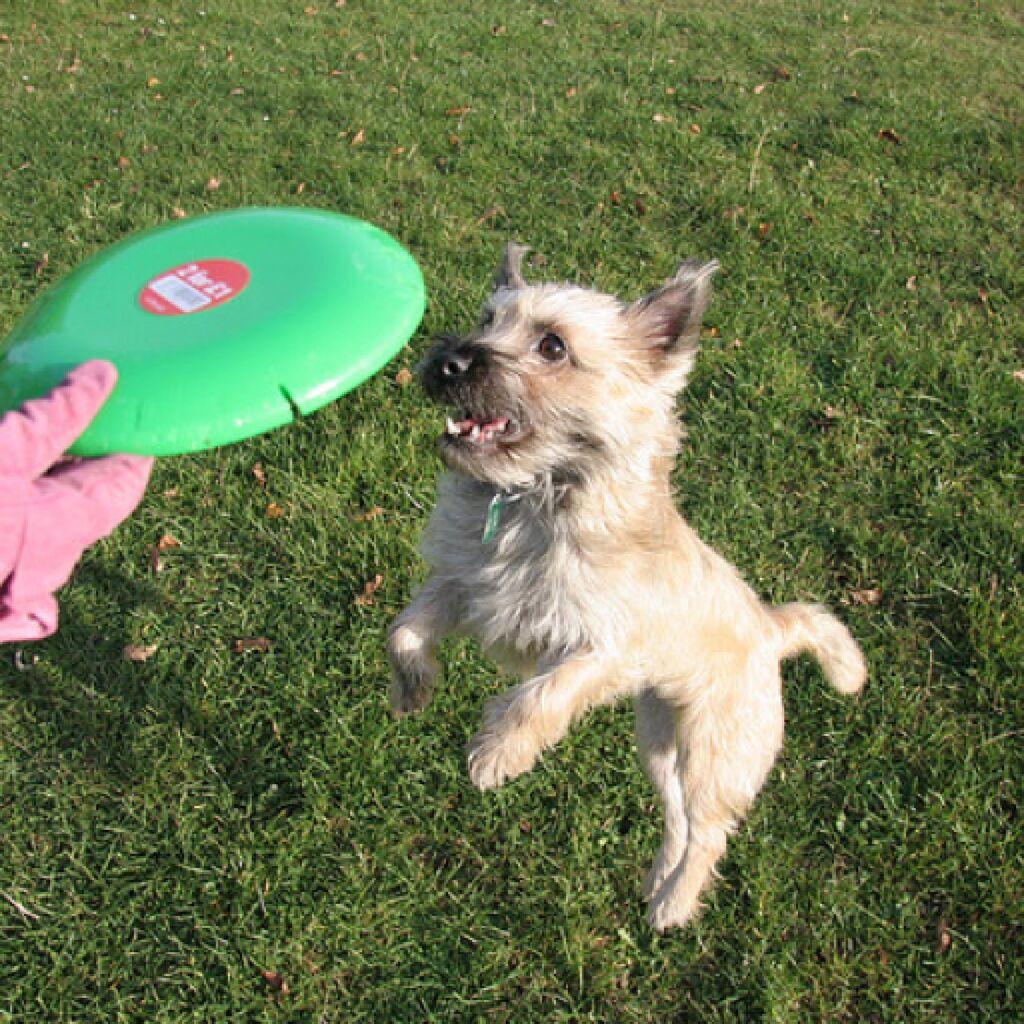}
    \end{subfigure}%
    \begin{subfigure}[b]{0.165\textwidth}
        \centering
        \includegraphics[width=\textwidth]{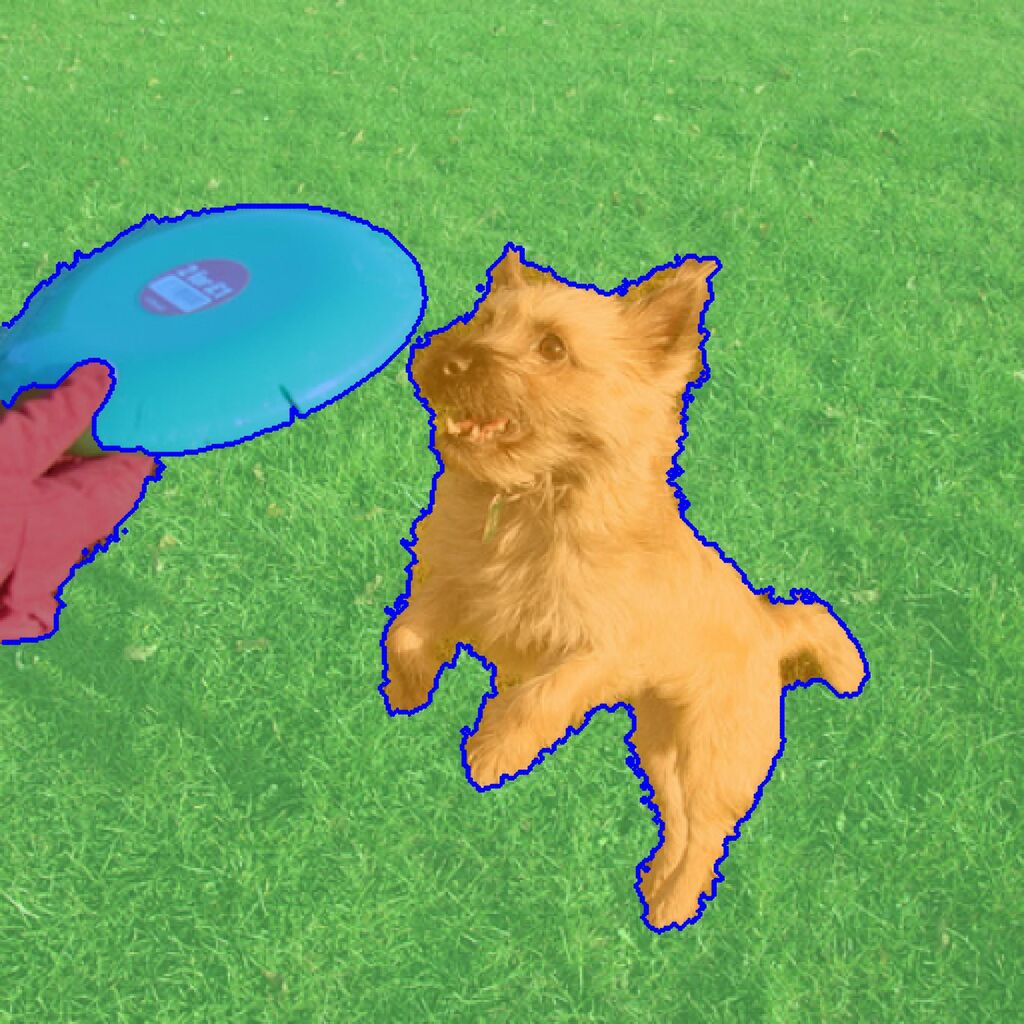}
    \end{subfigure}%
    \begin{subfigure}[b]{0.165\textwidth}
        \centering
        \includegraphics[width=\textwidth]{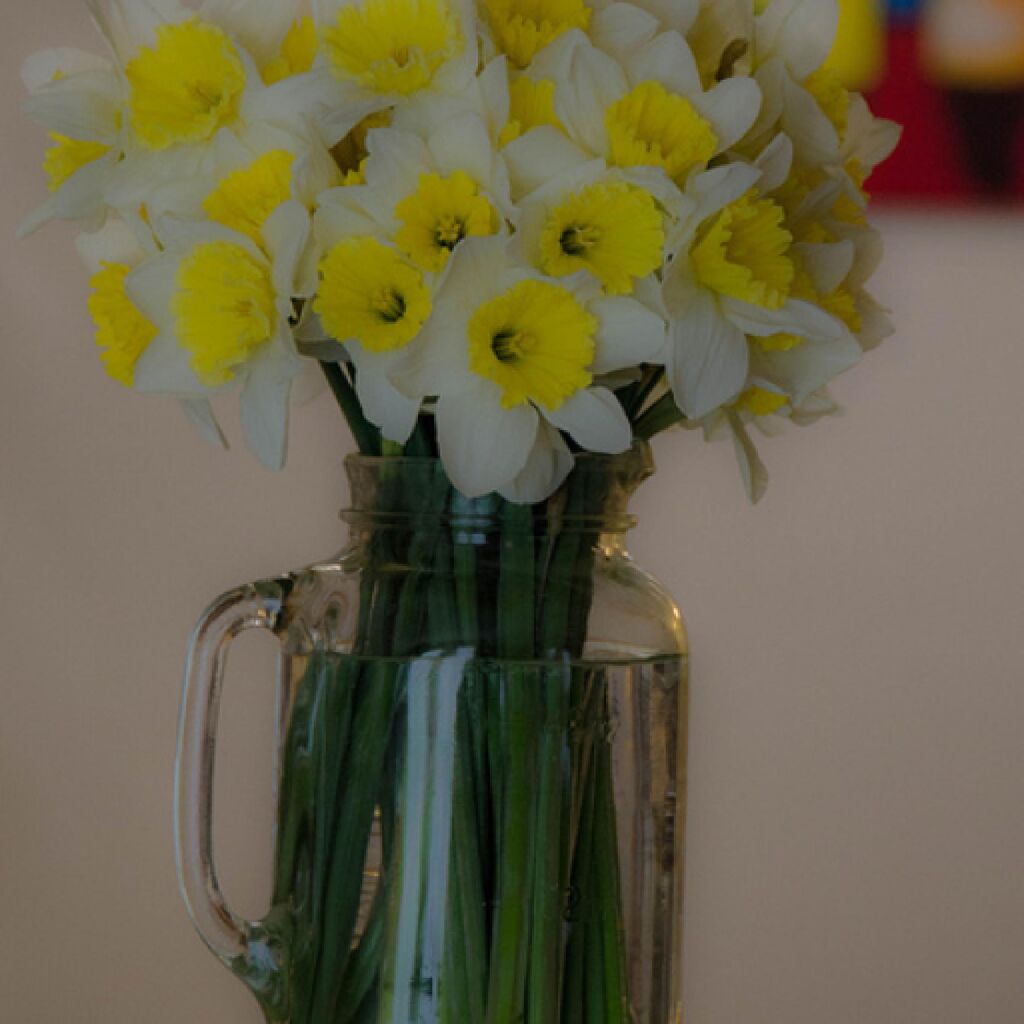}
    \end{subfigure}%
    \begin{subfigure}[b]{0.165\textwidth}
        \centering
        \includegraphics[width=\textwidth]{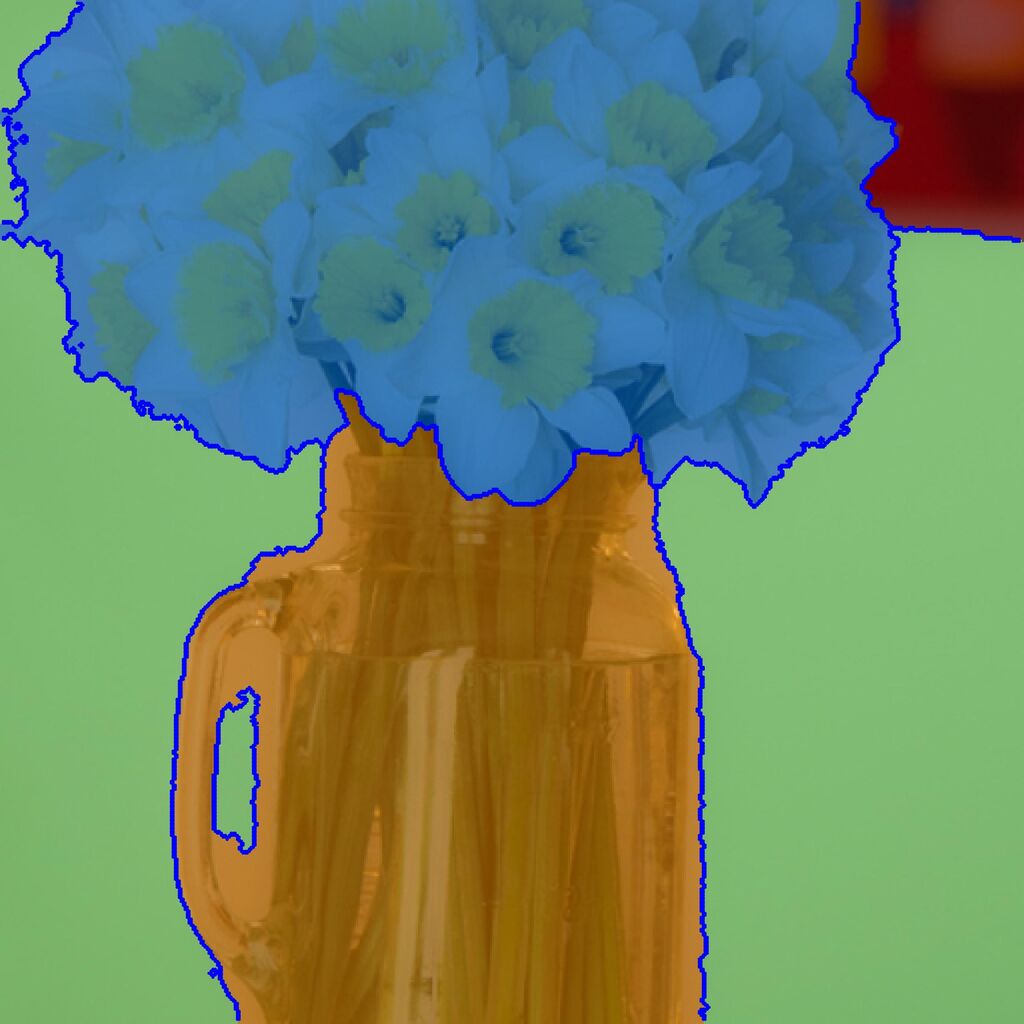}
    \end{subfigure}%
    \begin{subfigure}[b]{0.165\textwidth}
        \centering
        \includegraphics[width=\textwidth]{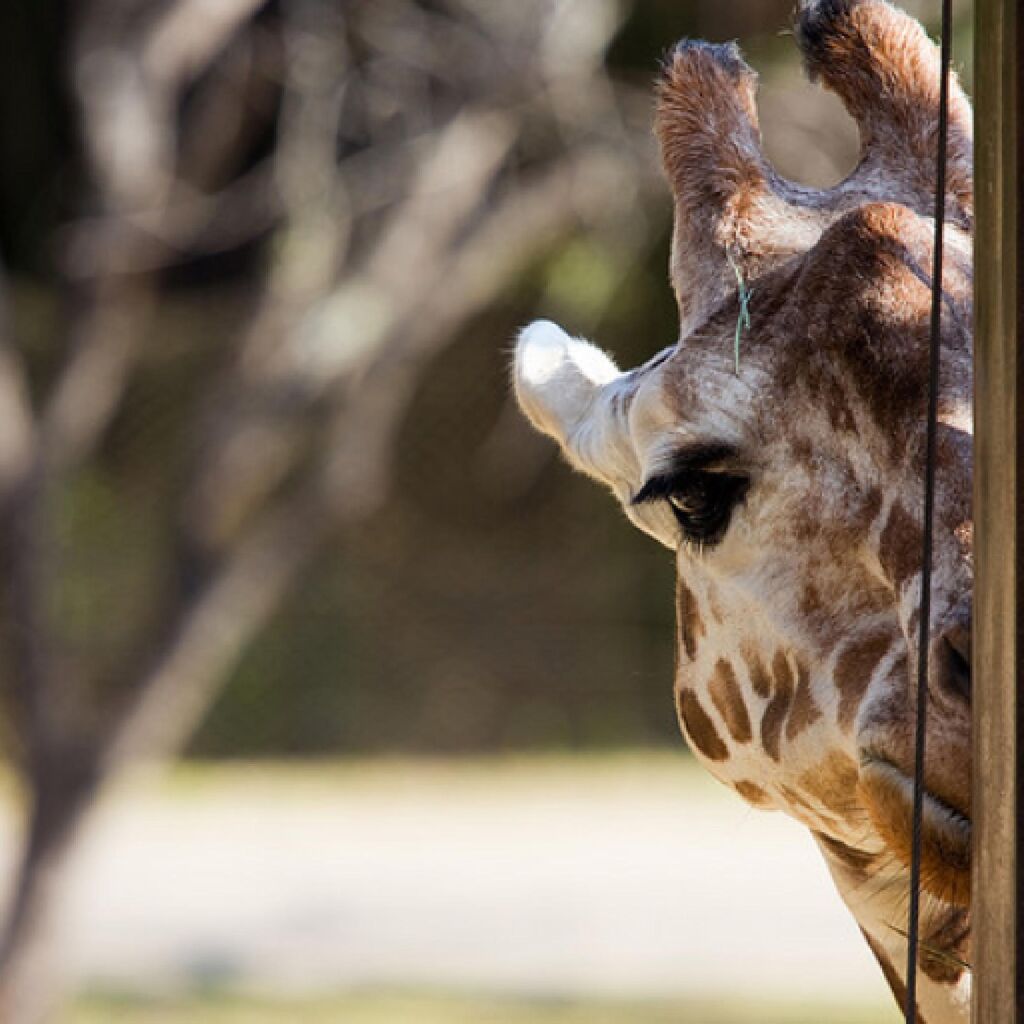}
    \end{subfigure}%
    \begin{subfigure}[b]{0.165\textwidth}
        \centering
        \includegraphics[width=\textwidth]{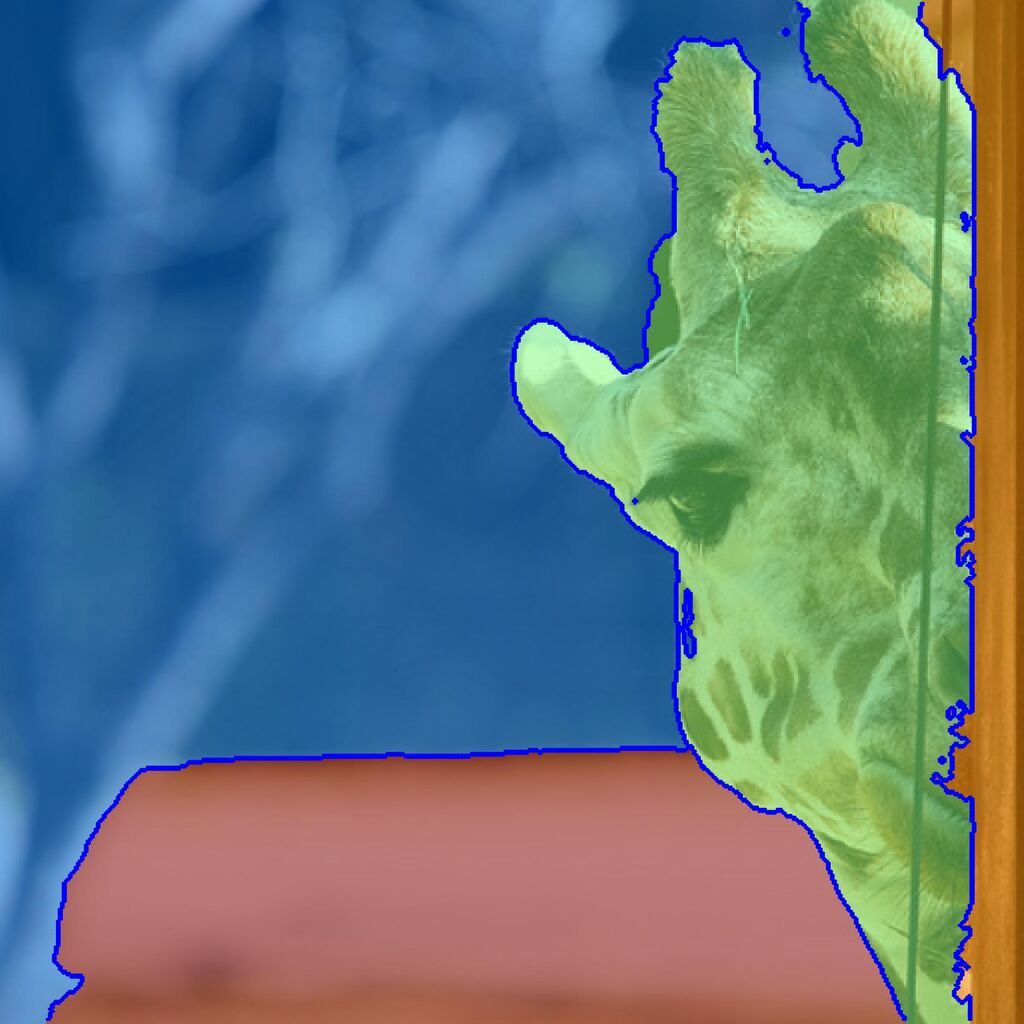}
    \end{subfigure}

    \vspace{-1em}

    \begin{subfigure}[b]{0.165\textwidth}
        \centering
        \includegraphics[width=\textwidth]{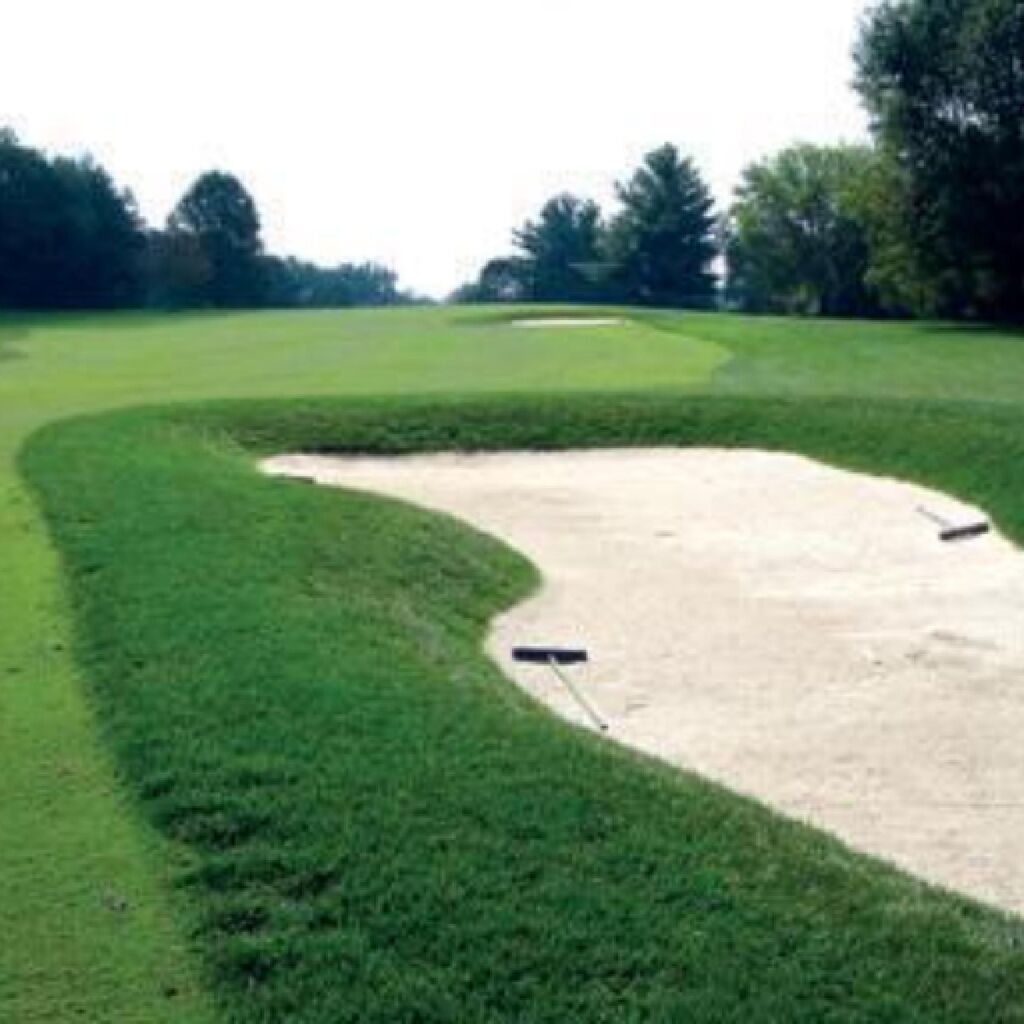}
    \end{subfigure}%
    \begin{subfigure}[b]{0.165\textwidth}
        \centering
        \includegraphics[width=\textwidth]{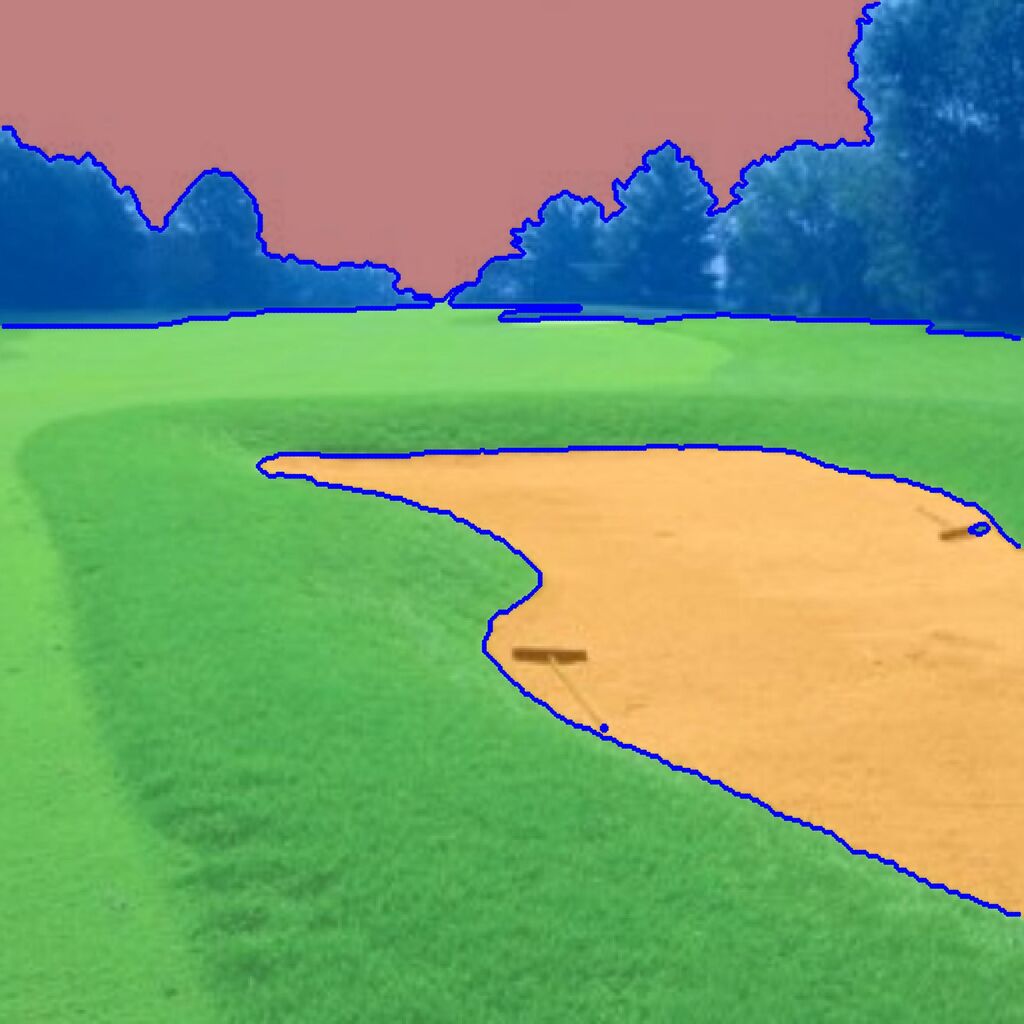}
    \end{subfigure}%
    \begin{subfigure}[b]{0.165\textwidth}
        \centering
        \includegraphics[width=\textwidth]{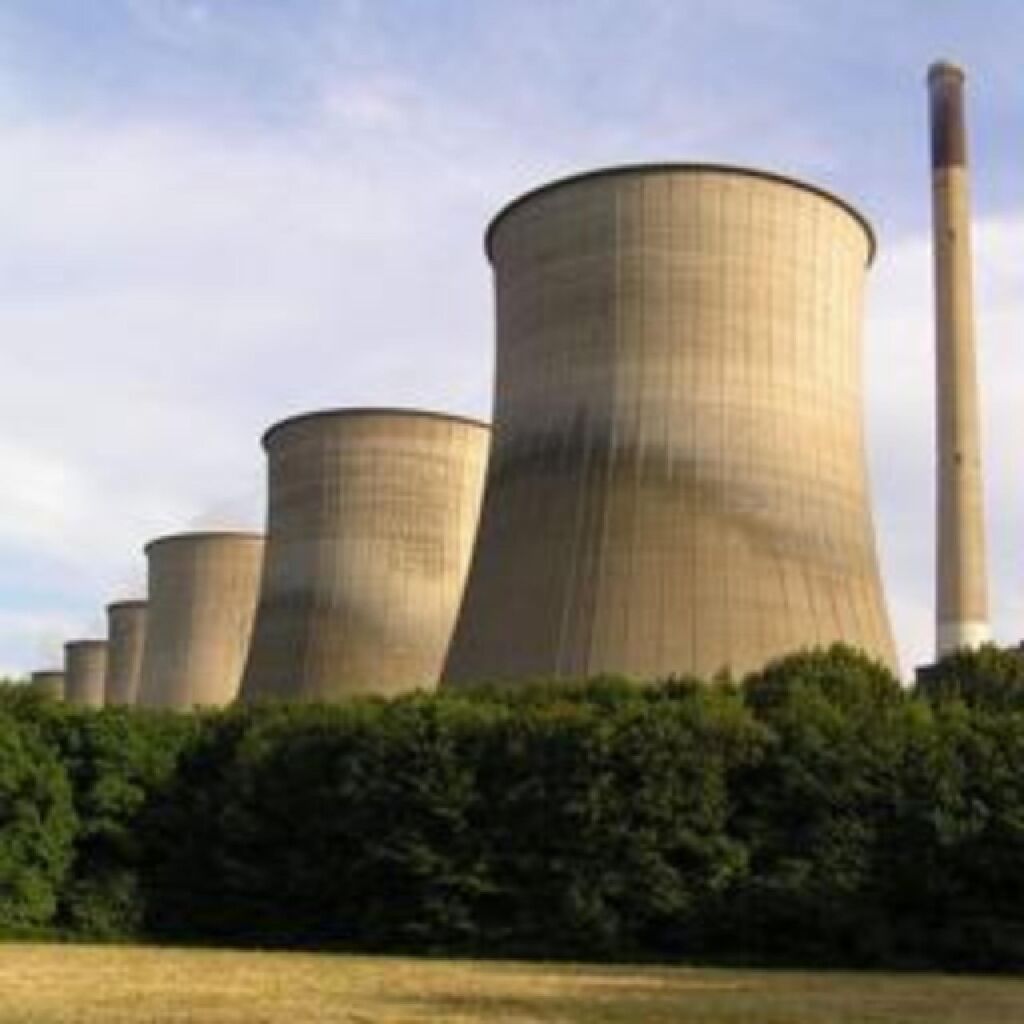}
    \end{subfigure}%
    \begin{subfigure}[b]{0.165\textwidth}
        \centering
        \includegraphics[width=\textwidth]{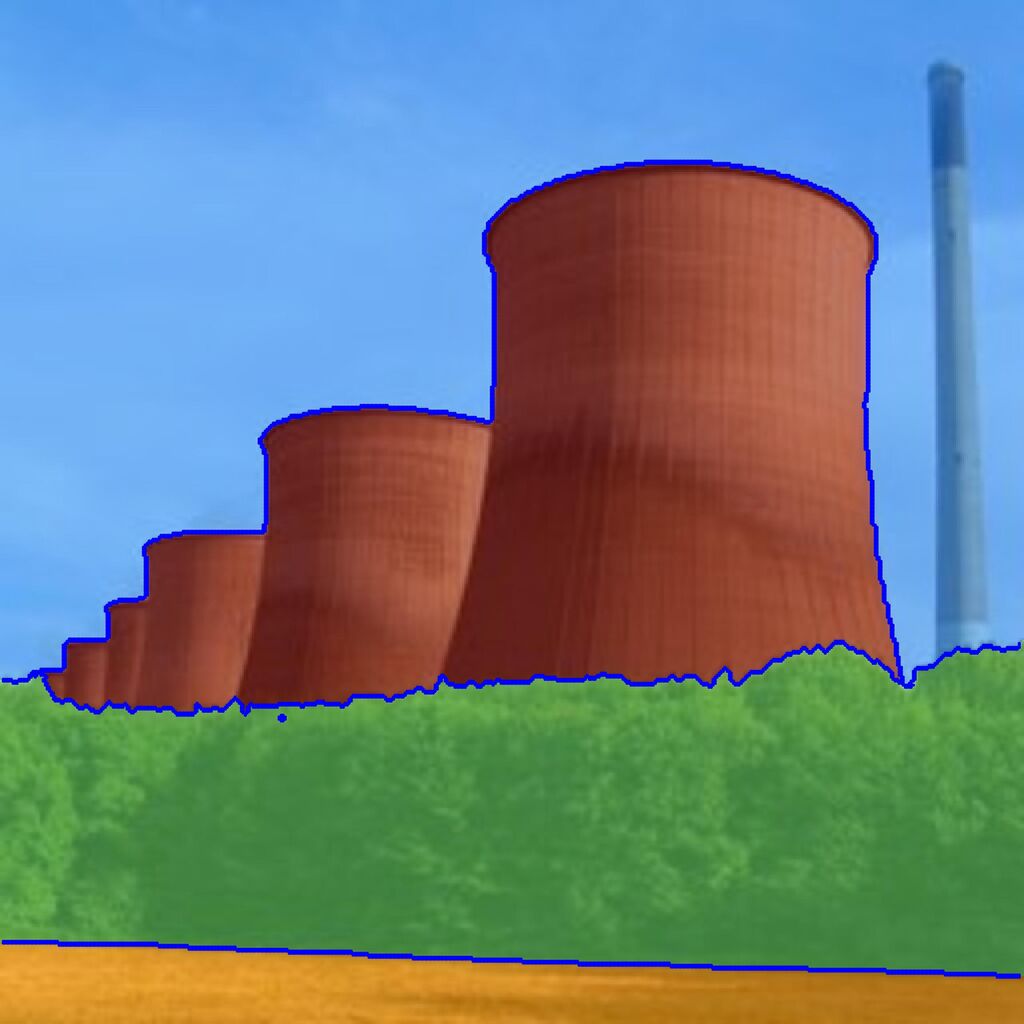}
    \end{subfigure}%
    \begin{subfigure}[b]{0.165\textwidth}
        \centering
        \includegraphics[width=\textwidth]{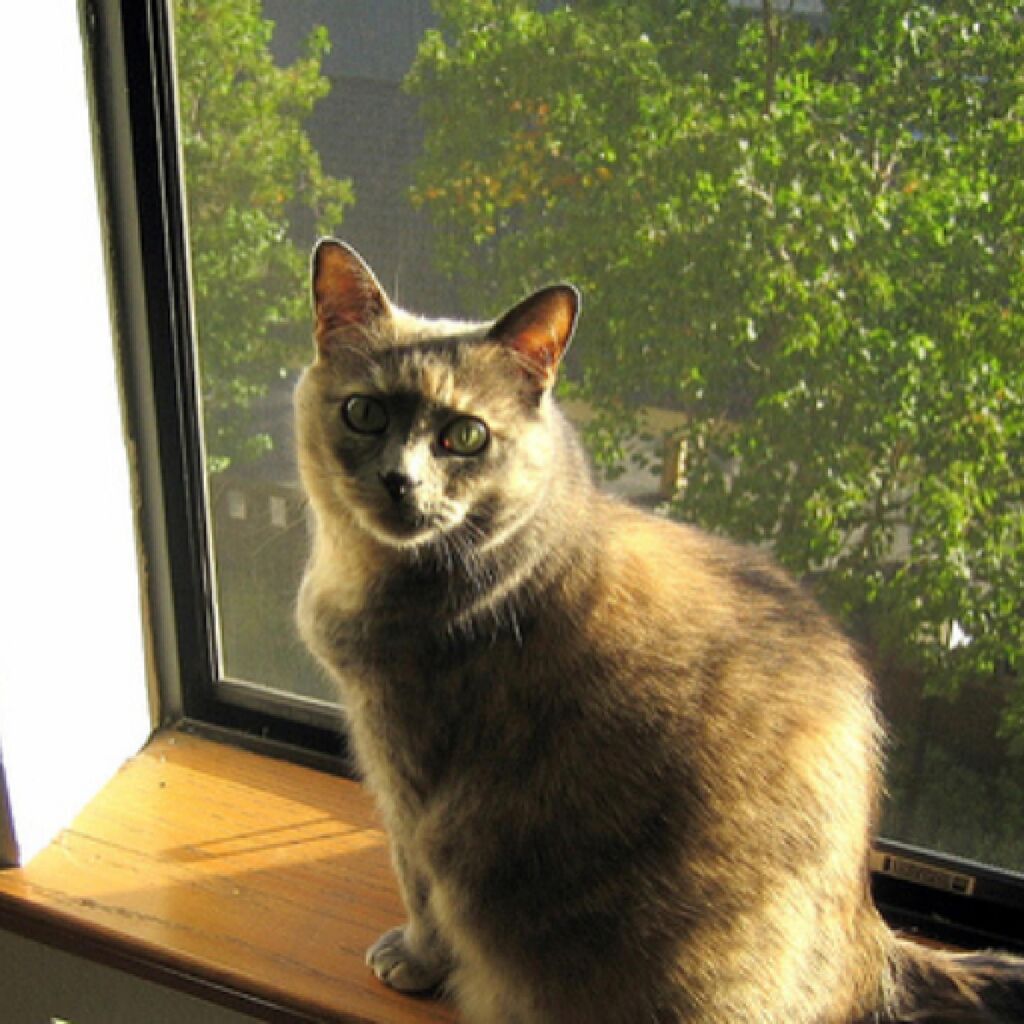}
    \end{subfigure}%
    \begin{subfigure}[b]{0.165\textwidth}
        \centering
        \includegraphics[width=\textwidth]{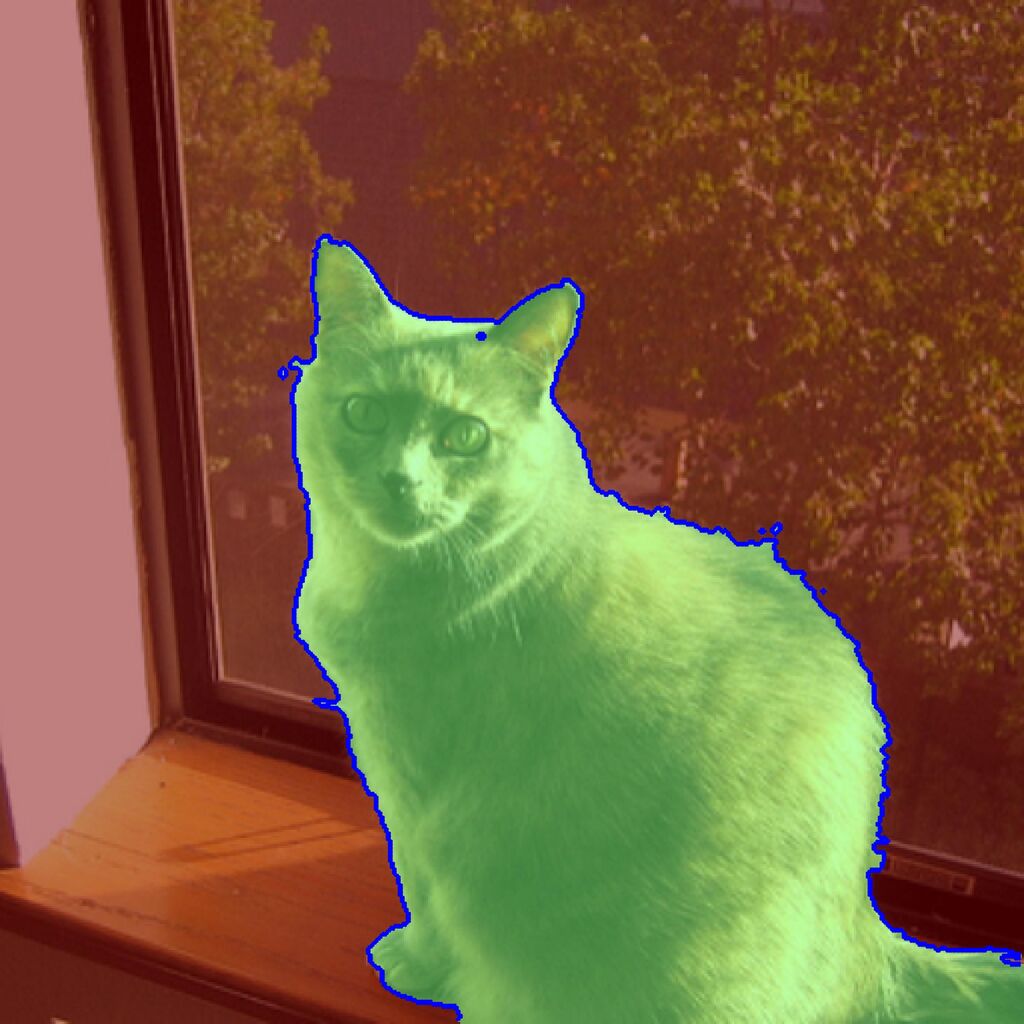}
    \end{subfigure}

    \caption{A gallery of images and their corresponding semantic segmentations generated by the proposed method.}
    \label{fig:gallery}
\end{figure*}
\begin{figure*}[h]
    \centering
    \begin{subfigure}[t]{0.225\textwidth}
        \caption*{\textbf{Image}}
        \centering
        \includegraphics[width=\textwidth]{figure/supp/png/qualitative_comparison/img/000000002153_img.jpg}
    \end{subfigure}%
    \begin{subfigure}[t]{0.225\textwidth}
        \caption*{\textbf{DiffCut}}
        \centering
        \includegraphics[width=\textwidth]{figure/supp/png/qualitative_comparison/diffcut/000000002153.jpg}
    \end{subfigure}%
    \begin{subfigure}[t]{0.225\textwidth}
        \caption*{\textbf{DiffSeg}}
        \centering
        \includegraphics[width=\textwidth]{figure/supp/png/qualitative_comparison/diffseg/000000002153.jpg}
    \end{subfigure}%
    \begin{subfigure}[t]{0.225\textwidth}
        \caption*{\textbf{Ours}}
        \centering
        \includegraphics[width=\textwidth]{figure/supp/png/qualitative_comparison/ours/000000002153_mask_refined.jpg}
    \end{subfigure}%
    
    \vspace{-1em}
    
    \begin{subfigure}[b]{0.225\textwidth}
        \centering
        \includegraphics[width=\textwidth]{figure/supp/png/qualitative_comparison/img/000000044699_img.jpg}
    \end{subfigure}%
    \begin{subfigure}[b]{0.225\textwidth}
        \centering
        \includegraphics[width=\textwidth]{figure/supp/png/qualitative_comparison/diffcut/000000044699.jpg}
    \end{subfigure}%
    \begin{subfigure}[b]{0.225\textwidth}
        \centering
        \includegraphics[width=\textwidth]{figure/supp/png/qualitative_comparison/diffseg/000000044699.jpg}
    \end{subfigure}%
    \begin{subfigure}[b]{0.225\textwidth}
        \centering
        \includegraphics[width=\textwidth]{figure/supp/png/qualitative_comparison/ours/000000044699_mask_refined.jpg}
    \end{subfigure}%

    \vspace{-1em}
    
    \begin{subfigure}[b]{0.225\textwidth}
        \centering
        \includegraphics[width=\textwidth]{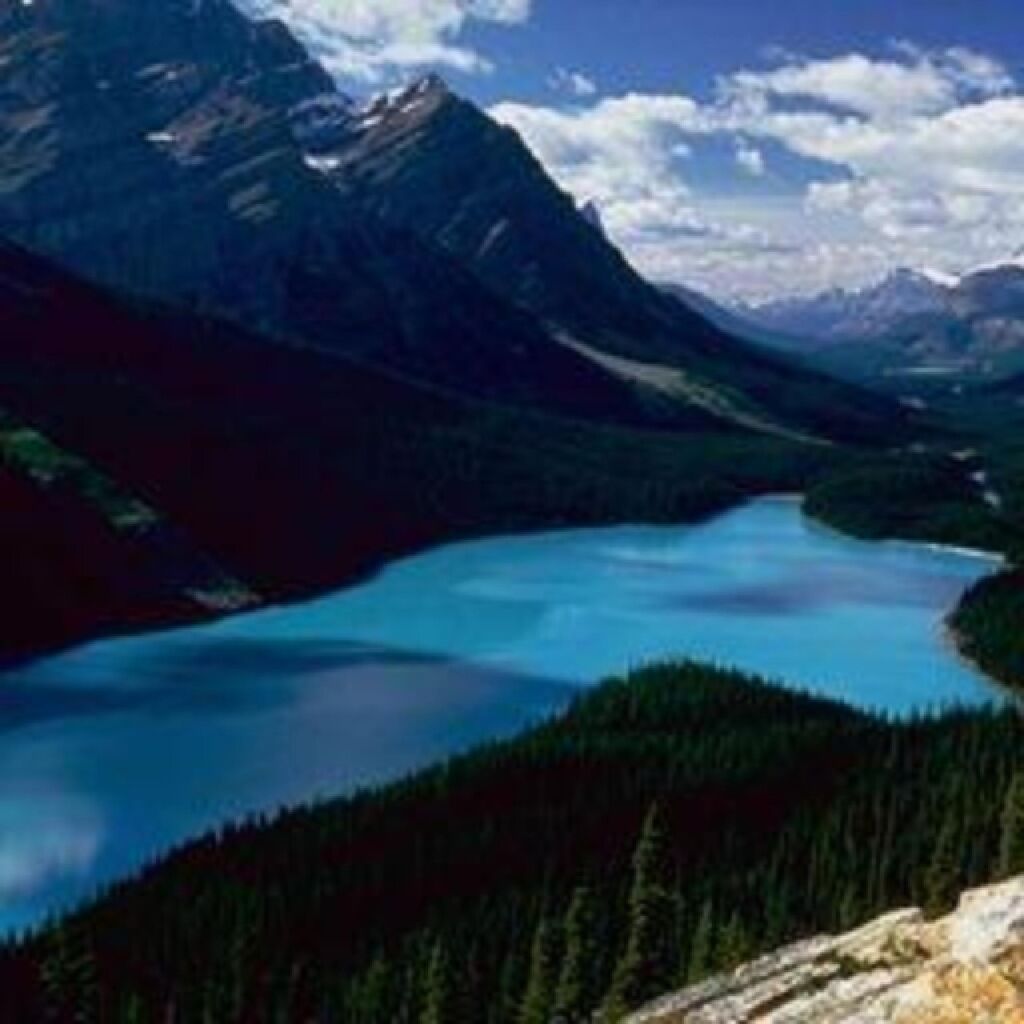}
    \end{subfigure}%
    \begin{subfigure}[b]{0.225\textwidth}
        \centering
        \includegraphics[width=\textwidth]{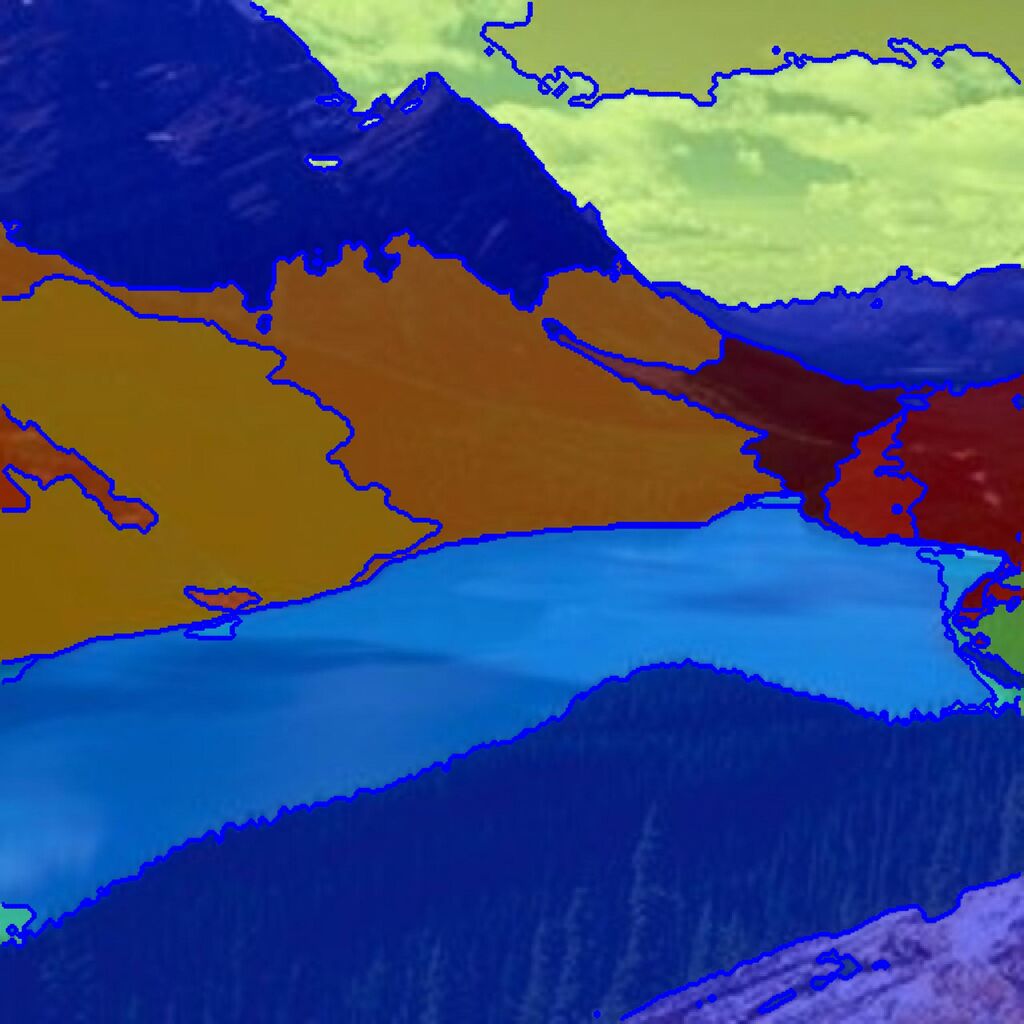}
    \end{subfigure}%
    \begin{subfigure}[b]{0.225\textwidth}
        \centering
        \includegraphics[width=\textwidth]{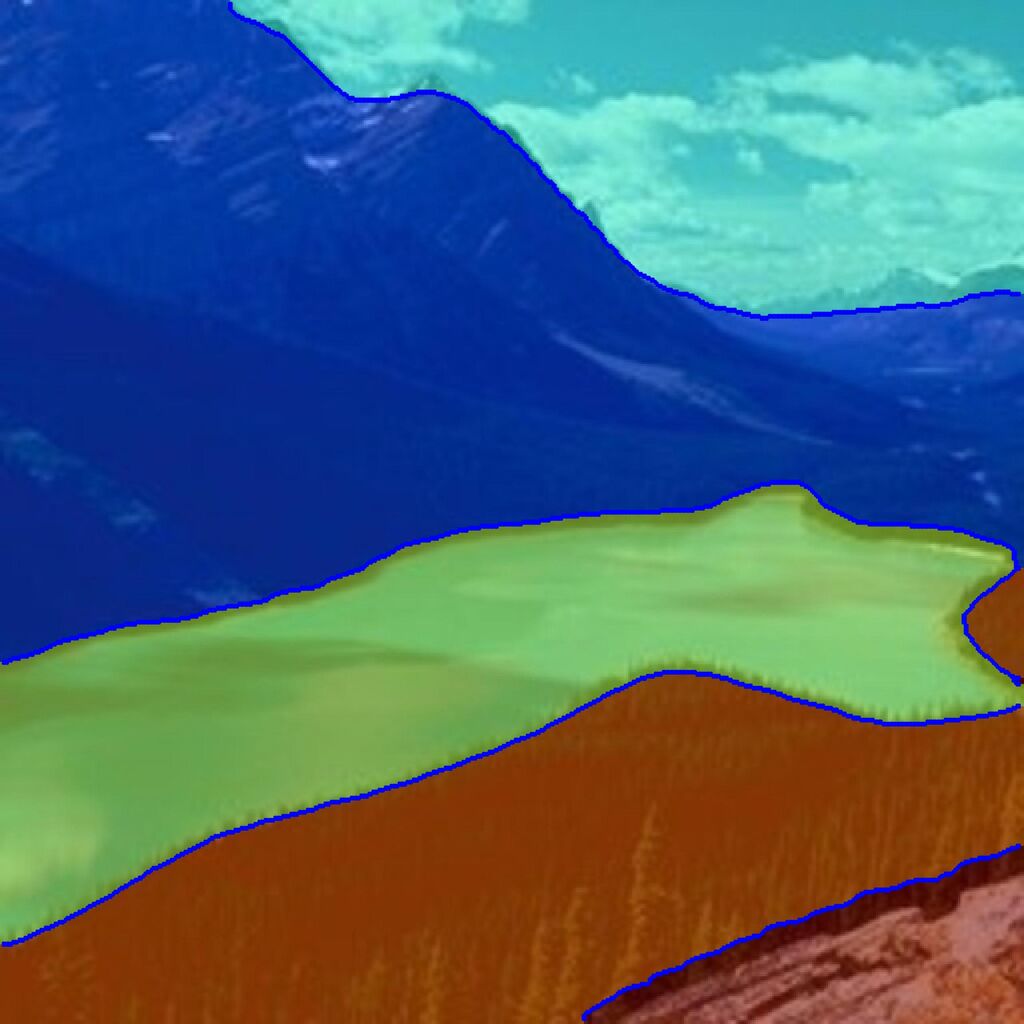}
    \end{subfigure}%
    \begin{subfigure}[b]{0.225\textwidth}
        \centering
        \includegraphics[width=\textwidth]{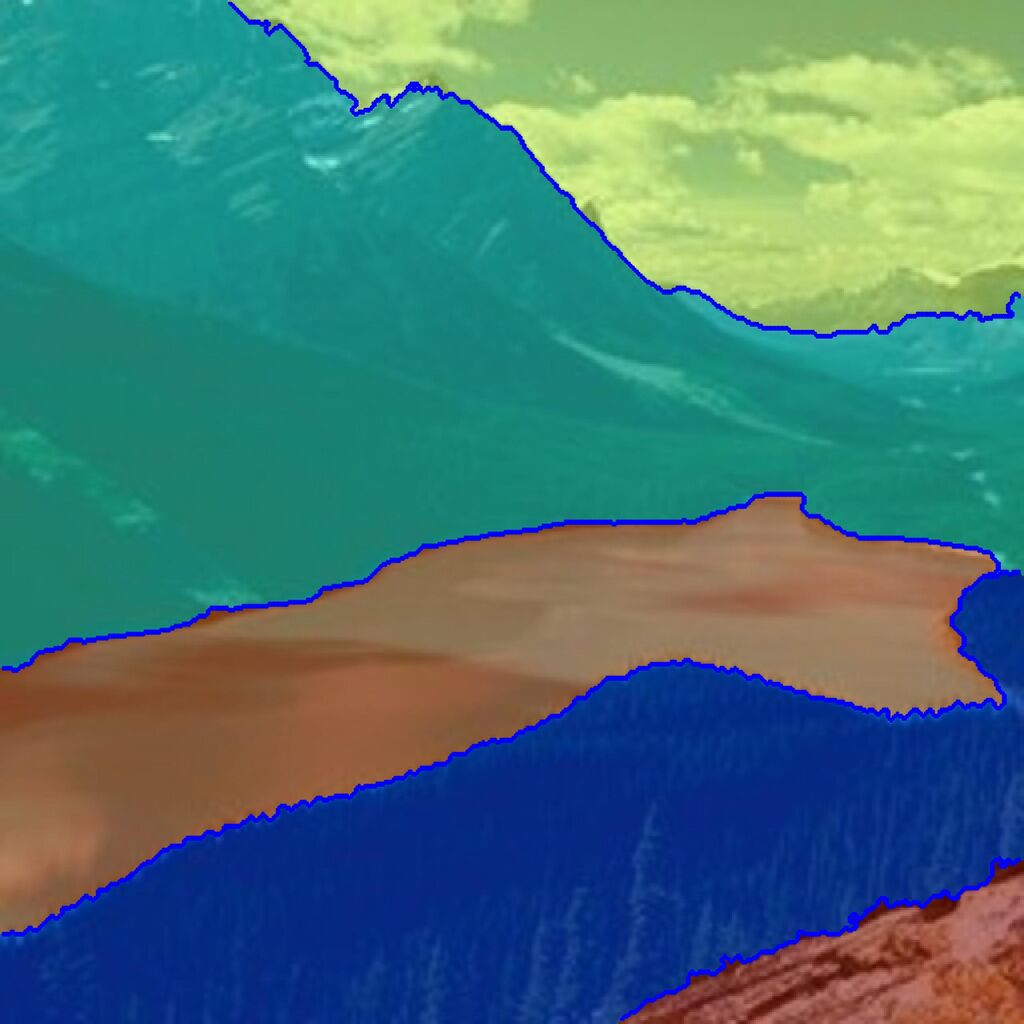}
    \end{subfigure}%

    \vspace{-1em}
    
    \begin{subfigure}[b]{0.225\textwidth}
        \centering
        \includegraphics[width=\textwidth]{figure/supp/png/qualitative_comparison/img/000000080273_img.jpg}
    \end{subfigure}%
    \begin{subfigure}[b]{0.225\textwidth}
        \centering
        \includegraphics[width=\textwidth]{figure/supp/png/qualitative_comparison/diffcut/000000080273.jpg}
    \end{subfigure}%
    \begin{subfigure}[b]{0.225\textwidth}
        \centering
        \includegraphics[width=\textwidth]{figure/supp/png/qualitative_comparison/diffseg/000000080273.jpg}
    \end{subfigure}%
    \begin{subfigure}[b]{0.225\textwidth}
        \centering
        \includegraphics[width=\textwidth]{figure/supp/png/qualitative_comparison/ours/000000080273_mask_refined.jpg}
    \end{subfigure}%

    \vspace{-1em}
    
    \begin{subfigure}[b]{0.225\textwidth}
        \centering
        \includegraphics[width=\textwidth]{figure/supp/png/qualitative_comparison/img/2007_003991_img.jpg}
    \end{subfigure}%
    \begin{subfigure}[b]{0.225\textwidth}
        \centering
        \includegraphics[width=\textwidth]{figure/supp/png/qualitative_comparison/diffcut/2007_003991.jpg}
    \end{subfigure}%
    \begin{subfigure}[b]{0.225\textwidth}
        \centering
        \includegraphics[width=\textwidth]{figure/supp/png/qualitative_comparison/diffseg/2007_003991.jpg}
    \end{subfigure}%
    \begin{subfigure}[b]{0.225\textwidth}
        \centering
        \includegraphics[width=\textwidth]{figure/supp/png/qualitative_comparison/ours/2007_003991_mask_refined.jpg}
    \end{subfigure}%

    \caption{Qualitative comparison of the proposed method against DiffSeg and DiffCut.}
    \label{fig:baseline_qualitative_comparison_1}
\end{figure*}
\begin{figure*}[h]
    \centering
    
    \begin{subfigure}[t]{0.225\textwidth}
        \caption*{\textbf{Image}}
        \centering
        \includegraphics[width=\textwidth]{figure/supp/png/qualitative_comparison/img/ADE_val_00000083_img.jpg}
    \end{subfigure}%
    \begin{subfigure}[t]{0.225\textwidth}
        \caption*{\textbf{DiffCut}}
        \centering
        \includegraphics[width=\textwidth]{figure/supp/png/qualitative_comparison/diffcut/ADE_val_00000083.jpg}
    \end{subfigure}%
    \begin{subfigure}[t]{0.225\textwidth}
        \caption*{\textbf{DiffSeg}}
        \centering
        \includegraphics[width=\textwidth]{figure/supp/png/qualitative_comparison/diffseg/ADE_val_00000083.jpg}
    \end{subfigure}%
    \begin{subfigure}[t]{0.225\textwidth}
        \caption*{\textbf{Ours}}
        \centering
        \includegraphics[width=\textwidth]{figure/supp/png/qualitative_comparison/ours/ADE_val_00000083_mask.jpg}
    \end{subfigure}%

    \vspace{-1em}

    \begin{subfigure}[t]{0.225\textwidth}
        \centering
        \includegraphics[width=\textwidth]{figure/supp/png/qualitative_comparison/img/ADE_val_00000660_img.jpg}
    \end{subfigure}%
    \begin{subfigure}[t]{0.225\textwidth}
        \centering
        \includegraphics[width=\textwidth]{figure/supp/png/qualitative_comparison/diffcut/ADE_val_00000660.jpg}
    \end{subfigure}%
    \begin{subfigure}[t]{0.225\textwidth}
        \centering
        \includegraphics[width=\textwidth]{figure/supp/png/qualitative_comparison/diffseg/ADE_val_00000660.jpg}
    \end{subfigure}%
    \begin{subfigure}[t]{0.225\textwidth}
        \centering
        \includegraphics[width=\textwidth]{figure/supp/png/qualitative_comparison/ours/ADE_val_00000660_mask.jpg}
    \end{subfigure}%

    \vspace{-1em}

    \begin{subfigure}[t]{0.225\textwidth}
        \centering
        \includegraphics[width=\textwidth]{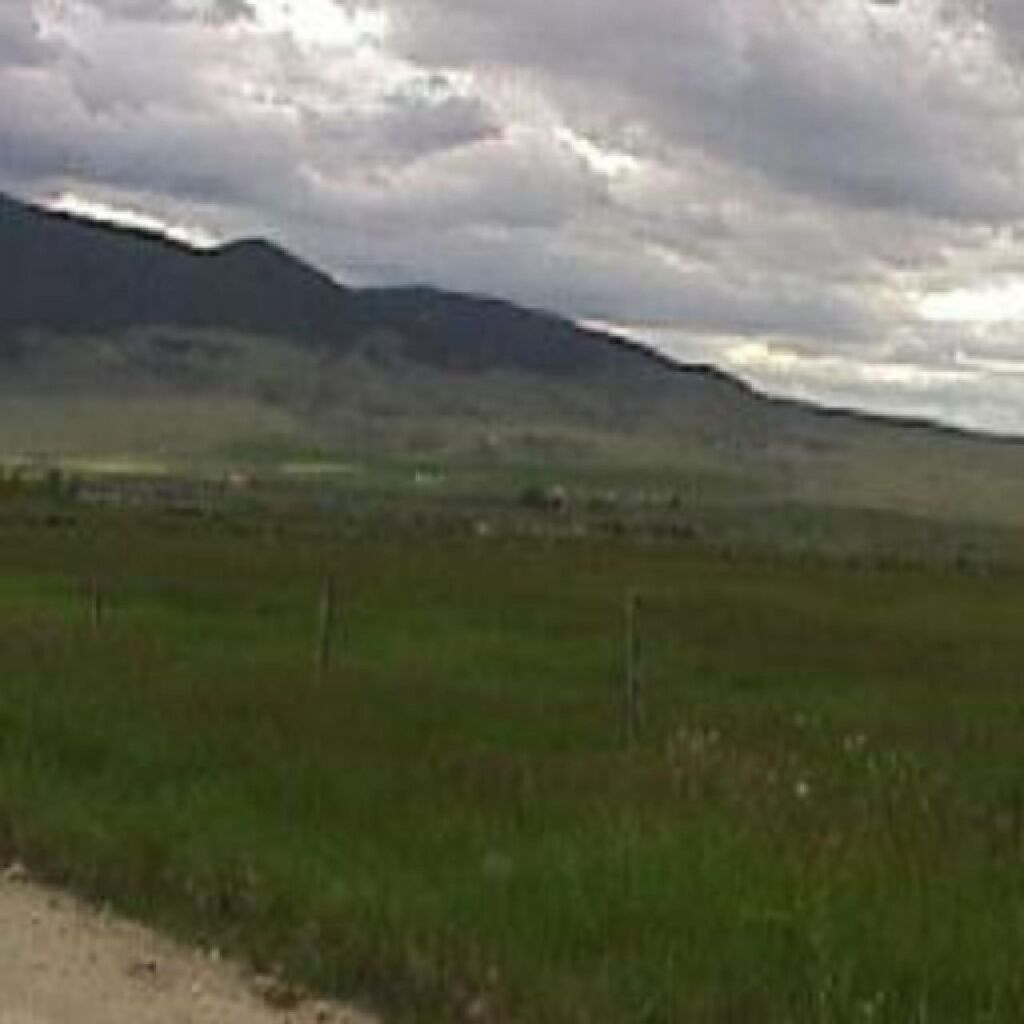}
    \end{subfigure}%
    \begin{subfigure}[t]{0.225\textwidth}
        \centering
        \includegraphics[width=\textwidth]{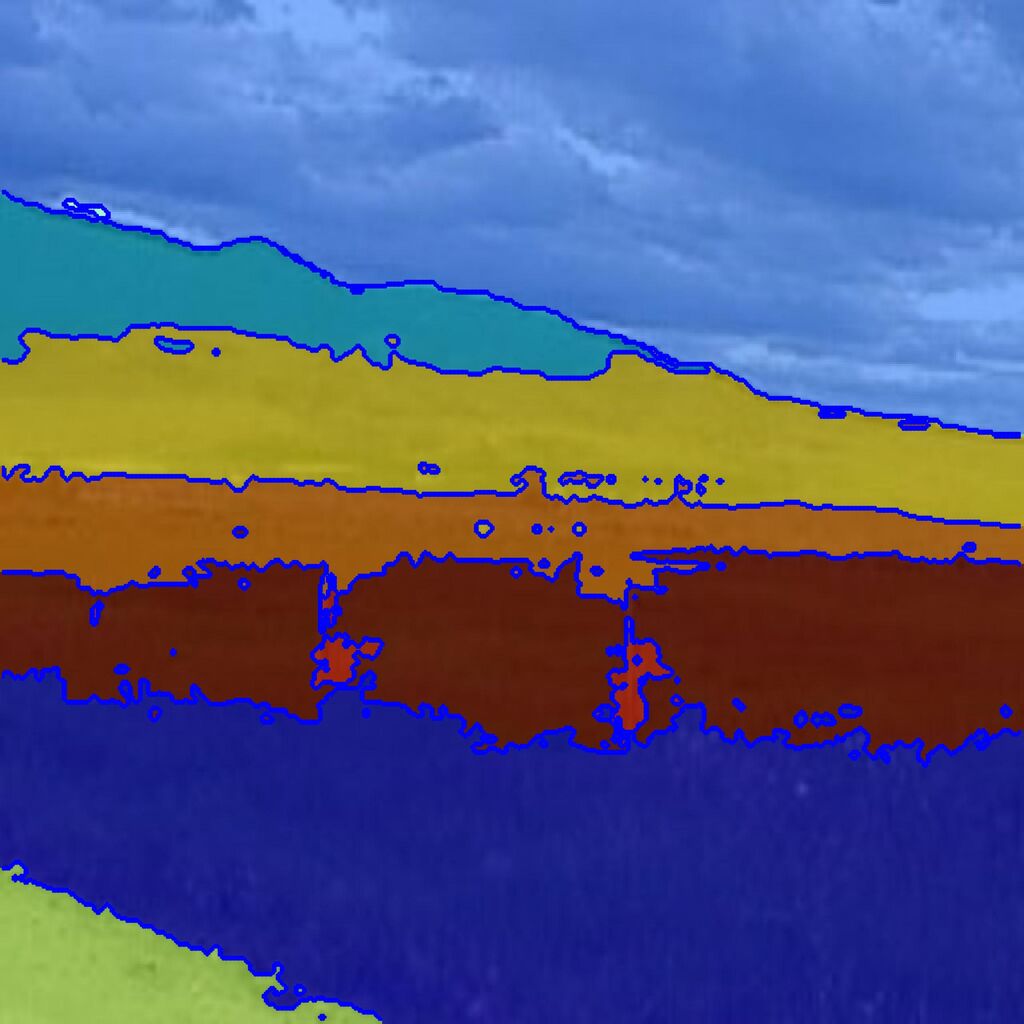}
    \end{subfigure}%
    \begin{subfigure}[t]{0.225\textwidth}
        \centering
        \includegraphics[width=\textwidth]{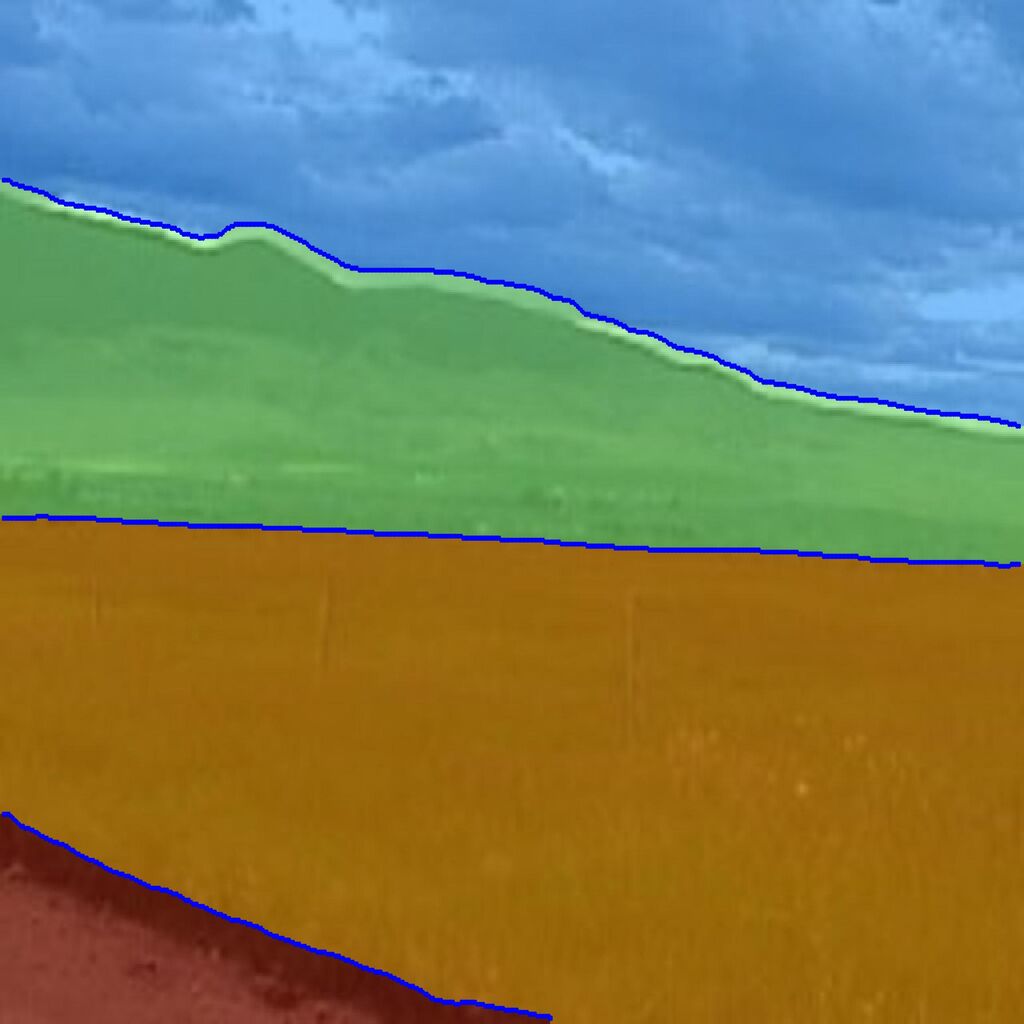}
    \end{subfigure}%
    \begin{subfigure}[t]{0.225\textwidth}
        \includegraphics[width=\textwidth]{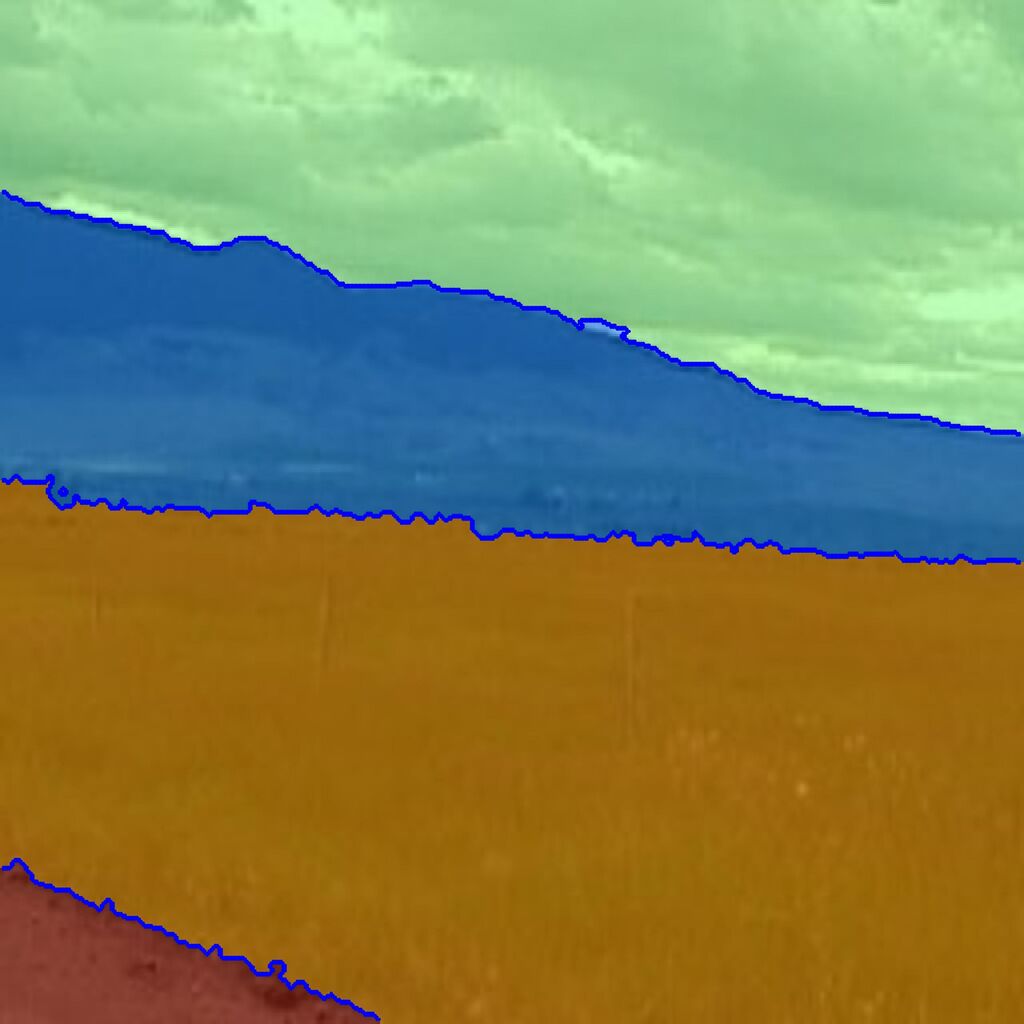}
    \end{subfigure}%

    \vspace{-1em}

    \begin{subfigure}[t]{0.225\textwidth}
        \centering
        \includegraphics[width=\textwidth]{figure/supp/png/qualitative_comparison/img/ADE_val_00000916_img.jpg}
    \end{subfigure}%
    \begin{subfigure}[t]{0.225\textwidth}
        \centering
        \includegraphics[width=\textwidth]{figure/supp/png/qualitative_comparison/diffcut/ADE_val_00000916.jpg}
    \end{subfigure}%
    \begin{subfigure}[t]{0.225\textwidth}
        \centering
        \includegraphics[width=\textwidth]{figure/supp/png/qualitative_comparison/diffseg/ADE_val_00000916.jpg}
    \end{subfigure}%
    \begin{subfigure}[t]{0.225\textwidth}
        \centering
        \includegraphics[width=\textwidth]{figure/supp/png/qualitative_comparison/ours/ADE_val_00000916_mask.jpg}
    \end{subfigure}%

    \vspace{-1em}

    \begin{subfigure}[t]{0.225\textwidth}
        \centering
        \includegraphics[width=\textwidth]{figure/supp/png/qualitative_comparison/img/ADE_val_00000918_img.jpg}
    \end{subfigure}%
    \begin{subfigure}[t]{0.225\textwidth}
        \centering
        \includegraphics[width=\textwidth]{figure/supp/png/qualitative_comparison/diffcut/ADE_val_00000918.jpg}
    \end{subfigure}%
    \begin{subfigure}[t]{0.225\textwidth}
        \centering
        \includegraphics[width=\textwidth]{figure/supp/png/qualitative_comparison/diffseg/ADE_val_00000918.jpg}
    \end{subfigure}%
    \begin{subfigure}[t]{0.225\textwidth}
        \centering
        \includegraphics[width=\textwidth]{figure/supp/png/qualitative_comparison/ours/ADE_val_00000918_mask.jpg}
    \end{subfigure}%

    \vspace{-1em}
    
    \caption{Qualitative comparison of the proposed method against DiffSeg and DiffCut.}
    \label{fig:baseline_qualitative_comparison_2}
\end{figure*}
\begin{figure*}[h]
    \centering
    
    \begin{subfigure}[t]{0.225\textwidth}
        \caption*{\textbf{Image}}
        \centering
        \includegraphics[width=\textwidth]{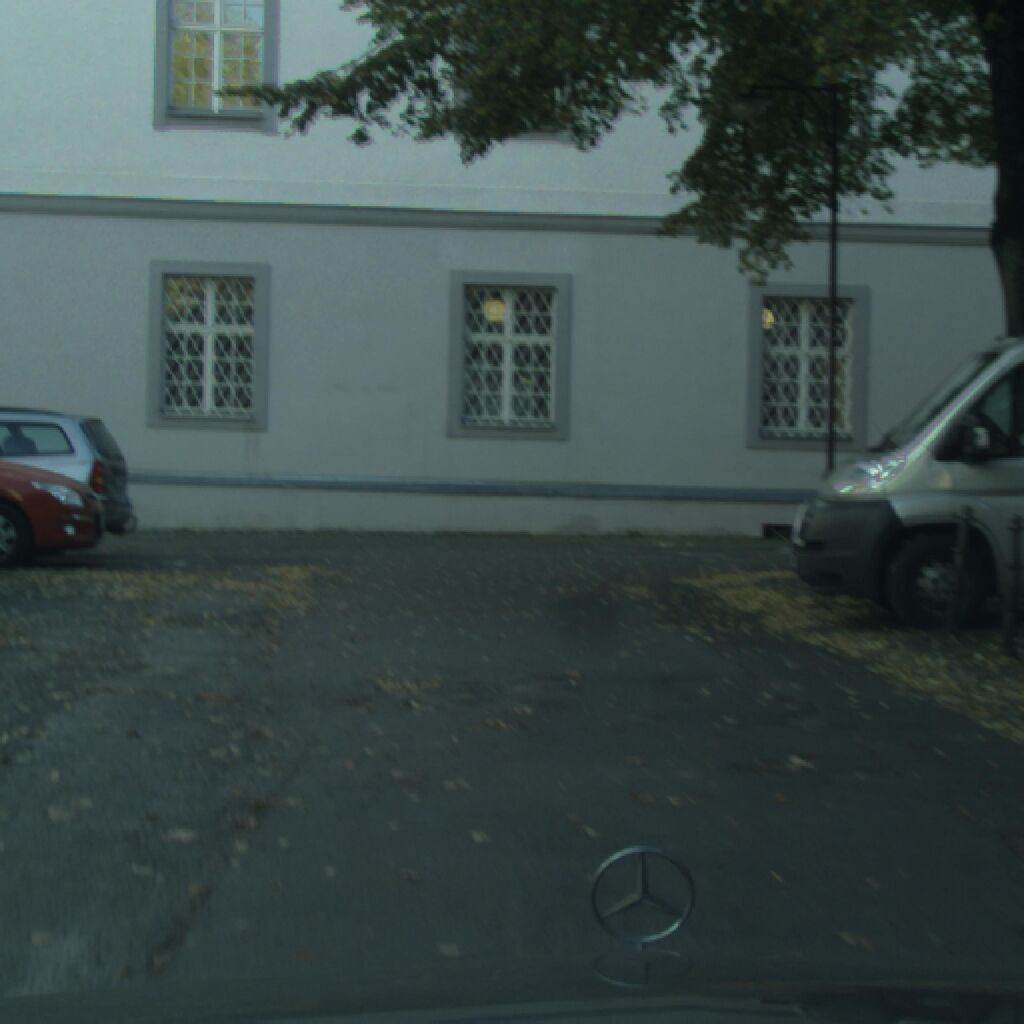}
    \end{subfigure}%
    \begin{subfigure}[t]{0.225\textwidth}
        \caption*{\textbf{DiffCut}}
        \centering
        \includegraphics[width=\textwidth]{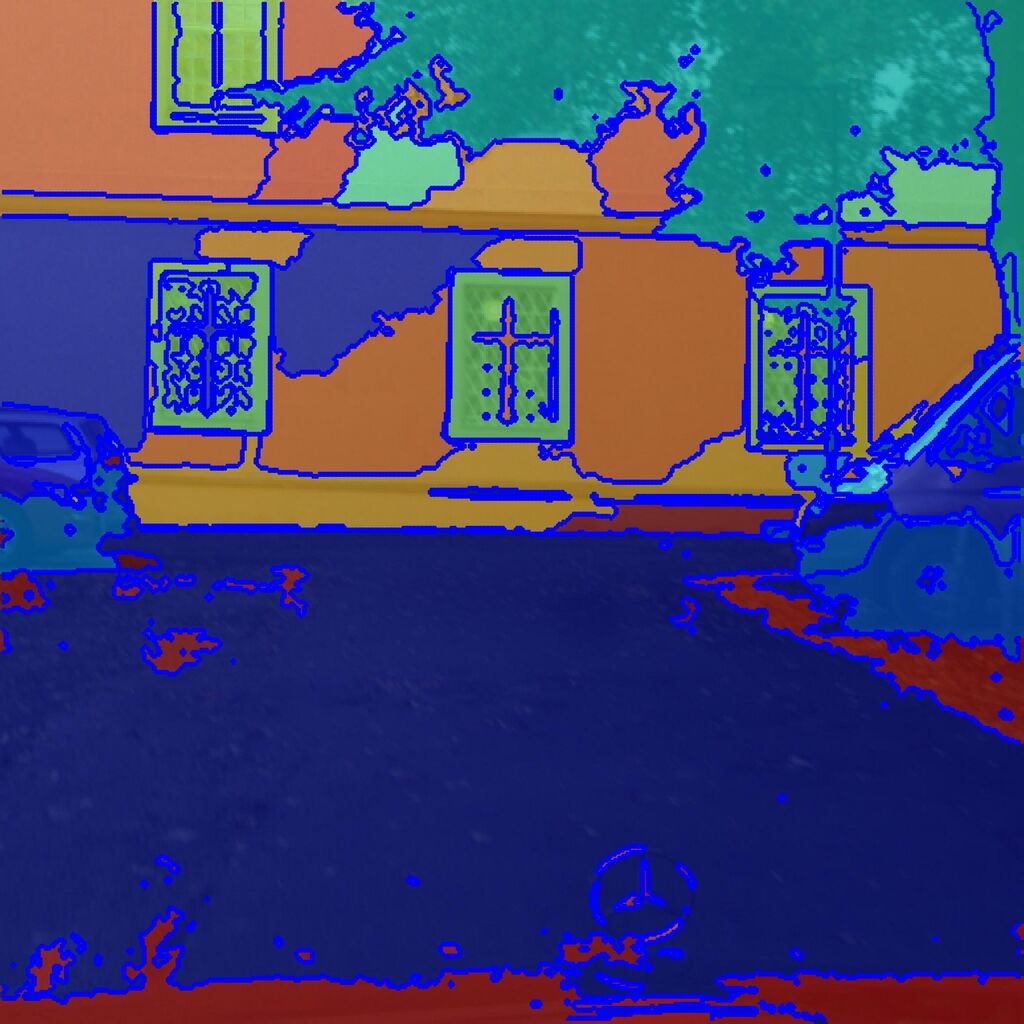}
    \end{subfigure}%
    \begin{subfigure}[t]{0.225\textwidth}
        \caption*{\textbf{DiffSeg}}
        \centering
        \includegraphics[width=\textwidth]{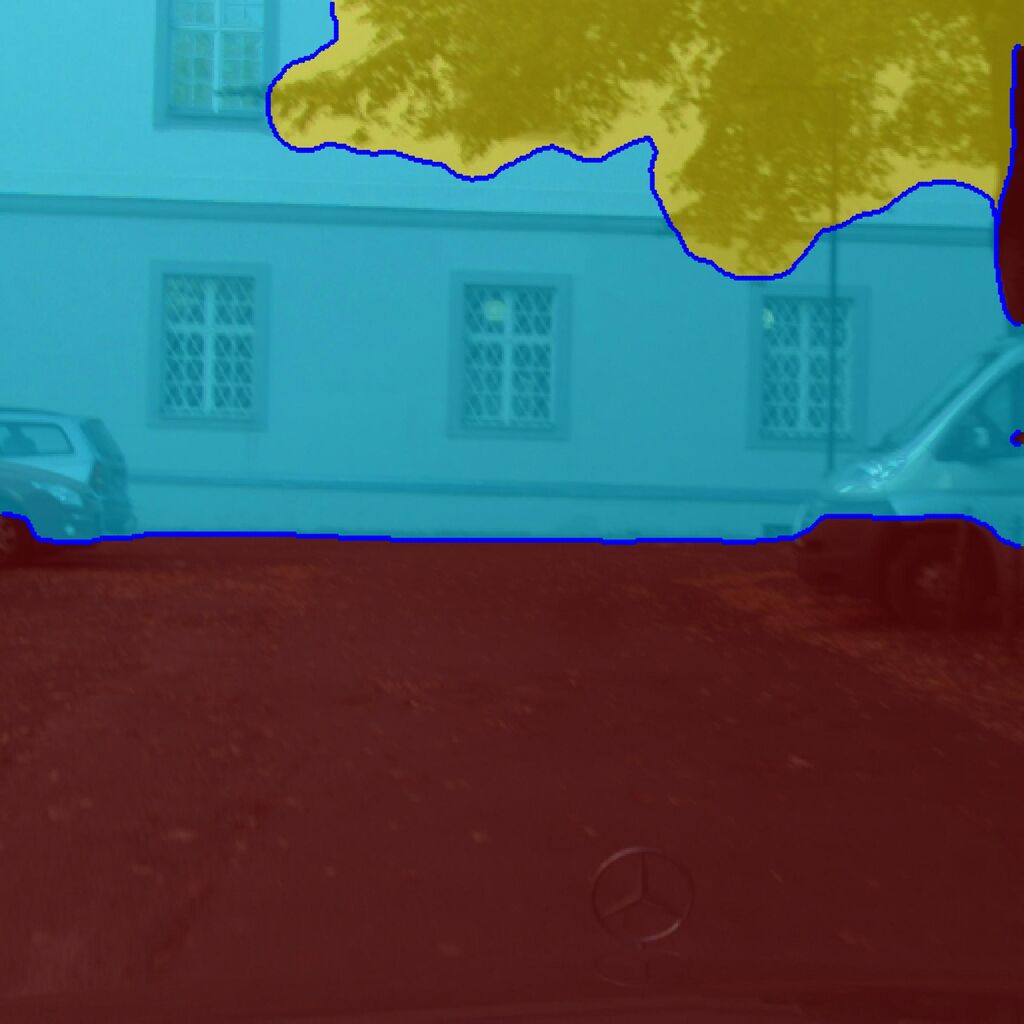}
    \end{subfigure}%
    \begin{subfigure}[t]{0.225\textwidth}
        \caption*{\textbf{Ours}}
        \centering
        \includegraphics[width=\textwidth]{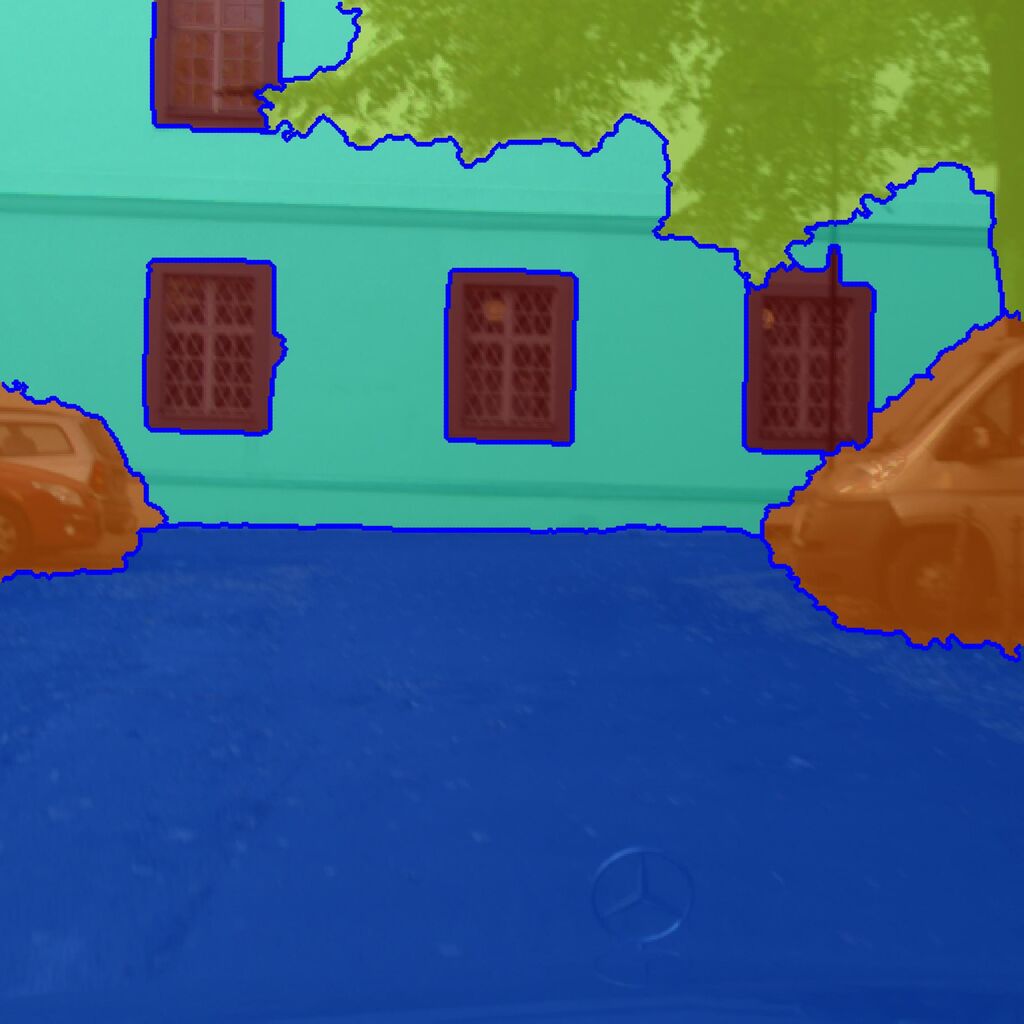}
    \end{subfigure}%

    \vspace{-1em}
    
    \begin{subfigure}[b]{0.225\textwidth}
        \centering
        \includegraphics[width=\textwidth]{figure/supp/png/qualitative_comparison/img/2008_001478_img.jpg}
    \end{subfigure}%
    \begin{subfigure}[b]{0.225\textwidth}
        \centering
        \includegraphics[width=\textwidth]{figure/supp/png/qualitative_comparison/diffcut/2008_001478.jpg}
    \end{subfigure}%
    \begin{subfigure}[b]{0.225\textwidth}
        \centering
        \includegraphics[width=\textwidth]{figure/supp/png/qualitative_comparison/diffseg/2008_001478.jpg}
    \end{subfigure}%
    \begin{subfigure}[b]{0.225\textwidth}
        \centering
        \includegraphics[width=\textwidth]{figure/supp/png/qualitative_comparison/ours/2008_001478_mask_refined.jpg}
    \end{subfigure}%

    \vspace{-1em}

    \begin{subfigure}[t]{0.225\textwidth}
        \centering
        \includegraphics[width=\textwidth]{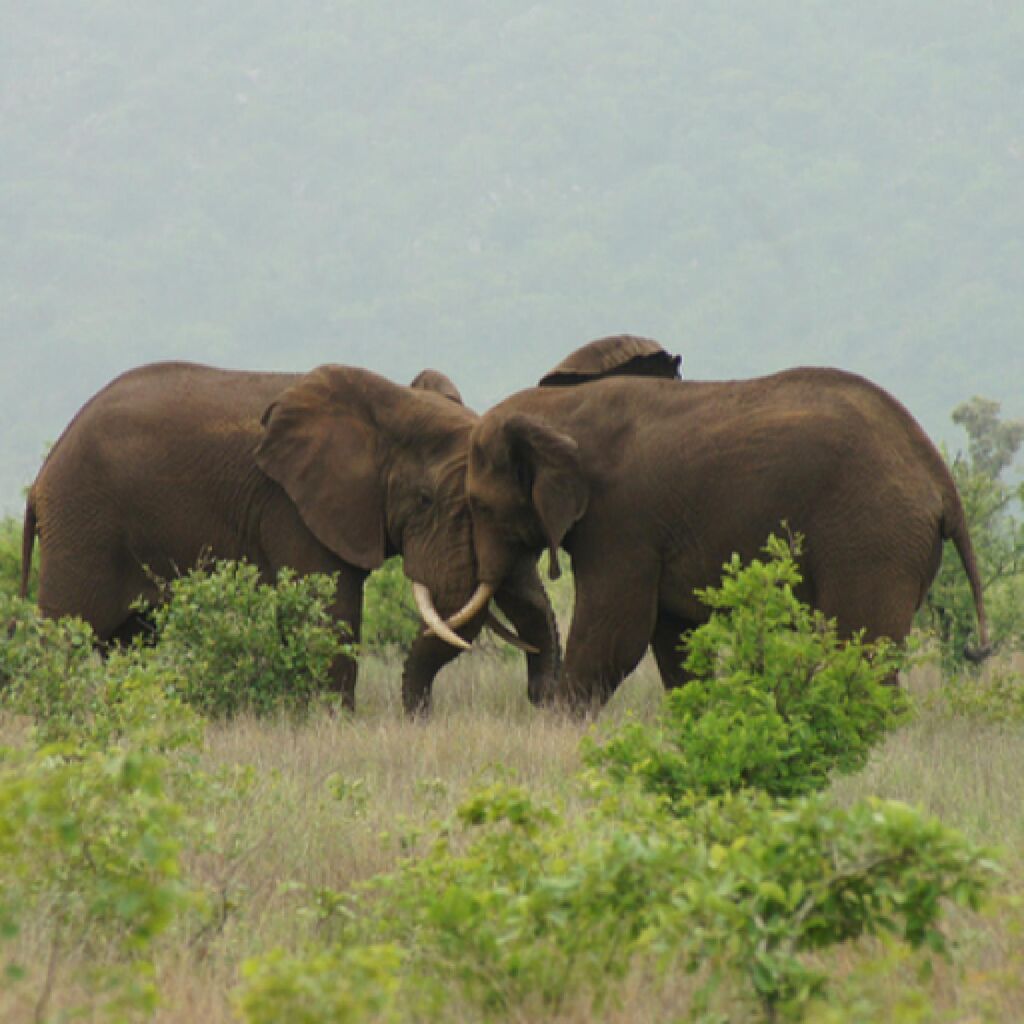}
    \end{subfigure}%
    \begin{subfigure}[t]{0.225\textwidth}
        \centering
        \includegraphics[width=\textwidth]{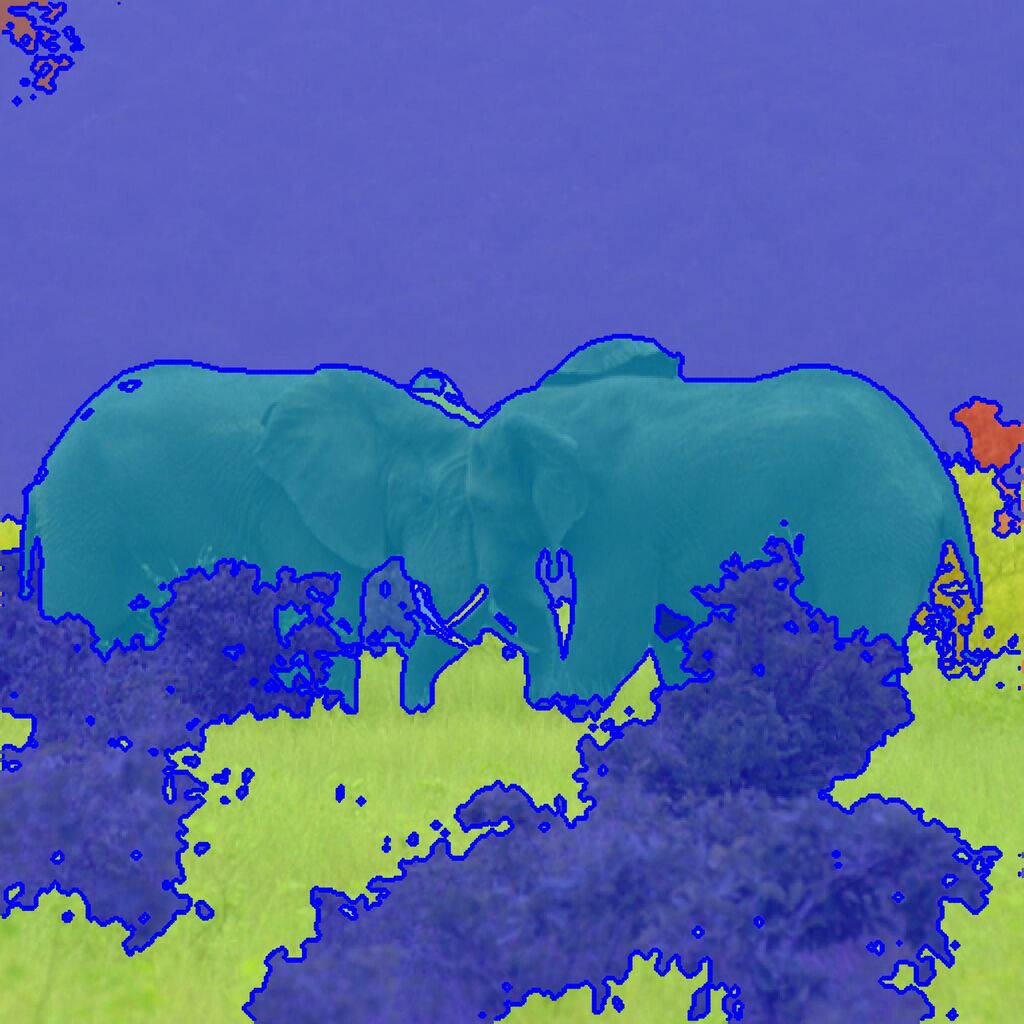}
    \end{subfigure}%
    \begin{subfigure}[t]{0.225\textwidth}
        \centering
        \includegraphics[width=\textwidth]{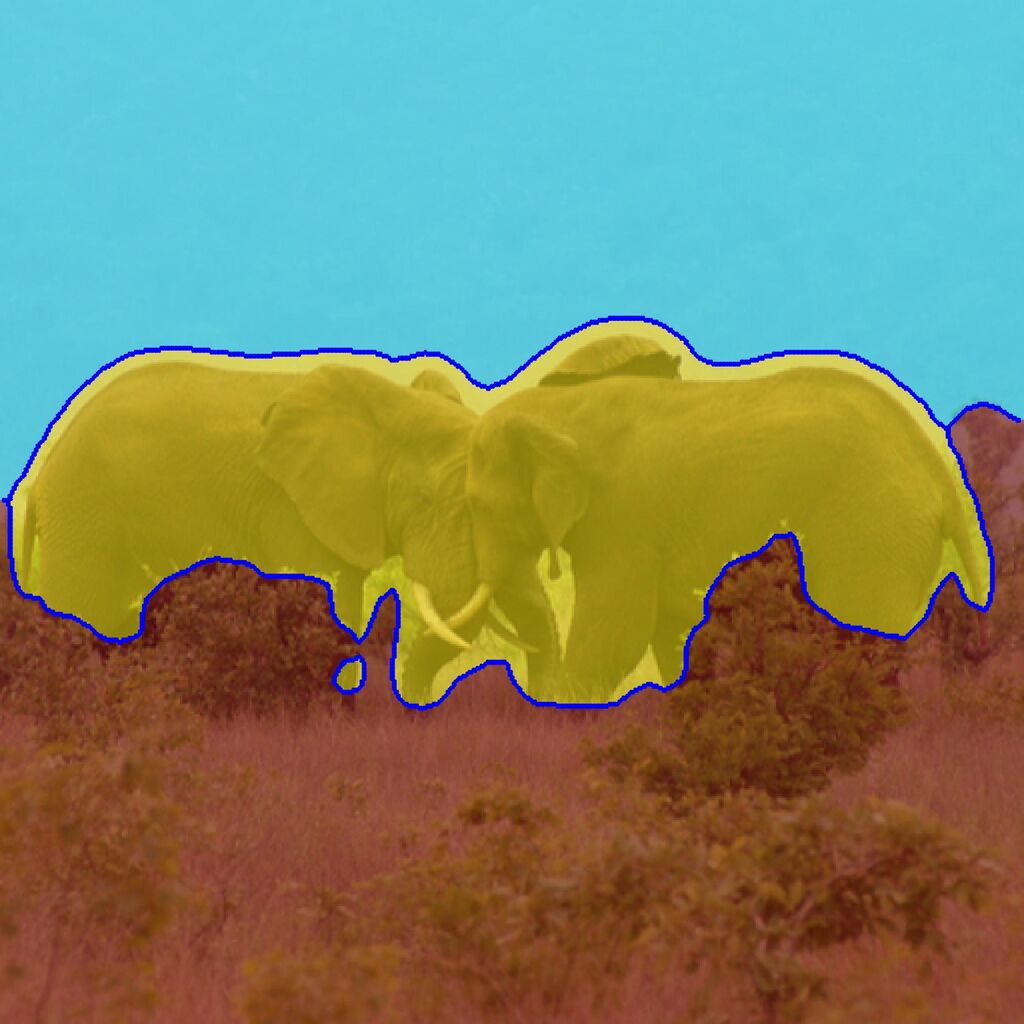}
    \end{subfigure}%
    \begin{subfigure}[t]{0.225\textwidth}
        \centering
        \includegraphics[width=\textwidth]{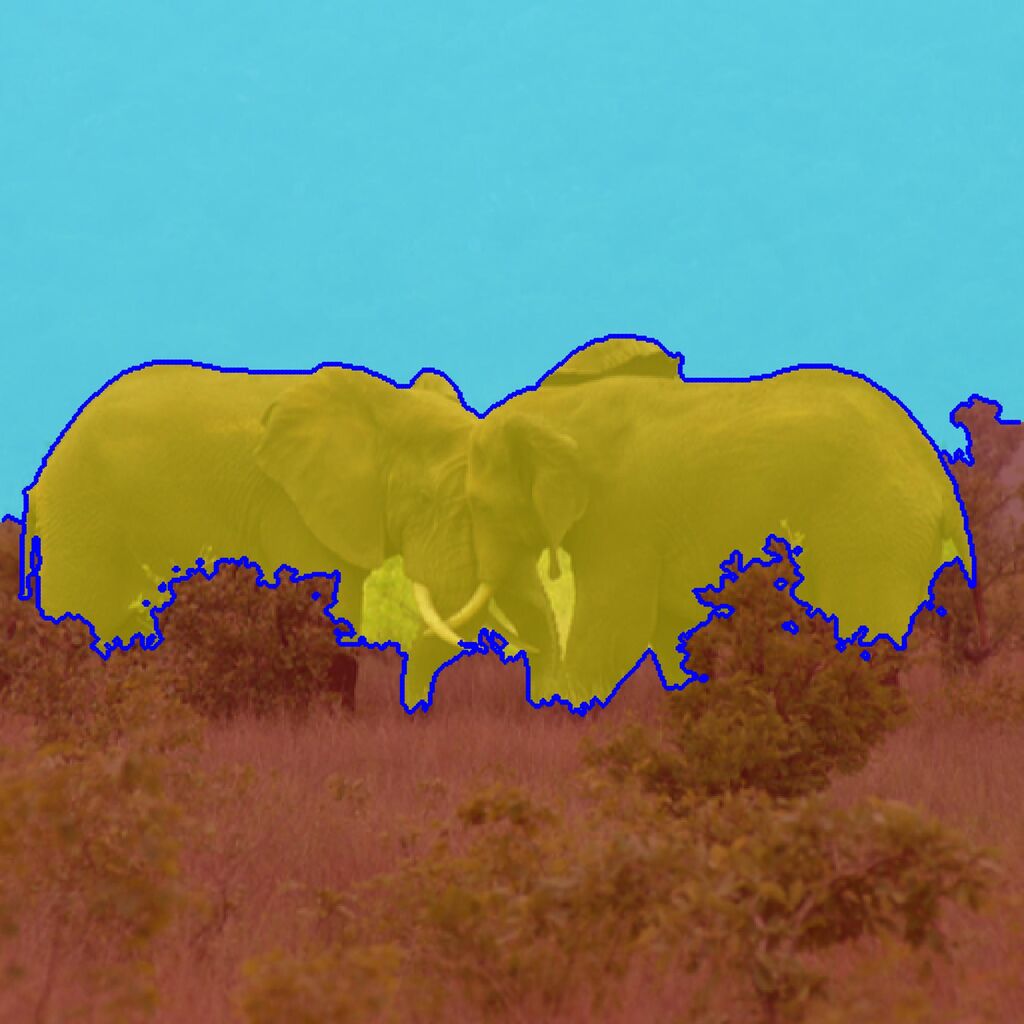}
    \end{subfigure}%

    \vspace{-1em}

    \begin{subfigure}[t]{0.225\textwidth}
        \centering
        \includegraphics[width=\textwidth]{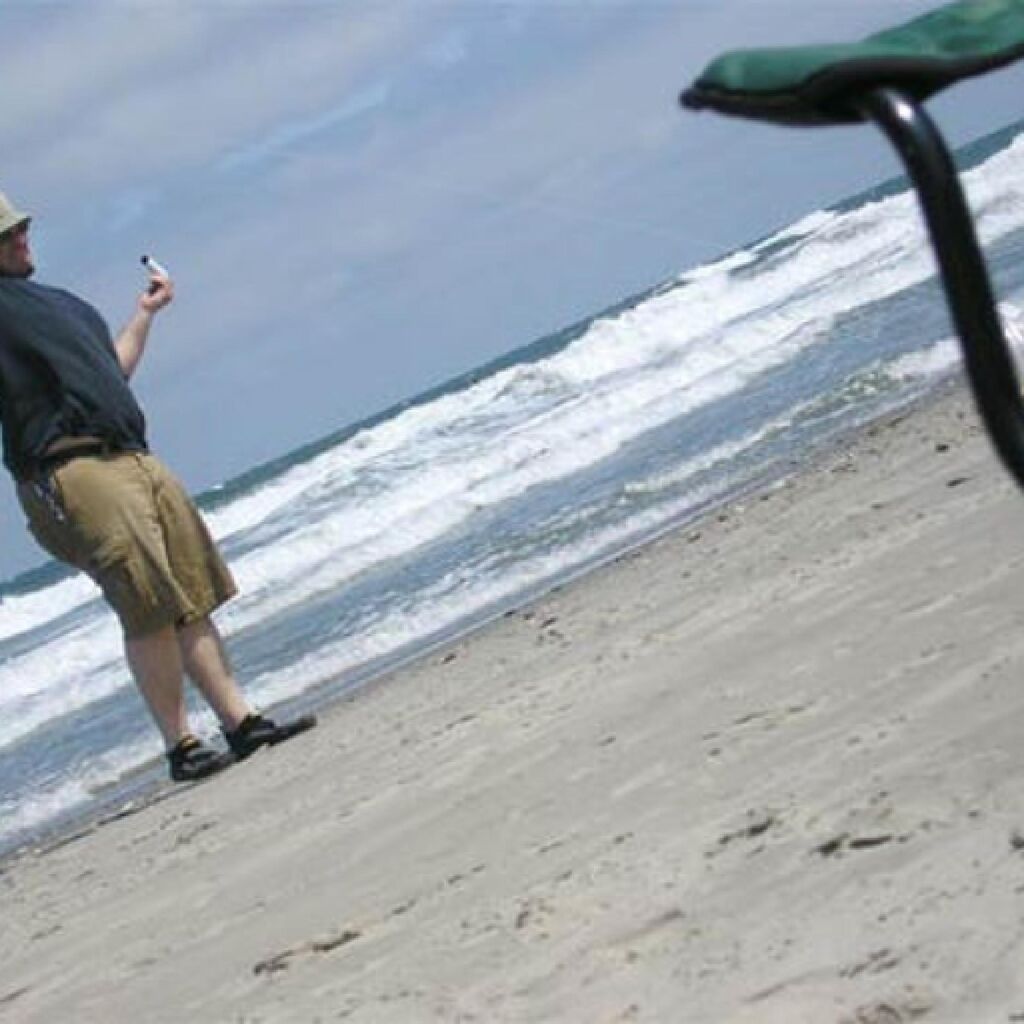}
    \end{subfigure}%
    \begin{subfigure}[t]{0.225\textwidth}
        \centering
        \includegraphics[width=\textwidth]{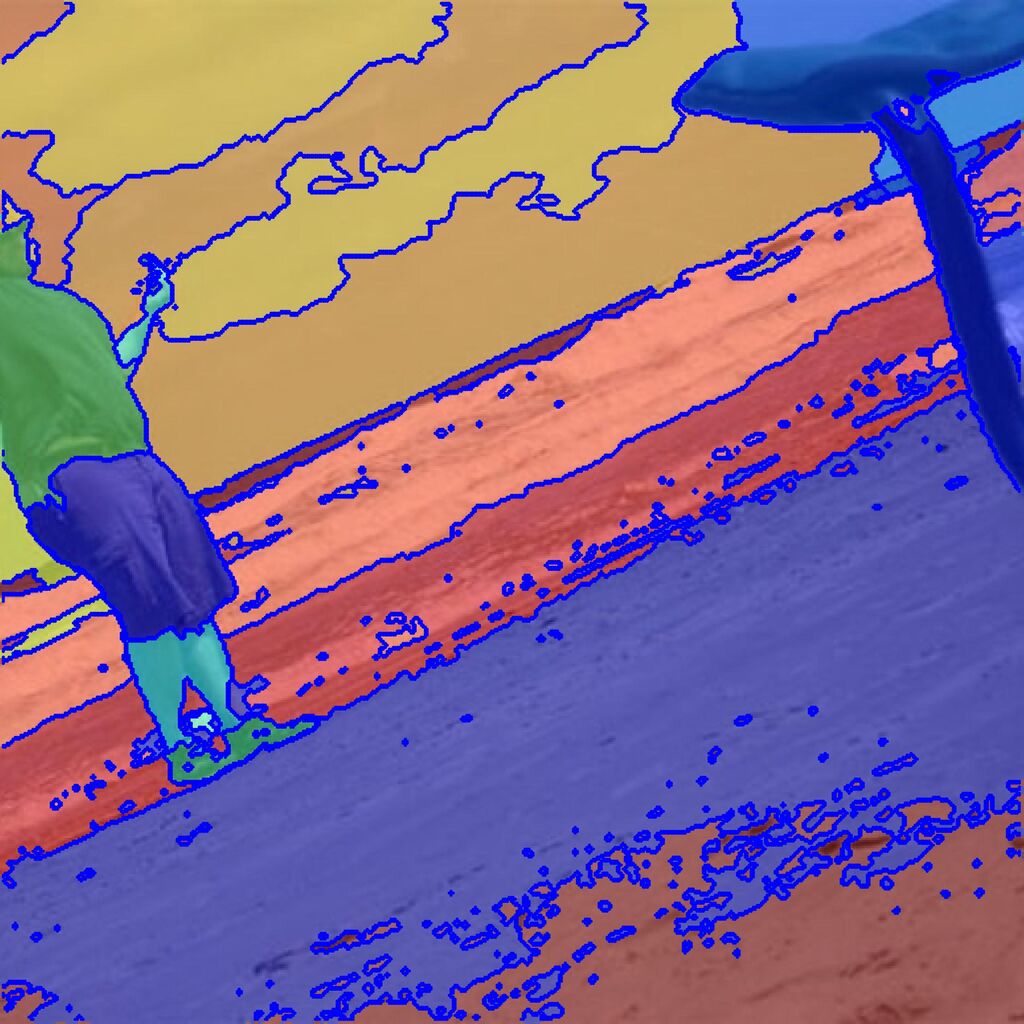}
    \end{subfigure}%
    \begin{subfigure}[t]{0.225\textwidth}
        \centering
        \includegraphics[width=\textwidth]{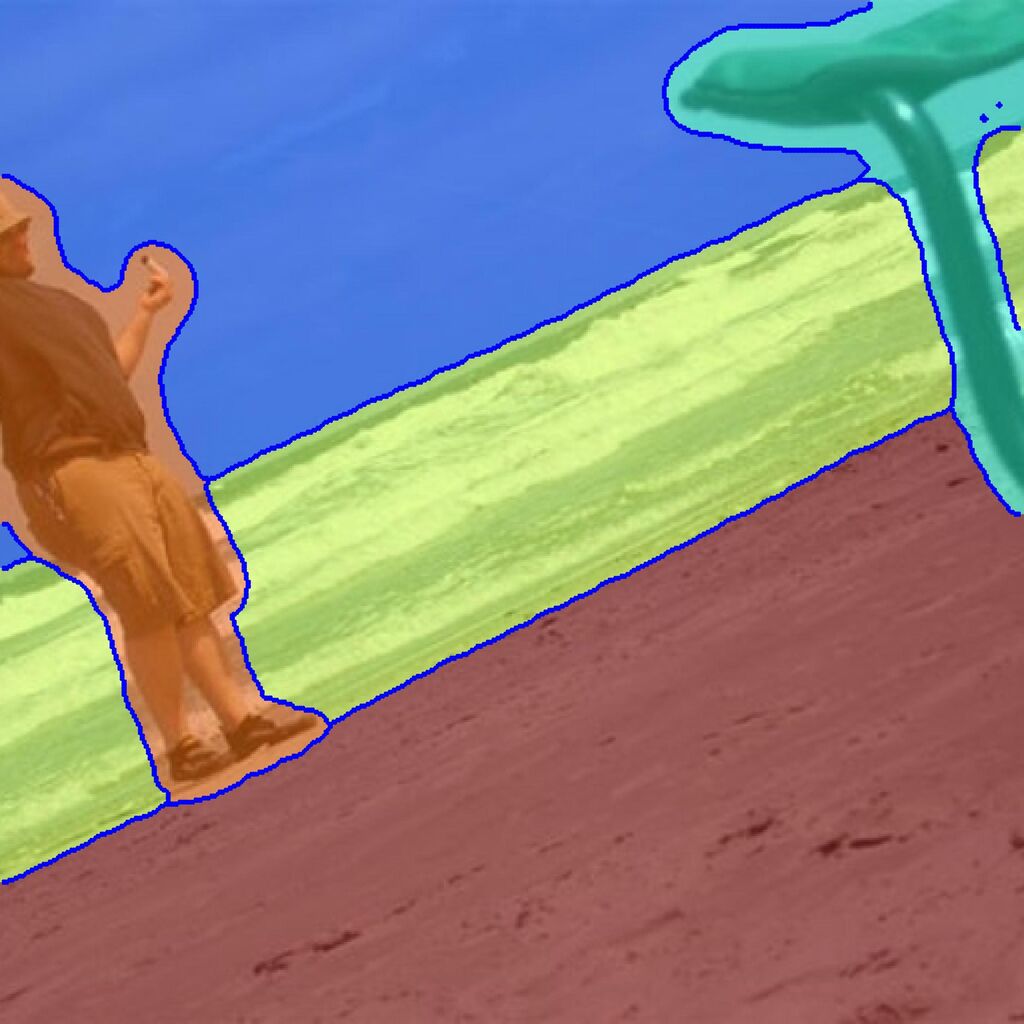}
    \end{subfigure}%
    \begin{subfigure}[t]{0.225\textwidth}
        \centering
        \includegraphics[width=\textwidth]{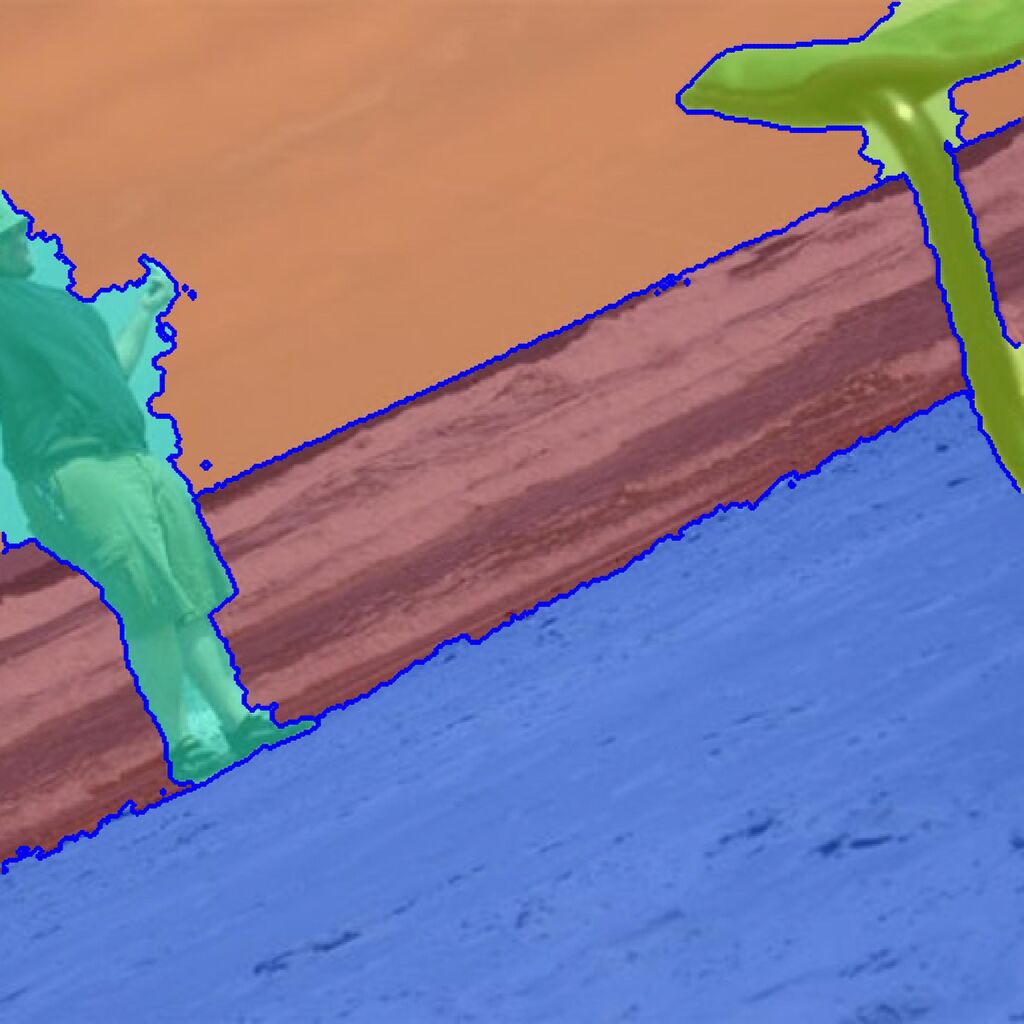}
    \end{subfigure}%

    \vspace{-1em}

    \begin{subfigure}[t]{0.225\textwidth}
        \centering
        \includegraphics[width=\textwidth]{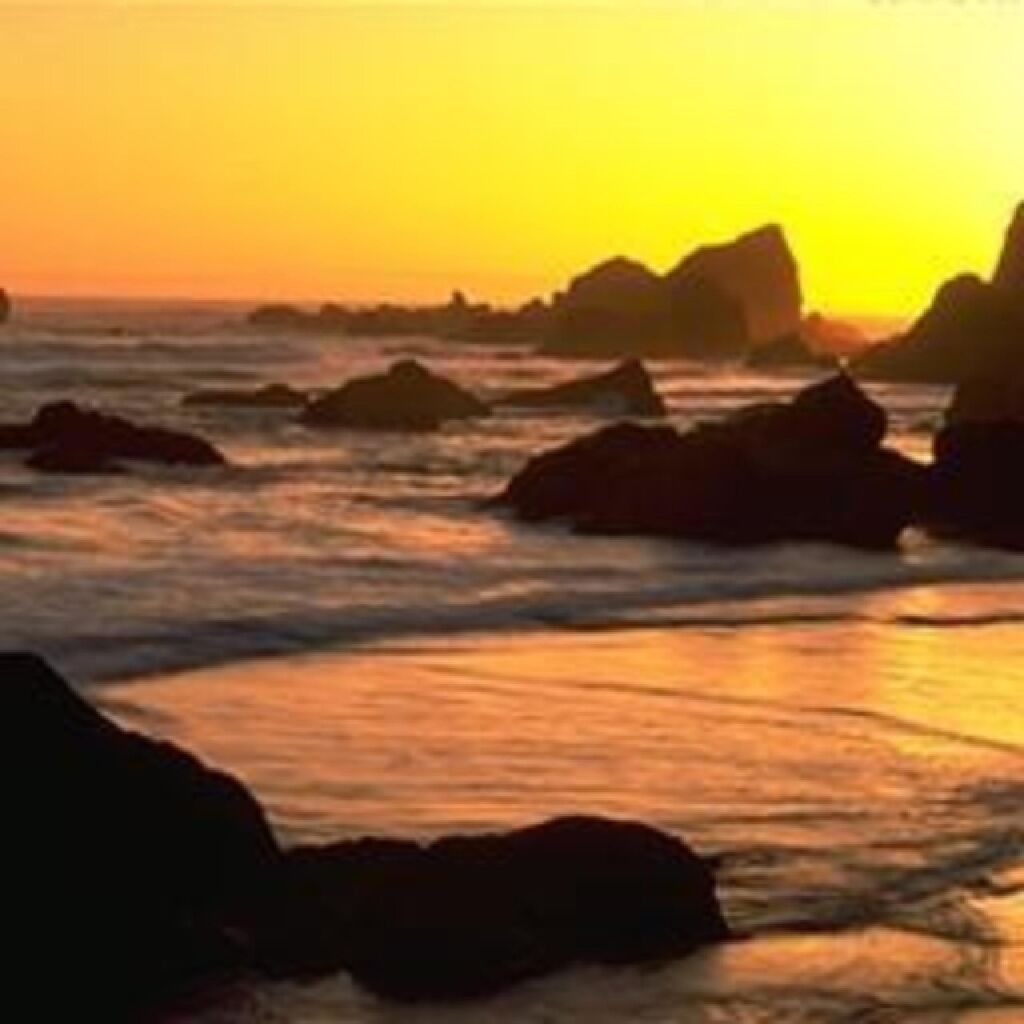}
    \end{subfigure}%
    \begin{subfigure}[t]{0.225\textwidth}
        \centering
        \includegraphics[width=\textwidth]{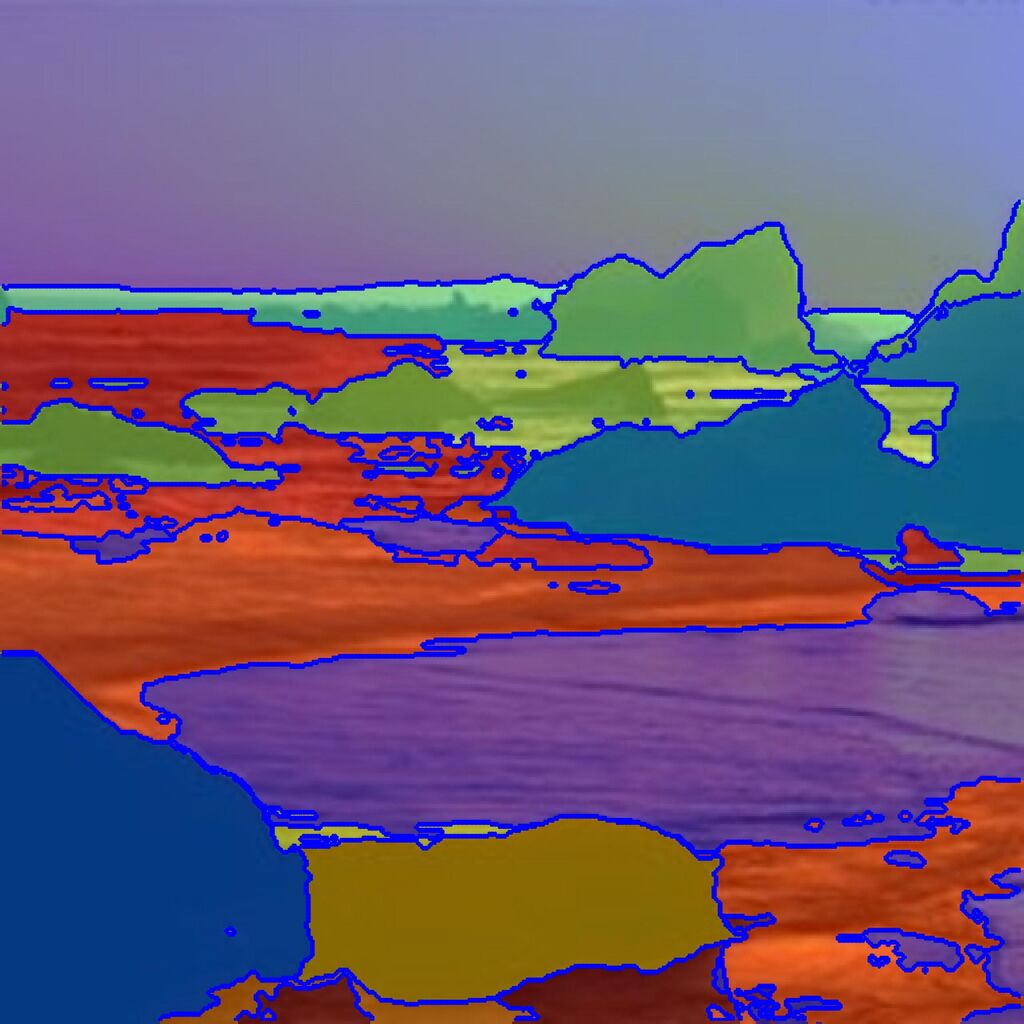}
    \end{subfigure}%
    \begin{subfigure}[t]{0.225\textwidth}
        \centering
        \includegraphics[width=\textwidth]{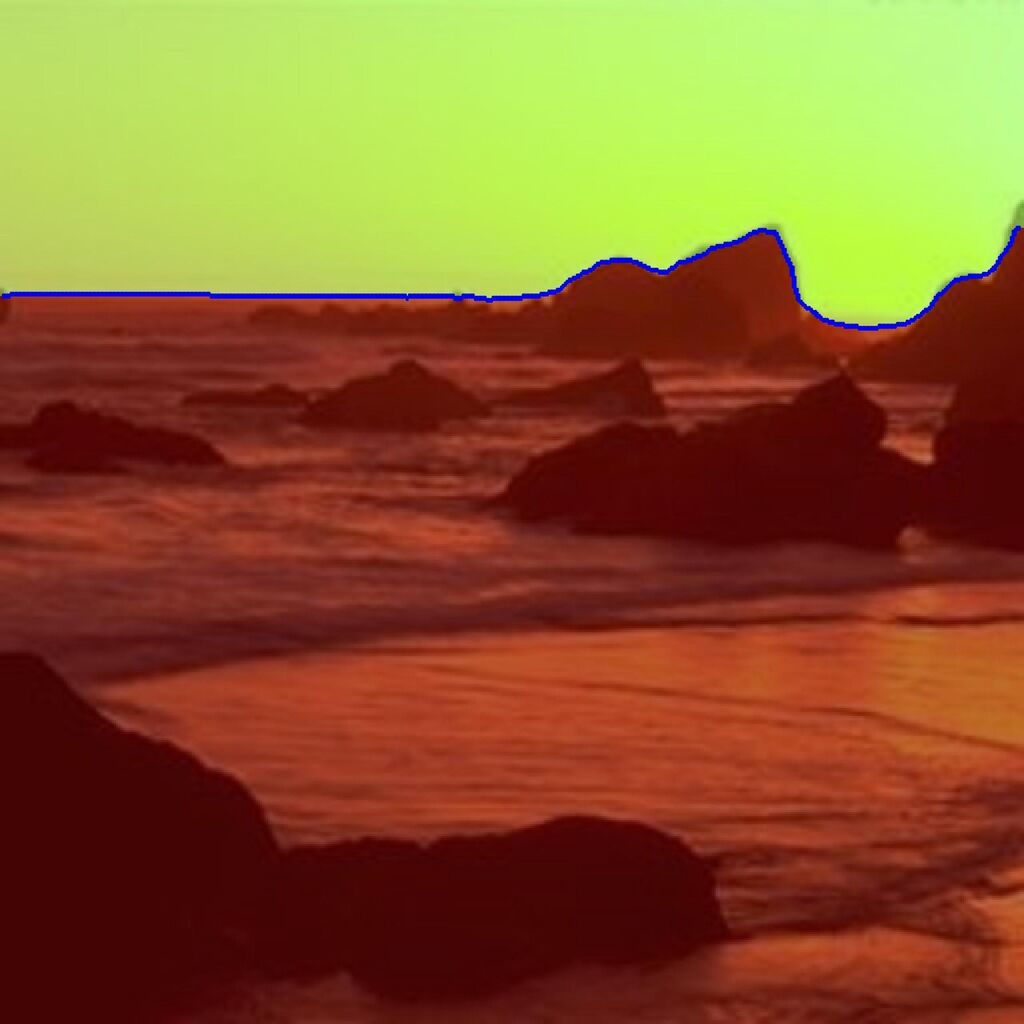}
    \end{subfigure}%
    \begin{subfigure}[t]{0.225\textwidth}
        \centering
        \includegraphics[width=\textwidth]{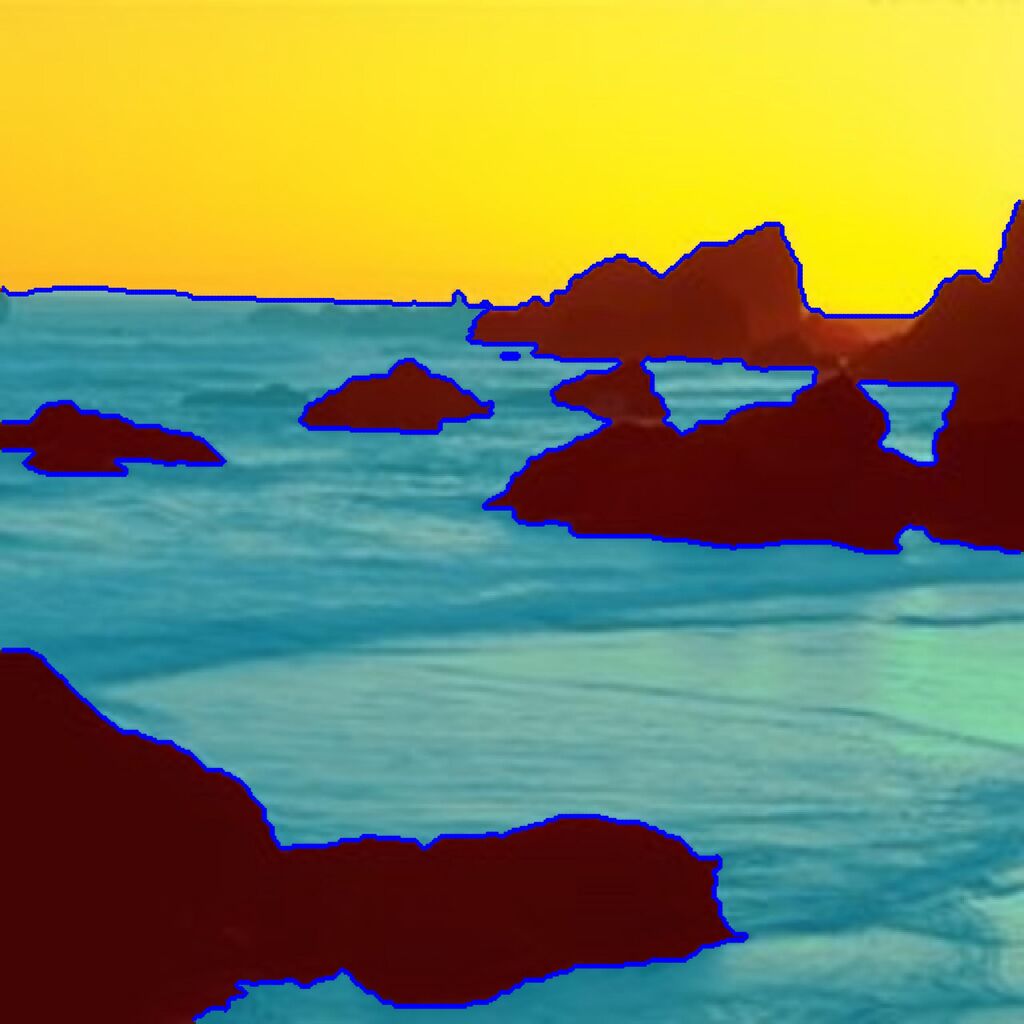}
    \end{subfigure}%

    \caption{Qualitative comparison of the proposed method against DiffSeg and DiffCut.}
    \label{fig:baseline_qualitative_comparison_3}
\end{figure*}

\section{Open-Vocabulary Extension}
\label{sec:supp_open_vocabulary}

We also expanded our model to an open-vocabulary setting, where the identified masks were assigned to an arbitrary set of textual semantic labels. Specifically, after obtaining the segmentation masks using the proposed framework, we applied a frozen CLIP model~\cite{radford2021learning} to the original input image. We then computed the average CLIP feature vector within each object mask identified by our model. \replaced{The pre-trained visual}{A} projection layer \added{of CLIP} was then used to map each averaged CLIP feature to the text space, where they can be compared to the embedding representation of predefined labels in each dataset. For evaluation, the predicted labels were compared to the ground truth labels using the mIoU score. Supp. Tab.~\ref{tab:open_vocabulary_results} compares the performance of our model against DiffCut, showing a better performance across datasets.

\section{Failure Cases}

\begin{figure*}[h!]
    \centering
    \begin{minipage}[c]{0.15\textwidth}
        \centering
        \includegraphics[width=\textwidth]{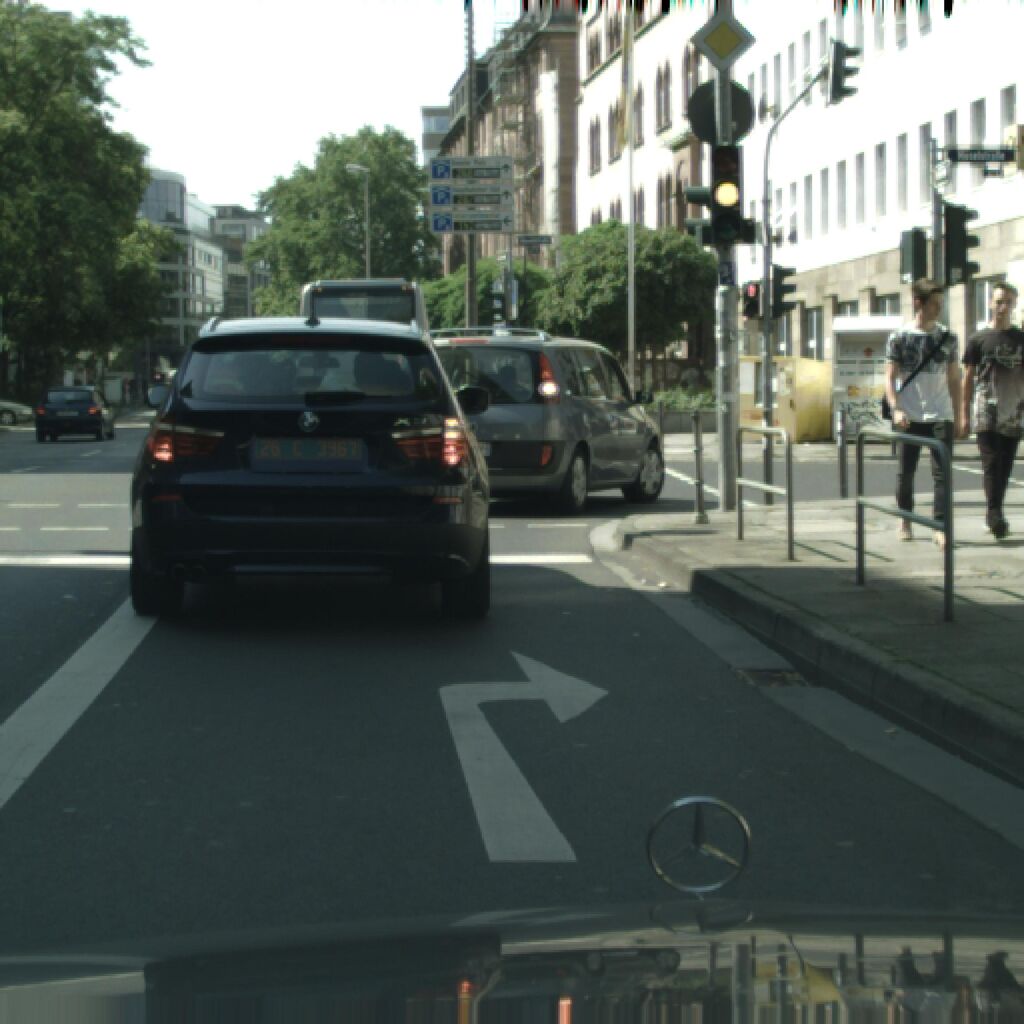}
    \end{minipage}%
    \begin{minipage}[c]{0.15\textwidth}
        \centering
        \includegraphics[width=\textwidth]{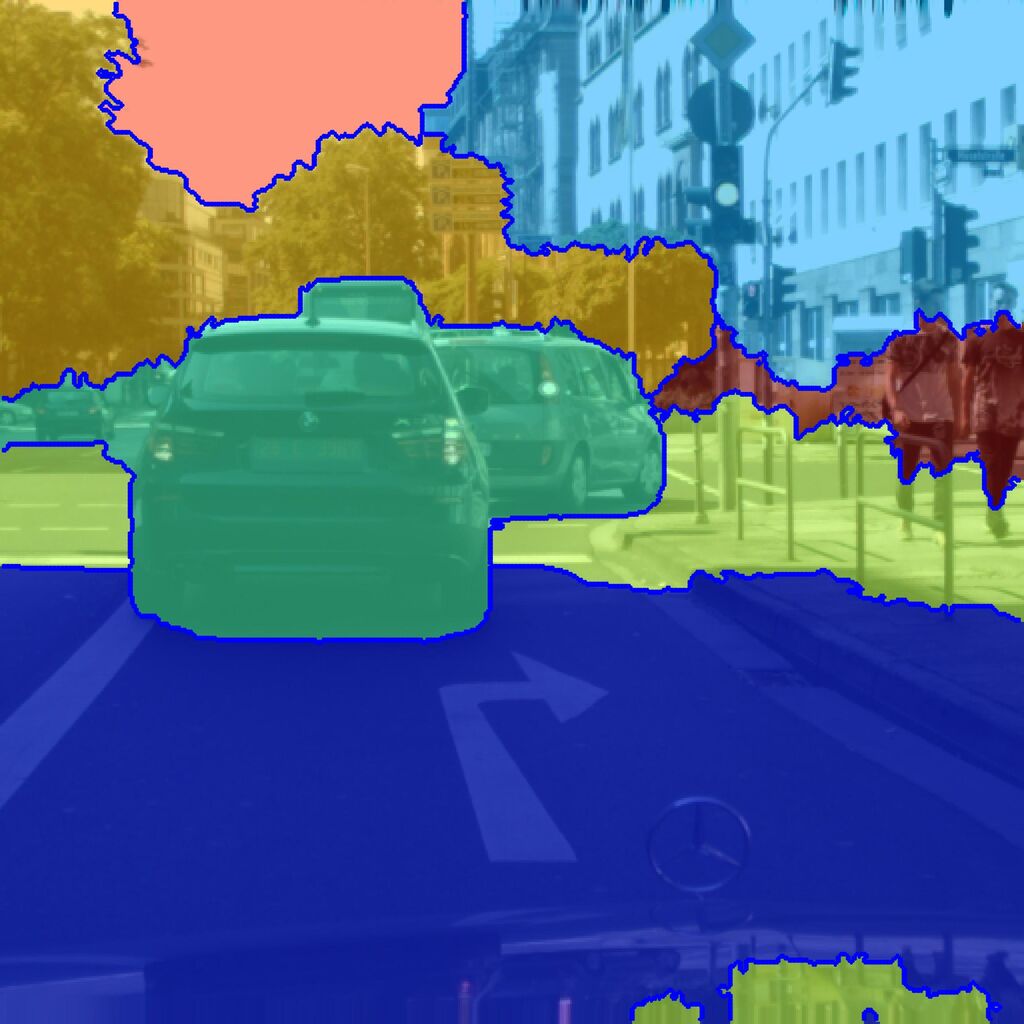}
    \end{minipage}%


    \begin{minipage}[c]{0.15\textwidth}
        \centering
        \includegraphics[width=\textwidth]{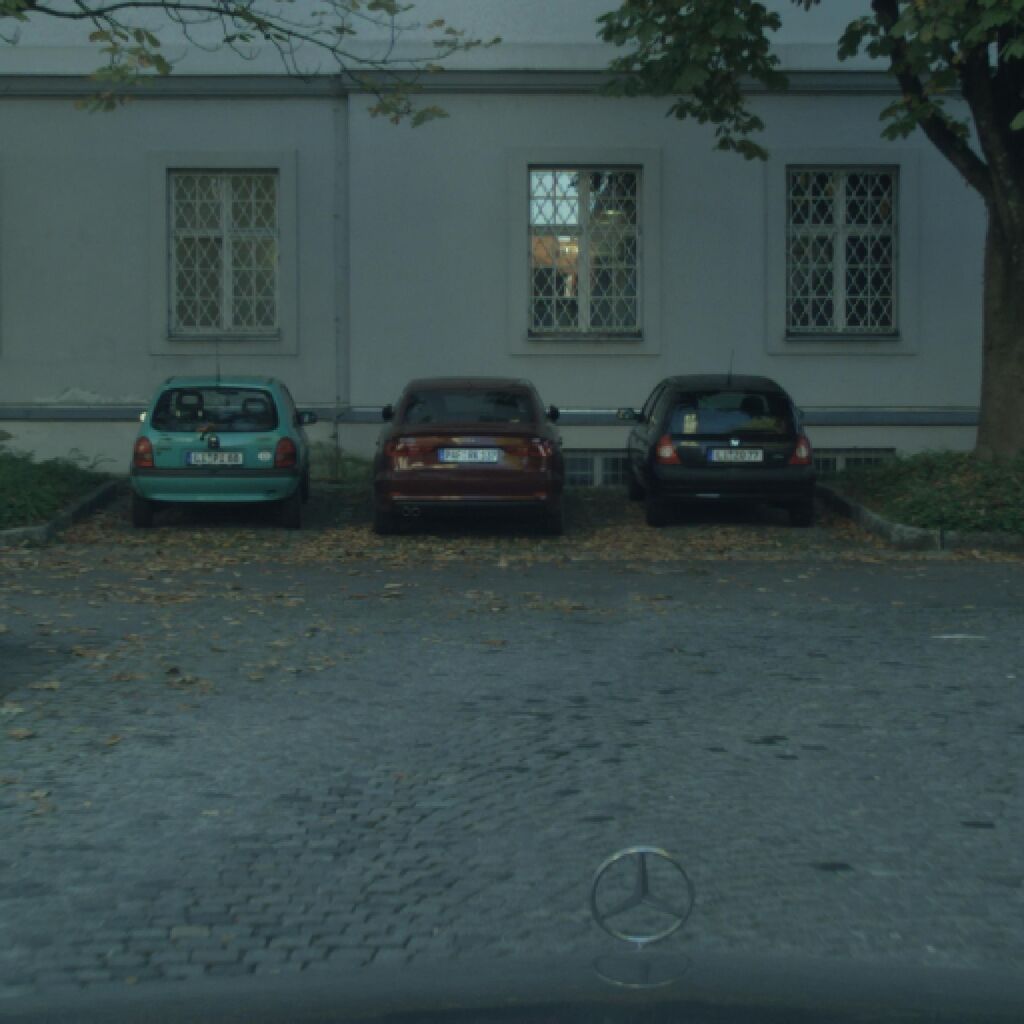}
    \end{minipage}%
    \begin{minipage}[c]{0.15\textwidth}
        \centering
        \includegraphics[width=\textwidth]{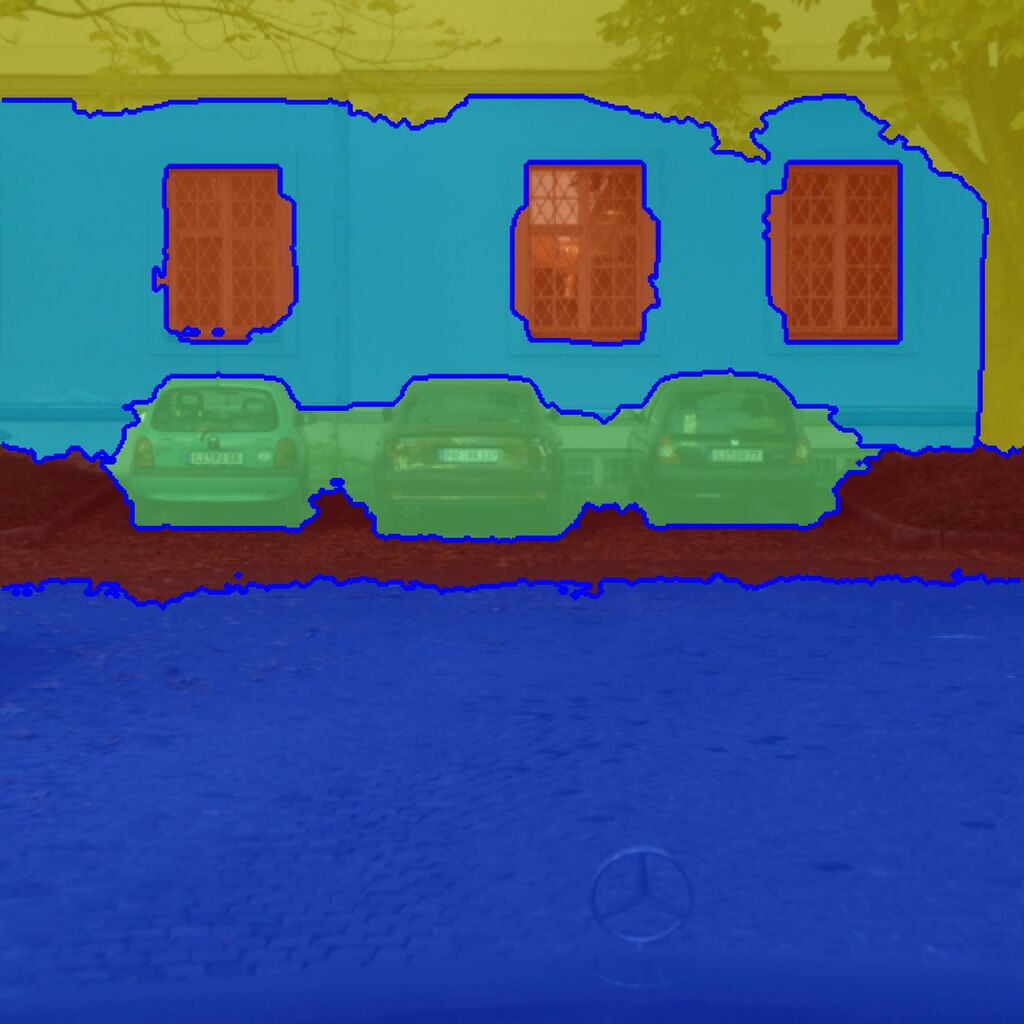}
    \end{minipage}%


    \begin{minipage}[c]{0.15\textwidth}
        \centering
        \includegraphics[width=\textwidth]{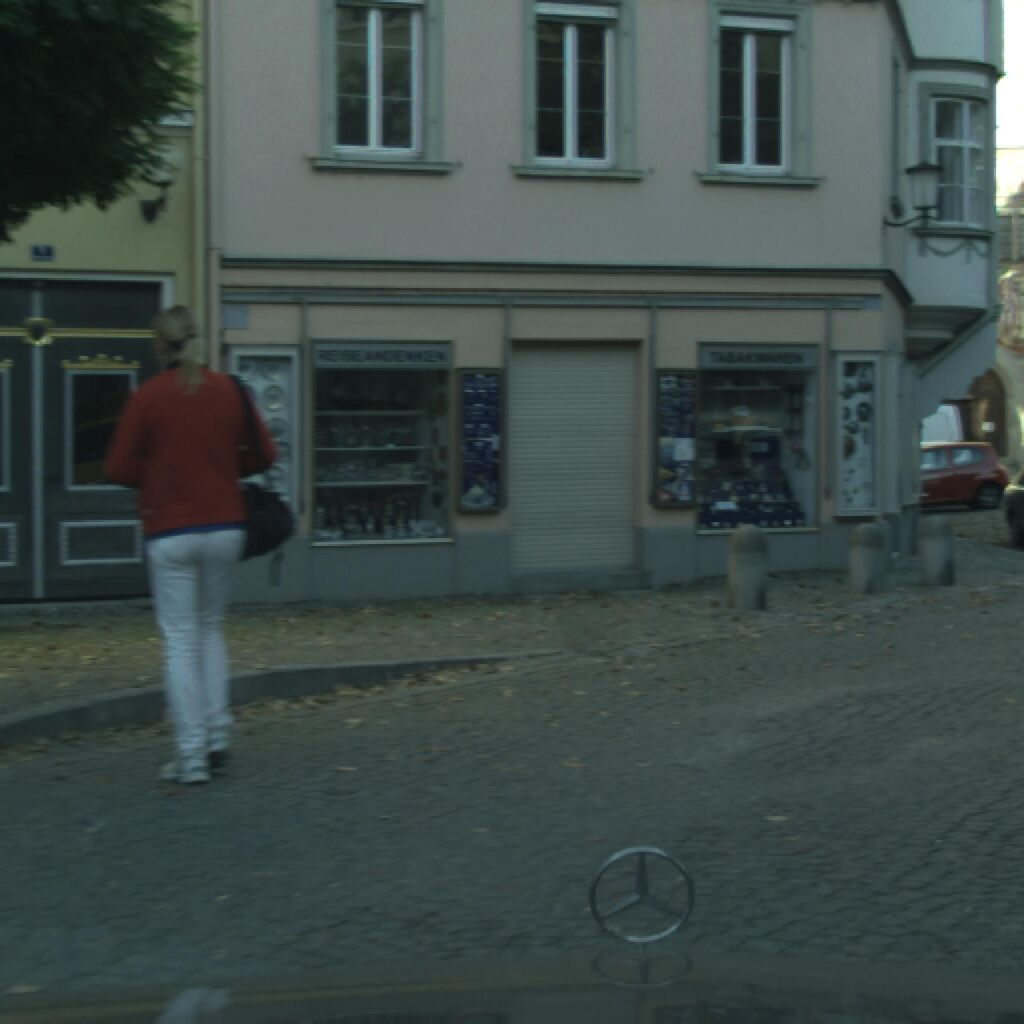}
    \end{minipage}%
    \begin{minipage}[c]{0.15\textwidth}
        \centering
        \includegraphics[width=\textwidth]{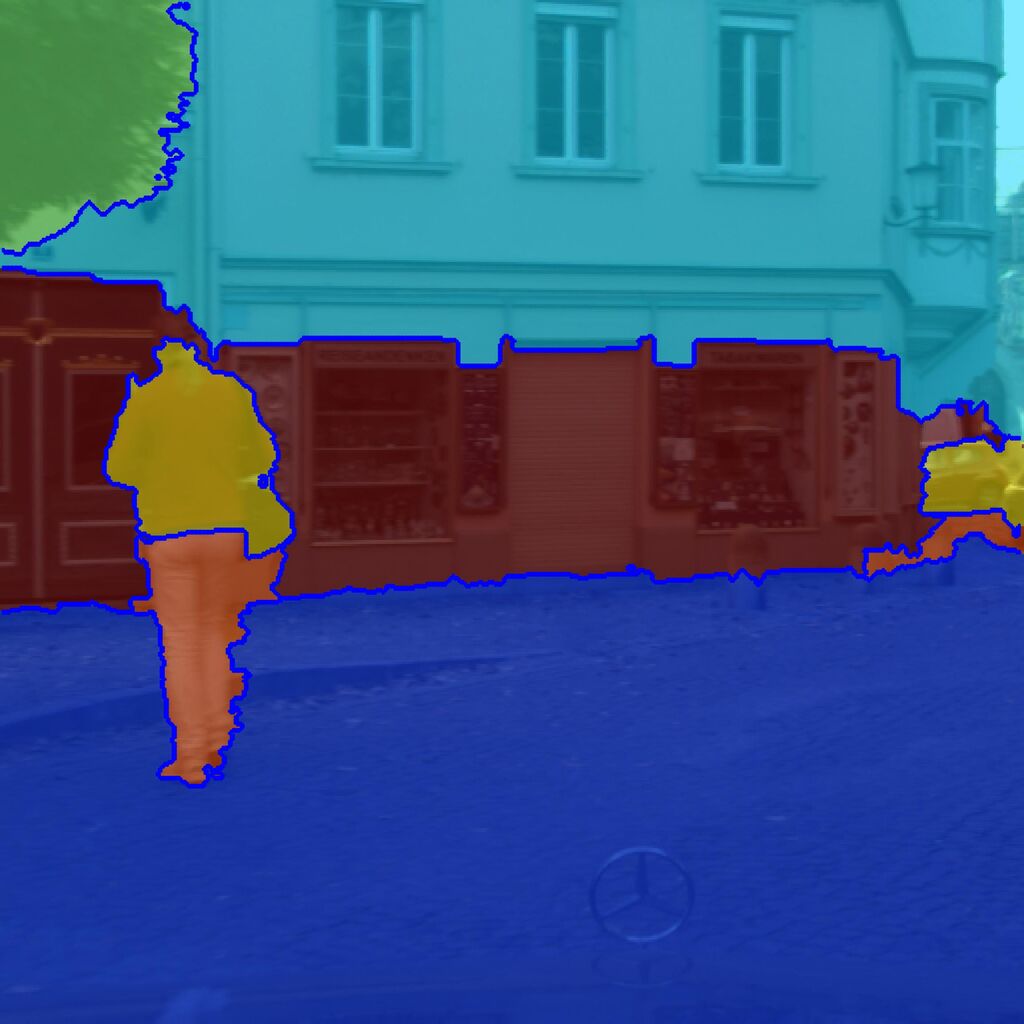}
    \end{minipage}%
    
    \caption{Examples of segmentation failures. Each row shows an input image and its corresponding segmentation mask. These examples highlight instances where our model exhibits failure cases such as using color for segmentation of different sections, particularly in cluttered scenes, shadows, and cluttered objects with multiple coloring schemes.}
    \label{fig:qualitative_limitations}
\end{figure*}

Supp. Fig.~\ref{fig:qualitative_limitations} shows some failure cases of the proposed method. These examples illustrate failure cases where the model uses colors for segmentation, particularly in cluttered scenes, shadows, and cluttered objects where multiple coloring schemes can be found. 


\begin{figure*}[h!]
    \centering
    \includegraphics[width=\linewidth]{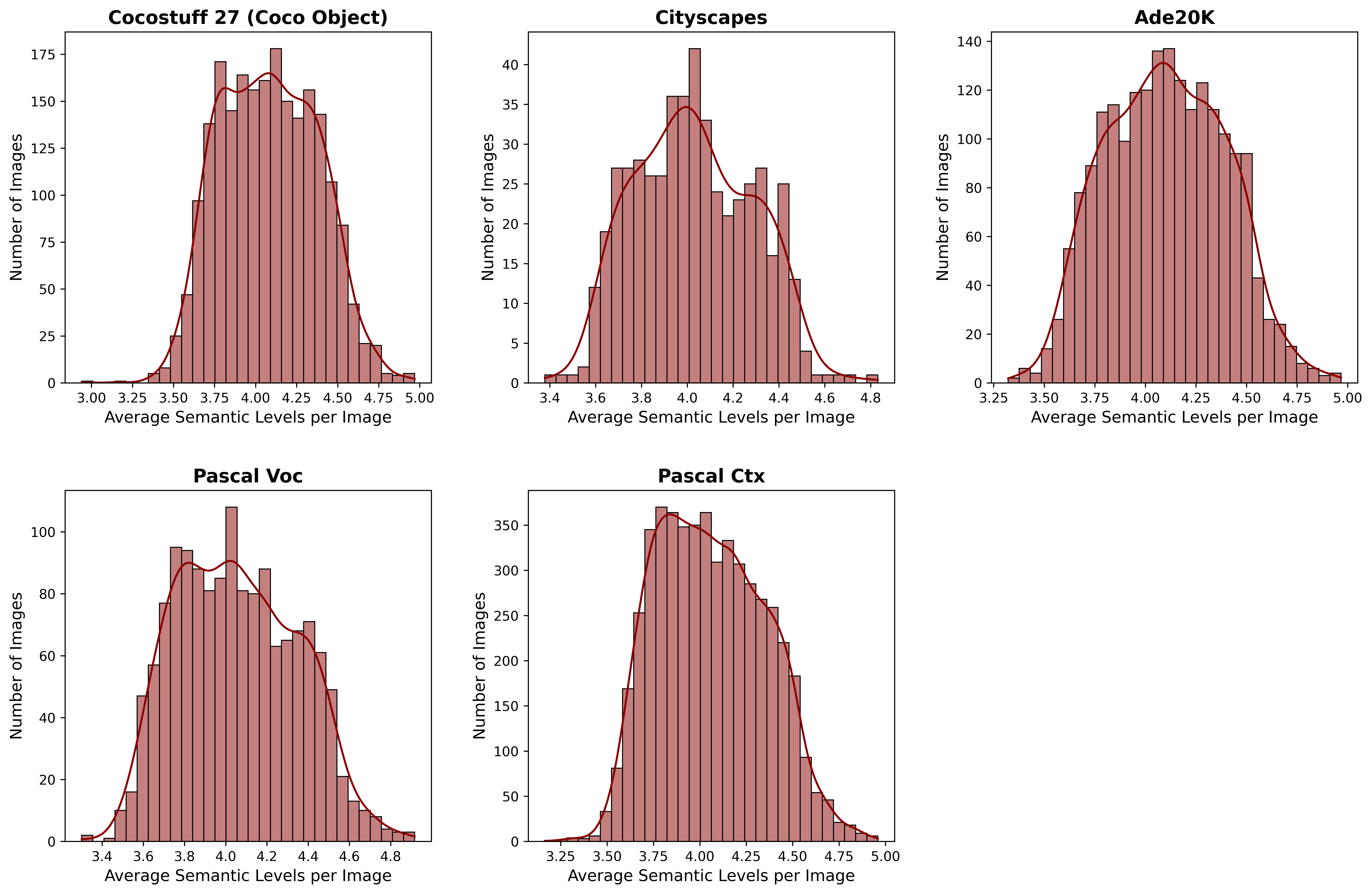}
    \caption{Average number of semantic levels identified by our method in each image across different datasets. Majority of the images fall into four semantics levels with minimum of 3 and a maximum of 5 levels.}
    \label{fig:semantic_level_average_per_dataset}
\end{figure*}

\section{Difference with Self-Attention features}

This section provides a detailed clarification on the architectural novelty of the proposed Contextual Similarity Map and distinguishes it from the canonical self-attention mechanism. While our method leverages the outputs of self-attention blocks in the U-Net architecture, its core operation is a cross-layer feature fusion strategy.


A standard self-attention mechanism, as widely used in Transformer architectures, is an intra-layer operation. Given a set of input features for a single layer, it computes Query (Q), Key (K), and Value (V) matrices. The attention scores are calculated based on the similarity between queries and keys, and these scores are then used to produce a weighted sum of the values.

The key characteristic of this mechanism is that it is self-contained. It refines a feature set by modeling the internal relationships \textit{within that same set of features} at a single layer of the architecture.


Our proposed method operates on a different principle. Instead of refining features within a single layer, we create a new, enriched feature representation by fusing outputs from \textit{multiple, disparate layers} of the U-Net.

Our formulation is expressed as:
\[
V_t = M_t \times F_t .
\]
Here, the components are fundamentally different in their origin and semantic meaning:
\begin{itemize}
    \item \textbf{$F_t$:} These are the semantically rich features extracted from the final, deepest transformer block of the U-Net encoder. While they are the \textit{output} of a self-attention operation, they represent the highest level of contextual abstraction in the network, capturing a holistic understanding of the scene at a low spatial resolution.
    
    \item \textbf{$M_t$:} This is a unified matrix created by aggregating the multi-resolution self-attention maps from the various layers of the U-Net. Unlike the values in a standard attention mechanism, these maps are not a learned projection of the input features. Instead, they serve as a collection of pre-computed, high-resolution spatial and structural relationship scores.
\end{itemize}


The novelty of our approach lies in explicitly fusing different levels of feature abstraction to overcome a critical trade-off.

\begin{enumerate}
    \item \textbf{Cross-Layer Fusion, Not Intra-Layer Refinement:} We are not calculating new attention scores. We are using the attention maps from shallow and deep layers ($M_t$) as a high-resolution guidance mechanism to re-weight and aggregate the high-level semantic features from the network's bottleneck ($F_t$). This is a \textit{cross-layer fusion} that bridges different parts of the U-Net architecture.
    
    \item \textbf{Synergy of Context and Resolution:} The primary motivation for our method is to solve the inherent trade-off between spatial resolution and contextual information. The high-resolution attention maps ($M_t$) provide precise, fine-grained boundary details but lack robust semantic context. Conversely, the deep encoder features ($F_t$) offer rich contextual understanding but have a significantly reduced spatial resolution, leading to coarse segmentation. Our method creates a synergistic feature representation that possesses both of these desired qualities simultaneously.
\end{enumerate}

In summary, the CSM is not a self-attention mechanism itself, but a novel feature fusion and re-weighting framework. It leverages the outputs of multiple self-attention blocks across the network's hierarchy to construct a more comprehensive per-pixel representation, which is suited for high-quality, zero-shot semantic segmentation.

\section{Random Seed Robustness}
\label{sec:supp_random_seed_robustness}

The robustness of our method was analyzed against the random initialization seed used for noise generation during the feature extraction process. Supp. Tab.~\ref{tab:supp_ablation_seeds} shows that the performance of the model remains roughly consistent across different number of seeds.

\begin{table}[h!]
    \centering
    \small
    \caption{Ablation on the number of seeds used during inference on the Cityscapes dataset.}
    \begin{tabular}{lccccc}
        \toprule
        \textbf{Method} & \textbf{1 Seed} & \textbf{2 Seeds} & \textbf{3 Seeds} & \textbf{4 Seeds} & \textbf{5 Seeds}\\
        \midrule
        mIoU & 28.0 & 27.5 & 27.9 & 27.4 & 27.8 \\
        \bottomrule
    \end{tabular}
    \label{tab:supp_ablation_seeds}
\end{table}

\section{PELT Threshold Analysis}

We investigated the effect of varying the threshold $\beta$ within the PELT algorithm. As shown in Supp. Tab.~\ref{tab:abaltion_pelt_threshold_equal_rows}, the performance of our method is largely consistent for values of $\beta$ up to $1.0$. However, performance noticeably declines when $\beta$ exceeds this value.  This decrease is due to the PELT algorithm anticipating larger changes in the Fiedler values, which prevents it from identifying the correct semantic boundaries and instead defaults to using the final timestep as the target for subsequent analysis.

\begin{table}[h!]
  \centering
  \small
  \caption{Performance across different PELT thresholds on the CityScapes dataset.}
  \label{tab:abaltion_pelt_threshold_equal_rows}
      \begin{tabular}{lrrrrrrr}
        \toprule
        \textbf{$\mathbf{\beta}$} & 0.1 & 0.2 & 0.3 & 0.4 & 0.5 & 0.6 & 0.7 \\
        \textbf{mIoU} & 27.5 & 27.6 & \textbf{27.8} & 27.5 & 27.2 & 27.4 & 27.2 \\
        \midrule
        \textbf{$\mathbf{\beta}$} & 0.8 & 0.9 & 1.0 & 5.0 & 10.0 & 20.0 & 50.0 \\
        \textbf{mIoU} & 26.9 & 27.1 & 26.8 & 21.8 & 11.8 & 11.8 & 11.8 \\
        \bottomrule
      \end{tabular}
\end{table}

\section{Attention Resolution Analysis}

We investigated the effect of varying attention resolutions on our methodology by exclusively utilizing self-attention maps from a single, specific spatial resolution. As detailed in Supp. Tab.~\ref{tab:supp_attention_resolution}, an inverse relationship can be seen: lower spatial resolutions for the attention maps typically result in a decrease in performance. This finding underscores the inadequacy of solely depending on U-Net features and highlights the crucial role of attention map resolution in achieving superior semantic comprehension from a pre-trained Stable Diffusion.

\begin{table}[htbp]
\centering
\caption{Ablation of self-attention resolutions on the CityScapes dataset.}
\label{tab:supp_attention_resolution} 
\begin{tabular}{lccccc}
\toprule
\textbf{Resolution} & $All$ & $8 \times 8$ & $16 \times 16$ & $32 \times 32$ & $64 \times 64$ \\
\midrule
\textbf{mIoU} & \textbf{27.8} & 14.7 & 26.48 & 27.59 & 26.92 \\
\bottomrule
\end{tabular}%
\end{table}

\section{Additional Analysis}


Supp. \cref{fig:supp_hierarchical_progress_sample3,fig:supp_hierarchical_progress_sample5,fig:supp_hierarchical_progress_sample1,fig:supp_hierarchical_progress_sample14,fig:supp_hierarchical_progress_sample7} show some input images with their corresponding Temporal Stability Matrices (TSMs) and some of their corresponding Contextual Similarity Maps (CSMs) at different denoising timesteps.

Supp. Fig.~\ref{fig:supp_hierarchical_progress_sample3} highlights the dynamic nature of the semantics during the denoising process as the optimal timestep for the red point is around $t=500$ while for the blue point is around $t=40$.

The TSMs in Supp. Fig.~\ref{fig:supp_hierarchical_progress_sample5} illustrate distinct hierarchical abstraction levels within the image.
For instance, the CSM of the blue point at time $t=20$ emphasizes the fridge door. Subsequently, at time $t=250$, the focus transitions to encompass the entire fridge. Finally, at higher timesteps, the scope broadens to include the entire room.
It is noticeable that both TSMs (for the red and the blue point) show 5 different semantics levels.

Supp. Fig.~\ref{fig:supp_hierarchical_progress_sample1} shows that early CSMs are primarily concentrated on the frontal wall of the building. As the denoising progresses, the CSMs expand their focus to the whole building (e.g. $t=650$), and eventually encompass most of the foreground objects.
 Similarly, Supp. Fig.~\ref{fig:supp_hierarchical_progress_sample14} shows that CSMs from early timesteps ($t=20$) mainly focus on the dog's ears. As the denoising progresses to larger timesteps ($t = 350$), the CSMs expand to the whole body of the dog, and in later stages, they cover the foreground objects in the image. These observations demonstrate the hierarchical progression of semantic information throughout the denoising process.

TSMs in Supp. Fig.~\ref{fig:supp_hierarchical_progress_sample7} show patterns with different number of semantic levels. For example, the TSM associated with the red query point has three semantic levels, whereas the blue query point's TSM displays four levels.







\section{Implementation Details}
\label{sec:supp_implementation_details}

Following ~\cite{tian2024diffuse}, we used SDv1.4~\cite{rombach2022latent} as the pre-trained Stable Diffusion model. To extract the features and the self-attention maps at a specific timestep $t$, noise was added to the input image, as follows:
\begin{equation}
    \label{eq:noise_schedule}
    x_t = \sqrt{\bar{\alpha}_t} x_0 + \sqrt{1 - \bar{\alpha}_t}\epsilon .
\end{equation}


The noisy image was then passed to the Stable Diffusion U-Net to extract the self-attention maps and the encoder features. The attention maps consisted of 5 maps with $64 \times 64$ resolution, 5 at $32 \times 32$, 5 at $16 \times 16$, and 1 at $8 \times 8$, with each containing 8 heads. For each map, we averaged the heads, upscaled the result to $64 \times 64$, and then averaged them in the spatial dimension based on the normalized weights given by the square root of the spatial size, as in \cite{tian2024diffuse}. Features were extracted from the last self-attention block of the U-Net encoder, resulting in a $16 \times 16 \times 1280$ embedding. Finally, these features and self-attention maps were fused to construct the CSM for each timestep $t$ and each pixel, as described in Sec.~\ref{sec:contextual_similarity_map}. \added{For DiT architectures, we used the middle transformer layer for feature extraction (e.g., layer 12 out of a 24-layer DiT) while the self-attention from all layers was used.}


After selecting the target timestep, as instructed in Sec.~\ref{sec:semantic_boundary_detection}, the CSMs went through the two-phase aggregation in Eq.~\ref{eq:hierarchical_aggregation} with $\tau_{s}$ set to $0.05$. The resulting feature vector, $Q$, was upscaled to the original image resolution 
The NCut algorithm partitioned the input feature vector into multiple segments through an iterative process with a maximum of $10$ iterations. This process minimized the normalized cut objective, which simultaneously ensured high similarity within the resulting segments and maximal disparity between them. The algorithm was executed using the parameters $\alpha = 1.0$ and $\tau = 0.5$. The yielded labels were subsequently refined using a post-processing method, PAMR, as detailed in \cite{araslanov2020single}. 


The denoising process was analyzed across high \replaced{(550,600,650,700,750,800,850,900)}{(550-900)}, medium \replaced{(100, 150, 200, 250, 300, 350, 400, 450, 500)}{(100-500)}, and low timesteps \replaced{(10, 20, 30, 40, 50)}{(10-50)}. Input images were cropped and resized to $512\times 512$ resolution. For the PELT algorithm, we used a non-linear RBF (Radial Basis Function) cost function with penalty parameter $\beta$ empirically set to $0.3$. Attention maps and encoder features were aggregated across 5 random seeds to enhance the robustness of the CSMs. All experiments were conducted on a single NVIDIA A100 GPU.

\section{Object vs. Part Soft-IoU Evaluation Details}
\label{sec:supp_object_vs_part}

To perform our hierarchical semantic evaluation, we select images from PASCAL-Part dataset \cite{chen2014detect} that contain at least one object annotated with three or more distinct semantic parts. We evaluate a diverse set of diffusion backbones, including SDv1.4, SDv2.1, SDXL, SSD-1B, PixArt-$\alpha$, and Lumina-Next-SFT, standardizing the input resolution to $512 \times 512$ across all architectures. For each image, we extract self-attention maps across 14 distinct timesteps spanning the full denoising trajectory. To isolate the conceptual evolution, ground-truth object/part masks are resized using nearest-neighbor interpolation to match the spatial resolution of the extracted attention maps.

For each pixel in the object, we calculate the Point-Affinity Soft-IoU between the attention maps and the object/part ground-truth annotation mask. Let $A_i \in \mathbb{R}^{H \times W}$ be the attention map for pixel $i$ \added{(normalized with min-max)} and $M_i \in \mathbb{R}^{H \times W}$ be the binary mask from ground-truth object or part annotation for the same pixel, where values of $1$ show regions that have the same semantic labels and $0$ shows otherwise. We then compute the Soft-Intersection over Union (Soft-IoU) against a ground-truth binary mask $M_i$ as: $\text{Soft-IoU}(A_i, M_i) = \frac{\sum (A_i \odot M_i)}{\sum (A_i + M_i - A_i \odot M_i)}$, where $\odot$ denotes element-wise multiplication. For each pixel, this metric is computed twice: once against its localized part-level mask and once against its parent object-level mask. The final reported Part-IoU and Object-IoU scores are aggregated by averaging across pixels in the object.

\end{document}